\documentclass{these-dbl}

\hypersetup{
   pdfauthor   = {Clément BONET},
}

\usepackage{minitoc}
\mtcsetfeature{minitoc}{after}{\vspace{15pt}}
\mtcsetfeature{minitoc}{before}{\vspace{-35pt}}

\usepackage{bookmark}

\usepackage{amsmath,amsfonts,amssymb,mathtools}
\usepackage{bbold}
\usepackage{bm}

\usepackage{amsthm}
\newtheorem{definition}{Definition} 
\newtheorem{theorem}{Theorem}
\newtheorem{proposition}{Proposition}

\newtheorem{remark}{Remark}
\newtheorem{lemma}{Lemma}

\usepackage{graphicx}
\usepackage{float}
\usepackage{subfig}
\usepackage{wrapfig}

\usepackage{booktabs} 
\usepackage{multirow}
\newcommand{\rotcell}[1]{%
  \rotatebox[origin=c]{90}{\begin{tabular}{c}#1\end{tabular}}%
}
\usepackage{algorithm}
\usepackage{algorithmic}

\usepackage{cleveref}
\usepackage{appendix}

\usepackage{macros}

\DeclareMathOperator*{\argmin}{argmin}
\DeclareMathOperator*{\argmax}{argmax}
\DeclareMathOperator*{\arccosh}{arccosh}
\DeclareMathOperator*{\arctanh}{arctanh}

\makeatletter
\newcommand{\proofpart}[2]{%
  \par
  \addvspace{\medskipamount}%
  \noindent\emph{Part #1: #2}\par\nobreak
  \addvspace{\smallskipamount}%
  \@afterheading
}
\makeatother

\setcitestyle{authoryear,round,citesep={;},aysep={,},yysep={;}}

\begin{document}

\selectlanguage{french}


\makeatletter
\newcommand\addcase[3]{\expandafter\def\csname\string#1@case@#2\endcsname{#3}}
\newcommand\makeswitch[2][]{%
  \newcommand#2[1]{%
    \ifcsname\string#2@case@##1\endcsname\csname\string#2@case@##1\endcsname\else#1\fi%
  }%
}
\makeatother

\newcommand\hauteurlogos[3]{
    \hauteurlogoecole{#1}
    \hauteurlogoetablissementA{#2}
    \hauteurlogoetablissementB{#3}
}


\newcommand\addecoledoctorale[5]{\direcole{#1}\numeroecole{#2}\definecolor{color-ecole}{RGB}{#3}\nomecoleA{#4}\nomecoleB{#5}}

\makeswitch[default]\ecoledoctorale{}

\addcase\ecoledoctorale{MathSTIC}{\addecoledoctorale
    {MathSTIC}
    {644}
    {159,182,217} 
    {Math\'{e}matiques et Sciences et Technologies}
    {de l'Information et de la Communication en Bretagne Océane}
}


\newcommand\addetablissement[4]{\logoetablissementB{#1}\nometablissementC{#2}\nometablissementD{#3}\nometablissementE{#4}}

\makeswitch[default]\etablissement{}

\addcase\etablissement{ENIB}{\addetablissement
    {ENIB}
    {}
    {L'\'{E}COLE NATIONALE}
    {D'ING\'{E}NIEURS DE BREST}
}

\addcase\etablissement{UBO}{\addetablissement
    {UBO}
    {}
    {}
    {L'UNIVERSIT\'{E} DE BRETAGNE OCCIDENTALE}
}
\addcase\etablissement{UBS}{\addetablissement
    {UBS}
    {}
    {}
    {L'UNIVERSIT\'{E} DE BRETAGNE SUD}
}

\ecoledoctorale{MathSTIC}

\etablissement{UBS}

\spec{Mathématiques et leurs interactions}

\author{Clément BONET}

\title{Leveraging Optimal Transport via Projections on Subspaces for Machine Learning Applications}
\lesoustitre{}

\date{16 Novembre 2023}
\lieu{Vannes}

\uniterecherche{Laboratoire de Mathématiques de Bretagne Atlantique}

\numthese{676} 

\jury{
{\normalTwelve \textbf{Rapporteurs avant soutenance :}}\\ \newline
\footnotesizeTwelve
\begin{tabular}{@{}ll}
Gabriel Peyré & Directeur de recherche CNRS, DMA, École Normale Supérieure \\ 
Gabriele Steidl & Professor at Technische Universität Berlin \\
\end{tabular}

\vspace{\baselineskip}
{\normalTwelve \textbf{Composition du Jury :}}\\
\footnotesizeTwelve
\begin{tabular}{@{}lll}

Pr\'{e}sidente :        & Julie Delon & Professeure des universités, Université Paris Cité \\
Examinateurs :         
                       & Frank Nielsen & Sony Computer Science Laboratories Inc \\
                       & David Alvarez-Melis & Assistant Professor, Harvard SEAS \\
Dir. de th\`{e}se :    & François Septier & Professeur des universités, Université Bretagne Sud \\
Co-dir. de th\`{e}se : & Nicolas Courty & Professeur des universités, Université Bretagne Sud  \\
 & Lucas Drumetz & Maître de conférence, IMT Atlantique
\end{tabular}

\vspace{\baselineskip}
{\normalTwelve \textbf{Invit\'{e}(s) :}}\\ \newline
\footnotesizeTwelve
\begin{tabular}{@{}ll}
Rémi Flamary & Professeur des universités, Ecole Polytechnique \\
\end{tabular}
}

\maketitle

\selectlanguage{english}

\clearemptydoublepage
\cleartooddpage[\thispagestyle{empty}]
\chapter*{Acknowledgement}

Je tiens tout d’abord à remercier mes directeurs de thèse, François Septier, Nicolas Courty et Lucas Drumetz pour m’avoir permis de faire cette thèse sur un sujet si intéressant, pour l’accompagnement durant ces trois années et notamment les nombreuses idées et pistes de recherches que vous m’avez proposées, mais aussi pour m’avoir laissé la liberté de développer mes idées et creuser dans différentes directions. François, je tenais à te remercier pour l’accueil au LMBA et pour toujours avoir été présent. Lucas, je te remercie pour nos nombreux échanges. Merci également à Nicolas pour m’avoir fait découvrir le transport optimal, pour avoir toujours été disponible pour discuter recherche, pour tes très nombreuses idées et ton enthousiasme, et pour m’avoir poussé à regarder des sujets que je n’aurais pas forcément abordé de prime abord.

Then, I would like to thank Gabriel Peyré and Gabriele Steidl for their very nice report and for the time taken to read the manuscript. I would also like to thank Julie Delon, David Alvarez-Melis, Frank Nielsen and Rémi Flamary for being members of the jury. It was a high honor to defend in front of researchers whom I admire for their works. 

Ensuite, je souhaiterais remercier les deux laboratoires d’accueil. D’une part, je remercie tous les membres du LMBA pour l’accueil et la bonne humeur quotidienne. En particulier, merci aux habitués du RU. D'autre part, je tiens aussi à remercier l’ensemble de l’équipe Obélix que j’ai eu la chance de croiser durant ces 3 années, notamment durant les nombreux séminaires très sympathiques.

Je souhaite aussi remercier les doctorants des deux laboratoires que j’ai cotoyés durant ma thèse, que ce soit les plus anciens du LMBA  et ceux arrivés après moi, mais aussi les doctorants de l’équipe Obélix. D’ailleurs, Huy et Guillaume, c’est toujours un plaisir de discuter transport optimal avec vous. Merci aussi à Nassab, Corentin et Thi Trang pour votre aide pour organiser le pot de thèse. Enfin, je veux remercier en particulier l’équipe originelle du bureau F196 : Arthur et Yann. Sans vous, je n’aurais certainement pas autant apprécié mon séjour à Vannes. Je garderai un très bon souvenir de toutes les soirées passées ensemble. Arthur, nos discussions sur le sport, la politique, et les autres sujets me manqueront, ainsi que notre super double de badminton.

Je veux  aussi remercier tous ceux avec qui j’ai eu la chance de collaborer durant ma thèse : Titouan Vayer, Paul Berg, Minh-Tan Pham, Laetitia Chapel, Alain Rakotomamonjy, Guillaume Mahey, Gilles Gasso, Elsa Cazelles, Thibault Séjourné, Kilian Fatras et Kimia Nadjahi. Au plaisir de recollaborer dans le futur. Je remercie aussi Eric Lemoine pour avoir accepté de relire mon manuscrit, pour ses suggestions et pour la correction des (nombreuses) erreurs d’anglais dans la version initiale.

Je voudrais aussi remercier les différents professeurs que j’ai eu la chance d’avoir durant mes études, que ce soit au lycée, en classes préparatoires ou à Télécom Paris et au MVA, et qui m’ont poussé et donné envie de poursuivre en thèse. Je vous dois beaucoup. Merci aussi à Pierre Latouche et Raphaël Lachièze-Rey pour m’avoir permis de faire un super stage en M2 puis pour m’avoir encouragé à poursuivre en thèse.

Je remercie ensuite le groupe de Télécom avec qui  je prends toujours du plaisir à échanger quand je retourne à Paris. En particulier, merci à Benoît Malézieux pour nos échanges réguliers sur la thèse, nos discussions scientifiques et nos discussions sur d’autres sujets. Je suis très heureux que nous ayons pu en plus collaborer. Cela a été un plaisir et cela a mené à un super article ! J’espère que nous aurons encore l’occasion de travailler ensemble.

Pour finir, je remercie toute ma famille pour le soutien au quotidien.

\dominitoc

\frontmatter
\clearemptydoublepage
\cleartooddpage[\thispagestyle{empty}]
\renewcommand{\contentsname}{Table of Contents}

\numberwithin{definition}{chapter}
\numberwithin{proposition}{chapter}
\numberwithin{lemma}{chapter}
\numberwithin{theorem}{chapter}
\numberwithin{algorithm}{chapter}


{
  \pdfbookmark{\contentsname}{toc}
  \hypersetup{linkcolor=black}
  \tableofcontents
}

\clearemptydoublepage
\cleartooddpage[\thispagestyle{empty}]
\section*{Notations}

\begin{tabular}{p{3cm}p{11cm}p{1cm}}
\textbf{Abbreviations} & \\
    ML & Machine Learning \\
    OT & Optimal Transport \\
    UOT & Unbalanced Optimal Transport \\
    SW & Sliced-Wasserstein \\
    KL & Kullback-Leibler divergence \\
    MMD & Maximum Mean Discrepancy \\
    AI & Affine-Invariant \\
    LE & Log-Euclidean \\
    NF & Normalizing Flow \\
    AE & Autoencoder \\
    JKO & Jordan-Kinderlehrer-Otto \\
    ICNN & Input Convex Neural Network \\
    WGF & Wasserstein Gradient Flow \\
    SWGF & Sliced-Wassertein Gradient Flow \\
    PDE & Partial Differential Equation \\
    SDE & Stochastic Differential Equation \\
    ULA & Unadjusted Langevin Algorithm \\
    FID & Fréchet Inception Distance \\
    PCA & Principal Component Analysis \\
\end{tabular}

\medskip

\begin{tabular}{p{3cm}p{11cm}p{1cm}}
\textbf{Linear algebra} & \\
    $\langle\cdot, \cdot\rangle$ & Euclidean inner product \\
    $\|\cdot\|_2$ & Euclidean norm \\
    $\|\cdot\|_F$ & Frobenius norm \\
    $S_d(\mathbb{R})$ & Symmetric matrices \\
    $S_d^{+}(\mathbb{R})$ & Symmetric positive semi-definite matrices \\
    $S_d^{++}(\mathbb{R})$ & Symmetric positive definite matrices \\
    $GL_d(\mathbb{R})$ & Invertibles matrices in $\mathbb{R}^{d\times d}$ \\
    $SO_d(\mathbb{R})$ & Group of rotations \\
\end{tabular}

\medskip

\begin{tabular}{p{3cm}p{11cm}p{1cm}}
\textbf{Functions} & \\
    $\nabla$ & Gradient operator \\
    $C(X)$ & The set of continuous functions from $X$ to $\mathbb{R}$ \\
    $C_0(X)$ & The set of continuous functions from $X$ to $\mathbb{R}$ that vanish at infinity \\
    $C_b(X)$ & The set of continuous bounded functions from $X$ to $\mathbb{R}$ \\
    $L^p(X)$ & The set of $p$-integrable function \\
    $\mathrm{prox}_f(x)$ & Proximal operator of $f$ defined as $\mathrm{prox}_f(x) = \argmin_y \frac{\|x-y\|_2^2}{2} + f(y)$ \\
    $\divop$ & Divergence operator, $\divop(f) = \langle \nabla, f\rangle$ \\
    $\Delta$ & Laplacian operator, $\Delta f = \langle \nabla, \nabla f\rangle$
\end{tabular}

\begin{tabular}{p{3cm}p{11cm}p{1cm}}
\textbf{Measure theory} & \\
    $\mathcal{P}(X)$ & The set of probability measures on the space $X$ \\
    $\mathcal{P}_p(X)$ & The set of probability measures with moments of order $p$ on the space $X$ \\
    $\mathcal{M}(X)$ & The set of measures on $X$ \\
    $\mathcal{M}_+(X)$ & The set of positive measures on $X$ \\
    $\mathcal{P}_{ac}(X), \mathcal{M}_{ac}(X)$ & The set of absolutely continuous probability distributions and measures respectively \emph{w.r.t.} the Lebesgue measure \\
    $\#$ & Push-forward operator \\
    $\mathcal{B}(\mathbb{R}^d)$ & Borelians \\
    $W_p(\mu,\nu)$ & $p$-Wasserstein distance between $\mu$ and $\nu$ \\
    $GW_p(\mu,\nu)$ & $p$-Gromov-Wasserstein distance between $\mu$ and $\nu$  \\  
    $\sw_p(\mu,\nu)$ & Sliced-Wasserstein distance between $\mu$ and $\nu$ \\
    $\kl(\mu||\nu)$ & Kullback-Leibler divergence between $\mu$ and $\nu$ \\
    $\delta_x$ & Dirac measure, \emph{i.e.} $\delta_x(A) = 1$ if $x\in A$, $0$ otherwise \\
    $\lambda$ & Uniform measure on $S^{d-1}$ \\
    $\Pi(\mu,\nu)$ & Set of couplings \\
    $\pi^i$ & Projection on the $i$-th marginal, \emph{i.e.} $\pi^i:(x_1,\dots,x_d)\mapsto x_i$ \\
    $\mu\otimes \nu$ & Independent coupling, \emph{i.e.} $\mu\otimes\nu(A\times B) = \mu(A)\nu(B)$ \\
    $(\id, T)_\#\mu$ & Measure $\mu\otimes T_\#\mu$ \\
    $F_\mu$ & Cumulative distribution function of $\mu$ \\
    $F_\mu^{-1}$ & Quantile function of $\mu$ \\
    $\mathcal{N}(m,\Sigma)$ & Gaussian distribution with mean $m$ and covariance $\Sigma$ \\
    $BW(\mathbb{R}^d)$ & Bures-Wasserstein space \\
    $\mu\ll\nu$ & $\mu$ absolutely continuous \emph{w.r.t.} $\nu$, \emph{i.e.} $\nu(A)=0$ implies $\mu(A)=0$ \\
    $x\sim\mu$ & $x$ is a sample drawn from $\mu$ \\
    $\mathcal{F}$ & Fourier transform \\
    $R$ & Radon transform \\
    $\frac{\delta\mathcal{F}}{\delta\mu}$ & First variation of $\mathcal{F}$ \\
    $\nabla_{W_2}\mathcal{F}$ & Wasserstein gradient of $\mathcal{F}$, $\nabla_{W_2}\mathcal{F} = \nabla \frac{\delta \mathcal{F}}{\delta \mu}$ \\
\end{tabular}

\medskip

\begin{tabular}{p{3cm}p{11cm}p{1cm}}
\textbf{Manifolds} & \\
    $T_x\mathcal{M}$ & Tangent space at $x\in\mathcal{M}$ \\
    $T\mathcal{M}$ & Tangent bundle \\
    $B^\gamma$ & Busemann function associated with the geodesic ray $\gamma$ \\
    $S^{d-1}$ & Unit sphere on $\mathbb{R}^d$, \emph{i.e.} $S^{d-1} = \{x\in\mathbb{R}^d,\ \|x\|_2 = 1\}$ \\
    $\mathbb{V}_{d,k}$ & Stiefel manifold, \emph{i.e.} $\mathbb{V}_{d,k} = \{U\in\mathbb{R}^{d\times k},\ U^TU = k\}$ \\
    $\mathcal{G}_{d,k}$ & Grassmann manifold, \emph{i.e.} $\mathcal{G}_{d,k} = \{E\subset \mathbb{R}^d,\ \mathrm{dim}(E)=k\}$ \\
    $\mathbb{H}^d$ & d-dimensional Hyperbolic space \\
    $\langle x, y\rangle_\mathbb{L}$ & Minkowski pseudo inner-product, \emph{i.e.} $\forall x,y\in\mathbb{R}^{d+1},\ \langle x,y\rangle_\mathbb{L} = -x_0 y_0 + \sum_{i=1}^d x_i y_i$ \\
    $\mathbb{L}^d$ & Lorentz model in $\mathbb{R}^{d+1}$, \emph{i.e.} $\mathbb{L}^d = \{(x_0,\dots,x_d)\in\mathbb{R}^{d+1},\ \langle x,x\rangle_\mathbb{L}=-1,\ x_0>0\}$ \\    $\mathbb{B}^d$ & Poincaré ball \\
    $d_\mathbb{L}, d_\mathbb{B}$ & Geodesic distances on respectively the Lorentz model and the Poincaré ball \\
    $P_{\mathbb{L}\to\mathbb{B}}, P_{\mathbb{B}\to\mathbb{L}}$ & Isometric projection from $\mathbb{L}^d$ to $\mathbb{B}^d$ and from $\mathbb{B}^d$ to $\mathbb{L}^d$ \\
    $d_{AI}, d_{LE}$ & Geodesic distance on $S_d^{++}(\mathbb{R})$ endowed respectively with the Affine-Invariant metric and the Log-Euclidean metric \\
    $P^v$ & Coordinate projection on the geodesic with direction $v$ \\
    $\Tilde{P}^v$ & Projection on the geodesic with direction $v$ \\
    $\gchsw$ & Geodesic Cartan-Hadamard Sliced-Wasserstein \\
    $\hchsw$ & Horospherical Cartan-Hadamard Sliced-Wasserstein \\
    $\chsw$ & $\gchsw$ or $\hchsw$ \\
    $\ghsw$ & Geodesic Hyperbolic Sliced-Wasserstein \\
    $\hhsw$ & Horospherical Hyperbolic Sliced-Wasserstein \\
    $\hsw$ & $\ghsw$ or $\hhsw$ \\
    $\lespdsw$ & Sliced-Wasserstein on $S_d^{++}(\mathbb{R})$ endowed with the Log-Euclidean metric \\
    $\aispdsw$ & Horospherical Sliced-Wasserstein on $S_d^{++}(\mathbb{R})$ endowed with the Affine-Invariant metric \\
    $\ssw$ & Spherical Sliced-Wasserstein \\
\end{tabular}



\clearemptydoublepage
\mainmatter
\cleartooddpage[\thispagestyle{empty}]
\chapter{Introduction}
{
    \hypersetup{linkcolor=black} 
    \minitoc 
}

\chaptermark{Introduction}

In Machine Learning (ML), we aim at learning the best possible model for a given task from a training set of data. The data can have different structures, from point clouds to images or graphs, and can lie on different spaces. A convenient way to model the data is to assume they follow an underlying unknown probability distribution. Thus it is important to develop tools to cope with probability distributions such as metrics to be able to compare them, or algorithms to learn them, as well as developing efficient ways to model them. Moreover, considering the amount of data available and their potential high dimensionality, these methods need to be able to scale well with the number of samples in the data and with the dimension.



For instance, generative modeling is a popular task in ML, which has received a lot of attention lately through Large Language Models (LLMs) which aim at generating text \citep{brown2020language, touvron2023llama, openai2023gpt4}, or through diffusion models which aim at generating images \citep{rombach2022high, ramesh2022hierarchical, saharia2022photorealistic}. Typically, the objective of these tasks is to learn the unknown distribution of the data in order to be able to sample new examples. This amounts to minimizing a well chosen discrepancy between probability distributions. To model the unknown probability distribution, practitioners leverage Deep Learning using neural networks. Popular frameworks include Generative Adversarial Networks (GANs) \citep{goodfellow2014generative}, Variational Autoencoders (VAEs) \citep{kingma2013auto}, Normalizing Flows (NFs) \citep{papamakarios2021normalizing} or more recently Score-Based generative models \citep{sohl2015deep, song2019generative}.

A typical loss to minimize in order to learn a probability distribution is the Kullback-Leibler divergence (KL), which is tightly related with the Maximum Likelihood learning task widely used in statistics to find a good estimator of the data. For example, Normalizing Flows leverage invertible neural networks and the change-of-variable formula to minimize the KL. VAEs instead use arbitrary architectures and are trained by minimizing a lower bound on the KL. GANs are a popular alternative, which use adversarial training, and minimize the Jensen-Shannon divergence. Score-based models also indirectly minimize the KL by learning the score of the data and then performing a diffusion scheme which minimizes the KL. Many alternatives to the KL divergence have been considered such as more general $f$-divergences \citep{nowozin2016f} or Maximum Mean Discrepancies (MMDs) \citep{li2017mmd, binkowski2018demystifying, mroueh2021convergence}.


However, these different objective discrepancies usually require both distributions to have densities, to share the same support or/and do not necessarily respect well the geometry of the data \citep{arjovsky2017wasserstein}. A popular alternative for handling probability distributions while respecting the geometry of the data through a specified ground cost and for being able to compare distributions which do not necessarily have the same support is Optimal Transport (OT) \citep{villani2009optimal}, which allows comparing distributions by finding the cheapest way to move mass from one distribution to another. Thus, OT losses have been used for generative modeling as another alternative to the KL through \emph{e.g.} the Wasserstein GANs \citep{arjovsky2017wasserstein} or the Wasserstein Autoencoders \citep{tolstikhin2018wasserstein}.

Yet, in its original formulation, OT suffers from a computational bottleneck and from the curse of dimensionality, which can hinder its usability in ML applications, in particular for large scale problems. Thus, this thesis will focus on the development and analysis of efficient OT methods with the objective to apply them on Machine Learning problems.


\section{Optimal Transport for Machine Learning}

Optimal Transport \citep{villani2009optimal} provides a principled way to compare probability distributions while taking into account their underlying geometry. This problem, first introduced by \citet{monge1781memoire}, originally consists of finding the best way to move a probability distribution to another with respect to some cost function. This provides two quantities of interest. The first one is the Optimal Transport map (and more generally the OT plan), which allows to push a source distribution onto a target distribution, and the second one is the optimal value of the underlying problem, which quantifies how close two probability distributions are and actually defines a distance between them usually called the Wasserstein distance (when using a well chosen cost).

Keeping in mind these two items
, the Optimal Transport problem has received a lot of attention in the last few years. On the one hand, the OT map, also called the Monge map, can be used effectively in many practical problems such as domain adaptation \citep{courty2016optimal}, where we aim at classifying the data from a target probability distribution from which we do not have training examples through another dataset which we use as training set. Thus, the OT map helps to align the source dataset towards the target dataset, which then allows to use a classifier learned on the source dataset. It has also been useful for text alignment, such as translation, where we want to align two embeddings of different languages \citep{grave2019unsupervised}, in computational biology \citep{schiebinger2019optimal,bunne2021learning,bunne2022supervised}, in computer vision \citep{makkuva2020optimal} or in physics applications such as cosmology \citep{nikakhtar2022optimal, panda2023semi}. 
However, finding this map can be challenging \citep{perrot2016mapping}. A recent line of works models the Monge map with neural networks \citep{seguy2018large, makkuva2020optimal, korotin2021wasserstein, rout2022generative, fan2022scalable, uscidda2023monge, morel2023turning}. This allows to link arbitrary samples of two distributions which can be interesting in some situations \citep{bunne2022supervised,panda2023semi} or to be used for generative modeling tasks where we aim at sampling from some complicated target distribution (for example a distribution of images) given samples from a tractable standard distribution \citep{makkuva2020optimal, huang2021convex}.

In this thesis, we will mostly  be interested in the distance properties of the OT problem. As it provides a principled way to compare probability distributions, it has been used \emph{e.g.} to classify documents which can be seen as probability distributions over words \citep{kusner2015word, huang2016supervised}, to perform dimensionality reductions for datasets of histograms or more generally of probability distributions using Principal Component Analysis (PCA) \citep{seguy2015principal,bigot2017geodesic, cazelles2018geodesic} or Dictionary Learning \citep{rolet2016fast, schmitz2018wasserstein, mueller2022geometric}, or to perform clustering \citep{cuturi2014fast} with \emph{e.g.} Wasserstein K-Means \citep{domazakis2019clustering, zhuang2022wasserstein}. 
It also provides effective losses for supervised learning problems \citep{frogner2015learning} or generative modeling tasks with Wasserstein GANs \citep{arjovsky2017wasserstein, gulrajani2017improved, genevay2017gan} or Wasserstein Autoencoders \citep{tolstikhin2018wasserstein}. The OT cost has also been used in order to obtain straighter trajectories of flows leading to faster and better inference \citep{finlay2020train, onken2021ot, tong2023conditional}. Furthermore, the space of probability measures endowed with the Wasserstein distance has a geodesic structure \citep{otto2001geometry}, which allows to derive a complete theory of gradient flows \citep{ambrosio2008gradient}. It led to the derivation of many algorithms which provide meaningful ways to minimize functionals on the space of probability measures \citep{arbel2019maximum, salim2020wasserstein, glaser2021kale, altekruger2023neural} and which are linked with sampling algorithms derived for example in the Markov chain Monte-Carlo (MCMC) community \citep{jordan1998variational, wibisono2018sampling}.

\section{Motivations}


In practical Machine Learning, we may have to deal with large scale problems, where large amounts of data are at hand. In this case, one of the main bottleneck of OT is the computational complexity \emph{w.r.t.} the number of samples to compute the OT distance. To alleviate this computational burden, different solutions have been proposed in the last decade, which made OT very popular in ML.

\paragraph{Alternatives to the original OT problem.}

\looseness=-1 \citet{cuturi2013sinkhorn} proposed to add an entropic regularization to the classical OT problem, which led to a tractable algorithm with a better computational complexity and usable on GPUs \citep{feydy2020geometric}, hence significantly popularizing OT in the ML community \citep{torres2021survey}. 
This objective has notably been used for generative modeling using autodifferentiation \citep{genevay2018learning}. For learning problems, where we aim at learning implicitly the distribution of the data, another popular alternative widely used in Deep Learning is the minibatch approach \citep{genevay2016stochastic, fatras2020learning, fatras2021minibatch} which only uses at each step a small portion of the data. Another family of approaches uses alternatives to the classical OT problem by considering projections on subspaces. These approaches can be motivated on one hand on the fact that high-dimensional distributions are often assumed to be supported on a lower dimensional subspace, or that two distributions on such space only differ on a lower dimensional subspace \citep{niles2022estimation}. On the other hand, these approaches can be computed more efficiently than the classical OT problem while keeping many interesting properties of Optimal Transport and often having better statistical properties in high dimensional settings. In this thesis, we will mostly be interested in methods which rely on projections on subspaces.

\paragraph{Sliced-Wasserstein.} \looseness=-1 In order to take advantage of appealing forms of OT on low dimensional spaces, these methods project the measures on subspaces. The main example of such method is the Sliced-Wasserstein distance (SW) \citep{rabin2012wasserstein, bonnotte2013unidimensional, bonneel2015sliced}, which is defined as the average of the Wasserstein distance between one dimensional projections of the measures over all directions. This distance enjoys many nice properties, and among others, has a low computational complexity. It has proven to be a suitable alternative to the classical Wasserstein distance or to the entropic regularized OT discrepancy. As it is a differentiable loss, it was used in many learning problems such as generative modeling by learning the latent space of autoencoders with Sliced-Wasserstein Autoencoders \citep{kolouri2018sliced}, by learning generators with Sliced-Wasserstein generators \citep{deshpande2018generative, wu2019sliced, lezama2021run}, by training Normalizing Flows \citep{coeurdoux2022sliced, coeurdoux2023learning}, for Variational Inference \citep{yi2023sliced}, or as an objective for non-parametric algorithms \citep{liutkus2019sliced, dai2021sliced, du2023nonparametric}. It has also been used in wide different applications such as texture synthesis \citep{tartavel2016wasserstein, heitz2021sliced}, domain adaptation \citep{lee2019sliced, rakotomamonjy2021statistical, xu2022unsupervised}, approximate bayesian computation \citep{nadjahi2020approximate}, point-cloud reconstructions \citep{nguyen2023self}, two-sample tests \citep{wang2021two, wang2021two_kernel, xu2022central} or to evaluate the performance of GANs \citep{karras2018progressive}. 
Besides, it is a Hilbertian distance and hence can be used to define kernels between probability distributions which can then be plugged in kernel methods \citep{hofmann2008kernel}, which has been done \emph{e.g.} for kernel K-Means, PCA, SVM \citep{kolouri2016sliced} or for regression \citep{meunier2022distribution}.

Since SW became very popular, many variants were designed, either to deal with specific data \citep{nguyen2022revisiting} or to improve its discriminative power by sampling more carefully the directions of the projections \citep{deshpande2019max, rowland2019orthogonal, nguyen2020distributional, nguyen2020improving, dai2021sliced, nguyen2023markovian, nguyen2023energy, ohana2022shedding}, changing the way to project \citep{kolouri2019generalized, chen2020augmented, nguyen2022hierarchical} or the subspaces on which to project \citep{paty2019subspace, lin2021projection, li2022hilbert}. Other works proposed estimators of the SW distance, either to reduce the variance \citep{nguyen2023control} or to alleviate the curse of dimensionality with respect to the projections \citep{nadjahi2021fast}.

\looseness=-1 The slicing process has also received much attention for other types of discrepancies. \citet{nadjahi2020statistical} studied properties of sliced probability divergences, covering for example the Sliced-Wasserstein distance, the Sliced-Sinkhorn divergence or the Sliced-Maximum Mean Discrepancy. It was used \emph{e.g.} to provide a tree sliced variant of the Wasserstein distance \citep{le2019tree}, to generalize divergences which are only well defined between one dimensional distributions to higher dimensional distributions such as the Cramér distance \citep{Kolouri2020Sliced} or to alleviate the curse of dimensionality of the kernelized Stein discrepancy \citep{gong2021sliced}, of the mutual information \citep{goldfeld2021sliced, goldfeld2022k} or of the Total Variation and the Kolmogorov-Smirnov distances to compare MCMC chains \citep{grenioux2023sampling}. It can also be used for score matching tasks \citep{song2020sliced} which was recently put under the spotlight through the diffusion and score-based generative models. It was also extended for many OT based problems such as the multi-marginal problems \citep{cohen2021sliced} or the partial OT problem \citep{figalli2010optimal} in \citep{bonneel2019spot, bai2022sliced} which can deal with measures of different mass and which is a particular case of Unbalanced Optimal Transport problems \citep{benamou2003numerical}.

These previous lines of works focused mainly on Euclidean spaces. However, many data have a known structure which does not suit Euclidean spaces. Indeed, by the manifold hypothesis, it is widely accepted that the data usually lie on a lower dimensional manifold \citep{dong2012nonlinear, bengio2013representation, fefferman2016testing, pope2021the} or a union of lower dimensional manifolds \citep{brown2023verifying}. In some cases, it is possible to know exactly the Riemannian structure of the data. For example, earth data lie on a sphere or hierarchical data can be efficiently embedded in Hyperbolic spaces \citep{nickel2017poincare}. Fortunately, OT is well defined on such spaces \citep{villani2009optimal}. Hence, in ML, OT has recently received attention for data lying on Riemannian manifolds \citep{alvarez2020unsupervised, hoyos2020aligning}. But the focus has been on  using the Wasserstein distance or the entropic regularized OT problem, instead of methods relying on projections on subspaces. In order to bridge this gap, one of the main objectives of the thesis will be to develop Sliced-Wasserstein distances on Riemannian manifolds.

One of the limitations of SW is the lack of OT plan, which can be very useful in many applications such as domain adaptation \citep{courty2016optimal}, word embedding alignments with Wasserstein Procrustes \citep{grave2019unsupervised, ramirez2020novel}, single cell alignment \citep{demetci2022scot} or cross-domain retrieval \citep{chuang2022infoot}. To overcome this, one might resort to barycentric projection, which however might not give a good plan as many projections are not meaningful. Finding an OT plan requires us to solve the OT problem, which can be intractable in practice for large scale settings. \citet{muzellec2019subspace} proposed to project the distributions on a subspace, and then to rely on the disintegration of measures to recover an OT plan. In another line of work, \citet{bernton2019approximate, li2022hilbert} instead use the possibly suboptimal OT plan obtained between projections on Hilbert curves.

\paragraph{OT between incomparable data.}

When dealing with incomparable data, \emph{i.e.} data which can not be represented in the same space or which cannot be meaningfully compared between them with distances, for example because of invariances between the data which are not taken into account by the distance, the classical OT problem is not applicable anymore, or at least not successful. While it has been proposed to simultaneously learn latent global transformations along computing the OT distance \citep{alvarez2019towards} or to embed both distributions in a common Euclidean space \citep{alaya2021heterogeneous, alaya2022theoretical}, a popular framework which directly takes into account these invariances while allowing to compare distributions lying on different spaces is the Gromov-Wasserstein distance \citep{memoli2011gromov}. This distance has recently attracted considerable interests 
in ML, for example to compare genomics data \citep{demetci2022scot} or graphs \citep{xu2019scalable, chowdhury2021generalized}. However, it suffers from an even bigger computational cost compared to the original OT problem \citep{peyre2016gromov}, and hence can hardly be used in large scale contexts. While it does not always have a closed-form in one dimension \citep{dumont2022existence, beinert2022assignment}, in some particular cases, a closed-form is available \citep{vayer2020contribution} and a sliced version has been proposed \citep{vayer2019sliced}.

\paragraph{Objectives.}

Here, we sum up some of the objectives of the thesis before describing in the next section more precisely the contributions.

\begin{itemize}
    \item[$\bullet$] First, as many data have a Riemannian structure, we will aim at defining new Sliced-Wasserstein distances on Riemannian manifolds in order to be able to deal efficiently with such data.
    \item[$\bullet$] As SW provides an efficient distance between probability distributions which shares many properties with the Wasserstein distance, a natural question is to study the properties of the underlying gradient flows compared to the Wasserstein gradient flows.
    \item[$\bullet$] Motivated by the robustness properties of the Unbalanced Optimal Transport and the recently proposed Sliced Partial OT methods, we will explore how to extend the slicing process to Unbalanced Optimal Transport in order to be able to compare positive measures.
    \item[$\bullet$] Another objective of the thesis will be to provide new tools to project on subspaces of the space of probability measures, aiming to deal with datasets composed of probability distributions.
    \item[$\bullet$] As a limitation of SW is to not provide an OT plan, we will explore how to compute efficiently OT plans between incomparable spaces using the Gromov-Wasserstein problem.
\end{itemize}

\section{Outline of the Thesis and Contributions}


The focus of this thesis is on OT distances which are based on projections on subspaces. \Cref{chapter:bg_ot} provides the general background on Optimal Transport required to understand the rest of the thesis as well as an overview of the related literature. 

Then, \Cref{part:sw_riemannian} introduces Sliced-Wasserstein distances on Riemannian manifolds and applies it to different Machine Learning problems and on different manifolds. \Cref{part:ot} covers either applications of Optimal Transport based on the Wasserstein distance, or variants of Optimal Transport which are based on projections on subspaces. We detail now in more depth the content and contributions of each chapter. We additionally mention collaborators outside the author's hosting laboratories.


\subsection{\Cref{part:sw_riemannian}: Sliced-Wasserstein on Riemannian Manifolds}

In \Cref{part:sw_riemannian}, we study the extension of the Sliced-Wasserstein distance, originally well defined on Euclidean spaces, to Riemannian manifolds. More precisely, we introduce first in \Cref{chapter:sw_hadamard} a way to construct Sliced-Wasserstein distances on (Cartan-)Hadamard manifolds and introduce some of its properties. Then, we leverage in \Cref{chapter:hsw} and \Cref{chapter:spdsw} this general construction to build Sliced-Wasserstein distances on specific Hadamard manifolds: Hyperbolic spaces and the space of Symmetric Positive Definite (SPD) matrices. Finally, in \Cref{chapter:ssw}, we study the case of the sphere, which does not enter the previous framework as it is not a Hadamard manifold.

\subsubsection{\Cref{chapter:sw_hadamard}: Sliced-Wasserstein on Cartan-Hadamard Manifolds} 

In this chapter, by seeing $\mathbb{R}^d$ 
as a particular case of Riemannian manifold, we derive the tools to extend Sliced-Wasserstein distances on geodesically complete Riemannian manifolds. More precisely, we identify lines as geodesics, and propose to project measures on geodesics of manifolds.

We focus here on geodesically complete Riemannian manifolds of non-positive curvature, which have the appealing property that their geodesics are isometric to $\mathbb{R}$. This allows projecting the measures on the real line where the Wasserstein distance can be easily computed. Moreover, we propose to use two different ways to project on the real line. Both ways are natural extensions of the projection in the Euclidean case. The first one is the geodesic projection, which projects by following the shortest paths, and which allows to define the Geodesic Cartan-Hadamard Sliced-Wasserstein distance ($\gchsw$). The second one is the horospherical projection, which projects along horospheres using the level sets of the Busemann function, and which allows to define the Horospherical Cartan-Hadamard Sliced-Wasserstein distance ($\hchsw$).


Then, we analyze theoretically these two constructions and show that many important properties of the Euclidean Sliced-Wasserstein distance still hold on Hadamard manifolds. More precisely, we discuss their distance properties, derive their first variations and show that they can be embedded in Hilbert spaces. Then, we derive their projection complexity as well as their sample complexity, which similarly as in the Euclidean case, are independent of the dimension.

\subsubsection{\Cref{chapter:hsw}: Hyperbolic Sliced-Wasserstein}

In this chapter, we leverage the general constructions derived in \Cref{chapter:sw_hadamard} and apply it to Hyperbolic spaces, which are particular cases of Hadamard manifolds, as they are of (constant) negative curvature. 

Since there are different (equivalent) parameterizations of Hyperbolic spaces, we study the case of the Lorentz model and of the Poincaré ball, and derive the closed-form formulas to define and compute efficiently the Geodesic Hyperbolic Sliced-Wasserstein distance ($\ghsw$) and Horospherical Hyperbolic Sliced-Wasserstein distance ($\hhsw$). We also show that these two formulations can be used equivalently in both the Poincaré ball and the Lorentz model.

Then, we compare the behavior of $\ghsw$, $\hhsw$ and the Euclidean Sliced-Wasserstein distance on the Poincaré ball and on the Lorentz model on different tasks such as gradient descent or classification problems with deep neural networks.

This chapter is based on \citep{bonet2022hyperbolic} and has been presented at the workshop on Topology, Algebra and Geometry in Machine Learning (TAG-ML) of the International Conference of Machine Learning (ICML 2023). The code is open sourced at \url{https://github.com/clbonet/Hyperbolic_Sliced-Wasserstein_via_Geodesic_and_Horospherical_Projections}.

\subsubsection{\Cref{chapter:spdsw}: Sliced-Wasserstein on Symmetric Positive Definite Matrices}

\looseness=-1 In this chapter, we introduce Sliced-Wasserstein distances on the space of Symmetric Positive Definite matrices (SPD). Endowed with specific metrics, the space of SPDs is of non-positive curvature and hence a Hadamard manifold. Thus, we can also use the theory introduced in \Cref{chapter:sw_hadamard} to define Sliced-Wasserstein distances.

\looseness=-1 We study the space of SPDs endowed with two specific metrics: the Affine-Invariant metric and the Log-Euclidean metric. With the Affine-Invariant metric, the space of SPDs is of non-positive and variable curvature. As deriving a closed-form of the geodesic projection is challenging, we first focus on the Busemann projection and introduce the Horospherical SPD Sliced-Wasserstein distance ($\aispdsw$). However, $\aispdsw$ is computationally costly to compute in practice. Thus, it motivates to use the Log-Euclidean metric, which can be seen as a first-order approximation of the Affine-Invariant metric \citep{arsigny2005fast, pennec2020manifold} and which is easier to compute in practice. Endowed with this metric, the space of SPDs is of null curvature and we can derive the counterpart SPD Sliced-Wasserstein distance $\lespdsw$.

We derive some complementary properties for $\lespdsw$. And then, we apply this distance to problems of Magnetoencephalography and of Electroencephalography (M/EEG) such as brain-age prediction or domain adaptation for Brain Computer Interface applications.

This chapter is based on \citep{bonet2023sliced} and has been accepted at the International Conference of Machine Learning (ICML 2023). The code is in open source and can be accessed at \url{https://github.com/clbonet/SPDSW}. This work was made in collaboration with Benoît Malézieux (Inria).

\subsubsection{\Cref{chapter:ssw}: Spherical Sliced-Wasserstein}

We study in this chapter a way to define a Sliced-Wasserstein distance on the sphere. Contrary to the previous chapters, the sphere is of positive curvature and hence is not a Hadamard manifold. Thus, we cannot leverage directly the framework introduced in \Cref{chapter:sw_hadamard}.

Taking into account the particularities of the sphere, we introduce a Spherical Sliced-Wasserstein distance ($\ssw$) by projecting the measures on any great circle, which are the geodesics of the sphere. For the practical implementation, we derive a closed-form of the geodesic projection, and we use the algorithm of \citet{delon2010fast} to compute the Wasserstein distance on the circle. Moreover, we also introduce a closed-form to compute the Wasserstein distance on the circle between an arbitrary measure and the uniform distribution on $S^1$. On the theoretical side, we study some connections with a spherical Radon transform, known as the semicircle transform on $S^2$, allowing us to investigate its distance properties. 

Then, we illustrate the use of this discrepancy on Machine Learning tasks such as sampling, density estimation and generative modeling.

This chapter is based on \citep{bonet2023spherical} and has been accepted at the International Conference of Learning Representations (ICLR 2023). The code has been released at \url{https://github.com/clbonet/Spherical_Sliced-Wasserstein}. Moreover, we implemented it in the open-source library Python Optimal Transport (POT) \citep{flamary2021pot}.

\subsection{\Cref{part:ot}: Optimal Transport and Variants through Projections}

In \Cref{part:ot}, we study different problems which involve projections on subspaces and Optimal Transport. Firstly, in \Cref{chapter:swgf}, we investigate gradient flows in the space of probability measures endowed with the Sliced-Wasserstein distance compared with when endowed with the Wasserstein distance. Then, in \Cref{chapter:usw}, we develop a framework to compare positive measures with Sliced Optimal Transport methods. In \Cref{chapter:busemann}, we investigate the Busemann function in the space of probability measures endowed with the Wasserstein distance. And finally, in \Cref{chapter:gw}, we develop a subspace detour based approach for the Gromov-Wasserstein problem.

\subsubsection{\Cref{chapter:swgf}: Gradient Flows in Sliced-Wasserstein Space}

A way to minimize functionals on the space of probability measures is to use Wasserstein gradient flows, which can be approximated through the backward Euler scheme, also called the Jordan-Kinderlehrer-Otto (JKO) scheme. However, this can be computationally costly to compute in practice. Hence, in this chapter, we propose to replace the Wasserstein distance in the backward Euler scheme by the Sliced-Wasserstein distance to alleviate the computational burden. This amounts to computing gradient flows in the space of probability measures endowed with the Sliced-Wasserstein distance. Modeling probability distributions through neural networks, we propose to approximate the trajectory of the Sliced-Wasserstein gradient flows of particular functionals, and to compare their trajectory with their Wasserstein gradient flows.

We study different types of functionals. First, we study the Kullback-Leibler divergence which requires to use invertible neural networks - called Normalizing Flows - in order to be able to approximate it in practice. With a Gaussian target, we know exactly its Wasserstein gradient flow, and we therefore compare its trajectory with the approximated Sliced-Wasserstein gradient flow. Then, we also study the capacity of our method to approximate the target measure on real data in a setting of Bayesian logistic regression. Furthermore, we study the minimization of 
the Sliced-Wasserstein distance to learn high-dimensional target measure such as distribution of images.

This chapter is based on \citep{bonet2022efficient} and has been published in the journal Transactions on Machine Learning Research (TMLR). The code is available online at \url{https://github.com/clbonet/Sliced-Wasserstein_Gradient_Flows}.

\subsubsection{\Cref{chapter:usw}: Unbalanced Optimal Transport Meets Sliced-Wasserstein}

In some cases, it can be beneficial to compare positive measures instead of probability distributions. This led to the development of the Unbalanced Optimal Transport (UOT) problem which relaxes the OT cost to be able to deal with positive measures.

We study in this chapter how to efficiently slice these methods in two ways. First, we naively propose to average the UOT problem between the projected measures, hence extending \citep{bonneel2019spot, bai2022sliced} to more general UOT problems and denoted SUOT. As one of the main feature of UOT is to remove outliers of the original marginals, we also introduce the Unbalanced Sliced-Wasserstein distance (USW), which performs the regularization on the original marginals. The practical implementation is made using the Frank-Wolfe algorithm building upon \citep{sejourne2022faster}.

This chapter is based on a paper under review \citep{sejourne2023unbalanced}, and is a collaborative effort with Thibault Séjourné (EPFL), Kimia Nadjahi (MIT), Kilian Fatras (Mila) and Nicolas Courty. The main contribution of the author of the thesis is on the experiment side, where we show on a document classification task the benefits of using USW instead of SUOT. The algorithm is also flexible enough to deal with any sliced OT problem, and we illustrate it by computing the Unbalanced Hyperbolic Sliced-Wasserstein distance which builds upon \Cref{chapter:hsw}.

\subsubsection{\Cref{chapter:busemann}: Busemann Function in Wasserstein Space}

The Busemann function, associated to well chosen geodesics, provides (in some sense) a natural generalization of the inner product on manifolds. Thus, its level sets can be seen as a natural counterpart of hyperplanes. It has been recently extensively used on Hadamard manifolds such as Hyperbolic spaces in order to perform PCA or classification tasks \citep{chami2021horopca, ghadimi2021hyperbolic}.

To deal with datasets composed of probability measures, this chapter studies the Busemann function on the space of probability measures endowed with the Wasserstein distance (Wasserstein space). 
In the Wasserstein space, it is not defined for every geodesic. Hence, we first identify for which geodesics this function is well defined. Then, we provide closed-form formulas in particular cases: probability measures on the real line and Gaussian distributions. We also illustrate the use of this function on a Principal Component Analysis application on one dimensional distributions.

This work is done in collaboration with Elsa Cazelles (IRIT).

\subsubsection{\Cref{chapter:gw}: Subspace Detours Meet Gromov-Wasserstein}

In this chapter, we are interested in alleviating the computational cost of the Gromov-Wasserstein problem, while still being able to compute an OT plan between the original measures. Thus, we proposed to extend the subspace detour approach, originally introduced by \citet{muzellec2019subspace} for the OT problem, to the Gromov-Wasserstein problem. As the Gromov-Wasserstein problem requires only to compute distances in each space, we propose to project on a different subspace the source and the target, which can allow to better conserve the true OT plan. We derive some theoretical properties of the problem, and notably a closed-form formula for the coupling based on the subspace detour approach when both measures are Gaussians and the problem is restricted to Gaussian couplings. Then, we illustrate this approach on a shape matching problem.

In a second part, we introduce a new OT cost, which shares the property of the original OT problem to be formally connected to the Knothe-Rosenblatt coupling under a degenerated cost.

\looseness=-1 This chapter is based on \citep{bonet2021subspace} and has been presented at the Neurips workshop OTML2021 and published in the journal Algorithms. It was made in collaboration with Titouan Vayer (Inria).

\clearemptydoublepage
\cleartooddpage[\thispagestyle{empty}]
\clearemptydoublepage

\chapter{Background on Optimal Transport} \label{chapter:bg_ot}

{
    \hypersetup{linkcolor=black} 
    \minitoc 
}


In this chapter, we provide some background knowledge on Optimal Transport, which is required to motivate and understand the contributions in the next chapters. More precisely, in \Cref{sec:general_ot}, we will start with a general description of the OT problem, from the Monge problem to the Kantorovich problem, with some of its variants such as the Gromov-Wasserstein problem. Then, in \Cref{sec:computational_ot}, we will discuss how we can solve the problem in practice by presenting different possibilities to model probability distributions, along the computational methods and variants. Last but not least, in \Cref{sec:sw}, we will introduce the Sliced-Wasserstein distance, another alternative to the classical OT problem which will be of most interest in the rest of the thesis.

For more details about Optimal Transport, we refer to the books of \citet{villani2021topics, villani2009optimal} or of \citet{santambrogio2015optimal}. For the computational aspect, we refer to \citep{peyre2019computational}.

\section{General Optimal Transport Problem} \label{sec:general_ot}

\subsection{Optimal Transport Problem}


\paragraph{Monge and Kantorovich problem.}

\looseness=-1 Optimal transport is a problem which consists originally of moving a source probability distribution towards a target probability distribution in an optimal way. This was first introduced by \citet{monge1781memoire} and is known nowadays as the Monge problem. Let $\mu,\nu\in \mathcal{P}(\mathbb{R}^d)$ be two probability distributions, then moving the source $\mu$ towards the target $\nu$ can be formalized as finding a transport map $T:\mathbb{R}^d\to \mathbb{R}^d$ such as $T_\#\mu = \nu$ where $\#$ is the push-forward operator, defined as,
\begin{equation}
    \int h\big(T(x)\big)\ \mathrm{d}\mu(x) = \int h(y)\ \mathrm{d}(T_\#\mu)(y),
\end{equation}
for all continuous maps $h$. Equivalently, it can be characterized through measurable sets as 
\begin{equation}
    \forall A\in\mathcal{B}(\mathbb{R}^d),\ \nu(A) = \mu\big(T^{-1}(A)\big),
\end{equation}
where $\mathcal{B}(\mathbb{R}^d)$ is the set of all Borelians. 
Now that we know how to formally push measures, we can find the optimal way to move measures using the Monge problem defined as
\begin{equation} \label{eq:monge_problem}
    M_c(\mu,\nu) = \inf_{T_\#\mu=\nu}\ \int c\big(x,T(x)\big)\ \mathrm{d}\mu(x),
\end{equation}
where $c:\mathbb{R}^d\times \mathbb{R}^d\to \mathbb{R}$ denotes some cost, which will characterize in which way the transformation is optimal. Various cost functions give different OT costs. Typically, in this manuscript, the OT cost will be chosen as a distance. Unfortunately, this problem might not always have a solution. For example, in the simple case where $\mu=\delta_x$ and $\nu = \frac12 \delta_{y_1} + \frac12 \delta_{y_2}$ with $y_1\neq y_2$ and where $\delta$ denotes the Dirac measure, there is no transformation $T$ such that $T_\#\mu=\nu$ and thus the cost is infinite. A solution to this issue was introduced by \citet{kantorovich1942translocation}, and consists of relaxing the problem by looking for an optimal coupling instead of an optimal map, and hence allowing to split the mass. This defines the Kantorovich problem
\begin{equation} \label{eq:Kantorovich_problem}
    W_c(\mu,\nu) = \inf_{\gamma\in\Pi(\mu,\nu)}\ \int c(x,y)\ \mathrm{d}\gamma(x,y),
\end{equation}
where $\Pi(\mu,\nu) = \{\gamma\in\mathcal{P}(\mathbb{R}^d\times \mathbb{R}^d),\ \pi^1_\#\gamma=\mu, \pi^2_\#\gamma=\nu\}$ denotes the set of couplings between $\mu$ and $\nu$, and with $\pi^1:(x,y)\mapsto x$ and $\pi^2:(x,y)\mapsto y$ the projections on the marginals. As $\Pi(\mu,\nu)$ always contains at least the independent coupling $\mu\otimes \nu$ (defined as $\mu\otimes \nu(A\times B) = \mu(A)\nu(B)$ for all Borelians $A,B\in\mathcal{B}(\mathbb{R}^d)$), the set of constraints is never empty. Under assumptions on the cost $c$, there is always a solution to this problem \citep[Theorem 1.7]{santambrogio2015optimal}.

Furthermore, when the solution is of the form $(\id,T)_\#\mu$, the solutions of the Monge problem and of the Kantorovich problem coincide. It is also easy to see that the Kantorovich problem gives a lower bound of the Monge problem as the set of measures $\{(\id,T)_\#\mu,\ T_\#\mu=\nu\}$ is included in the set of couplings $\Pi(\mu,\nu)$. Many works have been devoted to characterizing when both solutions coincide. An important theorem of \citet{brenier1991polar} states that it is \emph{e.g.} the case when $c(x,y) = \frac12\|x-y\|_2^2$ and $\mu$ is absolutely continuous with respect to the Lebesgue measure. Furthermore, he characterizes the solution as the gradient of a convex function.
\begin{theorem}[Brenier's Theorem] \label{th:brenier}
    Let $\mu,\nu\in\mathcal{P}_2(\mathbb{R}^d)$ and $c(x,y)=\frac12\|x-y\|_2^2$. Suppose that $\mu$ is absolutely continuous with respect to the Lebesgue measure. Then, there exists a unique optimal coupling $\gamma^*$ solution of \eqref{eq:Kantorovich_problem} of the form $\gamma^*=(\id, T^*)_\#\mu$ where $T^*$ is the unique solution ($\mu$-almost everywhere) of \eqref{eq:monge_problem}. Furthermore, $T^*$ is of the form $T^* = \nabla \varphi$ where $\varphi:\mathbb{R}^d\to\mathbb{R}$ is a convex function.
\end{theorem}
Such a convex function $\varphi$ is called a Brenier potential.
This result can further be extended under conditions on the cost such as being strictly convex \citep{santambrogio2015optimal}.

\paragraph{Wasserstein distance.}

For now, we have only been interested in the optimal solution. However, the value of the problem can also prove itself interesting as such value characterizes how far the distributions are from one another. In particular, in the case where the cost is chosen as $c(x,y)=\|x-y\|_2^p$ for $p\ge 1$, then it defines a finite distance on $\mathcal{P}_p(\mathbb{R}^d)=\{\mu\in\mathcal{P}(\mathbb{R}^d),\ \int \|x\|_2^p\ \mathrm{d}\mu(x) < \infty\}$, the space of probability measures with moments of order $p$, called the Wasserstein distance. 
\begin{definition}[Wasserstein distance]
    Let $p\ge 1$ and $\mu,\nu\in\mathcal{P}_p(\mathbb{R}^d)$. The $p$-Wasserstein distance between $\mu$ and $\nu$ is defined as
    \begin{equation}
        W_p(\mu,\nu) = \left(\inf_{\gamma\in\Pi(\mu,\nu)}\ \int \|x-y\|_2^p\ \mathrm{d}\gamma(x,y)\right)^{\frac{1}{p}}.
    \end{equation}
\end{definition}
\begin{theorem}[Wasserstein distance]
    For any $p \ge 1$, $W_p$ is a finite distance on $\mathcal{P}_p(\mathbb{R}^d)$, \emph{i.e.} for all $\mu,\nu\in\mathcal{P}_p(\mathbb{R}^d)$,  $W_p(\mu,\nu)<\infty$ and
    \begin{enumerate}
        \item $\forall \mu,\nu\in\mathcal{P}_p(\mathbb{R}^d)$, $W_p(\mu,\nu)=W_p(\nu,\mu)$ (symmetry)
        \item $W_p(\mu,\nu) = 0 \iff \mu=\nu$ (indiscernible property)
        \item $\forall \mu,\nu,\alpha\in\mathcal{P}_p(\mathbb{R}^d)$, $W_p(\mu,\nu) \le W_p(\mu,\alpha) + W_p(\alpha,\nu)$ (triangular inequality)
    \end{enumerate}
\end{theorem}
In particular, this distance has many interesting properties which make it very useful to compare probability distributions. For instance, contrary to usual divergences used in ML such as the KL divergence, it can compare probability distributions which do not share the same support. It also provides a geodesic space structure \citep{otto2001geometry}, which can be interesting \emph{e.g.} to interpolate between measures \citep{mccann1997convexity}. More precisely, a geodesic curve between $\mu_0$ and $\mu_1\in\mathcal{P}_p(\mathbb{R}^d)$ is of the form $\mu_t = ((1-t)\pi^1 + t\pi^2)_\#\gamma$ for $t\in [0,1]$ where $\gamma\in\Pi(\mu_0,\mu_1)$ is an optimal coupling. This curve is also called McCann's interpolation and satisfies for all $s,t\in [0,1]$, $W_p(\mu_s,\mu_t)=|t-s| W_p(\mu_0,\mu_1)$.

Note also that there are other equivalent formulations of the Wasserstein distance, such as the Benamou-Brenier dynamic formulation \citep{benamou2000computational}, or the dual formulation.
\begin{proposition}[Dual formulation]
    Let $p\ge 1$ and $\mu,\nu\in \mathcal{P}_p(\mathbb{R}^d)$, then
    \begin{equation}
        W_p^p(\mu,\nu) = \sup_{(\psi,\phi)\in \mathcal{C}}\ \int \psi\ \mathrm{d}\mu + \int\phi\ \mathrm{d}\nu,
    \end{equation}
    where $\mathcal{C} = \{(\psi,\phi)\in L^1(\mu)\times L^1(\nu),\ \psi(x)+\phi(y) \le \|x-y\|_2^p \text{ for }\mu\otimes\nu\text{-almost every }(x,y)\}$.
\end{proposition}
$\psi$ and $\phi$ are known as Kantorovich potentials. Note also that they can be related with the optimal coupling as, for example for $p=2$ and $(x,y)\in\supp(\gamma^*)$, $\nabla\psi(x) = x-y$ \citep[Section 1.3]{santambrogio2015optimal}. In the particular case where there is a Monge map $T$, we have for $\mu$-almost every $x$, $T(x) = x-\nabla\psi(x)=\nabla\varphi(x)$ where $\varphi(x) = \frac{\|x\|_2^2}{2} - \psi(x)$.

\paragraph{Other OT problems.}

Changing the cost, we can obtain very different OT problems. We can also change the whole objective to deal either with more general problems, or with specific problems which cannot be handled by the original formulation. To provide some examples, let us first define the disintegration of a measure.
\begin{definition}[Disintegration of a measure] \label{def:disintegration}
    Let $(Y,\mathcal{Y})$ and $(Z,\mathcal{Z})$ be measurable spaces, and~$(X,\mathcal{X})=(Y\times Z,\mathcal{Y}\otimes\mathcal{Z})$ the product measurable space. Then, for~$\mu\in\mathcal{P}(X)$, we denote the marginals as $\mu_Y = \pi^Y_\#\mu$ and $\mu_Z=\pi^Z_\#\mu$, where $\pi^Y$ (respectively $\pi^Z$) is the projection on $Y$ (respectively Z). Then, a~family $\big(K(y,\cdot)\big)_{y\in\mathcal{Y}}$ is a disintegration of $\mu$ if for all $y\in Y$, $K(y,\cdot)$ is a measure on $Z$, for~all $A\in\mathcal{Z}$, $K(\cdot,A)$ is measurable and:
    \begin{equation*}
        \forall g\in C(X),\ \int_{Y\times Z} g(y,z)\ \mathrm{d}\mu(y,z) = \int_Y\int_Z g(y,z)K(y,\mathrm{d}z)\ \mathrm{d}\mu_Y(y),
    \end{equation*}
    where $C(X)$ is the set of continuous functions on $X$. We can note $\mu=\mu_Y\otimes K$. $K$ is a probability kernel if for all $y\in Y$, $K(y,Z)=1$.
\end{definition}
The~disintegration of a measure actually corresponds to conditional laws in the context of probabilities. In the case where $X=\mathbb{R}^d$, we have existence and uniqueness of the disintegration (see \citep[Box 2.2]{santambrogio2015optimal} or \citep[Chapter 5]{ambrosio2008gradient} for the more general case).

Then, disintegrating $\gamma\in\Pi(\mu,\nu)\subset\mathcal{P}(\mathbb{R}^d\times\mathbb{R}^d)$ as $\gamma = \mu \otimes K$ where $K$ is a probability kernel, we can write the OT problem as
\begin{equation}
    W_c(\mu,\nu) = \inf_{\gamma\in\Pi(\mu,\nu)}\ \iint c(x,y)\ K(x,\mathrm{d}y)\ \mathrm{d}\mu(x).
\end{equation}
Then, noting $C\big(x, K(x,\cdot)\big) = \int c(x,y) K(x,\mathrm{d}y)$, $W_c$ writes as 
\begin{equation}
    W_c(\mu,\nu) = \inf_{\gamma\in\Pi(\mu,\nu)} \int C\big(x,K(x,\cdot)\big)\ \mathrm{d}\mu(x),
\end{equation}
and changing $C$, we can obtain very different OT cost. This formulation is called the weak OT formulation \citep{gozlan2017kantorovich}. An example of cost is the barycentric weak OT \citep{backhoff2019existence, cazelles2021novel} defined with the following ground cost:
\begin{equation}
    C\big(x,K(x,\cdot)\big) = \left\|x-\int y\ K(x,\mathrm{d}y)\right\|_2^2.
\end{equation}
Another OT problem which allows to deal with incomparable data is the Gromov-Wasserstein problem \citep{memoli2011gromov, sturm2012space}, which can be seen as an extension of the Gromov-Hausdorff distance between spaces \citep{memoli2014gromov}, and which is defined as
\begin{equation}
    GW_c(\mu,\nu) = \inf_{\gamma\in\Pi(\mu,\nu)}\ \iint L\big(c(x,x'),c(y,y')\big)\ \mathrm{d}\gamma(x,y)\mathrm{d}\gamma(x',y'),
\end{equation}
where $L:\mathbb{R}\times \mathbb{R} \to \mathbb{R}$ is some loss function. As it only involves a cost metric computed in each space, it can be used to compare distributions lying in different spaces. Even more generally, we can define the general OT problem \citep{asadulaev2022neural} as minimizing a functional $\mathcal{F}:\mathcal{P}(X\times Y)\to \mathbb{R}$, where $X$ and $Y$ are some spaces, as
\begin{equation}
    \inf_{\gamma\in\Pi(\mu,\nu)}\ \mathcal{F}(\gamma).
\end{equation}



\subsection{Particular Cases with Closed-Forms} \label{section:closed_forms}

In general, we need to solve the infimum problem over the set of couplings, which is not possible between arbitrary measures. However, there are some particular cases in which we know how to solve the problem in closed-form.

\paragraph{One dimensional case.}

First, let us define the cumulative distribution function $F_\mu$ of a measure $\mu\in\mathcal{P}(\mathbb{R})$ as 
\begin{equation}
    \forall t \in \mathbb{R},\ F_\mu(t) = \mu\big(]-\infty, t]\big) = \int \mathbb{1}_{]-\infty, t]}(x)\ \mathrm{d}\mu(x).
\end{equation}
It is well known that $F_\mu$ is a càdlàg function, \emph{i.e.} ``continue à droite, limite à gauche'' (right continuous with left limits). While not always invertible, we can define its pseudo-inverse $F_\mu^{-1}$, also called the quantile function, as
\begin{equation}
    \forall u\in [0,1],\ F_\mu^{-1}(u) = \inf \{x\in\mathbb{R},\ F_\mu(x)\ge u\}.
\end{equation}
Then, we have the following closed-form for the $p$-Wasserstein distance \citep[Proposition 2.17]{santambrogio2015optimal}.
\begin{proposition} \label{prop:wasserstein_1d}
    Let $p\ge 1$, $\mu,\nu\in\mathcal{P}_p(\mathbb{R})$. Then,
    \begin{equation} \label{eq:wasserstein_1d}
        W_p^p(\mu,\nu) = \int_0^1 |F_{\mu}^{-1}(u)-F_{\nu}^{-1}(u)|^p\ \mathrm{d}u.
    \end{equation}
\end{proposition}
If $\mu$ is atomless, as $(F_\mu)_\#\mu = \mathrm{Unif}([0,1])$ \citep[Lemma 2.4]{santambrogio2015optimal}, using the change of variable formula, we know that we have
\begin{equation}
    W_p^p(\mu,\nu) = \int \left|x-F_{\nu}^{-1}\big(F_\mu(x)\big)\right|^p\ \mathrm{d}\mu(x).
\end{equation}
Hence, from this equality, we recognize that the Monge map between $\mu$ (atomless) and $\nu\in\mathcal{P}_p(\mathbb{R})$ is of the form $T(x) = F_{\nu}^{-1}\big(F_\mu(x)\big)$. This function is also known as the increasing rearrangement map. Furthermore, we see also that the derivative of the Kantorovich potential is of the form $\psi'(x) = x-T(x) = x-F_{\nu}^{-1}\big(F_\mu(x)\big)$. More generally, for arbitrary $\mu$, the OT plan is given by $(F_\mu^{-1}, F_\nu^{-1})_\#\mathrm{Unif}([0,1])$ \citep[Theorem 2.9]{santambrogio2015optimal}.

In the light of these closed-forms, the one dimensional case is particularly attractive. Moreover, we observe that the $p$-Wasserstein distance is actually the $L^p$ norm between the quantile functions, and hence a Hilbertian metric. In particular, for $p=2$, the space $\mathcal{P}_2(\mathbb{R})$ endowed with $W_2$ is a Hilbert space. This is actually not the case in higher dimensions, as the Wasserstein space is in general of positive curvature (in the sense of Alexandrov) \citep[Section 7.3]{ambrosio2008gradient}. And it is known that it cannot be embedded in Hilbert spaces in higher dimensions \citep[Section 8.3]{peyre2019computational}.



\paragraph{Gaussian case.} Another particularly interesting case where we have closed-forms is when both measures are Gaussians \citep{givens1984class, gelbrich1990formula, takatsu2008wasserstein}. 

\begin{proposition} \label{prop:closed_forms_gaussians}
    Let $\mu=\mathcal{N}(m_\mu,\Sigma_\mu)$ and $\nu=\mathcal{N}(m_\nu, \Sigma_\nu)$ with $m_\mu,m_\nu\in\mathbb{R}^d$ and $\Sigma_\mu, \Sigma_\nu \in S_d^+(\mathbb{R})$ positive semi-definite matrices. Then, 
    \begin{equation} \label{eq:closed_form_gaussians}
        W_2^2(\mu,\nu) = \|m_\mu-m_\nu\|_2^2 + \tr\left(\Sigma_\mu+\Sigma_\nu - 2(\Sigma_\mu^\frac12 \Sigma_\nu \Sigma_\mu^\frac12)^\frac12\right).
    \end{equation}
    Furthermore, the Monge map is of the form $T:x\mapsto m_\nu + A(x-m_\mu)$ where 
    \begin{equation}
        A = \Sigma_\mu^{-\frac12}(\Sigma_\mu^\frac12 \Sigma_\nu \Sigma_\mu^\frac12)^\frac12\Sigma_\mu^{-\frac12}.
    \end{equation}
\end{proposition}
The second part of \eqref{eq:closed_form_gaussians} actually defines a distance between positive semi-definite matrices known in the literature of quantum information as the Bures distance \citep{bhatia2019bures}. Thus, we often call the Wasserstein distance between Gaussians the Bures-Wasserstein distance.

These results are also true when considering elliptical distributions \citep{gelbrich1990formula, muzellec2018generalizing} or restricting to the Linear Monge problem \citep{flamary2019concentration}. Note also that \eqref{eq:closed_form_gaussians} is always a lower bound of the Wasserstein distance \citep{gelbrich1990formula}.

Restricting the space to Gaussian measures endowed with the Wasserstein distance, we obtain the Bures-Wasserstein space $BW(\mathbb{R}^d)$ \citep{bhatia2019bures}, which is a Riemannian manifold (contrary to the Wasserstein space which has only a Riemannian structure \citep{otto2001geometry}) and has received many attention recently \citep{lambert2022variational, diao2023forward, brechet2023critical}. 


\paragraph{Tree metrics.} 
For particular choices of metrics, the computation of the Wasserstein distance can be alleviated. An example is the one of tree metrics for which all elements where the metric is defined are included in the nodes of a tree and the distance between two points is the length of the path between two nodes \citep{le2019tree, takezawa2022fixed}. Formally, let $\mathcal{T}= (V,E)$ be a tree with $v_0$ as root. For any $v\in V\setminus\{v_0\}$, denote $w_v$ the length of the edge between $v$ and its parent node and denote by $d_\mathcal{T}:V\times V \to \mathbb{R}_+$ the tree metric. Then, denoting $\Gamma(v)$ the set of nodes contained in the subtree rooted at $v$, the 1-Wasserstein distance with cost $d_\mathcal{T}$ between $\mu,\nu\in\mathcal{P}(V)$ is given by \citep[Proposition 1]{evans2012phylogenetic, le2019tree}
\begin{equation}
    W_{d_\mathcal{T}}(\mu,\nu) = \sum_{v\in V} w_v \big|\mu\big(\Gamma(v)\big)-\nu\big(\Gamma(v)\big)\big|.
\end{equation}

\section{Computational Optimal Transport} \label{sec:computational_ot}

In this section, we discuss how to approximate the Wasserstein distance in practice. The first step is to approximate the probability distributions as we generally do not have access to its real form in general. Then, one must see how to obtain the Wasserstein distance numerically. For a more thorough overview of the computational methods to solve the OT problem, we refer to \citep{peyre2019computational}.

\subsection{Modeling Probability Distributions}


\paragraph{Model data as probability distributions.}

In general, we have only access to samples $x_1,\dots,x_n\in\mathbb{R}^d$ and we need to approximate the probability distributions, generally unknown, followed by these samples in order to use the Optimal Transport framework.

\looseness=-1 A first way to approximate the OT distance between samples could be to first approximate their mean and covariance matrix as 
\begin{equation}
    \begin{aligned}
        \hat{m}_n &= \frac{1}{n} \sum_{i=1}^n \delta_{x_i}, \\
        \hat{\Sigma}_n &= \frac{1}{n-1} \sum_{i=1}^n (x_i-\hat{m}_n)(x_i-\hat{m}_n)^T,
    \end{aligned}
\end{equation}
and then approximate the underlying distribution $\mu$ by $\hat{\mu} = \mathcal{N}(\hat{m}_n, \hat{\Sigma}_n)$. This can be a good approximation for high-dimensional datasets for example \citep{bonneel2023survey}. Then, leveraging \Cref{prop:closed_forms_gaussians}, we can easily compute the OT map with complexity $O(nd^2+d^3)$. It has been used for example for color transfer \citep{pitie2007linear}, but also as a quantifier of the quality of generated images, called the Fréchet Inception distance (FID), by comparing the features of Inception models \citep{heusel2017gans}, or to compare graphs \citep{petric2019got} or datasets \citep{alvarez2020geometric}. However, this approximation only uses the two first moments, and can be costly to compute in high dimensional scenarios.

Other approaches directly use the discrete samples to approximate the distribution $\mu$. First, using an Eulerian representation, one can discretize the space with a grid. Then, the approximated distribution is $\hat{\mu}_N = \sum_{i=1}^N \alpha_i \delta_{\hat{x}_i}$ where $(\hat{x}_i)_{i=1}^N$ represents a regular grid of the space, and $\alpha_i$ represents the number of samples $x_j$ which closest point on the grid is $\hat{x}_i$. Note that to have probability distributions, the $(\alpha_i)_{i=1}^N$ are normalized such that $\sum_{i=1}^N \alpha_i = 1$. While these methods can approximate fairly well distributions in low dimensional spaces, they do not scale well with the dimension since the size of the grid augments exponentially with it. Instead, one can use the Lagrangian representation, which maps each point $x_i$ to a Dirac $\delta_{x_i}$ and approximates the distribution as $\hat{\mu}_n = \frac{1}{n}\sum_{i=1}^n \delta_{x_i}$. This is maybe the most straightforward way to approximate the underlying distribution.

\subsection{Estimating the Wasserstein Distance} \label{section:estimation_wasserstein}



As we saw in the previous section, we are often required in practice to approximate the probability distributions with discrete distributions. When approximating $\mu$ and $\nu$ by their sample counterparts $\hat{\mu}_n = \frac{1}{n}\sum_{i=1}^n \delta_{x_i}$ and $\hat{\nu}_n = \frac{1}{n}\sum_{i=1}^n \delta_{y_i}$ where $x_1,\dots,x_n\sim \mu$ and $y_1,\dots,y_n\sim \nu$, it is common practice to approximate the Wasserstein distance $W_p(\mu,\nu)$ by the plug-in estimator  $W_p(\hat{\mu}_n, \hat{\nu}_n)$ \citep{manole2021plugin}. Thus, we discuss in this section how to compute the Wasserstein distance between discrete samples. 

\paragraph{Wasserstein distance as a linear program.} \looseness=-1 Let $\mu = \sum_{i=1}^n \alpha_i \delta_{x_i}$ and $\nu = \sum_{j=1}^m \beta_j \delta_{y_j}$ where for all $i,j$, $x_i,y_j\in\mathbb{R}^d$ and $\alpha  = (\alpha_1,\dots,\alpha_n) \in \Sigma_n$, $\beta = (\beta_1,\dots,\beta_m)\in\Sigma_m$ with $\Sigma_n = \{ \alpha \in \mathbb{R}_+^n,\ \sum_{i=1}^n \alpha_i = 1\}$ the probability simplex. Let's note $C\in\mathbb{R}^{n\times m}$ the matrix such that for any $i,j$, $C_{i,j} = \|x_i-y_j\|_2^p$. The Wasserstein distance then can be written as 
\begin{equation}
    W_p^p(\mu,\nu) = \inf_{P\in \Pi(\alpha,\beta)}\ \langle C, P\rangle,
\end{equation}
where $\Pi(\alpha,\beta) = \{P\in \mathbb{R}_+^{n\times m},\ P\mathbb{1}_m=\alpha,\ P^T\mathbb{1}_n = \beta\}$ is the set of couplings between $\alpha$ and $\beta$. Under this form, the Optimal Transport problem is a linear program \citep{dantzig2002linear} and classical algorithms can be used (see \emph{e.g.} \citep[Section 3]{peyre2019computational}). However, the main bottleneck is that the computational complexity is in general super-cubic $O(n^3\log n)$ with respect to the number of samples $n$ \citep{pele2009fast}. This prevents the computation of the Wasserstein distance in large scale problems. Furthermore, approximating the Wasserstein distance by the plug-in estimator suffers from the curse of dimensionality as $W_p^p(\hat{\mu}_n,\hat{\nu}_n)$ converges toward $W_p^p(\mu,\nu)$ in $O(n^{-\frac{1}{d}})$ \citep{boissard2014mean, niles2022estimation}. This result means that the number of samples required to have an approximation of the same order when augmenting the dimension must increase exponentially. These different drawbacks motivated several approximations which we describe now.


\paragraph{Entropic regularized Optimal Transport.} \looseness=-1 To alleviate the computational cost of computing the Wasserstein distance, it is possible to regularize the problem. Several such regularizations are possible \citep{flamary2014optimal,blondel2018smooth, liu2023sparsityconstrained, lindback2023bringing}, and give different properties. Here, we discuss the most popular which uses the entropic regularization \citep{cuturi2013sinkhorn}. Let $\epsilon>0$, $\mu,\nu\in\mathcal{P}(\mathbb{R}^d)$, then the entropic regularized OT problem is defined as
\begin{equation}
    W_\epsilon(\mu, \nu) = \inf_{\gamma\in\Pi(\mu,\nu)}\ \int c(x,y) \ \mathrm{d}\gamma(x,y) + \epsilon \kl(\pi||\mu\otimes \nu),
\end{equation}
where $\kl$ denotes the Kullback-Leibler divergence (or relative entropy) and is defined as 
\begin{equation}
    \kl(\pi||\mu\otimes\nu) = \left\{
    \begin{array}{ll}
        \int \log\left(\frac{\mathrm{d}\pi(x,y)}{\mathrm{d}\mu(x)\mathrm{d}\nu(y)}\right)\ \mathrm{d}\pi(x,y) & \mbox{ if } \pi \ll \mu\otimes \nu \\
        +\infty & \mbox{ otherwise}.
    \end{array}\right.
\end{equation}
The parameter $\epsilon$ quantifies how much the problem is regularized by the KL divergence. One important feature of this problem is that it can be solved in $O(n^2)$ using the Sinkhorn algorithm \citep{cuturi2013sinkhorn}, which can be implemented efficiently on GPUs. Moreover it converges to the OT problem when $\epsilon\to 0$ \citep{carlier2017convergence} and it is differentiable \citep[Theorem 2]{luise2018differential} contrary to the Wasserstein distance. However, it is known that transport plans are more blurred than the true OT plan \citep{blondel2018smooth}.

One major bottleneck is that this quantity is not a distance nor a divergence as it is biased ($W_\epsilon(\mu,\mu)\neq 0$) \citep{feydy2019interpolating}. This motivated to introduce a correction term and to define the Sinkhorn divergence \citep{ramdas2017wasserstein, genevay2018learning} as 
\begin{equation}
    S_\epsilon(\mu,\nu) = W_\epsilon(\mu,\nu) - \frac12 W_\epsilon(\mu,\mu) - \frac12 W_\epsilon(\nu,\nu).
\end{equation}
This divergence actually interpolates between $W_c(\mu,\nu)$ (as $\epsilon \to 0$) and $\mmd(\mu,\nu)$ for a particular kernel (as $\epsilon\to\infty$), and is a convex, smooth divergence metrizing the weak convergence \citep{feydy2019interpolating}. It was also shown to be useful as an estimator of the squared Wasserstein distance \citep{chizat2020faster}.


\paragraph{Minibatch Optimal Transport.} In Deep Learning applications, loading the whole data on GPUs is typically intractable and practitioners rely on batches of data to optimize the neural networks through stochastic gradient descent. It has naturally been used with OT objectives \citep{genevay2018learning, damodaran2018deepjdot}. \citet{fatras2020learning} formalized the minibatch OT problem as 
\begin{equation}
    MW_p(\mu,\nu) = \mathbb{E}_{X_1,\dots,X_m\sim \mu, Y_1,\dots,Y_m\sim\nu}\left[W_p\left(\frac{1}{m}\sum_{i=1}^m \delta_{X_i}, \frac{1}{m}\sum_{j=1}^m \delta_{Y_j}\right)\right].
\end{equation}
Furthermore, \citet{fatras2020learning} studied the transport plan of the minibatch OT and \citet{fatras2021minibatch} developed the analysis for other OT problems. The computational complexity of solving this problem is in $O(km^3\log m)$ with $k$ the number of mini-batches sampled and $m$ the size of the batches. Note that in Deep Learning applications, we typically choose $k=1$.

\paragraph{Low-rank OT.}

Some recent works proposed to restrain the set of couplings to the ones having low-rank constraints \citep{forrow2019statistical, scetbon2021low, scetbon2022low}. For $r\ge 1$, denote $\Pi_r(\mu,\nu) = \{\gamma\in\Pi(\mu,\nu),\ \exists (\mu_i)_{i=1}^r, (\nu_i)_{i=1}^r \in \mathcal{P}_p(\mathbb{R}^d)^r,\ \lambda\in \Sigma_r^*,\ \text{such that }\gamma = \sum_{i=1}^r \lambda_i (\mu_i\otimes \nu_i)\}$ the set of rank-r coupling, with $\Sigma_r^*$ the subset of the simplex with positive vectors of $\mathbb{R}_+^d$. Then, the low rank OT cost between $\mu,\nu\in\mathcal{P}(\mathbb{R}^d)$ is defined as 
\begin{equation}
    LROT_{c,r}(\mu,\nu) = \inf_{\gamma\in\Pi_r(\mu,\nu)}\ \int c(x,y)\ \mathrm{d}\gamma(x,y).
\end{equation}
\citet{scetbon2021low} showed that using a mirror-descent scheme, this can be solved in $O(nrd)$ and \citet{scetbon2022low} studied some of its statistical properties.

\paragraph{Tree variants.} As we saw earlier, computing the 1-Wasserstein distance with a tree metric can be done efficiently as we have access to a closed-form. Thus, by approximating the Euclidean metric by a tree metric (see \emph{e.g.} \citep{bartal1998approximating}), it is possible to approximate the Wasserstein distance efficiently \citep{backurs2020scalable}. Additionally, using a partition-based tree metric $d_\mathcal{T}^H$ of depth $H$, it can be shown that \citep{le2019tree}
\begin{equation}
    W_2(\hat{\mu}_n, \hat{\nu}_n) \le \frac12 W_{d_\mathcal{T}^H}(\hat{\mu}_n, \hat{\nu}_n) + \beta \frac{\sqrt{d}}{2^H},
\end{equation}
with $\beta$ the side of hypercubes.

\paragraph{Neural estimators for continuous solvers.} While previous methods focus on computing or approximating the OT problem between discrete distributions, some works proposed instead to approximate it directly between continuous distributions. For example, \citet{makkuva2020optimal, korotin2021wasserstein, rout2022generative} leverage the dual formulation and model potential with neural networks before solving an underlying minimax problem. More recently, \citet{uscidda2023monge} proposed to use the Monge gap defined as $\mathcal{M}_\mu(T) = \int c\big(x,T(x)\big)\ \mathrm{d}\mu(x) - M_c(\mu, T_\#\mu)$ as a regularizer to enforce the optimality which can be more efficient to solve and which extends to general costs. These different solvers require neural networks to approximate the Wasserstein distance and are thus more computationally intensive. Moreover, the objective being to compute the Wasserstein distance between general distributions while computing the OT map, the objectives are different from the ones considered in this thesis.

\section{Sliced-Wasserstein Distance} \label{sec:sw}




\begin{figure*}[t]
    \centering
    \hspace*{\fill}
    \subfloat[Samples and directions]{\label{fig:samples_illustration_sw}\includegraphics[width={0.4\linewidth}]{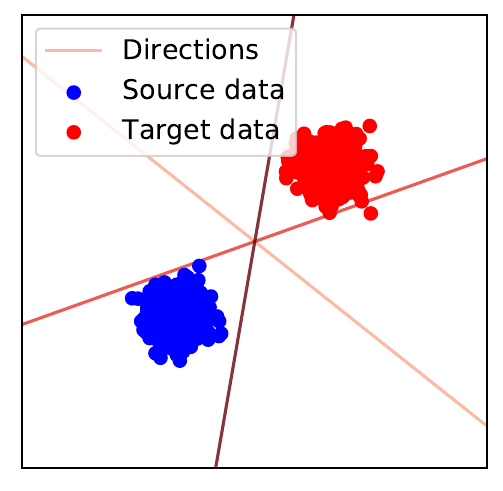}} \hfill
    \subfloat[One dimensional densities]{\label{fig:density}\includegraphics[width={0.4\linewidth}]{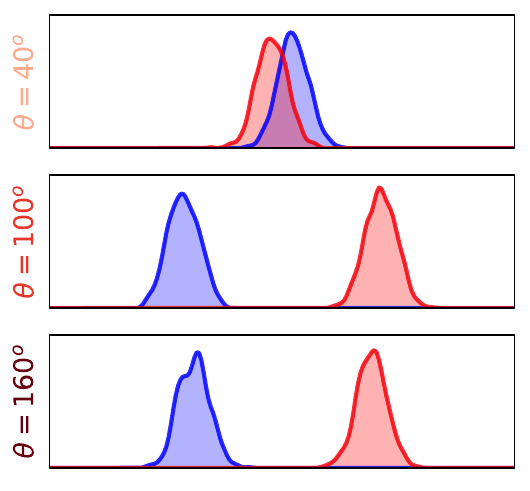}} \hfill
    \hspace*{\fill}
    \caption{Illustration of the projection of distributions on different lines.}
    \label{fig:illustration_sw}
\end{figure*}

Another alternative to the original Wasserstein problem is to consider proxy distances which have similar behaviors while being efficient to compute and having better scalability with respect to the number of samples and with the dimension. We introduce here the Sliced-Wasserstein distance and discuss some of its properties and variants.

\subsection{Definition and Computation}

\paragraph{Definition.} The Sliced-Wasserstein distance, first introduced in \citep{rabin2012wasserstein} to approximate barycenters, and then studied in \citep{bonnotte2013unidimensional, bonneel2015sliced}, leverages the one dimensional formulation of the Wasserstein distance \eqref{eq:wasserstein_1d} by computing the average of the Wasserstein distance between measures projected in one dimensional spaces in all possible directions. We illustrate the projection process of 2D densities in \Cref{fig:illustration_sw}.

\begin{definition}[Sliced-Wasserstein]
    Let $p\ge 1$, $\mu,\nu\in\mathcal{P}_p(\mathbb{R}^d)$. Then, the Sliced-Wasserstein distance is defined as
    \begin{equation} \label{eq:sw}
        \sw_p^p(\mu,\nu) = \int_{S^{d-1}} W_p^p(P^\theta_\#\mu, P^\theta_\#\nu)\ \mathrm{d}\lambda(\theta),
    \end{equation}
    where $\lambda$ is the uniform measure on the hypersphere $S^{d-1} = \{x\in\mathbb{R}^d,\ \|x\|_2^2=1\}$ and $P^\theta:x\mapsto \langle x,\theta\rangle$ is the coordinate projection on the line $\mathrm{span}(\theta)$.
\end{definition}

\begin{figure*}[t]
    \centering
    \hspace*{\fill}
    \subfloat[SW]{\label{fig:proj_sw}\includegraphics[width={0.4\linewidth}]{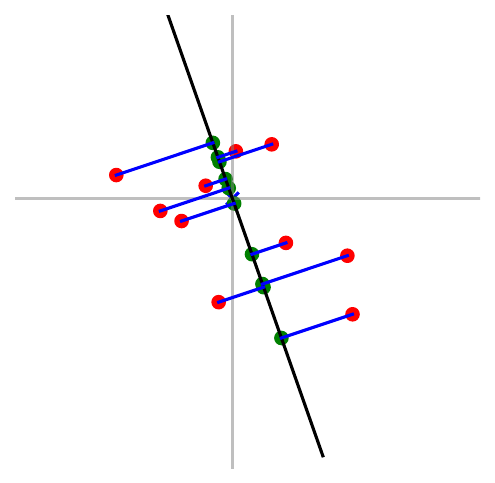}} \hfill
    \subfloat[Circular GSW]{\label{fig:proj_cgsw}\includegraphics[width={0.4\linewidth}]{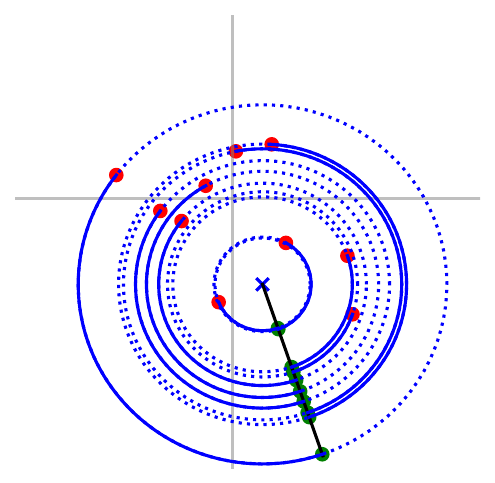}} \hfill
    \hspace*{\fill}
    \caption{Projection of (red) points onto the (black) line. The projected points are in green. The level sets along which the points are projected are plotted in blue.}
    \label{fig:projections_sw_vs_cgsw}
    \vspace{-10pt}
\end{figure*}

\paragraph{Computation.} In practice, when approximating the distributions $\mu$ and $\nu$ by their counterpart empirical distributions $\hat{\mu}_n=\frac{1}{n}\sum_{i=1}^n \delta_{x_i}$ and $\hat{\nu}_m = \frac{1}{m} \sum_{j=1}^m \delta_{y_i}$, the projected distributions are $P^\theta_\#\hat{\mu}_n = \frac{1}{n} \sum_{i=1}^n \delta_{\langle x_i,\theta\rangle}$ and $P^\theta_\#\hat{\nu}_m = \frac{1}{m} \sum_{j=1}^m \delta_{\langle y_j, \theta\rangle}$ for any $\theta\in S^{d-1}$. Hence, it amounts at projecting each point with the orthogonal projection on the line $\mathrm{span}(\theta)$ and getting the corresponding coordinate. We illustrate this in \Cref{fig:proj_sw}. Then, we need to compute the one dimensional Wasserstein distance between $P^\theta_\#\hat{\mu}_n$ and $P^\theta_\#\hat{\nu}_m$, as 
\begin{equation}
    W_p^p(P^\theta_\#\hat{\mu}_n, P^\theta_\#\hat{\nu}_m) = \int_0^1 | F_{P^\theta_\#\hat{\mu}_n}^{-1}(u)- F_{P^\theta_\#\hat{\nu}_m}^{-1}(u)|^p\ \mathrm{d}u.
\end{equation}
This integral can be easily approximated using \emph{e.g.} a rectangle method or a Monte-Carlo approximation. Note that in the particular case of $n=m$ with uniform weights,
\begin{equation}
    W_p^p(P^\theta_\#\hat{\mu}_n, P^\theta_\#\hat{\nu}_n) = \frac{1}{n} \sum_{i=1}^n \big|\langle \theta, x_{\sigma_\theta(i)} - y_{\tau_\theta(i)}\rangle\big|^p,
\end{equation}
where $\sigma_\theta$ (respectively $\tau_\theta$) is the permutation sorting $\big(\langle \theta, x_i\rangle\big)_i$ (respectively $\big(\langle \theta, y_i\rangle\big)_i$), \emph{i.e.} $\langle \theta, x_{\sigma_\theta(1)}\rangle \le \dots \le \langle \theta, x_{\sigma_\theta(n)}\rangle$ (respectively $\langle \theta, y_{\tau_\theta(1)}\rangle \le \dots \le \langle \theta, y_{\tau_\theta(n)}\rangle$). Thus, we only need to sort the projections of each measure in order to get the order statistics and to compute SW.

\looseness=-1 To approximate the outer integral with respect to $\lambda$, we use a Monte-Carlo approximation by first sampling $L$ directions $\theta_1,\dots,\theta_L\sim \lambda$, which can be done using the stochastic representation of $\lambda=\mathrm{Unif}(S^{d-1})$ \citep{fang2018symmetric} and amounts at first sampling $Z_\ell \sim \mathcal{N}(0, I_d)$ and then defining $\theta_\ell = Z/\|Z\|_2 \sim \lambda$ for $\ell\in\{1,\dots,L\}$. Finally, the Sliced-Wasserstein distance between $\mu$ and $\nu$ is approximated by
\begin{equation} \label{eq:approx_sw}
    \widehat{\sw}_p^p(\hat{\mu}_n, \hat{\nu}_m) = \frac{1}{L} \sum_{\ell=1}^L W_p^p(P^{\theta_\ell}_\#\hat{\mu}_n, P^{\theta_\ell}_\#\hat{\nu}_m).
\end{equation}
We sum up the procedure in \Cref{alg:sw}. The overall complexity is in $O\big(Ln(d+\log n)\big)$ ($Lnd$ operations for the projections and $Ln\log n$ for the sorting operations).

\begin{algorithm}[tb]
   \caption{Computing $\sw$}
   \label{alg:sw}
    \begin{algorithmic}
       \STATE {\bfseries Input:} $(x_i)_{i=1}^n\sim \mu$, $(y_j)_{j=1}^n\sim \nu$, $(\alpha_i)_{i=1}^n$, $(\beta_j)_{j=1}^n\in \Delta_n$, $L$ the number of projections, $p$ the order
       \FOR{$\ell=1$ {\bfseries to} $L$}
       \STATE Draw $\theta\in S^{d-1}$
       \STATE $\forall i,j,\ \hat{x}_i^{\ell}=\langle \theta, x_i\rangle$, $\hat{y}_j^\ell=\langle \theta, y_j\rangle$
       \STATE Compute $W_p^p(\sum_{i=1}^n \alpha_i \delta_{\hat{x}_i^\ell}, \sum_{j=1}^n \beta_j \delta_{\hat{y}_j^\ell})$
       \ENDFOR
       \STATE Return $\frac{1}{L}\sum_{\ell=1}^L W_p^p(\sum_{i=1}^n \alpha_i \delta_{\hat{x}_i^\ell}, \sum_{j=1}^n \beta_j \delta_{\hat{y}_j^\ell})$
    \end{algorithmic}
\end{algorithm}

\paragraph{Differentiability.} 

Independently from the low computational complexity, it is also 
differentiable with respect to the position of the particles which justifies its use in many learning tasks, and to perform gradient descent over particles. This property relies on the fact that the 1D Wasserstein distance is differentiable almost everywhere \citep[Section 3.2.4]{feydy2020geometric}, which is justified as the sort operation is differentiable almost everywhere \citep{blondel2020fast}, and in particular well differentiable when all the values are different.

\subsection{Properties}

In this Section, we discuss and sum up important properties of the Sliced-Wasserstein distance, which motivate its use as a proxy of the Wasserstein distance.

\paragraph{Distance.} First, the SW distance is indeed a distance, which justifies its use to compare probability distributions regardless of its connections with the Wasserstein distance.

\begin{proposition}[Distance]
    Let $p\ge 1$, $\sw_p$ is a finite distance on $\mathcal{P}_p(\mathbb{R}^d)$.
\end{proposition}

\begin{proof}
    See \citep[Proposition 5.1.2]{bonnotte2013unidimensional}.
\end{proof}

The pseudo-distance properties (symmetry and triangular inequality) rely on the slicing process. For the indiscernible property, it is possible to use the injectivity of the Fourier transform $\mathcal{F}$ to demonstrate it as, for $\mu,\nu\in\mathcal{P}_p(\mathbb{R}^d)$, $\sw_p(\mu,\nu)=0$ implies that for $\lambda$-almost every $\theta\in S^{d-1}$, $P^\theta_\#\mu=P^\theta_\#\nu$. Then, using that for all $s\in\mathbb{R}$, $\mathcal{F}(P^\theta_\#\mu)(s) = \mathcal{F}\mu(s\theta)$, we get the result. Another way of seeing it is to link $\sw$ with the Radon transform \citep{bonneel2015sliced}. This transform was introduced by \citet{Radon17} and has been very popular \emph{e.g.} in tomography \citep{helgason2011integral}.

\begin{definition}[Radon transform] \label{def:radon} \leavevmode
    \begin{enumerate}
        \item The Radon transform operator $R:L^1(\mathbb{R}^d)\to L^1(\mathbb{R}\times S^{d-1})$ is defined as, for all $f\in L^1(\mathbb{R}^d)$,
        \begin{equation}
            \forall t\in\mathbb{R}, \theta\in S^{d-1},\ Rf(t,\theta) = \int f(x) \mathbb{1}_{\{\langle x,\theta\rangle = t\}}\ \mathrm{d}x.
        \end{equation}
        \item The back-projection operator (dual transform) $R^*:C_0(\mathbb{R}\times S^{d-1})\to C_0(\mathbb{R}^d)$, where $C_0$ denotes the set of continuous functions that vanish at infinity, is defined as, for all $g\in C_0(\mathbb{R}\times S^{d-1})$, 
        \begin{equation}
            \forall x\in \mathbb{R}^d,\ R^*g(x) = \int_{S^{d-1}} g(\langle x,\theta\rangle, \theta)\ \mathrm{d}\lambda(\theta).
        \end{equation}
        \item The Radon transform on the set of measures $R:\mathcal{M}(\mathbb{R}^d)\to \mathcal{M}(\mathbb{R}\times S^{d-1})$ is defined, for $\mu\in\mathcal{M}(\mathbb{R}^d)$, as the measure $R\mu$ which satisfies, for all $g\in C_0(\mathbb{R}\times S^{d-1})$,
        \begin{equation}
            \int_{\mathbb{R}\times S^{d-1}} g(t,\theta) \ \mathrm{d}(R\mu)(t,\theta) = \int_{\mathbb{R}^d} R^*g(x)\ \mathrm{d}\mu(x).
        \end{equation}
    \end{enumerate}
\end{definition}

As the Radon transform of a measure is a measure on $\mathbb{R}\times S^{d-1}$, we can use the disintegration (\Cref{def:disintegration}) with respect to $\lambda$. Thus, we have $R\mu = \lambda \otimes K$, where $K$ is a probability kernel on $S^{d-1}\times \mathcal{B}(\mathbb{R})$. This kernel is actually exactly the orthogonal projection of $\mu$, \emph{i.e.} for $\lambda$-almost every $\theta\in S^{d-1}$, $K(\theta,\cdot) = P^\theta_\#\mu$ \citep[Proposition 6]{bonneel2015sliced}. Thus, the $\sw$ distance can be written using the Radon transform. For clarity, we will write $K(\theta,\cdot) = (R\mu)^\theta$.

\begin{proposition}[Relation with Radon transform] \label{prop:sw_radon}
    Let $p\ge 1$. For any $\mu,\nu\in \mathcal{P}_p(\mathbb{R}^d)$,
    \begin{equation}
        \sw_p^p(\mu,\nu) = \int_{S^{d-1}} W_p^p\big((R\mu)^\theta, (R\nu)^\theta\big)\ \mathrm{d}\lambda(\theta).
    \end{equation}
\end{proposition}

Using the injectivity of the Radon transform on the set of measures (see \emph{e.g.} \citep[Theorem A]{boman2009support}), we can also conclude that SW is a distance.

\paragraph{Topological Properties.}

Besides being a distance, we can also link its topological properties with the ones of the Wasserstein distance. This motivates further its use as a proxy since it has a relatively similar behavior. First, \citet{bonnotte2013unidimensional} showed that the two distances are actually weakly equivalent on distributions supported on compact sets.

\begin{proposition}[Equivalence with Wasserstein]
    Let $p\ge 1$ and denote $B(0,r) = \{x\in\mathbb{R}^d,\ \|x\|_2<r\}$ the open ball centered in 0 and of radius $r>0$. Then, for $\mu,\nu\in\mathcal{P}_p\big(B(0,r)\big)$, there exist constants $0 < c_{d,p}\le 1$ and $C_{d,p} > 0$ such that
    \begin{equation}
        \sw_p^p(\mu,\nu) \le c_{d,p}^p W_p^p(\mu,\nu) \le C_{d,p}^p r^{p-1/(d+1)} \sw_p(\mu,\nu)^{1/(d+1)},
    \end{equation}
    with $c_{d,p}^p = \frac{1}{d} \int_{S^{d-1}} \|\theta\|_p^p\ \mathrm{d}\lambda(\theta)$.
\end{proposition}

\begin{proof}
    See \citep[Theorem 5.1.5]{bonnotte2013unidimensional}.
\end{proof}

\citet[Theorem 2.1]{bayraktar2021strong} showed that for $d\ge 2$ and $p=1$, $\sw_1$ and the Wasserstein distance are not strongly equivalent, \emph{i.e.} we cannot find a constant $c$ for which the Wasserstein distance is upperbounded by $c\cdot \sw_1$.

The fact that the Wasserstein distance metrizes the weak convergence is well-known (see \emph{e.g.} \citep[Theorem 6.8]{villani2009optimal}). This last proposition shows that $\sw$ also metrizes the weak convergence for compactly supported measures. \citet{nadjahi2019asymptotic} showed that it holds on the general domain. Hence both metrics are topologically equivalent. We recall that a sequence of probability measures $(\mu_k)_{k\in\mathbb{N}}$ converges weakly to $\mu$ if, for any continuous and bounded function $f$, 
\begin{equation}
    \lim_{k\to\infty} \int f\ \mathrm{d}\mu_k = \int f \ \mathrm{d}\mu.
\end{equation}
In this case, we note $\mu_k\xrightarrow[k\to\infty]{\mathcal{L}} \mu$.

\begin{proposition}[Weak convergence]
    Let $p\ge 1$, and $(\mu_k)_{k\in\mathbb{N}}$ be a sequence of probability measures in $\mathcal{P}_p(\mathbb{R}^d)$. Then, $\mu_k$ converges weakly to $\mu$ if and only if $\lim_{k\to\infty}\sw_p(\mu_k,\mu) = 0$.
\end{proposition}

\begin{proof}
    See \citep[Theorem 1]{nadjahi2019asymptotic}.
\end{proof}

\paragraph{Statistical Properties.}

\looseness=-1 Besides being more computationally efficient than the Wasserstein distance, the Sliced-Wasserstein distance also happens to have a better behavior in high dimensional settings when approximated with the plug-in estimator. This has been studied in \citep{nadjahi2020statistical}, in which the sample complexity has been investigated by providing the convergence rate of $\sw_p(\hat{\mu}_n,\hat{\nu}_n)$ towards $\sw_p(\mu,\nu)$. \citet{nadjahi2020statistical} showed that thanks to the slicing process and contrary to the Wasserstein distance, the sample complexity is independent of the dimension. Thus, to have the same approximation, we do not need more samples in higher dimensions. This is a major property which also motivates to use the Sliced-Wasserstein distance for generative modeling, where the data can typically be of very high dimension and where, because of the limited memory of GPUs, small batches need to be used.

\begin{proposition}[Sample complexity]
    Let $p\ge 1$, $q>p$, $\mu,\nu\in\mathcal{P}_p(\mathbb{R}^d)$. Let $x_1,\dots,x_n\sim \mu$ and $y_1,\dots,y_n\sim \nu$, and denote $\hat{\mu}_n = \frac{1}{n} \sum_{i=1}^n \delta_{x_i}$, $\hat{\nu}_n = \frac{1}{n}\sum_{i=1}^n\delta_{y_i}$. Let $M_q(\mu)=\int \|x\|_2^q \ \mathrm{d}\mu(x)$ the moments of order $q$. Then, there exists a constant $C_{p,q}$ depending only on $p$ and $q$ such that
    \begin{equation}
        \mathbb{E}\big[|\sw_p(\hat{\mu}_n,\hat{\nu}_n) - \sw_p(\mu,\nu)|\big] \le C_{p,q}^{1/p} \big(M_q^{1/q}(\mu) + M_q^{1/q}(\nu)\big) \left\{ 
        \begin{array}{ll}
          n^{-1/(2p)} & \mbox{ if } q > 2p, \\
          n^{-1/(2p)} \log(n)^{1/p} & \mbox{ if } q = 2p, \\
          n^{-(q-p)/(pq)} & \mbox{ if } q \in (p,2p).
        \end{array}
        \right.
    \end{equation}
\end{proposition}

\begin{proof}
    See \citep[Corollary 2]{nadjahi2020statistical}.
\end{proof}

However, there is a second approximation done in practice as the integral \emph{w.r.t} the uniform distribution on $S^{d-1}$ is intractable. Thus, we also perform a Monte-Carlo approximation to approximate this integral and use \eqref{eq:approx_sw}. \citet{nadjahi2020statistical} provided a bound to quantify this error which depends on the number of projections used for the Monte-Carlo approximation as well as the variance, which depends implicitly on the dimension. This can hinder the approximation in high dimensional settings. 

\begin{proposition}[Projection complexity]
    Let $p\ge 1$, $\mu,\nu\in\mathcal{P}_p(\mathbb{R}^d)$. Then, 
    \begin{equation}
        \mathbb{E}_\theta\left[|\widehat{\sw}_{p,L}^p(\mu,\nu) - \sw_p^p(\mu,\nu)|\right]^2 \le \frac{1}{L} \mathrm{Var}_\theta\big[W_p^p(P^\theta_\#\mu, P^\theta_\#\nu)\big].
    \end{equation}
\end{proposition}

\begin{proof}
    See \citep[Theorem 6]{nadjahi2020statistical}.
\end{proof}

We also mention \citep[Proposition 5]{nietert2022statistical} which provided an explicit convergence rate by bounding the variance in terms of the parameters of the problem.

\citet[Proposition 4]{xu2022central} further showed a concentration result allowing to quantify the number of projections needed to have a small enough Monte-Carlo error.

\begin{proposition}
    Let $p\ge 1$, $\epsilon>0, \delta>0$ and $\mu,\nu\in\mathcal{P}_p(\mathbb{R}^d)$. When the number of projections $L$ satisfies $L\ge \frac{2K^2}{(d-1)\epsilon^2} \log(2/\delta)$ with $K = p W_p^{p-1}(\mu,\nu) \big(M_p(\mu) + M_p(\nu)\big)$ with $M_p$ the moments of order $p$, then
    \begin{equation}
        \mathbb{P}\big(|\widehat{\sw}_{p,L}^p(\mu,\nu) - \sw_p^p(\mu,\nu)|\ge \epsilon\big) \le \delta.
    \end{equation}
\end{proposition}

\begin{proof}
    See \citep[Proposition 4]{xu2022central}.
\end{proof}

\citet{manole2022minimax} also derived confidence intervals while \citet{goldfeld2022statistical, xu2022central, xi2022distributional} derived central limit theorems for $\sw$. For generative model tasks, \citet{nadjahi2019asymptotic} provided asymptotic guarantees for using $\sw$. For these types of problems, it was noted that using a small amount of projections was enough, which might be connected to the stochastic approximations process \citep{delyon2000stochastic}. More recently, \citet{tanguy2023properties, tanguy2023reconstructing, tanguy2023convergence} analyzed in more depth properties of the empirical Sliced-Wasserstein distance between discrete measures and studied the convergence of stochastic gradient descent with $\sw$ as objective.

\paragraph{Geodesics in Sliced-Wasserstein Space.} \label{paragraph:geodesics_sw}

\looseness=-1 It is well-known that the Wasserstein space is a geodesic space \citep{otto2001geometry}. Thus, a natural question is whether or not we have similar properties when endowing the space of probability measures with SW. This was studied by \citet{candau_tilh}, who showed that, surprisingly, it is not a geodesic space, but rather a pseudo-geodesic space whose geodesics are related to the Wasserstein distance.

We recall here first some notions in metric spaces. Let $(X,d)$ be some metric space. In our case, we will have $X=\mathcal{P}_2(\Omega)$ with $\Omega$ a bounded, open convex set and $d=\sw_2$. We first need to define an absolutely continuous curve.

\begin{definition}[Absolutely continuous curve]
    A curve $w:[0,1]\to X$ is said to be absolutely continuous if there exists $g\in L^1([0,1])$ such that
    \begin{equation}
        \forall t_0<t_1,\ d\big(w(t_0),w(t_1)\big) \le \int_{t_0}^{t_1}g(s)\mathrm{d}s.
    \end{equation}
\end{definition}
We denote by $AC(X, d)$ the set of absolutely continuous measures and by $AC_{x,y}(X,d)$ the set of curves in $AC(X,d)$ starting at $x$ and ending at $y$. Then, we can define the length of an absolutely continuous curve $w\in AC(X,d)$ as
\begin{equation}
    L_d(w) = \sup\left\{\sum_{k=0}^{n-1} d\big(w(t_k),w(t_{k+1})\big),\ n\ge 1,\ 0=t_0<t_1<\dots < t_n = 1\right\}.
\end{equation}
Then, we say that a space $X$ is a geodesic space if for any $x,y\in X$,
\begin{equation}
    d(x,y)=\min\left\{L_d(w),\ w\in AC(X,d), w(0)=x, w(1)=y\right\}.
\end{equation}

\citet{candau_tilh} showed in Theorem 2.4 that $(\mathcal{P}_2(\Omega),\sw_2)$ is not a geodesic space but rather a pseudo-geodesic space since for $\mu,\nu\in\mathcal{P}_2(\Omega)$,
\begin{equation}
    \inf\big\{L_{\sw_2}(w),\ w\in AC_{\mu,\nu}(\mathcal{P}_2(\Omega),\sw_2)\big\} = c_{d,2} W_2(\mu,\nu).
\end{equation}

We see that the infimum of the length in the SW space is the Wasserstein distance. Hence, it suggests that the geodesics in SW space are related to the ones in Wasserstein space, which are well-known since they correspond to the McCann interpolation (see \emph{e.g.} \citep[Theorem 5.27]{santambrogio2015optimal}).

\subsection{Variants} \label{section:variants_sw}

While the Sliced-Wasserstein distance is an appealing proxy of the Wasserstein distance which can scale to large problems and has many nice properties, it suffers from some drawbacks. Hence, a whole line of works consists of developing variants of the Sliced-Wasserstein distance. We provide a (non exhaustive) introduction to some of these variants.

\paragraph{With different slicing distributions.} As the SW distance integrates over all possible directions, it also takes into account directions which are not relevant to discriminate the two distributions (see for example the direction $\theta=40$° in \Cref{fig:density}). This point is exacerbated in practice as we use a Monte-Carlo approximation to approximate the integral. Hence, many directions, for which the Wasserstein distance between the projected distributions is almost null, are actually irrelevant \citep{deshpande2019max}. 
A solution to this issue is to use a slicing distribution which will mainly draw relevant directions where the two distributions can be well discriminated. \citet{deshpande2019max} first proposed to only sample the direction which is the most discriminative, which motivated the max-SW distance
\begin{equation}
    \maxsw_p^p(\mu,\nu) = \max_{\theta\in S^{d-1}}\ W_p^p(P^\theta_\#\mu,P^\theta_\#\nu).
\end{equation}
This comes back at choosing for slicing distribution $\sigma = \delta_{\theta^*}$ where $\theta^* \in\argmax_{\theta\in S^{d-1}}\ W_p^p(P^\theta_\#\mu, P^\theta_\#\nu)$. However, choosing only the most important direction can miss some potentially relevant directions. Thus, \citet{dai2021sliced} proposed to sample the $K$ most informative directions as 
\begin{equation}
    \maxksw_p^p(\mu,\nu) = \max_{\theta_1,\dots,\theta_K\ \text{orthonormal}}\ \frac{1}{K} \sum_{k=1}^K W_p^p(P^{\theta_k}_\#\mu, P^{\theta_k}_\#\nu),
\end{equation}
while \citet{nguyen2023markovian} proposed to sample $\theta_1,\dots,\theta_K$ as samples from a Markov chain defined on $S^{d-1}$ with well chosen Markov kernel to specify the transitions. \citet{nguyen2020distributional} proposed instead to learn a distribution on $S^{d-1}$ (parameterized in practice with a neural network) and defined the Distributional SW distance as
\begin{equation}
    \dsw_p^p(\mu,\nu) = \sup_{\sigma\in \mathbb{M}_C}\ \int_{S^{d-1}} W_p^p(P^\theta_\#\mu, P^\theta_\#\nu)\ \mathrm{d}\sigma(\theta),
\end{equation}
where $\mathbb{M}_C = \{\sigma\in\mathcal{P}(S^{d-1}),\ \mathbb{E}_{\theta,\theta'\sim\sigma}[|\langle \theta, \theta'\rangle|] \le C\}$ for $C\ge 0$. As the distribution is approximated by a neural network, this is a parametric model. Some other parametric model specified the distribution such as in \citep{nguyen2020improving} where it is chosen as a von Mises-Fisher distribution or a mixture of von Mises-Fisher distributions. \citet{ohana2022shedding} proposed to find the best distribution among von Mises-Fisher distributions by optimizing a PAC-Bayes bound. More recently, \citet{nguyen2023energy} proposed a parameter-free slicing distribution by choosing an energy-based slicing distribution $\sigma_{\mu,\nu}(\theta,f) \propto f\big(W_p^p(P^\theta_\#\mu,P^\theta_\#\nu)\big)$ with $f$ a monotonically increasing function. \citet{ohana2022shedding} named these methods ``Adaptative Sliced-Wasserstein distances''. In order to alleviate the computational cost required by solving a min-max problem when using these losses into generative models, \citet{nguyen2022amortized} further proposed to use amortized optimization.


\paragraph{With different projections.} Another bottleneck of the SW distance is that it uses linear projections, which can provide a low projection efficiency, especially in high dimensional settings where the data \mbox{often} lie on manifolds. To reduce the number of projections needed, nonlinear projections were proposed by \citet{kolouri2019generalized}. Using the relation with the Radon transform (see \Cref{prop:sw_radon}), they proposed to replace the Radon transform with generalized Radon transforms \citep{ehrenpreis2003universality, homan2017injectivity}, which integrate along hypersurfaces instead of hyperplanes. Formally, generalized Radon transforms are defined for $f\in L^1(\mathbb{R}^d)$ as 
\begin{equation}
    \forall t\in \mathbb{R}, \theta\in S^{d-1},\ Gf(t,\theta) = \int f(x) \mathbb{1}_{\{g(x,\theta)=t\}}\ \mathrm{d}x,
\end{equation}
where $g:X\times (\mathbb{R}^d\setminus\{0\})\to\mathbb{R}$, with $X\subset \mathbb{R}^d$, is a defining function  which satisfies the following properties: (i) $g$ is $C^\infty$ and (ii) 1-homogeneous in $\theta$, \emph{i.e.} $g(x,\lambda\theta)=\lambda g(x,\theta)$ for all $\lambda\in\mathbb{R}$, (iii) $\frac{\partial g}{\partial x}(x,\theta)\neq 0$ and (iv) $\det\big((\frac{\partial^2 g}{\partial x_i \partial \theta_j})_{ij}\big)>0$. It includes the Radon transform for $g(x,\theta) = \langle x,\theta\rangle$ and \citet{kolouri2019generalized} proposed a polynomial variant with $g(x,\theta) = \sum_{|\alpha|=m} \theta_\alpha x^\alpha$ and a neural network version. Besides, while not in the framework of generalized Radon transforms as not homogeneous \emph{w.r.t} $\theta$, \citet{kolouri2019generalized} also proposed to use a circular projection with $g(x,\theta) = \|x - r\theta\|_2$ for $r>0$ (which we illustrate in \Cref{fig:proj_cgsw}). \citet{kolouri2019generalized} observed that the resulting $\sw$ discrepancy is a distance if and only if the corresponding Radon transform is injective. This is for example the case for the polynomial version \citep{rouviere2015} or for the circular Radon transform \citep{kuchment2006generalized}, but not necessarily with the neural network version. \citet{chen2020augmented} further observed that using an invertible neural network $f$ and projections of the form $g(x,\theta) = \langle \theta, f(x)\rangle$ allows to satisfy the distance property. Note that for projections of this form, we can see it as embedding the data in another space where they can be better discriminated, in a similar fashion as \emph{e.g.} kernel methods \citep{hofmann2008kernel}.

Changing the projections can also allow better handling of data structures. For instance, \citet{nguyen2022revisiting} introduced convolution projections on images to better capture the spatial structure of images compared to naively vectorizing them.

\paragraph{On different subspaces.} 

While the SW distance is computationally efficient as it leverages the 1D closed-form of the Wasserstein distance, one can wonder whether one could obtain better discriminative power by projecting on higher dimensional subspaces and hence extracting more geometric information \citep{lin2021projection}. This line of work was first introduced by \citet{paty2019subspace} with the Projection Robust Wasserstein (PRW) distance
\begin{equation}
    \prw_p^p(\mu,\nu) = \max_{E\in\mathcal{G}_{d,k}}\ W_p^p(P^E_\#\mu, P^E_\#\nu),
\end{equation}
where $\mathcal{G}_{d,k} = \{E\subset\mathbb{R}^d, \ \dim(E)=k\}$ is the Grassmannian and $P^E$ the orthogonal projection on $E\in\mathcal{G}_{d,k}$. This formulation can also alleviate the curse of dimensionality as it has a better sample complexity \citep{lin2021projection, niles2022estimation}. Riemannian optimization onto the Stiefel manifolds were proposed in \citep{lin2020projection, huang2021riemannian, jiang2022riemannian} to compute this problem more efficiently as it is more intricate to compute since it is a max-min problem. For $k=1$, it coincides with the $\maxsw$ distance. \citet{lin2021projection} also studied an integral version \emph{w.r.t} the uniform distribution on the Stiefel manifold analogue to the Sliced-Wasserstein distance.

\paragraph{To obtain better estimation.} 

As the SW distance is approximated using a Monte-Carlo approximation, it is possible to leverage the literature of Monte-Carlo to reduce the variance of the estimators. Hence, \citet{nguyen2023control, leluc2023speeding} used control variates to obtain a better estimation of the SW distance with less variance.

The Monte-Carlo approximation has an expected error bound in $O(L^{-d/2})$ \citep{portier2020lecture} and requires a sufficient number of projections to have a reasonably small error. Hence, in high dimensional settings, and typically when $n\ll d$ (\emph{e.g.} in Deep Learning settings when the limited memory of GPUs constrains the use of mini-batches), the main bottleneck of the computation of SW is the projection step which has a complexity in $O(Ldn)$ (as $\log n\ll d$). A solution was recently provided by \citet{nguyen2022hierarchical} by decomposing the projection process in a hierarchical way with fewer projections on the original space. Another solution in high dimensional settings is to approximate the measure by gaussians using the concentration of measures \citep{nadjahi2021fast}. This provides the following approximation of SW:
\begin{equation}
    \widehat{\sw}_2^2(\mu,\nu) = \frac{\big(m_2(\Bar{\mu})^{\frac12} - m_2(\Bar{\nu})^{\frac12}\big)^2 + \|m_\mu - m_\nu\|_2^2}{d},
\end{equation}
where $m_\mu = \int x\ \mathrm{d}\mu(x)$, $\Bar{\mu} = (T_{m_\mu})_\#\mu$ with $T_{m_\mu}: x \mapsto x-m_\mu$ is the centered distribution and $m_2(\mu) = \mathbb{E}_{X\sim \mu}[\|X\|_2^2]$. This type of results was also extended to some of the Generalized Sliced-Wasserstein distances in \citep{le2022fast}. This solution is particularly appealing as it removes the need to choose the number of projections. However, it is only a good approximation in very high dimensional scenarios.

\paragraph{Projected Wasserstein distance.} \looseness=-1 Finally, let us describe another alternative inspired from $\sw$. \citet{rowland2019orthogonal} introduced the projected Wasserstein distance (PWD), which leverages the one dimensional coupling obtained between the projected measures and plug it between the original points, \emph{i.e.} 
\begin{equation}
    \pwd_p^p(\hat{\mu}_n, \hat{\nu}_n) = \int_{S^{d-1}} \frac{1}{n}\sum_{i=1}^n \|x_{\sigma_\theta(i)}-y_{\tau_\theta(i)}\|_2^p \ \mathrm{d}\lambda(\theta),
\end{equation}
with $\sigma_\theta$ (respectively $\tau_\theta$) the permutation sorting the samples of $P^\theta_\#\hat{\mu}_n$ (respectively $P^\theta_\#\hat{\nu}_n$). As each coupling is not at all optimal, it is clear that it is an upper bound of the Wasserstein distance. Furthermore, some permutations can be highly irrelevant leading to an overestimation of the Wasserstein distance. Choosing only an optimal direction in the same spirit of max-SW has been studied in \citep{mahey2023fast}.

\paragraph{Hilbert Curves.}

We also mention the work of \citet{bernton2019approximate, li2022hilbert} in which distributions are projected on space filling curves such as Hilbert curves, such curves having the appealing property to be locally preserving and hence to better respect the distance between the original points once projected. By defining a cumulative distribution function and the related quantile function on the Hilbert curve, \citet{li2022hilbert} leverage this to obtain a coupling and then compute the distance between the distributions in the original space. Thus, it is another efficient to compute upper-bound of the Wasserstein distance. As it suffers also from the curse of dimensionality, the authors also proposed a sliced version to alleviate it.

\clearemptydoublepage
\cleartooddpage[\thispagestyle{empty}]
\part{Sliced-Wasserstein on Riemannian Manifolds} \label{part:sw_riemannian}

\clearemptydoublepage
\cleartooddpage[\thispagestyle{empty}]
\chapter{Sliced-Wasserstein Discrepancies on Cartan-Hadamard Manifolds} \label{chapter:sw_hadamard}

{
    \hypersetup{linkcolor=black} 
    \minitoc 
}

\looseness=-1 This chapter aims at providing a general recipe to construct intrinsic extensions of the Sliced-Wasserstein distance on Riemannian manifolds. While many Machine Learning methods were developed or transposed on Riemannian manifolds to tackle data with known non Euclidean geometry, Optimal Transport methods on such spaces have not received much  attention. The main OT tools on these spaces are the Wasserstein distance and its entropic regularization with geodesic ground cost, but with the same bottleneck as in the Euclidean space. Hence, it is of much interest to develop new OT distances on such spaces, which allow to alleviate the computational burden. This chapter introduces a general construction and will be followed by three chapters covering specific cases of Riemannian manifolds with Machine Learning applications. Namely, we will study the particular case of Hyperbolic spaces, of the space of Symmetric Positive Definite matrices and of the Sphere.


\section{Introduction}

Working directly on Riemannian manifolds has received a lot of attention in recent years. On the one hand, it is well-known that data have an underlying structure on a low dimensional manifold \citep{bengio2013representation}. However, it can be intricate to work directly on such manifolds. Therefore, most works only focus on the Euclidean space and do not take advantage of this representation. In some cases though, the data naturally lies on a manifold, or can be embedded on some known manifolds allowing one to take into account its intrinsic structure. In such cases, it has been shown to be beneficial to exploit such structures by working directly on the manifold. To name a few examples, directional or earth data - data for which only the direction provides information - naturally lie on the sphere \citep{mardia2000directional} and hence their structure can be exploited by using methods suited to the sphere. Another popular example is given by data having a known hierarchical structure. Then, such data benefit from being embedded into Hyperbolic spaces \citep{nickel2017poincare}.

Motivated by these examples, many works proposed new tools to handle data lying on Riemannian manifolds. To cite a few, \citet{fletcher2004principal, huckemann2006principal} developed PCA to perform dimension reduction on manifolds while \citet{le2019approximation} studied density approximation, \citet{feragen2015geodesic, jayasumana2015kernel,fang2021kernel} studied kernel methods and \citet{azangulov2022stationary, azangulov2023stationary} developed Gaussian processes on (homogeneous) manifolds. More recently, there has been many interests into developing new neural networks with architectures taking into account the geometry of the ambient manifold \citep{bronstein2017geometric} such as Residual Neural Networks \citep{katsmann2022riemannian}, discrete Normalizing Flows \citep{rezende2020normalizing, rezende2021implicit} or Continuous Normalizing Flows \citep{mathieu2020riemannian, lou2020neural, rozen2021moser, yataka2022grassmann}. In the generative model literature, we can also cite the recent \citep{chen2023riemannian} which extended the flow matching training of Continuous Normalizing Flows to Riemannian manifolds, or \citet{bortoli2022riemannian} who performed score based generative modeling and \citet{thornton2022riemannian} who studied Schrödinger bridges on manifolds.

To compare probability distributions or perform generative modeling tasks, one usually needs suitable discrepancies or distances. In Machine Learning, classical divergences used are for example the Kullback-Leibler divergence or the Maximum Mean Discrepancy. While these distances are well defined for distributions lying on Riemannian manifolds, taking an extra care for the choice of the kernel in MMD, see \emph{e.g.} \citep{feragen2015geodesic}, other possible distances which take more into account the geometry of the underlying space are Optimal Transport based distances. While the Wasserstein distance can be well defined on manifolds, and has been studied in many works theoretically, see \emph{e.g.} \citep{mccann2001polar, villani2009optimal}, it suffers from computational burden as in the Euclidean case (see \Cref{section:estimation_wasserstein}). While on Euclidean cases, the Sliced-Wasserstein distance is a tractable alternative allowing to work in large scale settings, extending this construction on manifolds has not yet received much attention. Hence, as underlined in the conclusion of the thesis of \citet{nadjahi2021sliced}, deriving new SW based distance on manifolds could be of much interest.

In this chapter, we start by providing some background on Riemannian manifolds. Then, we introduce different ways to construct intrinsically Sliced-Wasserstein discrepancies on geodesically complete Riemannian manifolds with non-positive curvatures. Then, we derive some theoretical properties common to any sliced discrepancy on these Riemannian manifolds.

\section{Background on Riemannian Manifolds}

In this Section, we introduce some backgrounds on Riemannian manifolds. We refer to \citep{gallot1990riemannian,lee2006riemannian,lee2012smooth} for more details.

\subsection{Riemannian Manifolds} \label{sec:bg_riemnnian_manifolds}

\paragraph{Definition.}

A Riemannian manifold $(\mathcal{M}, g)$ of dimension $d$ is a space that behaves locally as a linear space diffeomorphic to $\mathbb{R}^d$, called a tangent space. To any $x\in \mathcal{M}$, one can associate a tangent space $T_x\mathcal{M}$ endowed with a inner product $\langle\cdot,\cdot\rangle_x : T_x\mathcal{M}\times T_x\mathcal{M}\to \mathbb{R}$ which varies smoothly with $x$. This inner product is defined by the metric $g_x$ associated to the Riemannian manifold as $g_x(u,v) = \langle u,v\rangle_x$ for any $x\in \mathcal{M}$, $u,v\in T_x\mathcal{M}$. We note $G(x)$ the matrix representation of $g_x$ defined such that
\begin{equation}
    \forall u,v \in T_x\mathcal{M},\ \langle u,v\rangle_x = g_x(u,v) = u^T G(x) v.
\end{equation}
For some spaces, different metrics can give very different geometries. We call tangent bundle the disjoint union of all tangent spaces $T\mathcal{M} = \{(x,v),\ x\in\mathcal{M} \text{ and } v\in T_x\mathcal{M}\}$, and we call a vector field a map $V:\mathcal{M}\to T\mathcal{M}$ such that $V(x)\in T_x\mathcal{M}$ for all $x\in\mathcal{M}$.

\paragraph{Geodesics.}

A generalization of straight lines in Euclidean spaces to Riemannian manifolds can be geodesics, which are smooth curves connecting two points with the minimal length, \emph{i.e.} curves $\gamma:[0,1]\to \mathbb{R}$ which minimize the length $\mathcal{L}$ defined as
\begin{equation}
    \mathcal{L}(\gamma) = \int_0^1 \|\gamma'(t)\|_{\gamma(t)}\ \mathrm{d}t,
\end{equation}
where $\|\gamma'(t)\|_{\gamma(t)} = \sqrt{\langle \gamma'(t), \gamma'(t)\rangle_{\gamma(t)}}$. In this work, we will focus on geodesically complete Riemannian manifolds, in which case there is always a geodesic between two points $x,y\in\mathcal{M}$. Furthermore, all geodesics are actually geodesic lines, \emph{i.e.} can be extended to $\mathbb{R}$. Let $x,y\in \mathcal{M}$, $\gamma:[0,1]\to\mathbb{R}$ a geodesic between $x$ and $y$ such that $\gamma(0)=x$ and $\gamma(1)=y$, then the value of the length defines actually a distance $(x,y)\mapsto d(x,y)$ between $x$ and $y$, which we call the geodesic distance:
\begin{equation}
    d(x,y) = \inf_\gamma\ \mathcal{L}(\gamma). 
\end{equation}
Note that for a geodesic $\gamma$ between $x$ and $y$, we have for any $s,t\in [0,1]$, $d\big(\gamma(t), \gamma(s)\big) = |t-s| d(x,y)$. And it is true for $s,t\in\mathbb{R}$ for geodesic lines.

\paragraph{Exponential map.}

Let $x\in\mathcal{M}$, then for any $v\in T_x\mathcal{M}$, there exists a unique geodesic $\gamma_{(x,v)}$ starting from $x$ with velocity $v$, \emph{i.e.} such that $\gamma_{(x,v)}(0) = x$ and $\gamma_{(x,v)}'(0) = v$ \citep{sommer2020introduction}. Now, we can define the exponential map as $\exp:T\mathcal{M}\to \mathcal{M}$ which for any $x\in \mathcal{M}$, maps tangent vectors $v\in T_x\mathcal{M}$ back to the manifold at the point reached by the geodesic $\gamma_{(x,v)}$ at time 1:
\begin{equation}
    \forall (x,v)\in T\mathcal{M},\ \exp_x(v) = \gamma_{(x,v)}(1).
\end{equation}
On geodesically complete manifolds, the exponential map is defined on the entire tangent space, but is not necessarily a bijection. When it is one, we note $\log_x$ the inverse of $\exp_x$, which can allow to map elements from the manifold to the tangent space.


\paragraph{Sectional curvature.} A notion which allows studying the geometry as well as the topology of a given Riemannian manifold is the sectional curvature. Let $x\in\mathcal{M}$, and $u,v\in T_x\mathcal{M}$ two linearly independent vectors. Then, the sectional curvature $\kappa_x(u,v)$ is defined geometrically as the Gaussian curvature of the plane $E=\mathrm{span}(u,v)$ \citep{zhang2016riemannian}, \emph{i.e.}
\begin{equation}
    \kappa_x(u,v) = \frac{\langle R(u,v)u, v\rangle_x}{\langle u,u\rangle_x\langle v,v\rangle_x - \langle u,v\rangle_x^2},
\end{equation}
where $R$ is the Riemannian curvature tensor. We refer to \citep{lee2006riemannian} for more details. The behavior of geodesics changes given the curvature of the manifold. For instance, they usually diverge on manifolds of negative sectional curvature and converge on manifolds of positive sectional curvature \citep{hu2023riemannian}. Important examples of Riemannian manifolds are Euclidean spaces which are of constant null curvature, the sphere which is of positive constant curvature and Hyperbolic spaces which are of negative constant curvature (\emph{i.e.} have the same value at any point $x\in\mathcal{M}$ and for any 2-planes $E$). We can also cite the torus which have some points of positive curvature, some points of negative curvature and some points of null curvature \citep{borde2023latent}. In this chapter, we will mostly focus on Cartan-Hadamard manifolds which are complete connected Riemannian manifolds of non-positive sectional curvature.

\begin{figure*}[t]
    \centering
    \hspace*{\fill}
    \subfloat[Negatively curved]{\label{fig:triangle_neg}\includegraphics[width={0.2\linewidth}]{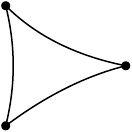}} \hfill
    \subfloat[No curvature]{\label{fig:triangle}\includegraphics[width={0.2\linewidth}]{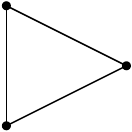}} \hfill
    \subfloat[Positively curved]{\label{fig:triangle_pos}\includegraphics[width={0.2\linewidth}]{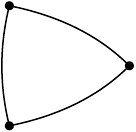}}
    \hspace*{\fill}
    \caption{Triangles in different curvatures. For negative curvatures ($k<0$), the sum of angles is lower than $\pi$, and for positive curvature ($k>0$), the sum of angles is greater than $\pi$.}
    \label{fig:triangles}
\end{figure*}

\paragraph{CAT($0$) space.} Let us also introduce the more general notion of CAT($0$) space \citep[Part II, Section 1.1]{bridson2013metric}. Let $(X,d)$ be a geodesic complete metric space. A geodesic triangle $\Delta(x,y,z)$ with vertices $x,y,z\in X$ is the union of three geodesic segments $[x,y]$, $[y,z]$ and $[z,x]$. Then, we call a comparison triangle $\Bar{\Delta}(\Bar{x}, \Bar{y}, \Bar{z})$ for $\Delta(x,y,z)$ a triangle in $\mathbb{R}^2$ such that $\Bar{x}, \Bar{y}, \Bar{z}\in\mathbb{R}^2$ and $d(x,y)=|\Bar{x}-\Bar{y}|$, $d(y,z)=|\Bar{y}-\Bar{z}|$ and $d(x,z)=|\Bar{x}-\Bar{z}|$. Similarly, $\Bar{w}\in[\Bar{x},\Bar{y}]$ is a comparison point for $w\in [x,y]$ if $d(x,w) = |\Bar{x}-\Bar{w}|$. Then, the geodesic metric space $(X,d)$ is a CAT(0) space if for every geodesic triangle $\Delta(x,y,z)$ and for any $p,q\in [x,y]$ and comparison points $\Bar{p}, \Bar{q}\in [\Bar{x},\Bar{y}]$, $d(p,q) \le |\Bar{p}-\Bar{q}|$. Note that we can extend the definition to CAT($k$) spaces for $k\in\mathbb{R}$ by changing $\mathbb{R}^2$ by the sphere $S^2$ for $k>0$ and the hyperbolic space $\mathbb{H}^2$ for $k<0$ (using the right geodesic distance instead of the absolute distance).

We illustrate the triangles for different values of $k$ in \Cref{fig:triangles}. This is actually a more general notion of curvature than the sectional curvature, see \citep[Part II, Appendix of Chapter 1]{bridson2013metric}. In particular, CAT(0) spaces are called Hadamard spaces and encompass for example Cartan-Hadamard manifolds.

\subsection{Optimization on Riemannian Manifolds}

We are often interested in solving optimization problems for variables which lie on manifolds. Common examples include Principal Component Analysis where we optimize over the Stiefel manifold, Procruste problems optimizing on rotations or maximum likelihood for densities such as Gaussians. In our context, we are often interested in learning distributions on some manifold. This can be done by either learning a set of particles directly lying on the manifold, or using neural networks well suited to the manifold, which often involve parameters also on the manifold \citep{ganea2018hyperbolic, shimizu2021hyperbolic,fei2023survey}. Hence, for many reasons, we need to be able to optimize directly over manifolds. We refer to the books of \citet{absil2009optimization} or \citet{boumal2023introduction} for more details.

Fortunately, similarly as in the Euclidean case, one can use first order optimization methods such as gradient descents. As the analog of straight lines are geodesics, we will follow the geodesic in the direction which minimizes the functional as fast as possible. Let $f:\mathcal{M}\to \mathbb{R}$ be a functional, which we suppose (geodesically) convex, \emph{i.e.} for any geodesic curve $\gamma$ linking $x\in\mathcal{M}$ to $y\in \mathcal{M}$, $f$ satisfies
\begin{equation}
    \forall t \in [0,1],\ f\big(\gamma(t)\big) \le (1-t) f(x) + t f(y).
\end{equation}
Furthermore, we will suppose that the functional is differentiable. Then, let us define the Riemannian gradient of $f$.
\begin{definition}[Gradient]
    We define the Riemannian gradient of $f$ as the unique vector field $\mathrm{grad}_\mathcal{M}f:\mathcal{M}\to T\mathcal{M}$ satisfying
    \begin{equation}
        \forall (x,v)\in T\mathcal{M},\ \frac{\mathrm{d}}{\mathrm{d}t}f\big(\exp_x(tv)\big)\Big|_{t=0} = \langle v, \mathrm{grad}_\mathcal{M}f(x)\rangle_x.
    \end{equation}
\end{definition}
As the gradient belongs to the tangent space, we can use the exponential map to project it back to the manifold. Therefore, the gradient descent algorithm reads as, starting from $x_0\in\mathcal{M}$ and with gradient step $\tau>0$,
\begin{equation}
    \forall k\ge 0,\ x_{k+1} = \exp_{x_k}\big(-\tau\ \mathrm{grad}_\mathcal{M}f(x_k)\big).
\end{equation}
Note that in the Euclidean case, since $\exp_{x}(y) = x+y$ and $\mathrm{grad}f(x)=\nabla f(x)$, it reads as $x_{k+1} = x_k - \tau \nabla f(x_k)$ which coincides well with the regular gradient descent algorithm. In some cases, the exponential map can be intractable or hard to compute. Then, it is possible to use instead a retraction, which is a smooth map $R:T\mathcal{M}\to\mathcal{M}$ such that each curve $c(t) = R_x(tv)$ satisfies $c(0)=x$ and $c'(0)=v$ \citep[Section 3.6]{boumal2023introduction}. 

Similar variants as in the Euclidean space can be derived and used. For instance, one can use the stochastic version \citep{bonnabel2013stochastic}, backward versions \citep{ferreira2002proximal, bento2017iteration}, Nesterov accelerated methods \citep{kim2022nesterov}, or adaptative moment methods such as Riemannian Adam \citep{becigneul2018riemannian}. A recent line of work also studies optimization algorithms which do not use any retractions as they can be computationally expensive \citep{ablin2022fast, gao2022optimization,ablin2023infeasible}.

\subsection{Probability Distributions on Riemannian Manifolds}

\paragraph{Probability distribution.}

Let $(\mathcal{M}, g)$ be a Riemannian manifold. For $x\in \mathcal{M}$, $G(x)$ induces an infinitesimal change of volume on the tangent space $T_x\mathcal{M}$, and thus a measure on the manifold,
\begin{equation}
    \mathrm{d}\vol(x) = \sqrt{|G(x)|}\ \mathrm{d}x.
\end{equation}
Here, we denote by $\mathrm{d}x$ the Lebesgue measure. We refer to \citep{pennec2006intrinsic} for more details on distributions on manifolds. Now, we discuss some possible distributions on Riemannian manifolds, which can be seen as generalizations of Gaussian distributions. 

The first way of naturally generalizing Gaussian distributions to Riemannian manifolds is to use the geodesic distance in the density, which becomes
\begin{equation}
    f(x) \propto \exp\left(-\frac{1}{2\sigma^2}d(x,\mu)^2\right),
\end{equation}
for $\mu\in \mathcal{M}$, $\sigma\in \mathbb{R}$. This was first introduced in \citep{pennec2006intrinsic} and then further considered and theoretically studied on particular Riemannian manifolds in \citep{said2017riemannian, said2017gaussian}. Notably, an important property required to use such a density is that the normalization factor must not depend on the mean parameter $\mu$, which might not always be the case. In particular, it holds on Riemannian symmetric spaces \citep{said2017riemannian}. However, it is not straightforward to sample from such a distribution.


More convenient distributions, on which we can use the reparameterization trick, are wrapped distributions \citep{chevallier2020wrapped,chevallier2022exponential,galaz2022wrapped}. The idea is to push-forward a distribution $\mu\in\mathcal{P}(T_x\mathcal{M})$ onto $\mathcal{P}(\mathcal{M})$. A natural function to use is the exponential map when it is invertible over the whole tangent space. This has received much attention \emph{e.g.} on hyperbolic spaces with the wrapped normal distribution \citep{nagano2019wrapped, cho2022rotated}, which samples from a Gaussian in the tangent space, as it gives a very convenient way to sample on the manifold, while all transformations are differentiable, and can hence be used in variational autoencoders for instance.

Another solution to sample on a manifold is to condition the samples to belong to the manifold. When restricting an isotropic distribution to lie on the unit sphere, this gives for example the well-known von Mises-Fisher distribution \citep{hauberg2018directional}.

\paragraph{Optimal Transport.}

Optimal Transport is also well defined on Riemannian manifolds using appropriate ground costs into the problem. Using the geodesic distance at the power $p\ge 1$, we recover the $p$-Wasserstein distance
\begin{equation}
    W_p^p(\mu,\nu) = \inf_{\gamma\in \Pi(\mu,\nu)}\ \int_{\mathcal{M}\times \mathcal{M}} d(x,y)^p\ \mathrm{d}\gamma(x,y),
\end{equation}
where $\mu,\nu\in\mathcal{P}_p(\mathcal{M}) = \{\mu\in \mathcal{P}(\mathcal{M}),\ \int_\mathcal{M} d(x,o)^p\ \mathrm{d}\mu(x)<\infty\}$, with $o\in \mathcal{M}$ some origin which can be arbitrarily chosen (because of the triangular inequality).

This problem has received much attention, see \emph{e.g.} \citep{villani2009optimal, bianchini2011optimal}. In particular, Brenier's theorem was extended by \citet{mccann2001polar} on Riemannian manifolds. For $\mu,\nu \in\mathcal{P}_2(\mathcal{M})$ when the source measure $\mu$ is absolutely continuous \emph{w.r.t} the volume measure on $\mathcal{M}$, then there exists a unique OT map $T$ such that $T_\#\mu=\nu$ and $T$ is given by, for $\mu$-almost every $x\in \mathcal{M}$, $T(x) = \exp_x\big(-\mathrm{grad}_\mathcal{M} \psi(x)\big)$ with $\psi$ a c-concave map.

\section{Intrinsic Riemannian Sliced-Wasserstein} \label{section:irsw}

\looseness=-1 In this Section, we propose natural generalizations of the Sliced-Wasserstein distance on probability distributions supported on Riemannian manifolds by using tools intrinsically defined on them. To do that, we will first consider the Euclidean space as a Riemannian manifold. Doing so, we will be able to generalize it naturally to other geodesically complete Riemannian manifolds. We will first focus on manifolds of non-positive curvatures. Then, we will discuss some challenges inherent to Riemannian manifolds with positive curvatures.

\subsection{Euclidean Sliced-Wasserstein as a Riemannian Sliced-Wasserstein Distance} \label{sec:sw_esw_rsw}

It is well known that the Euclidean space can be viewed as a Riemannian manifold of null constant curvature \citep{lee2006riemannian}. From that point of view, we can translate the elements used to build the Sliced-Wasserstein distance as Riemannian elements, and identify how to generalize it to more general Riemannian manifolds.

First, let us recall that the $p$-Sliced-Wasserstein distance for $p\ge 1$ between $\mu,\nu\in\mathcal{P}_p(\mathbb{R}^d)$ is defined as
\begin{equation}
    \sw_p^p(\mu,\nu) = \int_{S^{d-1}} W_p^p(P^\theta_\#\mu, P^\theta_\#\nu)\ \mathrm{d}\lambda(\theta),
\end{equation}
where $P^\theta(x) = \langle x, \theta\rangle$ and $\lambda$ is the uniform distribution $S^{d-1}$. Geometrically, we saw in \Cref{sec:sw} that it amounts to project the distributions on every possible line going through the origin $0$. Hence, we see that we need first to generalize lines passing through the origin, while being still able to compute the Wasserstein distance on these subsets. Furthermore, we also need to generalize the projection.

\paragraph{Lines.} From a Riemannian manifold point of view, straight lines can be seen as geodesics, which are, as we saw in \Cref{sec:bg_riemnnian_manifolds}, curves minimizing the distance between any two points on it. For any direction $\theta\in S^{d-1}$, the geodesic passing through $0$ in direction $\theta$ is described by the curve $\gamma_\theta:\mathbb{R} \to \mathbb{R}^d$ defined as $\gamma_\theta(t) = t \theta = \exp_0(t\theta)$ for any $t\in \mathbb{R}$, and the geodesic is $\mathcal{G}_\theta = \mathrm{span}(\theta)$. Hence, when it makes sense, a natural generalization to straight lines would be to project on geodesics passing through an origin.

\paragraph{Projections.} The projection $P^\theta(x)$ of $x\in\mathbb{R}^d$ can be seen as the coordinate of the orthogonal projection on the geodesic $\mathcal{G}_\theta$. Indeed, the orthogonal projection $\Tilde{P}$ is formally defined as 
\begin{equation}
    \Tilde{P}^\theta(x) = \argmin_{y \in \mathcal{G}_\theta}\ \|x-y\|_2 = \langle x, \theta\rangle\theta.
\end{equation}
From this formulation, we see that $\Tilde{P}^\theta$ is a metric projection, which can also be called a geodesic projection on Riemannian manifolds as the metric is a geodesic distance. 
Then, we see that its coordinate on $\mathcal{G}_\theta$ is $t=\langle x,\theta\rangle = P^\theta(x)$, which can be also obtained by first giving a direction to the geodesic, and then computing the distance between $\Tilde{P}^\theta(x)$ and the origin $0$, as
\begin{equation}
    P^\theta(x) = \sign(\langle x,\theta\rangle)\|\langle x,\theta\rangle \theta - 0 \|_2 = \langle x,\theta\rangle.
\end{equation}
Note that this can also be recovered by solving
\begin{equation}
    P^\theta(x) = \argmin_{t \in \mathbb{R}} \ \|\exp_0(t\theta) - x\|_2.
\end{equation}
This formulation will be useful to generalize it to more general manifolds by replacing the Euclidean distance by the right geodesic distance.

Note also that the geodesic projection can be seen as a projection along hyperplanes, \emph{i.e.} the level sets of the projection function $g(x,\theta) = \langle x, \theta\rangle$ are (affine) hyperplanes. This observation will come useful in generalizing SW to manifolds of non-positive curvature.

\paragraph{Wasserstein distance.} The Wasserstein distance between measures lying on the real line has a closed-form which can be computed very easily (see \Cref{section:closed_forms}). On more general Riemannian manifolds, as the geodesics will not necessarily be lines, we will need to check how to compute the Wasserstein distance between the projected measures.

\subsection{On Manifolds of Non-Positive Curvature}

In this part, we focus on complete connected Riemannian manifolds of non-positive curvature, which can also be called Hadamard manifolds or Cartan-Hadamard manifolds \citep{lee2006riemannian, robbin2011introduction, lang2012fundamentals}. These spaces actually include Euclidean spaces, but also spaces with constant negative curvature such as Hyperbolic spaces, or with variable non-positive curvatures such as the space of Symmetric Positive Definite matrices and product of manifolds with constant negative curvature \citep[Lemma 1]{gu2019learning}. We refer to \citep{ballmann2006manifolds} or \citep{bridson2013metric} for more details. These spaces share many properties with Euclidean spaces \citep{bertrand2012geometric} which make it possible to extend the Sliced-Wasserstein distance on them. We will denote $(\mathcal{M}, g)$ a Hadamard manifold in the following. The particular cases of Hyperbolic spaces and the spaces of Symmetric Positive Definite matrices will be further studied respectively in \Cref{chapter:hsw} and \Cref{chapter:spdsw}.

\paragraph{Properties of Hadamard Manifolds.} First, as a Hadamard manifold is a complete connected Riemannian manifold, then by the Hopf-Rinow theorem \citep[Theorem 6.13]{lee2006riemannian}, it is also geodesically complete. Therefore, any geodesic curve $\gamma:[0,1]\to\mathcal{M}$ connecting $x\in \mathcal{M}$ to $y\in \mathcal{M}$ can be extended on $\mathbb{R}$ as a geodesic line. Furthermore, by Cartan-Hadamard theorem \citep[Theorem 11.5]{lee2006riemannian}, Hadamard manifolds are diffeomorphic to the Euclidean space $\mathbb{R}^d$, and the exponential map at any $x\in\mathcal{M}$ from $T_x\mathcal{M}$ to $\mathcal{M}$ is bijective with the logarithm map as inverse. Moreover, their injectivity radius is infinite and thus, its geodesics are aperiodic, and can be mapped to the real line, which will allow to find coordinates on the real line, and hence to compute the Wasserstein distance between the projected measures efficiently. The SW discrepancy on such spaces is therefore very analogous to the Euclidean case. Note that Hadamard manifolds belong to the more general class of CAT(0) metric spaces, and hence inherit their properties described in \citep{bridson2013metric}. Now, let us discuss two different possible projections, which both generalize the Euclidean orthogonal projection.

\paragraph{Geodesic Projections.} As we saw in \Cref{sec:sw_esw_rsw}, a natural projection on geodesics is the geodesic projection. Let's note $\mathcal{G}$ a geodesic passing through an origin point $o\in \mathcal{M}$. Such origin will often be taken naturally on the space, and corresponds to the analog of the $0$ in $\mathbb{R}^d$. Then, the geodesic projection on $\mathcal{G}$ is obtained naturally as
\begin{equation}
    \forall x\in \mathcal{M},\ \Tilde{P}^\mathcal{G}(x) = \argmin_{y\in \mathcal{G}}\ d(x, y).
\end{equation}
From the projection, we can get a coordinate on the geodesic by first giving it a direction and then computing the distance to the origin. By noting $v\in T_o\mathcal{M}$ a vector in the tangent space at the origin, such that $\mathcal{G} = \mathcal{G}^v = \{\exp_o(tv),\ t\in\mathbb{R}\}$, we can give a direction to the geodesic by computing the sign of the inner product in the tangent space of $o$ between $v$ and the log of $\Tilde{P}^\mathcal{G}$. Analogously to the Euclidean space, we can restrict $v$ to be of unit norm, \emph{i.e.} $\|v\|_o=1$. Now, we will use $v$ in index of $\Tilde{P}$ and $P$ instead of $\mathcal{G}$. Hence, we obtain the coordinates using
\begin{equation}
    P^v(x) = \sign\big(\langle \log_o\big(\Tilde{P}^v(x)\big), v\big\rangle_o\big)\ d\big(\Tilde{P}^v(x), o\big).
\end{equation}
We show in the next Proposition that the map $t^v: \mathcal{G}^v \to \mathbb{R}$ defined as
\begin{equation} \label{eq:coordmap}
    \forall x\in \mathcal{G}^v,\ t^v(x) = \sign\big(\langle \log_o(x), v\rangle_o\big)\ d(x, o),
\end{equation}
is an isometry.
\begin{proposition} \label{prop:isometry}
    Let $(\mathcal{M}, g)$ be a Hadamard manifold with origin $o$. Let $v\in T_o\mathcal{M}$, then, the map $t^v$ defined in \eqref{eq:coordmap} is an isometry from $\mathcal{G}^v = \{\exp_o(tv),\ t\in\mathbb{R}\}$ to $\mathbb{R}$.
\end{proposition}
\begin{proof}
    See \Cref{proof:prop_isometry}.
\end{proof}
Note that to get directly the coordinate from $x\in \mathcal{M}$, we can also solve directly the following problem:
\begin{equation} \label{eq:geod_proj}
    P^v(x) = \argmin_{t\in \mathbb{R}}\ d\big(\exp_o(tv), x\big).
\end{equation}


Using that Hadamard manifolds belong to the more general class of CAT(0) metric spaces, by \citep[II. Proposition 2.2]{bridson2013metric}, the geodesic distance is geodesically convex. Hence, $t\mapsto d\big(\exp_o(tv), x\big)$ is a coercive convex problem, and hence admits a unique solution. Therefore, \eqref{eq:geod_proj} is well defined. Moreover, we have the following characterization for the optimum:
\begin{proposition} \label{prop:charac_geod_proj}
    Let $(\mathcal{M},g)$ be a Hadamard manifold with origin $o$. Let $v\in T_o\mathcal{M}$, and note $\gamma(t) = \exp_o(tv)$ for all $t\in\mathbb{R}$. Then, for any $x\in\mathcal{M}$,
    \begin{equation}
        P^v(x) = \argmin_{t\in\mathbb{R}}\ d(\gamma(t), x) \iff \left\langle \gamma'\big(P^v(x)\big), \log_{\gamma\big(P^v(x)\big)}(x)\right\rangle_{\gamma\big(P^v(x)\big)} = 0.
    \end{equation}
\end{proposition}

\begin{proof}
    See \Cref{proof:prop_charac_geod_proj}.
\end{proof}

In the Euclidean case $\mathbb{R}^d$, as geodesics are of the form $\gamma(t) = t\theta$ for any $t\in\mathbb{R}$ and for a direction $\theta\in S^{d-1}$, and as $\log_x(y) = y-x$ for $x,y\in\mathbb{R}^d$, we recover the projection formula:
\begin{equation}
    \left\langle \gamma'\big(P^\theta(x)\big), \log_{\gamma\big(P^\theta(x)\big)}(x)\right\rangle_{\gamma\big(P^\theta(x)\big)} = 0 \iff \langle \theta, x-P^\theta(x)\theta\rangle = 0 \iff P^\theta(x) = \langle \theta, x\rangle.
\end{equation}

\begin{figure}
    \centering
    \includegraphics[width=0.25\linewidth]{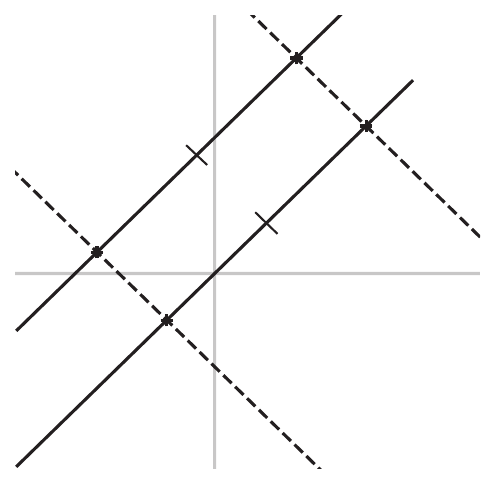}
    \caption{On Euclidean spaces, the distance between the projections of two points belonging to a geodesic with the same direction is conserved. This might not be the case on more general Riemannian manifolds.}
    \label{fig:parallel_proj}
\end{figure}

\paragraph{Busemann Projections.} \looseness=-1 The level sets of previous projections are geodesic subspaces. It has been shown that projecting along geodesics is not always the best solution as it might not preserve distances well between the original points \citep{chami2021horopca}. Indeed, on Euclidean spaces, as mentioned earlier, the projections are actually along hyperplanes, which tends to preserve the distance between points belonging to another geodesic with the same direction better (see \Cref{fig:parallel_proj}). On Hadamard manifolds, there are analogs of hyperplanes, which can be obtained through the level sets of the Busemann function which we introduce now.

Let $\gamma$ be a geodesic line, then the Busemann function associated to $\gamma$ is defined as \citep[II. Definition 8.17]{bridson2013metric}
\begin{equation}
    \forall x\in\mathcal{M},\ B^\gamma(x) = \lim_{t\to\infty}\ \left(d\big(x,\gamma(t)\big) - t\right).
\end{equation}
On Hadamard manifolds, and more generally on CAT(0) spaces with $\gamma$ a geodesic ray, the limit does exist \citep[II. Lemma 8.18]{bridson2013metric}. This function returns a coordinate on the geodesic $\gamma$, which can be understood as a normalized distance to infinity towards the direction given by $\gamma$ \citep{chami2021horopca}. The level sets of this function are called horospheres. On spaces of constant curvature (\emph{i.e.} Euclidean or Hyperbolic spaces), horospheres are of constant null curvature and hence very similar to hyperplanes. We illustrate horospheres in Hyperbolic spaces in \Cref{fig:projections_hsw}.

For example, in the Euclidean case, we can show that the Busemann function associated to $\mathcal{G}^\theta = \mathrm{span}(\theta)$ is given by
\begin{equation}
    \forall x\in \mathbb{R}^d,\ B^\theta(x) = -\langle x, \theta\rangle.
\end{equation}
It actually coincides with the inner product, which can be seen as a coordinate on the geodesic $\mathcal{G}^\theta$. Moreover, its level sets in this case are (affine) hyperplanes orthogonal to $\theta$.

Hence, the Busemann function gives a principled way to project measures on a Hadamard manifold to the real line provided that we can compute its closed-form. To find the projection on the geodesic $\gamma$, we can solve the equation in $s\in\mathbb{R}$, $B^\gamma(x) = B^\gamma\big(\gamma(s)\big) = -s$, and we find that the projection on $\gamma$ is $\Tilde{B}^\gamma(x)=\exp_o\big(- B^\gamma(x) v\big)$ if $\gamma(t) = \exp_o(tv)$.


\paragraph{Wasserstein Distance on Geodesics.} We saw that we can obtain projections on $\mathbb{R}$. Hence, it is analogous to the Euclidean case as we can use the one dimensional Wasserstein distance on the real line to compute it. In the next proposition, as a sanity check, we verify that the Wasserstein distance between the coordinates is as expected equal to the Wasserstein distance between the measures projected on geodesics. This relies on the isometry property of $t^v$ derived in \Cref{prop:isometry}.

\begin{proposition} \label{prop:eq_wasserstein}
    Let $(\mathcal{M},g)$ a Hadamard manifold, $p\ge 1$ and $\mu,\nu \in \mathcal{P}_p(\mathcal{M})$. Let $v\in T_o\mathcal{M}$ and $\mathcal{G}^v=\{\exp_o(tv),\ t\in\mathbb{R}\}$ the geodesic on which the measures are projected. Then,
    \begin{equation}
        W_p^p(\Tilde{P}^v_\#\mu, \Tilde{P}^v_\#\nu) = W_p^p(P^v_\#\mu, P^v_\#\nu).
    \end{equation}
\end{proposition}

\begin{proof}
    See \Cref{proof:prop_eq_wasserstein}.
\end{proof}

Observing that $t^v\circ \Tilde{B}^v = -B^v$, we obtain a similar result for the Busemann projection.

\begin{proposition} \label{prop:eq_wasserstein_busemann}
    Let $(\mathcal{M},g)$ a Hadamard manifold, $p\ge 1$ and $\mu,\nu \in \mathcal{P}_p(\mathcal{M})$. Let $v\in T_o\mathcal{M}$ and $\mathcal{G}^v=\{\exp_o(tv),\ t\in\mathbb{R}\}$ the geodesic on which the measures are projected. Then,
    \begin{equation}
        W_p^p(\Tilde{B}^v_\#\mu, \Tilde{B}^v_\#\nu) = W_p^p(B^v_\#\mu, B^v_\#\nu).
    \end{equation}
\end{proposition}

\begin{proof}
    See \Cref{proof:prop_eq_wasserstein_busemann}.
\end{proof}

From these properties, we can work equivalently in $\mathbb{R}$ and on the geodesics when using the Busemann projection (also called horospherical projection) or the geodesic projection of measures.

\begin{figure}
    \centering
    \includegraphics[width=0.7\linewidth]{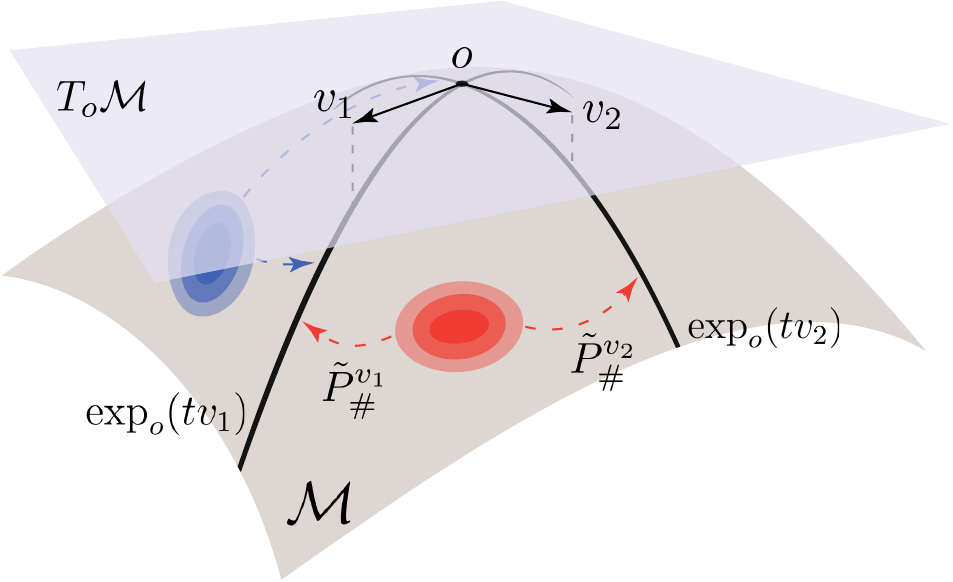}
    \caption{Illustration of the projection process of measures on geodesics $t\mapsto \exp_o(tv_1)$ and $t\mapsto \exp_o(tv_2)$.}
    \label{fig:illustration_rsw}
\end{figure}

\paragraph{Sliced-Wasserstein on Hadamard Manifolds.} We are ready to define the Sliced-Wasserstein distance on Hadamard manifolds. For directions, we will sample from the uniform measure on $\{v\in T_o\mathcal{M},\ \|v\|_o=1\}$. Note that other distributions might be used such as a Dirac in the maximum direction similarly as max-SW \citep{deshpande2019max} for example or any variant using different slicing distributions described in \Cref{section:variants_sw}. But to define a strict generalization of SW, we choose the uniform one in this work. 


\begin{definition}[Cartan-Hadamard Sliced-Wasserstein]
    Let $(\mathcal{M},g)$ a Hadamard manifold with $o$ its origin. Denote $\lambda$ the uniform distribution on $S_o=\{v\in T_o\mathcal{M},\ \|v\|_o=1\}$. Let $p\ge 1$, then we define the $p$-Geodesic Cartan-Hadamard Sliced-Wasserstein distance between $\mu,\nu\in\mathcal{P}_p(\mathcal{M})$ as
    \begin{equation}
        \gchsw_p^p(\mu,\nu) = \int_{S_o} W_p^p(P^v_\#\mu, P^v_\#\nu)\ \mathrm{d}\lambda(v).
    \end{equation}
    Likewise, we define the $p$-Horospherical Cartan-Hadamard Sliced-Wasserstein distance between $\mu,\nu\in\mathcal{P}_p(\mathcal{M})$ as
    \begin{equation}
        \hchsw_p^p(\mu,\nu) = \int_{S_o} W_p^p(B^v_\#\mu, B^v_\#\nu)\ \mathrm{d}\lambda(v).
    \end{equation}
\end{definition}
In the following, when we want to mention both $\gchsw$ and $\hchsw$, for example for properties satisfied by both, we will use the term Cartan-Hadamard Sliced-Wasserstein abbreviated as $\chsw$. Then, we will write without loss of generality
\begin{equation}
    \chsw_p^p(\mu, \nu) = \int_{S_o} W_p^p(P^v_\#\mu, P^v_\#\nu)\ \mathrm{d}\lambda(v),
\end{equation}
with $P^v$ either denoting the geodesic or the horospherical projection. We illustrate the projection process on \Cref{fig:illustration_rsw}.

\subsection{On Manifolds with Non-Negative Curvature}

It is more challenging to develop a unifying theory for manifolds of non-negative curvatures as their geometry can be very different. For example, by Bonnet's theorem \citep[Theorem 11.7]{lee2006riemannian}, spaces whose sectional curvature is bounded below by a positive constant, and which are hence of positive curvature, are compact. It is known that on any compact Riemannian manifold $\mathcal{M}$, there is at least one geodesic which is periodic \citep{gromoll1969periodic}.

In \Cref{chapter:ssw}, we will study the case of the hypersphere, which has constant positive curvature and for which all geodesics are periodic. We can use several constructions to define a sliced method. For example, similarly as for Hadamard manifolds, one might fix an origin, \emph{e.g.} the north pole, and integrate over all geodesics passing through it, by sampling the directions in the tangent space. As the origin on the sphere is arbitrary, we can also choose to integrate over all geodesics which we will do in \Cref{chapter:ssw}.

We leave for future works extending such constructions to other spaces with non-negative curvature, such as the Stiefel manifold \citep{chakraborty2017statistics}, the Grassmannian manifold \citep{wang2023online} or projectives spaces \citep{ziller2007examples}. 



\subsection{Related Works}




\paragraph{Intrinsic Sliced-Wasserstein.}

To the best of our knowledge, the only attempt to define a generalization of the Sliced-Wasserstein distance on Riemannian manifolds was made by \citet{rustamov2020intrinsic}. In this work, they restricted their analysis to compact spaces and proposed to use the eigendecomposition of the Laplace-Beltrami operator (see \citep[Definition 4.7]{gallot1990riemannian}). Let $(\mathcal{M}, g)$ be a compact Riemannian manifold. For $\ell\in\mathbb{N}$, denote $\lambda_\ell$ the eigenvalues and $\phi_\ell$ the eigenfunctions of the Laplace-Beltrami operator sorted by increasing eigenvalues. Then, we can define spectral distances as 
\begin{equation}
    \forall x,y\in\mathcal{M},\ d_\alpha(x,y) = \sum_{\ell\ge 0} \alpha(\lambda_\ell) \big(\phi_\ell(x)-\phi_\ell(y)\big)^2,
\end{equation}
where $\alpha:\mathbb{R}_+\to\mathbb{R}_+$ is a monotonically decreasing function. Then, they define the Intrinsic Sliced-Wasserstein (ISW) distance between $\mu,\nu\in\mathcal{P}_2(\mathcal{M})$ as 
\begin{equation}
    \isw_2^2(\mu,\nu) = \sum_{\ell\ge 0} \alpha(\lambda_\ell) W_2^2\big((\phi_\ell)_\#\mu, (\phi_\ell)_\#\nu\big).
\end{equation}
The eigenfunctions are used to map the measures to the real line, which make it very efficient to compute in practice. The eigenvalues are sorted in increasing order, and the series is often truncated by keeping only the $L$ smallest eigenvalues. 

This distance cannot be applied on Hadamard manifolds as these spaces are not compact. On compact spaces such as the sphere, this provides an alternate sliced distance. In \Cref{chapter:ssw}, we will define the sliced distance by integrating and projecting over all geodesics as we choose to work on the sphere endowed by the geodesic distance with the same tools as in the Euclidean space. We note that $\isw$ is more in the spirit of a max-K Sliced-Wasserstein distance \citep{dai2021sliced}, which projects over the $K$ maximal directions, than the Sliced-Wasserstein distance.

However, on general geometries, the geodesic distance and the geodesic projection can be difficult to compute efficiently, as we may not always have closed-forms. In these situations, using the spectral distance can be beneficial as being more practical to compute but also more robust to noise and geometry aware \citep{lipman2010biharmonic, chen2023riemannian}. Nonetheless, we note that the computation of this spectrum is often impossible \citep{gallot1990riemannian, pennec2006intrinsic}, and that in particular cases where it is possible such as the sphere, computing the eigenfunctions can become numerically unstable in dimension $d\ge 10$ \citep[Appendix A]{dutordoir2020sparse}.

\paragraph{Generalized Sliced-Wasserstein.} A very related distance is the Generalized Sliced-Wasserstein distance \citep{kolouri2019generalized} that we introduced in \Cref{section:variants_sw}. First, the main difference lies in the fact that GSW focuses on probability distributions lying in Euclidean space by projecting the measures along nonlinear hypersurfaces. That said, adapting the definition of GSW to handle probability measures on Riemannian manifolds, and the properties that need to be satisfied by the defining function $g$ such as the homogeneity, then we can write the $\chsw$ in the framework of GSW using $g:(x,v)\mapsto P^v(x)$. We will discuss in the next Section with more details the relations with the Radon transforms.

\section{Properties} \label{section:chsw_properties}

In this Section, we derive theoretical properties of the Cartan-Hyperbolic Sliced-Wasserstein distance. First, we will study its topology and the conditions required to have that $\chsw$ is a true distance. Then, we will study some of its statistical properties.

\subsection{Topology}

\paragraph{Distance Property.} First, we are interested in the distance properties of $\chsw$. From the properties of the Wasserstein distance and of the slicing process, we can show that it is a pseudo-distance, \emph{i.e.} that it satisfies the positivity, the positive definiteness, the symmetry and the triangular inequality.

\begin{proposition} \label{prop:chsw_pseudo_distance}
    Let $p\ge 1$, then $\chsw_p$ is a finite pseudo-distance on $\mathcal{P}_p(\mathcal{M})$.
\end{proposition}

\begin{proof}
    See \Cref{proof:prop_chsw_pseudo_distance}.
\end{proof}

For now, the lacking property is the one of indiscernibility, \emph{i.e.} that $\chsw_p(\mu,\nu)=0$ implies that $\mu=\nu$. We conjecture that it holds but we were not able to show it yet. In the following, we derive a sufficient condition on a related Radon transform to have this property to hold.

Let $f\in L^1(\mathcal{M})$, and let us define, analogously to the Euclidean Radon transform, the Cartan-Hadamard Radon transform $\chr: L^1(\mathcal{M}) \to L^1(\mathbb{R}\times S_o)$ as
\begin{equation}
    \forall t\in\mathbb{R},\ \forall v\in S_o,\ \chr f(t,v) = \int_\mathcal{M} f(x) \mathbb{1}_{\{t = P^v(x)\}}\ \mathrm{d}\vol(x).
\end{equation}
Then, we can also define its dual operator $\chr^*:C_0(\mathbb{R}\times S_o)\to C_b(\mathcal{M})$ for $g\in C_0(\mathbb{R}\times S_o)$ where $C_0(\mathbb{R}\times S_o)$ is the space of continuous functions on $\mathbb{R}\times S_o$ that vanish at infinity, as
\begin{equation}
    \forall x\in \mathcal{M},\ \chr^*g(x) = \int_{S_o} g(P^v(x), v)\ \mathrm{d}\lambda(v).
\end{equation}
\begin{proposition} \label{prop:chsw_dual_rt}
    $\chr^*$ is the dual operator of $\chr$, \emph{i.e.} for all $f\in L^1(\mathcal{M})$, $g\in C_0(\mathbb{R}\times S_o)$,
    \begin{equation}
        \langle \chr f, g\rangle_{\mathbb{R}\times S_o} = \langle f, \chr^* g\rangle_\mathcal{M}.
    \end{equation}
\end{proposition}
\begin{proof}
    See \Cref{proof:prop_chsw_dual_rt}.
\end{proof}
$\chr^*$ maps $C_0(\mathbb{R}\times S_o)$ to $C_b(\mathcal{M})$ because $g$ is necessarily bounded as a continuous function which vanish at infinity. Note that $\chr^*$ actually maps $C_0(\mathbb{R}\times S_o)$ to $C_0(\mathcal{M})$.
\begin{proposition} \label{prop:chr_vanish}
    Let $g\in C_0(\mathbb{R}\times S_o)$, then $\chr^*g \in C_0(\mathcal{M})$.
\end{proposition}
\begin{proof}
    See \Cref{proof:prop_chr_vanish}.
\end{proof}

Using the dual operator, we can define the Radon transform of a measure $\mu$ in $\mathcal{M}$ as the measure $\chr \mu$ satisfying 
\begin{equation}
    \forall g\in C_0(\mathbb{R}\times S_0),\ \int_{\mathbb{R}\times S_o} g(t,v)\ \mathrm{d}(\chr\mu)(t, v) = \int_\mathcal{M} \chr^*g(x)\ \mathrm{d}\mu(x).
\end{equation}
$\chr \mu$ being a measure on $\mathbb{R}\times S_o$, we can disintegrate it \emph{w.r.t} the uniform measure on $S_o$ as $\chr\mu = \lambda \otimes K$ where $K$ is a probability kernel on $S_o\times \mathcal{B}(\mathbb{R})$. In the following proposition, we show that for $\lambda$-almost every $v\in S_o$, $K(v,\cdot)_\#\mu$ coincides with $P^v_\#\mu$.
\begin{proposition} \label{prop:chsw_disintegration}
    Let $\mu$ be a measure on $\mathcal{M}$, then for $\lambda$-almost every $v\in S_o$, $K(v,\cdot)_\#\mu = P^v_\#\mu$.
\end{proposition}
\begin{proof}
    See \Cref{proof:prop_chsw_disintegration}.
\end{proof}
All these derivations allow to link the Cartan-Hadamard Sliced-Wasserstein distance with the corresponding Radon transform. Then, $\chsw_p(\mu,\nu)=0$ implies that for $\lambda$-almost every $v\in S_o$, $P^v_\#\mu=P^v_\#\nu$. Showing that the Radon transform is injective would allow to conclude that $\mu=\nu$.

Actually, here we derived two different Cartan-Hadamard Radon transforms. Using $P^v$ as the geodesic projection, the Radon transform integrates over geodesic subspaces of dimension $\mathrm{dim}(\mathcal{M})-1$. Such spaces are totally geodesic subspaces, and are related to the more general geodesic Radon transform \citep{rubin2003notes}. In the case where the geodesic subspace is of dimension one, \emph{i.e.} it integrates only over geodesics, this coincides with the X-ray transform, and it has been studied \emph{e.g.} in \citep{lehtonen2018tensor}. Here, we are interested in the case of dimension $\mathrm{dim}(\mathcal{M})-1$, which, to the best of our knowledge, has only been studied in \citep{lehtonen2016geodesic} in the case where $\mathrm{dim}(\mathcal{M})=2$ and hence when the geodesic Radon transform and the X-ray transform coincide. However, no results on the injectivity over the sets of measures is yet available. In the case where $P^v$ is the Busemann projection, the set of integration is a horosphere. General horospherical Radon transforms on Cartan-Hadamard manifolds have not yet been studied to the best of our knowledge.



\paragraph{Link with the Wasserstein Distance.} An important property of the Sliced-Wasserstein distance on Euclidean spaces is that it is topologically equivalent to the Wasserstein distance, \emph{i.e.} it metrizes the weak convergence. Such results rely on properties of the Fourier transform which do not translate straightforwardly to manifolds. Hence, deriving such results will require further investigation. We note that a possible lead for the horospherical case is the connection between the Busemann function and the Fourier-Helgason transform \citep{biswas2018fourier, sonoda2022fully}. Using that the projections are Lipschitz functions, we can still show that $\chsw$ is a lower bound of the geodesic Wasserstein distance.

\begin{proposition} \label{prop:chsw_upperbound}
    Let $\mu,\nu\in \mathcal{P}_p(\mathcal{M})$, then
    \begin{equation}
        \chsw_p^p(\mu,\nu) \le W_p^p(\mu,\nu).
    \end{equation}
\end{proposition}

\begin{proof}
    See \Cref{proof:prop_chsw_upperbound}.
\end{proof}

\looseness=-1 This property means that it induces a weaker topology compared to the Wasserstein distance, which can be computationally beneficial but which also comes with less discriminative powers \citep{nadjahi2020statistical}.


\paragraph{First Variations.} Being discrepancies on Hadamard manifolds, CHSWs can be used to learn distributions by minimizing it. An elegant solution could be to use Wasserstein gradient flows of $\mathcal{F}(\mu) = \frac12 \chsw_2^2(\mu,\nu)$ where $\nu$ is some target distribution. As we will see in \Cref{chapter:swgf}, there are many possibilities to solve such a problem. For example, using a JKO-ICNN scheme, we could solve it with well chosen neural networks. Another elegant solution to get samples from $\nu$ is to use the forward Euler scheme, as done previously in \citep{liutkus2019sliced}, which requires to compute its first variation. The first variation can also be used to analyze theoretically the convergence of the Wasserstein gradient flow. As a first step towards computing Wasserstein gradient flows of $\chsw$ on Hadamard spaces, and analyzing them, we derive in \Cref{prop:chsw_1st_variation} the first variation of $\mathcal{F}$. 

\begin{proposition} \label{prop:chsw_1st_variation}
    Let $K$ be a compact subset of $\mathcal{M}$, $\mu,\nu\in\mathcal{P}_2(K)$ with $\mu\ll \vol$. Let $v\in S_o$, denote $\psi_v$ the Kantorovich potential between $P^v_\#\mu$ and $P^v_\#\nu$ for the cost $c(x,y)=\frac12 d(x,y)^2$. Let $\xi$ be a diffeomorphic vector field on $K$ and denote for all $\epsilon\ge 0$, $T_\epsilon : K \to \mathcal{M}$ defined as $T_\epsilon(x) = \exp_x\big(\epsilon \xi(x)\big)$ for all $x\in K$. Then,
    \begin{equation}
        \lim_{\epsilon\to 0^+}\ \frac{\chsw_2^2\big((T_\epsilon)_\#\mu,\nu\big) - \chsw_2^2(\mu,\nu)}{2\epsilon} = \int_{S_o} \int_{\mathcal{M}} \psi_v'\big(P^v(x)\big)\langle \mathrm{grad}_\mathcal{M} P^v(x),\xi(x)\rangle_x \ \mathrm{d}\mu(x)\ \mathrm{d}\lambda(v).
    \end{equation}
\end{proposition}
\begin{proof}
    See \Cref{proof:prop_chsw_1st_variation}.
\end{proof}
In the Euclidean case, we recover well the first variation formula for $\sw$ first derived in \citep[Proposition 5.1.7]{bonnotte2013unidimensional} as in this case, for $x\in \mathbb{R}^d$, $T_\epsilon(x) = x + \epsilon \xi(x)$, and for $\theta\in S^{d-1}$, $P^\theta(x) = \langle x,\theta\rangle$ and thus $\mathrm{grad}P^\theta(x) =\nabla P^\theta(x) = \theta$, and we recover
\begin{equation}
    \lim_{\epsilon\to 0^+}\ \frac{\sw_2^2\big((\mathrm{Id}+\epsilon\xi)_\#\mu, \nu\big)- \sw_2^2(\mu,\nu)}{2\epsilon} = \int_{S^{d-1}} \int_{\mathbb{R}^d} \psi_\theta'\big(P^\theta(x)\big) \big\langle \theta, \xi(x)\big\rangle\ \mathrm{d}\mu(x)\ \mathrm{d}\lambda(\theta).
\end{equation}

\paragraph{Hilbert Embedding.} $\chsw$ also comes with the interesting properties that it can be embedded in Hilbert spaces. This is in contrast with the Wasserstein distance which is known to not be Hilbertian \citep[Section 8.3]{peyre2019computational} except in one dimension where it coincides with its sliced counterpart.

\begin{proposition} \label{prop:chsw_hilbertian}
    Let $p\ge 1$ and $\mathcal{H}=L^p([0,1]\times S_o, \mathrm{Leb}\otimes \lambda)$. We define $\Phi$ as 
    \begin{equation}
        \begin{aligned}
        \Phi : \ &\mathcal{P}_p(\mathcal{M}) \rightarrow \mathcal{H}\\
        &\mu \mapsto \big( (q, v) \mapsto F^{-1}_{P^v_{\#}\mu}(q) \big),
        \end{aligned}
    \end{equation}
    where $F^{-1}_{P^v_\#\mu}$ is the quantile function of $P^v_\#\mu$. Then $\chsw_p$ is Hilbertian and for all $\mu,\nu\in \mathcal{P}_p(\mathcal{M})$,
    \begin{equation}
        \chsw_p^p(\mu,\nu) = \|\Phi(\mu)-\Phi(\nu)\|_\mathcal{H}^p.
    \end{equation}
\end{proposition}

\begin{proof}
    See \Cref{proof:prop_chsw_hilbertian}.
\end{proof}

This is a nice property which allows to define a valid positive definite kernel for measures such as the Gaussian kernel \citep[Theorem 6.1]{jayasumana2015kernel}, and hence to use kernel methods \citep{hofmann2008kernel}. This can allow for example to perform distribution clustering, classification \citep{kolouri2016sliced, carriere2017sliced} or regression \citep{meunier2022distribution}.

\begin{proposition}
    Define the kernel $K:\mathcal{P}_2(\mathcal{M})\times \mathcal{P}_2(\mathcal{M})\to \mathbb{R}$ as $K(\mu,\nu) = \exp\big(-\gamma \chsw_2^2(\mu,\nu)\big)$ for $\gamma>0$. Then $K$ is a positive definite kernel. 
\end{proposition}

\begin{proof}
    Apply \citep[Theorem 6.1]{jayasumana2015kernel}.
\end{proof}

Note that to show that the Gaussian kernel is universal, \emph{i.e.} that the resulting Reproducing Kernel Hilbert Space (RKHS) is powerful enough to approximate any continuous function \citep{meunier2022distribution}, we would need additional results such as that it metrizes the weak convergence and that $\chsw_2$ is a distance, as shown in \citep[Proposition 7]{meunier2022distribution}.

\subsection{Statistical Properties}

\paragraph{Sample Complexity.} In practical settings, we usually cannot directly compute the closed-form between $\mu,\nu\in\mathcal{P}_p(\mathcal{M})$, but we have access to samples $x_1,\dots,x_n\sim \mu$ and $y_1,\dots,y_n\sim \nu$. Then, it is common practice to estimate the discrepancy with the plug-in estimator $\chsw(\hat{\mu}_n,\hat{\nu}_n)$ \citep{manole2021plugin,manole2022minimax,niles2022estimation} where $\hat{\mu}_n = \frac{1}{n}\sum_{i=1}^n \delta_{x_i}$ and $\hat{\nu}_n = \frac{1}{n}\sum_{i=1}^n \delta_{y_i}$ are empirical estimations of the measures. We are interested in characterizing the speed of convergence of the plug-in estimator towards the true distance. Relying on the proof of \citet{nadjahi2020statistical}, we derive in \Cref{prop:chsw_sample_complexity} the sample complexity of $\chsw$. As in the Euclidean case, we find that the sample complexity does not depend on the dimension, which is an important and appealing property of sliced divergences \citep{nadjahi2020statistical} compared to the Wasserstein distance, which has a sample complexity in $O(n^{-1/d})$ \citep{niles2022estimation}.

\begin{proposition} \label{prop:chsw_sample_complexity}
    Let $p\ge 1$, $q>p$ and $\mu,\nu\in\mathcal{P}_p(\mathcal{M})$. Denote $\hat{\mu}_n$ and $\hat{\nu}_n$ their counterpart empirical measures and $M_q(\mu) = \int_\mathcal{M} d(x,o)^q\ \mathrm{d}\mu(x)$ their moments of order $q$. Then, there exists $C_{p,q}$ a constant depending only on $p$ and $q$ such that 
    \begin{equation}
        \mathbb{E}\big[|\chsw_p(\hat{\mu}_n, \hat{\nu}_n) - \chsw_p(\mu,\nu)|\big] \le 2C_{p,q}^{1/p} \big(M_q(\mu)^{1/q} + M_q(\nu)^{1/q}\big) 
        \left\{ 
        \begin{array}{ll}
          n^{-1/(2p)} & \mbox{ if } q > 2p, \\
          n^{-1/(2p)} \log(n)^{1/p} & \mbox{ if } q = 2p, \\
          n^{-(q-p)/(pq)} & \mbox{ if } q \in (p,2p).
        \end{array}
        \right.
    \end{equation}
\end{proposition}

\begin{proof}
    See \Cref{proof:prop_chsw_sample_complexity}.
\end{proof}

This property is very appealing in practical settings as it allows to use the same number of samples while having the same convergence rate in any dimension. In practice though, we cannot compute exactly $\chsw_p(\hat{\mu}_n, \hat{\nu}_n)$ as the integral on $S_o$ \emph{w.r.t.} the uniform measure $\lambda$ is intractable.


\paragraph{Projection Complexity.} Thus, to compute it in practice, we usually rely on a Monte-Carlo approximation, by drawing $L\ge 1$ projections $v_1,\dots,v_L$ and approximating the distance by $\widehat{\chsw}_{p,L}$ defined between $\mu,\nu\in\mathcal{P}_p(\mathcal{M})$ as
\begin{equation}
    \widehat\chsw_{p,L}^p(\mu,\nu) = \frac{1}{L} \sum_{\ell=1}^L W_p^p(P^{v_\ell}_\#\mu, P^{v_\ell}_\#\nu).
\end{equation}
In the following proposition, we derive the Monte-Carlo error of this approximation, and we show that we recover the classical rate of $O(1/\sqrt{L})$.

\begin{proposition} \label{prop:chsw_proj_complexity}
    Let $p\ge 1$, $\mu,\nu\in\mathcal{P}_p(\mathcal{M})$. Then, the error made by the Monte-Carlo estimate of $\chsw_p$ with $L$ projections can be bounded as follows
    \begin{equation}
        \mathbb{E}_v\left[|\widehat{\chsw}_{p,L}^p(\mu,\nu) - \chsw_p^p(\mu,\nu)|\right]^2 \le \frac{1}{L} \mathrm{Var}_v\left[W_p^p(P^v_\#\mu, P^v_\#\nu)\right].
    \end{equation}
\end{proposition}

\begin{proof}
    See \Cref{proof:prop_chsw_proj_complexity}.
\end{proof}

We note that here the dimension actually intervenes in the term of variance $\mathrm{Var}_v\left[W_p^p(P^v_\#\mu, P^v_\#\nu)\right]$.

\paragraph{Computational Complexity.} As we project on the real line, the complexity of computing the Wasserstein distances between each projected sample is in $O(Ln\log n)$. Then, we add the complexity of computing the projections, which will depend on the spaces and whether or not we have access to a closed-form.

\section{Future Works and Discussions}

In this chapter, we introduced formally a way to generalize the Sliced-Wasserstein distance on Riemannian manifolds of non-positive curvature. In the next two chapters, we will study these constructions in two particular cases of such manifolds: Hyperbolic spaces and the space of Symmetric Positive-Definite matrices. Further works might include constructing SW type distances on geodesically complete Riemannian manifolds of non-negative curvature. Such spaces have more complicated geometries which makes it harder to build a general construction. Hence, we will focus in \Cref{chapter:ssw} on the particular case of the hypersphere, which is a space of positive constant curvature.

Besides constructing SW distances on Riemannian manifolds, one could also be interested in extending the constructions on more general metric spaces. A particular class of such space with appealing properties, and which encloses Hadamard manifolds, are CAT(0) spaces \citep{bridson2013metric}. Optimal transport on these classes of metric spaces have recently received some attention \citep{berdellima2023existence}. We could also study generalization of Riemannian manifolds such as Finsler manifolds \citep{shen2001lectures} which have recently received some attention in Machine Learning \citep{lopez2021symmetric, pouplin2022identifying}.

For the projections, we study two natural generalizations of the projection used in Euclidean spaces. We could also study other projections which do not follow geodesics subspaces or horospheres, but are well suited to Riemannian manifolds, in the same spirit of the Generalized Sliced-Wasserstein. Other subspaces could also be used, such as Hilbert curves \citep{li2022hilbert} adapted to manifolds, or higher dimensional subspaces \citep{paty2019subspace, chami2021horopca}. Finally, we could also define other variations of $\chsw$ such as $\text{max-CHSW}$ for instance and more generally adapt many of the variants described in \Cref{section:variants_sw} to the case of Riemannian manifolds. Note also that the Busemann function is an example of a more broad class of functions called horofunctions. On Hadamard manifolds, horofunctions are necessarily Busemann functions, but it might not be the case on more general metric spaces.

On the theoretical side, we still need to show that these Sliced-Wasserstein discrepancies are proper distances by showing the indiscernible property. It might also be interesting to study whether statistical properties for the Euclidean SW distance derived in \emph{e.g.} \citep{nietert2022statistical, manole2022minimax, goldfeld2022statistical, xu2022central, xi2022distributional} still hold more generally for $\chsw$ on any Cartan-Hadamard manifold.



\clearemptydoublepage
\cleartooddpage[\thispagestyle{empty}]
\chapter{Hyperbolic Sliced-Wasserstein} \label{chapter:hsw}

{
    \hypersetup{linkcolor=black} 
    \minitoc 
}

In this chapter, based on \citep{bonet2022hyperbolic}, we study the Sliced-Wasserstein distance on a particular case of Hadamard manifold: Hyperbolic spaces. Hyperbolic space embeddings have been shown beneficial for many learning tasks where data have an underlying hierarchical structure. Consequently, many machine learning tools were extended to such spaces, but only few discrepancies exist to compare probability distributions defined over those spaces. Among the possible candidates, Optimal Transport distances are well defined on such Riemannian manifolds and enjoy strong theoretical properties, but suffer from high computational cost. On Euclidean spaces,  Sliced-Wasserstein distances, which leverage a closed-form solution of the Wasserstein distance in one dimension, are more computationally efficient, but are not readily available on Hyperbolic spaces. In this work, we propose to derive novel Hyperbolic Sliced-Wasserstein discrepancies. These constructions use projections on the underlying geodesics either along horospheres or geodesics. We study and compare them on different tasks where hyperbolic representations are relevant, such as sampling or image classification.

\section{Introduction}

\looseness=-1 In recent years, Hyperbolic spaces have received a lot of attention in machine learning (ML) as they allow to efficiently process data 
that present a hierarchical structure \citep{nickel2017poincare,nickel2018learning}. This encompasses data such as graphs \citep{gupte2011finding}, words \citep{tifrea2018poincare} or images \citep{khrulkov2020hyperbolic}. Embedding in Hyperbolic spaces has been proposed for various applications such as drug embedding \citep{yu2020semi}, image clustering \citep{park2021unsupervised, ghadimi2021hyperbolic}, zero-shot recognition \citep{liu2020hyperbolic}, remote sensing \citep{hamzaoui2021hyperbolic} or reinforcement learning \citep{cetin2022hyperbolic}. Hence, many works proposed to develop tools to be used on such spaces, such as generalization of Gaussian distributions \citep{nagano2019wrapped, cho2022rotated, galaz2022wrapped}, neural networks \citep{ganea2018hyperbolic, liu2019hyperbolic} or Normalizing Flows \citep{lou2020neural, bose2020latent}.


As we saw in \Cref{chapter:sw_hadamard}, the theoretical study of the Wasserstein distance on Riemannian manifolds is well developed \citep{mccann2001polar, villani2009optimal}. When it comes to Hyperbolic spaces, some Optimal Transport attempts aimed at aligning distributions of data which have been embedded in a Hyperbolic space \citep{alvarez2020unsupervised,hoyos2020aligning}. To the best of our knowledge, the SW distance has not been extended yet to Hyperbolic spaces. Hence, in this chapter, we leverage the general theory derived in \Cref{chapter:sw_hadamard} and apply it in this particular case.

\paragraph{Contributions.} We extend Sliced-Wasserstein to data living in Hyperbolic spaces. Analogously to Euclidean SW, we project the distributions on geodesics passing through the origin. Interestingly enough, different projections can be considered, leading to several new SW constructions that exhibit different theoretical properties and empirical benefits. We make connections with Radon transforms already defined in the literature and we show that Hyperbolic SW are (pseudo-) distances. We provide the algorithmic procedure and discuss its complexity. We illustrate the benefits of these new Hyperbolic SW distances on several tasks such as sampling or image classification.

\section{Background on Hyperbolic Spaces}

Hyperbolic spaces are Riemannian manifolds of negative constant curvature \citep{lee2006riemannian} and are particular cases of Hadamard manifolds studied in \Cref{chapter:sw_hadamard}. They have recently received a surge of interest in machine learning as they allow embedding data with a hierarchical structure efficiently \citep{nickel2017poincare,nickel2018learning}. A thorough review of the recent use of hyperbolic spaces in machine learning can be found in \citep{peng2021hyperbolic} and in \citep{mettes2023hyperbolic}.

There are five usual parameterizations of a hyperbolic manifold \citep{peng2021hyperbolic}. They are equivalent (isometric) and one can easily switch from one formulation to the other. Hence, in practice, we use the one which is the most convenient, either given the formulae to derive or the numerical properties. In machine learning, the two most used models are the Poincaré ball and the Lorentz model (also known as the hyperboloid model). Each of these models has its own advantages compared to the other. For example, the Lorentz model has a distance which behaves better \emph{w.r.t.} numerical issues compared to the distance of the Poincaré ball. 
However, the Lorentz model is unbounded, contrary to the Poincaré ball. We introduce in the following these two models as we will use both of them 
in our work.

\subsection{Lorentz Model}

First, we introduce the Lorentz model $\mathbb{L}^d\subset \mathbb{R}^{d+1}$ of a $d$-dimensional hyperbolic space. It can be defined as
\begin{equation}
    \mathbb{L}^d = \big\{(x_0,\dots,x_d)\in \mathbb{R}^{d+1},\ \langle x,x\rangle_\mathbb{L}=-1,\ x_0>0\big\}
\end{equation}
where
\begin{equation}
    \forall x,y\in\mathbb{R}^{d+1},\ \langle x,y\rangle_\mathbb{L} = -x_0y_0 + \sum_{i=1}^{d} x_iy_i
\end{equation}
is the Minkowski pseudo inner-product \citep[Chapter 7]{boumal2023introduction}. The Lorentz model can be seen as the upper sheet of a two-sheet hyperboloid. In the following, we will denote $x^0 = (1,0,\dots,0)\in\mathbb{L}^d$ the origin of the hyperboloid. The geodesic distance in this manifold, which denotes the length of the shortest path between two points, can be defined as
\begin{equation}
    \forall x,y\in\mathbb{L}^d,\ d_{\mathbb{L}}(x,y) = \arccosh\big(-\langle x,y\rangle_\mathbb{L}\big).
\end{equation}

At any point $x\in\mathbb{L}^d$, we can associate a subspace of $\mathbb{R}^{d+1}$ orthogonal in the sense of the Minkowski inner product. These spaces are called tangent spaces and are described formally as $T_x\mathbb{L}^d=\{v\in\mathbb{R}^{d+1},\ \langle v,x\rangle_\mathbb{L}=0\}$. Note that on tangent spaces, the Minkowski inner-product is a real inner product. In particular, on $T_{x^0}\mathbb{L}^d$, it is the usual Euclidean inner product, \emph{i.e.} for $u,v\in T_{x^0}\mathbb{L}^d$, $\langle u,v\rangle_\mathbb{L}=\langle u,v\rangle$. Moreover, for all $v\in T_{x^0}\mathbb{L}^d$, $v_0=0$.

We can draw a connection with the sphere. Indeed, by endowing $\mathbb{R}^{d+1}$ with $\langle\cdot,\cdot\rangle_\mathbb{L}$, we obtain $\mathbb{R}^{1,d}$ the so-called Minkowski space. Then, $\mathbb{L}^d$ is the analog in the Minkowski space of the sphere $S^{d}$ in the regular Euclidean space \citep{bridson2013metric}.

\subsection{Poincaré Ball}

The second model of hyperbolic space we will be interested in is the Poincaré ball $\mathbb{B}^d\subset\mathbb{R}^d$. This space can be obtained as the stereographic projection of each point $x\in\mathbb{L}^d$ onto the hyperplane $\{x\in\mathbb{R}^{d+1},\ x_0=0\}$. More precisely, the Poincaré ball is defined as
\begin{equation}
    \mathbb{B}^d = \{x\in\mathbb{R}^d,\ \|x\|_2 < 1\},
\end{equation}
with geodesic distance, for all $x,y\in\mathbb{B}^d$,
\begin{equation}
    d_\mathbb{B}(x,y) = \arccosh\left(1+2\frac{\|x-y\|_2^2}{(1-\|x\|_2^2)(1-\|y\|_2^2)}\right).
\end{equation}
We see in this formulation that the distance can be subject to numerical instabilities when one of the points is too close to the boundary of the ball.
\par
We can switch from Lorentz to Poincaré using the following isometric projection \citep{nickel2018learning}:
\begin{equation} \label{eq:proj_lorentz_to_poincare}
    \forall x\in\mathbb{L}^d,\ P_{\mathbb{L}\to\mathbb{B}}(x) = \frac{1}{1+x_0} (x_1,\dots, x_d)
\end{equation}
and from Poincaré to Lorentz by
\begin{equation} \label{eq:proj_poincare_to_lorentz}
    \forall x \in \mathbb{B}^d,\ P_{\mathbb{B}\to\mathbb{L}}(x) = \frac{1}{1-\|x\|_2^2}(1+\|x\|_2^2, 2x_1,\dots, 2x_d).
\end{equation}

\section{Hyperbolic Sliced-Wasserstein Distances} \label{section:hsw}

\begin{figure}[t]
    \centering
    \hspace*{\fill}
    \subfloat[Euclidean.]{\label{fig:proj_euc}\includegraphics[width={0.15\linewidth}]{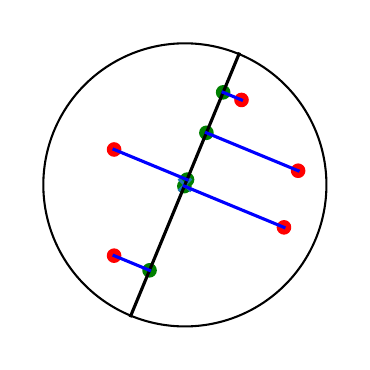}} \hfill
    \subfloat[Geodesics.]{\label{fig:proj_geods}\includegraphics[width={0.15\linewidth}]{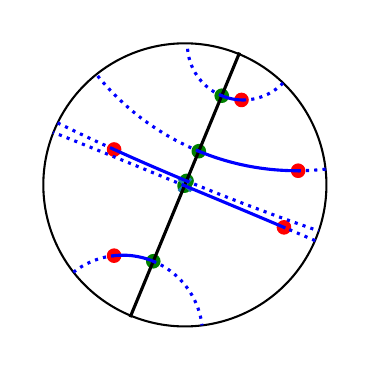}} \hfill
    \subfloat[Horospheres.]{\label{fig:proj_horo}\includegraphics[width={0.15\linewidth}]{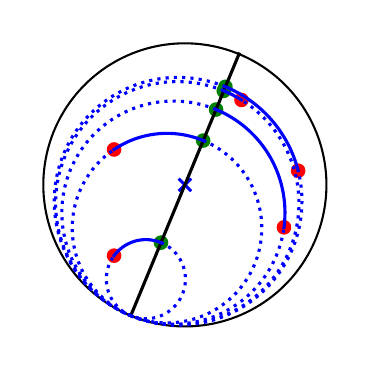}} \hfill
    \subfloat[Euclidean.]{\label{fig:proj_euc_l}\includegraphics[width=0.15\linewidth]{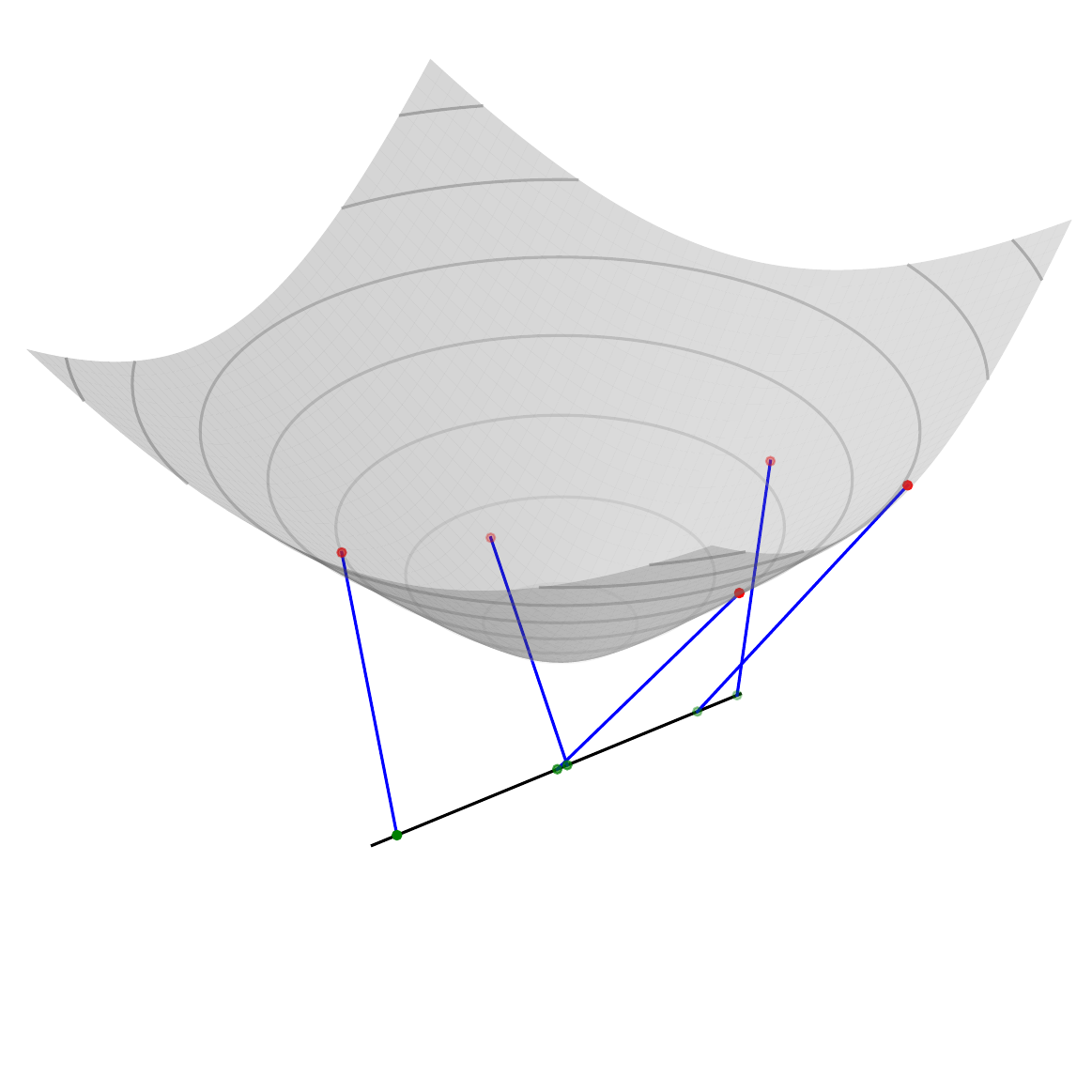}} 
    \subfloat[Geodesics.]{\label{fig:proj_lorentz}\includegraphics[width=0.15\linewidth]{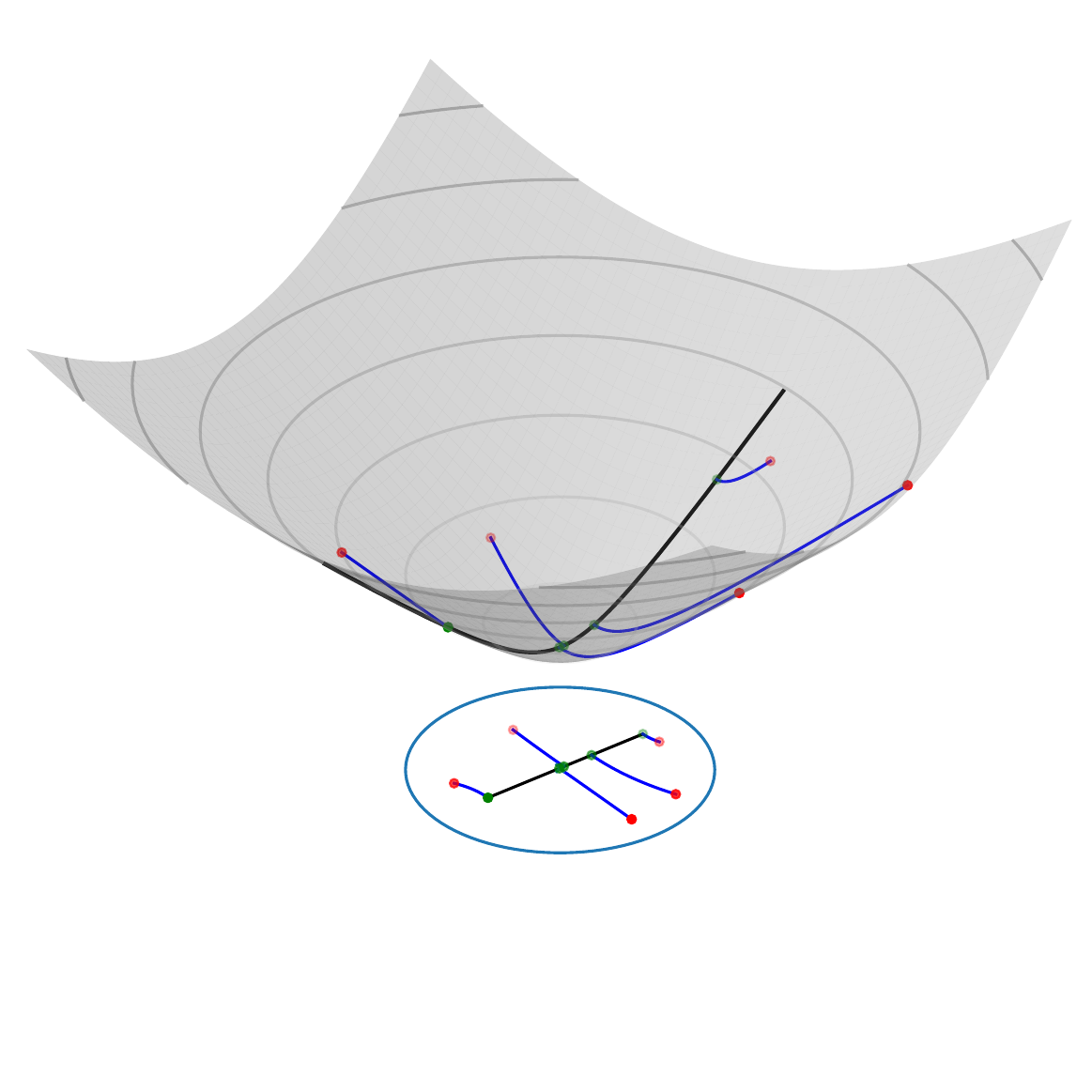}} \hfill
    \subfloat[Horospheres.]{\label{fig:proj_lorentz_horo}\includegraphics[width=0.15\linewidth]{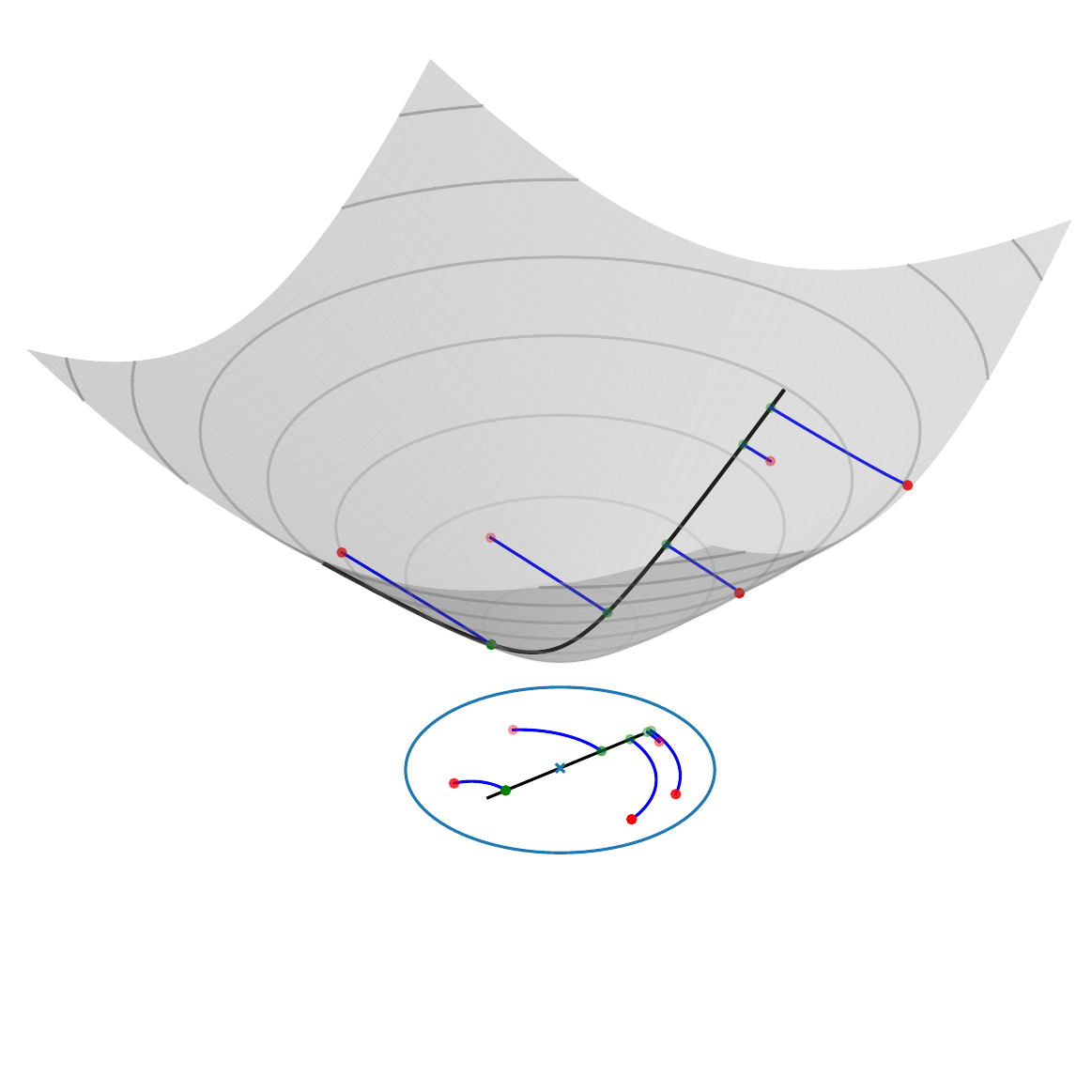}} \hfill
    \hspace*{\fill}
    \caption{Projection of (red) points on a geodesic (black line) in the Poincaré ball and in the Lorentz model along  Euclidean lines, geodesics or horospheres (in blue). Projected points on the geodesic are shown in green.}
    \label{fig:projections_hsw}
\end{figure}

\looseness=-1 In this section, we aim at introducing Sliced-Wasserstein type of distances on Hyperbolic spaces. Interestingly enough, several constructions can be performed, depending on the projections that are involved. The first solution we consider is the extension of Euclidean SW between distributions whose support lies on Hyperbolic spaces. Then, we consider variants involving the geodesic distance, and derived as particular cases of $\chsw$ on hyperbolic spaces. The different projection processes are illustrated in \Cref{fig:projections_hsw}. 

\subsection{Euclidean Sliced-Wasserstein on Hyperbolic Spaces}

The support of distributions lying on Hyperbolic space are included in the ambient spaces $\mathbb{R}^{d}$ (Poincaré ball) or $\mathbb{R}^{d+1}$ (Lorentz model). As such, Euclidean SW can be used for such data. On the Poincaré ball, the projections lie onto the manifold as geodesics passing through the origin are straight lines (see \Cref{sec:hsw_proj}), but the initial geometry of the data might not be fully taken care of as the orthogonal projection does not respect the Poincaré geodesics. On the Lorentz model though, the projections lie out of the manifold. We will denote SWp and SWl the Poincaré ball and Lorentz model version. These formulations allow inheriting from the properties of SW, such as being a distance.

\subsection{Hyperbolic Sliced-Wasserstein} \label{sec:hsw_proj}

As Hyperbolic spaces are particular cases of Hadamard manifolds, we leverage the constructions proposed in \Cref{chapter:sw_hadamard}. We saw in this chapter two different constructions of sliced distances on such spaces. Both of them involve projecting the measures on geodesics passing through an origin, and differ with which projection is used. First, we describe the geodesics in these spaces. Then, we derive the closed-form for the geodesic projection and for the horospherical projection for both models.

\paragraph{Geodesics in the Lorentz model.} In the Lorentz model, geodesics passing through the origin $x^0$ can be obtained by taking the intersection between $\mathbb{L}^d$ and a 2-dimensional plane containing $x^0$ \citep[Proposition 5.14]{lee2006riemannian}. Any such plane can be obtained as $\mathrm{span}(x^0,v)$ where $v\in T_{x^0}\mathbb{L}^d\cap S^{d} = \{v\in S^d,\ v_0=0\}$. The corresponding geodesic can be described by a geodesic line \citep[Corollary 2.8]{bridson2013metric}, \emph{i.e.} a map $\gamma:\mathbb{R}\to\mathbb{L}^d$ satisfying for all $t,s\in\mathbb{R}$, $d_\mathbb{L}(\gamma(s),\gamma(t)) = |t-s|$, of the form 
\begin{equation}
    \forall t\in\mathbb{R},\ \gamma(t) = \cosh(t) x^0 + \sinh(t) v.
\end{equation}

\paragraph{Geodesics in the Poincaré ball.} On the Poincaré ball, geodesics are circular arcs perpendicular to the boundary $S^{d-1}$ \citep[Proposition 5.14]{lee2006riemannian}. In particular, geodesics passing through the origin are straight lines. Hence, they can be characterized by a point $\Tilde{v}$ on the border $S^{d-1}$. Such points will be called ideal points. Formally, they can be described as 
\begin{equation}
    \forall t\in\mathbb{R},\ \gamma(t) = \exp_0(t\Tilde{v}) = \tanh\left(\frac{t}{2}\right) \Tilde{v}.
\end{equation}


\paragraph{Geodesic projections in Hyperbolic spaces.} Now, let us derive the geodesic projections. We provide the different formulas for both the Lorentz model and the Poincaré ball. First, let us recall that geodesic projections are defined as 
\begin{equation}
    \forall x\in \mathbb{L}^d,\ \Tilde{P}^\mathcal{G}(x) = \argmin_{y\in\mathcal{G}}\ d_\mathbb{L}(x,y),
\end{equation}
for a geodesic $\mathcal{G}$.

\begin{proposition}[Geodesic projection] \label{prop:hsw_geodesic_proj} \leavevmode
    \begin{enumerate}
        \item Let $\mathcal{G}^v=\mathrm{span}(x^0,v)\cap \mathbb{L}^d$ where $v\in T_{x^0}\mathbb{L}^d\cap S^d$. Then, the geodesic projection $\Tilde{P}^v$ on $\mathcal{G}^v$ of $x\in\mathbb{L}^d$ is
        \begin{equation}
            \begin{aligned}
                \Tilde{P}^v(x) &= \frac{1}{\sqrt{\langle x,x^0\rangle_\mathbb{L}^2-\langle x, v\rangle_\mathbb{L}^2}} \big(-\langle x,x^0\rangle_\mathbb{L}x^0 + \langle x,v\rangle_\mathbb{L} v\big) \\
                &= \frac{P^{\mathrm{span}(x^0, v)}(x)}{\sqrt{-\langle P^{\mathrm{span}(x^0, v)}(x), P^{\mathrm{span}(x^0, v)}(x)\rangle_\mathbb{L}}},
            \end{aligned}
        \end{equation}
        where $P^{\mathrm{span}(x^0,v^0)}$ is the linear orthogonal projection on the subspace $\mathrm{span}(x^0,v)$.
        \item Let $\Tilde{v}\in S^{d-1}$ be an in ideal point. Then, the geodesic projection $\Tilde{P}^{\Tilde{v}}$ on the geodesic characterized by $\Tilde{v}$ of $x\in \mathbb{B}^d$ is
        \begin{equation}
            \Tilde{P}^{\Tilde{v}}(x) = s(x) \Tilde{v},
        \end{equation}
        where
        \begin{equation}
            s(x) = \left\{\begin{array}{ll} \frac{1+\|x\|_2^2 - \sqrt{(1+\|x\|_2^2)^2 - 4 \langle x, \Tilde{v}\rangle^2}}{2 \langle x, \Tilde{v}\rangle} & \mbox{ if } \langle x,\Tilde{v}\rangle \neq 0 \\
            0 & \mbox{ if } \langle x,\Tilde{v}\rangle = 0.
            \end{array}\right.
        \end{equation}
    \end{enumerate}
\end{proposition}

\begin{proof}
    See \Cref{proof:prop_hsw_geodesic_proj}.
\end{proof}

We observe that on the Lorentz model, the projection on the geodesic can be done by first projecting on the subspace $\mathrm{span}(x^0,v)$ and then by projecting on the hyperboloid by normalizing. This is analogous to the spherical case studied later in \Cref{chapter:ssw}. Note that while it is analogous, the constructions have some differences since the sphere is not a Hadamard manifold.

For practical implementations, we can also derive in closed-form the coordinate on a geodesic.

\pagebreak

\begin{proposition}[Coordinate of the geodesic projection] \label{prop:hsw_coord_geod_proj} \leavevmode
    \begin{enumerate}
        \item Let $\mathcal{G}^v = \mathrm{span}(x^0, v)\cap \mathbb{L}^d$ where $v\in T_{x^0}\mathbb{L}^d\cap S^d$. Then, the coordinate $P^v$ of the geodesic projection on $\mathcal{G}^v$ of $x\in \mathbb{L}^d$ is
        \begin{equation}
            P^v(x) = \arctanh\left(-\frac{\langle x, v\rangle_\mathbb{L}}{\langle x,x^0\rangle_\mathbb{L}}\right).
        \end{equation}
        \item Let $\Tilde{v}\in S^{d-1}$ be an ideal point. Then, the coordinate $P^{\Tilde{v}}$ of the geodesic projection on the geodesic characterized by $\Tilde{v}$ of $x\in \mathbb{B}^d$ is
        \begin{equation}
            P^{\Tilde{v}}(x) = 2 \arctanh\big(s(x)\big),
        \end{equation}
        where $s$ is defined as in \Cref{prop:hsw_geodesic_proj}.
    \end{enumerate}
\end{proposition}

\begin{proof}
    See \Cref{proof:prop_hsw_coord_geod_proj}.
\end{proof}

Now, following the construction of the Geodesic Cartan-Hadamard Sliced-Wasserstein distance, we have all the tools in closed-form to define the Geodesic Hyperbolic Sliced-Wasserstein distance (GHSW) between $\mu,\nu\in\mathcal{P}_p(\mathbb{L}^d)$ as, for $p \ge 1$,
\begin{equation} \label{eq:ghsw_lorentz}
    \ghsw_p^p(\mu,\nu) = \int_{T_{x^0}\mathbb{L}^d \cap S^{d}} W_p^p(P^v_\#\mu, P^v_\#\nu)\ \mathrm{d}\lambda(v).
\end{equation}
Note that $T_{x^0}\mathbb{L}^d\cap S^d \cong S^{d-1}$ and that $v$ can be drawn by first sampling $\Tilde{v}\sim\mathrm{Unif}(S^{d-1})$ and then adding a $0$ in the first coordinate, \emph{i.e.} $v=(0,\Tilde{v})$ with $\Tilde{v}\in S^{d-1}$. 

Similarly, we can define GHSW between $\mu,\nu\in\mathcal{P}(\mathbb{B}^d)$ as
\begin{equation} \label{eq:ghsw_poincare}
    \ghsw_p^p(\mu,\nu) = \int_{S^{d-1}} W_p^p(P^{\Tilde{v}}_\#\mu, P^{\Tilde{v}}_\#\nu)\ \mathrm{d}\lambda(\Tilde{v}).
\end{equation}

\paragraph{Horospherical projections in Hyperbolic spaces.} Now, we derive horospherical projections by the mean of the Busemann function. We recall that Busemann functions are defined as 
\begin{equation}
    B^\gamma(x) = \lim_{t\to\infty}\ \big( d(x,\gamma(t))- t\big),
\end{equation}
for $\gamma$ a geodesic ray, and that they can be seen as a generalization of the inner product on manifolds. Moreover, its level sets are horospheres, which can be seen as generalization of hyperplanes or also as spheres of infinite radius \citep{izumiya2009horospherical}, and along which we will project the measures in this part. Now, let us state the closed-form of the Busemann function in the Lorentz model and in the Poincaré ball.

\begin{proposition}[Busemann function on hyperbolic space] \label{prop:busemann_closed_forms} \leavevmode
    \begin{enumerate}
        \item On $\mathbb{L}^d$, for any direction $v\in T_{x^0}\mathbb{L}^d\cap S^d$, 
        \begin{equation}
            \forall x\in\mathbb{L}^d,\ B^v(x) = \log\big(-\langle x,x^0+v\rangle_\mathbb{L}\big).
        \end{equation}
        \item On $\mathbb{B}^d$, for any ideal point $\Tilde{v}\in S^{d-1}$,
        \begin{equation}
            \forall x\in \mathbb{B}^d,\ B^{\Tilde{v}}(x) = \log\left(\frac{\|\Tilde{v}-x\|_2^2}{1-\|x\|_2^2}\right).
        \end{equation}
    \end{enumerate}
\end{proposition}

\begin{proof}
    See \Cref{proof:prop_busemann_closed_forms}.
\end{proof}

\looseness=-1 To conserve Busemann coordinates on Hyperbolic spaces, it has been proposed by \citet{chami2021horopca} to project points on a subset following its level sets which are horospheres. In the Poincaré ball, a horosphere is a Euclidean sphere tangent to an ideal point. \citet{chami2021horopca} argued that this projection is beneficial against the geodesic projection as it tends to better preserve the distances. This motivates us to project on geodesics following the level sets of the Busemann function in order to conserve the Busemann coordinates, \emph{i.e.} we want to have $B_{\Tilde{v}}(x) = B_{\Tilde{v}}\big(P^{\Tilde{v}}(x)\big)$ (resp. $B_v(x)=B_v\big(P^v(x)\big)$) on the Poincaré ball (resp. Lorentz model) where $\Tilde{v}\in S^{d-1}$ (resp. $v\in T_{x^0}\mathbb{L}^d\cap S^d$) is characterizing the geodesic. In the next Proposition, we derive for completeness a closed-form for the projection in both the Poincaré ball and Lorentz model. Note that for a practical implementation, we will use the Busemann coordinates directly. 

\begin{proposition}[Horospherical projection] \label{prop:horospherical_projection} \leavevmode
    \begin{enumerate}
        \item Let $v\in T_{x^0}\mathbb{L}^d\cap S^d$ be a direction and $\mathcal{G}^v=\mathrm{span}(x^0,v)\cap \mathbb{L}^d$ the corresponding geodesic passing through $x^0$. Then, for any $x\in\mathbb{L}^d$, the projection on $\mathcal{G}^v$ along the horosphere is given by
        \begin{equation}
            \Tilde{B}^v(x) = \frac{1+u^2}{1-u^2} x^0 + \frac{2u}{1-u^2} v,
        \end{equation}
        where $u = \frac{1+\langle x, x^0+v\rangle_\mathbb{L}}{1-\langle x, x^0+v\rangle_\mathbb{L}}$.
        \item Let $\Tilde{v}\in S^{d-1}$ be an ideal point. Then, for all $x\in \mathbb{B}^d$,
        \begin{equation}
            \Tilde{B}^{\Tilde{v}}(x) = \left(\frac{1-\|x\|_2^2-\|\Tilde{v}-x\|_2^2}{1-\|x\|_2^2+\|\Tilde{v}-x\|_2^2}\right)\Tilde{v}.
        \end{equation}
    \end{enumerate}
\end{proposition}

\begin{proof}
    See \Cref{proof:prop_horospherical_projections}.
\end{proof}

Now, we have also all the tools to construct, similarly as the Horospherical Cartan-Hadamard Sliced-Wasserstein, the Horospherical Hyperbolic Sliced-Wasserstein distance (HHSW) between $\mu, \nu \in \mathcal{P}_p(\mathbb{L}^d)$ as, for $p\ge 1$,
\begin{equation}
    \hhsw_p^p(\mu,\nu) = \int_{T_{x^0}\mathbb{L}^d\cap S^d} W_p^p(B^v_\#\mu, B^v_\#\nu)\ \mathrm{d}\lambda(v).
\end{equation}
We also provide a formulation on the Poincaré ball between $\mu,\nu\in\mathcal{P}_p(\mathbb{B}^d)$ as 
\begin{equation}
\hhsw_p^p(\mu,\nu) = \int_{S^{d-1}} W_p^p(B^{\Tilde{v}}_\#\mu, B^{\Tilde{v}}_\#\nu)\ \mathrm{d}\lambda(\Tilde{v}).
\end{equation}

Using that the projections formula between $\mathbb{L}^d$ and $\mathbb{B}^d$ are isometries, we show in the next proposition that the two formulations are equivalent. Hence, we choose in practice the formulation which is the more suitable, either from the nature of data or from a numerical stability viewpoint.

\begin{proposition} \label{prop:equality_hhsw}
    For $p\ge 1$, let $\mu,\nu\in\mathcal{P}_p(\mathbb{B}^d)$ and denote $\Tilde{\mu} = (P_{\mathbb{B}\to\mathbb{L}})_\#\mu$, $\Tilde{\nu} = (P_{\mathbb{B}\to\mathbb{L}})_\#\nu$. Then,
    \begin{align}
        \hhsw_p^p(\mu,\nu) = \hhsw_p^p(\Tilde{\mu},\Tilde{\nu}), \\
        \ghsw_p^p(\mu,\nu) = \ghsw_p^p(\Tilde{\mu}, \Tilde{\nu}).
    \end{align}
\end{proposition}

\begin{proof}
    See \Cref{proof:prop_equality_hhsw}.
\end{proof}

\subsection{Properties}

As particular cases of the Cartan-Hadamard Sliced-Wasserstein discrepancies, Hyperbolic Sliced-Wasserstein (HSW) discrepancies satisfy all the properties derived in \Cref{section:chsw_properties}. In particular, they are pseudo distances. Here, we discuss the connection in the literature with known Radon transforms.

\paragraph{Geodesic Hyperbolic Sliced-Wasserstein.} First, we can connect $\ghsw$ with a Radon transform, defined as in \Cref{section:chsw_properties}, as 
\begin{equation}
    \forall t\in \mathbb{R}, v\in T_{x^0}\mathbb{L}^d\cap S^d,\  Rf(t,v) = \int_{\mathbb{L}^d} f(x) \mathbb{1}_{\{P^v(x) = t\}} \ \mathrm{d}\vol(x),
\end{equation}
where $f\in L^1(\mathbb{L}^d)$. Then, defining it on measures through its dual, and disintegrating \emph{w.r.t} $\lambda$, we can also show that
\begin{equation}
    \forall \mu,\nu \in \mathcal{P}_p(\mathbb{L}^d),\ \ghsw_p^p(\mu,\nu) = \int_{T_{x^0}\mathbb{L}^d \cap S^d} W_p^p\big((R\mu)^v, (R\nu)^v\big)\ \mathrm{d}\lambda(v).
\end{equation}
Now, let us precise the set of integration of this Radon transform.

\begin{proposition}[Set of integration] \label{prop:hsw_integration_set}
    Let $t\in\mathbb{R}$, $v\in T_{x^0}\mathbb{L}^d \cap S^d$, and $z\in\mathrm{span}(x^0,v)\cap\mathbb{L}^d$ the unique point on the geodesic $\mathrm{span}(x^0,v)\cap\mathbb{L}^d$ such that $t^v(z) = t$ where $t^v$ is the isometry defined in \eqref{eq:coordmap}. Then, the integration set of $R$ is,
    \begin{equation}
        \{x\in\mathbb{L}^d,\ P^v(x)=t\} = \mathrm{span}(v_z)^\bot \cap \mathbb{L}^d,
    \end{equation}
    where $v_z = R_z v$ with $R_z$ a rotation matrix 
    in the plan $\mathrm{span}(v,x^0)$ such that $\langle v_z, z\rangle = 0$.
\end{proposition}

\begin{proof}
    See \Cref{proof:prop_hsw_integration_set}.
\end{proof}

From the previous proposition, in the Lorentz model, we see that the Radon transform $R$ integrates over hyperplanes intersected with $\mathbb{L}^d$, which are totally geodesic submanifolds. This is illustrated in the case $d=2$ in \Cref{fig:proj_lorentz}. This corresponds actually to the hyperbolic Radon transform first introduced by \citet{helgason1959differential} and studied more recently for example in \citep{berenstein1999radon, rubin2002radon, berenstein2004totally}. However, to the best of our knowledge, its injectivity over the set of measures has not been studied yet.

\paragraph{Radon transform for HHSW.} We can derive a Radon transform associated to HHSW in the same way. Moreover, the integration set can be intuitively derived as the level set of the Busemann function, since we project on the only point on the geodesic which has the same Busemann coordinate. Since the level sets of the Busemann functions correspond to horospheres, the associate Radon transform is the horospherical Radon transform. It has been for example studied by \citet{bray1999inversion, bray2019radon} on the Lorentz model, and by \citet{casadio2021radon} on the Poincaré ball. Note that it is also known as the Gelfand-Graev transform \citep{gelfand1966generalized}.

\pagebreak

\subsection{Implementation}

\begin{algorithm}[tb]
   \caption{Guideline of GHSW}
   \label{alg:hsw}
    \begin{algorithmic}
       \STATE {\bfseries Input:} $(x_i)_{i=1}^n\sim \mu$, $(y_j)_{j=1}^n\sim \nu$, $(\alpha_i)_{i=1}^n$, $(\beta_j)_{j=1}^n\in \Delta_n$, $L$ the number of projections, $p$ the order
       \FOR{$\ell=1$ {\bfseries to} $L$}
       \STATE Draw $\Tilde{v}\sim\mathrm{Unif}(S^{d-1})$, let $v=[0,\Tilde{v}]$
       \STATE $\forall i,j,\ \hat{x}_i^{\ell}=P^v(x_i)$, $\hat{y}_j^\ell=P^v(y_j)$
       \STATE Compute $W_p^p(\sum_{i=1}^n \alpha_i \delta_{\hat{x}_i^\ell}, \sum_{j=1}^n \beta_j \delta_{\hat{y}_j^\ell})$
       \ENDFOR
       \STATE Return $\frac{1}{L}\sum_{\ell=1}^L W_p^p(\sum_{i=1}^n \alpha_i \delta_{\hat{x}_i^\ell}, \sum_{j=1}^n \beta_j \delta_{\hat{y}_j^\ell})$
    \end{algorithmic}
\end{algorithm}


\begin{wrapfigure}{R}{0.5\textwidth}
    \centering
    \includegraphics[width=\linewidth]{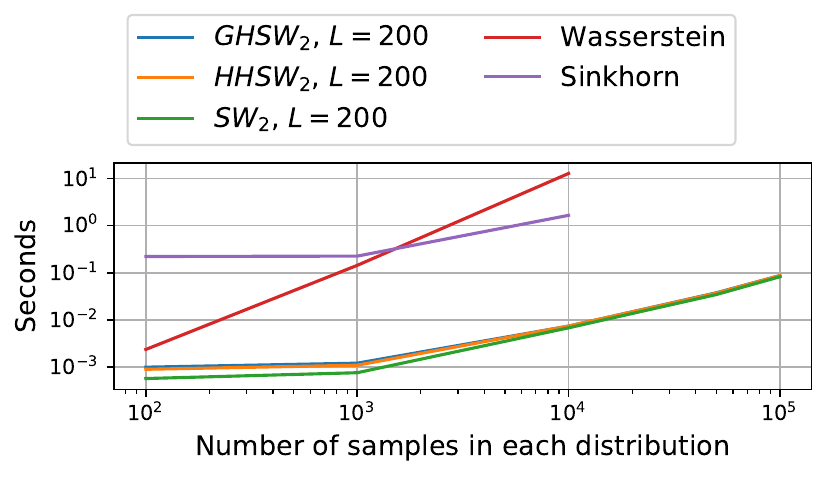}
    \caption{Runtime comparison in log-log scale between Wasserstein and Sinkhorn using the geodesic distance, $\sw_2$, $\ghsw_2$ and $\hhsw_2$ with 200 projections, including the computation time of the cost matrices.}
    \label{fig:runtime_hsw}
\end{wrapfigure}

\paragraph{Implementation.} In practice, we only have access to discrete distributions $\hat{\mu}_n=\sum_{i=1}^n \alpha_i\delta_{x_i}$ and $\hat{\nu}_n = \sum_{i=1}^n \beta_i \delta_{y_i}$ where $(x_i)_i$ and $(y_i)_i$ are sample locations in Hyperbolic space, and $(\alpha_i)_i$ and $(\beta_i)_i$ belong to the simplex $\Delta_n = \{\alpha\in [0,1]^n,\ \sum_{i=1}^n \alpha_i = 1\}$. We approximate the integral by a Monte-Carlo approximation by drawing a finite number $L$ of projection directions $(v_\ell)_{\ell=1}^L$ in $S^{d-1}$. Then, computing GHSW and HHSW amount at first getting the coordinates on $\mathbb{R}$ by using the corresponding projections, and computing the 1D Wasserstein distance between them. We summarize the procedure in Algorithm \ref{alg:hsw} for GHSW.

\paragraph{Complexity.} For both GHSW and HHSW, the projection procedure has a complexity of $O(nd)$. Hence, for $L$ projections, the complexity is in $O\big(Ln(d+\log n)\big)$ which is the same as for SW. In \Cref{fig:runtime_hsw}, we compare the runtime between GHSW, HHSW, SW, Wasserstein and Sinkhorn with geodesic distances in $\mathbb{L}^2$ for $n\in \{10^2,10^3,10^4,5\cdot 10^4, 10^5\}$ samples which are drawn from wrapped normal distributions \citep{nagano2019wrapped}, and $L=200$ projections. We used the POT library \citep{flamary2021pot} to compute SW, Wasserstein and Sinkhorn. We observe  the quasi-linearity complexity of GHSW and HHSW. When we only have a few samples, the cost of the projection is higher than computing the 1D Wasserstein distance, and SW is the fastest.

\section{Application}

In this Section, we perform several experiments which aim at comparing GHSW, HHSW, SWp and SWl. First, we 
study the evolution of the different distances between wrapped normal distributions which move along geodesics. Then, we illustrate the ability to fit distributions on $\mathbb{L}^2$ using gradient flows. Finally, we use HHSW and GHSW for an image classification problem where they are used to fit a prior in the embedding space. We add more information about distributions and optimization in hyperbolic spaces in Appendix \ref{appendix:hyperbolic_space}. Complete details of the experimental settings are reported in Appendix \ref{appendix:hsw_xps}. 

\begin{figure*}[t]
    \centering
    \hspace*{\fill}
    \subfloat[SW on Poincaré (SWp), GHSW]{\label{fig:w_wnd}\includegraphics[width=0.3\linewidth]{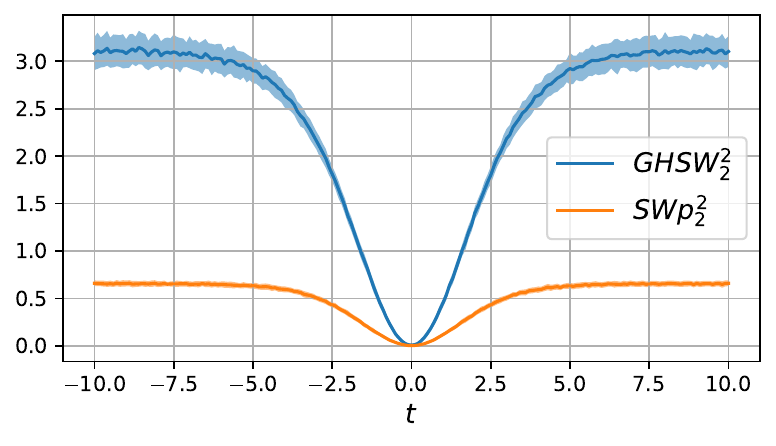}} \hfill
    \subfloat[HHSW and Wasserstein]{\label{fig:hsw_wnd}\includegraphics[width=0.3\linewidth]{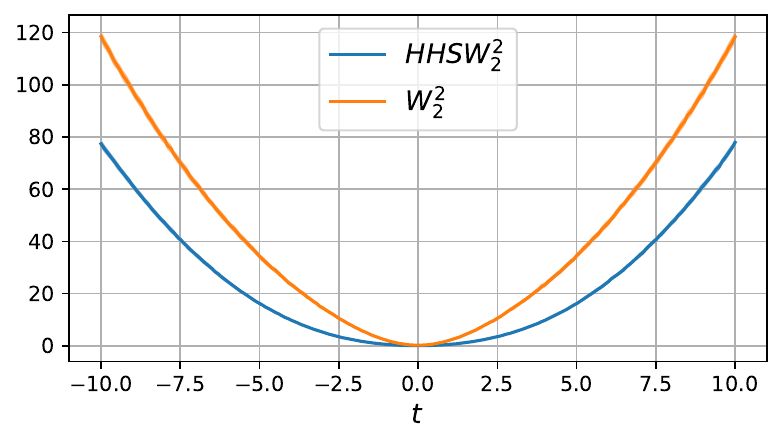}} \hfill
    \subfloat[SW on Lorentz (SWl)]{\label{fig:sw_wnd_lorentz}\includegraphics[width=0.3\linewidth]{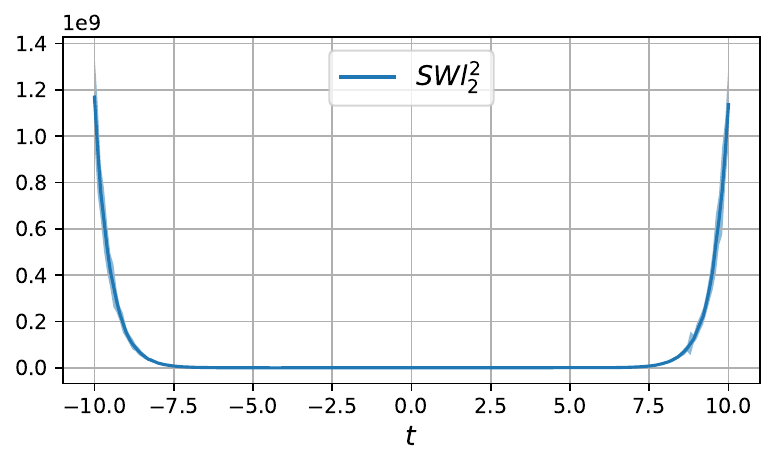}} \hfill
    \hspace*{\fill} \\
    \caption{Comparison of the Wasserstein distance (with the geodesic distance as cost), GHSW, HHSW and SW between Wrapped Normal distributions. We gather the discrepancies together by scale of the values. SW on the Poincaré model has very small values as it operates on the unit ball, while on the Lorentz model, it can take very high values. GHSW returns small values as the geodesic projections tend to project the points close to the origin. HHSW has values which are closer to the geodesic Wasserstein distance as the horospherical projection tends to better keep the distance between points.}
    \label{fig:comparison_wnd}
\end{figure*}

\subsection{Comparisons of the Different Hyperbolical SW Discrepancies} \label{section:hsw_comparison_sw}

On Figure \ref{fig:comparison_wnd}, we compare the evolutions of GHSW, HHSW, SW and Wasserstein with the geodesic distance between Wrapped Normal Distributions (WNDs), where one is centered and the other moves along a geodesic. More precisely, by denoting $\mathcal{G}(\mu,\Sigma)$ a WND, we plot the evolution of the distances between $\mathcal{G}(x^0,I_2)$ and $\mathcal{G}(x_t, I_2)$ where $x_t = \cosh(t)x^0 + \sinh(t)v$ for $t\in [-10,10]$ and $v\in T_{x^0}\mathbb{L}^2\cap S^2$. We observe first that SW on the Lorentz model explodes when the two distributions are getting far from each other. Then, we observe that $\hhsw_2$ has values with a scale similar to $W_2$. We argue that it comes from the observation of \citet{chami2021horopca} which stated that the horospherical projection better preserves the distance between points compared to the geodesic projection. As SWp operates on the unit ball using Euclidean distances, the distances are very small, even for distributions close to the border. Interestingly, as geodesic projections tend to project points close to the origin, GHSW tends also to squeeze the distance between distributions far from the origin. This might reduce numerical instabilities when getting far from the origin, especially in the Lorentz model. This experiment also allows to observe that, at least for WNDs, the indiscernible property is observed in practice as we only obtain one minimum when both measures coincide. Hence, it would suggest that GHSW and HHSW are proper distances.

\subsection{Gradient Flows}

We now assess the ability to learn distributions by minimizing the Hyperbolic SW discrepancies (HSW). We suppose that we have a target distribution $\nu$ from which we have access to samples $(x_i)_{i=1}^n$. Therefore, we aim at learning $\nu$ by solving the following optimization problem: $\min_{\mu} \ \hsw\big(\mu,\frac{1}{n}\sum_{i=1}^n \delta_{x_i}\big)$. We model $\mu$ as a set of $n=500$ particles and propose to perform a Riemannian gradient descent \citep{boumal2023introduction} to learn the distribution. 

\begin{figure}[t]
    \centering
    \includegraphics[width=\columnwidth]{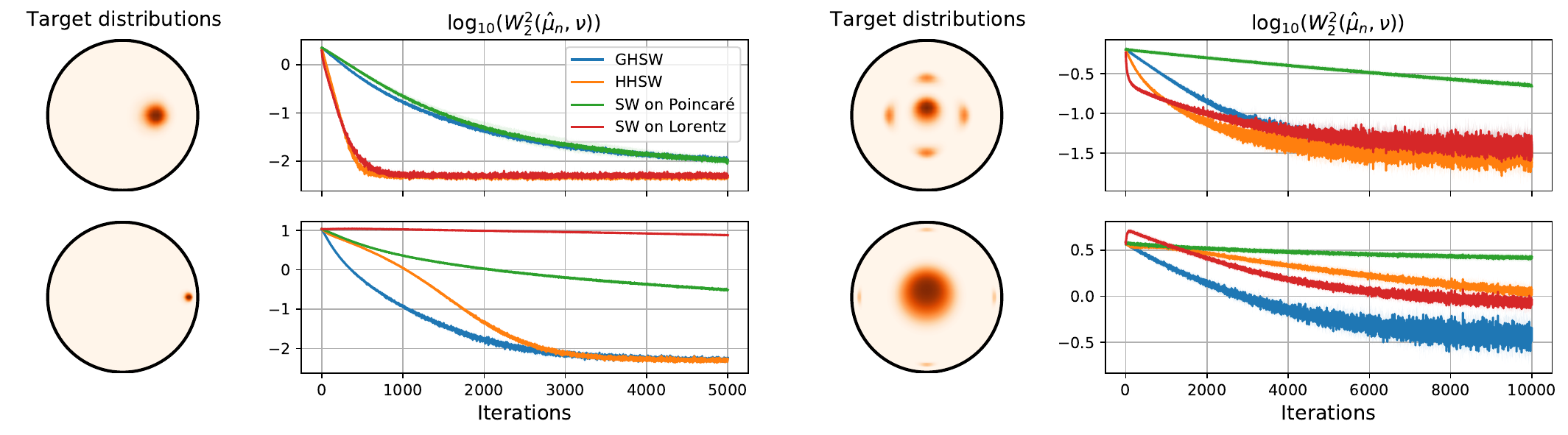}
    \caption{Log 2-Wasserstein between a target and the gradient flow of GHSW, HHSW and SW (averaged over 5 runs).}
    \label{fig:comparison_gradientflows}
\end{figure}

To compare the dynamics of the different discrepancies, we plot on Figure \ref{fig:comparison_gradientflows} the evolution of the exact log $2$-Wasserstein distance, with geodesic distance as ground cost, between the learned distribution at each iteration and the target, with the same learning rate. We use as targets wrapped normal distributions and mixtures of WNDs. For each type of target, we consider two settings, one in which the distribution is close to the origin and another in which the distribution lies closer to the border. We observe different behaviors in the two settings. When the target is lying close to the origin, SWl and HHSW, which present the biggest magnitude, are the fastest to converge. As for distant distributions however, GHSW converges the fastest. Moreover, SWl suffers from many numerical instabilities, as the projections of the gradients do not necessarily lie on the tangent space when points are too far off the origin. This requires to lower the learning rate, and hence to slow down the convergence. Interestingly, SWp is the slowest to converge in both settings.

\subsection{Deep Classification with Prototypes}

We now turn to a classification use case with real world data. Let $\{(x_i,y_i)_{i=1}^n\}$ be a training set where $x_i\in\mathbb{R}^m$ and $y_i\in\{1,\dots,C\}$ denotes a label. \citet{ghadimi2021hyperbolic} perform classification on the Poincaré ball by assigning to each class $c\in\{1,\dots,C\}$ a prototype $p_c\in S^{d-1}$, and then by learning an embedding on the hyperbolic space using a neural network $f_\theta$ followed by the exponential map. Then, by denoting by $z=\exp_0\big(f_\theta(x)\big)$ the output, the loss to be minimized is, for a regularization parameter $s\ge 0$,
\begin{equation}
    \ell(\theta) = \frac{1}{n}\sum_{i=1}^n \Big(B^{p_{y_i}}\big(z_i\big) - sd\cdot \log\big(1-\|z_i\|_2^2\big)\Big).
\end{equation}
The first term is the Busemann function which will draw the representations of $x_i$ towards the prototype assigned to the class $y_i$, while the second term penalizes the overconfidence and pulls back the representation towards the origin. \citet{ghadimi2021hyperbolic} showed that the second term can be decisive to improve the accuracy. Then, the classification of an input is done by solving $y^* = \argmax_{c}\ \langle \frac{z}{\|z\|}, p_c\rangle$.

\looseness=-1 We propose to replace the second term by a global prior on the distribution of the representations. More precisely, we add a discrepancy $D$ between the distribution $(\exp_0\circ f_\theta)_\# p_X$, where $p_X$ denotes the distribution of the training set, and a mixture of $C$ WNDs where the centers are chosen as $(\alpha p_c)_{c=1}^C$, with $(p_c)_c$ the prototypes and $0<\alpha<1$. In practice, we use $D=\ghsw_2^2$, $D=\hhsw_2^2$, $D=\swp_2^2$ and $D=\swl_2^2$ to assess their usability on a real problem. We also compare the results when using $D=W_2^2$ or $D=MMD$ where the MMD is taken with Laplacian kernel \citep{feragen2015geodesic}. Let $(w_i)_{i=1}^n$ be a batch of points drawn from this mixture, then the loss we minimize is
\begin{equation} \label{eq:loss_hsw}
    \ell(\theta) = \frac{1}{n}\sum_{i=1}^n B^p(z_i)  + \lambda D\left(\frac{1}{n}\sum_{i=1}^n \delta_{z_i}, \frac{1}{n}\sum_{i=1}^n \delta_{w_i}\right).
\end{equation}
On Table \ref{tab:acc_pebuse}, we report the classification accuracy on the test set for CIFAR10 and CIFAR100 \citep{krizhevsky2009learning}, using the exact same setting as \citep{ghadimi2021hyperbolic}. We rerun their method, called PeBuse here and we report results averaged over 3 runs. We observe that the proposed penalization outperforms the original method for all the different dimensions.

\begin{table}[t]
    \centering
    \caption{Test Accuracy on deep classification with prototypes (best performance in bold)}
    \small
    \resizebox{\columnwidth}{!}{
        \begin{tabular}{ccccccccc}
            & \multicolumn{4}{c}{CIFAR10} & & \multicolumn{3}{c}{CIFAR100} \\ 
            \toprule
            Dimensions & 2 & 3 & 4 & 5 & & 3 & 5 & 10 \\ \midrule
            PeBuse & $90.64_{\pm 0.06}$ & $90.32_{\pm 0.43}$ & $90.59_{\pm 0.11}$ & $90.55_{\pm 0.09}$ & & $49.28_{\pm 1.95}$ & $53.44_{\pm 0.76}$ & $59.19_{\pm 0.39}$ \\
            GHSW & $91.39_{\pm 0.23}$ & $\bold{91.86}_{\pm 0.38}$ & $91.66_{\pm 0.27}$ & $91.70_{\pm 0.14}$ & & $\bold{53.97}_{\pm 1.35}$ & $60.64_{\pm 0.87}$ & $61.45_{\pm 0.41}$ \\
            HHSW & $91.28_{\pm 0.26}$ & $91.73_{\pm 0.38}$ & $\bold{91.98}_{\pm 0.05}$ & $\bold{92.09_{\pm 0.05}}$ & & $53.88_{\pm 0.06}$ & $\bold{60.69}_{\pm 0.25}$ & $\bold{62.80}_{\pm 0.09}$ \\
            SWp & $\bold{91.84}_{\pm 0.31}$ & $91.74_{\pm 0.05}$ & $91.68_{\pm 0.10}$ & $91.43_{\pm 0.40}$ & & $53.25_{\pm 3.27}$ & $59.77_{\pm 0.81}$ & $60.36_{\pm 1.26}$ \\
            SWl & $91.13_{\pm 0.14}$ & $91.57_{\pm 0.10}$ & $91.74_{\pm 0.12}$ & $91.61_{\pm 0.40}$ & & $53.88_{\pm 0.02}$ & $60.62_{\pm 0.39}$ & $62.30_{\pm 0.23}$ \\
            W & $91.67_{\pm 0.18}$ & $91.82_{\pm 0.19}$ & $91.83_{\pm 0.21}$ & $91.43_{\pm 0.40}$ & & $50.07_{\pm 4.58}$ & $57.49_{\pm 0.94}$ & $58.82_{\pm 1.66}$ \\
            MMD & $91.47_{\pm 0.10}$ & $91.65_{\pm 0.17}$ & $91.68_{\pm 0.09}$ & $91.54_{\pm 0.09}$ & & $50.59_{\pm 4.44}$ & $58.10_{\pm 0.73}$ & $58.91_{\pm 0.91}$ \\
            \bottomrule
        \end{tabular}
    }
    \vspace{-5pt}
    \label{tab:acc_pebuse}
\end{table}

\section{Conclusion and Discussion}

In this work, we propose different Sliced-Wasserstein discrepancies between distributions lying in Hyperbolic spaces. In particular, we introduce two new SW discrepancies which are intrinsically defined on Hyperbolic spaces. They are built by first identifying a closed-form for the Wasserstein distance on geodesics, and then by using different projections on the geodesics. We compare these metrics on multiple tasks such as sampling and image classification. We observe that, while Euclidean SW in the ambient space still works, it suffers from either slow convergence on the Poincaré ball or numerical instabilities on the Lorentz model when distributions are lying far from the origin. On the other hand, geodesic versions exhibit the same complexity and converge generally better for gradient flows. Further works will look into other tasks where hyperbolic embeddings and distributions have been shown to be beneficial, such as persistent diagrams \citep{carriere2017sliced, kyriakis2021learning}. Besides further applications, proving that these discrepancies are indeed distances, and deriving statistical results are interesting directions of work. One might also consider different subspaces on which to project, such as horocycles which are circles of infinite radius and which can be seen as another analog object to lines in Hyperbolic spaces \citep{casadio2021radon}. Another direction of research could be to define sliced distances on generalizations of hyperbolic spaces such as pseudo-Riemannian spaces known as ultrahyperbolic spaces \citep{law2021ultrahyperbolic} or Finsler manifolds such as the Siegel space \citep{lopez2021symmetric} or the Hilbert simplex \citep{nielsen2022non}.


\clearemptydoublepage
\cleartooddpage[\thispagestyle{empty}]
\chapter{Sliced-Wasserstein on Symmetric Positive Definite Matrices} \label{chapter:spdsw}

{
    \hypersetup{linkcolor=black} 
    \minitoc 
}

This chapter is based on \citep{bonet2023sliced} and studies particular cases of Hadamard manifolds of Symmetric Positive Definite matrices applied to magneto and encephalogram (M/EEG) signals. Indeed, when dealing with electro or magnetoencephalography records, many supervised prediction tasks are solved by working with covariance matrices to summarize the signals.
Learning with these matrices requires using Riemanian geometry to account for their structure. We propose a new method to deal with distributions of covariance matrices and demonstrate its computational efficiency on M/EEG multivariate time series. More specifically, we define a Sliced-Wasserstein distance between measures of Symmetric Positive Definite matrices that comes with strong theoretical guarantees. For the numerical computation, we propose a simple way for uniform sampling of the unit-norm SDP matrix set and the projection along geodesics. Then, we take advantage of its properties and kernel methods to apply this distance to brain-age prediction from MEG data and compare it to state-of-the-art algorithms based on Riemannian geometry. Finally, we show that it is an efficient surrogate to the Wasserstein distance in domain adaptation for Brain Computer Interface applications.

\section{Introduction}

Magnetoencephalography and electroencephalography (M/EEG) are non-invasive techniques for recording the electrical activity of the brain \citep{hamalainen1993magnetoencephalography}.
The data consist of multivariate time series output by sensors placed around the head, which capture the intensity of the magnetic or electric field with high temporal resolution.
Those measurements provide information on cognitive processes as well as the biological state of a subject.

Successful Machine Learning (ML) techniques that deal with M/EEG data often rely on covariance matrices estimated from band-passed filtered signals in several frequency bands \citep{blankertz2007optimizing}. The main difficulty that arises when processing such covariance matrices is that the set of Symmetric Positive Definite (SPD) matrices is not a linear space, but a Riemannian manifold \citep{bhatia2009positive, bridson2013metric}. Therefore, specific algorithms have to be designed to take into account the non Euclidean structure of the data.
The usage of Riemannian geometry on SPD matrices has become increasingly popular in the ML community \citep{huang2017riemannian, chevallier2017kernel, ilea2018covariance, brooks2019riemannian}. 
In particular, these tools have proven to be very effective on prediction tasks with M/EEG data in Brain Computer Interface (BCI) applications \citep{barachant2011multiclass, barachant2013classification, gaur2018multi} or more recently in brain-age prediction \citep{sabbagh2019manifold, sabbagh2020predictive, engemann2022reusable}. As covariance matrices sets from M/EEG data are often modeled as samples from a probability distribution -- for instance in domain adaptation for BCI \citep{yair2019domain} -- it is of great interest to develop efficient tools that work directly on those distributions.

Optimal transport (OT) \citep{villani2009optimal,peyre2019computational} provides a powerful theoretical framework and computational toolbox to compare probability distributions while respecting the geometry of the underlying space. It is well defined on Riemannian manifolds \citep{mccann2001polar, cui2019spherical,alvarez2020unsupervised} and in particular on the space of SPD matrices that is considered in M/EEG learning tasks \citep{brigant2018optimal, yair2019domain, ju2022deep}. To alleviate the computational burden of the original OT problem, we propose to leverage the constructions of Sliced-Wasserstein distances proposed in \Cref{chapter:sw_hadamard} in the particular case of Symmetric Positive Definite matrices.

\paragraph{Contributions.} In order to benefit from the advantages of SW in the context of M/EEG, we propose new SW distances on the manifold of SPD matrices endowed by two different metrics. First, we study the case of the Affine-Invariant metric. Then, we study in more detail the case of the Log-Euclidean metric and introduce $\lespdsw$. We derive theoretical results, including topological, statistical, and computational properties. In particular, we prove that $\lespdsw$ is a distance topologically equivalent to the Wasserstein distance in this context. We extend the distribution regression with SW kernels to the case of SPD matrices, apply it to brain-age regression with MEG data, and show that it performs better than other methods based on Riemannian geometry. Finally, we show that $\lespdsw$ is an efficient surrogate to the Wasserstein distance in domain adaptation for BCI.

\section{Background on SPD matrices} \label{sec:bg_spd}

Let $S_d(\mathbb{R})$ be the set of symmetric matrices of $\mathbb{R}^{d \times d}$, and $S_d^{++}(\mathbb{R})$ be the set of SPD matrices of $\mathbb{R}^{d \times d}$, \emph{i.e.} matrices $M\in S_d(\mathbb{R})$ satisfying
\begin{equation}
    \forall x\in\mathbb{R}^d\setminus\{0\},\ x^T M x > 0.
\end{equation}
$S_d^{++}(\mathbb{R})$ is a Riemannian manifold \citep{bhatia2009positive}, meaning that it behaves locally as a linear space, called a tangent space. 
Each point $M \in S_d^{++}(\mathbb{R})$ defines a tangent space $T_M S_d^{++}(\mathbb{R})$, which can be given an inner product $\langle \cdot, \cdot \rangle_M : T_M S_d^{++}(\mathbb{R}) \times T_M S_d^{++}(\mathbb{R}) \rightarrow \mathbb{R}$, and thus a norm.
The choice of this inner-product induces different geometry on the manifold.
One example is the geometric and Affine-Invariant (AI) metric \citep{pennec2006riemannian}, where the inner product is defined as
\begin{equation}
    \forall M \in S_d^{++}(\mathbb{R}),\ A,B\in T_M S_d^{++}(\mathbb{R}),\ \langle A,B\rangle_M = \tr(M^{-1}AM^{-1}B).
\end{equation}
Denoting by $\tr$ the Trace operator, the corresponding geodesic distance $d_{AI}(\cdot,\cdot)$ is given by
\begin{equation}
    \forall X, Y \in S_d^{++}(\mathbb{R}),\ d_{AI}(X,Y) = \sqrt{\tr\big(\log(X^{-1}Y)^2\big)}.
\end{equation}
An interesting property justifying the use of the Affine-Invariant metric is that $d_{AI}$ satisfies the affine-invariant property: for any $g\in GL_d(\mathbb{R})$, where $GL_d(\mathbb{R})$ denotes the set of invertible matrices in $\mathbb{R}^{d\times d}$,
\begin{equation}
    \forall X,Y\in S_d^{++}(\mathbb{R}),\ d_{AI}(g\cdot X, g\cdot Y) = d_{AI}(X,Y),
\end{equation}
where $g\cdot X = gXg^T$.

Another example is the Log-Euclidean (LE) metric \citep{arsigny2005fast,arsigny2006log} for which, 
\begin{equation}
        \forall M \in S_d^{++}(\mathbb{R}),\ A,B\in T_M S_d^{++}(\mathbb{R}),\ \langle A,B\rangle_M = \langle D_M\log A, D_M \log B\rangle,
\end{equation}
with $\log$ the matrix logarithm and $D_M\log A$ the directional derivative of the $\log$ at $M$ along $A$ \citep{huang2015log}.
This definition provides another geodesic distance \citep{arsigny2006log}
\begin{equation}
    \forall X, Y \in S_d^{++}(\mathbb{R}), \ d_{LE}(X,Y) = \|\log X - \log Y\|_F,
\end{equation}
which is simply an Euclidean distance in $S_d(\mathbb{R})$ as $\log$ is a diffeomorphism from $S_d^{++}(\mathbb{R})$ to $S_d(\mathbb{R})$, whose inverse is $\exp$. For $S_d^{++}(\mathbb{R})$, note that $T_MS_d^{++}(\mathbb{R})$ is diffeormorphic with $S_d(\mathbb{R})$ \citep[Theorem 3]{arsigny2005fast}. For the AI metric, geodesic lines passing through $X$ and $Y\in S_d^{++}(\mathbb{R})$ are of the form
\begin{equation}
    \forall t\in \mathbb{R},\ \gamma(t) = X^{\frac12} \exp\big(t\log(X^{-\frac12}YX^{-\frac12})\big)X^{\frac12}.
\end{equation}
For the LE metric, they are of the form
\begin{equation}
    \forall t\in\mathbb{R},\ \gamma(t) = \exp\big((1-t)\log X + t\log Y\big).
\end{equation}

$S_d^{++}(\mathbb{R})$ endowed with the Affine-Invariant metric is a Riemannian manifold of non-constant and non-positive curvature \citep{bhatia2009positive, bridson2013metric} while $S_d^{++}(\mathbb{R})$ with the Log-Euclidean metric is of constant null curvature \citep{pennec2020manifold}. In particular, The Log-Euclidean distance is actually a lower bound of the Affine-Invariant distance, and they coincide when the matrices commute. The Log-Euclidean metric can actually be seen as a good first order approximation of the Affine-Invariant metric \citep{arsigny2005fast,pennec2020manifold} which motivated the proposal of this metric. Notably, they share the same geodesics passing through the identity \citep[Section 3.6.1]{pennec2020manifold}, which are of the form $t\mapsto \exp(tA)$ for $A\in S_d(\mathbb{R})$. To span all such geodesics, we can restrict to $A$ with unit Frobenius norm, \emph{i.e.} $\| A \|_F = 1$.




\section{Sliced-Wasserstein on SPD Matrices} \label{section:spdsw}

In this Section, we introduce SW discrepancies on SPD matrices and provide a theoretical analysis of their properties and behavior. Following the general framework introduced in \Cref{chapter:sw_hadamard}, we first discuss how to project on geodesics passing through the origin $I_d$ with the different metrics. Then we define the different Sliced-Wasserstein distances, and present the additional theoretical properties which are not deduced from the general framework. 

\subsection{Projections on Geodesics} \label{section:proj_spdsw}

As origin, we choose the identity $I_d$ and we aim to project on geodesics passing through $I_d$ which are of the form $\mathcal{G}_A = \{\exp(tA),\ t\in\mathbb{R}\}$ where $A\in S_d(\mathbb{R})$. We derive the different projections in closed-form first when $S_d^{++}(\mathbb{R})$ is endowed with the Log-Euclidean metric, and then when it is endowed with the Affine-Invariant metric.

\paragraph{With Log-Euclidean Metric.} First, we derive in \Cref{prop:spdsw_geodesic_projection} the closed-form of the geodesic projection on $\mathcal{G}_A$, which we recall is defined as
\begin{equation}
    \forall M\in S_d^{++}(\mathbb{R}),\ \geodproj^{\mathcal{G}_A}(M) = \argmin_{X\in\mathcal{G}_A}\ d_{LE}(X,M).
\end{equation}

\begin{proposition} \label{prop:spdsw_geodesic_projection}
     Let $A\in S_d(\mathbb{R})$ with $\|A\|_F = 1$, and let $\mathcal{G}_A$ be the associated geodesic line.
     Then, for any $M\in S_d^{++}(\mathbb{R})$, the geodesic projection on $\mathcal{G}_A$ is
    \begin{equation}
        \geodproj^{\mathcal{G}_A}(M) = \exp\big(\tr(A\log M) A\big).
    \end{equation}
\end{proposition}

\begin{proof}
    See \Cref{proof:prop_spdsw_geodesic_projection}.
\end{proof}

Then, we also provide the coordinate on the geodesic $\mathcal{G}_A$, which we recall is obtained by giving an orientation to $\mathcal{G}_A$ and computing the distance between $\geodproj^{\mathcal{G}_A}(M)$ and the origin $I_d$, as follows
\begin{equation}
    \label{eqn:coordinate}
    \forall M \in S_d^{++}(\mathbb{R}),\ P^A(M) = \mathrm{sign}(\langle \log \geodproj^{\mathcal{G}_A}(M), A\rangle_F) d_{LE}(\geodproj^{\mathcal{G}_A}(M), I_d).
\end{equation}
The closed-form expression is given by \Cref{prop:spdsw_geodesic_coordinate}.

\begin{proposition} \label{prop:spdsw_geodesic_coordinate}
     Let $A\in S_d(\mathbb{R})$ with $\|A\|_F = 1$, and let $\mathcal{G}_A$ be the associated geodesic line.
     Then, for any $M\in S_d^{++}(\mathbb{R})$, the geodesic coordinate on $\mathcal{G}_A$ is
    \begin{equation}
        P^A(M) = \langle A, \log M\rangle_F = \mathrm{Tr}(A\log M).
    \end{equation}
\end{proposition}

\begin{proof}
    See \Cref{proof:prop_spdsw_geodesic_coordinate}.
\end{proof}

These two properties give a closed-form expression for the Riemannian equivalent of one-dimensional projection in a Euclidean space.
Note that coordinates on the geodesic might also be found using Busemann coordinates, and that they actually coincide here (up to a sign) as shown in the following proposition. This is due to the null curvature of the space, in which case horospheres and hyperplanes coincide.

\begin{proposition}[Busemann coordinates] \label{prop:spdsw_busemann_coords}
    Let $A\in S_d(\mathbb{R})$ such that $\|A\|_F=1$, and let $\mathcal{G}_A$ be the associated geodesic line. Then, the Busemann function associated to $\mathcal{G}_A$ is defined as 
    \begin{equation}
        \forall M \in S_d^{++}(\mathbb{R}),\ B^A(M) = - \mathrm{Tr}(A\log M).
    \end{equation}
\end{proposition}

\begin{proof}
    See \Cref{proof:prop_spdsw_busemann_coords}.
\end{proof}

\begin{figure}[t]
    \centering
    \includegraphics[width=0.7\columnwidth]{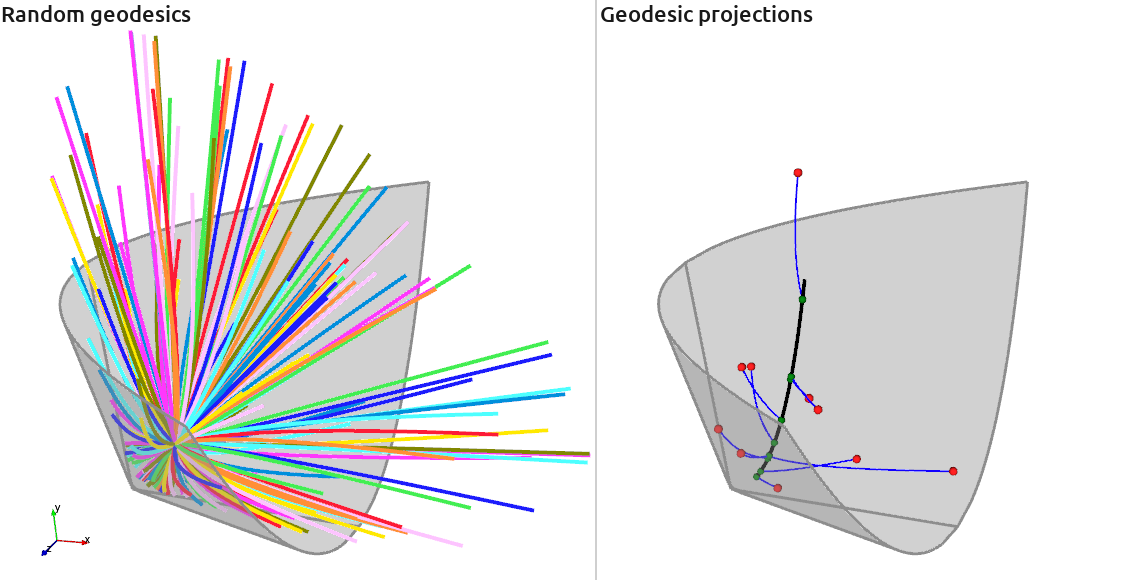}
    \caption{({\bf Left}) Random geodesics drawn in  $S_2^{++}(\mathbb{R})$. ({\bf Right}) Projections (green points) of covariance matrices (depicted as red points) over one geodesic (in black) passing through $I_2$ along the Log-Euclidean geodesics (blue lines).}
    \label{fig:spdsw_proj}
\end{figure}

In \Cref{fig:spdsw_proj}, we illustrate the projections of matrices $M\in S_2^{++}(\mathbb{R})$ embedded as vectors $(m_{11},m_{22},m_{12})\in\mathbb{R}^3$.
$S_2^{++}(\mathbb{R})$ is an open cone and we plot the projections of random SPD matrices on geodesics passing through $I_2$.

\paragraph{With Affine-Invariant Metric.}

For the Affine-Invariant case, to the best of our knowledge, there is no closed-form for the geodesic projection on $\mathcal{G}_A$, the difficulty being that the matrices do not necessarily commute. Hence, we will discuss here the horospherical projection which can be obtained with the Busemann function. For $A\in S_d(\mathbb{R})$ such that $\|A\|_F=1$, denoting $\gamma_A:t\mapsto \exp(tA)$ the geodesic line passing through $I_d$ with direction $A$, the Busemann function $B^A$ associated to $\gamma_A$ writes as 
\begin{equation}
    \forall M\in S_d^{++}(\mathbb{R}),\ B^A(M) = \lim_{t\to\infty}\ \big(d_{AI}\big(\exp(tA),M\big)-t\big).
\end{equation} 
Contrary to the Log-Euclidean case, we cannot directly compute this quantity by expanding the distance since $\exp(-tA)$ and $M$ are not necessarily commuting. The main idea to solve this issue is to first find a group $G\subset GL_d(\mathbb{R})$ which will leave the Busemann function invariant. Then, we can find an element of this group which will project $M$ on the space of matrices commuting with $\exp(A)$. This part of the space is of null curvature, \emph{i.e.} it is isometric to an Euclidean space. In this case, we can compute the Busemann function as in \Cref{prop:spdsw_busemann_coords} as the matrices are commuting. Hence, the Busemann function is of the form
\begin{equation}
    B^A(M) = -\left\langle A, \log \big(\pi_A (M)\big)\right\rangle_F, 
\end{equation}
where $\pi_A$ is a projection on the space of commuting matrices. In the next paragraph, we detail how we can proceed to obtain $\pi^A$.

When $A$ is diagonal with sorted values such that $A_{11} > \dots > A_{dd}$, then the group leaving the Busemann function invariant is the set of upper triangular matrices with ones on the diagonal \citep[II. Proposition 10.66]{bridson2013metric}, \emph{i.e.} for such matrix $g$, $B^A(M) = B^A(gMg^T)$. If the points are sorted in increasing order, then the group is the set of lower triangular matrices. Let's note $G_U$ the set of upper triangular matrices with ones on the diagonal. For a general $A\in S_d(\mathbb{R})$, we can first find an appropriate diagonalization $A=P\Tilde{A}P^T$, where $\Tilde{A}$ is diagonal sorted, and apply the change of basis $\Tilde{M}=P^TMP$ \citep{fletcher2009computing}. We suppose that all the eigenvalues of $A$ have an order of multiplicity of one, which is a reasonable hypothesis as we will see in \Cref{lemma:uniform_distribution}. By the affine-invariance property, the distances do not change, \emph{i.e.} $d_{AI}(\exp(tA),M) = d_{AI}(\exp(t\Tilde{A}),\Tilde{M})$ and hence, using the definition of the Busemann function, we have that $B^A(M) = B^{\Tilde{A}}(\Tilde{M})$. Then, we need to project $\Tilde{M}$ on the space of matrices commuting with $\exp(\Tilde{A})$ which we denote $F(A)$. By \citet[II. Proposition 10.67]{bridson2013metric}, this space corresponds to the diagonal matrices. Moreover, by \citet[II. Proposition 10.69]{bridson2013metric}, there is a unique pair $(g,D)\in G_U\times F(A)$ such that $\Tilde{M} = gDg^T$, and therefore, we can note $\pi_A(\Tilde{M})=D$. This decomposition actually corresponds to a UDU decomposition. If the eigenvalues of $A$ are sorted in increasing order, this would correspond to a LDL decomposition.

For more details about the Busemann function on the Affine-invariant space, we refer to \citet[Section II.10]{bridson2013metric} and \citet{fletcher2009computing, fletcher2011horoball}.

\subsection{Definitions of Sliced-Wasserstein Distances}

We are now ready to define Sliced-Wasserstein distances on both the Log-Euclidean space and the Affine-Invariant space.

\paragraph{SPDSW.} We start by defining an SW distance on the space of measures 
\begin{equation*}
    \mathcal{P}_p\big(S_d^{++}(\mathbb{R})\big)=\{\mu\in\mathcal{P}\big(S_d^{++}(\mathbb{R})\big),\ \int d_{LE}(X, M_0)^p\ \mathrm{d}\mu(X)<\infty,\ M_0\in S_d^{++}(\mathbb{R})\}    
\end{equation*}
which we call $\lespdsw$.

\begin{definition}
    \label{def:swspd}
    Let $\lambda_S$ be the uniform distribution on $\{A\in S_d(\mathbb{R}),\ \|A\|_F=1\}$. Let $p\ge 1$ and $\mu,\nu\in\mathcal{P}_p\big(S_d^{++}(\mathbb{R})\big)$, then the $\lespdsw$ discrepancy is defined as 
    \begin{equation}
        \lespdsw_p^p(\mu,\nu) = \int_{S_d(\mathbb{R})} W_p^p(P^A_\#\mu, P^A_\#\nu)\ \mathrm{d}\lambda_S(A).
    \end{equation}
\end{definition}

The coordinate of the projection on the geodesic $\mathcal{G}_A$ is provided by $P^A(\cdot) = \tr(A \log \cdot)$ defined in \Cref{prop:spdsw_geodesic_coordinate}.
The Wasserstein distance is easily computed using order statistics, and this leads to a natural extension of the SW distance in $S_d^{++}(\mathbb{R})$.
There exists a strong link between $\sw$ on distributions in $\mathbb{R}^{d \times d}$ and $\lespdsw$.
Indeed, \Cref{prop:equivalence_swlog} shows that $\lespdsw$ is equal to a variant of $\sw$ where projection parameters are sampled from unit norm matrices in $S_d(\mathbb{R})$ instead of the unit sphere, and where the distributions are pushed forward by the $\log$ operator.
\begin{proposition} \label{prop:equivalence_swlog}
    Let $\Tilde{\mu}, \Tilde{\nu} \in\mathcal{P}_p(S_d(\mathbb{R}))$, and $t^A(B) = \tr(A^T B)$ for $A,B\in S_d(\mathbb{R})$.
    We define
    \begin{equation}
        \symsw_p^p(\Tilde{\mu}, \Tilde{\nu}) = \int_{S_d(\mathbb{R})} W_p^p(t^A_\# \Tilde{\mu} ,t^A_\# \Tilde{\nu} )\ \mathrm{d}\lambda_S(A).
    \end{equation}
    Then, for $\mu,\nu\in\mathcal{P}_p(S_d^{++}(\mathbb{R}))$,
    \begin{equation}
        \lespdsw^p_p(\mu, \nu) = \symsw_p^p(\log_\#\mu, \log_\#\nu).
    \end{equation}
\end{proposition}
\begin{proof}
    See \Cref{proof:prop_equivalence_swlog}.
\end{proof}
Thus, it seems natural to compare the results obtained with $\lespdsw$ to the Euclidean counterpart $\logsw = \sw(\log_\# \cdot, \log_\# \cdot)$ where the distributions are made of projections in the $\log$ space and where the sampling is done with the uniform distribution on the sphere. This variant will provide an ablation over the integration set.

\paragraph{HSPDSW.} Similarly, using the horospherical projection introduced in the last Section, we can define a horospherical Sliced-Wasserstein distance on the space of measures on $S_d^{++}(\mathbb{R})$ endowed by the Affine-Invariant metric, \emph{i.e.} $\mathcal{P}_p^{AI}\big(S_d^{++}(\mathbb{R})\big) = \left\{ \mu \in \mathcal{P}\big(S_d^{++}(\mathbb{R})\big),\ \int d_{AI}(X,M_0)^p\ \mathrm{d}\mu(X) < \infty,\ M_0\in S_d^{++}(\mathbb{R})\right\}$.

\begin{definition}
    \label{def:aiswspd}
    Let $\lambda_S$ be the uniform distribution on $\{A\in S_d(\mathbb{R}),\ \|A\|_F=1\}$. Let $p\ge 1$ and $\mu,\nu\in\mathcal{P}_p^{AI}\big(S_d^{++}(\mathbb{R})\big)$, then the $\aispdsw$ discrepancy is defined as 
    \begin{equation}
        \aispdsw_p^p(\mu,\nu) = \int_{S_d(\mathbb{R})} W_p^p(B^A_\#\mu, B^A_\#\nu)\ \mathrm{d}\lambda_S(A),
    \end{equation}
    where $B^A(M) = -\tr\big(A \log (\pi_A(M))\big)$ with $\pi^A$ the projection derived in \Cref{section:proj_spdsw}.
\end{definition}


\paragraph{Sampling from $\lambda_S$.}

As shown by the definitions, being able to sample from $\lambda_S$ is one of the cornerstones of the computation of SPDSW. In \Cref{lemma:uniform_distribution}, we propose a practical way of uniformly sampling a symmetric matrix $A$. More specifically, we sample an orthogonal matrix $P$ and a diagonal matrix $D$ of unit norm and compute $A=PDP^T$ which is a symmetric matrix of unit norm. This is equivalent to sampling from $\lambda_S$ as the measures are equal up to a normalization factor $d!$ which represents the number of possible permutations of the columns of $P$ and $D$ for which $PDP^T=A$.
\begin{lemma} \label{lemma:uniform_distribution}
    Let $\lambda_O$ be the uniform distribution on $\mathcal{O}_d = \{P \in \mathbb{R}^{d \times d},\ P^TP = PP^T = I\}$ (Haar distribution), and $\lambda$ be the uniform distribution on $S^{d-1} = \{\theta \in \mathbb{R}^d,\ \|\theta\|_2=1\}$.
    Then $\lambda_S \in \mathcal{P}(S_d(\mathbb{R}))$, defined such that $\forall \ A = P \mathrm{diag}(\theta) P^T \in S_d(\mathbb{R})$, $\mathrm{d}\lambda_S(A) = d! \ \mathrm{d}\lambda_O(P) \mathrm{d}\lambda(\theta)$, is the uniform distribution on $\{A\in S_d(\mathbb{R}),\ \|A\|_F=1\}$.
\end{lemma}
\begin{proof}
    See \Cref{proof:lemma_uniform_distribution}.
\end{proof}
Note that since we sample the eigenvalues from the uniform distribution on $S^{d-1}$, the values are all different almost surely. Hence, the hypothesis made in \Cref{section:proj_spdsw} that all eigenvalues have an order of multiplicity of 1 is justified.

\subsection{Properties of SPDSW} \label{section:properties_spdsw}

As both constructions follow the framework of \Cref{chapter:sw_hadamard}, they satisfy the same properties derived in this chapter and we do not restate them. Notably, they are both pseudo-distances which can be embedded in Hilbert spaces, have a sample complexity independent of the dimension and a projection complexity with the same rate of Monte-Carlo estimators. In this Section, we add theoretical results obtained for $\lespdsw$ which rely on the null curvature of the Log-Euclidean space, and which notably allows to show well known results of the Euclidean SW distance: distance properties and metrization of the weak convergence.

\paragraph{Topology.} Following usual arguments which are valid for any sliced divergence with any projection, we can show that both $\lespdsw$ and $\aispdsw$ are pseudo-distances. However, $S_d^{++}(\mathbb{R})$ with the Log-Euclidean metric is of  null sectional curvature \citep{arsigny2005fast,xu2022unsupervised} and we have access to a diffeomorphism to a Euclidean space -- the $\log$ operator.
This allows us to show that $\lespdsw$ is a distance in \Cref{prop:spdsw_distance}.
\begin{theorem} \label{prop:spdsw_distance}
    Let $p\ge 1$, then $\lespdsw_p$ is a finite distance on $\mathcal{P}_p\big(S_d^{++}(\mathbb{R})\big)$.
\end{theorem}
\begin{proof}
    See \Cref{proof:prop_spdsw_distance}.
\end{proof}
For $\aispdsw$, as the projection $\log \circ \pi_A$ is not a diffeomorphism, whether the indiscernible property holds or not remains an open question and could be studied via the related Radon transform.

An important property which justifies the use of the SW distance in place of the Wasserstein distance in the Euclidean case is that they both metrize the weak convergence \citep{bonnotte2013unidimensional}.
We show in \Cref{prop:spdsw_weakcv} that this is also the case with $\lespdsw$ in $\mathcal{P}_p\big(S_d^{++}(\mathbb{R})\big)$. 
\begin{theorem} \label{prop:spdsw_weakcv}
    For $p\ge 1$, $\lespdsw_p$ metrizes the weak convergence, \emph{i.e.} for $\mu\in\mathcal{P}_p\big(S_d^{++}(\mathbb{R})\big)$ and a sequence $(\mu_k)_k$ in $\mathcal{P}_p\big(S_d^{++}(\mathbb{R})\big)$, $\lim_{k\to\infty}\lespdsw_p(\mu_k,\mu) = 0$ if and only if $(\mu_k)_k$ converges weakly to $\mu$.
\end{theorem}
\begin{proof}
    See \Cref{proof:prop_spdsw_weakcv}.
\end{proof}
Moreover, $\lespdsw_p$ and $W_p$ -- the $p$-Wasserstein distance with Log-Euclidean ground cost -- are also weakly equivalent on compactly supported measures of $\mathcal{P}_p\big(S_d^{++}(\mathbb{R})\big)$, as demonstrated in \Cref{prop:spdsw_bound}.
\begin{theorem} \label{prop:spdsw_bound}
    Let $p\ge 1$, let $\mu,\nu\in\mathcal{P}_p\big(S_d^{++}(\mathbb{R})\big)$.
    Then
    \begin{equation}
        \lespdsw_p^p(\mu,\nu) \le c_{d,p}^p W_p^p(\mu,\nu),
    \end{equation}
    where $c_{d,p}^p = \frac{1}{d}\int \|\theta\|_p^p\ \mathrm{d}\lambda(\theta)$.
    Let $R>0$ and $B(I,R)=\{M\in S_d^{++}(\mathbb{R}),\ d_{LE}(M, I_d)=\|\log M\|_F \le R\}$ be a closed ball. Then there exists a constant $C_{d,p,R}$ such that for all $\mu,\nu\in\mathcal{P}_p\big(B(I,R)\big)$,
    \begin{equation}
        W_p^p(\mu,\nu) \le C_{d,p,R} \lespdsw_p(\mu,\nu)^{\frac{2}{d(d+1)+2}}.
    \end{equation}
\end{theorem}
\begin{proof}
    See \Cref{proof:prop_spdsw_bound}.
\end{proof}
The theorems above highlight that $\lespdsw_p$ behaves similarly to $W_p$ on $\mathcal{P}_p\big(S_d^{++}(\mathbb{R})\big)$. Thus, it is justified to use $\lespdsw_p$ as a surrogate of Wasserstein and to take advantage of the statistical and computational benefits that we present in the next Section.

We note that we recover the same constant $c_{d,p}^p$ in the upper bound as for the Euclidean SW distance \citep{bonnotte2013unidimensional,candau_tilh}.
In particular, for $p=2$, we have
\begin{equation}
    \lespdsw_2^2(\mu,\nu) \le \frac{1}{d} W_2^2(\mu,\nu).
\end{equation}
Moreover, denoting by $\Tilde{W}_p$ the $p$-Wasserstein distance with Affine-Invariant ground cost, we have
\begin{equation}
    \lespdsw_p^p(\mu,\nu) \le c_{d,p}^p W_p^p(\mu,\nu) \le c_{d,p}^p \Tilde{W}_p^p(\mu,\nu),
\end{equation}
since the Log-Euclidean geodesic distance is a lower bound of the Affine-Invariant one \citep[Theorem 6.14]{bhatia2009positive}.

\begin{algorithm}[tb]
   \caption{Computation of $\lespdsw$}
   \label{alg:spdsw}
    \begin{algorithmic}
       \STATE {\bfseries Input:} $(X_i)_{i=1}^n\sim \mu$, $(Y_j)_{j=1}^m\sim \nu$, $L$ the number of projections, $p$ the order
       \FOR{$\ell=1$ {\bfseries to} $L$}
       \STATE Draw $\theta\sim\mathrm{Unif}(S^{d-1})=\lambda$
       \STATE Draw $P\sim \mathrm{Unif}(O_d(\mathbb{R}))=\lambda_O$
       \STATE $A=P\mathrm{diag}(\theta)P^T$
       \STATE $\forall i,j,\ \hat{X}_i^{\ell}=P^A(X_i)$, $\hat{Y}_j^\ell=P^A(Y_j)$
       \STATE Compute $W_p^p(\frac{1}{n}\sum_{i=1}^n \delta_{\hat{X}_i^\ell}, \frac{1}{m}\sum_{j=1}^m \delta_{\hat{Y}_j^\ell})$
       \ENDFOR
       \STATE Return $\frac{1}{L}\sum_{\ell=1}^L W_p^p(\frac{1}{n}\sum_{i=1}^n \delta_{\hat{X}_i^\ell}, \frac{1}{m}\sum_{j=1}^m \delta_{\hat{Y}_j^\ell})$
    \end{algorithmic}
\end{algorithm}

\begin{algorithm}[tb]
   \caption{Computation of $\aispdsw$}
   \label{alg:hspdsw}
    \begin{algorithmic}
       \STATE {\bfseries Input:} $(X_i)_{i=1}^n\sim \mu$, $(Y_j)_{j=1}^m\sim \nu$, $L$ the number of projections, $p$ the order
       \FOR{$\ell=1$ {\bfseries to} $L$}
       \STATE Draw $\theta\sim\mathrm{Unif}(S^{d-1})=\lambda$
       \STATE Draw $P\sim \mathrm{Unif}(O_d(\mathbb{R}))=\lambda_O$
       \STATE Get $Q$ the permutation matrix such that $\Tilde{\theta} = Q\theta$ is sorted in decreasing order
       \STATE Set $A=\mathrm{diag}(\Tilde{\theta})$, $\Tilde{P}=PQ^T$
       \STATE $\forall i,j$, $\Tilde{X}_i^{\ell} = \Tilde{P}^T X_i \Tilde{P}$, $\Tilde{Y}_j^{\ell} = \Tilde{P}^T Y_j \Tilde{P}$
       \STATE $\forall i, j$, $D_i^\ell=UDU(\Tilde{X}_i^\ell)$, $\Delta_j^\ell = UDU(\Tilde{Y}_j^\ell)$
       \STATE  $\forall i,j,\ \hat{X}_i^{\ell}=P^A(D^\ell_i)$, $\hat{Y}_j^\ell=P^A(\Delta^\ell_j)$
       \STATE Compute $W_p^p(\frac{1}{n}\sum_{i=1}^n \delta_{\hat{X}_i^\ell}, \frac{1}{m}\sum_{j=1}^m \delta_{\hat{Y}_j^\ell})$
       \ENDFOR
       \STATE Return $\frac{1}{L}\sum_{\ell=1}^L W_p^p(\frac{1}{n}\sum_{i=1}^n \delta_{\hat{X}_i^\ell}, \frac{1}{m}\sum_{j=1}^m \delta_{\hat{Y}_j^\ell})$
    \end{algorithmic}
\end{algorithm}

\paragraph{Computational Complexity and Implementation.}

Let $\mu,\nu \in \mathcal{P}_p\big(S_d^{++}(\mathbb{R})\big)$ and $(X_i)_{i=1}^n$ (resp. $(Y_j)_{j=1}^m$) samples from $\mu$ (resp. from $\nu$).
We approximate $\lespdsw_p^p(\mu,\nu)$ by $\widehat{\lespdsw}_{p,L}^p(\hat{\mu}_n,\hat{\nu}_m)$ where $\hat{\mu}_n = \frac{1}{n}\sum_{i=1}^n \delta_{X_i}$ and $\hat{\nu}_m = \frac{1}{m}\sum_{j=1}^m \delta_{Y_j}$.
Sampling from $\lambda_O$ requires drawing a matrix $Z\in\mathbb{R}^{d\times d}$ with i.i.d normally distributed coefficients, and then taking the QR factorization with positive entries on the diagonal of $R$ \citep{mezzadri2006generate}, which needs $O(d^3)$ operations \citep[Section 5.2]{golub2013matrix}.
Then, computing $n$ matrix logarithms takes $O(nd^3)$ operations.
Given $L$ projections, the inner-products require $O(Lnd^2)$ operations, and the computation of the one-dimensional Wasserstein distances is done in $O(Ln\log n)$ operations.
Therefore, the complexity of $\lespdsw$ is $O\big(Ln(\log n + d^2) + (L+n)d^3\big)$.
The procedure is detailed in \Cref{alg:spdsw}.
In practice, when it is required to call $\lespdsw$ several times in optimization procedures, the computational complexity can be reduced by drawing projections only once at the beginning.

For $\aispdsw$, it requires an additional projection step with a UDU decomposition for each sample and projection. Hence the overall complexity becomes $O\big(Ln(\log n + d^3)\big)$ where the $O(Lnd^3)$ comes from the UDU decomposition. In practice, it takes more time than $\lespdsw$ for results which are quite similar. We detail the procedure to compute $\aispdsw$ in \Cref{alg:hspdsw}.

Note that it is possible to draw symmetric matrices with complexity $O(d^2)$ by taking $A = \frac{Z + Z^T}{\|Z + Z^T\|_F}$.
Although this is a great advantage from the point of view of computation time, we leave it as an open question to know whether this breaks the bounds in \Cref{prop:spdsw_bound}.

\begin{figure}[t]
    \centering
    \includegraphics[width=0.5\columnwidth]{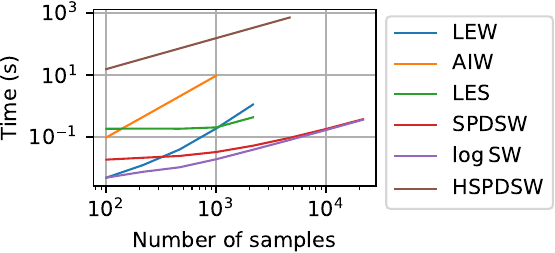}
    \caption{Runtime of $\lespdsw$, $\aispdsw$ and $\logsw$ (200 proj.) compared to alternatives based on Wasserstein between samples from a Wishart distribution in dimension $d=20$.
    Sliced discrepancies can scale to larger distributions in $S_d^{++}(\mathbb{R})$.
    }
    \label{fig:spdsw_runtime}
\end{figure}

We illustrate the computational complexity \emph{w.r.t} samples in \Cref{fig:spdsw_runtime}.
The computations have been performed on a GPU NVIDIA Tesla V100-DGXS 32GB using \texttt{PyTorch} \citep{pytorch}\footnote{Code is available at \url{https://github.com/clbonet/SPDSW}.}.
We compare the runtime to the Wasserstein distance with Affine-Invariant (AIW) and Log-Euclidean (LEW) metrics, and to Sinkhorn algorithm (LES) which is a classical alternative to Wasserstein to reduce the computational cost.
When enough samples are available, then computing the Wasserstein distance takes more time than computing the cost matrix, and $\lespdsw$ is faster to compute. The computational burden of the UDU decomposition for $\aispdsw$ is huge and it takes even more time than computing the Wasserstein distance. Hence, in the following, we will focus on $\lespdsw$ which we show is a computationally efficient alternative to Wasserstein on $\mathcal{P}\big(S_d^{++}(\mathbb{R})\big)$ as it is topologically equivalent while having a better computational complexity and being better conditioned for regression of distributions.

\section{From Brain Data to Distributions in $S_d^{++}(\mathbb{R})$}

M/EEG data consists of multivariate time series $X \in \mathbb{R}^{N_C \times T}$, with $N_C$ channels, and $T$ time samples.
A widely adopted model assumes that the measurements $X$ are linear combinations of $N_S$ sources $S \in \mathbb{R}^{N_S \times T}$ degraded by noise $N \in \mathbb{R}^{N_C \times T}$.
This leads to $X = AS + N$, where $A \in \mathbb{R}^{N_C \times N_S}$ is the forward linear operator \citep{hamalainen1993magnetoencephalography}.
A common practice in statistical learning on M/EEG data is to consider that the target is a function of the power of the sources, \emph{i.e.} $\mathbb{E}[SS^T]$ \citep{blankertz2007optimizing, dahne2014spoc, sabbagh2019manifold}.
In particular, a broad range of methods rely on second-order statistics of the measurements, \emph{i.e.} covariance matrices of the form $C = \frac{XX^T}{T}$, which are less costly and uncertain than solving the inverse problem to recover $S$ before training the model.
After proper rank reduction to turn the covariance estimates into SPD matrices \citep{harandi2017dimensionality}, and appropriate band-pass filtering to stick to specific physiological patterns \citep{blankertz2007optimizing}, Riemannian geometry becomes an appropriate tool to deal with such data.

In this section, we propose two applications of $\lespdsw$ to prediction tasks from M/EEG data.
More specifically, we introduce a new method to perform brain-age regression, building on the work of \citet{sabbagh2019manifold} and \citet{meunier2022distribution}, and another for domain adaptation in BCI.

\subsection{Distributions Regression for Brain-age Prediction}
\label{sec:brain_age}

Learning to predict brain age from population-level neuroimaging data-sets can help characterize biological aging and disease severity \citep{spiegelhalter2016old, cole2017predicting, cole2018brain}.
Thus, this task has encountered more and more interest in the neuroscience community in recent years \citep{xifra2021estimating, peng2021accurate, engemann2022reusable}.
In particular, \citet{sabbagh2019manifold} take advantage of Riemannian geometry for feature engineering and prediction with the following steps.
First, one covariance estimate is computed per frequency band from each subject recording.
Then these covariance matrices are projected onto a lower dimensional space to make them full rank, for instance with a PCA.
Each newly obtained SPD matrix is projected onto the $\log$ space to obtain a feature after vectorization and aggregation among frequency bands.
Finally, a Ridge regression model predicts brain age.
This white-box method achieves state-of-the-art brain age prediction scores on MEG datasets like Cam-CAN \citep{taylor2017cambridge}.

\paragraph{MEG recordings as distributions of covariance matrices.}
Instead of modeling each frequency band by a unique covariance matrix, we propose to use a distribution of covariance matrices estimated from small time frames.
Concretely, given a time series $X \in \mathbb{R}^{N_C \times T}$ and a time-frame length $t < T$, a covariance matrix is estimated from each one of the $n = \lfloor \frac{T}{t} \rfloor$ chunks of signal available.
This process models each subject by as many empirical distributions of covariance estimates $(C_i)_{i=1}^n$ as there are frequency bands.
Then, all samples are projected on a lower dimensional space with a PCA, as done in \citet{sabbagh2019manifold}.
Here, we study whether modeling a subject by such distributions provides additional information compared to feature engineering based on a unique covariance matrix.
In order to perform brain age prediction from these distributions, we extend recent results on distribution regression with SW kernels \citep{kolouri2016sliced,meunier2022distribution} to SPD matrices, and show that $\lespdsw$ performs well on this prediction task while being easy to implement.

\paragraph{$\lespdsw$ kernels for distributions regression.}
As shown in \Cref{section:properties_spdsw}, $\lespdsw$ is a well-defined distance on distributions in $S_d^{++}(\mathbb{R})$.
The most straightforward way to build a kernel from this distance is to resort to well-known Gaussian kernels, \emph{i.e.} $K(\mu, \nu) = e^{-\frac{1}{2\sigma^2}\lespdsw_2^2(\mu, \nu)}$.

However, this is not sufficient to make it a proper positive kernel.
Indeed, we need $\lespdsw$ to be a Hilbertian distance \citep{hein2005hilbertian}.
A pseudo-distance $d$ on $\mathcal{X}$ is Hilbertian if there exists a Hilbert space $\mathcal{H}$ and a feature map $\Phi : \mathcal{X} \rightarrow \mathcal{H}$ such that $ \forall x, y \in \mathcal{X}, d(x, y) = \|\Phi(x) - \Phi(y) \|_{\mathcal{H}}$.
We now extend \citep[Proposition 5]{meunier2022distribution} to the case of $\lespdsw$ in \Cref{prop:hilbertian}.
\begin{proposition}
    \label{prop:hilbertian}
    Let $m$ be the Lebesgue measure and let $\mathcal{H} = L^2([0,1] \times S_d(\mathbb{R}), m \otimes \lambda_S)$.
    We define $\Phi$ as
    \begin{equation}
        \begin{aligned}
        \Phi : \ &\mathcal{P}_2(S_d^{++}(\mathbb{R})) \rightarrow \mathcal{H}\\
        &\mu \mapsto \big( (q, A) \mapsto F^{-1}_{P^A_{\#}\mu}(q) \big),
        \end{aligned}
    \end{equation}
    where $F^{-1}_{P^A_{\#}\mu}$ is the quantile function of $P^A_{\#}\mu$.
    Then, $\lespdsw_2$ is Hilbertian and for all $\mu, \nu \in\mathcal{P}_2(S_d^{++}(\mathbb{R}))$,
    \begin{equation}
        \lespdsw_2^2(\mu, \nu) = \| \Phi(\mu) - \Phi(\nu) \|_{\mathcal{H}}^2.
    \end{equation}
\end{proposition}
\begin{proof}
    This is a particular case of \Cref{prop:chsw_hilbertian}.
\end{proof}
The proof is similar to the one of \citet{meunier2022distribution} for $\sw$ in Euclidean spaces and highlights two key results.
The first one is that $\lespdsw$ extensions of Gaussian kernels are valid positive definite kernels, as opposed to what we would get with the Wasserstein distance \citep[Section 8.3]{peyre2019computational}.
The second one is that we have access to an explicit and easy-to-compute feature map that preserves $\lespdsw$, making it possible to avoid inefficient quadratic algorithms on empirical distributions from very large data.
In practice, we rely on the finite-dimensional approximation of projected distributions quantile functions proposed in \citet{meunier2022distribution} to compute the kernels more efficiently with the $\ell_2$-norm.
Then, we leverage Kernel Ridge regression for prediction \citep{murphy2012machine}.
Let $0 < q_1 < \dots < q_M < 1$, and $(A_1, \dots, A_L) \in S_d(\mathbb{R})^L$.
The approximate feature map has a closed-form expression in the case of empirical distributions and is defined as
\begin{equation}
    \hat{\Phi}(\mu) = \left(\frac{1}{\sqrt{ML}}F^{-1}_{t^{A_i}_{\#}\mu}(q_j)\right)_{1 \le j \le M, 1 \le i \le L}.
\end{equation}

Regarding brain-age prediction, we model each couple of subject $s$ and frequency band $f$ as an empirical distribution $\mu_n^{s,f}$ of covariance estimates $(C_i)_{i=1}^n$.
Hence, our data-set consists of the set of distributions in $S_d^{++}(\mathbb{R})$
\begin{equation}
    \left(\mu_n^{s,f} = \frac{1}{n} \sum_{i=1}^n \delta_{C_i}\right)_{s, f}.
\end{equation}
First, we compute the associated features $\big( \hat{\Phi}(\mu_n^{s, f})\big)_{s,f}$ by loading the data and band-pass filtering the signal once per subject.
Then, as we are interested in comparing each subject in specific frequency bands, we compute one approximate kernel matrix per frequency $f$, as follows
\begin{equation}
    K^f_{i,j} = e^{-\frac{1}{2\sigma^2}\| \hat{\Phi}(\mu_n^{i, f}) -  \hat{\Phi}(\mu_n^{j, f})\|^2_2}.
\end{equation}
Finally, the kernel matrix obtained as a sum over frequency bands, \emph{i.e.} $K = \sum_f K^f$, is plugged into the Kernel Ridge regression of \texttt{scikit-learn} \citep{pedregosa2011scikit}.

\paragraph{Numerical results.}

We demonstrate the ability of our algorithm to perform well on brain-age prediction on the largest publicly available MEG data-set Cam-CAN \citep{taylor2017cambridge}, which contains recordings from 646 subjects at rest.
We take advantage of the benchmark provided by \citet{engemann2022reusable} -- available online\footnote{\url{https://github.com/meeg-ml-benchmarks/brain-age-benchmark-paper}} and described in \Cref{subsec:brain_age_prediction_details} -- to replicate the same pre-processing and prediction steps from the data, and thus produce a meaningful and fair comparison.

For each one of the seven frequency bands, we divide every subject time series into frames of fixed length.
We estimate covariance matrices from each timeframe with OAS \citep{chen2010shrinkage} and apply PCA for rank-reduction, as in \citep{sabbagh2019manifold}, to obtain SPD matrices of size $53 \times 53$.
This leads to distributions of 275 points per subject and per frequency band.
In \citep{sabbagh2019manifold}, the authors rely on Ridge regression on vectorized projections of SPD matrices on the tangent space.
We also provide a comparison to Kernel Ridge regression based on a kernel with the Log-Euclidean metric, \emph{i.e.} $K_{i,j}^{\log} = e^{-\frac{1}{2\sigma^2}\| \log C_i - \log C_j \|^2_F}$.

\begin{figure*}[t]
    \includegraphics[width=\linewidth]{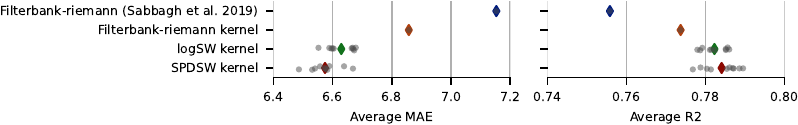}
    \caption{Average MAE and $R^2$ score for 10 random seeds on the Cam-CAN data-set with time-frames of 2s and 1000 projections.
    Kernel Ridge regression based on $\mathrm{SW}$ kernels performs best.
    $\mathrm{SPDSW}$ and $\log \mathrm{SW}$ are close to each other.
    Sampling from symmetric matrices offers a slight advantage but does not play a key role on performance.
    For information, Euclidean $\mathrm{SW}$ led to poor results on the task (MAE 9.7).
    }
    \label{fig:brain_age_average}
    \vspace{-5pt}
\end{figure*}

\Cref{fig:brain_age_average} shows that $\lespdsw$ and $\logsw$ (1000 projections, time-frames of 2s) perform best in average on 10-folds cross-validation for 10 random seeds, compared to the baseline with Ridge regression \citep{sabbagh2019manifold} and to Kernel Ridge regression based on the Log-Euclidean metric, with identical pre-processing.
We provide more details on scores for each fold on a single random seed in \Cref{fig:brain_age_boxplot}.
In particular, it seems that evaluating the distance between distributions of covariance estimates instead of just the average covariance brings more information to the model in this brain-age prediction task, and allows to improve the score.
Moreover, while $\lespdsw$ gives the best results, $\logsw$ actually performs well compared to baseline methods.
Thus, both methods seem to be usable in practice, even though sampling symmetric matrices and taking into account the Riemannian geometry improves the performances compared to $\logsw$.
Also note that Log-Euclidean Kernel Ridge regression works better than the baseline method based on Ridge regression \citep{sabbagh2019manifold}. 
Then, \Cref{fig:variance_swspd} in the appendix shows that $\lespdsw$ does not suffer from variance with more than 500 projections in this use case with matrices of size 53 $\times$ 53.
Finally, \Cref{fig:meg_timeframes} shows that there is a trade-off to find between smaller time-frames for more samples per distribution and larger time-frames for less noise in the covariance estimates and that this is an important hyper-parameter of the model.

\begin{table*}[t]
    \centering
    \caption{Accuracy and Runtime for Cross Session.}
    \small
    \resizebox{\linewidth}{!}{
        \begin{tabular}{ccccccccccccc}
             Subjects & Source & AISOTDA & & SPDSW & LogSW & LEW & LES & & SPDSW & LogSW & LEW & LES \\
             & & \citep{yair2019domain} & & \multicolumn{4}{c}{Transformations in $S_d^{++}(\mathbb{R})$} & & \multicolumn{4}{c}{Descent over particles}\\ \toprule
            1 & 82.21 & 80.90 & & 84.70 & 84.48 & 84.34 & 84.70 & & 85.20 & 85.20 & 77.94 & 82.92 \\
            3 & 79.85 & 87.86 & & 85.57 & 84.10 & 85.71 & 86.08 & & 87.11 & 86.37 & 82.42 & 81.47 \\
            7 & 72.20 & 82.29 & & 81.01 & 76.32 & 81.23 & 81.23 & & 81.81 & 81.73 & 79.06 & 73.29 \\
            8 & 79.34 & 83.25 & & 83.54 & 81.03 & 82.29 & 83.03 & & 84.13 & 83.32 & 80.07 & 85.02 \\
            9 & 75.76 & 80.25 & & 77.35 & 77.88 & 77.65 & 77.65 & & 80.30 & 79.02 & 76.14 & 70.45 \\
            \midrule 
            Avg. acc. & 77.87 & 82.93 & & 82.43 & 80.76 & 82.24 & 82.54 & & 83.71 & 83.12 & 79.13 & 78.63 \\
            Avg. time (s) & - & - & & \textbf{4.34} & \textbf{4.32} & 11.41 & 12.04 & & \textbf{3.68} & \textbf{3.67} & 8.50 & 11.43 \\
            \bottomrule
        \end{tabular}
    }
    \label{tab:cross_session}
\end{table*}

\subsection{Domain Adaptation for Brain Computer Interface}

BCI consists in establishing a communication interface between the brain and an external device, in order to assist or repair sensory-motor functions \citep{daly2008brain, nicolas2012brain, wolpaw2013brain}.
The interface should be able to correctly interpret M/EEG signals and link them to actions that the subject would like to perform.
One challenge of BCI is that ML methods are generally not robust to the change of data domain, which means that an algorithm trained on a particular subject will not be able to generalize to other subjects.
Domain adaptation (DA) 
\citep{bendavidDA} offers a solution to this problem by taking into account the distributional shift between source and target domains.
Classical DA techniques employed in BCI involve projecting target data on source data or vice versa, or learning a common embedding that erases the shift, sometimes with the help of Optimal Transport \citep{courty2016optimal}. As Riemannian geometry works well on BCI \citep{barachant2013classification}, DA tools have been developed for SPD matrices \citep{yair2019domain, ju2022deep}.

\paragraph{$\lespdsw$ for domain adaptation on SPD matrices.}
We study two training frameworks on data from $\mathcal{P}\big(S_d^{++}(\mathbb{R})\big)$.
In the first case, a push forward operator $f_\theta$ is trained to change a distribution $\mu_S$ in the source domain into a distribution $\mu_T$ in the target domain by minimizing a loss of the form $L(\theta) = \mathcal{L}\big((f_\theta)_\#\mu_S, \mu_T\big)$, where $\mathcal{L}$ is a transport cost like Wasserstein on $\mathcal{P}\big(S_d^{++}(\mathbb{R})\big)$ or $\mathrm{SPDSW}$.
The model $f_\theta$ is a sequence of simple transformations in $S_d^{++}(\mathbb{R})$ \citep{rodrigues2018riemannian}, \emph{i.e.} $T_W(C) = W^T C W$ for $W \in S_d^{++}(\mathbb{R})$ (translations) or $W \in SO_d(\mathbb{R})$ (rotations), potentially combined to specific non-linearities \citep{huang2017riemannian}.
The advantage of such models is that they provide a high level of structure with a small number of parameters.

In the second case, we directly align the source on the target by minimizing $\mathcal{L}$ with a Riemannian gradient descent directly over the particles \citep{boumal2023introduction}, \emph{i.e.} by denoting $\mu_S \big((x_i)_{i=1}^{|X_S|}\big)=\frac{1}{|X_S|}\sum_{i=1}^{|X_S|}\delta_{x_i}$ with $X_S=\{x_i^S\}_i$ the samples of the source, we initialize at $(x_i^S)_{i=1}^{|X_S|}$ and minimize
$L\big((x_i)_{i=1}^{|X_S|}\big) = \mathcal{L}\left(\mu_S\big((x_i)_{i=1}^{|X_S|}\big), \mu_T\right)$.

We use \texttt{Geoopt} \citep{kochurov2020geoopt} and \texttt{Pytorch} \citep{paszke2017automatic} to optimize on manifolds.
Then, an SVM is trained on the vectorized projections of $X_S$ in the $\log$ space, \emph{i.e.} from couples $\big(\mathrm{vect}(\log x_i^S), y_i\big)_{i=1}^{|X_S|}$, and we evaluate the model on the target distribution.

\begin{figure}[t]
	\centering
	\hspace*{\fill}
	\subfloat{\includegraphics[width=0.225\columnwidth]{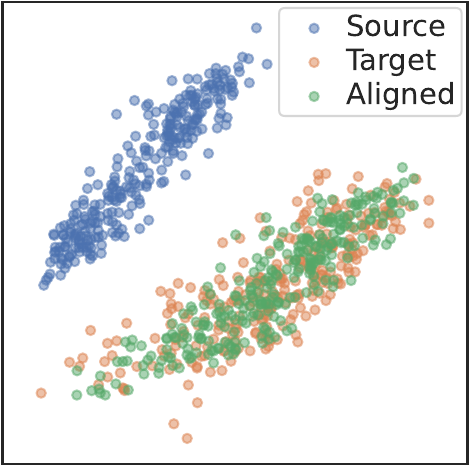}} \hfill
	\subfloat{\includegraphics[width=0.235\columnwidth]{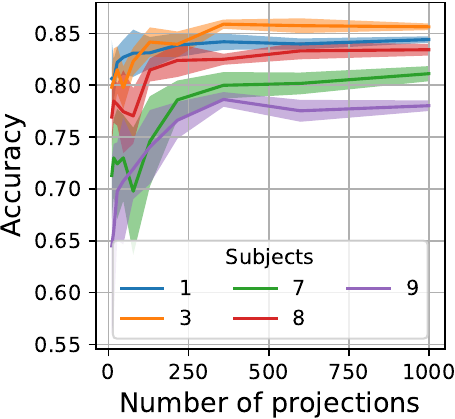}}
	\hspace*{\fill}
	\caption{(\textbf{Left}) PCA on BCI data before and after alignment. Minimizing $\lespdsw$ with enough projections allows aligning sources on targets.
		(\textbf{Right}) Accuracy \emph{w.r.t} num. of projections for the cross-session task with transformations. Here, there is no need for too many projections to converge.}
	\label{fig:pca_acc_projs}
\end{figure}

\paragraph{Numerical results.}
In \Cref{tab:cross_session}, we focus on cross-session classification for the BCI IV 2.a Competition dataset \citep{brunner2008bci} with 4 target classes and about 270 samples per subject and session.
We compare accuracies and runtimes for several methods run on a GPU Tesla V100-DGXS-32GB.
The distributions are aligned by minimizing different discrepancies, namely $\lespdsw$, $\logsw$, Log-Euclidean Wasserstein (LEW) and Sinkhorn (LES), computed with \texttt{POT}~\citep{flamary2021pot}.
Note that we did not tune hyper-parameters on each particular subject and discrepancy, but only used a grid search to train the SVM on the source data-set, and optimized each loss until convergence, \emph{i.e.} without early stopping.
We compare this approach to the naive one without DA, and to the barycentric OTDA \citep{courty2016optimal} with Affine-Invariant metric reported from \citep{yair2019domain}. 
We provide further comparisons on cross-subject in \Cref{appendix:da}.
Our results show that all discrepancies give equivalent accuracies.
As expected, $\lespdsw$ has an advantage in terms of computation time compared to other transport losses.
Moreover, transformations in $S_d^{++}(\mathbb{R})$ and descent over the particles work almost equally well in the case of $\lespdsw$.
We illustrate the alignment we obtain by minimizing $\lespdsw$ in \Cref{fig:pca_acc_projs}, with a PCA for visualization purposes.
Additionally, \Cref{fig:pca_acc_projs} shows that $\lespdsw$ does not need too many projections to reach optimal performance.
We provide more experimental details in \Cref{sec:exp_details}.


\section{Conclusion}

We introduced in this Chapter two new discrepancies between distributions of SPD matrices. The first, $\aispdsw$, is defined using the Affine-Invariant metric but is computationally heavy to compute. The second, $\lespdsw$, uses the Log-Euclidean metric and has appealing properties such as being a distance and metrizing the weak convergence.
Being a Hilbertian metric, it can be plugged as is into Kernel methods, as we demonstrate for brain age prediction from MEG data.
Moreover, it is usable in loss functions dealing with distributions of SPD matrices, for instance in domain adaptation for BCI, with less computational complexity than its counterparts. Beyond M/EEG data, our discrepancy is of interest for any learning problem that involves distributions of SPD matrices, and we expect to see other applications of $\lespdsw$ in the future.

One might also be interested in using other metrics on positive definite or semi-definite matrices such as the Bures-Wasserstein metric, with the additional challenges that this space is positively curved and not geodesically complete \citep{thanwerdas2021n}. In particular, the Log-Euclidean metric belongs to the family of pullback metrics \citep[Theorem 3.1]{chen2023adaptive}. Thus, it would be of interest to compare the results on different tasks using other pullback metrics such as the Log-Cholesky metric \citep{lin2019riemannian} or the Adaptative metric introduced in \citep{chen2023adaptive} which could be learned given the data. Moreover, the Affine-Invariant metric can be derived as a particular instance of vector-valued distances \citep{lopez2021vector} which also encompass the symmetric Stein divergence \citep{cherian2011efficient, sra2012new, sra2016positive} and Finsler distances, and which could be of interests to study.


Further works could also be done to improve the design of the kernel used in the brain-age regression task, as they are taken as the sum over all frequencies. A natural lead forward would be to perform a non uniform linear combination of each frequency by learning weights, for example using the Multiple Kernel Learning framework \citep{rakotomamonjy2008simplemkl}.

\clearemptydoublepage
\cleartooddpage[\thispagestyle{empty}]
\chapter{Spherical Sliced-Wasserstein} \label{chapter:ssw}

{
    \hypersetup{linkcolor=black} 
    \minitoc 
}

This chapter is based on \citep{bonet2023spherical} and aims at defining a new Sliced-Wasserstein discrepancy on the sphere. The sphere seen as a Riemannian manifold is of unit curvature, and hence does not enter into the general framework on manifolds of non-positive curvature developed in \Cref{chapter:sw_hadamard}, which poses additional challenges. Hence, we define a novel SW discrepancy, which we call Spherical Sliced-Wasserstein, for probability distributions lying on the sphere. Our construction is notably based on closed-form solutions of the Wasserstein distance on the circle, together with a spherical Radon transform. Along with efficient algorithms and the corresponding implementations, we illustrate its properties in several Machine Learning use cases where spherical representations of data are at stake: sampling on the sphere, density estimation on real earth data or hyperspherical auto-encoders.

\section{Introduction}


Although embedded in larger dimensional Euclidean spaces, data generally lie in practice on manifolds \citep[]{fefferman2016testing}. A simple manifold, but with lots of practical applications, is the hypersphere $S^{d-1}$. Several types of data are by essence spherical: a good example is found in directional data \citep[]{mardia2000directional,pewsey2021recent} for which dedicated Machine Learning solutions are being developed~\citep[]{sra2018directional}, but other applications concern for instance geophysical data \citep{di2014nonparametric}, meteorology \citep{besombes2021producing}, cosmology \citep{perraudin2019deepsphere} or extreme value theory for the estimation of spectral measures \citep{guillou2015folding}. 
Remarkably, in a more abstract setting, considering hyperspherical latent representations of data is becoming more and more common, see {\em e.g.} \citep{Liu_2017_CVPR,xu2018spherical,davidson2018hyperspherical}. For example, in the context of Variational Autoencoders \citep[]{kingma2013auto}, using priors on the sphere has been demonstrated to be beneficial \citep[]{davidson2018hyperspherical}. Also, in the context of Self-Supervised Learning (SSL), where one wants to learn discriminative representations in an unsupervised way, the hypersphere is usually considered for the latent representation~\citep[]{wu2018unsupervised,chen2020simple,wang2020understanding,grill2020bootstrap,caron2020unsupervised}. It is thus of primary importance to develop Machine Learning tools that accommodate well with this specific geometry.

The OT theory on manifolds is well developed \citep{mccann2001polar,villani2009optimal, figalli2011optimal} and several works started to use it in practice, with a focus mainly on the approximation of OT maps. For example, \citet{cohen2021riemannian, rezende2021implicit} approximate the OT map to define Normalizing Flows on the sphere and \citet{cui2019spherical, hamfeldt2021convergent, hamfeldt2022convergence} derive algorithms to approximate the OT map on the sphere.
However, the computational bottleneck to compute the Wasserstein distance on such spaces remains. 
Notably, \citet{rustamov2020intrinsic} proposed a variant of SW, based on the spectral decomposition of the Laplace-Beltrami operator, which generalizes to manifolds given the availability of the eigenvalues and eigenfunctions. However, it is not directly related to the original SW on Euclidean spaces.

\paragraph{Contributions.} \looseness=-1 Therefore, by leveraging properties of the Wasserstein distance on the circle \citep{rabin2011transportation}, we define the first, to the best of our knowledge, natural generalization of the original SW discrepancy on the sphere $S^{d-1}$, and hence we make a first step towards defining SW distances on Riemannian manifolds of positive curvature. We make connections with a new spherical Radon transform and analyze some of its properties. We discuss the underlying algorithmic procedure, and notably provide an efficient implementation when computing the discrepancy against the uniform distribution. Then, we show that we can use this discrepancy on different tasks such as sampling, density estimation or generative modeling.


\section{A Sliced-Wasserstein Discrepancy on the Sphere} \label{section:ssw}

\looseness=-1 Our goal here is to define a Sliced-Wasserstein distance on the sphere $S^{d-1}=\{x\in \mathbb{R}^d,\ \|x\|_2=1\}$. To that aim, we proceed analogously to the classical Euclidean space. However, contrary to Euclidean spaces or Cartan-Hadamard manifolds, geodesics are actually great circles, \emph{i.e.} circles with the same diameter as the sphere, and there is no clear origin. Hence, we propose to integrate over all possible great circles, which play the role of the real line for the hypersphere, instead of all great circles passing through some origin. Then, we propose to project distributions lying on the sphere to great circles, and we rely on the nice properties of the Wasserstein distance on the circle \citep{rabin2011transportation}. In this section, we first describe the OT problem on the circle before defining a new Sliced-Wasserstein discrepancy on the sphere. 

\subsection{Optimal Transport on the Circle}
    
On the circle $S^1 = \mathbb{R}/\mathbb{Z}$ equipped with the geodesic distance $d_{S^1}$, an appealing formulation of the Wasserstein distance is available \citep{delon2010fast}. First, let us parametrize $S^1$ by $[0,1[$, then the geodesic distance can be written as, for all $x,y\in [0,1[,\ d_{S^1}(x,y)= \min(|x-y|, 1-|x-y|)$ \citep{rabin2011transportation}.
Then, for the cost function $c(x,y)=h\big(d_{S^1}(x,y)\big)$ with $h:\mathbb{R}\to\mathbb{R}^+$ an increasing convex function, the Wasserstein distance between $\mu\in\mathcal{P}(S^1)$ and $\nu\in\mathcal{P}(S^1)$ can be written as
\begin{equation}
    W_c(\mu,\nu) = \inf_{\alpha\in\mathbb{R}}\ \int_0^1 h\big(|F_\mu^{-1}(t)-(F_\nu-\alpha)^{-1}(t)|\big)\ \mathrm{d}t,
\end{equation}
where $F_\mu:[0,1[\to [0,1]$ denotes the cumulative distribution function (cdf) of $\mu$, $F_\mu^{-1}$ its quantile function and $\alpha$ is a shift parameter. The optimization problem over the shifted cdf $F_\nu-\alpha$ can be seen as looking for the best ``cut'' (or origin) of the circle in order to wrap it into the real line because of the 1-periodicity. Indeed, the proof of this result for discrete distributions in \citep{rabin2011transportation} consists in cutting the circle at the optimal point and wrapping it around the real line, for which the Optimal Transport map is the increasing rearrangement $F_\nu^{-1}\circ F_\mu$ which can be obtained for discrete distributions by sorting the points \citep{peyre2019computational}.
 
\citet{rabin2011transportation} showed that the minimization problem is convex and coercive in the shift parameter and \citet{delon2010fast} derived a binary search algorithm to find it. For the particular case of $h=\mathrm{Id}$, it can further be shown \citep{werman1985distance, cabrelli1995kantorovich} that
\begin{equation}
    W_1(\mu,\nu) = \inf_{\alpha\in\mathbb{R}}\ \int_0^1 |F_\mu(t)-F_\nu(t)-\alpha|\ \mathrm{d}t.
\end{equation}
In this case, we know exactly the minimum which is attained at the level median \citep{hundrieser2022statistics}, defined as, for $f:[0,1[\to\mathbb{R}$,
\begin{equation}
    \mathrm{LevMed}(f) = \min\left\{\argmin_{\alpha\in\mathbb{R}}\ \int_0^1 |f(t)-\alpha|\mathrm{d}t\right\} = \inf\left\{t\in\mathbb{R},\ \beta\big(\{x\in[0,1[,\ f(x)\le t\}\big)\ge \frac12 \right\},
\end{equation}
where $\beta$ is the Lebesgue measure. Therefore, we also have
\begin{equation} \label{eq:levmedformula}
    W_1(\mu,\nu) = \int_0^1 |F_\mu(t)-F_\nu(t)-\mathrm{LevMed}(F_\mu-F_\nu)|\ \mathrm{d}t.
\end{equation}
Since we know the minimum, we do not need the binary search and we can approximate the integral very efficiently as we only need to sort the samples to compute the level median and the cdfs.

Another interesting setting in practice is to compute $W_2$, \emph{i.e.} with $h(x)=x^2$, \emph{w.r.t.} the uniform distribution $\nu$ on the circle. We derive here the optimal shift $\hat{\alpha}$ for the Wasserstein distance between $\mu$ an arbitrary distribution on $S^1$ and $\nu$. We also provide a closed-form when $\mu$ is a discrete distribution.
\begin{proposition} \label{prop:w2_unif_circle}
    Let $\mu\in\mathcal{P}_2(S^1)$ and $\nu=\mathrm{Unif}(S^1)$. Then,
    \begin{equation} \label{eq:w2_unif_circle}
        W_2^2(\mu,\nu) = \int_0^1 |F_\mu^{-1}(t)-t-\hat{\alpha}|^2\ \mathrm{d}t \quad \text{with} \quad \hat{\alpha} = \int x\ \mathrm{d}\mu(x) - \frac12.
    \end{equation}
    In particular, if $x_1< \dots < x_n$ and $\mu_n = \frac{1}{n}\sum_{i=1}^n \delta_{x_i}$, then 
    \begin{equation} \label{eq:w2_unif_closedform}
        W_2^2(\mu_n,\nu) = \frac{1}{n}\sum_{i=1}^n x_i^2 - \Big(\frac{1}{n}\sum_{i=1}^n x_i\Big)^2 + \frac{1}{n^2} \sum_{i=1}^n (n+1-2i)x_i + \frac{1}{12}.
    \end{equation}
\end{proposition}
\begin{proof}
    See \Cref{proof:prop_w2_unif_circle}.
\end{proof}
This proposition offers an intuitive interpretation: the optimal cut point between an empirical and the uniform distribution is the antipodal point of the circular mean of the discrete samples. Moreover, a very efficient algorithm can be derived from this property, as it solely requires a sorting operation to compute the order statistics of the samples.

\subsection{Definition of SW on the Sphere}

\begin{wrapfigure}{R}{0.4\textwidth}
    \centering
    \includegraphics[width=\linewidth]{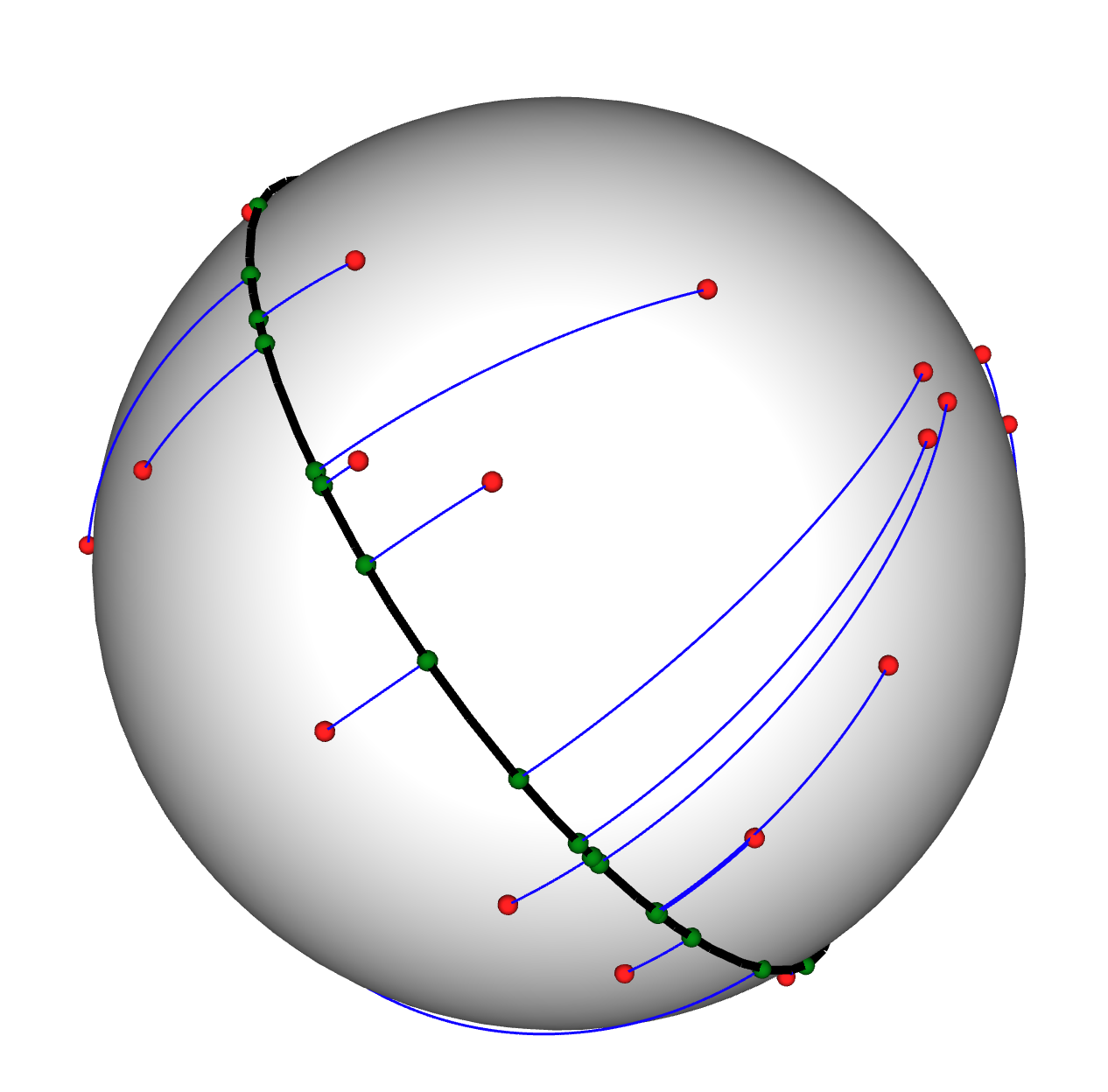}
    \caption{Illustration of the geodesic projections on a great circle (in black). In red, random points sampled on the sphere. In green the projections and in blue the trajectories.}
    \label{fig:geodesic_projs}
\end{wrapfigure}

On the hypersphere, the counterpart of straight lines are the great circles, which are circles with the same diameter as the sphere, and which correspond to the geodesics. Moreover, we can compute the Wasserstein distance on the circle fairly efficiently. Hence, to define a Sliced-Wasserstein discrepancy on this manifold, we propose, analogously to the classical SW distance, to project measures on great circles. The most natural way to project points from $S^{d-1}$ to a great circle $C$ is to use the geodesic projection \citep{fletcher2004principal, jung2021geodesic} defined as
\begin{equation}
    \forall x\in S^{d-1},\ P^C(x) = \argmin_{y\in C}\ d_{S^{d-1}}(x,y),
\end{equation}
where $d_{S^{d-1}}(x,y)=\arccos(\langle x,y\rangle)$ is the geodesic distance. See Figure \ref{fig:geodesic_projs} for an illustration of the geodesic projection on a great circle. Note that the projection is unique for almost every $x$ (see \citep[Proposition 4.2]{bardelli2017probability} and Appendix \ref{appendix:uniqueness_projection}) and hence the pushforward $P^C_\#\mu$ of $\mu\in\mathcal{P}_{p,ac}(S^{d-1})$, where $\mathcal{P}_{p,ac}(S^{d-1})$ denotes the set of absolutely continuous measures \emph{w.r.t.} the Lebesgue measure and with moments of order $p$, is well defined.

Great circles can be obtained by intersecting $S^{d-1}$ with a 2-dimensional plane \citep{jung2012analysis}. Therefore, to average over all great circles, we propose to integrate over the Grassmann manifold $\mathcal{G}_{d,2} =\{E\subset \mathbb{R}^d,\ \mathrm{dim}(E)=2\}$ \citep{absil2004riemannian, bendokat2020grassmann} and then to project the distribution onto the intersection with the hypersphere. Since the Grassmannian is not very practical, we consider the identification using the set of rank 2 projectors:
\begin{equation}
    \begin{aligned}
        \mathcal{G}_{d,2} &= \{P\in\mathbb{R}^{d\times d},\ P^T=P,\ P^2=P,\ \mathrm{Tr}(P)=2\} = \{UU^T,\ U\in\mathbb{V}_{d,2}\},
    \end{aligned}
\end{equation}
where $\mathbb{V}_{d,2}=\{U\in\mathbb{R}^{d\times 2},\ U^TU=I_2\}$ is the Stiefel manifold \citep{bendokat2020grassmann}. 

Finally, we can define the Spherical Sliced-Wasserstein distance (SSW) for $p\ge1$ between locally absolutely continuous measures \emph{w.r.t.} the Lebesgue measure $\mu,\nu\in\mathcal{P}_{p,\mathrm{ac}}(S^{d-1})$ as
\begin{equation} \label{eq:ssw}
    \ssw_p^p(\mu,\nu) = \int_{\mathbb{V}_{d,2}} W_p^p(P^U_\#\mu, P^U_\#\nu)\ \mathrm{d}\sigma(U), 
\end{equation}
where $\sigma$ is the uniform distribution over the Stiefel manifold $\mathbb{V}_{d,2}$, $P^U$ is the geodesic projection on the great circle generated by $U$ and then projected on $S^1$, \emph{i.e.} 
\begin{equation} \label{eq:proj_S1}
    \forall U\in\mathbb{V}_{d,2}, \forall x\in S^{d-1},\ P^U(x) = U^T\argmin_{y \in \mathrm{span}(UU^T)\cap S^{d-1}}\ d_{S^{d-1}}(x,y) = \argmin_{z\in S^1}\ d_{S^{d-1}}(x,Uz),
\end{equation}
and the Wasserstein distance is defined with the geodesic distance $d_{S^1}$. 
Moreover, we can derive a closed form expression which will be very useful in practice:

\begin{lemma} \label{lemma:proj_closeform}
    Let $U\in\mathbb{V}_{d,2}$ then for a.e. $x\in S^{d-1}$,
    \begin{equation}
        P^U(x) = \frac{U^T x}{\|U^Tx\|_2}.
    \end{equation}
\end{lemma}

\begin{proof}
    See \Cref{proof:lemma_proj_closedform_ssw}.
\end{proof}

Hence, we notice from this expression of the projection that we recover almost the same formula as \citet{lin2020projection} but with an additional $\ell^2$ normalization which projects the data on the circle. As in \citep{lin2020projection}, we could project on a higher dimensional subsphere by integrating over $\mathbb{V}_{d,k}$ with $k\ge 2$. However, we would lose the computational efficiency provided by the properties of the Wasserstein distance on the circle.

\section{A Spherical Radon Transform} \label{section:spherical_radon}


In this section, we investigate the distance properties of $\ssw$ through a related spherical Radon transform that we introduce. Similarly as for the Cartan-Hadamard Sliced-Wasserstein that we studied in \Cref{section:chsw_properties}, we can show easily that $\ssw$ is a pseudo distance using integration properties as well as properties of the Wasserstein distance.

\begin{proposition} \label{prop:ssw_distance}
    Let $p\ge 1$, $\ssw_p$ is a pseudo-distance on $\mathcal{P}_{p,ac}(S^{d-1})$.
\end{proposition}

\begin{proof}
    See \Cref{proof:prop_ssw_distance}.
\end{proof}

To show that it is a distance, we require to show that it satisfies the indiscernible property. One way of doing that is to study the injectivity of related Radon transforms.

\subsection{Spherical Radon Transforms}

\looseness=-1 Let us introduce a Spherical Radon transform related to $\ssw$. As for the classical SW distance, we can derive a second formulation using a Radon transform by integrating over the set of points on the sphere which are projected on $z\in S^1$: $\{x\in S^{d-1},\ P^U(x)=z\}$. Let us first identify formally the set of integration. 

\begin{figure}[t]
    \centering
    \includegraphics[width=0.5\linewidth]{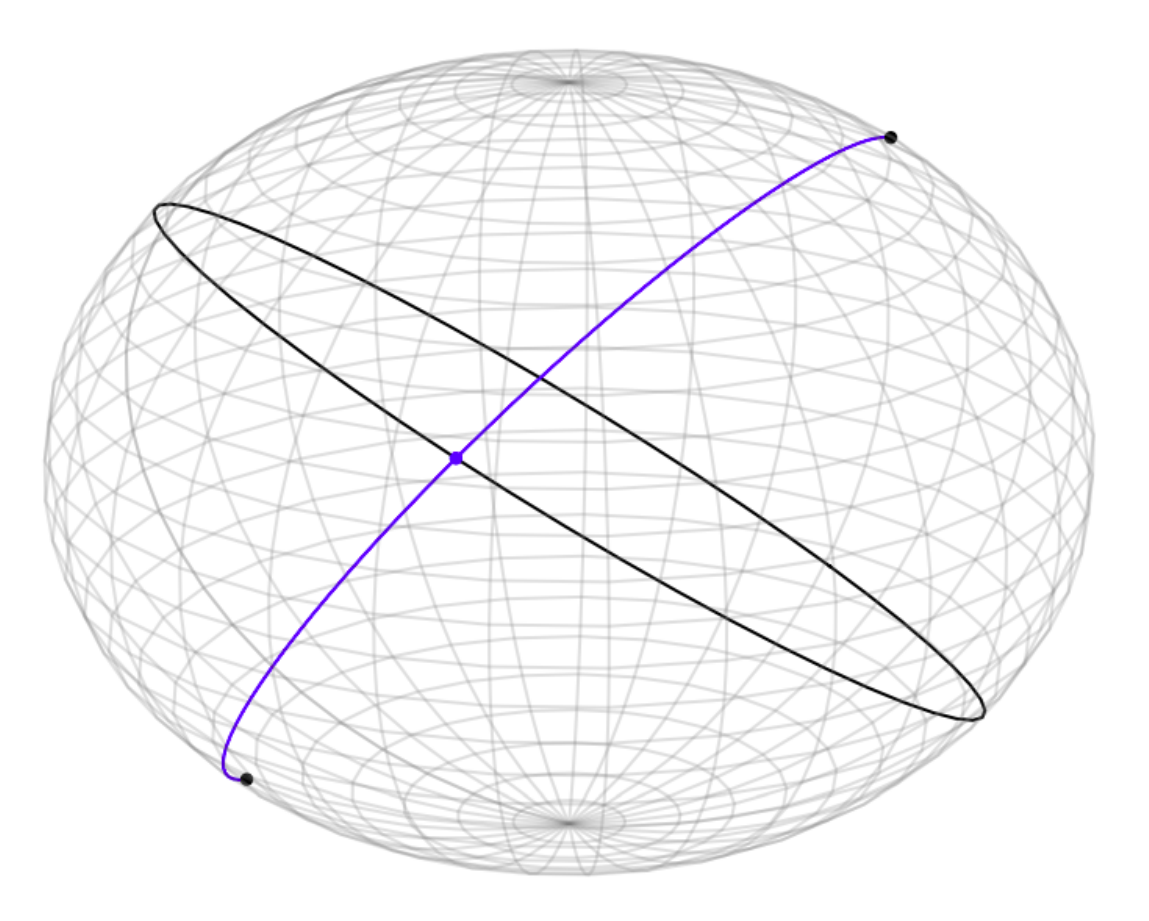}
    \caption{Set of integration of the spherical Radon transform \eqref{eq:spherical_RT}. The great circle is in black and the point $Uz\in \mathrm{span}(UU^T)\cap S^{d-1}$ on which we aim to project is in blue. Then, all the points on the semi-circle in blue are projected on $Uz$ and this semi-circle corresponds to the set of integration of \eqref{eq:spherical_RT}.}
    \label{fig:integration_set}
\end{figure}

\paragraph{Set of integration.} While the classical Radon transform integrates over hyperplanes of $\mathbb{R}^d$, the generalized Radon transform over hypersurfaces \citep{kolouri2019generalized} and the Minkowski-Funk transform over $(d-2)$-dimensional subsphere,  \emph{i.e.} the intersection between a hyperplane and $S^{d-1}$ \citep{rubin2003notes}, we show in \Cref{prop:integration_set_ssw} that the set of integration is a half of a $(d-2)$-subsphere. We illustrate the set of integration on $S^2$ in Figure \ref{fig:integration_set}. In this case, the intersection between a hyperplane and $S^2$ is a great circle, and hence it coincides with a $(d-2)$-subsphere.

\begin{proposition} \label{prop:integration_set_ssw}
    Let $U\in\mathbb{V}_{d,2}$, $z\in S^1$. The set of integration of the Radon transform \eqref{eq:spherical_RT} is 
    \begin{equation}
        \{x\in S^{d-1},\ P^U(x)=z\} = \{x\in F\cap S^{d-1},\ \langle x,Uz\rangle > 0\},
    \end{equation}
    where $F=\mathrm{span}(UU^T)^\bot \oplus \mathrm{span}(Uz)$.
\end{proposition}

\begin{proof}
    See \Cref{proof:prop_integration_set_ssw}.
\end{proof}

\paragraph{Radon transform.}

Let $f\in L^1(S^{d-1})$, we want to define a spherical Radon transform $\Tilde{R}:L^1(S^{d-1})\to L^1(S^1 \times \mathbb{V}_{d,2})$ which integrates the function $f$ over the set of integration described in the last Proposition. However, as communicated to us by Michael Quellmalz and presented in \citep{quellmalz2023sliced} on $S^2$, we cannot just integrate with respect to the volume measure as it would not project probability densities on probability densities. Thus, we need to integrate \emph{w.r.t} the right measure. 

To define properly such transform, let us first recall that the volume measure on $S^{d-1}$ is defined 
for any $f\in L^1(S^{d-1})$ by
\begin{equation}
    \int_{S^{d-1}} f(x)\ \mathrm{d}\vol(x) = \int_0^{2\pi} \int_{[0,\pi]^{d-2}} f\big(\varphi(\theta_1,\dots,\theta_{d-2},\theta_{d-1})\big)\ \left(\prod_{i=1}^{d-2} \sin(\theta_i)^{d-1-i}\right)\ \mathrm{d}\theta_1\dots\mathrm{d}\theta_{d-2}\mathrm{d}\theta_{d-1},
\end{equation}
where for $\theta_{d-1}\in [0,2\pi[$ and $\theta_i\in [0,\pi]$ for $i\in\{1,\dots,d-2\}$,
\begin{equation}
    \varphi(\theta_1,\dots,\theta_{d-1}) = \begin{pmatrix}
        \cos(\theta_1) \\
        \sin(\theta_1) \cos(\theta_2) \\
        \vdots \\
        \sin(\theta_1)\dots\sin(\theta_{d-2})\cos(\theta_{d-1}) \\
        \sin(\theta_1)\dots\sin(\theta_{d-1})
    \end{pmatrix}.
\end{equation}


Let $U_0$ be such that $\mathrm{span}(U_0U_0^T)=\mathrm{span}(e_{d-1},e_{d})$ with $(e_1,\dots,e_d)$ the canonical basis, and define the measure $\smeasure^z$ for $z\in S^1$ such that for any $f\in C_b(S^{d-1})$,
\begin{equation}
    \int_{S^{d-1}} f(x)\ \mathrm{d}\smeasure^z(x) = \int_0^{2\pi} \int_{[0,\pi]^{d-2}} f\big(\varphi(\theta_1,\dots,\theta_{d-2},\theta_{d-1})\big)\ \left(\prod_{i=1}^{d-2} \sin(\theta_i)^{d-1-i}\right)\ \mathrm{d}\theta_1\dots\mathrm{d}\theta_{d-2}\delta_{\mathrm{ang}(U_0 z)}(\mathrm{d}\theta_{d-1}).
\end{equation}
Here, $\mathrm{ang}(U_0 z)$ denotes the angle of $U_0z$ on the circle $\mathrm{span}(U_0U_0^T)\cap S^{d-1}$ which can be obtained using the atan2 function. Note that by integrating the last equation \emph{w.r.t} $z\in S^1$, we obtain by using the definition of the surface measure on $S^{d-1}$,
\begin{equation}
    \begin{aligned}
        &\int_{S^1}\int_{S^{d-1}} f(x)\ \mathrm{d}\sigma_d^z(x)\ \mathrm{d}\vol(z) \\ &= \int_0^{2\pi} \int_{[0,\pi]^{d-2}} f\big(\varphi(\theta_1,\dots,\theta_{d-2},\theta_{d-1})\big)\ \left(\prod_{i=1}^{d-2} \sin(\theta_i)^{d-1-i}\right)\ \mathrm{d}\theta_1\dots\mathrm{d}\theta_{d-2}\mathrm{d}\theta_{d-1} \\
        &= \int_{S^{d-1}} f(x)\ \mathrm{d}\vol(x).
    \end{aligned}
\end{equation}
In this case, if $f$ is a density with respect to the measure $\vol$, then we obtain well that it integrates to 1. Thus, we define the spherical Radon transform for $U_0$ as
\begin{equation}
    \forall z\in S^1,\ \Tilde{R}f(z, U_0) = \int_{S^{d-1}} f(x) \ \mathrm{d}\smeasure^z(x).
\end{equation}
For arbitrary $U\in\mathbb{V}_{d,2}$, denote $O_U\in SO(d)$ the rotation such that for all $z\in S^1$, $O_U Uz\in \mathrm{span}(e_{d-1} ,e_{d})$. Applying the change-of-variable $x=O_U^Ty$, and defining $\Tilde{\sigma}_d^z = (O_U^T)_\#\smeasure^z$, we can define
\begin{equation} \label{eq:spherical_RT}
    \begin{aligned}
        \forall z\in S^1, U\in\mathbb{V}_{d,2},\ \Tilde{R}f(z,U) &= \int_{S^{d-1}} f(x)\ \mathrm{d}\Tilde{\sigma}_d^z(x) \\
        &= \int_{S^{d-1}} f(O_U^T y)\ \mathrm{d}\smeasure^z(y). \\
    \end{aligned}
\end{equation}

Then, analogously to the classical Radon transform, we can define the back-projection operator $\Tilde{R}^*:C_b(S^1\times \mathbb{V}_{d,2}) \to C_b(S^{d-1})$, $C_b(S^{d-1})$ being the space of continuous bounded functions, for $g\in C_b(S^1\times \mathbb{V}_{d,2})$ as for a.e. $x\in S^{d-1}$,
\begin{equation}
    \Tilde{R}^*g(x) = \int_{\mathbb{V}_{d,2}} g\big(P^U(x),U\big)\ \mathrm{d}\sigma(U).
\end{equation}

\begin{proposition} \label{prop:radon_dual_ssw}
    $\Tilde{R}^*$ is the dual operator of $\Tilde{R}$, \emph{i.e.} for all $f\in L^1(S^{d-1})$, $g\in C_b(S^1\times \mathbb{V}_{d,2})$,
    \begin{equation}
        \langle \Tilde{R}f,g\rangle_{S^1\times \mathbb{V}_{d,2}} = \langle f, \Tilde{R}^*g\rangle_{S^{d-1}}.
    \end{equation}
\end{proposition}

\begin{proof}
    See \Cref{proof:prop_radon_dual_ssw}.
\end{proof}

Now that we have a dual operator, we can also define the Radon transform of an absolutely continuous measure $\mu\in\mathcal{M}_{ac}(S^{d-1})$ by duality \citep{boman2009support, bonneel2015sliced} as the measure $\Tilde{R}\mu$ satisfying 
\begin{equation}
    \forall g\in C_b(S^1\times \mathbb{V}_{d,2}),\ \int_{S^1\times \mathbb{V}_{d,2}} g(z,U) \ \mathrm{d}(\Tilde{R}\mu)(z,U) = \int_{S^{d-1}} \Tilde{R}^*g(x)\ \mathrm{d}\mu(x).
\end{equation}
Since $\Tilde{R}\mu$ is a measure on the product space $S^1\times \mathbb{V}_{d,2}$, $\Tilde{R}\mu$ can be disintegrated \citep[Theorem 5.3.1]{ambrosio2008gradient} \emph{w.r.t.} $\sigma$ as $\Tilde{R}\mu = \sigma \otimes K$ where $K$ is a probability kernel on $\mathbb{V}_{d,2}\times \mathcal{S}^1$ with $\mathcal{S}^1$ the Borel $\sigma$-field of $S^1$. We will denote for $\sigma$-almost every $U\in\mathbb{V}_{d,2}$, $(\Tilde{R}\mu)^U = K(U,\cdot)$ the conditional probability. 
\begin{proposition} \label{prop:radon_disintegrated}
    Let $\mu\in \mathcal{M}_{ac}(S^{d-1})$, then for $\sigma$-almost every $U\in\mathbb{V}_{d,2}$, $(\Tilde{R}\mu)^U=P^U_\#\mu$.
\end{proposition}
\begin{proof}
    See \Cref{proof:prop_radon_disintegrated}.
\end{proof}

Finally, we can write SSW \eqref{eq:ssw} using this Radon transform:
\begin{equation} \label{eq:ssw_radon}
    \forall \mu,\nu \in\mathcal{P}_{p,ac}(S^{d-1}),\ \ssw_p^p(\mu,\nu) = \int_{\mathbb{V}_{d,2}} W_p^p\big((\Tilde{R}\mu)^U,(\Tilde{R}\nu)^U\big)\ \mathrm{d}\sigma(U).
\end{equation} 

\subsection{Properties of the Spherical Radon Transform}


As observed by \citet{kolouri2019generalized} for the Generalized SW distances (GSW), studying the injectivity of the related Radon transforms allows to study the set on which SW is actually a distance. 





\paragraph{Link with Hemispherical transform.} Since the intersection between a hyperplane and $S^{d-1}$ is isometric to $S^{d-2}$ \citep{jung2012analysis}, we can relate $\Tilde{R}$ to the hemispherical transform $\mathcal{H}$ \citep{rubin2003notes} on $S^{d-2}$. First, the hemispherical transform of a function $f\in L^1(S^{d-1})$ is defined as
\begin{equation}
    \forall x\in S^{d-1},\ \mathcal{H}_{d-1}f(x) = \int_{S^{d-1}} f(y)\mathbb{1}_{\{\langle x,y\rangle >0\}}\ \mathrm{d}\vol(y).
\end{equation}
From \Cref{prop:integration_set_ssw}, we can write the spherical Radon transform \eqref{eq:spherical_RT} as a hemispherical transform on $S^{d-2}$.

\begin{proposition} \label{prop:link_hemispherical}
    Let $f\in L^1(S^{d-1})$, $U\in\mathbb{V}_{d,2}$ and $z\in S^1$, then
    \begin{equation}
        \Tilde{R}f(z, U) = \int_{S^{d-2}} \Tilde{f}_U(x)\mathbb{1}_{\{\langle x, \Tilde{U}z\rangle >0\}}\ \mathrm{d}\vol(x) = \mathcal{H}_{d-2}\Tilde{f}(\Tilde{U}z),
    \end{equation}
    where for all $x\in S^{d-2}$, $\Tilde{f}_U(x) = f(O_U^T J x)$ with $O_U\in SO(d)$ the rotation matrix such that for all $x\in F=\mathrm{span}(UU^T)^\bot \oplus \mathrm{span}(Uz)$, $O_U x\in \mathrm{span}(e_1,\dots,e_{d-1})$ where $(e_1,\dots,e_d)$ denotes the canonical basis, and $J=\begin{pmatrix}I_{d-1} \\ 0_{1, d-1} \end{pmatrix}$, and $\Tilde{U}=J^T O_U U \in \mathbb{R}^{(d-1)\times 2}$.
\end{proposition}

\begin{proof}
    See \Cref{proof:prop_link_hemispherical}.
\end{proof}

\paragraph{Kernel of $\Tilde{R}$.} By exploiting the formulation involving the hemispherical transform of \Cref{prop:link_hemispherical}, for which the kernel was computed in \citep[Lemma 2.3]{rubin1999inversion},  we can derive the kernel of $\Tilde{R}$ as the set of even measures which are null over all hyperplanes intersected with $S^{d-1}$.

\begin{proposition} \label{prop:kernel_radon_ssw}
    $\mathrm{ker}(\Tilde{R}) = \{\mu\in \mathcal{M}_{\mathrm{even}}(S^{d-1}),\ \forall H\in\mathcal{G}_{d,d-1},\ \mu(H\cap S^{d-1})=0\}$ where $\mu\in \mathcal{M}_{\mathrm{even}}$ if for all $f\in C(S^{d-1})$, $\langle \mu, f\rangle = \langle \mu, f_+\rangle$ with $f_+(x)= \big(f(x)+f(-x)\big)/2$ for all $x$. 
\end{proposition}

\begin{proof}
    See \Cref{proof:prop_kernel_radon_ssw}.
\end{proof}

We leave for future works checking whether this set is null or not. Hence, we conclude here that SSW is a pseudo-distance, but a distance on the sets of injectivity of $\Tilde{R}$ \citep{agranovskyt1996injectivity}.

\subsection{Spherical Radon Transforms from the Literature}

Note that a natural way to define SW distances can be through already known Radon transforms using the formulation \eqref{eq:ssw_radon}. It is for example what was done in \citep{kolouri2019generalized} using generalized Radon transforms \citep{ehrenpreis2003universality, homan2017injectivity} to define generalized SW distances, or in \citep{chen2020augmented} with the spatial Radon transform. 

In this work, we choose to extend the Sliced-Wasserstein distance by using analogous objects defined intrinsically on the sphere, such as great circles as counterparts of geodesics, and the geodesic projection. Constructing SSW like this, we obtained a spherical Radon transform $\Tilde{R}$ related to it. The transform $\Tilde{R}$ was  actually first introduced on $S^2$ by \citet{backus1964geographical} and has already been further studied in the literature \citep{groemer1998spherical, rubin2017determination, hielscher2018svd}. In particular, \citet{groemer1998spherical} noted the link with the hemispherical transform. 
More recently, building on \citep{bonet2023spherical}, \citet{quellmalz2023sliced} studied it on $S^2$ and notably showed that the counterpart Spherical Sliced-Wasserstein distance, which they call the Semi-circle Sliced-Wasserstein distance as $\Tilde{R}$ integrates over semi circles (see \Cref{fig:integration_set}), is well a distance.

However, we could also take the point of view of using a different spherical Radon transform already known in the literature, which we discuss now. The spherical Radon transform which is maybe the most natural is the Minkowski-Funk transform \citep{dann2010minkowski}, defined for $\theta\in S^{d-1}$ and $f\in L^1(S^{d-1})$ as
\begin{equation}
    Mf(\theta) = \int_{S^{d-1}} f(x)\mathbb{1}_{\{\langle x,\theta\rangle = 0\}}\ \mathrm{d}\vol(x).
\end{equation}
The Minkowki-Funk transform integrates over $\mathrm{span}(\theta)^\bot \cap S^{d-1}$, which is the intersection between the hyperplane $\mathrm{span}(\theta)^\bot$ and $S^{d-1}$, and is thus a $(d-2)$-subsphere. $(d-2)$-subspheres are actually totally geodesic submanifolds of dimension $d-2$, and hence can be seen as counterparts of hyperplanes from Euclidean spaces \citep[Chapter 3]{helgason2011integral}. Hence, from that point of view, the Minkowski-Funk transform can be seen as a strict generalization of the usual Euclidean Radon transform. Contrary to our spherical Radon transform which integrates over a half $(d-2)$-subsphere, the Minkowski-Funk transform integrates over full $(d-2)$-subspheres. Therefore, using these sets for a projection is not well defined when projecting on a geodesic, as there would be several possible projections. A possible way around would be to project on half great circles instead of great circles.

\looseness=-1 A second interesting transform on the sphere is the spherical slice transform, studied \emph{e.g.} in \citep{quellmalz2017generalization,quellmalz2020funk,rubin2019reconstruction,rubin2022spherical}, which integrates over affine hyperplanes passing through some point $a\in \mathbb{R}^d$ and intersected with $S^{d-1}$. In the particular case where $a=0$, this actually coincides with the Minkowski-Funk transform. Interestingly, it has different properties given $a\in S^{d-1}$ or $a\notin S^{d-1}$ and in particular, if $a\in S^{d-1}$, it is injective \citep{rubin2022spherical}. Thus, it might be of interest to derive projections from these Radon transforms in order to inherit from these properties. Recently, \citet{quellmalz2023sliced} proposed to use the vertical slice transform \citep{hielscher2016reconstructing, rubin2019vertical}, which corresponds to the limiting case $a=\infty$ when all cross-sections are parallel \citep{rubin2022spherical}, in order to define a Vertical Sliced-Wasserstein discrepancy, which is however only injective on the set of even measures.

\section{Properties and Implementation} \label{section:properties_ssw}

In this Section, we first provide some properties verified by SSW and which are similar with those expected for sliced divergences. Then, we detail the implementation in practice of SSW.

\subsection{Properties}

\paragraph{Convergence.} We begin by showing that $\ssw$ respects the weak convergence, which is straightforward from the properties of the Wasserstein distance. Showing the converse, \emph{i.e.} that the convergence \emph{w.r.t} $\ssw$ implies the weak convergence, is more intricate and is left for future works.

\begin{proposition} \label{prop:ssw_cv}
    Let $(\mu_k),\mu\in\mathcal{P}_p(S^{d-1})$ such that $\mu_k\xrightarrow[k\to\infty]{}\mu$, then
    \begin{equation}
        \ssw_p(\mu_k,\mu)\xrightarrow[k\to\infty]{}0.
    \end{equation}
\end{proposition}

\begin{proof}
    See \Cref{proof:prop_ssw_cv}
\end{proof}

\paragraph{Sample Complexity.}

We show here that the sample complexity is independent of the dimension. Actually, this is a well known property of sliced-based distances and it was studied first in \citep{nadjahi2020statistical}. To the best of our knowledge, the sample complexity of the Wasserstein distance on the circle has not been derived yet. We suppose in the next proposition that it is known as we mainly want to show that the sample complexity of SSW does not depend on the dimension.

\begin{proposition} \label{prop:samplecomplexity_ssw}
    Let $p\ge 1$. Suppose that for $\mu,\nu\in\mathcal{P}(S^1)$, with empirical measures $\hat{\mu}_n = \frac{1}{n}\sum_{i=1}^n \delta_{x_i}$ and $\hat{\nu}_n = \frac{1}{n}\sum_{i=1}^n \delta_{y_i}$, where $(x_i)_i \sim \mu$, $(y_i)_i \sim \nu$ are independent samples, we have
    \begin{equation}
        \mathbb{E}[|W_p^p(\hat{\mu}_n,\hat{\nu}_n)-W_p^p(\mu,\nu)|] \le \beta(p,n).
    \end{equation}
    Then, for any $\mu,\nu\in \mathcal{P}_{p,ac}(S^{d-1})$ with empirical measures $\hat{\mu}_n$ and $\hat{\nu}_n$, we have
    \begin{equation}
        \mathbb{E}[|\ssw_p^p(\hat{\mu}_n,\hat{\nu}_n) - \ssw_p^p(\mu,\nu)|] \le \beta(p,n).
    \end{equation}
\end{proposition}

\begin{proof}
    See \Cref{proof:prop_samplecomplexity_ssw}.
\end{proof}

\paragraph{Projection Complexity.}

We derive in the next proposition the projection complexity, which refers to the convergence rate of the Monte Carlo approximate \emph{w.r.t} of the number of projections $L$ towards the true integral. Note that we find the typical rate of Monte Carlo estimates, and that it has already been derived for sliced-based distances in \citep{nadjahi2020statistical}.

\begin{proposition} \label{prop:proj_complexity_ssw}
    Let $p\ge 1$, $\mu,\nu\in \mathcal{P}_{p,ac}(S^{d-1})$. Then, the error made with the Monte Carlo estimate of $\ssw_p$ can be bounded as
    \begin{equation}
        \begin{aligned}
            \mathbb{E}_U\left[|\widehat{\ssw}_{p,L}^p(\mu,\nu)-\ssw_p^p(\mu,\nu)|\right]^2 &\le \frac{1}{L} \int_{\mathbb{V}_{d,2}} \left(W_p^p(P^U_\#\mu,P^U_\#\nu)-SSW_p^p(\mu,\nu)\right)^2 \ \mathrm{d}\sigma(U) \\
            &= \frac{1}{L}\mathrm{Var}_U\left(W_p^p(P^U_\#\mu,P^U_\#\nu)\right),
        \end{aligned}
    \end{equation}
    where $\widehat{\ssw}_{p,L}^p(\mu,\nu) = \frac{1}{L}\sum_{i=1}^L W_p^p(P^{U_i}_\#\mu, P^{U^i}_\#\nu)$ with $(U_i)_{i=1}^L \sim \sigma$ independent samples.
\end{proposition}

\begin{proof}
    See \Cref{proof:prop_proj_complexity_ssw}.
\end{proof}

\subsection{Implementation}

In practice, we approximate the distributions with empirical approximations and, as for the classical SW distance, we rely on the Monte-Carlo approximation of the integral on $\mathbb{V}_{d,2}$. We first need to sample from the uniform distribution $\sigma\in\mathcal{P}(\mathbb{V}_{d,2})$. This can be done by first constructing $Z\in\mathbb{R}^{d\times 2}$ by drawing each of its component from the standard normal distribution $\mathcal{N}(0,1)$ and then applying the QR decomposition \citep{lin2021projection}. Once we have $(U_\ell)_{\ell=1}^L\sim \sigma$, we project the samples on the circle $S^1$ by applying Lemma \ref{lemma:proj_closeform} and we compute the coordinates on the circle using the $\mathrm{atan2}$ function. Finally, we can compute the Wasserstein distance on the circle by either applying the binary search algorithm of \citep[]{delon2010fast} or the level median formulation \eqref{eq:levmedformula} for $\ssw_1$. In the particular case in which we want to compute $\ssw_2$ between a measure $\mu$ and the uniform measure on the sphere $\nu=\mathrm{Unif}(S^{d-1})$, we can use the appealing fact that the projection of $\nu$ on the circle is uniform, \emph{i.e.} $P^U_\#\nu=\mathrm{Unif}(S^1)$ (particular case of Theorem 3.1 in \citep[]{jung2021geodesic}, see \Cref{appendix:vmf}). Hence, we can use the Proposition \ref{prop:w2_unif_circle} to compute $W_2$, which allows a very efficient implementation either by the closed-form \eqref{eq:w2_unif_closedform} or approximation by rectangle method of \eqref{eq:w2_unif_circle}. This will be of particular interest for applications in \Cref{section:applications_ssw} such as autoencoders. We sum up the procedure in \Cref{alg:ssw}.

\begin{algorithm}[tb]
   \caption{SSW}
   \label{alg:ssw}
    \begin{algorithmic}
       \STATE {\bfseries Input:} $(x_i)_{i=1}^n\sim \mu$, $(y_j)_{j=1}^m\sim \nu$, $L$ the number of projections, $p$ the order
       \FOR{$\ell=1$ {\bfseries to} $L$}
       \STATE Draw a random matrix $Z\in\mathbb{R}^{d\times 2}$ with for all $i,j,\ Z_{i,j}\sim\mathcal{N}(0,1)$
       \STATE $U=\mathrm{QR}(Z) \sim \sigma$
       \STATE Project on $S^1$ the points: $\forall i,j,\ \hat{x}_i^{\ell}=\frac{U^T x_i}{\|U^Tx_i\|_2}$, $\hat{y}_j^\ell=\frac{U^T y_j}{\|U^T y_j\|_2}$
       \STATE Compute the coordinates on the circle $S^1$: $\forall i,j,\ \Tilde{x}_{i}^\ell = (\pi+\mathrm{atan2}(-x_{i,2}, -x_{i,1}))/(2\pi)$, $\Tilde{y}_j^\ell = (\pi+\mathrm{atan2}(-y_{j,2}, -y_{j,1}))/(2\pi)$
       \STATE Compute $W_p^p(\frac{1}{n}\sum_{i=1}^n  \delta_{\Tilde{x}_i^\ell}, \frac{1}{m}\sum_{j=1}^m \delta_{\Tilde{y}_j^\ell})$ by binary search or \eqref{eq:levmedformula} for $p=1$
       \ENDFOR
       \STATE Return $SSW_p^p(\mu,\nu)\approx \frac{1}{L}\sum_{\ell=1}^L W_p^p(\frac{1}{n}\sum_{i=1}^n \delta_{\Tilde{x}_i^\ell}, \frac{1}{m}\sum_{j=1}^m \delta_{\Tilde{y}_j^\ell})$
    \end{algorithmic}
\end{algorithm}

\begin{wrapfigure}{R}{0.5\textwidth}
    \centering
    \includegraphics[width=\linewidth]{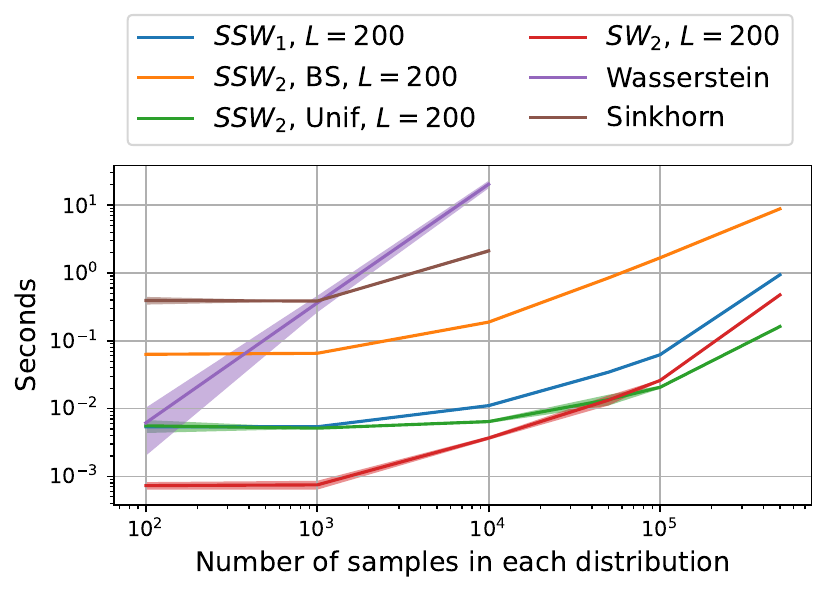}
    \caption{Runtime comparison in log-log scale between W, Sinkhorn with the geodesic distance, $\sw_2$, $\ssw_2$ with the binary search (BS) and uniform distribution \eqref{eq:w2_unif_circle} and $\ssw_1$ with formula \eqref{eq:levmedformula} between two distributions on $S^2$. The time includes the calculation of the distance matrices.}
    \label{fig:runtime_ssww}
\end{wrapfigure}

\paragraph{Complexity.} Let us note $n$ (resp. $m$) the number of samples of $\mu$ (resp. $\nu$), and $L$ the number of projections. First, we need to compute the QR factorization of $L$ matrices of size $d\times 2$. This can be done in $O(Ld)$ by using \emph{e.g.} Householder reflections \citep[Chapter 5.2]{golub2013matrix} or the Scharwz-Rutishauser algorithm \citep{gander1980algorithms}. 
Projecting the points on $S^1$ by Lemma \ref{lemma:proj_closeform} is in $O\big((n+m)dL\big)$ since we need to compute $L(n+m)$ products between $U_\ell^T \in\mathbb{R}^{2\times d}$ and $x\in\mathbb{R}^d$. For the binary search or particular case formula \eqref{eq:levmedformula} and \eqref{eq:w2_unif_closedform}, we need first to sort the points. But the binary search also adds a cost of $O\big((n+m)\log(\frac{1}{\epsilon})\big)$ to approximate the solution with precision $\epsilon$ \citep{delon2010fast} and the computation of the level median requires to sort $(n+m)$ points. Hence, for the general $\ssw_p$, the complexity is $O\big(L(n+m)(d+\log(\frac{1}{\epsilon})) + Ln\log n + Lm\log m\big)$ versus $O\big(L(n+m)(d+\log(n+m))\big)$ for $SSW_1$ with the level median and $O\big(Ln(d + \log n)\big)$ for $\ssw_2$ against a uniform with the particular advantage that we do not need uniform samples in this case.

\paragraph{Runtime Comparison.}

Here, We perform some runtime comparisons. Using \texttt{Pytorch} \citep{pytorch}, we implemented the binary search algorithm of \citep{delon2010fast} and used it with $\epsilon=10^{-6}$. We also implemented $\ssw_1$ using the level median formula \eqref{eq:levmedformula} and $\ssw_2$ against the uniform measure \eqref{eq:w2_unif_circle}. All experiments are conducted on GPU.

On Figure \ref{fig:runtime_ssww}, we compare the runtime between two distributions on $S^2$ between SSW, SW, the Wasserstein distance and the entropic approximation using the Sinkhorn algorithm \citep{cuturi2013sinkhorn} with the geodesic distance as cost function. The distributions were approximated using $n\in \{10^2,10^3,10^4,5\cdot 10^4,10^5,5\cdot 10^5\}$ samples of each distribution and we report the mean over 20 computations. We use the Python Optimal Transport (POT) library \citep{flamary2021pot} to compute the Wasserstein distance and the entropic approximation. For large enough batches, we observe that SSW is much faster than its Wasserstein counterpart, and it also scales better in terms of memory because of the need to store the $n\times n$ cost matrix. For small batches, the computation of SSW actually takes longer because of the computation of the QR factorizations, of the projections and of the binary search. For bigger batches, it is bounded by the sorting operation and we recover the quasi-linear slope. Furthermore, as expected, the fastest algorithms are $\ssw_1$ with the level median and $\ssw_2$ against a uniform as they have a quasilinear complexity. 

\section{Experiments} \label{section:applications_ssw}

\begin{figure}[t]
    \centering
    \begin{minipage}{.45\textwidth}
        \centering
        \hspace*{\fill}
        \subfloat[Target]{\label{target_mixture_vmf}\includegraphics[width=0.45\columnwidth]{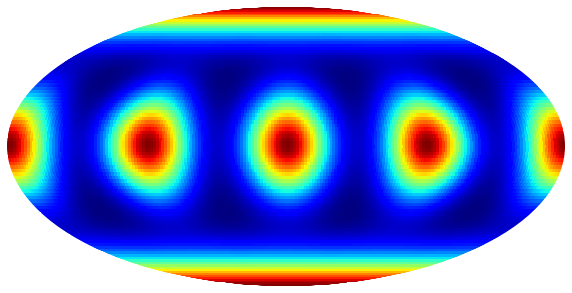}} \hfill
        \subfloat[KDE estimate]{\label{particles_mixture_vmf}\includegraphics[width=0.45\columnwidth]{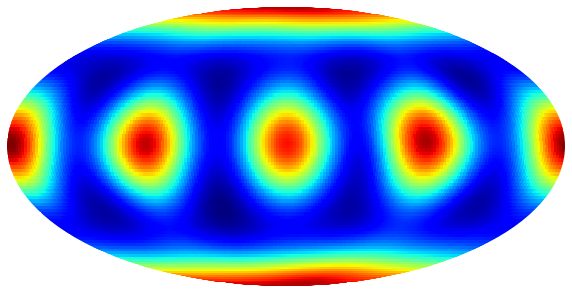}}
        \hspace*{\fill}
        \caption{Minimization of SSW with respect to a mixture of vMF.}
        \label{fig:gradient_flow_mixture_vmf}
    \end{minipage}
    \hspace{0.03\textwidth}
    \begin{minipage}{.45\textwidth}
        \centering
        \hspace*{\fill}
        \subfloat[SWAE]{\label{latent_space_swae}\includegraphics[width=0.45\columnwidth]{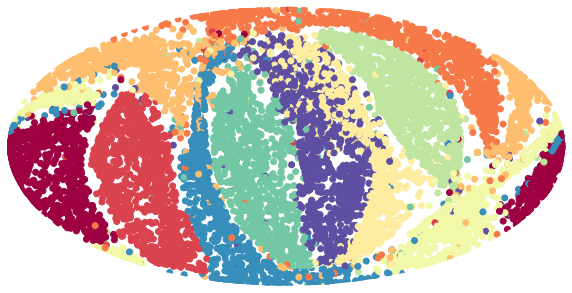}} \hfill 
        \subfloat[SSWAE]{\label{latent_space_sswae}\includegraphics[width=0.45\columnwidth]{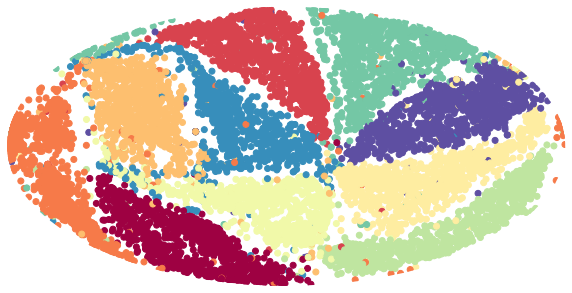}}
        \caption{Latent space of SWAE and SSWAE on MNIST for a uniform prior on $S^2$.}
        \hspace*{\fill}
        \label{fig:latent_space_unif}
    \end{minipage}
\end{figure}

\looseness=-1 Apart from showing that SSW is an effective discrepancy for learning problems defined over the sphere, the objective of this experimental Section is to show that it behaves better than using the more immediate SW in the embedding space. We first illustrate the ability to approximate different distributions by minimizing SSW \emph{w.r.t.} some target distributions on $S^2$ and by performing density estimation experiments on real earth data. Then, we apply SSW for generative modeling tasks using the framework of Sliced-Wasserstein Autoencoder and we show that we obtain competitive results with other Wasserstein Autoencoder based methods using a prior on higher dimensional hyperspheres. Complete details about the experimental settings and optimization strategies are given in \Cref{appendix:experiments_ssw}. 
The code is available online\footnote{\url{https://github.com/clbonet/Spherical_Sliced-Wasserstein}}.

\subsection{SSW as a Loss} \label{xp:gradient_flows_ssw} 

\paragraph{Gradient flow on toy data.}

We verify on the first experiments that we can learn some target distribution $\nu\in\mathcal{P}(S^{d-1})$ by minimizing SSW, \emph{i.e.} we consider the minimization problem $\argmin_{\mu}\ \ssw_p^p(\mu,\nu)$. We suppose that we have access to the target distribution $\nu$ through samples, \emph{i.e.} through $\hat{\nu}_m=\frac{1}{m}\sum_{j=1}^m \delta_{{y}_j}$ where $(y_j)_{j=1}^m$ are i.i.d samples of $\nu$. 
As target distribution, we choose a mixture of 6 well separated von Mises-Fisher distributions \citep{mardia1975statistics}. This is a fairly challenging distribution since there are 6 modes which are not connected. We show on Figure \ref{fig:gradient_flow_mixture_vmf} the Mollweide projection of the density approximated by a kernel density estimator for a distribution with 500 particles. To optimize directly over particles, we perform a Riemannian gradient descent on the sphere \citep{absil2009optimization}.

\begin{wraptable}{r}{0.4\linewidth}
    \centering
    \small
    \caption{Negative test log likelihood.} 
    \resizebox{\linewidth}{!}{
        \begin{tabular}{cccc}
            \toprule
            & Earthquake & Flood & Fire \\ \toprule
            SSW & $\bm{{0.84}_{\pm 0.07}}$ & $\bm{{1.26}_{\pm 0.05}}$ & $\bm{{0.23}_{\pm 0.18}}$ \\
            SW & ${0.94}_{\pm 0.02}$ & ${1.36}_{\pm 0.04}$ & ${0.54}_{\pm 0.37}$ \\
            Stereo & ${1.91}_{\pm 0.1}$ & ${2.00}_{\pm 0.07}$ & ${1.27}_{\pm 0.09}$ \\
            \bottomrule
        \end{tabular}
    }
    \label{tab:nll}
\end{wraptable}

\paragraph{Density estimation on earth data.} We perform density estimation on datasets first gathered by \citet{mathieu2020riemannian} which contain locations of wildfires \citep{firedataset}, floods \citep{flooddataset} or earthquakes \citep{earthquakedataset}. We use exponential map Normalizing Flows introduced in \citep{rezende2020normalizing} (see \Cref{appendix:nf_sphere}) which are invertible transformations mapping the data to some prior that we need to enforce. Here, we choose as prior the uniform distribution on $S^2$ and we learn the model using SSW. These transformations allow to evaluate exactly the density at any point. More precisely, let $T$ be such transformation, let $p_Z$ be a prior distribution on $S^2$ and $\mu$ the measure of interest, which we know from samples, \emph{i.e.} through $\hat{\mu}_n=\frac{1}{n}\sum_{i=1}^n \delta_{x_i}$. Then, we solve the following optimization problem $\min_{T}\ \ssw_2^2(T_\#\mu, p_Z)$. Once it is fitted, then the learned density $f_\mu$ can be obtained by
\begin{equation}
    \forall x\in S^2,\ f_\mu(x) = p_Z\big(T(x)\big) |\det J_T(x)|,
\end{equation}
where we used the change of variable formula.

\begin{figure}[t]
    \centering
    \hspace*{\fill}
    \subfloat[Fire]{\label{ll_fire}\includegraphics[width=0.3\columnwidth]{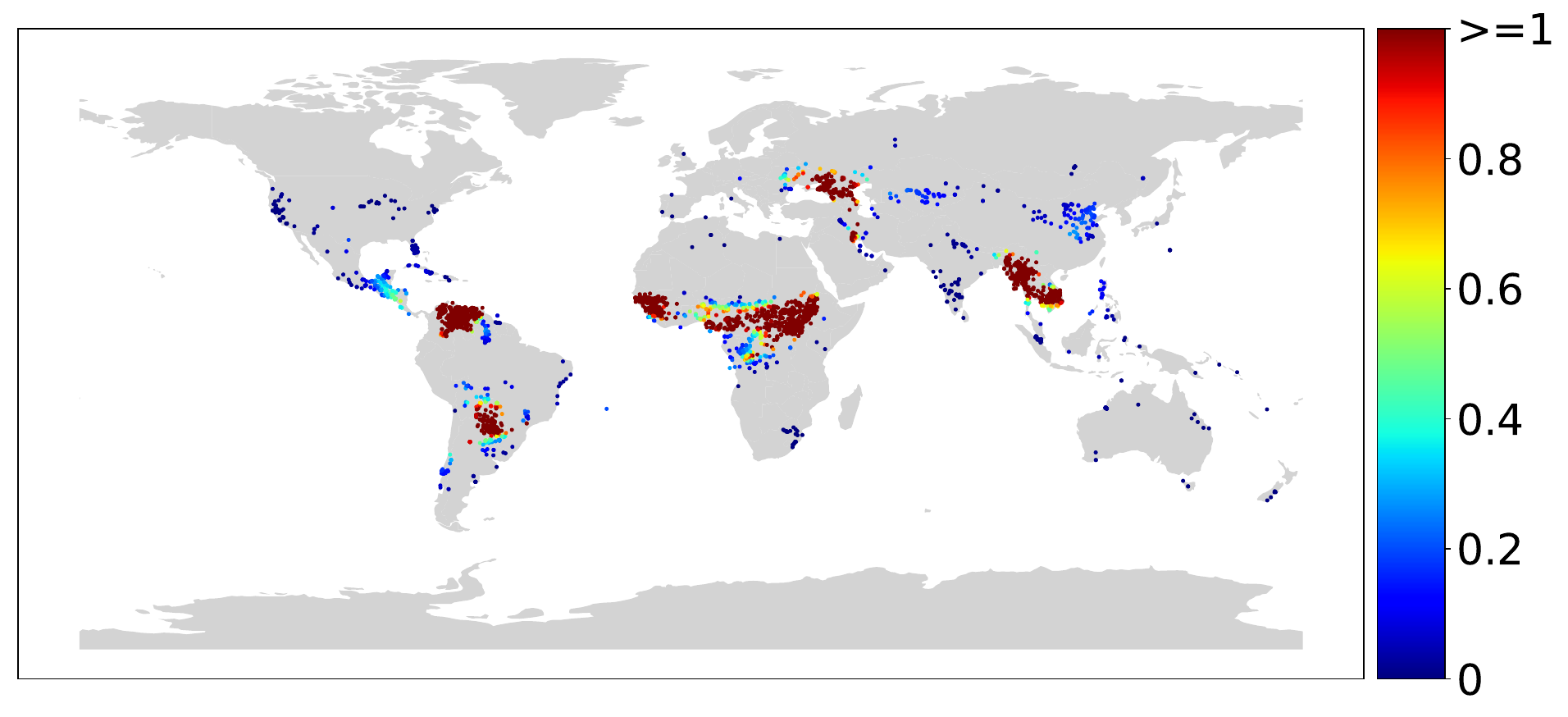}} \hfill
    \subfloat[Earthquake]{\label{ll_earthquake}\includegraphics[width=0.3\columnwidth]{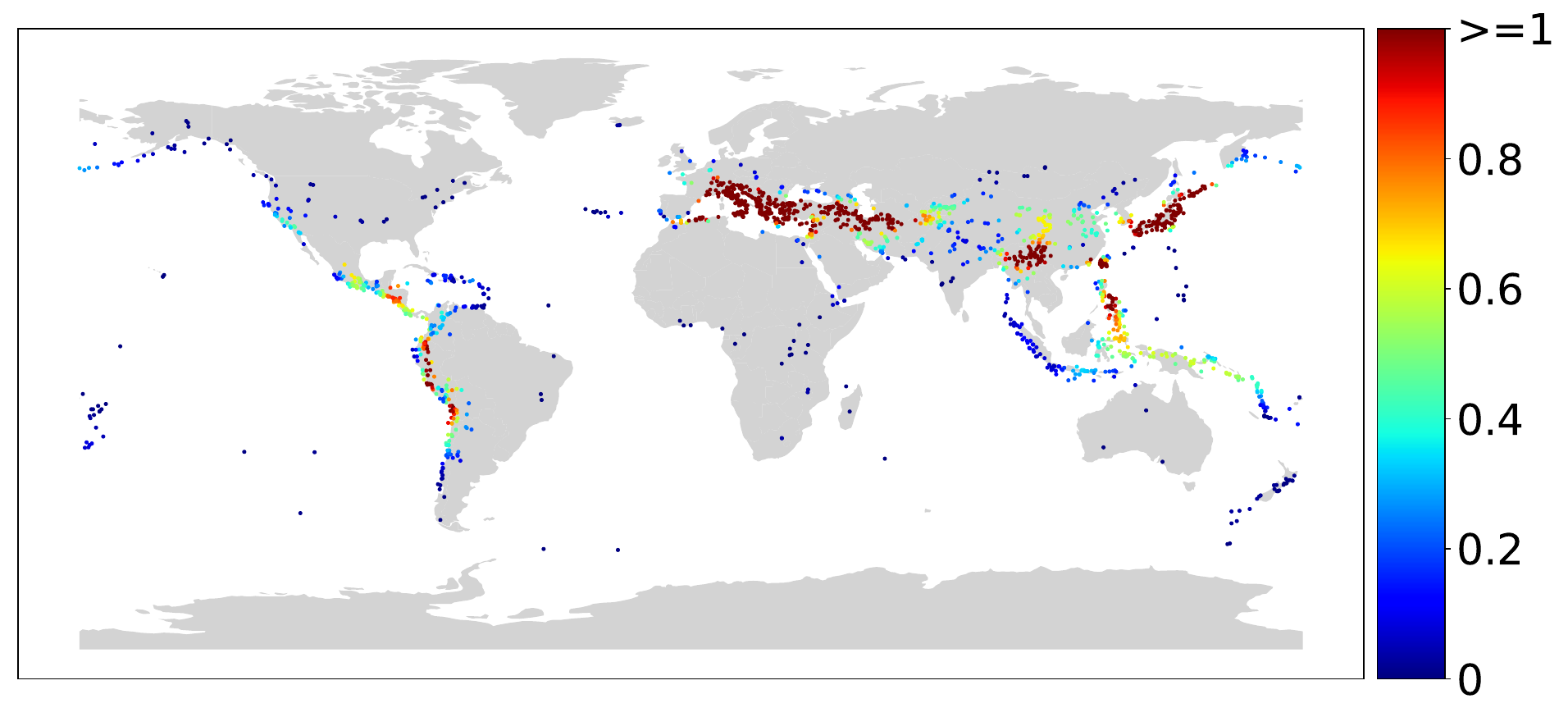}} \hfill 
    \subfloat[Flood]{\label{ll_flood}\includegraphics[width=0.3\columnwidth]{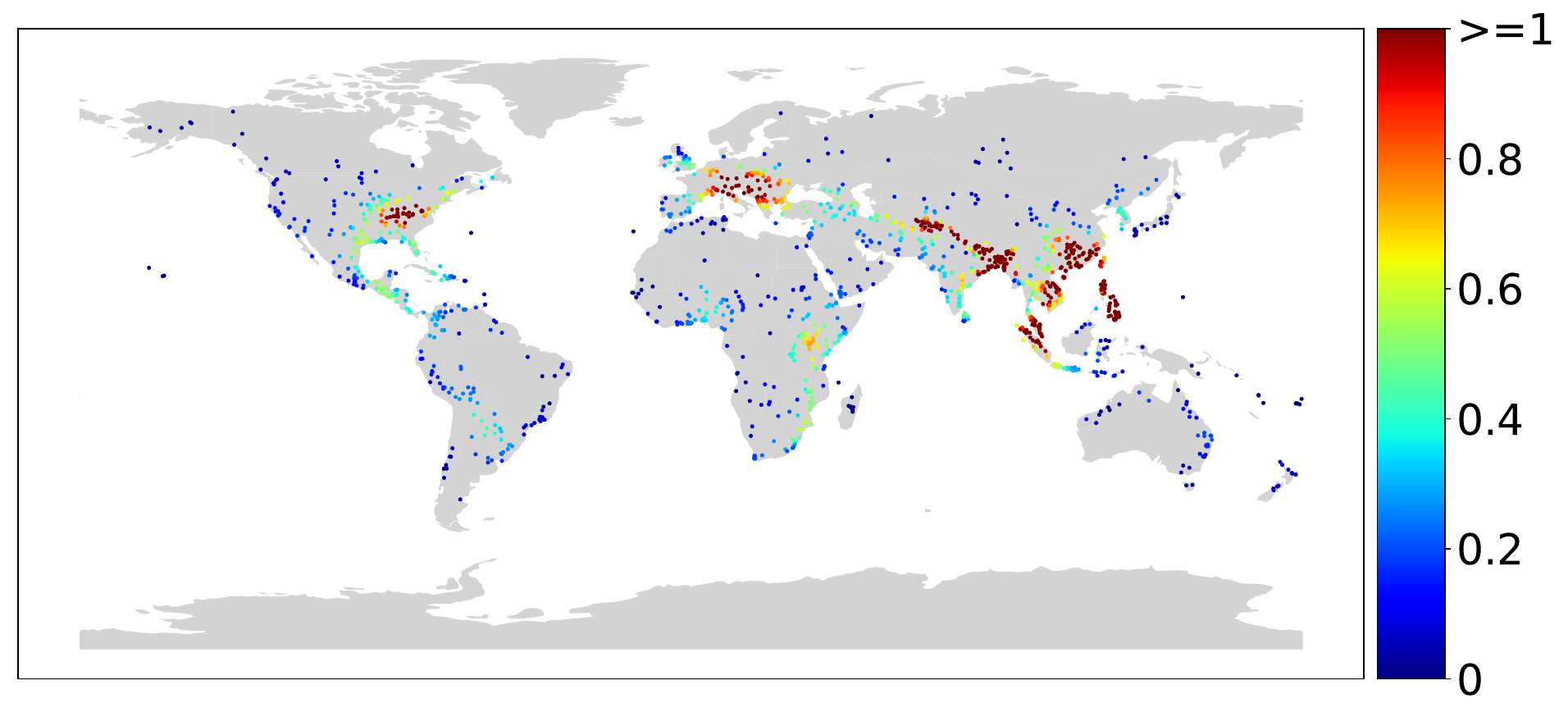}} \hfill
    \hspace*{\fill}
    \caption{Density estimation of models trained on earth data. We plot the density on the test data.}
    \label{fig:density_earth_data}
\end{figure}

We show on \Cref{fig:density_earth_data} the density of test data learned. 
We observe on this figure that the Normalizing Flows (NFs) put mass where most data points lie, and hence are able to somewhat recover the principle modes of the data. We also compare on \Cref{tab:nll} the negative test log likelihood, averaged over 5 trainings with different split of the data, between different OT metrics, namely SSW, SW and the stereographic projection model \citep{gemici2016normalizing} which first projects the data on $\mathbb{R}^2$ and use a regular NF in the projected space. We observe that SSW allows to better fit the data compared to the other OT based methods which are less suited to the sphere.

\subsection{SSW Autoencoders} \label{section:swae}

In this section, we use SSW to learn the latent space of Autoencoders (AE). We rely on the SWAE framework introduced by \citet{kolouri2018sliced}. Let $f$ be some encoder and $g$ be some decoder, denote $p_Z$ a prior distribution, then the loss minimized in SWAE is
\begin{equation}
    \mathcal{L}(f,g) = \int c\big(x,g(f(x))\big)\mathrm{d}\mu(x) + \lambda SW_2^2(f_\#\mu,p_Z),
\end{equation}
where $\mu$ is the distribution of the data for which we have access to samples. 
While VAEs \citep{kingma2013auto} rely on Variational Inference which necessitates a simple reference distribution, Sliced-Wasserstein Autoencoders, and more generally Wasserstein Autoencoders \citep{tolstikhin2018wasserstein}, can take any reference prior as no parametrization trick is needed.


In several concomitant works, it was shown that using a prior on the hypersphere can improve the results \citep{davidson2018hyperspherical, xu2018spherical}. Hence, we propose in the same fashion as \citep{kolouri2018sliced, kolouri2019generalized, patrini2020sinkhorn} to replace SW by SSW, which we denote SSWAE, and to enforce a prior on the sphere. In the following, we use the MNIST \citep{lecun-mnisthandwrittendigit-2010}, FashionMNIST \citep{xiao2017fashion} and CIFAR10 \citep{krizhevsky2009learning} datasets, and we put an $\ell^2$ normalization at the output of the encoder. As a prior, we use the uniform distribution on $S^{10}$ for MNIST and FashionMNIST, and on $S^{64}$ for CIFAR10. We compare in Table \ref{tab:fid} the Fréchet Inception Distance (FID) \citep{heusel2017gans}, for 10000 samples and averaged over 5 trainings, obtained with the Wasserstein Autoencoder (WAE) \citep{tolstikhin2018wasserstein}, the classical SWAE \citep{kolouri2018sliced}, the Sinkhorn Autoencoder (SAE) \citep{patrini2020sinkhorn} and circular GSWAE \citep{kolouri2019generalized}. We observe that we obtain fairly competitive results on the different datasets. We add on Figure \ref{fig:latent_space_unif} the latent space obtained with a uniform prior on $S^2$ on MNIST. We notably observe a better separation between classes for SSWAE.

\begin{table}[t] 
    \centering
    \small
    \caption{FID (Lower is better).} 
    \resizebox{0.5\linewidth}{!}{
        \begin{tabular}{cccc}
            \toprule
            Method / Dataset & MNIST & Fashion & CIFAR10 \\
            \toprule
            SSWAE  &  $\bm{{14.91}}_{\pm 0.32}$ & $\bm{43.94}_{\pm 0.81}$ & ${98.57}_{\pm 035}$ \\
            SWAE & ${15.18}_{\pm 0.32}$ & ${44.78}_{\pm 1.07}$ & $\bm{{98.5}_{\pm 0.45}}$ \\
            WAE-MMD IMQ &  ${18.12}_{\pm 0.62}$ & ${68.51}_{\pm 2.76}$ & ${100.14}_{\pm 0.67}$ \\
            WAE-MMD RBF & ${20.09}_{\pm 1.42}$ & ${70.58}_{\pm 1.75}$ & ${100.27}_{\pm 0.74}$\\
            SAE & ${19.39}_{\pm 0.56}$ & ${56.75}_{\pm 1.7}$ & ${99.34}_{\pm 0.96}$\\
            Circular GSWAE & ${15.01}_{\pm 0.26}$ &  ${44.65}_{\pm 1.2}$ & ${98.8}_{\pm 0.68}$ \\
            \bottomrule
        \end{tabular}
    }
    \label{tab:fid}
\end{table}

\section{Conclusion and Discussion}

In this chapter, we derive a new Sliced-Wasserstein discrepancy on the hypersphere, that comes with practical advantages when computing Optimal Transport distances on hyperspherical data. We notably showed that it is competitive or even sometimes better than other metrics defined directly on $\mathbb{R}^d$ on a variety of Machine Learning tasks, including density estimation or generative models. This work is the first, up to our knowledge, to adapt the classical Sliced-Wasserstein framework to non-trivial manifolds. The three main ingredients are: {\em i)} a closed-form for Wasserstein on the circle, {\em ii)} a closed-form solution to the projection onto great circles, and {\em iii)} a Radon transform on the Sphere. An immediate follow-up of this work would be to examine asymptotic properties as well as statistical and topological aspects. While we postulate that results comparable to the Euclidean case might be reached, the fact that the manifold is closed might bring interesting differences and justify further use of this type of discrepancies rather than their Euclidean counterparts.

\clearemptydoublepage
\cleartooddpage[\thispagestyle{empty}]
\part{Optimal Transport and Variants through Projections} \label{part:ot}

\clearemptydoublepage
\cleartooddpage[\thispagestyle{empty}]
\chapter{Gradient Flows in Sliced-Wasserstein Space} \label{chapter:swgf}

{
    \hypersetup{linkcolor=black} 
    \minitoc 
}

This chapter is based on \citep{bonet2022efficient} and aims at minimizing functionals in the space of probability measures. Such a task can traditionally be done with Wasserstein gradient flows. To solve them numerically, a possible approach is to rely on the Jordan–Kinderlehrer–Otto (JKO) scheme which is analogous to the proximal scheme in Euclidean spaces. However, it requires solving a nested optimization problem at each iteration, and is known for its computational challenges, especially in high dimension. To alleviate it, recent works propose to approximate the JKO scheme leveraging Brenier's theorem, and using gradients of Input Convex Neural Networks to parameterize the density (JKO-ICNN). However, this method comes with a high computational cost and stability issues. Instead, this work proposes to use gradient flows in the space of probability measures endowed with the Sliced-Wasserstein distance. We argue that this method is more flexible than JKO-ICNN, since SW enjoys a closed-form approximation. Thus, the density at each step can be parameterized by any generative model which alleviates the computational burden and makes it tractable in higher dimensions.

\section{Introduction}

Minimizing functionals with respect to probability measures is a ubiquitous problem in Machine Learning. Important examples are generative models such as Generative Adversarial Networks (GANs) \citep{goodfellow2014generative, arjovsky2017wasserstein}, Variational Autoencoders (VAEs) \citep{kingma2013auto} or Normalizing Flows (NFs) \citep{papamakarios2021normalizing}.

To that aim, one can rely on Wasserstein gradient flows (WGF) \citep{ambrosio2008gradient} which are curves decreasing the functional as fast as possible~\citep{santambrogio2017euclidean}. For particular functionals, these curves are known to be characterized by the solution of some partial differential equation (PDE) \citep{jordan1998variational}. Hence, to solve Wasserstein gradient flows numerically, we can solve the related PDE when it is available. However, solving a PDE can be a difficult and computationally costly task, especially in high dimensions \citep{han2018solving}. Fortunately, several alternatives exist in the literature. For example, one can approximate instead a counterpart stochastic differential equation (SDE) related to the PDE followed by the gradient flow. For the Kullback-Leibler divergence, it comes back to the so called unadjusted Langevin algorithm (ULA) \citep{roberts1996exponential, wibisono2018sampling}, but it has also been proposed for other functionals such as the Sliced-Wasserstein distance with an entropic regularization \citep{liutkus2019sliced}.

Another way to solve Wasserstein gradient flows numerically is to approximate the curve in discrete time. By using the well-known forward Euler scheme, particle schemes have been derived for diverse functionals such as the Kullback-Leibler (KL) divergence \citep{feng2021relative, wang2022projected, wang2022optimal}, the Maximum Mean Discrepancy (MMD) \citep{arbel2019maximum}, the kernel Stein discrepancy \citep{korba2021kernel} or lower bounds of the KL \citep{glaser2021kale}. \citet{salim2020wasserstein} propose instead a forward-backward discretization scheme analogously to the proximal gradient algorithm \citep{bauschke2011convex}. Yet, these methods only provide samples approximately following the gradient flow, but without any information about the underlying density.

Another time discretization possible is the so-called JKO scheme introduced by \citet{jordan1998variational}, which is analogous in probability space to the well-known proximal operator \citep{parikh2014proximal} in Hilbertian space and which corresponds to the backward Euler scheme. However, as a nested minimization problem, it is a difficult problem to handle numerically. Some works use a discretization in space (\emph{e.g.} a grid) and the entropic regularization of the Wasserstein distance~\citep{peyre2015entropic, carlier2017convergence}, which benefits from specific resolution strategies. However, those approaches do not scale to high dimensions, as the discretization of the space scales exponentially with the dimension. Very recently, it was proposed in several concomitant works \citep{mokrov2021largescale, bunne2021jkonet, alvarez-melis2022optimizing} to take advantage of Brenier's theorem \citep{brenier1991polar} and model the Optimal Transport map (Monge map) as the gradient of a convex function with Input Convex Neural Networks (ICNN) \citep{amos2017input}. By solving the JKO scheme with this parameterization, these models, called JKO-ICNN, handle higher dimension problems well. Yet, a drawback of JKO-ICNN is the training time due to a number of evaluations of the gradient of each ICNN that is quadratic in the number of JKO iterations. It also requires to backpropagate through the gradient which is challenging in high dimensions, even though stochastic methods were proposed in \citep{huang2021convex} to alleviate it. Moreover, it has also been observed in several works that ICNNs have a poor expressiveness \citep{korotin2021wasserstein, korotin2021neural, rout2022generative} and that we should rather directly estimate the gradient of convex functions by neural networks \citep{saremi2019approximating, richterpowell2021input, chaudhari2023learning}.
Other recent works proposed to use the JKO scheme by either exploiting variational formulations of functionals in order to avoid the evaluation of densities and allowing to use more general neural networks in \citep{fan2022variational}, or by learning directly the density in \citep{hwang2021deep}.

In parallel, it was proposed to endow the space of probability measures with other distances than Wasserstein. For example, \citet{gallouet2017jko} study a JKO scheme in the space endowed by the Kantorovich-Fisher-Rao distance. However, this still requires a costly JKO step. Several particle schemes were derived as gradient flows into this space \citep{lu2019accelerating, zhang2021dpvi}. We can also cite Kalman-Wasserstein gradient flows \citep{garbuno2020interacting} or the Stein variational gradient descent \citep{liu2016stein, liu2017stein, duncan2019geometry} which can be seen as gradient flows in the space of probabilities endowed by a generalization of the Wasserstein distance. However, the JKO schemes of these different metrics are not easily tractable in practice.

\paragraph{Contributions.}

\looseness=-1 In the following, we propose to study the JKO scheme in the space of probability distributions endowed with the Sliced-Wasserstein (SW) distance \citep{rabin2012wasserstein}. This novel and simple modification of the original problem comes with several benefits, mostly linked to the fact that this distance is easily differentiable and computationally more tractable than the Wasserstein distance. We first derive some properties of this new class of flows and discuss links with Wasserstein gradient flows. Notably, we observe empirically for both gradient flows the same dynamic, up to a time dilation of parameter the dimension of the space. 
Then, we show that it is possible to minimize functionals and learn the stationary distributions in high dimensions, on toy datasets as well as real image datasets, using \emph{e.g.} neural networks. In particular, we propose to use Normalizing Flows for functionals which involve the density, such as the negative entropy.
Finally, we exhibit several examples for which our strategy performs better than JKO-ICNN, either \emph{w.r.t.} to computation times and/or \emph{w.r.t.} the quality of the final solutions.

\section{Background on Gradient Flows}

In this chapter, we are interested in finding a numerical solution to gradient flows in probability spaces. Such problems generally arise when minimizing a functional $\mathcal{F}$ defined on $\mathcal{P}(\mathbb{R}^d)$:
\begin{equation}
    \min_{\mu\in\mathcal{P}(\mathbb{R}^d)}\ \mathcal{F}(\mu),
\end{equation}
but they can also be defined implicitly through their dynamics, expressed as partial differential equations. JKO schemes are implicit optimization methods that operate on particular discretizations of these problems and consider the natural metric of $\mathcal{P}(\mathbb{R}^d)$ to be the Wasserstein distance. Recalling our goal is to study similar schemes with an alternative, computationally friendly metric (SW), we start by formally defining the notion of gradient flows in Euclidean spaces, before switching to probability spaces. We finally give a rapid overview of existing numerical schemes.

\subsection{Gradient Flows in Euclidean Spaces}

Let $F:\mathbb{R}^d\to\mathbb{R}$ be a functional. A gradient flow of $F$ is a curve (\emph{i.e.} a continuous function from $\mathbb{R}_+$ to $\mathbb{R}^d$) which decreases $F$ as much as possible along it. If $F$ is differentiable, then a gradient flow $x:[0,T]\to\mathbb{R}^d$ solves the following Cauchy problem \citep{santambrogio2017euclidean}
\begin{equation}
    \begin{cases}
        \frac{\mathrm{d}x(t)}{\mathrm{d}t} = -\nabla F(x(t)), \\
        x(0) = x_0.
    \end{cases}
\end{equation}
Under conditions on $F$ (\emph{e.g.} $\nabla F$ Lipschitz continuous, $F$ convex or semi-convex), this problem admits a unique solution which can be approximated using numerical schemes for ordinary differential equations such as the explicit or the implicit Euler scheme. For the former, we recover the regular gradient descent, and for the latter, we recover the proximal point algorithm \citep{parikh2014proximal}: let $\tau>0$,
\begin{equation} \label{eq:prox}
    x_{k+1}^\tau \in \argmin_{x}\  \frac{\|x-x_k^\tau\|_2^2}{2\tau}+F(x) = \mathrm{prox}_{\tau F}(x_k^\tau).
\end{equation}

This formulation does not use any gradient, and can therefore be used in any metric space by replacing $\|x-x_k^\tau\|_2^2 = d(x,x_k^\tau)^2$ 
with the right squared distance.

\subsection{Gradient Flows in Probability Spaces}

To define gradient flows in the space of probability measures, we first need a metric. We restrict our analysis to probability measures with finite moments of order 2: $\mathcal{P}_2(\mathbb{R}^d)=\{\mu\in\mathcal{P}(\mathbb{R}^d),\ \int \|x\|^2 \mathrm{d}\mu(x)<+\infty\}$. Then, a possible distance on $\mathcal{P}_2(\mathbb{R}^d)$ is the Wasserstein distance. 

Now, by endowing the space of measures with $W_2$, we can define the Wasserstein gradient flow of a functional $\mathcal{F}:\mathcal{P}_2(\mathbb{R}^d)\to\mathbb{R}$ by plugging $W_2$ in (\ref{eq:prox}) which becomes
\begin{equation} \label{eq:jko}
    \mu_{k+1}^\tau \in \argmin_{\mu\in \mathcal{P}_2(\mathbb{R}^d)}\ \frac{W_2^2(\mu,\mu_k^\tau)}{2\tau}+\mathcal{F}(\mu).
\end{equation}
The gradient flow is then the limit of the sequence of minimizers when $\tau\to0$. This scheme was introduced in the seminal work of Jordan, Kinderlehrer and Otto \citep{jordan1998variational} and is therefore referred to as the JKO scheme. In this work, the authors showed that gradient flows are linked to PDEs, and in particular with the Fokker-Planck equation when the functional $\mathcal{F}$ is of the form
\begin{equation} \label{eq:FokkerPlanckFunctional}
    \mathcal{F}(\mu)=\int V\mathrm{d}\mu + \mathcal{H}(\mu)
\end{equation}
where $V$ is some potential function and $\mathcal{H}$ is the negative entropy: let $\sigma$ denote the Lebesgue measure,
\begin{equation}
    \mathcal{H}(\mu) = \left\{
    \begin{array}{ll}
        \int \log\big(\rho(x)\big) \rho(x)\ \mathrm{d}x & \mbox{ if } \mathrm{d}\mu=\rho\mathrm{d}\sigma \\
        +\infty & \mbox{ otherwise}.
    \end{array}\right.
\end{equation}
Then, the limit of $(\mu^\tau)_\tau$ when $\tau\to 0$ is a curve $t\mapsto\mu_t$ such that for all $t>0$, $\mu_t$ has a density $\rho_t$. The curve $\rho$ satisfies (weakly) the Fokker-Planck PDE
\begin{equation} \label{eq:FokkerPlanck}
    \frac{\partial \rho}{\partial t} = \mathrm{div}(\rho\nabla V)+\Delta \rho.
\end{equation}
By satisfying weakly the PDE, we mean that for all test functions $\xi\in C^\infty_c(]0,+\infty[\times \mathbb{R}^d)$ (smooth with compact support),
\begin{equation}
    \int_0^{+\infty}\int_{\mathbb{R}^d} \left(\frac{\partial \xi}{\partial t}(t,x) + \langle \nabla V(x), \nabla_x\xi(t,x)\rangle - \Delta\xi(t,x) \right)\ \mathrm{d}\rho_t(x) \mathrm{d}t = - \int \xi(0,x)\ \mathrm{d}\rho_0(x).
\end{equation}

Note that many other functional can be plugged in \eqref{eq:jko}, defining different PDEs. We introduce here the Fokker-Planck PDE as a classical example, since the functional is connected to the Kullback-Leibler (KL) divergence, as taking a target distribution $\nu$ with a density $q(x) \propto e^{-V(x)}$,
\begin{equation}
    \begin{aligned}
        \kl(\mu||\nu) &= \mathbb{E}_\mu\left[\log\left(\frac{\rho(X)}{q(X)}\right)\right] \\
        &= \int \log\big(\rho(x)\big)\rho(x)\ \mathrm{d}x - \int \log \big(q(x)\big)\ \mathrm{d}\mu(x) \\
        &= \mathcal{H}(\mu) + \int V(x)\ \mathrm{d}\mu(x) + cst,
    \end{aligned}
\end{equation}
and its Wasserstein gradient flow is connected to many classical  algorithms such as the unadjusted Langevin algorithm (ULA) \citep{wibisono2018sampling}. But we will also use other functionals in Section \ref{section:xps} such as SW or the interaction functional, defined for regular enough $W$ as
\begin{equation} \label{eq:interaction_functional}
    \mathcal{W}(\mu) = \frac12 \iint W(x-y)\ \mathrm{d}\mu(x)\mathrm{d}\mu(y),
\end{equation}
which admits as Wasserstein gradient flow the aggregation equation
 \citep[Chapter 8]{santambrogio2015optimal}
\begin{equation}
    \frac{\partial\rho}{\partial t} = \mathrm{div}\big(\rho(\nabla W*\rho)\big)
\end{equation}
where $*$ denotes the convolution operation.


\subsection{Numerical Methods to solve the JKO Scheme}


Being composed of two nested optimization problems, solving \Cref{eq:jko} is not simple as it requires solving an Optimal Transport problem as each step.

Several strategies have been used to tackle this difficulty. For example, \citet{laborde2016interacting} rewrites (\ref{eq:jko}) as a convex minimization problem using the Benamou-Brenier dynamic formulation of the Wasserstein distance \citep{benamou2000computational}. \citet{peyre2015entropic} approximates the JKO scheme by using the entropic regularization and rewriting the problem with respect to the Kullback-Leibler proximal operator. The problem becomes easier to solve using Dykstra's algorithm \citep{dykstra1985iterative}. This scheme was proved to converge to the right PDE in \citep{carlier2017convergence}. Note that one might also consider using the Sinkhorn divergence \citep{ramdas2017wasserstein, feydy2019interpolating} with \emph{e.g.} neural networks to parameterize the distributions as it is differentiable, and it was shown to be a good approximation of the Wasserstein distance \citep{chizat2020faster}. It was proposed to use the dual formulation in other works such as \citep{caluya2019proximal} or \citep{frogner2020approximate}. \citet{cances2020variational} proposed to linearize the Wasserstein distance using the weighted Sobolev approximation \citep{villani2021topics, peyre2018comparison}.

More recently, \citet{mokrov2021largescale} and \citet{alvarez-melis2022optimizing}, following \citet{benamou2016discretization}, have proposed to exploit Brenier's theorem by rewriting the JKO scheme as
\begin{equation}
    u_{k+1}^\tau\in\argmin_{u\ \mathrm{convex}}\ \frac{1}{2\tau}\int\|\nabla u(x)-x\|_2^2\ \mathrm{d}\mu_k^\tau(x) + \mathcal{F}\big((\nabla u)_\#\mu_k^\tau\big)         
\end{equation}
and by modeling the probability measures as $\mu_{k+1}^\tau = (\nabla u_{k+1}^\tau)_\#\mu_k^\tau$. 
Then, to solve it numerically, they model convex functions using ICNNs \citep{amos2017input}:
\begin{equation}
    \theta_{k+1}^\tau\in\argmin_{\theta\in\{\theta,u_\theta\in\mathrm{ICNN}\}}\ \frac{1}{2\tau}\int\|\nabla_x u_\theta(x)-x\|_2^2\ \mathrm{d}\mu_k^\tau(x) + \mathcal{F}\big((\nabla_x u_\theta)_\#\mu_k^\tau\big).
\end{equation}
In the remainder, this method is denoted as JKO-ICNN. \citet{bunne2021jkonet} also proposed to use ICNNs into the JKO scheme, but with a different objective of learning the functional from samples trajectories along the timesteps. Lastly, \citet{fan2022variational} proposed to learn directly the Monge map $T$ by solving at each step the following problem:
\begin{equation}
    T_{k+1}^\tau \in \argmin_T\ \frac{1}{2\tau} \int \|T(x)-x\|_2^2\ \mathrm{d}\mu_k^\tau(x) + \mathcal{F}(T_\#\mu_k^\tau)
\end{equation}
and by using variational formulations for functionals involving the density. This formulation requires only to use samples from the measure. However, it needs to be derived for each functional, and involves minimax optimization problems which are notoriously hard to train \citep{arjovsky2017towards,bond2021deep}.


\subsection{More General Background on Wasserstein Gradient Flows} \label{sec:bg_wgf}

Before diving into Sliced-Wasserstein gradient flows, let us introduce other ways to compute Wasserstein gradient flows in practice. This Section is not necessary to understand our contributions in the remainder of the chapter, but will present some methods, different from JKO-ICNN, which we will use as baselines. First, we will introduce formally the Wasserstein gradient flows, and notably the Wasserstein gradient which we will use to present the forward Euler scheme.

\paragraph{Gradient Flows in Wasserstein Space.}

\looseness=-1 First, let us formalize the characterization of gradient flows in Wasserstein space. We mentioned earlier that the limit $\tau\to 0$ of the JKO scheme satisfies a PDE. This PDE is called a continuity equation. More precisely, let $T>0$ and $\mu:[0,T]\to \mathcal{P}_2(\mathbb{R}^d)$ a curve, then it satisfies a continuity equation if there exists a velocity field $(v_t)_{t\in [0,T]}$, such that $v_t\in L^2(\mu_t)$ and satisfies weakly (in the distributional sense)
\begin{equation}
    \frac{\partial \mu_t}{\partial t} + \divop (\mu_t v_t) = 0,
\end{equation}
\emph{i.e.} for all $\xi \in C_c^\infty([0,T[\times \mathbb{R}^d)$,
\begin{equation}
    \int_0^T \int_{\mathbb{R}^d} \left(\frac{\partial \xi}{\partial t}(t,x) - \langle v_t(x), \nabla_x \xi(t,x)\rangle\right)\ \mathrm{d}\mu_t(x) \mathrm{d}t = 0.
\end{equation}
This equation describes the evolution of the density along time. This is equivalent in the Lagrangian formulation as seeing that the particles $x_t \sim \mu_t$ are driven by the velocity vector field $v_t$, \emph{i.e.} they satisfy the following ODE $\frac{\mathrm{d}x_t}{\mathrm{d}t} = v_t(x_t)$ \citep[Proposition 8.1.8]{ambrosio2008gradient}. We refer to \citep[Chapter 5.3]{santambrogio2015optimal} for more details such as the existence of such a velocity field. In particular, when we aim at minimizing a functional $\mathcal{F}$, a suitable velocity field is the Wasserstein gradient which we introduce now.

\looseness=-1 For a functional $\mathcal{F}$, we call $\frac{\delta\mathcal{F}}{\delta \mu}(\mu)$ the first variation of $\mathcal{F}$ \citep[Definition 7.12]{santambrogio2015optimal}, if it exists, the unique function (up to additive constants) such that
\begin{equation}
    \frac{\mathrm{d}\mathcal{F}}{\mathrm{d}t}(\mu+t\chi)\Big|_{t=0} = \lim_{t\to 0} \ \frac{\mathcal{F}(\mu+t\chi)-\mathcal{F}(\mu)}{t} = \int \frac{\delta\mathcal{F}}{\delta\mu}(\mu)\ \mathrm{d}\chi,
\end{equation}
where for $\Tilde{\mu}\in\mathcal{P}_2(\mathbb{R}^d)$, $\chi = \Tilde{\mu} - \mu$ is a perturbation around $\mu$ which satisfies $\int \mathrm{d}\chi = 0$.  Then, we define the Wasserstein gradient of $\mathcal{F}$, which we denote $\nabla_{W_2}\mathcal{F}$, as $\nabla_{W_2}\mathcal{F}(\mu) = \nabla \frac{\delta\mathcal{F}}{\delta \mu}(\mu)$. Now, we say that $\mu:[0,T]\to \mathcal{P}_2(\mathbb{R}^d)$ is a Wasserstein gradient flow of $\mathcal{F}$ if it satisfies distributionally the following continuity equation:
\begin{equation}
    \frac{\partial \mu_t}{\partial t} - \divop\big(\mu_t \nabla_{W_2}\mathcal{F}(\mu)\big) = 0.
\end{equation}
For more details on gradient flows in Wasserstein space, we refer to \citep[Chapter 10]{ambrosio2008gradient} and in particular to Lemma 10.4.1 for the Wasserstein gradient. Note that the Wasserstein gradient is different from the gradient in Wasserstein space, which is defined using its Riemannian structure \citep{otto2001geometry}.


\paragraph{Particle Scheme.} Now that we know how to find the PDE, let us discuss some methods to sample from its solution in practice. On one hand, we saw that using a backward Euler scheme, we can use the so-called JKO scheme. The other natural counterpart is to use the forward Euler scheme, which translates as 
\begin{equation}
    \forall k\ge 0,\ \mu_{k+1}^\tau = \big(\id - \tau \nabla_{W_2}\mathcal{F}(\mu_k^\tau)\big)_\#\mu_k^\tau.
\end{equation}
Using a Lagrangian approximation with particles with $\hat{\mu}_{k+1} = \frac{1}{n} \sum_{i=1}^n \delta_{x_i^{(k)}}$, then we obtain the following update rule
\begin{equation}
    \forall i \in \{1,\dots, n\},\ x_i^{(k+1)} = x_i^{(k)} - \tau \nabla_{W_2}\mathcal{F}(\hat{\mu_k})(x_i^{(k)}).
\end{equation}

Now, let us provide two examples which will be of much interest in the experiment section. First, we will study $\mathcal{F}(\mu) = \kl(\mu||\nu)$ where $\nu$ has a density $q\propto e^{-V}$. Let's note $p$ the density of $\mu$. Then, the Wasserstein gradient of $\mathcal{F}$ is (see \emph{e.g.} \citep{feng2021relative})
\begin{equation}
    \nabla_{W_2}\mathcal{F}(\mu) = \nabla \log \frac{p}{q} = \nabla (\log p + V).
\end{equation}
Hence, using the Forward-Euler scheme, we obtain for the update equation
\begin{equation}
    \forall i,\ x^{(k+1)}_i = x^{(k)}_i - \tau \left(\nabla \log p(x^{(k)}_i) + \nabla V(x^{(k)}_i)\right).
\end{equation}
However, the density $p_k$ of $\hat{\mu}_k$ is usually not available. Hence, several works propose to approximate it, either using kernel density estimators \citep{wang2022projected} or by approximating the log density ratios with neural networks \citep{feng2021relative, ansari2021refining, wang2022optimal, heng2023generative, yi2023monoflow}. There are other possible approximations. For example, restricting the velocity field to be in a Reproducing Kernel Hilbert Space (RKHS), we obtain the Stein Variational Gradient Descent \citep{liu2016stein} with the advantage that it does not involve evaluating the density.

Another solution to avoid evaluating unknown densities is to use that Fokker-Planck type PDEs have a counterpart SDE \citep{bogachev2022fokker, liutkus2019sliced}, which solutions follow the same dynamic. For example, for the KL divergence, we saw earlier that the Wasserstein gradient flow follows a Fokker-Planck equation, which admits as counterpart PDE the Langevin equation \citep{mackey2011time, wibisono2018sampling}
\begin{equation}
    \mathrm{d}X_t = - \nabla V(X_t) \mathrm{d}t + \sqrt{2}\mathrm{d}W_t,
\end{equation}
where $W_t$ is a standard Brownian motion. Using the Euler-Maruyama scheme, we can simulate from this SDE with the following particle scheme
\begin{equation}
    \forall i,\ x^{(k+1)}_i = x^{(k)}_i - \tau \nabla V(x^{(k)}_i) + \sqrt{2\tau} Z_i,
\end{equation}
where $Z\sim\mathcal{N}(0,I_d)$. This particle scheme is also well known as the Unadjusted Langevin Algorithm (ULA), which has been extensively studied in the Markov chain Monte-Carlo (MCMC) community \citep{roberts1996exponential, durmus2017nonasymptotic, dalalyan2017theoretical, altschuler2022resolving}. Notably, the Wasserstein gradient flow point of view helped to derive new convergence rates \citep{cheng2018convergence, durmus2019analysis, balasubramanian2022towards}.

Here, we focused on the (reverse) Kullback-Leibler divergence as a discrepancy to minimize with respect to a target measure which we aim to learn. Actually, there are many different discrepancies which can be used instead of the KL. For example, one might consider more generally f-divergences \citep{gao2019deep}. But these functionals can only be used when we have access to the density of the target distribution up to a constant. When we have access to samples from the target, we can use other functionals such as the MMD or the Sliced-Wasserstein distance, on which we will focus now, and that we introduced in \Cref{sec:sw}. 
Let $\mathcal{F}(\mu) = \frac12 \sw_2^2(\mu,\nu)$, \citet{bonnotte2013unidimensional} first studied its Wasserstein gradient flow and found the continuity equation it follows. \citet{liutkus2019sliced} extended the result to $\mathcal{F}(\mu) = \frac12 \sw_2^2(\mu,\nu) + \lambda \mathcal{H}(\mu)$, where they additionally added the negative entropy as regularization in order to introduce the noise inherent to generative models. Then, under mild conditions, they showed that the Wasserstein gradient flow $\rho$ of $\mathcal{F}$ satisfies the following continuity equation:
\begin{equation}
    \frac{\partial \rho_t}{\partial t} + \divop(\rho_t v_t) = \Delta \rho_t,
\end{equation}
where
\begin{equation}
    v_t(x) = - \int_{S^{d-1}} \psi_{t,\theta}'\big(\langle x,\theta\rangle\big)\theta\ \mathrm{d}\lambda(\theta),    
\end{equation}
with $\psi_{t,\theta}$ the Kantorovich potential between $P^\theta_\#\mu_t$ and $P^\theta_\#\nu$.

From this continuity equation, they further derived the counterpart SDE
\begin{equation}
    \mathrm{d}X_t = v_t(X_t) \mathrm{d}t + \sqrt{2\lambda} \mathrm{d}W_t,
\end{equation}
and approximated it with the Euler-Maruyama scheme. The final particle scheme approximating the Wasserstein gradient flow of $\mathcal{F}$ can then be obtained by 
\begin{equation}
    \forall i,\ x^{(k+1)}_i = x^{(k)}_i + \tau \hat{v}_k(x^{(k)}_i) + \sqrt{2\lambda\tau} Z_i,
\end{equation}
where $Z\sim \mathcal{N}(0,I_d)$, and the velocity field is approximated by 
\begin{equation}
    \hat{v}_k(x) = - \frac{1}{L} \sum_{\ell=1}^L \psi_{k,\theta_\ell}'\big(\langle \theta_\ell, x\rangle\big) \theta_\ell,
\end{equation}
with $\psi_{k,\theta}$ the Kantorovich potential between $P^\theta_\# \mu_k$ and $P^\theta_\#\nu$. In the following, we will call this scheme the Sliced-Wasserstein flows (SWF).

While these previous methods work in practice, they also suffer from some drawbacks. First of all, a scheme needs to be derived individually for each functional. Second, if we want new samples, we must run the whole scheme again or learn an amortized representation \citep{wang2016learning}. Moreover, 
the Euler-Maruyama discretization of SDEs does not necessarily converge to the right stationary measure as it is a biased algorithm \citep{roberts1996exponential, durmus2018efficient}, often requiring an additional correction step such as a Metropolis-Hasting step \citep{metropolis1953equation, hastings1970monte}. Therefore, in this chapter, we advocate using the Backward Euler scheme.

\section{Sliced-Wasserstein Gradient Flows} \label{sec:SWGFs}

As seen in the previous section, solving numerically (\ref{eq:jko}) is a challenging problem. To tackle high-dimensional settings, one could benefit from neural networks, such as generative models, that are known to model high-dimensional distributions accurately. The problem being not directly differentiable, previous works relied on Brenier's theorem and modeled convex functions through ICNNs, which results in JKO-ICNN. However, this method is very costly to train. For a JKO scheme of $k$ steps, 
it requires $O(k^2)$ evaluations of gradients \citep{mokrov2021largescale} which can be a huge price to pay when the dynamic is very long. Moreover, it requires to backpropagate through gradients, and to compute the determinant of the Jacobian when we need to evaluate the likelihood (assuming the ICNN is strictly convex). The method of \citet{fan2022variational}, while not using ICNNs, also requires $O(k^2)$ evaluations of neural networks, as well as to solve a minimax optimization problem at each step.

Here, we propose instead to use the space of probability measures endowed with the Sliced-Wasserstein (SW) distance by modifying adequately the JKO scheme. Surprisingly enough, this class of gradient flows, which are very easy to compute, has never been considered numerically in the literature. 

In this Section, we first recall some motivations to use SW as a proxy of the Wasserstein distance for the gradient flow problem. 
We then study some properties of the scheme and discuss links with Wasserstein gradient flows. Since this metric is known in closed-form, the JKO scheme is more tractable numerically and can be approximated in several ways that we describe in Section \ref{section:swjko_practice}.

\subsection{Motivations}



\paragraph{Computational Properties.}

Firstly, $\sw_2$ is very easy to compute by a Monte-Carlo approximation (see \Cref{sec:sw}). It is also differentiable, and hence using \emph{e.g.} the Python Optimal Transport (POT) library \citep{flamary2021pot}, we can backpropagate \emph{w.r.t.} parameters or weights parameterizing the distributions (see \Cref{section:swjko_practice}). Note that some libraries allow to directly backpropagate through Wasserstein. However, theoretically, we only have access to a subgradient in that case \citep[Proposition 1]{cuturi2014fast}, and the computational complexity is bigger ($O(n^3\log n)$ versus $O(n\log n)$ for SW with $n$ the number of samples). Besides, libraries such as POT first compute the optimal plan and then differentiate, and hence cannot use the GPU. Moreover, contrary to $W_2$, the sample complexity of SW does not depend on the dimension \citep{nadjahi2020statistical} which is important to overcome the curse of dimensionality. However, it is known to be hard to approximate in high-dimension \citep{deshpande2019max} since the error of the Monte-Carlo estimates is impacted by the number of projections in practice \citep{nadjahi2020statistical}. Nevertheless, several variants could also be used. Moreover, a deterministic approach using a concentration of measure phenomenon (and hence being more accurate in high dimension) was recently proposed by \citet{nadjahi2021fast} to approximate $\sw_2$.

\paragraph{Link with Wasserstein.}

The Sliced-Wasserstein distance also has many properties related to the Wasserstein distance. First, they actually induce the same topology \citep{nadjahi2019asymptotic, bayraktar2021strong} which might justify using SW as a proxy of Wasserstein. Moreover, as showed in Chapter 5 of \citet{bonnotte2013unidimensional}, they can be related on compact sets by the following inequalities, let $R>0$, for all $\mu,\nu\in\mathcal{P}(B(0,R))$,
\begin{equation}
    \sw_2^2(\mu,\nu) \le c_{d}^2 W_2^2(\mu,\nu)\le C_{d}^2 \sw_2^{\frac{1}{d+1}}(\mu,\nu),
\end{equation}
with $c_d^2 = \frac{1}{d}$ and $C_d$ some constant.

\looseness=-1 Hence, from these properties, we can wonder whether their gradient flows are related or not, or even better, whether they are the same or not. This property was initially conjectured by Filippo Santambrogio\footnote{in private communications}. Some previous works started to gather some hints on this question. For example, \citet{candau_tilh} showed that, while $(\mathcal{P}_2(\mathbb{R}^d),\sw_2)$ is not a geodesic space, the minimal length (in metric space, Definition 2.4 in \citep{santambrogio2017euclidean}) connecting two measures is $W_2$ up to a constant (which is actually $c_{d}$). We refer to \Cref{paragraph:geodesics_sw} for more details about these results.

\subsection{Definition and Properties of Sliced-Wasserstein Gradient Flows} \label{section:discussion_swgf}

Instead of solving the regular JKO scheme (\ref{eq:jko}), we propose to introduce a SW-JKO scheme, let $\mu_0\in\mathcal{P}_2(\mathbb{R}^d)$,
\begin{equation} \label{eq:swjko}
    \forall k\ge 0,\ \mu_{k+1}^\tau \in \argmin_{\mu\in\mathcal{P}_2(\mathbb{R}^d)}\ \frac{\sw_2^2(\mu,\mu_k^\tau)}{2\tau}+\mathcal{F}(\mu)
\end{equation}
in which we replaced the Wasserstein distance by $\sw_2$. 

To study gradient flows and show that they are well defined, we first have to check that discrete solutions of the problem \eqref{eq:swjko} indeed exist. Then, we have to check that we can pass to the limit $\tau\to 0$ and that the limit satisfies gradient flows properties. These limit curves will be called Sliced-Wasserstein gradient flows (SWGFs).

In the following, we restrain ourselves to measures on $\mathcal{P}_2(K)$ where $K\subset \mathbb{R}^d$ is a compact set. We report some properties of the scheme \eqref{eq:swjko} such as the existence and uniqueness of the minimizer. 

\begin{proposition} \label{prop:minimizer}
    Let $\mathcal{F}:\mathcal{P}_2(K)\to \mathbb{R}$ be a lower semi continuous functional, then the scheme (\ref{eq:swjko}) admits a minimizer. Moreover, it is unique if $\mu_k^\tau$ is absolutely continuous and $\mathcal{F}$ convex or if $\mathcal{F}$ is strictly convex.
\end{proposition}

\begin{proof}
    See \Cref{proof:prop_minimizer}.
\end{proof}

This proposition shows that the problem is well defined for convex lower semi continuous functionals since we can find at least a minimizer at each step. The assumptions on $\mathcal{F}$ are fairly standard and will apply for diverse functionals such as for example \eqref{eq:FokkerPlanckFunctional} or \eqref{eq:interaction_functional} for $V$ and $W$ regular enough.

\begin{proposition} \label{prop:nonincreasing}
    The functional $\mathcal{F}$ is non increasing along the sequence of minimizers $(\mu_k^\tau)_k$.
\end{proposition}

\begin{proof}
    Proof of \Cref{proof:prop_nonincreasing}.
\end{proof}

As the ultimate goal is to find the minimizer of the functional, this proposition assures us that the solution will decrease $\mathcal{F}$ along it at each step. If $\mathcal{F}$ is bounded below, then the sequence $\big(\mathcal{F}(\mu_k^\tau)\big)_k$ will converge (since it is non increasing).

More generally, by defining the piecewise constant interpolation as $\mu^\tau(0)=\mu_0$ and for all $k\ge 0$, $t\in]k\tau,(k+1)\tau]$, $\mu^\tau(t)=\mu_{k+1}^\tau$, we can show that for all $t<s$, $\sw_2\big(\mu^\tau(t),\mu^\tau(s)\big)\le C\big(|t-s|^\frac12 + \tau^\frac12\big)$. Following \citet{santambrogio2017euclidean}, we can apply the Ascoli-Arzelà theorem \citep[Box 1.7]{santambrogio2015optimal} and extract a converging subsequence. However, the limit when $\tau\to 0$ is possibly not unique and has no \emph{a priori} relation with $\mathcal{F}$. 
Since $(\mathcal{P}_2(\mathbb{R}^d), \sw_2)$ is not a geodesic space, but rather a ``pseudo-geodesic'' space whose true geodesics are $c_d W_2$ \citep{candau_tilh} (see \Cref{paragraph:geodesics_sw}), we cannot directly apply the theory introduced in \citep{ambrosio2008gradient}.
We leave for future work the study of the theoretical properties of the limit. Nevertheless, we conjecture that in the limit $t\to\infty$, SWGFs converge toward the same measure as for WGFs. We will study it empirically in \Cref{section:xps} by showing that we are able to find as good minima as WGFs for different functionals.


\paragraph{Limit PDE.} \label{paragraph:limit_pde}

Here, we discuss some possible links between SWGFs and WGFs. \citet{candau_tilh} shows that the Euler-Lagrange equation of the functional (\ref{eq:FokkerPlanckFunctional}) has a similar form (up to the first variation of the distance) for the JKO and the SW-JKO schemes, \emph{i.e.} $\mu_{k+1}^\tau$ the optimal solution of \eqref{eq:jko} satisfies
\begin{equation}
    \log(\rho_{k+1}^\tau)+V+\frac{\psi}{\tau} = \mathrm{constant}\ \mathrm{a.e.},
\end{equation}
where $\rho_{k+1}^\tau$ is the density of $\mu_{k+1}^\tau$ and $\psi$ is the Kantorovich potential from $\mu_{k+1}^\tau$ to $\mu_k^\tau$, while $\Tilde{\mu}_{k+1}^\tau$ the optimal solution of \eqref{eq:swjko} satisfies
\begin{equation}
    \log(\Tilde{\rho}_{k+1}^\tau) + V + \frac{1}{\tau} \int_{S^{d-1}} \psi_\theta\circ P^\theta \ \mathrm{d}\lambda(\theta) = \mathrm{constant}\ \mathrm{a.e.},
\end{equation}
where for $\theta\in S^{d-1}$, $\psi_\theta$ is the Kantorovich potential form $P^\theta_\#\mu_{k+1}^\tau$ to $P^\theta_\#\mu_k^\tau$. Hence, he conjectures that there is a correlation between the two gradient flows. We identify here some cases for which we can relate the Sliced-Wasserstein gradient flows to the Wasserstein gradient flows.

We first notice that for one dimensional supported measures, $W_2$ and $\sw_2$ are the same up to a constant $\sqrt{d}$, \emph{i.e.} let $\mu,\nu\in\mathcal{P}_2(\mathbb{R}^d)$ be supported on the same line, then $\sw_2^2(\mu,\nu)=W_2^2(\mu,\nu)/d$. Interestingly enough, this is the same constant as between geodesics. This property is actually still true in any dimension for Gaussians with a covariance matrix of the form $cI_d$ with $c>0$. Therefore, we argue that for these classes of measures, provided that the minimum at each step stays in the same class, we would have a dilation of factor $d$ between the WGF and the SWGF. For example, for the Fokker-Planck functional, the PDE followed by the SWGF would become $\frac{\partial \rho}{\partial t}= d\big(\mathrm{div}(\rho\nabla V) + \Delta \rho\big).$ And, by correcting the SW-JKO scheme  as
\begin{equation} \label{eq:sw_jko_d}
        \mu_{k+1}^\tau \in \argmin_{\mu\in\mathcal{P}_2(\mathbb{R}^d)}\ \frac{d}{2\tau} \sw_2^2(\mu,\mu_k^\tau)+\mathcal{F}(\mu),
\end{equation}
we would have the same dynamic. For more general measures, it is not the case anymore. But, by rewriting $\sw_2^2$ and $W_2^2$ \emph{w.r.t.} the means $m_\mu=\int x \ \mathrm{d}\mu(x)$ and $m_\nu=\int x\ \mathrm{d}\nu(x)$ and the centered measures $\Bar{\mu}$ and $\Bar{\nu}$, obtained as $\Bar{\mu} = (T_{m_\mu})_\#\mu$ and $\Bar{\nu} = (T_{m_\nu})_\#\nu$ where $T_{m_\mu}:x\mapsto x - m_\mu$, we have:
\begin{equation}
    W_2^2(\mu,\nu) = \|m_\mu-m_\nu\|_2^2 + W_2^2(\Bar{\mu},\Bar{\nu}),\ \quad \sw_2^2(\mu,\nu) = \frac{\|m_\mu-m_\nu\|_2^2}{d} + \sw_2^2(\Bar{\mu}, \Bar{\nu}).
\end{equation}
Hence, for measures characterized by their mean and variance (\emph{e.g.} Gaussians), there will be a constant $d$ between the optimal mean of the SWGF and of the WGF. However, such a direct relation is not available between variances, even on simple cases like Gaussians. We report in Appendix \ref{appendix:sw} the details of the calculations.

\subsection{Solving the SW-JKO Scheme in Practice} \label{section:swjko_practice}

As a Monte-Carlo approximation of SW can be computed in closed-form, \eqref{eq:swjko} is not a nested minimization problem anymore and is differentiable. We present here a few possible parameterizations of probability distributions which we can use in practice through SW-JKO to approximate the gradient flow. 
We further state, as an example, how to approximate the Fokker-Planck functional \eqref{eq:FokkerPlanckFunctional}. Indeed, classical other functionals can be approximated using the same method since they often only require to approximate an integral \emph{w.r.t.} the measure of interests and to evaluate its density as for \eqref{eq:FokkerPlanckFunctional}. Then, from these parameterizations, we can apply gradient-based optimization algorithms by using backpropagation over the loss at each step.

\paragraph{Discretized Grid.} \label{paragraph:discrete_grid}

A first proposition is to model the distribution on a regular fixed grid, as it is done \emph{e.g.} in \citep{peyre2015entropic}. If we approximate the distribution by a discrete distribution with a fixed grid on which the different samples are located, then we only have to learn the weights. Let us denote $\mu_k^\tau = \sum_{i=1}^N \rho_i^{(k)}\delta_{x_i}$ where we use $N$ samples located at $(x_i)_{i=1}^N$, and $\sum_{i=1}^N\rho_i=1$. Let $\Sigma_N$ denote the simplex, then the optimization problem \eqref{eq:swjko} becomes:
\begin{equation}
    \min_{(\rho_i)_i\in\Sigma_N}\ \frac{\sw_2^2\left(\sum_{i=1}^N\rho_i\delta_{x_i},\mu_k^\tau\right)}{2\tau}+\mathcal{F}\left(\sum_{i=1}^N\rho_i\delta_{x_i}\right).
\end{equation}
The entropy is only defined for absolutely continuous distributions. However, following \citep{peyre2015entropic, carlier2017convergence}, we can approximate the Lebesgue measure as: $L=l\sum_{i=1}^N\delta_{x_i}$ where $l$ represents a volume of each grid point (we assume that each grid point represents a volume element of uniform size). In that case, the Lebesgue density can be approximated by $(\frac{\rho_i}{l})_i$. Hence, for the Fokker-Planck \eqref{eq:FokkerPlanckFunctional} example, we approximate the potential and internal energies as
\begin{equation}
    \mathcal{V}(\mu) = \int V(x)\rho(x)\ \mathrm{d}x \approx \sum_{i=1}^N V(x_i)\rho_i, \quad \mathcal{H}(\mu) = \int \log\big(\rho(x)\big)\rho(x)\ \mathrm{d}x \approx \sum_{i=1}^N \log\big(\frac{\rho_i}{l}\big)\rho_i.
\end{equation}
To stay on the simplex, we use a projected gradient descent \citep{condat2016fast}. A drawback of discretizing the grid is that it becomes intractable in high dimensions.

\paragraph{With Particles.} \label{paragraph:particles_scheme}

We can also optimize over the position of a set of particles, assigning them uniform weights: $\mu_k^\tau=\frac{1}{n}\sum_{i=1}^n \delta_{x_i^{(k)}}$. The problem \eqref{eq:swjko} becomes:
\begin{equation}
    \min_{(x_i)_i}\ \frac{\sw_2^2\left(\frac{1}{n}\sum_{i=1}^n \delta_{x_i}, \mu_k^\tau\right)}{2\tau}+\mathcal{F}\left(\frac{1}{n}\sum_{i=1}^n \delta_{x_i}\right).
\end{equation}
In that case however, we do not have access to the density and cannot directly approximate $\mathcal{H}$ (or more generally internal energies). A workaround is to use non-parametric estimators \citep{beirlant1997nonparametric}, which is however impractical in high dimensions.

Additionally, using such a scheme requires to run the whole scheme at each time we want new samples which is not very practical. Using particles is more interesting when relying on the forward Euler scheme, in which case we do not need the extra minimization step performed by gradient descent.

\paragraph{Generative Models.} \label{paragraph:generative_models}

\begin{algorithm}[tb]
  \caption{SW-JKO with Generative Models}
  \label{alg:swjko_generativ_model}
    \begin{algorithmic}
      \STATE {\bfseries Input:} $\mu_0$ the initial distribution, $K$ the number of SW-JKO steps, $\tau$ the step size, $\mathcal{F}$ the functional, $N_e$ the number of epochs to solve each SW-JKO step, $n$ the batch size
      \FOR{$k=1$ {\bfseries to} $K$}
      \STATE Initialize a neural network $g_\theta^{k+1}$ \emph{e.g.} with $g_\theta^k$
      \FOR{$i=1$ {\bfseries to} $N_e$}
      \STATE Sample $z_j^{(k)},z_j^{(k+1)}\sim p_Z$ i.i.d
      \STATE $x_j^{(k)}=g_\theta^k(z_j^{(k)})$, $x_j^{(k+1)}=g_\theta^{k+1}(z_j^{(k+1)})$
      \STATE // Denote $\hat{\mu}_k^\tau = \frac{1}{n}\sum_{j=1}^{n} \delta_{x_j^{(k)}}$, $\hat{\mu}_{k+1}^\tau = \frac{1}{n}\sum_{j=1}^{n} \delta_{x_j^{(k+1)}}$
      \STATE $J(\hat{\mu}_{k+1}^\tau) = \frac{1}{2\tau} \sw_2^2(\hat{\mu}_k^\tau, \hat{\mu}_{k+1}^\tau)+\mathcal{F}(\hat{\mu}_{k+1}^\tau)$
      \STATE Backpropagate through $J$ \emph{w.r.t.} $\theta$
      \STATE Perform a gradient step using \emph{e.g.} Adam
      \ENDFOR
      \ENDFOR
    \end{algorithmic}
\end{algorithm}

To overcome these limitations, an interesting method is to use neural networks to model probability distributions, which have the advantage that we can obtain as many new samples as we want once it is trained, without needing to run it through the JKO scheme again. Moreover, it can also deal with high dimensional data and is known to generalize well.

Let us denote $g_\theta:\mathcal{Z}\to \mathcal{X}$ a generative model, with $\mathcal{Z}$ a latent space, $\theta$ the parameters of the model that will be learned, and let $p_Z$ be a simple distribution (\emph{e.g.} Gaussian). Then, we will denote $\mu_{k+1}^\tau = (g_\theta^{k+1})_\# p_Z$. The SW-JKO scheme \eqref{eq:swjko} will become in this case
\begin{equation}
    \min_\theta\ \frac{\sw_2^2\big((g_\theta^{k+1})_\# p_Z,\mu_k^\tau\big)}{2\tau} + \mathcal{F}\big((g_\theta^{k+1})_\# p_Z\big).
\end{equation}

To approximate the negative entropy, we have to be able to evaluate the density. A straightforward choice that we use in our experiments is to use invertible neural networks with a tractable density such as Normalizing Flows \citep{papamakarios2021normalizing, kobyzev2020normalizing}. Another solution could be to use the variational formulation as in \citep{fan2022variational} as we only need samples in that case, but at the cost of solving a minimax problem.

To perform the optimization, we can sample points of the different distributions at each step and use a Monte-Carlo approximation in order to approximate the integrals. Let $z_i\sim p_Z$ i.i.d, then $g_\theta(z_i)\sim (g_\theta)_\# p_Z = \mu $ and
\begin{equation} \label{eq:approx_fokker_planck_nf}
    \mathcal{V}(\mu) \approx \frac{1}{N}\sum_{i=1}^N V\big(g_\theta(z_i)\big), \quad \mathcal{H}(\mu) \approx \frac{1}{N}\sum_{i=1}^N \big(\log(p_Z(z_i))-\log|\det(J_{g_\theta}(z_i))|\big).
\end{equation}
using the change of variable formula in $\mathcal{H}$.

We sum up the procedure when modeling distributions with generative models in Algorithm \ref{alg:swjko_generativ_model}. We provide the algorithms for the discretized grid and for the particles in Appendix \ref{algorithms_swjko}.

\paragraph{Complexity.} Denoting by $d$ the dimension, $K$ the number of outer iterations, $N_e$ the number of inner optimization step, $N$ the batch size and $L$ the number of projections to approximate SW, SW-JKO has a complexity of $O(K  N_e L N\log N)$ versus $O\big(K N_e ((K+d)N + d^3)\big)$ for JKO-ICNN \citep{mokrov2021largescale} and $O(K^2 N_e N N_m)$ for the variational formulation of \citet{fan2022variational} where $N_m$ denotes the number of maximization iteration. Hence, we see that the SW-JKO scheme is more appealing for problems which will require very long dynamics.

\paragraph{Direct Minimization.} A straightforward way to minimize a functional $\mu\mapsto \mathcal{F}(\mu)$ would be to parameterize the distributions as described in this section and then to perform a direct minimization of the functional by performing a gradient descent on the weights, \emph{i.e.} for instance with a generative model, solving $\min_\theta \ \mathcal{F}\big((g_\theta)_\# p_z\big)$. While it is a viable solution, we noted that this is not much discussed in related papers implementing Wasserstein gradient flows with neural networks via the JKO scheme. This problem is theoretically not well defined as a gradient flow on the space of probability measures. And hence, it has less theoretical guarantees of convergence than Wasserstein gradient flows. In our experiments, we noted that the direct minimization suffers from more numerical instabilities in high dimensions, while SW acts as a regularizer. For simpler problems however, the performances can be quite similar.

\section{Empirical Dynamic of the Sliced-Wasserstein Gradient Flows}

In this Section, we compare empirically the trajectory of Sliced-Wasserstein Gradient Flows and of Wasserstein Gradient Flows on several examples in order to verify some of the hypotheses derived previously. More specifically, we start by drawing the trajectories of particles for an aggregation equation. Then, we focus on the Fokker-Planck equation with Gaussians measures, in which case we actually know exactly the Wasserstein gradient flow.

\begin{figure}[t]
    \centering
    \hspace*{\fill}
    \subfloat[$\mathcal{F}(\mu)=W_2^2(\mu,\frac{1}{n}\sum_{i=1}^n\delta_{x_i})$]{\label{a_traj_w}\includegraphics[width=0.45\columnwidth]{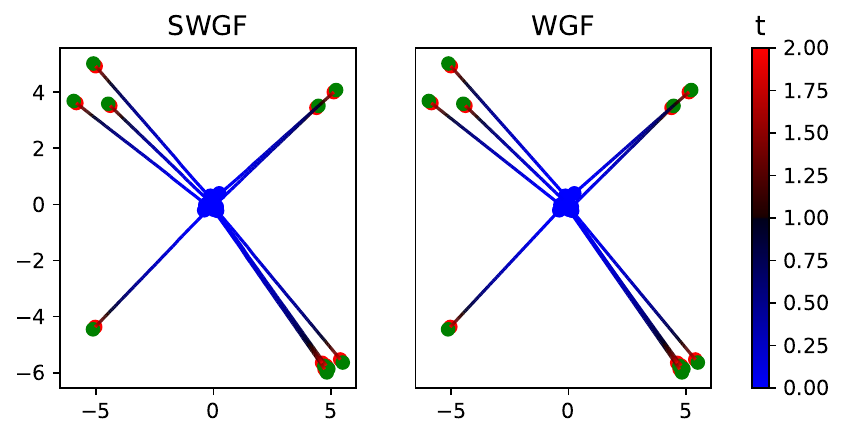}} \hfill
    \subfloat[Interaction Functional (\ref{eq:interaction_functional})]{\label{b_traj_sw}\includegraphics[width=0.45\columnwidth]{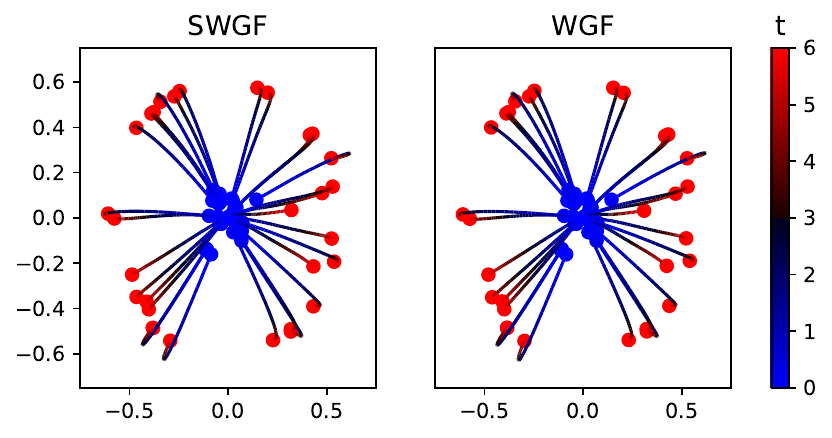}}
    \hspace*{\fill}
    \caption{Comparison of the trajectories of (dilated by $d=2$) Sliced-Wasserstein gradient flows (SWGF) and Wasserstein gradient flows (WGF) of different functionals. (\textbf{Left}) The stationary solution is a uniform discrete distributions. (\textbf{Right}) The stationary solution is a Dirac ring of radius 0.5. Blue points represent the initial positions, red points the final positions, and green points the target particles.} 
    \label{fig:comparison_trajectories}
\end{figure}

\subsection{Minimization of the Interaction Functional and of the Wasserstein Distance}

To compare the trajectory of particles following the WGFs and the SWGFs, we propose to compare with two different functionals. For the first, we choose a discrete target distribution $\frac{1}{n}\sum_{i=1}^n \delta_{x_i}$ with $x_i \in \mathbb{R}^2$, which we aim to learn. To do so, we propose to use as functional the Wasserstein distance \emph{w.r.t.} this distribution, \emph{i.e.} $\mathcal{F}(\mu) = W_2^2(\mu, \frac{1}{n}\sum_{i=1}^n \delta_{x_i})$. In this case, the target is a discrete measure with uniform weights and, using the same number of particles in the approximation $\hat{\mu}_n$, and performing gradient descent on the particles as explained in \Cref{section:swjko_practice}, we expect the Wasserstein gradient flow to push each particle on the closest target particle. This is indeed what we observe on \Cref{a_traj_w}.

For the second distribution, we use the interaction functional \eqref{eq:interaction_functional} which we recall:
\begin{equation}
    \mathcal{W}(\mu) = \iint W(x-y)\ \mathrm{d}\mu(x)\mathrm{d}\mu(y),
\end{equation}
with $W(x) = \frac{\|x\|_2^4}{4}-\frac{\|x\|_2^2}{2}$. In this case, we know that the stationary distribution is a Dirac ring \citep{carrillo2021primal}, as further explained in \Cref{section:agg_eq}. We draw on \Cref{b_traj_sw} the trajectories of some particles initially sampled from $\mathcal{N}(0,0.005 I_2)$ of the SWGF and WGF.

In both cases, by using a dilation parameter of $d$, we observe almost the same trajectories between the Sliced-Wasserstein gradient flows and the Wasserstein gradient flows, which is an additional support of the conjecture that the trajectories of the gradient flows in both spaces are alike.


\begin{figure}[t]
    \centering
    \begin{minipage}{0.49\linewidth}
        \centering
        \includegraphics[width=\columnwidth]{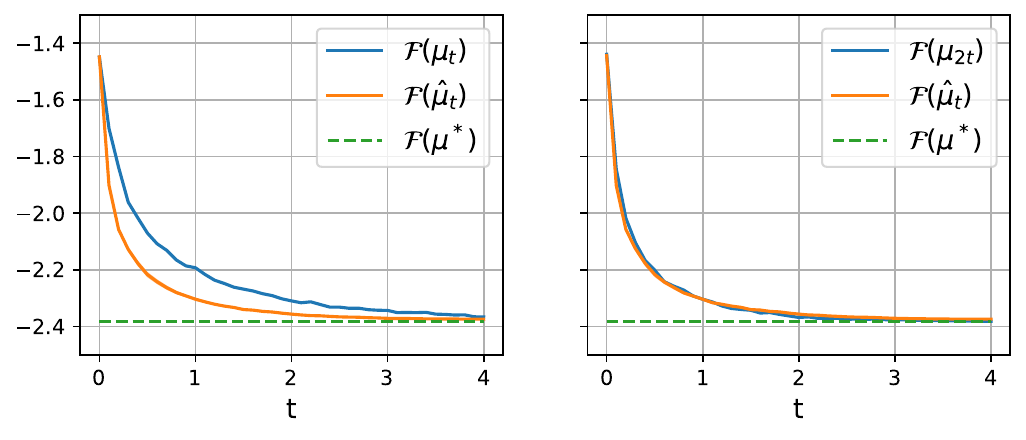}
        \caption{Evolution of the functional \eqref{eq:FokkerPlanckFunctional} along the WGF $\mu_t$, the learned SWGF $\hat{\mu}_t$, and the stationary measure $\mu^*$. We observe a dilation of parameter 2 between the WGF and the SWGF.}
        \label{fig:dilation}
    \end{minipage}
    \hfill
    \begin{minipage}{0.49\linewidth}
        \centering
        \includegraphics[width=\columnwidth]{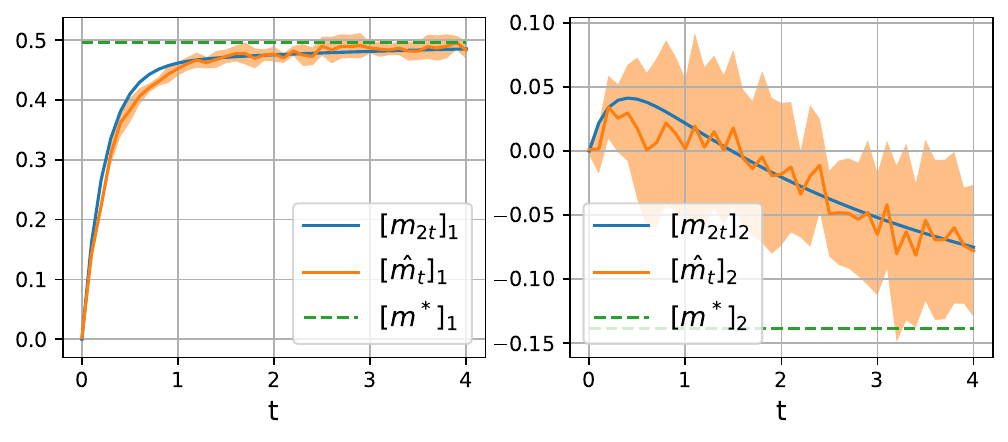}
        \caption{Evolution of the mean. $m$ denotes the true mean of WGF, $\hat{m}$ the mean obtained through SW-JKO \eqref{eq:swjko} with $\tau=0.1$ and $m^*$ the mean of the stationary measure.}
        \label{fig:means_gaussians_appendix}
    \end{minipage}
\end{figure}

\subsection{Ornstein-Uhlenbeck Process}

Now, let us focus on a case for which we know exactly the Wasserstein gradient flow. Here, we will use the Fokker-Planck functional \eqref{eq:FokkerPlanckFunctional} which we recall is defined as
\begin{equation}
    \mathcal{F}(\mu) = \int V \ \mathrm{d}\mu + \mathcal{H}(\mu).
\end{equation}
For 
\begin{equation}
    V(x)=\frac12 (x-m)^TA(x-m),
\end{equation}
with $A$ symmetric and positive definite, we obtain an Ornstein-Uhlenbeck process \citep[Chapter 8]{le2016brownian}. If we choose $\mu_0$ as a Gaussian $\mathcal{N}(m_0,\Sigma_0)$, then we know the Wasserstein gradient flow $\mu_t$ in closed form \citep{wibisono2018sampling,vatiwutipong2019alternative}, for all $t>0$, $\mu_t=\mathcal{N}(m_t,\Sigma_t)$ with
\begin{equation}
    \begin{cases}
        m_t = m + e^{-tA}(m_0-m) \\
        \Sigma_t = e^{-tA}\Sigma_0 (e^{-tA})^T + A^{-\frac12}(I-e^{-2tA})(A^{-\frac12})^T.
    \end{cases}
\end{equation}

As we know exactly the trajectory of the Wasserstein gradient flow, we propose to compare it with the trajectory of a Sliced-Wasserstein gradient flow learned using the SW-JKO scheme \eqref{eq:swjko}. It will allow us to visualize the influence of the dilation parameter on the value of the functional, and to monitor the evolution of the Sliced-Wasserstein gradient flow compared to the true distribution. Here, we do not make an extra hypothesis that the results are necessarily Gaussians, but we use Normalizing Flows to model implicitly the distributions, as discussed in \Cref{section:swjko_practice}.

More precisely, for this experiment, we model the density using RealNVPs \citep{dinh2017density} with 5 affine coupling layers, using fully connected neural networks (FCNN) for the scaling and shifting networks with 100 hidden units and 5 layers. We start the scheme with $\mu_0=\mathcal{N}(0,I_d)$ and take $L = 500$ projections to approximate the Sliced-Wasserstein distance. We randomly generate a target Gaussian (using ``make\_spd\_matrix'' from \texttt{scikit-learn} \citep{scikit-learn} to generate a random covariance with 42 as seed). We report all the results averaged over 5 trainings, with 95\% confidence intervals.

\begin{figure}[t]
    \centering
    \includegraphics[width=\columnwidth]{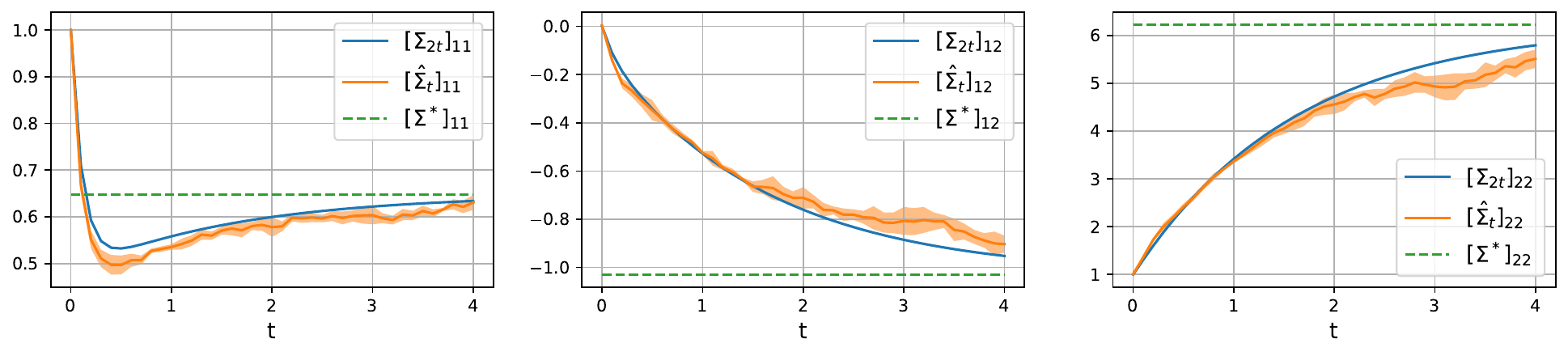}
    \caption{Evolution of the components of the covariance matrix taking into account the dilation parameter. $\Sigma$ denotes the true covariance matrix of WGF, $\hat{\Sigma}$ the covariance matrix obtained through SW-JKO \eqref{eq:swjko} with $\tau=0.1$ and $\Sigma^*$ the covariance matrix of the stationary distribution. We observe some differences between WGF and SWGF.}
    \label{fig:var_gaussians}
\end{figure}

We look at the evolution of the distributions learned between $t=0$ and $t=4$ with a time step of $\tau=0.1$. We compare it with the true Wasserstein gradient flow. On \Cref{fig:dilation}, we plot the values of the functional along the flow and we observe that when taking into account the dilation factor, the two curves are matching. Furthermore, we observed the same behavior in higher dimensions. Even though we cannot conclude on the PDE followed by SWGFs, this reinforces the conjecture that the SWGF obtained with a step size of $\frac{\tau}{d}$ (\emph{i.e.} using the scheme (\ref{eq:sw_jko_d})) is very close to the WGF obtained with a step size of $\tau$. We also report the evolution of the empirical mean (Fig. \ref{fig:means_gaussians_appendix}) and empirical covariance (Fig. \ref{fig:var_gaussians}) estimated with $10^4$ samples and averaged over 5 trainings. For the mean, it follows as expected the same diffusion. For the variance, it is less clear but it is hard to conclude since there are potentially optimization errors.

\section{Minimizing Functionals with Sliced-Wasserstein Gradient Flows} \label{section:xps}

In this section, we show that by approximating Sliced-Wasserstein gradient flows using the SW-JKO scheme \eqref{eq:swjko}, we are able to minimize functionals as well as Wasserstein gradient flows approximated by the JKO-ICNN scheme and with a better computational complexity. We first evaluate the ability to learn the stationary density for the Fokker-Planck equation \eqref{eq:FokkerPlanck} in the Gaussian case, and in the context of Bayesian Logistic Regression. Then, we evaluate it on an Aggregation equation. Finally, we use SW as a functional with image datasets as target, and compare the results with Sliced-Wasserstein flows introduced in \citep{liutkus2019sliced}.

For these experiments, we mainly use generative models. When it is required to evaluate the density (\emph{e.g.} to estimate $\mathcal{H}$), we use Real Non Volume Preserving (RealNVP) Normalizing Flows \citep{dinh2017density}. Our experiments were conducted using \texttt{PyTorch} \citep{pytorch}.

\subsection{Convergence to Stationary Distribution for the Fokker-Planck Equation} \label{section:xp_stationary_FKP}

\begin{figure}[t]
    \centering
    \includegraphics[width=0.75\columnwidth]{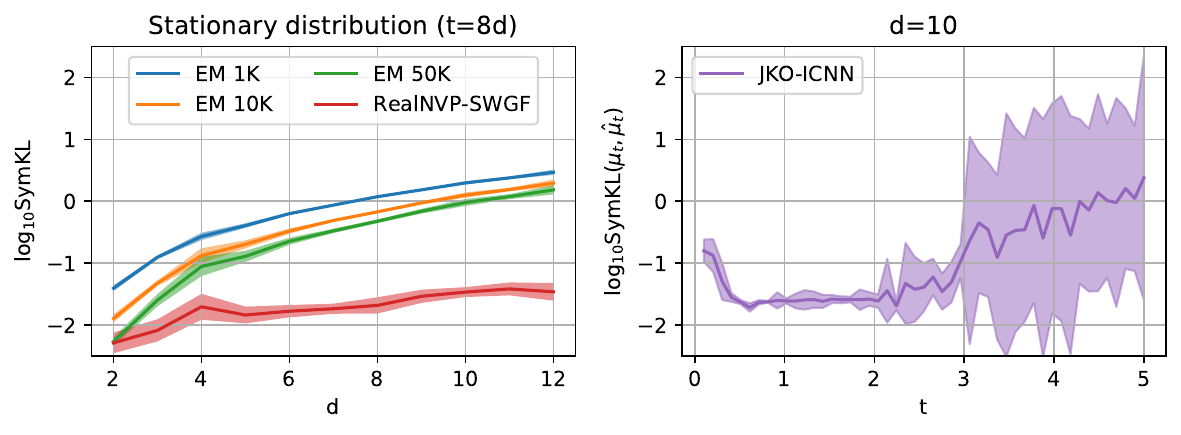}
    \caption{(\textbf{Left}) SymKL divergence between solutions at time $t=8d$ (using $\tau=0.1$ and 80 steps in \eqref{eq:swjko}) and stationary measure. (\textbf{Right}) SymKL between the true WGF $\mu_t$ and the approximation with JKO-ICNN $\hat{\mu}_t$, run through 3 Gaussians with $\tau=0.1$. We observe instabilities at some point.}
    \label{fig:unstabilities_jkoicnn}
\end{figure}

We first focus on the functional \eqref{eq:FokkerPlanckFunctional}. Its Wasserstein gradient flow is the solution of a PDE of the form of \eqref{eq:FokkerPlanck}. In this case, it is well known that the solution converges as $t\to\infty$ towards a unique stationary measure $\mu^*\propto e^{-V}$ \citep{risken1996fokker}. Hence, we focus here on learning this target distribution. First, we will choose a Gaussian as target, and then in a second experiment, we will learn a posterior distribution in a Bayesian Logistic Regression setting.

\paragraph{Gaussian Case.} \looseness=-1 Taking $V$ of the form $V(x) = \frac12(x-m)^T A (x-b)$ for all $x\in\mathbb{R}^d$, with $A$ a symmetric positive definite matrix and $m\in\mathbb{R}^d$, then the stationary distribution is $\mu^* = \mathcal{N}(m,A^{-1})$. We plot in Figure \ref{fig:unstabilities_jkoicnn} the symmetric Kullback-Leibler (SymKL) divergence over dimensions between approximated distributions and the true stationary distribution. We choose $\tau=0.1$ and performed 80 SW-JKO steps. We take the mean over 15 random gaussians for dimensions $d\in\{2,\dots,12\}$ for randomly generated positive semi-definite matrices $A$ using ``make\_spd\_matrix'' from \texttt{scikit-learn} \citep{scikit-learn}. Moreover, we use RealNVPs in SW-JKO. We compare the results with the Unadjusted Langevin Algorithm (ULA) \citep{roberts1996exponential}, called Euler-Maruyama (EM) since it is the EM approximation of the Langevin equation, which corresponds to the counterpart SDE of the PDE \eqref{eq:FokkerPlanck}. We see that, in dimension higher than 2, the results of the SWGF with RealNVP are better than with this particle scheme obtained with a step size of $10^{-3}$ and with either $10^3$, $10^4$ or $5\cdot 10^4$ particles. We do not plot the results for JKO-ICNN as we observe many instabilities (right plot in Figure \ref{fig:unstabilities_jkoicnn}). Moreover, we notice a very long training time for JKO-ICNN. We add more details in \Cref{appendix_cv_stationary}. We further note that SW acts here as a regularizer. Indeed, by training Normalizing Flows with the reverse KL (which is equal to \eqref{eq:FokkerPlanckFunctional} up to a constant), we obtain similar results, but with much more instabilities in high dimensions.


\begin{figure}[t]
    \centering
    \includegraphics[width=0.5\textwidth]{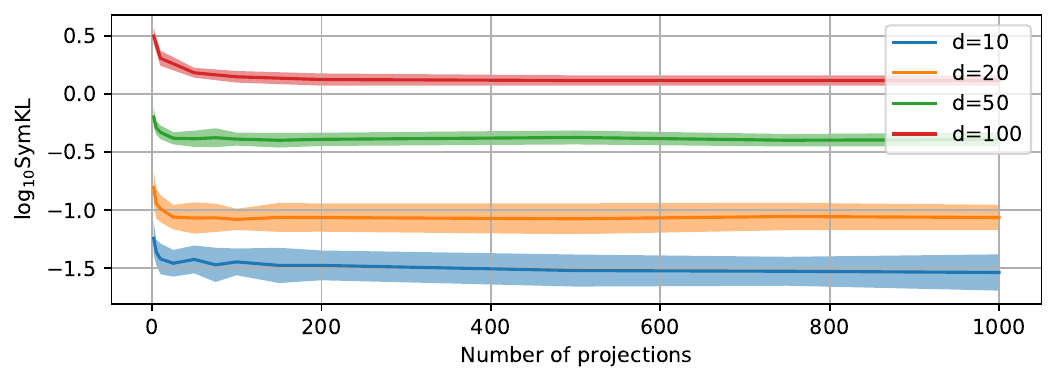}
    \caption{Impact of the number of projections for a fixed number of epochs.}
    \label{fig:impact_projs}
\end{figure}

\paragraph{Curse of Dimensionality.}

Even though the Sliced-Wasserstein distance sample complexity does not suffer from the curse of dimensionality, it appears through the Monte-Carlo approximation \citep{nadjahi2020statistical}. Here, since SW plays a regularizer role, the objective is not necessarily to approximate it well but rather to minimize the given functional. Nevertheless, the number of projections can still have an impact on the minimization, and we report on Figure \ref{fig:impact_projs} the evolution of the found minima \emph{w.r.t.} the number of projections, averaged over 15 random Gaussians. We observe that we do not need many projections to have fairly good results, even in higher dimensions. Indeed, with more than 200 projections, the performances stay relatively stable.


\paragraph{Bayesian Logistic Regression.} 

\begin{wraptable}{r}{0.5\linewidth}
    \centering
    \small
    \caption{Accuracy and Training Time for Bayesian Logistic Regression over 5 runs}
    \resizebox{0.5\columnwidth}{!}{
        \begin{tabular}{ccccc}
         & \multicolumn{2}{c}{JKO-ICNN} & \multicolumn{2}{c}{SWGF+RealNVP}  \\ \toprule
         Dataset & Acc & t & Acc & t \\ \midrule
         covtype & 0.755 $\pm 5\cdot 10^{-4}$ & 33702s & 0.755 $\pm 3\cdot 10^{-3}$ & 103s\\
         german & 0.679 $\pm 5\cdot 10^{-3}$ & 2123s & \textbf{0.68 $\pm 5\cdot 10^{-3}$} & 82s \\
         diabetis & 0.777 $\pm 7\cdot 10^{-3}$ & 4913s & \textbf{0.778 $\pm 2\cdot 10^{-3}$} & 122s \\
         twonorm & 0.981 $\pm 2\cdot 10^{-4}$ & 6551s & 0.981 $\pm 6\cdot 10^{-4}$ & 301s \\
         ringnorm & 0.736 $\pm 10^{-3}$ & 1228s & \textbf{0.741 $\pm 6\cdot 10^{-4}$} & 82s \\
         banana & 0.55 $\pm 10^{-2}$ & 1229s & \textbf{0.559 $\pm 10^{-2}$} & 66s \\
         splice & 0.847 $\pm 2\cdot 10^{-3}$ & 2290s & \textbf{0.85 $\pm 2\cdot 10^{-3}$} & 113s \\
         waveform & \textbf{0.782 $\pm 8\cdot 10^{-4}$} & 856s & 0.776 $\pm 8\cdot 10^{-4}$ & 120s \\
         image & \textbf{0.822 $\pm 10^{-3}$} & 1947s & 0.821 $\pm 3\cdot 10^{-3}$ & 72s \\ \bottomrule
        \end{tabular}
    }
    \label{tab:blr}
\end{wraptable}

Following the experiment of \cite{mokrov2021largescale} in Section 4.3, we propose to tackle the Bayesian Logistic Regression problem using SWGFs. For this task, we want to sample from $p(x|D)$ where $D$ represent data and $x=(w,\log \alpha)$ with $w$ the regression weights on which we apply a Gaussian prior $p_0(w|\alpha)=\mathcal{N}(w;0,\alpha^{-1})$ and with $p_0(\alpha)=\Gamma(\alpha;1,0.01)$. In that case, we use $V(x)=-\log p(x|D)$ to learn $p(x|D)$. We refer to \Cref{Appendix_BLR} for more details on the experiments, as well as hyperparameters.
We report in Table \ref{tab:blr} the accuracy results obtained on different datasets with SWGFs and compared with JKO-ICNN. We also report the training time and see that SWGFs allow to obtain results as good as with JKO-ICNN for most of the datasets but for shorter training times which underlines the better complexity of our scheme.

\begin{figure*}[t]
    \centering
    \hspace*{\fill}
    \subfloat[Steady state on the discretized grid]{\label{a}\includegraphics[width=0.2\columnwidth]{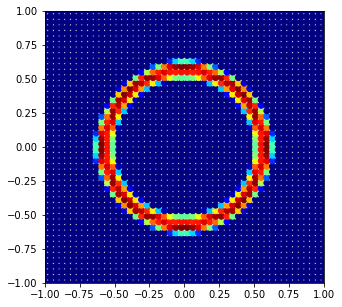}} \hfill
    \subfloat[Steady state for the fully connected neural network]{\label{b}\includegraphics[width=0.2\columnwidth]{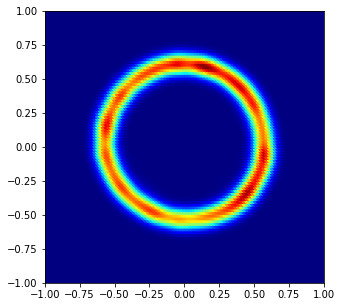}}   \hfill
    \subfloat[Steady state for particles]{\label{c}\includegraphics[width=0.2\columnwidth]{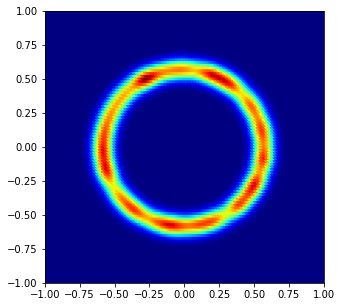}}
    \hfill
    \subfloat[Steady state for JKO-ICNN (with $\tau=0.1$)]{\label{d:JKO-ICNN_density}\includegraphics[width=0.2\columnwidth]{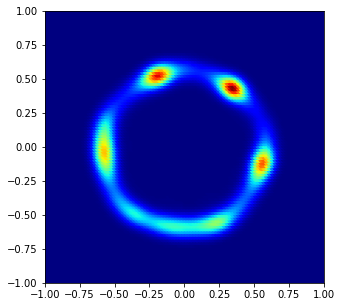}}
    \hspace*{\fill}
    \caption{Steady state of the aggregation equation for $a=4$, $b=2$. From left to right, we plot it for the discretized grid, for the FCNN, for particles and for JKO-ICNN. We observe that JKO-ICNN does not recover the ring correctly as the particles are not evenly distributed on it.}
    \label{fig:aggregation_equation1}
    \vspace{-10pt}
\end{figure*}

\subsection{Convergence to Stationary Distribution for an Aggregation Equation} \label{section:agg_eq}

We also show the possibility to find the stationary solution of different PDEs than Fokker-Planck. For example, using an interaction functional of the form
\begin{equation} 
    \mathcal{W}(\mu) = \frac12 \iint W(x-y)\ \mathrm{d}\mu(x)\mathrm{d}\mu(y).
\end{equation}
We notice here that we do not need to evaluate the density. Therefore, we can apply any neural network. For example, in the following, we will use a simple fully connected neural network (FCNN) and compare the results obtained with JKO-ICNN. We also show the results when learning directly over the particles and when learning weights over a regular grid.

\citet{carrillo2021primal} use a repulsive-attractive interaction potential $W(x) = \frac{\|x\|_2^4}{4}-\frac{\|x\|^2_2}{2}$. In this case, they showed empirically that the solution is a Dirac ring with radius 0.5 and centered at the origin when starting from $\mu_0=\mathcal{N}(0,0.25^2 I_2)$. With $\tau=0.05$, we show on Figure \ref{fig:aggregation_equation1} that we recover this result with SWGFs for different parameterizations of the probabilities. More precisely, we first use a discretized grid of $50\times 50$ samples of $[-1,1]^2$. Then, we show the results when directly learning the particles and when using a FCNN. We also compare them with the results obtained with JKO-ICNN. The densities reported for the last three methods are obtained through a kernel density estimator (KDE) with a bandwidth manually chosen since we either do not have access to the density, or we observed for JKO-ICNN that the likelihood exploded. 
It may be due to the fact that the stationary solution does not admit a density with respect to the Lebesgue measure. For JKO-ICNN, we observe that the ring shape is recovered, but the samples are not evenly distributed on it.

We report the solution at time $t=10$, and use $\tau=0.05$ for SW-JKO and $\tau=0.1$ for JKO-ICNN. As JKO-ICNN requires $O(k^2)$ evaluations of gradients of ICNNs, the training is very long for such a dynamic. 
Here, the training took around 5 hours on a RTX 2080 TI (for 100 steps), versus 20 minutes for the FCNN and 10 minutes for 1000 particles (for 200 steps). 

This underlines again the better training complexity of SW-JKO compared to JKO-ICNN, which is especially appealing when we are only interested in learning the optimal distribution. One such task is generative modeling in which we are interested in learning a target distribution $\nu$ which we have access to through samples.

\subsection{Application on Real Data}

\begin{figure}[t]
    \centering
    \begin{minipage}{0.49\linewidth}
        \centering
        \hspace*{\fill}
        \subfloat{\label{a_MNIST}\includegraphics[width=0.32\columnwidth]{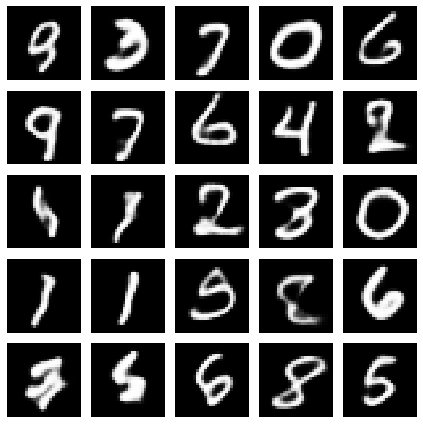}} \hfill
        \subfloat{\label{b_FashionMNIST}\includegraphics[width=0.32\columnwidth]{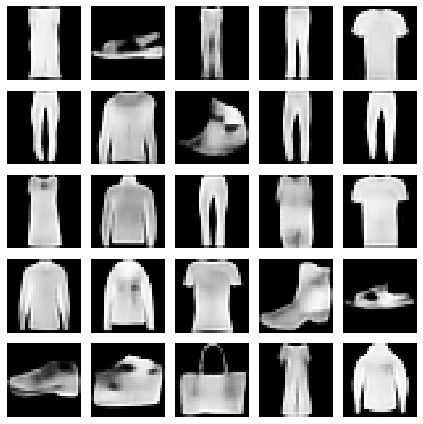}}
        \hfill
        \subfloat{\label{c_CelebA}\includegraphics[width=0.32\columnwidth]{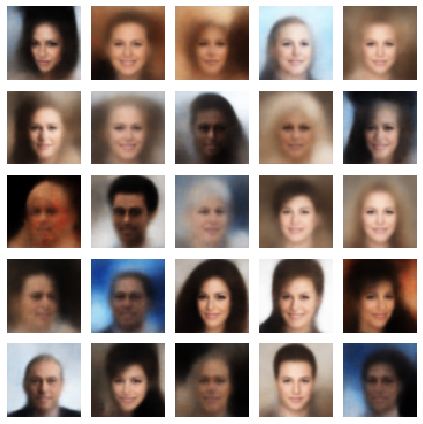}}
        \hspace*{\fill}
        \caption{Generated sample obtained through a pretrained decoder + RealNVP.}
        \label{fig:mnist_AE}
    \end{minipage}
    \hfill
    \begin{minipage}{0.49\linewidth}
        \centering
        \small
        \captionof{table}{FID scores on some datasets (lower is better)}
        \resizebox{\columnwidth}{!}{
            \begin{tabular}{cccccc}
                & \textbf{Methods}  &\textbf{MNIST} & \textbf{Fashion} & \textbf{CelebA}\\
                \toprule
                \multirow{4}{*}{\rotcell{\scriptsize Ambient \\ Space}} 
                & SWF \citep{liutkus2019sliced} & 225.1 & 207.6 & - \\
                & SWGF + RealNVP & 88.1 & {\bf 95.5} & - \\
                & SWGF + CNN & {\bf 69.3} & 102.3 & -\\
                & & & \\
                \midrule
                \multirow{4}{*}{\rotcell{Latent \\ Space}} 
                & AE (golden score) & 15.55 & 31 & 77 \\ \cmidrule{2-5}
                & SWGF + AE + RealNVP & {\bf 17.8} & {\bf 40.6} & 90.9 \\
                & SWGF + AE + FCNN & 18.3 & 41.7 & \textbf{88} \\
                & SWF & 22.5 & 56.4 & 91.2 \\
                \bottomrule
            \end{tabular}
            }
        \label{tab:table_FID}
    \end{minipage}
\end{figure}

In what follows, we show that the SW-JKO scheme can generate real data, and perform better than the associated particle scheme obtained by the associated SDE (see \Cref{sec:bg_wgf}). To perform generative modeling, we can use different functionals. For example, GANs use the Jensen-Shannon divergence \citep{goodfellow2014generative} and WGANs the Wasserstein-1 distance \citep{arjovsky2017wasserstein}. To compare with an associated particle scheme, we focus here on the regularized SW distance as functional, defined as
\begin{equation} \label{eq:swf}
    \mathcal{F}(\mu) = \frac12 \sw_2^2(\mu,\nu)+\lambda\mathcal{H}(\mu),
\end{equation}
where $\nu$ is some target distribution, for which we should have access to samples. The Wasserstein gradient flow of this functional was first introduced and studied by \citet{bonnotte2013unidimensional} for $\lambda=0$, and by \citet{liutkus2019sliced} with the negative entropy term. \citet{liutkus2019sliced} showcased a particle scheme called SWF (Sliced Wasserstein Flow) to approximate the WGF of \eqref{eq:swf}. 
Applied on images such as MNIST \citep{lecun-mnisthandwrittendigit-2010}, FashionMNIST \citep{xiao2017fashion} or CelebA \citep{liu2015deep}, SWFs need a very long convergence due to the curse of dimensionality and the trouble approximating SW. Hence, they used instead a pretrained autoencoder (AE) and applied the particle scheme in the latent space. Likewise, we use the AE proposed by \citet{liutkus2019sliced} with a latent space of dimension $d=48$, and we perform SW-JKO steps on those images. We report on Figure \ref{fig:mnist_AE} samples obtained with RealNVPs and on Table \ref{tab:table_FID} the Fréchet Inception distance (FID) \citep{heusel2017gans} obtained between $10^4$ samples. We denote ``golden score'' the FID obtained with the pretrained autoencoder. Hence, we cannot obtain better results than this. We compared the results in the latent and in the ambient space with SWFs and see that we obtain fairly better results using generative models within the SW-JKO scheme, especially in the ambient space, although the results are not really competitive with state-of-the-art methods. This may be due more to the curse of dimensionality in approximating the objective SW than in approximating the regularizer SW. Note that in a more recent work \citep{du2023nonparametric}, it was shown that changing the projections to take into account the specificities of images, \emph{e.g.} translation invariance, with convolutions, allowed to obtain very nice results with SWFs, even in the space of images.

To sum up, an advantage of the SW-JKO scheme is to be able to use easier, yet powerful enough, architectures to learn the dynamic. 
This is cheaper in training time and less memory costly. Furthermore, we can tune the architecture with respect to the characteristics of the problem and add inductive biases (\emph{e.g.} using CNN for images) or learn directly over the particles for low dimensional problems.

\section{Conclusion and Discussion}

\looseness=-1 In this chapter, we derive a new class of gradient flows in the space of probability measures endowed with the Sliced-Wasserstein metric, and the corresponding algorithms. 
To the best of our knowledge, and despite its simplicity, this is the first time that this class of flows is proposed in a Machine Learning context. We showed that it has several advantages over state-of-the-art approaches such as the recent JKO-ICNN. Aside from being less computationally intensive, it is more versatile \emph{w.r.t.} the different practical solutions for modeling probability distributions, such as Normalizing Flows, generative models or sets of evolving particles.

Regarding the theoretical aspects, several challenges remain ahead: First, its connections with Wasserstein gradient flows are still unclear. 
Second, one needs to understand if, regarding the optimization task, convergence speeds or guarantees are changed with this novel formulation, revealing potentially interesting practical properties. Lastly, it is natural to study if popular variants of the Sliced-Wasserstein distance 
such as Max-sliced \citep{deshpande2019max}, Distributional Sliced \citep{nguyen2020distributional}, Subspace robust \citep{paty2019subspace}, generalized Sliced \citep{kolouri2019generalized} or projection Wasserstein distances \citep{rowland2019orthogonal} can  also be used in similar gradient flow schemes. The study of higher-order approximation schemes such as BDF2 \citep{plazotta2018bdf2, matthes2019variational, natale2022geodesic} could also be of interest.




\clearemptydoublepage
\cleartooddpage[\thispagestyle{empty}]
\chapter{Unbalanced Optimal Transport Meets Sliced-Wasserstein} \label{chapter:usw}

{
    \hypersetup{linkcolor=black} 
    \minitoc 
}

Optimal Transport (OT) has emerged as a powerful framework to compare probability measures, a fundamental task in many statistical and Machine Learning problems.
Substantial advances have been made over the last decade in designing OT variants which are either computationally and statistically more efficient, or more robust to the measures/datasets to compare.
Among them, Sliced-Wasserstein distances have been extensively used to mitigate Optimal Transport's cubic algorithmic complexity and curse of dimensionality. In parallel, unbalanced OT was designed to allow comparisons of more general positive measures, while being more robust to outliers. In this chapter, based on \citep{sejourne2023unbalanced}, we propose to combine these two concepts, namely slicing and unbalanced OT, to develop a general framework for efficiently comparing positive measures. We propose two new loss functions based on the idea of slicing unbalanced OT, and study their induced topology and statistical properties. We then develop a fast Frank-Wolfe-type algorithm to compute these loss functions, and show that the resulting methodology is modular as it encompasses and extends prior related work. We finally conduct an empirical analysis of our loss functions and methodology on both synthetic and real datasets, to illustrate their relevance and applicability. 

\section{Introduction}

Positive measures are ubiquitous in various fields, including data sciences and Machine Learning (ML) where they commonly serve as data representations.
A common example is the density fitting task, which arises in generative modeling~\citep{arjovsky2017wasserstein,de2021diffusion}: the observed samples can be represented as a discrete positive measure $\nu$ and the goal is to find a parametric measure $\mu_\eta$ which fits the best $\nu$. 
This can be achieved by training a model that minimizes a loss function over $\eta$, usually defined as a distance between $\nu$ and $\mu_\eta$. 
Therefore, it is important to choose a meaningful discrepancy with desirable statistical, robustness and computational properties. 
In particular, some settings require comparing arbitrary positive measures, \emph{i.e.}~measures whose total mass can have an arbitrary value, as opposed to probability distributions, whose total mass is equal to 1. 
In cell biology~\citep{schiebinger2019optimal}, for example, measures are used to represent and compare gene expressions of cell populations, and the total mass represents the population size. 

\paragraph{(Unbalanced) Optimal Transport.} 
Optimal Transport has been chosen as a loss function in various ML applications.
OT defines a distance between two positive measures of same mass $\mu$ and $\nu$ (\emph{i.e.}~$m(\mu)=m(\nu)$) by moving the mass of $\mu$ toward the mass of $\nu$ with least possible effort. 
The mass equality can nevertheless be hindered by imposing a normalization of $\mu$ and $\nu$ to enforce $m(\mu) = m(\nu)$, which is potentially spurious and makes the problem less interpretable. 
In recent years, OT has then been extended to settings where measures have different masses, leading to the \emph{unbalanced OT} (UOT) framework \citep{kondratyev2016fitness, chizat2018unbalanced, liero2018optimal}. 
An appealing outcome of this new OT variant is its robustness to outliers which is achieved by discarding them before transporting $\mu$ to $\nu$.
UOT has been useful for many theoretical and practical applications, \emph{e.g.} theory of deep learning \citep{chizat2018global,rotskoff2019global}, biology \citep{schiebinger2019optimal, demetci2022scotv2} and domain adaptation \citep{fatras2021unbalanced}. 
We refer to~\citep{sejourne2022unbalanced} for an extensive survey of UOT. 
Computing OT requires solving a linear program whose complexity is cubical in the number $n$ of samples ($O(n^3 \log n)$). 
Besides, accurately estimating OT distances through empirical distributions is challenging as OT suffers from the curse of dimensionality \citep{dudley1969speed}.
A common workaround is to rely on OT variants with lower complexities and better statistical properties. Among the most popular, we can list entropic OT~\citep{cuturi2013sinkhorn}, minibatch OT \citep{fatras2021minibatch} and Sliced-Wasserstein \citep{rabin2011transportation,bonneel2015sliced}. 

When it comes  to slicing unbalanced OT, it has been applied to partial OT \citep{bonneel2019spot, sato2020fast, bai2022sliced}, a particular case of UOT, in order to speed up comparisons of unnormalized measures at large scale. However, while (sliced) partial OT allows to compare measures with different masses, it assumes that each input measure is discrete and supported on points that all share the same mass (typically 1).
In contrast, the Gaussian-Hellinger-Kantorovich (GHK) distance~\citep{liero2018optimal} (also known as the Wasserstein-Fisher-Rao distance \citep{chizat2018interpolating}), another popular formulation of UOT, allows to compare measures with different masses \emph{and} supported on points with varying masses, and has not been studied jointly with slicing. 

\paragraph{Contributions.} 

In this chapter, we present the first general framework combining UOT and slicing. 
Our main contribution is the introduction of two novel sliced variants of UOT, respectively called \emph{Sliced UOT ($\SUOT$)} and \emph{Unbalanced Sliced-Wasserstein ($\RSOT$)}. 
$\SUOT$ and $\RSOT$ both leverage one-dimensional projections and the newly-proposed implementation of UOT in 1D \citep{sejourne2022faster}, but differ in the penalization used to relax the constraint on the equality of masses: $\RSOT$ essentially performs a global reweighting of the inputs measures $(\mu,\nu)$, while $\SUOT$ reweights each projection of $(\mu, \nu)$. 
Our work builds upon the Frank-Wolfe-type method~\citep{frank1956algorithm} recently proposed in~\citep{sejourne2022faster} to efficiently compute GHK between univariate measures, an instance of UOT which has not yet been combined with slicing. We derive the associated theoretical properties, along with the corresponding fast and GPU-friendly algorithms. We demonstrate its versatility and efficiency on challenging experiments, where slicing is considered on a non-Euclidean hyperbolic manifold, as a similarity measure for document classification, or for computing barycenters of geoclimatic data.


\section{Background on Unbalanced Optimal Transport}  \label{sec:usw_background} 


We denote by $\Mmp(\Rd)$ the set of all positive Radon measures on $\mathbb{R}^d$. For any $\mu \in \Mmp(\mathbb{R}^d)$, $\mathrm{supp}(\mu)$ is the support of $\mu$ and $m(\mu) = \int_{\mathbb{R}^d} \mathrm{d} \mu(x)$ the mass of $\mu$. 
We recall the standard formulation of unbalanced OT \citep{liero2018optimal}. Here, we focus for the regularization on the Kullback-Leibler divergence, defined between $\mu,\nu\in\Mmp(\mathbb{R}^d)$ as 
\begin{equation}
    \kl(\mu||\nu) = \left\{\begin{array}{ll}
        \int_{\mathbb{R}^d} \log \left(\frac{\mathrm{d}\mu}{\mathrm{d}\nu}(x)\right)\ \mathrm{d}\mu(x) + \int_{\mathbb{R}^d} \mathrm{d}\nu(x) - \int_{\mathbb{R}^d}\mathrm{d}\mu(x) & \mbox{ if } \mu\ll\nu \\
        +\infty & \mbox{ otherwise},
    \end{array}\right.
\end{equation}
and on a cost of the form $c(x,y)=\|x-y\|_2^p$ for $p\ge 1$. This corresponds to the GHK setting \citep{liero2018optimal}. The framework and some results can be generalized to more general $\varphi$-divergences, and we refer to \citep{sejourne2023unbalanced} for more details. In particular, when choosing the Total Variation distance, we recover the partial Optimal Transport problem \citep{figalli2010optimal}.

\begin{definition}[Unbalanced OT] \label{def:phi-div}
    Let $\mu, \nu \in \Mmp(\mathbb{R}^d)$. Given $\rho_1,\rho_2\ge 0$ and a cost $c: \mathbb{R}^d \times \mathbb{R}^d \to \mathbb{R}$, the unbalanced OT problem between $\mu$ and $\nu$ reads
    \begin{equation} \label{eq:primal-uot}
        \UOT(\mu, \nu) = \inf_{\gamma\in\Mmp(\Rd \times \Rd)} \int c(x,y)\ \mathrm{d}\gamma(x,y) + \rho_1\kl(\pi^1_\#\gamma||\mu) + \rho_2\kl(\pi^2_\#\gamma||\nu),
    \end{equation}
    with $\pi^1:(x,y)\mapsto x$ and $\pi^2:(x,y)\mapsto y$.
\end{definition}
We note that we recover the regular OT problem $W_c$ \eqref{eq:Kantorovich_problem} when $\rho_1\to\infty$ and $\rho_2\to\infty$ as in this case, the marginals are fully enforced.

The UOT problem has been shown to admit an equivalent formulation obtained by deriving the dual of \eqref{eq:primal-uot} and proving strong duality. Based on \Cref{prop:strong_duality_uot}, computing $\UOT(\mu, \nu)$ consists in optimizing a pair of continuous functions $(f,g)$.

\begin{proposition}[Corollary 4.12 in \citep{liero2018optimal}] \label{prop:strong_duality_uot}
    The $\UOT$ problem~\eqref{eq:primal-uot} can equivalently be written as
    \begin{equation} \label{eq:dual-uot} 
        \UOT(\mu, \nu) = \sup_{f\oplus g\leq c}\ \int \varphi^\circ_1\big(f(x)\big)\ \mathrm{d}\mu(x) + \int \varphi^\circ_2\big(g(y)\big)\ \mathrm{d}\nu(y), 
    \end{equation}
    where for $i \in \{1, 2\}$, $\varphi_i^\circ(x)=\rho_i(1-e^{-x/\rho_i})$, and $f\oplus g\leq c$ means that for $(x,y) \sim \mu \otimes \nu$, $f(x)+g(y)\leq c(x,y)$.
\end{proposition}

$\UOT(\mu, \nu)$ is known to be computationally intensive \citep{pham2020unbalanced}, thus motivating the development of methods that can scale to dimensions and sample sizes encountered in ML applications. Therefore, it is appealing to develop sliced workarounds to overcome the computational bottleneck. 

$\sw_p(\mu, \nu)$ is defined in terms of the Kantorovich formulation of OT, hence inherits the following drawbacks: $\sw_p(\mu, \nu) < +\infty$ only when $m(\mu) = m(\nu)$, and may not provide meaningful comparisons in presence of outliers. 
To overcome such limitations, prior works have proposed sliced versions of partial OT \citep{bonneel2019spot,bai2022sliced}, a particular instance of UOT. However, their contributions only apply to measures whose samples have constant mass.
In the next section, we 
generalize their line of work and propose a new way of combining sliced OT and unbalanced OT.

\section{Sliced Unbalanced OT and Unbalanced Sliced OT} \label{sec:usw_theory}

\subsection{Definition}

We propose two strategies to make unbalanced OT scalable, by leveraging sliced OT. We formulate two loss functions (\Cref{def:suot_usot}), then study their theoretical properties and discuss their implications.  

\begin{definition} \label{def:suot_usot}
    Let $\mu, \nu \in \Mmp(\Rd)$ and $p\ge 1$. The \textbf{Sliced Unbalanced OT} loss ($\SUOT$) and the \textbf{Unbalanced Sliced-Wasserstein} loss ($\RSOT$) between $\mu$ and $\nu$ are defined as, 
    \begin{align}
        \SUOT(\mu, \nu) &= \int_{S^{d-1}} \UOT(P^\theta_\#\mu, P^\theta_\#\nu)\ \mathrm{d} \lambda(\theta), \label{eq:suot_primal} \\
        \RSOT_p^p(\mu, \nu) &= \inf_{(\pi_1,\pi_2)\in\Mmp(\Rd)\times\Mmp(\Rd)}\  \sw_p^p(\pi_1,\pi_2) + \rho_1\kl(\pi_1||\mu) + \rho_2\kl(\pi_2||\beta), \label{eq:def-rsot}
    \end{align}
    where $P^\theta(x) = \langle x, \theta\rangle$ and $\lambda$ is the uniform measure on $S^{d-1}$.
\end{definition}

$\SUOT(\mu, \nu)$ compares $\mu$ and $\nu$ by solving the UOT problem between $P^\theta_\#\mu$ and $P^\theta_\#\nu$ for $\theta \sim \lambda$. Note that when using the Total Variation distance instead of the KL divergence, $\SUOT$ becomes the sliced partial OT problem \citep{bonneel2019spot,bai2022sliced}. 
On the other hand, $\RSOT$ is a completely novel approach and stems from the following property on $\UOT$ \citep[Equations (4.21)]{liero2018optimal}: 
\begin{equation}
    \UOT(\mu, \nu) = \inf_{(\pi_1,\pi_2)\in\Mmp(\Rd)\times \Mmp(\Rd)}\ \OT(\pi_1,\pi_2) + \rho_1\kl(\pi_1||\mu) + \rho_2\kl(\pi_2||\nu).
\end{equation}

\begin{figure*}[t]
    \centering
    \hspace*{\fill}
    \subfloat[Samples and directions]{\label{fig:samples_illustration_usw}\includegraphics[width={0.3\linewidth}]{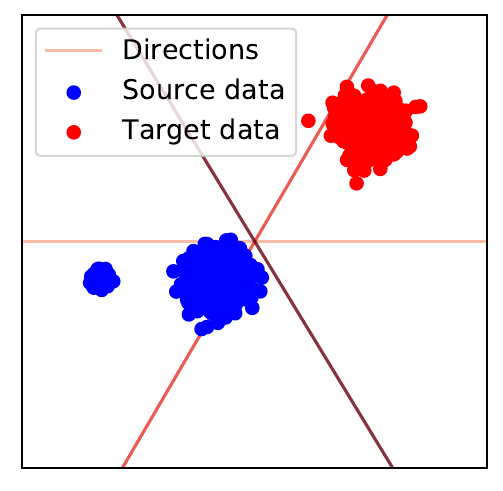}} \hfill
    \subfloat[$\SUOT$ ($\rho_1=0.01$, $\rho_2=1)$]{\label{fig:1d_density_suot}\includegraphics[width={0.3\linewidth}]{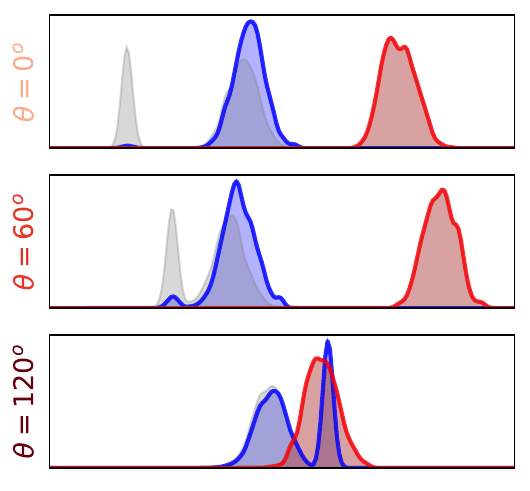}} \hfill
    \subfloat[$\RSOT$ ($\rho_1=1$, $\rho_2=1$)]{\label{fig:density_rsot}\includegraphics[width={0.3\linewidth}]{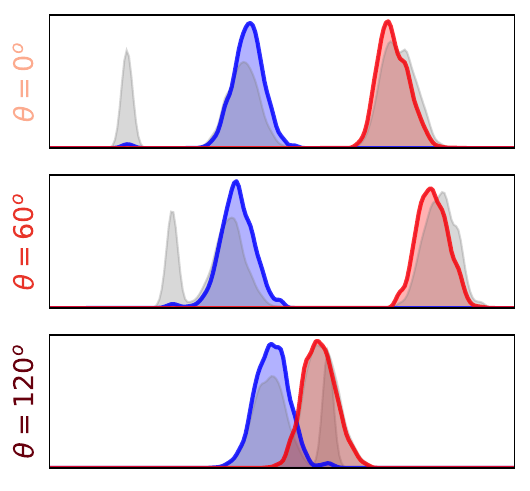}} \hfill
    \hspace*{\fill}
    \caption{{\em Toy illustration} on the behaviors of $\SUOT$ and $\RSOT$. (\textbf{Left}) Original 2D samples and slices used for illustration. KDE density estimations of the projected samples: {\color{Gray} grey}, original distributions, {\color{Rhodamine} colored}, distributions reweighed by $\SUOT$  (\textbf{Center}), and reweighed by $\RSOT$ (\textbf{Right}).}
    \label{fig:illustration_suot}
\end{figure*}

\paragraph{$\SUOT$ vs. $\RSOT$.} As outlined in \Cref{def:suot_usot}, $\SUOT$ and $\RSOT$ differ in how the transportation problem is penalized: $\SUOT(\mu, \nu)$ regularizes the marginals of $\gamma_\theta$ for $\theta \sim \lambda$ where $\gamma_\theta$ denotes the solution of $\UOT(P^\theta_\#\mu, P^\theta_\#\nu)$, while $\RSOT(\mu, \nu)$ operates a geometric normalization directly on $(\mu, \nu)$. We illustrate this difference in the following practical setting: we consider $\mu,\nu\in\Mmp(\rset^2)$ where $\mu$ is polluted with some outliers, and we compute $\SUOT(\mu, \nu)$ and $\RSOT(\mu, \nu)$. We plot the input measures and the sampled projections $\{\theta_k\}_k$ (\Cref{fig:illustration_suot}, left), the marginals of $\gamma_{\theta_k}$ for $\SUOT$ and the marginals $P^{\theta_k}_\# \mu$ and $P^{\theta_k}_\#\nu$ for $\RSOT$ (\Cref{fig:illustration_suot}, right). 
As expected, $\SUOT$ marginals change for each $\theta_k$.
We also observe that the source outliers have successfully been removed for any $\theta$ when using $\RSOT$, while they may still appear with $\SUOT$ (\emph{e.g.} for $\theta=120^\circ$): this is a direct consequence of the penalization terms $\kl$ in $\RSOT$, which operate on $(\mu, \nu)$ rather than on their projections.

\subsection{Theoretical Properties}

\looseness=-1 In this section, we report a set of theoretical properties of $\SUOT$ and $\RSOT$. All proofs are provided in \citep{sejourne2023unbalanced}. 
First, the infimum is attained in $\UOT(P^\theta_\#\mu, P^\theta_\#\nu)$ for $\theta \in S^{d-1}$ and in $\RSOT(\mu, \nu)$, see \citep[Proposition A.1]{sejourne2023unbalanced}. 
We also show that these optimization problems are convex, both $\SUOT$ and $\RSOT$ are jointly convex \emph{w.r.t.} their input measures, and that strong duality holds (\Cref{thm:duality-rsot}). 

Next, we prove that both $\SUOT$ and $\RSOT$ preserve some topological properties of $\UOT$, starting with the metric axioms as stated in the next proposition.

\begin{proposition}[Metric properties] \label{prop:metric-suot} \leavevmode
    \begin{enumerate}
        \item Suppose $\UOT$ is non-negative, symmetric and/or definite on $\Mmp(\rset) \times \Mmp(\rset)$. Then, $\SUOT$ is respectively non-negative, symmetric and/or definite on $\Mmp(\Rd) \times \Mmp(\Rd)$. If there exists $p \in [1, +\infty)$ s.t. for any $(\mu,\nu,\gamma)\in\Mmp(\rset)$, $\UOT^{1/p}(\mu,\nu)\leq \UOT^{1/p}(\mu,\gamma) + \UOT^{1/p}(\gamma,\nu)$, then $\SUOT^{1/p}(\mu,\nu)\leq \SUOT^{1/p}(\mu,\gamma) + \SUOT^{1/p}(\gamma,\nu)$.
        \item For any $\mu, \nu \in \Mmp(\Rd)$ and $p\ge 1$, $\RSOT_p^p(\mu, \nu) \geq 0$, $\RSOT_p^p$ is symmetric and is definite.
    \end{enumerate}
\end{proposition}

\begin{proof}
    See \citep[Proposition 3.2]{sejourne2023unbalanced}.
\end{proof}

By \Cref{prop:metric-suot} 1., establishing the metric axioms of $\UOT$ between \emph{univariate} measures (e.g., as detailed in \citep[Section 3.3.1]{sejourne2022unbalanced}) suffices to prove the metric axioms of $\SUOT$ between \emph{multivariate} measures. Since e.g. GHK~\citep[Theorem 7.25]{liero2018optimal} is a metric for $p=2$, then so is the associated $\SUOT$.

In our next theorem, we show that $\SUOT$, $\RSOT$ and $\UOT$ are equivalent, under certain assumptions on input measures $(\mu, \nu)$.

\begin{theorem}[Equivalence of $\SUOT, \RSOT, \UOT$] \label{thm:equiv-loss}
    Let $\mathsf{X}$ be a compact subset of~$\Rd$ with radius $R$. Let $p \in [1, +\infty)$ and assume $c(x,y) = \| x-y \|_2^p$. Then, for any $\mu, \nu \in \Mmp(\mathsf{X})$,
    \begin{equation} \label{eq:equiv_loss}
        \SUOT(\mu,\nu)\leq\RSOT_p^p(\mu,\nu) \leq \UOT(\mu,\nu) \leq c(m(\mu), m(\nu), \rho, R) \SUOT(\mu, \nu)^{1/(d+1)} \,,
    \end{equation}
    where $c(m(\mu), m(\nu), \rho, R)$ is a constant depending on $m(\mu), m(\nu), \rho, R$, which is non-decreasing in $m(\mu)$ and $m(\nu)$.
    Additionally, assume there exists $M > 0$ s.t. $m(\mu) \leq M, m(\nu) \leq M$. Then, $c(m(\mu), m(\nu), \rho, R)$ no longer depends on $m(\mu), m(\nu)$, which proves the equivalence of $\SUOT$, $\RSOT$ and $\UOT$.
\end{theorem}

\begin{proof}
    See \citep[Theorem 3.3]{sejourne2023unbalanced}.
\end{proof}

The equivalence of $\SUOT, \RSOT$ and $\UOT$ is useful to prove that $\SUOT$ and $\RSOT$ \emph{metrize the weak convergence} when $\UOT$ does, \emph{e.g.} in the GHK setting \citep[Theorem 7.25]{liero2018optimal}.

\begin{theorem}[Weak metrization] \label{thm:weak-cv}
    Let $p \in [1, +\infty)$ and consider $c(x,y)=\|x-y\|_2^p$. Let $(\mu_n)$ be a sequence of measures in $\Mmp(\mathsf{X})$ and $\mu\in\Mmp(\mathsf{X})$, where $\mathsf{X} \subset \Rd$ is compact with radius $R > 0$. Then, $\SUOT$ and $\RSOT$ metrize the weak convergence, \emph{i.e.}
    \begin{equation}
        \mu_n \xrightarrow[n\to\infty]{\mathcal{L}} \mu \iff \lim_{n\to\infty} \SUOT(\mu_n,\mu) = 0 \iff \lim_{n\to\infty} \RSOT_p^p(\mu_n,\mu) = 0.
    \end{equation}
\end{theorem}

\begin{proof}
    See \citep[Theorem 3.4]{sejourne2023unbalanced}.
\end{proof}

The metrization of weak convergence is an important property when comparing measures. For instance, it can be leveraged to justify the well-posedness of approximating an unbalanced Wasserstein gradient flow~\citep{ambrosio2008gradient} using $\SUOT$, as done in \citep{candau_tilh} and in \Cref{chapter:swgf} for $\sw$. Unbalanced Wasserstein gradient flows have been a key tool in deep learning theory, \emph{e.g.} to prove global convergence of 1-hidden layer neural networks~\citep{chizat2018global,rotskoff2019global}.

We move on to the statistical properties and prove that $\SUOT$ offers important statistical benefits, as it lifts the \emph{sample complexity} of $\UOT$ from one-dimensional setting to multi-dimensional ones. In what follows, for any $\mu \in \Mmp(\Rd)$, we use $\hat{\mu}_n$ to denote the empirical approximation of $\mu$ over $n \geq 1$ i.i.d. samples, \emph{i.e.}~$\hat{\mu}_n = \frac1n \sum_{i=1}^n \delta_{Z_i}$, $Z_i \sim \mu$ for $i = 1, \dots, n$.

\begin{theorem}[Sample complexity] \label{thm:sample-comp} \leavevmode
    \begin{enumerate}
        \item If for $\alpha,\beta\in\Mmp(\rset)$, $\mathbb{E}\big[|\UOT(\alpha,\beta) -  \UOT(\hat\alpha_n,\hat\beta_n)|\big]\leq \kappa(n)$, then for $\mu,\nu\in\Mmp(\Rd)$, 
        \begin{equation}
            \mathbb{E}\big[|\SUOT(\mu,\nu) -  \SUOT(\hat\mu_n,\hat\nu_n)|\big]\leq \kappa(n).
        \end{equation}
        \item If for $\alpha,\beta\in\Mmp(\rset)$, $\mathbb{E}\big[|\UOT(\alpha,\hat\beta_n)|\big]\leq \xi(n)$, then for $\mu,\nu\in\Mmp(\Rd)$, 
        \begin{equation}
            \mathbb{E}\big[|\SUOT(\mu,\hat\mu_n)|\big]\leq \xi(n).
        \end{equation}
    \end{enumerate}
\end{theorem}

\begin{proof}
    See \citep[Theorem 3.6]{sejourne2023unbalanced}.
\end{proof}

\Cref{thm:sample-comp} means that $\SUOT$ enjoys a \emph{dimension-free} sample complexity, even when comparing multivariate measures: this advantage is recurrent of sliced divergences \citep{nadjahi2020statistical} and further motivates their use on high-dimensional settings. 
The sample complexity rates $\kappa(n)$ or $\xi(n)$ can be deduced from the literature on $\UOT$ for univariate measures, for example we refer to~\citep{vacher2022stability} for the GHK setting. Establishing the statistical properties of $\RSOT$ may require extending the analysis in \citep{nietert2022outlier}: we leave this question for future work.

We conclude this section by deriving the dual formulations of $\SUOT, \RSOT$ and proving that strong duality holds. We will consider that $\lambda$ is approximated with $\hat{\lambda}_K = \frac{1}{K}\sum_{k=1}^K\delta_{\theta_k}$, $\theta_k \sim \lambda$. This corresponds to the routine case in practice, as practitioners usually resort to a Monte Carlo approximation to estimate the expectation \emph{w.r.t.} $\lambda$ defining $\sw$.

\begin{theorem}[Strong duality] \label{thm:duality-rsot}
    Define $\calE=\{ \forall\theta\in\mathrm{supp}(\lambda_K),\, f_\theta\oplus g_\theta\leq c \}$.
    Let $f_{avg}=\int_{S^{d-1}} f_\theta\ \mathrm{d}\hat{\lambda}_K(\theta)$, $g_{avg} = \int_{S^{d-1}} g_\theta\ \mathrm{d}\hat{\lambda}_K(\theta)$.
    Then, $\SUOT$~\eqref{eq:suot_primal} and $\RSOT$ \eqref{eq:def-rsot} can be equivalently written for $\mu, \nu \in \Mmp(\Rd)$ as,
    \begin{align}
        \SUOT(\mu,\nu) &= \sup_{(f_\theta),(g_\theta)\in\calE}\
        \int_{S^{d-1}} \Big(\int\varphi_1^\circ\big(f_\theta\circ P^\theta(x)\big)\ \mathrm{d}\mu(x)
        + \int\varphi_2^\circ\big(g_\theta\circ P^\theta(y)\big)\ \mathrm{d}\nu(y)\Big)\ \mathrm{d}\hat{\lambda}_K(\theta) \label{eq:dual-suot} \\
        \RSOT_p^p(\mu,\nu) &= \sup_{(f_\theta),(g_\theta)\in\calE}\ 
        \int \varphi_1^\circ\big(f_{avg}\circ P^\theta(x)\big)\ \mathrm{d}\mu(x)
        + \int \varphi_2^\circ\big(g_{avg}\circ P^\theta(y)\big)\ \mathrm{d}\nu(y),\label{eq:dual-rsot}
    \end{align} 
    with $\varphi_i$ defined as in \Cref{prop:strong_duality_uot} for $i\in\{1,2\}$.
\end{theorem}

\begin{proof}
    See \citep[Theorem 5]{sejourne2023unbalanced}.
\end{proof}

We conjecture that strong duality also holds for $\lambda$ Lebesgue over $S^{d-1}$. \Cref{thm:duality-rsot} has important practical implications, since it justifies the Frank-Wolfe-type algorithms that we develop in \Cref{sec:implem} to compute $\SUOT$ and $\RSOT$ in practice.

\section{Computing SUOT and USW with Frank-Wolfe algorithms} \label{sec:implem}

In this section, we explain how to implement $\SUOT$ and $\RSOT$. We propose two algorithms by leveraging our strong duality result (\Cref{thm:duality-rsot}) along with a Frank-Wolfe algorithm~(FW) \citep{frank1956algorithm} introduced in \citep{sejourne2022faster} to optimize the $\UOT$ dual~\ref{eq:dual-uot}. We refer to \citep{sejourne2023unbalanced} for more details on the technical implementation and theoretical justification of our methodology.

FW is an iterative procedure which aims at maximizing a functional $\calH$ over a compact convex set $\calE$, by maximizing a linear approximation $\nabla\calH$: given iterate $x^t$, FW solves the linear oracle 
$r^{t+1}\in\argmax_{r\in\calE}\ps{\nabla\calH(x^t)}{r}$ and performs a convex update $x^{t+1} = (1-\gamma_{t+1})x^t +\gamma_{t+1}r^{t+1}$, with $\gamma_{t+1}$ typically chosen as $\gamma_{t+1} = 2 / (2+ t +1)$. We call this step \texttt{FWStep} 
in our pseudo-code. When applied in~\citep{sejourne2022faster} to compute $\UOT(\mu, \nu)$ dual~\eqref{eq:dual-uot}, \texttt{FWStep} updates $(f_t,g_t)$ s.t. $f_t\oplus g_t\leq c$, and the linear oracle is the balanced dual of $\OT(\mu_t,\nu_t)$ where $(\mu_t,\nu_t)$ are normalized versions of $(\mu,\nu)$. Updating $(\mu_t, \nu_t)$ involves $(f_t,g_t)$ and $\bm{\rho}=(\rho_1,\rho_2)$: we refer to this routine as $\texttt{Norm}(\mu,\nu,f_t,g_t, \bm{\rho})$ 
and the closed-form updates are reported in \citep[Appendix B]{sejourne2023unbalanced}. In other words, computing $\UOT$ amounts to solving a sequence of $\OT$ problems, which can efficiently be done for univariate measures \citep{sejourne2022faster}.

Analogously to $\UOT$, and by \Cref{thm:duality-rsot}, we propose to compute $\SUOT(\mu,\nu)$ and $\RSOT(\mu,\nu)$ based on their dual forms. 
FW iterates consists in solving a sequence of sliced OT problems. 
We derive the updates for the \texttt{FWStep} tailored for $\SUOT$ and $\RSOT$ in \citep[Appendix B]{sejourne2023unbalanced}, and re-use the aforementioned \texttt{Norm} routine. For $\RSOT$, we implement an additional routine called $\texttt{\textbf{AvgPot}}\big((f_\theta)\big)$ to compute $\int f_\theta \ \mathrm{d}\hat{\lambda}_K(\theta)$ given the sliced potentials $(f_\theta)$.

A crucial difference is the need of $\SOT$ dual potentials $(r_\theta,s_\theta)$ to call $\texttt{Norm}$. However, past implementations only return the loss $\SOT(\mu,\nu)$ for \emph{e.g.} training models~\citep{deshpande2019max, nguyen2020distributional}.
Thus we designed two novel (GPU) implementations in \texttt{PyTorch}~\citep{pytorch} which return them.
The first one leverages that the gradient of $\OT(\mu,\nu)$ \emph{w.r.t.} $(\mu,\nu)$ are optimal $(f, g)$, which allows to backpropagate $\OT(P^\theta_\#\mu,P^\theta_\#\nu)$ \emph{w.r.t.} $(\mu,\nu)$ to obtain $(r_\theta,s_\theta)$.
The second implementation computes them in parallel on GPUs using their closed form, which to the best of our knowledge is a new sliced algorithm.
We call $\texttt{SlicedDual}(P^\theta_\#\mu, P^\theta_\#\nu)$ 
the step returning optimal $(r_\theta,s_\theta)$ solving $\OT(P^\theta_\#\mu, P^\theta_\#\nu)$ for all $\theta$.
Both routines preserve the $O(N\log N)$ per slice time complexity and can be adapted to any sliced Optimal Transport variant.
Thus, our FW approach is modular in that one can reuse the $\SOT$ literature.
We illustrate this by computing the Unbalanced $\ghsw$ between distributions in the hyperbolic Poincaré disk (\Cref{fig:usw_hsw}).

\paragraph{Algorithmic complexity.} FW algorithms and its variants have been widely studied theoretically.
Computing \texttt{SlicedDual} has a complexity $O(KN\log N)$, where $N$ is the number of samples, and $K$ the number of projections of $\hat{\lambda}_K$.
The overall complexity of $\SUOT$ and $\RSOT$ is thus $O(FKN\log N)$, where $F$ is the number of FW iterations needed to reach convergence.
Our setting falls under the assumptions of~\cite[Theorem 8]{lacoste2015global}, thus ensuring fast convergence of our methods.
We plot in \citep[Appendix B]{sejourne2023unbalanced} empirical evidence that a few iterations of FW ($F\leq 20$) suffice to reach numerical precision.

\begin{figure}[t]
    \begin{minipage}[t]{.47\textwidth}
        \begin{algorithm}[H]
            \caption{-- $\SUOT$} \label{algo:suot-pseudo}
            \small{
                \textbf{Input:} 
                $\mu$, $\nu$, $F$, $(\theta_k)_{k=1}^K$, $\bm{\rho}=(\rho_1, \rho_2)$ \\
                \textbf{Output:}~$\SUOT(\mu,\nu)$, $(f_\theta,g_\theta)$\\
                \vspace*{-1.0em}
                \begin{algorithmic}
                    \STATE 
                    $(f_\theta, g_\theta)\leftarrow (0,0)$
                    \FOR{$t = 0, 1, \dots, F-1$, \textbf{for} $\theta \in (\theta_k)_{k=1}^K$ }
                        \STATE \ali{3em}{$(\mu_\theta, \nu_\theta)$}              ${}\leftarrow \texttt{Norm}(P^\theta_\#\mu, P^\theta_\#\nu,f_\theta,g_\theta, \bm{\rho})$
                        \STATE \ali{3em}{$(r_\theta, s_\theta)$} ${}\leftarrow \texttt{SlicedDual}(\mu_\theta, \nu_\theta)$
                        \STATE \ali{3em}{$(f_\theta,g_\theta)$}      ${}\leftarrow \texttt{FWStep}(f_\theta,g_\theta,r_\theta,s_\theta,\gamma_t)$
                    \ENDFOR
                    \STATE Return  $\SUOT(\mu,\nu)$, $(f_\theta,g_\theta)$ as in~\eqref{eq:dual-suot}
                    \end{algorithmic}
            }
        \end{algorithm}
    \end{minipage} \hfill
    \begin{minipage}[t]{.52\textwidth}
        \begin{algorithm}[H]
            \caption{-- $\RSOT$}\label{algo:rsot-pseudo}
            \small{
                \textbf{Input:} 
                $\mu$, $\nu$, $F$, $(\theta_k)_{k=1}^K$, $\bm{\rho}=(\rho_1, \rho_2)$, $p$ \\
                \textbf{Output:}~$\RSOT(\mu,\nu)$, $(f_{avg}, g_{avg})$\\
                \vspace*{-1.0em}
                \begin{algorithmic}
                    \STATE 
                    $(f_\theta,g_\theta,f_{avg},g_{avg})\leftarrow(0,0,0,0)$
                    \FOR{$t = 0, 1, \dots, F-1$, \textbf{for} $\theta \in (\theta_k)_{k=1}^K$}
                        \STATE \ali{4em}{$(\pi_1,\pi_2)$} ${}\leftarrow \texttt{Norm}(\mu,\nu,f_{avg},g_{avg}, \bm{\rho})$
                        \STATE \ali{4em}{$(r_\theta,s_\theta)$} ${}\leftarrow \texttt{SlicedDual}(P^\theta_\#\pi_1,P^\theta_\#\pi_2)$ 
                        \STATE \ali{4em}{$r_{avg},s_{avg}$} ${}\leftarrow \texttt{AvgPot}(r_\theta), \texttt{AvgPot}(s_\theta)$ 
                        \STATE \ali{4em}{$(f_{avg}, g_{avg})$} ${}\leftarrow \texttt{FWStep}(f_{avg},g_{avg},r_{avg},s_{avg},\gamma_t)$ 
                    \ENDFOR
                    \STATE Return $\RSOT_p^p(\mu,\nu)$, $(f_{avg},g_{avg})$ as in~\eqref{eq:dual-rsot}
                \end{algorithmic}
            }
        \end{algorithm}
    \end{minipage}
\end{figure}

\paragraph{Outputing marginals of $\SUOT$ and $\RSOT$.}

The optimal primal marginals of $\UOT$ (and a fortiori $\SUOT$ and $\RSOT$) are geometric normalizations of inputs $(\mu,\nu)$ with discarded outliers.
Their computation involves the $\texttt{Norm}$ routine, using optimal dual potentials.
This is how we compute marginals in Figures~(\ref{fig:illustration_suot}, \ref{fig:usw_hsw}, \ref{fig:bary}).
We refer to \citep[Appendix B]{sejourne2023unbalanced} for more details and formulas.

\paragraph{Stochastic $\RSOT$.}
In practice, the measure $\hat{\lambda}_K=\frac{1}{K}\sum_i^K \delta_{\theta_i}$ is fixed, and $(f_{avg},g_{avg})$ are computed w.r.t. $\hat{\lambda}_K$.
However, the process of sampling $\hat{\lambda}_K$ satisfies $\mathbb{E}_{\theta_k\sim\lambda}[\hat{\lambda}_K] = \lambda$.
Thus, assuming \Cref{thm:duality-rsot} still holds for $\lambda$, it yields $\mathbb{E}_{\theta_k\sim\lambda}[f_{avg}(x)]=\int f_\theta\big( P^\theta(x)\big)\ \mathrm{d}\lambda(\theta)$ if we sample a new $\hat{\lambda}_K$ at each FW step. 
We call this approach \emph{Stochastic $\RSOT$}.
It outputs a more accurate estimate of the true $\RSOT$ \emph{w.r.t.} $\lambda$.
It is more expensive, as we need to sort projected data \emph{w.r.t} new projections at each iteration, 
More importantly, for balanced OT ($\varphi^\circ(x)=x$), one has $\RSOT=\SOT$ and this idea remains valid for sliced OT. See \Cref{sec:usw_xp} for applications.

\section{Experiments} \label{sec:usw_xp}

This section presents a set of numerical experiments, which illustrate the effectiveness, computational efficiency and versatility of $\SUOT$ and $\RSOT$, as implemented by \Cref{algo:suot-pseudo,algo:rsot-pseudo}.
We first evaluate $\RSOT$ between measures supported on hyperbolic data leveraging the Geodesic Hyperbolic Sliced-Wasserstein distance defined in \Cref{chapter:hsw}, and investigate the influence of the hyperparameters $\rho_1$ and $\rho_2$. 
Then, we solve a document classification problem with $\SUOT$ and $\RSOT$, and compare their performance (in terms of accuracy and computational complexity) against classical OT losses. 
Our last experiment is conducted on large-scale datasets from a real-life application: we deploy $\RSOT$ to compute barycenters of climate datasets in a robust and efficient manner.

\subsection{Comparing Hyperbolic Datasets.}

\newcommand{\myrot}[1]{ \rotatebox{90}{\small \hspace{2mm} #1}}
\newcommand{\myimg}[1]{\includegraphics[width=0.16\textwidth]{#1}}

\begin{wrapfigure}{R}{0.2\textwidth}
\centering
\vspace{-1.0em}
	\centering
		\begin{tabular}{@{}c@{}}
        Inputs $(\mu,\nu)$ \\
		\myimg{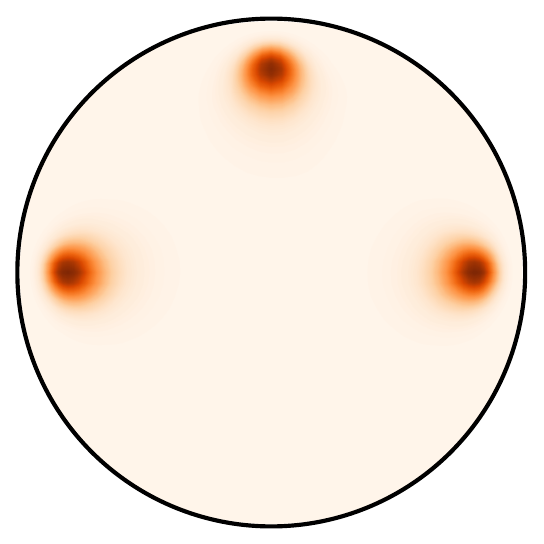} \\  %
        \myimg{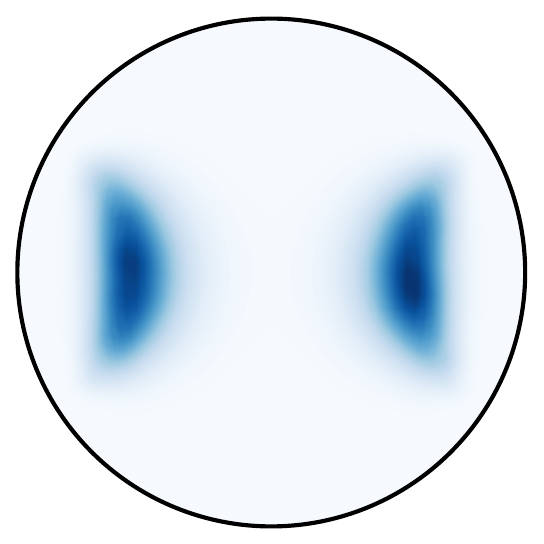} 
		\end{tabular}
\vspace{-2.0em}
\end{wrapfigure}

\begin{figure}[t]
    \centering
    \begin{tabular}{@{}c@{}c@{}c@{}c@{}c@{}c@{}c@{}}
    & \multicolumn{2}{c}{$\rho=10^{-3}$} & \multicolumn{2}{c}{$\rho=10^{-1}$} & \multicolumn{2}{c}{$\rho=10^{1}$} \\
    \myrot{$\quad(\pi_1,\pi_2)$ }  & \myimg{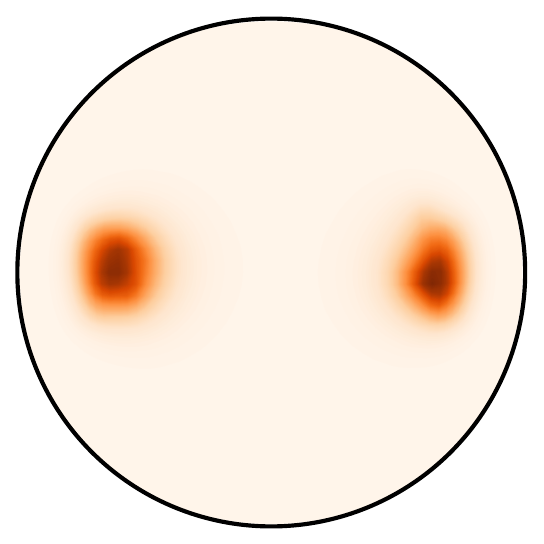} &  %
        \myimg{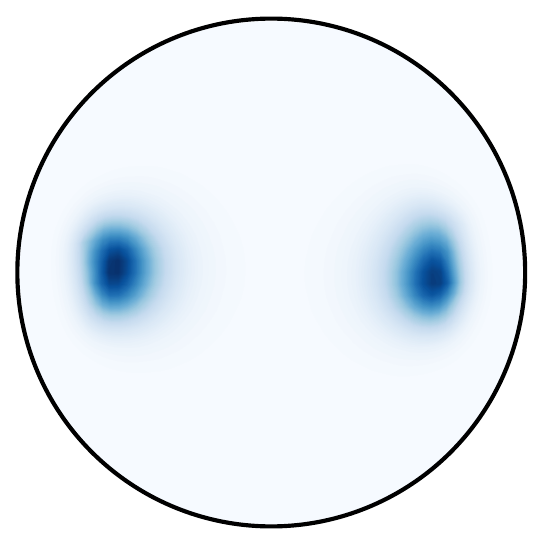} & %
        \myimg{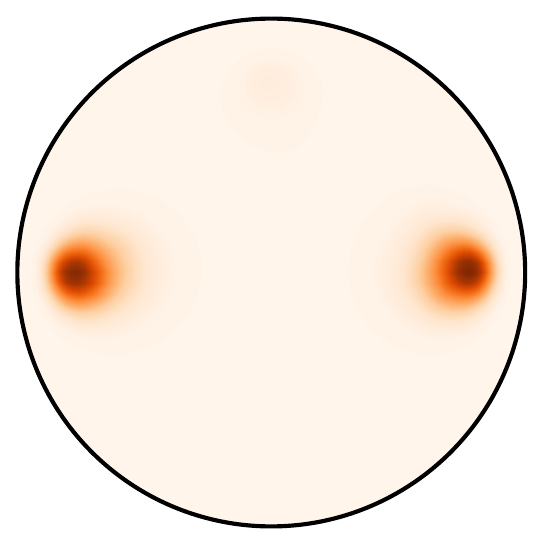} &  %
        \myimg{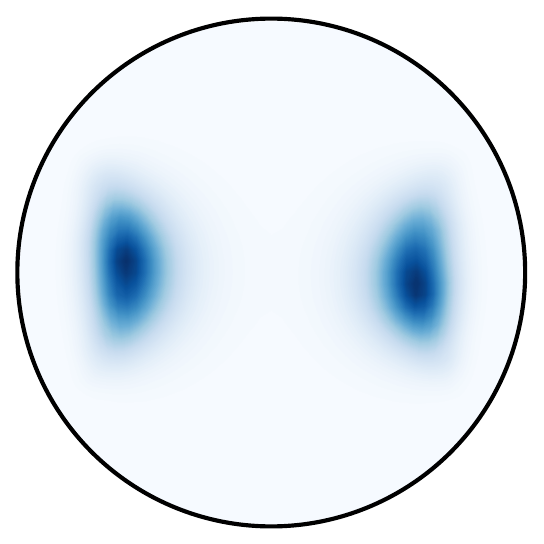} & %
        \myimg{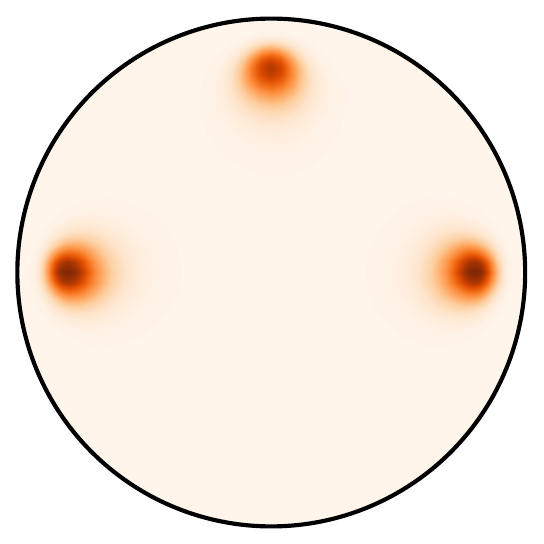} &  %
        \myimg{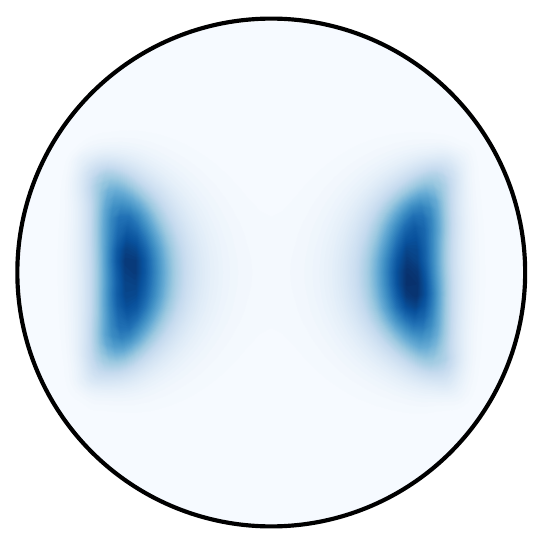} 
    \end{tabular}
    \caption{KDE estimation (kernel $e^{-d_{\mathbb{B}}^2/\sigma}$) of optimal $(\pi_1,\pi_2)$ of $\mathrm{UGHSW}_2^2(\mu,\nu)$.}
    \label{fig:usw_hsw}
\end{figure}

We display in \Cref{fig:usw_hsw} the impact of the parameter $\rho=\rho_1=\rho_2$ on the optimal marginals of $\RSOT$. To illustrate the modularity of our FW algorithm, our inputs are synthetic mixtures of Wrapped Normal Distribution on the $2$-hyperbolic manifold $\mathbb{B}^2$~\citep{nagano2019wrapped}, so that the FW oracle is $\ghsw$ defined in \eqref{eq:ghsw_poincare} instead of $\sw$.

We display the $2$-hyperbolic manifold on the Poincaré disk.
The measure $\mu$ (in red) is a mixture of 3 isotropic normal distributions, with a mode at the top of the disc playing the role of an outlier.
The measure $\nu$ is a mixture of two anisotropic normal distributions, whose means are close to two modes of $\mu$, but are slightly shifted at the disk's center.

\looseness=-1 We wish to illustrate several take-home messages, stated in \Cref{sec:usw_theory}.
First, the optimal marginals $(\pi_1,\pi_2)$ are renormalisation of $(\mu,\nu)$ accounting for their geometry, which are able to remove outliers for properly tuned $\rho$.
When $\rho$ is large, $(\pi_1,\pi_2)\simeq(\mu,\nu)$ and we retrieve $\SOT$.
When $\rho$ is too small, outliers are removed, but we see a shift of the modes, so that modes of $(\pi_1,\pi_2)$ are closer to each other, but do not exactly correspond to those of $(\mu,\nu)$.
Second, note that such plot cannot be made with $\SUOT$, since the optimal marginals depend on the projection direction (see \Cref{fig:illustration_suot}).
Third, we emphasize that we are indeed able to reuse any variant of $\SOT$ existing in the literature.

\subsection{Document Classification}

\begin{figure}[t]
    \centering
    \begin{minipage}{0.67\linewidth}
        \centering
        \captionof{table}{Accuracy on document classification}
        \resizebox{\columnwidth}{!}{
            \begin{tabular}{ccccc}
                & \textbf{BBCSport} & \textbf{Movies} & \textbf{Goodreads genre} & \textbf{Goodreads like} \\ \toprule
                $W_2$ & 94.55 & 74.44 & 55.22 & 71.00 \\
                UOT & 96.73 & - & - & - \\
                Sinkhorn UOT & 95.45 & 72.48 & 53.55 & 67.81\\
                $\sw_2$ & $89.39_{\pm 0.76}$ & $66.95_{\pm 0.45}$ & $50.09_{\pm 0.51}$ & $65.60_{\pm 0.20}$ \\
                SUOT & $90.12_{\pm 0.15}$ & $67.84_{\pm 0.37}$ & $50.15_{\pm 0.04}$ & $66.72_{\pm 0.38}$ \\
                $\RSOT_2$ & $92.36_{\pm 0.07}$ & $69.21_{\pm 0.37}$ &  $51.87_{\pm 0.56}$ & $67.41_{\pm 1.06}$\\
                S$\RSOT_2$ & $92.45_{\pm 0.39}$ & $69.53_{\pm 0.53}$ & $51.93_{\pm 0.53}$ & $67.33_{\pm 0.26}$ \\
                \bottomrule
            \end{tabular}
            }
        \label{tab:table_acc}
    \end{minipage}
    \hfill
    \begin{minipage}{0.31\linewidth}
        \centering
        \includegraphics[width=\linewidth]{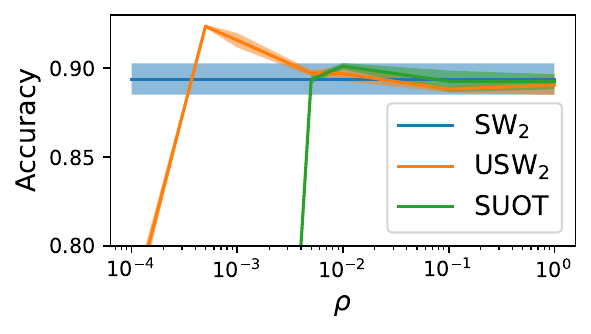}
        \caption{Ablation on BBCSport of the parameter $\rho$.}
        \label{fig:ablation_rho}
    \end{minipage}
\end{figure}

To show the benefits of our proposed losses over $\sw$, we consider a document classification problem \citep{kusner2015word}. Documents are represented as distributions of words embedded with \emph{word2vec} \citep{mikolov2013distributed} in dimension $d=300$. Let $D_k$ be the $k$-th document and $x_1^k,\dots,x_{n_k}^k\in \Rd$ be the set of words in $D_k$. Then, $D_k = \sum_{i=1}^{n_k} w_i^k \delta_{x_i^k}$ where $w_i^k$ is the frequency of $x_i^k$ in $D_k$ normalized \emph{s.t.} $\sum_{i=1}^{n_k} w_i^k = 1$. Given a loss function $\mathrm{L}$, the document classification task is solved by computing the matrix $\big(\mathrm{L}(D_k, D_\ell)\big)_{k,\ell}$, then using a k-nearest neighbor classifier. Since a word typically appears several times in a document, the measures are not uniform and sliced partial OT \citep{bonneel2019spot, bai2022sliced} cannot be used in this setting. The aim of this experiment is to show that by discarding possible outliers using a well chosen parameter $\rho$, $\RSOT_2$ is able to outperform $\sw_2$ and $\SUOT$ on this task while scaling better for large-scale documents compared to $W_2$ and UOT. We consider three different datasets, BBCSport \citep{kusner2015word}, Movies reviews \citep{Pang+Lee+Vaithyanathan:02a} and the Goodreads dataset \citep{maharjan2017multi}. For the latter, we perform two classification tasks by predicting the genre (8 classes) as well as the likability (2 classes) which is defined as in \citep{maharjan2017multi}. The two first datasets are not composed of large documents, and hence there is no real computational benefit compared to computing the Wasserstein distance, but we report them in order to illustrate the benefits of using $\RSOT$ over $\sw$ or $W$. The Goodreads dataset is composed of parts of books, and contains 1491 words on average. In this setting, there is indeed a computational benefit. We report in \Cref{appendix:doc_classif} more details on the experiment and on the datasets.

We report in \Cref{tab:table_acc} the accuracy of $\SUOT$, $\RSOT_2^2$ and the stochastic $\RSOT_2^2$ (S$\RSOT_2^2$) compared with $\sw_2^2$, $W_2^2$ and UOT computed with the majorization minimization algorithm \citep{chapel2021unbalanced} or approximated with the Sinkhorn algorithm \citep{pham2020unbalanced}. All results reported are the mean over 5 different train/test set. 
All the benchmark methods are computed using the POT library \citep{flamary2021pot}. For sliced methods ($\sw$, SUOT, $\RSOT$ and S$\RSOT$), we average over 3 computations of the loss matrix and report the standard deviation in \Cref{tab:table_acc}. The number of neighbors was selected via cross validation. The results in \Cref{tab:table_acc} are reported for $\rho$ yielding the best accuracy, and we display an ablation of this parameter on the BBCSport dataset in \Cref{fig:ablation_rho}. 
We observe that when $\rho$ is tuned, $\RSOT$ outperforms SOT, just as UOT outperforms OT.

\paragraph{Runtime.}

\begin{figure}[t]
    \centering
    \includegraphics[width=0.4\linewidth]{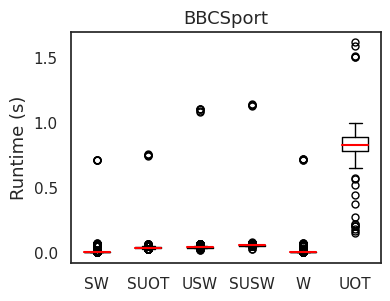}
    \includegraphics[width=0.4\linewidth]{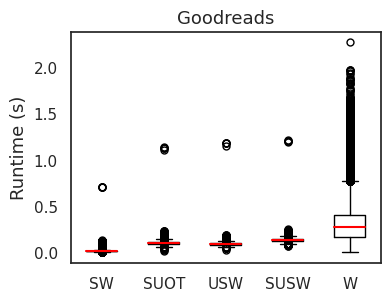}
    \caption{Runtime on the BBCSport dataset (\textbf{Left}) and on the Goodreads dataset (\textbf{Right}).}
    \label{fig:docs_runtime}
\end{figure}

We report in \Cref{fig:docs_runtime} the runtime of computing the different discrepancies between each pair of documents. On the BBCSport dataset, the documents have 116 words on average, thus the main bottleneck is the projection step for sliced OT methods. Hence, we observe that $W$ runs slightly faster than $\sw$ and the sliced unbalanced counterparts. Goodreads is a dataset with larger documents, with on average 1491 words per document. Therefore, as OT scales cubically with the number of samples, we observe here that all sliced methods run faster than OT, which confirms that sliced methods scale better \emph{w.r.t.} the number of samples.
In particular, computing the OT matrix entirely took 3 times longer than computing the $\RSOT$ matrix on GPU.
In this setting, we were not able to compute UOT with the POT implementation in a reasonable time. Computations have been performed with a NVIDIA A100 GPU.

\paragraph{Ablations.}

\begin{figure}[t]
    \centering
    \includegraphics[width=0.4\linewidth]{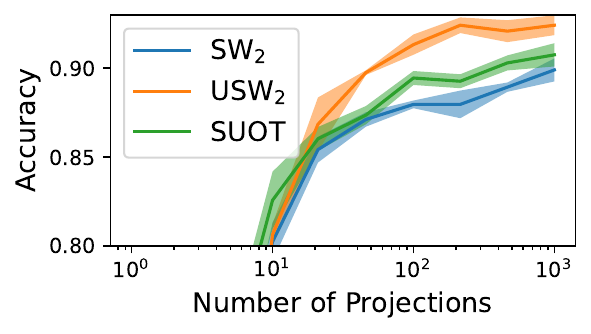}
    \includegraphics[width=0.4\linewidth]{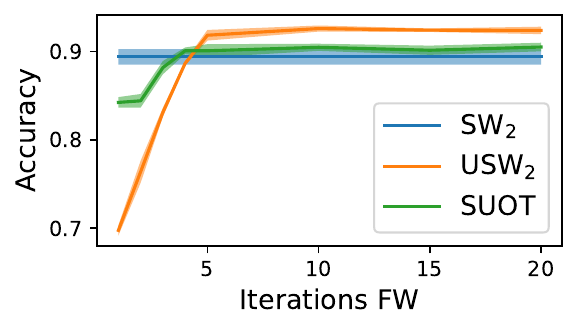}
    \caption{Ablation on BBCSport of the number of projections (\textbf{Left}) and of the number of Frank-Wolfe iterations (\textbf{Right}).}
    \label{fig:ablations_docs}
\end{figure}

We plot in \Cref{fig:ablations_docs} accuracy as a function of the number of projections and the number of iterations of the Frank-Wolfe algorithm. We averaged the accuracy obtained with the same setting described in \Cref{xp_docs:technicals}, with varying number of projections $K\in \{4, 10, 21, 46, 100, 215, 464, 1000\}$ and number of FW iterations $F\in\{1,2,3,4,5,10,15,20\}$. Regarding the hyperparameter $\rho$, we selected the one returning the best accuracy, \emph{i.e.}~$\rho=5\cdot 10^{-4}$ for $\RSOT$ and $\rho=10^{-2}$ for $\SUOT$.

\subsection{Barycenter on Geophysical Data.}

\begin{figure}[t]
  \centering
  \includegraphics[width=\textwidth]{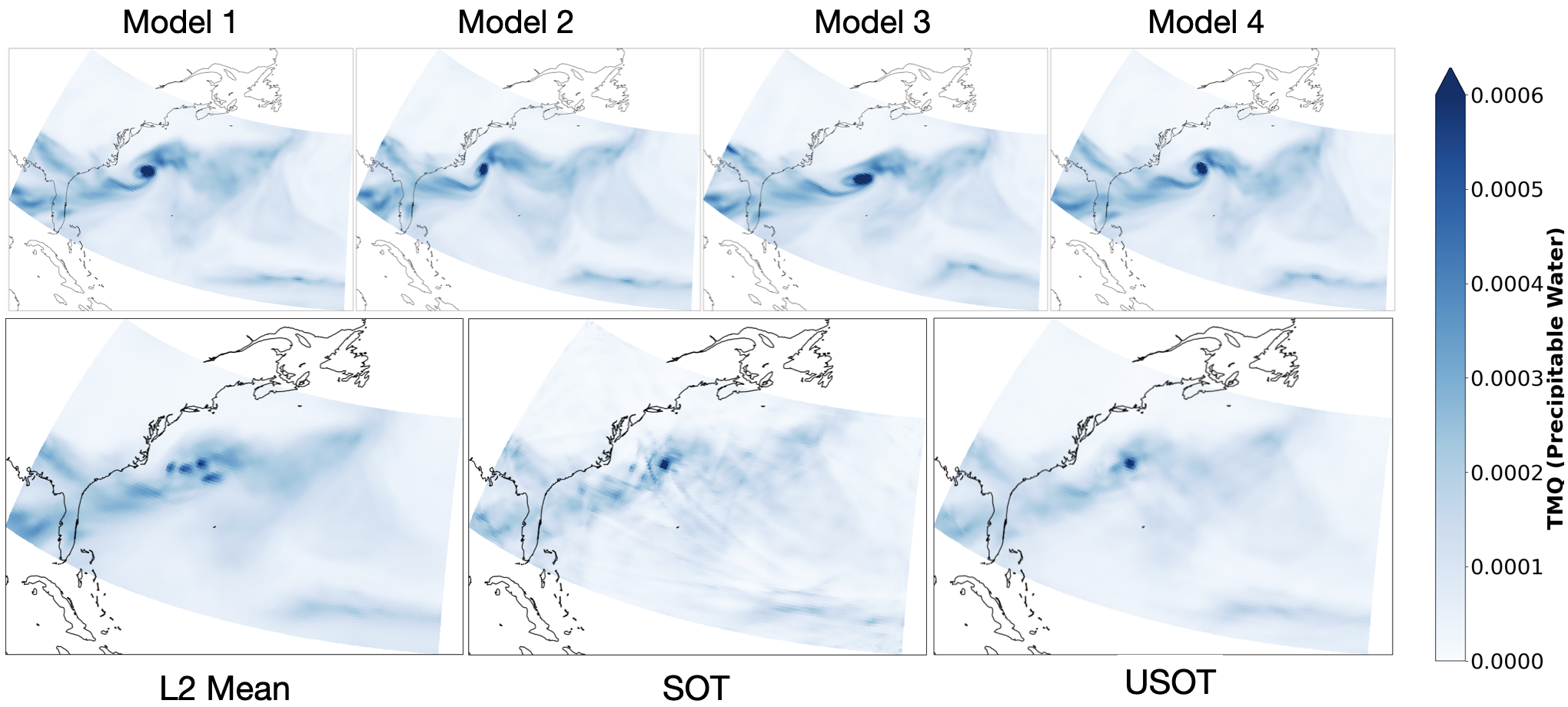}
  \caption{{\em Barycenter of geophysical data}. ({\bf First row}) Simulated output of 4 different climate models depicting different scenarios for the evolution of a tropical cyclone ({\bf Second row}) Results of different averaging/aggregation strategies.\vspace*{-0.5em}}
  \label{fig:bary}
\end{figure}

OT barycenters are an important topic of interest~\citep{le2021robust} for their ability to capture mass changes and spatial deformations over several reference measures. In order to compute barycenters under the $\RSOT$ geometry on a fixed grid, we employ a mirror-descent strategy similar to~\citep[Algorithm (1)]{cuturi2014fast} and described more in depth in \citep[Appendix C]{sejourne2023unbalanced}. We showcase unbalanced sliced OT barycenter using climate model data. Ensembles of multiple models are commonly employed to reduce biases and evaluate uncertainties in climate projections ({\em e.g.}~\citep{sanderson2015,thao2022combining}). The commonly used Multi-Model Mean approach assumes models are centered around true values and averages the ensemble with equal or varying weights. However, spatial averaging may fail in capturing specific characteristics of the physical system at stake. We propose to use the $\RSOT$ barycenter here instead. We use data from the ClimateNet dataset~\citep{kashinath2021climatenet}, and more specifically the TMQ (precipitable water) indicator. The ClimateNet dataset is a human-expert-labeled curated dataset that captures notably tropical cyclones (TCs). In order to simulate the output of several climate models, we take a specific instant (first date of 2011) and deform the data with the elastic deformation from \texttt{TorchVision}~\citep{pytorch}, in an area located close to the eastern part of the United States of America. As a result, we obtain 4 different TCs, as shown in the first row of \Cref{fig:bary}. The classical L2 spatial mean is displayed on the second row of \Cref{fig:bary} and, as can be expected, reveal 4 different TCs centers/modes, which is undesirable. As the total TMQ mass in the considered zone varies between the different models, a direct application of $\SOT$ is impossible, or requires a normalization of the mass that has undesired effect as can be seen on the second picture of the second row. Finally, we show the result of the $\RSOT$ barycenter with $\rho_1 = 1e1$ (related to the data) and $\rho_2 = 1e4$ (related to the barycenter). As a result, the corresponding barycenter has only one apparent mode which is the expected behavior. The considered measures have a size of $100\times 200$, and we run the barycenter algorithm for $500$ iterations (with $K=64$ projections), which takes $3$ minutes on a commodity GPU. $\UOT$ barycenters for this size of problems are intractable, and to the best of our knowledge, this is the first time such large scale unbalanced OT barycenters can be computed. This experiment encourages an in-depth analysis of the relevance of this aggregation strategy for climate modeling and related problems, which we will investigate as future work.

\section{Conclusion and Discussion}

We proposed two losses merging unbalanced and sliced OT altogether, with theoretical guarantees and an efficient Frank-Wolfe algorithm which allows to reuse any sliced OT variant. We highlighted experimentally the performance improvement over SOT, and described novel applications of unbalanced OT barycenters of positive measures, with a new case study on geophysical data. 
These novel results and algorithms pave the way to numerous new applications of sliced variants of OT, and we believe that our contributions will motivate practitioners to further explore their use in general ML applications, without the requirements of manipulating probability measures.

On the limitations side, an immediate drawback arises from the induced additional computational cost \emph{w.r.t.} $\sw$. 
While the above experimental results show that $\SUOT$ and $\RSOT$ improve performance significantly over $\sw$, and though the complexity is still sub-quadratic in number of samples, our FW approach uses $\sw$ as a subroutine, rendering it necessarily more expensive.
Additionally, another practical burden comes from the introduction of extra parameters $(\rho_1,\rho_2)$ which requires cross-validation when possible. 
Therefore, a future direction would be to derive efficient strategies to tune $(\rho_1,\rho_2)$, maybe \emph{w.r.t.} the applicative context, and further complement the possible interpretations of $\rho$ as a ``threshold'' for the geometric information encoded by the costs.

On the theoretical side, while OT between univariate measures has been shown to define a reproducing kernel, and while sliced OT can take advantage of this property~\citep{kolouri2016sliced, carriere2017sliced}, some of our numerical experiments suggest this property no longer holds for $\UOT$ (and therefore, for $\SUOT$ and $\RSOT$). 
This negative result leaves as an open direction the design of OT-based kernel methods between arbitrary positive measures.

\clearemptydoublepage
\cleartooddpage[\thispagestyle{empty}]
\chapter{Busemann Function in Wasserstein Space} \label{chapter:busemann}

{
    \hypersetup{linkcolor=black} 
    \minitoc 
}

\looseness=-1 This chapter is more prospective and studies the problem of defining the Busemann function on the space of probability measures endowed by the Wasserstein distance. This function has received recently much attention on Riemannian manifolds where all geodesics can be extended infinitely. On the Wasserstein space, this is not the case, and hence the Busemann function is only well defined with respect to specific geodesics which we investigate in this chapter. Then, we provide closed-forms in particular cases such as one dimensional probability distributions or Gaussians. Finally, we propose an application to the problem of Principal Component Analysis (PCA) in the space of one dimensional probability distributions. 

\section{Introduction}

Many datasets are composed of probability distributions. For example, one can cite one dimensional histograms which can describe \emph{e.g.} empirical return distributions financial assets or age distributions of countries among others \citep{campbell2022efficient}, documents which can be modeled as distributions of words \citep{kusner2015word}, cells as distributions of genes \citep{bellazzi2021gene}, images which can be seen as a distribution over a 2D grid \citep{seguy2015principal}, or more broadly symbolic data \citep{verde2015dimension}. It has also been proposed in several works to embed data directly into probability spaces as they provide a rich geometry \citep{xiong2023geometric}. For instance, \citet{vilnis2015word} embedded words into Gaussians while \citet{wang2022dirie} embedded knowledge graphs into Dirichlet distributions.

\looseness=-1 Probability distributions can be naturally dealt with using Optimal Transport by endowing the space with the Wasserstein distance. With this metric, this space, called the Wasserstein space, enjoys many theoretical properties which have been extensively studied \citep{ambrosio2008gradient, villani2009optimal}. Leveraging these properties, it has been applied in Machine Learning in order to deal with data sets of probability distributions. For instance, \citet{agueh2011barycenters} investigated Wasserstein barycenters which provide a way to find an average of such datasets, \citet{domazakis2019clustering, zhuang2022wasserstein} used clustering of probability distributions by extending the K-Means algorithm and \citet{schmitz2018wasserstein} performed dictionary learning in order to sum up a dataset of distributions. Another line of works consists in extending Principal Component Analysis (PCA) to datasets of probability distributions in order to describe the main modes of variations by exploiting the geodesic structure of the Wasserstein space \citep{seguy2015principal,bigot2017geodesic, cazelles2018geodesic, pegoraro2022projected, beraha2023wasserstein}.

In this chapter, motivated by the recent proposal to use the Busemann function in order to perform PCA in Hyperbolic spaces \citep{chami2021horopca}, we propose to study this function in Wasserstein space. \citet{zhu2021busemann} recently provided a theoretical analysis of its existence in this space but did not detail how to compute it in practice. In particular, the Busemann function is associated with geodesic rays. However, the Wasserstein space is not geodesically complete, and these geodesics need to be chosen carefully in practice. Thus, we propose to bridge this gap by first analyzing conditions to obtain geodesic rays in the Wasserstein space. Then, we provide closed-forms for the Busemann function for one dimensional probability distributions and for Gaussians, \emph{i.e.} in the Bures-Wasserstein space. Finally, as an application, we perform PCA of 1D histograms.




\section{Geodesic Rays in Wasserstein Space} \label{section:geodesic_rays}

\subsection{Background on Wasserstein Space}

\looseness=-1 We start by providing some background on Wasserstein spaces $(\mathcal{P}_2(\mathbb{R}^d), W_2)$. First, it is well known that the Wasserstein space has a Riemannian structure \citep{otto2001geometry}. In particular, this is a geodesic space and between two measures $\mu_0,\mu_1\in\mathcal{P}_2(\mathbb{R}^d)$, the geodesic $t\mapsto \mu_t$ is the displacement interpolation, introduced by \citet{mccann1997convexity} and defined as, 
\begin{equation}
    \forall t\in [0,1],\ \mu_t = \big((1-t)\pi^1 + t \pi^2\big)_\#\gamma,
\end{equation}
where $\gamma\in\Pi(\mu_0,\mu_1)$ is an optimal coupling, $\pi^1:(x,y)\mapsto x$ and $\pi^2:(x,y)\mapsto y$.
In the case where there is a Monge map between $\mu_0$ and $\mu_1$, \emph{e.g.} if $\mu_0$ is absolutely continuous with respect to the Lebesgue measure by Brenier's theorem (see \Cref{th:brenier}), then the geodesic curve can be further written as
\begin{equation}
    \forall t\in [0,1],\ \mu_t=\big((1-t)\id + t T\big)_\#\mu_0,
\end{equation}
with $T$ the Optimal Transport map between $\mu_0$ and $\mu_1$. This geodesic is also a constant-speed geodesic \citep[Theorem 5.27]{santambrogio2015optimal}, \emph{i.e.} it satisfies
\begin{equation}
    \forall s,t\in [0,1],\ W_2(\mu_t,\mu_s) = |t-s| W_2(\mu_0,\mu_1).
\end{equation}
We call $\kappa_\mu = W_2(\mu_0, \mu_1)$ the speed of the geodesic. If the geodesic can be extended to any $t\in\mathbb{R}$, it is called a geodesic line and a geodesic ray when it can be extended to any $t\in \mathbb{R}_+$ \citep{bridson2013metric}. 

\citet{kloeckner2010geometric} started to study under which conditions on the measures $\mu_0$ and $\mu_1$ the geodesics can be extended. For instance, in \citep[Proposition 3.6]{kloeckner2010geometric}, it was shown that the geodesic curve $t\mapsto \mu_t$ is a geodesic line if and only if there is a vector $u\in\mathbb{R}^d$ such that $\mu_1 = T^u_\#\mu_0$ where $T^u(x) = x-u$ for any $x\in\mathbb{R}^d$, \emph{i.e.} the measures are translated. Geodesic rays, which will be of much interest in order to define Busemann functions, received some attention in \citep{zhu2021busemann} in which it was shown that for any $\mu_0\in\mathcal{P}_2(\mathbb{R}^d)$, there exists at least one unit-speed geodesic ray originating from it.

\subsection{Geodesic Rays} 

Now, let us discuss how to characterize geodesic rays in practice. First, we show that in the setting of Brenier's theorem, geodesic rays are obtained if the Monge map between $\mu_0$ and $\mu_1$ is the gradient of a 1-convex Brenier potential function $u$, \emph{i.e.} such that $x\mapsto u(x) - \frac{\|x\|_2^2}{2}$ is convex.

\begin{proposition} \label{prop:geodesic_rays}
    Let $\mu_0, \mu_1\in\mathcal{P}_2(\mathbb{R}^d)$ with $\mu_0$ absolutely continuous with respect to the Lebesgue measure and consider $c(x,y)=\frac12 \|x-y\|_2^2$. Then, the optimal transport map $T$ between $\mu_0$ and $\mu_1$ is the gradient of a 1-convex function $u$ if and only if $\mu_t = \big((1-t)\id + t T\big)_\#\mu_0$ is a geodesic ray.
\end{proposition}

\begin{proof}
    See \Cref{proof:prop_geodesic_rays}.
\end{proof}

This result is very connected with \citep[Section 4]{natale2022geodesic} in which it is stated that a geodesic can be extended on a segment $[0,\alpha]$ for $\alpha\ge 1$ if and only if $x\mapsto \alpha u(x) - (\alpha-1)\frac{\|x\|_2^2}{2}$ is convex (if and only if $x\mapsto u(x) - (1-\frac{1}{\alpha})\frac{\|x\|_2^2}{2}$ is convex). Taking the limit $\alpha\to\infty$, we recover the result of \Cref{prop:geodesic_rays}.

In \Cref{prop:geodesic_rays}, we restrict ourselves to absolutely continuous measures in order to be able to use Brenier's theorem to have access to an OT map. In the one dimensional case, we can obtain a result for a larger class of measures. In this case, the measures are fully characterized by their quantile functions and in particular, denoting $F_0^{-1}$ the quantile of $\mu_0\in\mathcal{P}_2(\mathbb{R})$, $F_1^{-1}$ the quantile of $\mu_1\in\mathcal{P}_2(\mathbb{R})$ and $F_t^{-1}$ the quantile of the geodesic between $\mu_0$ and $\mu_1$ at time $t\in [0,1]$, defined by $\mu_t = \big((1-t)\pi^1 + t \pi^2 \big)_\#\gamma$ with $\gamma=(F_0^{-1},F_1^{-1})_\#\mathrm{Unif}([0,1])$ the optimal coupling between $\mu_0$ and $\mu_1$, then it is well known (see \emph{e.g.} \citep[Equation 7.2.8]{ambrosio2008gradient}) that
\begin{equation}
    \forall t\in [0,1],\ F_t^{-1} = (1-t)F_0^{-1} + tF_1^{-1}.
\end{equation}
Then, as observed by \citet{kloeckner2010geometric}, as non-decreasing left continuous functions are the inverse cumulative distribution function of a probability distribution, we can extend the geodesic as long as $F_t^{-1}$ is non-decreasing.

\begin{proposition} \label{prop:1d_geodesic_rays}
    Let $\mu_0,\mu_1\in\mathcal{P}_2(\mathbb{R})$ and denote $F_0^{-1}$, $F_1^{-1}$ their quantile functions. Denote for any $t\in [0,1]$, $\mu_t = \big((1-t)\pi^1 + t\pi^2\big)_\#\gamma$ with $\gamma = (F_0^{-1},F_1^{-1})_\#\mathrm{Unif}([0,1])$ the optimal coupling between $\mu_0$ and $\mu_1$. Then, $t\mapsto \mu_t$ is a geodesic ray if and only if $F_1^{-1}-F_0^{-1}$ is non-decreasing.
\end{proposition}

\begin{proof}
    See \Cref{proof:prop_1d_geodesic_rays}.
\end{proof}

This result is actually equivalent with saying that $\mu_0$ is smaller than $\mu_1$ in the dispersive order \citep[Chapter 3.B]{shaked2007stochastic}. 
This is also equivalent with having the equality $V(\mu_0|\mu_1)=W_2^2(\mu_0,\mu_1)$ \citep[Theorem 2.6]{shu2020hopf} where $V$ is the weak (barycentric) Optimal Transport defined as 
\begin{equation}
    V(\mu_0|\mu_1) = \inf_{\gamma\in\Pi(\mu_0,\mu_1)} \int \left\| x-\int y\ \mathrm{d}\gamma_x(y)\right\|^2\ \mathrm{d}\mu_1(x),
\end{equation}
with $\gamma$ disintegrated as $\gamma=\mu_1\otimes \gamma_x$. \citet{shu2020hopf} also derived a condition for this equality to hold in higher dimensions, which is that the OT map is the gradient of a 1-convex function and satisfies an additional smoothness property on the Hessian. In 1D, this condition coincides with the OT map being the gradient of a 1-convex function, which is also equivalent with the difference of the quantiles being non-decreasing \citep[Remark 3.6]{shu2020hopf}. \citet{shu2020hopf} actually conjectures in Remark 3.6 that the result still holds without the smoothness assumption. In this case, it would exactly coincide with the conditions needed in \Cref{prop:geodesic_rays} to have geodesic rays.

Now, let us give some examples of measures $\mu_0$ and $\mu_1$ for which the resulting geodesic is a ray.

\paragraph{1D Gaussians.} We start by studying the one dimensional Gaussian case. Let $\mu_0=\mathcal{N}(m_0,\sigma_0^2)$ and $\mu_1=\mathcal{N}(m_1, \sigma_1^2)$ with $m_0,m_1,\sigma_0,\sigma_1\in\mathbb{R}$. It is well known that for $p\in [0,1]$, $F_{0}^{-1}(p) = m_0 + \sigma_0 \phi^{-1}(p)$ where $\phi^{-1}$ denotes the quantile function of the standard Gaussian distribution $\mathcal{N}(0,1)$. In this case, for $0<p<p'<1$, we observe that 
\begin{equation}
    F_0^{-1}(p')-F_0^{-1}(p) = \sigma_0 \big(\phi^{-1}(p')-\phi^{-1}(p)\big),
\end{equation}
and therefore
\begin{equation}
    \begin{aligned}
        (F_1^{-1}-F_0^{-1})(p') - (F_1^{-1}-F_0^{-1})(p) &= \big(F_1^{-1}(p')-F_1^{-1}(p)\big)-\big(F_0^{-1}(p')-F_0^{-1}(p)\big) \\
         &= (\sigma_1-\sigma_0)\big(\phi^{-1}(p')-\phi^{-1}(p)\big).
    \end{aligned}
\end{equation}
Since $\phi^{-1}$ is non-decreasing, $F_1^{-1}-F_0^{-1}$ is non-decreasing if and only if $\sigma_0\le \sigma_1$. Thus, by \Cref{prop:1d_geodesic_rays}, $\sigma_0\le\sigma_1$ is a sufficient condition to define a geodesic ray starting from $\mu_0$ and passing through $\mu_1$. We note that if $m_0=m_1$, this condition is equivalent with saying that $\mu_0$ is smaller than $\mu_1$ in the convex order \citep{muller2001stochastic}, noted $\mu_0 \preceq_{\mathrm{cx}} \mu_1$, and which means that for any convex function $f$,
\begin{equation}
    \int f \ \mathrm{d}\mu_0 \le \int f\ \mathrm{d}\mu_1.
\end{equation}
In practice, we are often interested in unit-speed geodesic rays. Thus, we need to have the additional condition $W_2^2(\mu_0,\mu_1) = (m_0-m_1)^2 + (\sigma_1-\sigma_0)^2 = 1$. 

We can also recover the result using \Cref{prop:geodesic_rays}. Indeed, the Monge map between $\mu_0$ and $\mu_1$ is 
\begin{equation}
    \forall x\in \mathbb{R},\ T(x) = \frac{\sigma_1}{\sigma_0}(x-m_0) + m_1 = \nabla u(x),
\end{equation}
where $u(x) = \frac{\sigma_1}{2\sigma_0}x^2 + (m_1-\frac{\sigma_1}{\sigma_0} m_0)x$. Denote $g(x) = u(x)-\frac{x^2}{2}$, then $u$ is 1-convex if and only if $g''(x) \ge 0$, \emph{i.e.}
\begin{equation}
    g''(x) = \frac{\sigma_1}{\sigma_0} -1 \ge 0 \iff \sigma_1 \ge \sigma_0.
\end{equation}

\paragraph{Empirical 1D Distributions.}

Let us take two finite distributions with the same number of particles $n$ and uniform weights: $\mu_0=\frac{1}{n}\sum_{i=1}^n \delta_{x_i}\in\mathcal{P}_2(\mathbb{R})$ and $\mu_1=\frac{1}{n}\sum_{i=1}^n \delta_{y_i}\in\mathcal{P}_2(\mathbb{R})$. We assume that $x_1<\dots<x_n$ and $y_1<\dots<y_n$. Then, $F_1^{-1}-F_0^{-1}$ is non-decreasing if and only if for all $j>i$,
\begin{equation}
    F_1^{-1}\left(\frac{i}{n}\right)-F_1^{-1}\left(\frac{j}{n}\right) = y_i-y_j \le x_i-x_j = F_0^{-1}\left(\frac{i}{n}\right)-F_0^{-1}\left(\frac{j}{n}\right).
\end{equation}
This result also coincides with the condition to have equality between the weak Optimal Transport and the Wasserstein distance in the discrete one dimensional case \citep[Theorem 2.22]{shu2020hopf}.

\paragraph{Starting from a Dirac.} Let $\mu_0=\delta_{x_0}$ with $x_0\in\mathbb{R}$ and $\mu_1\in\mathcal{P}_2(\mathbb{R})$ an arbitrary distribution. Then, since $F_0^{-1}(u) = x_0$ for any $u>0$ and is thus constant, necessarily, $F_1^{-1}-F_0^{-1}$ is non-decreasing and by \Cref{prop:1d_geodesic_rays}, the geodesic between $\mu_0$ and $\mu_1$ is a geodesic ray. This was first observed by \citet[Proposition 3.2]{kloeckner2010geometric}.

\paragraph{Gaussians.}

Let $\mu_0=\mathcal{N}(m_0,\Sigma_0)$ and $\mu_1=\mathcal{N}(m_1,\Sigma_1)$ with $m_0,m_1\in\mathbb{R}^d$ and $\Sigma_0,\Sigma_1$ symmetric positive definite matrices. The Monge map between $\mu_0$ and $\mu_1$ is \citep[Remark 2.31]{peyre2019computational} 
\begin{equation}
    \forall x\in \mathbb{R}^d,\ T(x) = A(x-m_0) + m_1,
\end{equation}
where $A=\Sigma_0^{-\frac12}\big(\Sigma_0^{\frac12}\Sigma_1\Sigma_0^{\frac12}\big)^\frac12 \Sigma_0^{-\frac12}$. Let $u:x\mapsto \frac12 \langle Ax,x\rangle + \langle m_1-Am_0,x\rangle = \frac12 \|A^{\frac12}x\|_2^2 + \langle m_1-Am_0,x\rangle$. Note that we have $\nabla u = T$. Let us denote $g:x\mapsto u(x) - \frac{\|x\|_2^2}{2}$. Then, $u$ is 1-convex if and only if $\nabla^2 g \succeq 0$ (with $\succeq$ the partial order, also called the Loewner order), \emph{i.e.}
\begin{equation} \label{eq:geodesic_rays_gaussians}
    \begin{aligned}
        \nabla^2 g(x) = A-I_d \succeq 0 &\iff A \succeq I_d \\
        &\iff \big(\Sigma_0^{\frac12}\Sigma_1\Sigma_0^{\frac12}\big)^{\frac12} \succeq \Sigma_0. \\
    \end{aligned}
\end{equation}
When $\Sigma_0$ and $\Sigma_1$ commute, the condition simplifies to $\Sigma_1^\frac12 \succeq \Sigma_0^\frac12$. In the general case, by Furata inequality \citep[Theorem 1.3 in the particular case $p=q=r=2$]{fujii2010furuta}, we have that $\Sigma_1^\frac12\succeq \Sigma_0^\frac12$ implies $(\Sigma_0^\frac12 \Sigma_1 \Sigma_0^\frac12)^\frac12 \succeq \Sigma_0$ but it is not an equivalence.

Furthermore, for completeness, we recall that the geodesic between the Gaussian distributions $\mu_0$ and $\mu_1$ is of the form $t\mapsto \mathcal{N}(m_t,\Sigma_t)$ \citep{altschuler2021averaging} where 
\begin{equation}
    \begin{cases}
        m_t = (1-t)m_0+t m_1 \\
        \Sigma_t = \big((1-t)I_d+t A\big)\Sigma_0 \big((1-t)I_d + t A\big).
    \end{cases}
\end{equation}

\paragraph{More general case.}

In general, using \Cref{prop:geodesic_rays}, we can study whether or not a geodesic is a ray by studying the 1-convexity of the Brenier potential associated to the Monge map. However, we note that such a map between two distributions does not always exist. \citet{paty2020regularity} proposed to enforce this property by finding the best possible 1-convex map by solving $f^*\in\argmin_{f \text{ 1-convex}} W_2(\nabla f_\#\mu,\nu)$, called the nearest Brenier potential, which could be used to define nearest geodesic rays. For arbitrary measures, no characterization is yet available to the best of our knowledge.

\section{Busemann Function} \label{section:busemann}

\subsection{Background on Busemann Functions}

On any geodesic metric space $(X,d)$ which has geodesic rays, the Busemann function  associated to a unit-speed geodesic ray $\gamma$ can be defined as \citep[II.8.17]{bridson2013metric}
\begin{equation}
    \forall x\in X,\ B^\gamma(x) = \lim_{t\to\infty}\ \big(d(\gamma(t), x) - t\big).
\end{equation}
This function has attracted interest on Riemannian manifolds as it provides a natural generalization of hyperplanes. Indeed, on Euclidean spaces, geodesic rays are of the form $\gamma(t) = t\theta$ for $\theta\in S^{d-1}$, and thus we can show that
\begin{equation}
    \forall x\in \mathbb{R}^d,\ B^\gamma(x)=-\langle x,\theta\rangle.
\end{equation}
Thus, it has recently received attentions on Hyperbolic spaces in order to perform Horospherical Principal Component Analysis \citep{chami2021horopca}, but also to characterize directions and perform classifications with prototypes \citep{ghadimi2021hyperbolic, durrant2023hmsn} or to define decision boundaries for classification \citep{fan2023horocycle}. It can also be used as a projection on a geodesic as we described in \Cref{chapter:sw_hadamard} and experimented in \Cref{chapter:hsw} and \Cref{chapter:spdsw}.

Therefore, to deal with data represented as probability distributions, it is interesting to investigate the Busemann function on the Wasserstein space. As it is not a geodesically complete space, it is well defined only on geodesic rays. Fortunately, as shown in \citep{zhu2021busemann}, there is always a geodesic ray starting from some distribution $\mu_0$. Furthermore, as developed in the previous section, we know how to characterize them in some situations. Thus, in the next section, by leveraging closed-forms of the Wasserstein distance, we study closed-forms for the Busemann function.

\subsection{Busemann Functions in Wasserstein Space}

\looseness=-1 Let $(\mu_t)_{t\ge 0}$ be a unit-speed geodesic ray. Then, the Busemann function associated with $(\mu_t)_{t\ge 0}$ is naturally defined as 
\begin{equation}
    \forall \nu\in\mathcal{P}_2(\mathbb{R}^d),\ B^\mu(\nu) = \lim_{t\to\infty}\ \big(W_2(\mu_t, \nu) - t\big).
\end{equation}
This was first studied by \citet{zhu2021busemann} from a theoretical point of view, and in particular, they showed that the limit does exist. But no closed-form was proposed. Thus, we provide here some closed-forms in two particular cases: one dimensional probability distributions and Gaussian measures (and more generally elliptical distributions). Indeed, deriving a closed-form for the Busemann function heavily relies on closed-forms of the Wasserstein distance, and thus we restrict the analysis for now to these cases.

\paragraph{One dimensional case.}

\looseness=-1 On the real line, we have several appealing properties. In particular, we recall that in this case, the Wasserstein distance between $\mu,\nu\in\mathcal{P}_2(\mathbb{R})$ can be computed in closed-form \citep[Remark 2.30]{peyre2019computational} as 
\begin{equation}
    W_2^2(\mu,\nu) = \int_0^1 | F_\mu^{-1}(u)-F_\nu^{-1}(u)|^2 \ \mathrm{d}u,
\end{equation}
where $F_\mu^{-1}$ and $F_\nu^{-1}$ are the quantile functions of $\mu$ and $\nu$.

\begin{proposition}[Closed-from for the Busemann function on $\mathcal{P}_2(\mathbb{R})$] \label{prop:busemann_closed_1d}
    Let $(\mu_t)_{t\ge 0}$ be a unit-speed geodesic ray in $\mathcal{P}_2(\mathbb{R})$, then
    \begin{equation}
        \begin{aligned}
            \forall \nu \in \mathcal{P}_2(\mathbb{R}),\ B^\mu(\nu) &= -\int_0^1 \big(F_{\mu_1}^{-1}(u)-F_{\mu_0}^{-1}(u)\big) \big(F_\nu^{-1}(u)-F_{\mu_0}^{-1}(u)\big)\ \mathrm{d}u \\
            &= -\langle F_{\mu_1}^{-1} - F_{\mu_0}^{-1}, F_\nu^{-1}-F_{\mu_0}^{-1}\rangle_{L^2([0,1])}.
        \end{aligned}
    \end{equation}
\end{proposition}

\begin{proof}
    See \Cref{proof:prop_busemann_closed_1d}.
\end{proof}

We observe that it corresponds up to a sign to the $L^2([0,1])$ inner product between $F_{\mu_1}^{-1}-F_{\mu_0}^{-1}$ and $F_{\nu}^{-1} - F_{\mu_0}^{-1}$, which are the quantiles centered around $F_{\mu_0}^{-1}$. This comes from the Hilbert properties of the one dimensional Wasserstein space.

For one dimensional Gaussians $\mu_0 = \mathcal{N}(\mo, \so^2)$ and $\mu_1=\mathcal{N}(\m1, \s1^2)$ such that $\sigma_1\ge \sigma_0$ and $W_2^2(\mu_0,\mu_1)=1$, using that $F_{\nu}^{-1}(u) = m + \sigma \phi^{-1}(u)$ for any $\nu=\mathcal{N}(m,\sigma^2)$, $\int_0^1 \phi^{-1}(u)\ \mathrm{d}u=0$ and $\int_0^1 \phi^{-1}(u)^2\ \mathrm{d}u = 1$, we obtain for any $\nu=\mathcal{N}(m,\sigma^2)$,
\begin{equation} \label{eq:1d_busemann_gaussian}
    B^\mu(\nu) = - (\m1-\mo)(m-\mo) - (\s1-\so)(\sigma-\so) = - \left\langle \begin{pmatrix}
        \m1-\mo \\ \s1-\so
    \end{pmatrix}, \begin{pmatrix}
        m-\mo \\ \sigma-\so
    \end{pmatrix} \right\rangle.
\end{equation}

\paragraph{Bures-Wasserstein case.}

When restricting the space of probability measures to Gaussians with positive definite covariance matrices and endowing it with the (Bures-)Wasserstein distance, we obtain a proper Riemannian manifold \citep{bhatia2019bures}. Moreover, we know in closed-form the Wasserstein distance in this case (see \Cref{prop:closed_forms_gaussians}) as well as the form of the geodesics. Thus, we can compute the closed-form of the Busemann function.

\begin{proposition}[Closed-form for the Busemann function on $BW(\mathbb{R}^d)$] \label{prop:closed_form_general_gaussian}
    Let $(\mu_t)_{t\ge 0}$ be a unit-speed geodesic ray characterized by $\mu_0 = \mathcal{N}(\mo, \So)$ and $\mu_1 = \mathcal{N}(\m1, \S1)$ (\emph{i.e.} such that $(\So^{\frac12}\S1\So^{\frac12})^\frac12 \succeq \So$ by \eqref{eq:geodesic_rays_gaussians}, and $W_2^2(\mu_0,\mu_1)=1$). Then, for any $\nu=\mathcal{N}(m,\Sigma)$, 
    \begin{equation} \label{eq:busemann_gaussians}
        B^\mu(\nu) = -\langle \m1-\mo, m-\mo\rangle + \tr\big(\So(A-I_d) \big) - \tr\big((\Sigma^\frac12(\So-\So A - A\So + \S1)\Sigma^\frac12)^\frac12\big),
    \end{equation}
    where $A=\So^{-\frac12}(\So^\frac12\S1\So^\frac12)^\frac12\So^{-\frac12}$.
\end{proposition}

\begin{proof}
    See \Cref{proof:prop_closed_form_general_gaussian}.
\end{proof}

When all the covariance matrices commute, \emph{e.g.} if they are all chosen as diagonal, \eqref{eq:busemann_gaussians} simplifies as
\begin{equation}
    \begin{aligned}
        B^\mu(\nu) &= -\langle\m1-\mo,m-\mo\rangle - \tr\big((\S1^\frac12-\So^\frac12)(\Sigma^\frac12-\So^\frac12)\big) \\
        &= -\langle\m1-\mo,m-\mo\rangle - \langle \S1^\frac12-\So^\frac12, \Sigma^\frac12-\So^\frac12\rangle_F.
    \end{aligned}
\end{equation}
For commuting matrices, this is just the inner product in the product space $\mathbb{R}^d \times S_d(\mathbb{R})$. Moreover, we recover \eqref{eq:1d_busemann_gaussian} in one dimension.

\looseness=-1 We note that these results could be extended to elliptical distributions as we also have the closed-form for the Wasserstein distance in this case \citep{gelbrich1990formula, muzellec2018generalizing}. Finding closed-forms in a more general setting or for other distributions is for now an open direction of research as our results heavily rely on closed-forms of the Wasserstein distance and of the geodesics, which are often not available.

\section{Applications to PCA} \label{section:pca}



Principal Component Analysis (PCA) is a classical statistical method used to capture the main modes of variation of datasets in order \emph{e.g.} to reduce their dimensionality. As the space of probability distributions is not an Euclidean space, extending PCA to such space is not straightforward, as it requires defining principal components and projections.
In this section, we aim at using the Busemann function in order to perform PCA on Wasserstein space. First, we describe a framework which allows us to use the Busemann function on this space in order to use PCA on probability distributions. Then, we perform an empirical analysis on one dimensional distributions, first on datasets of 1D Gaussians distributions for which we provide a closed-form, and then on one dimensional histograms.

\subsection{Busemann Wasserstein PCA}

There are two popular formulations of PCA \citep[Section 12.1]{bishop2006pattern}. The first aims at minimizing the reconstruction error while the second aims at maximizing the variance of the projected data onto the directions in order to choose the direction explaining most of the original data. In the following, we focus on the latter. Assuming the data $x_1,\dots,x_n\in\mathbb{R}^d$ are centered, the Euclidean PCA problem to be solved is 
\begin{equation}
    \forall i\ge 1,\ \theta_i\in\argmax_{\theta\in S^{d-1}\cap \mathrm{span}(\theta_1,\dots,\theta_{i-1})^\bot}\ \frac{1}{n}\sum_{k=1}^n \langle \theta, x_k\rangle^2.
\end{equation}
More generally, without assuming that the data are centered and noting $\Bar{x}$ the barycenter of the data, the problem can be written as
\begin{equation}
    \forall i\ge 1,\ \theta_i \in \argmax_{\theta\in \Bar{x}+S^{d-1}\cap \mathrm{span}(\theta_1,\dots,\theta_{i-1})^\bot}\ \mathrm{Var}\big((\langle \theta, x_k\rangle)_k\big) = \mathrm{Var}\left(\big(B^\theta(x_k)\big)_k\right).
\end{equation}

We propose to extend this formulation to the Wasserstein space using geodesic rays for the directions and the right Busemann function as a way to get coordinates on geodesic rays. For the concept of orthogonality, we follow \citep{seguy2015principal} and use orthogonality of vector fields. Indeed, assuming that a Monge map $T$ starting from $\mu_0$ exists, we know that a geodesic in Wasserstein space is of the form $\mu_t = \big((1-t)\id + t T\big)_\#\mu_0 = (\id + t v)_\#\mu_0$ where $v=T-\id\in L^2(\mu_0)$ lies in the tangent space at $\mu_0$. Thus, fixing an origin distribution $\mu_0$, and noting $\nu_1,\dots,\nu_n\in\mathcal{P}_2(\mathbb{R}^d)$ the dataset, we aim at solving with respect to $\mu_1$, as a geodesic is fully characterized by two distributions on its path, the following problem:
\begin{equation}
    \forall i\ge 1,\ \mu^{(i)}_1 \in \argmax_{\mu_1}\ \mathrm{Var}\left(\big(B^\mu(\nu_k)\big)_k\right) \quad \text{such that} \quad \begin{cases}
        W_2^2(\mu_0, \mu_1) = 1 \\
        t\mapsto \mu_t \text{ is a geodesic ray} \\
        v \in \mathrm{span}\big((v_j)_{1\le j\le i-1}\big)^\bot,
    \end{cases}
\end{equation}
where for each $i$, $v_i=T_i-\id$ with $T_i$ the Monge map between $\mu_0$ and $\mu_1^{(i)}$. The first two constraints impose $t\mapsto \mu_t$ to be a unit-speed geodesic ray while the third constraint imposes the orthogonality of the geodesic rays.
In the following, we will specify this problem in the case where all distributions are one dimensional Gaussian, and in the more general case where we deal with arbitrary one dimensional distributions. 

To project an arbitrary distribution $\nu$ onto a principal direction $t\mapsto \mu_t$, we need to find the coordinate $t$ such that $B^\mu(\mu_t) = B^\mu(\nu)$. Denoting by $\gamma^*$ an optimal coupling between $\mu_0$ and $\mu_1$, and as $B^\mu(\mu_t) = -t$, the projection is given by $P^\mu(\nu) = \mu_{-B^\mu(\nu)} = \big((1+B^\mu(\nu)) \pi^1 - B^\mu(\nu) \pi^2\big)_\#\gamma^*$. However, note that as the Wasserstein space is not geodesically complete, some distributions for which $B^\mu(\nu)>0$ may be projected out of the geodesic, and hence need to be dealt with carefully in practice. In the Gaussian one dimensional case, we will investigate this issue in more details (see \Cref{lemma:extension_1dgaussian_ray}).

\paragraph{Related works.}

\looseness=-1 Extending PCA to other spaces has received a lot of attention over the years as there are several possible generalizations. \citet{fletcher2004principal} first proposed to generalize PCA on Riemannian manifolds using Principal Geodesic Analysis by projecting on subspaces using a geodesic projection. \citet{huckemann2006principal, huckemann2010intrinsic} proposed a variant named Geodesic PCA by choosing principal geodesics orthogonally. \citet{pennec2018barycentric} instead proposed to project on barycentric subspaces. Then, some works focused on developing efficient PCA methods adapted to specific Riemannian manifolds such as the space of SPDs \citep{horev2016geometry} or Hyperbolic spaces \citep{chami2021horopca}. In particular, \citet{chami2021horopca} proposed to project on geodesics submanifolds using the horospherical projection.

As the Wasserstein space possesses a weak Riemannian structure, PCA has been naturally extended to this space. We can split the different methods into two types, extrinsic and intrinsic ones. Intrinsic methods exploit the geodesic structure of the Wasserstein space and include for example the Geodesic PCA introduced by \citet{bigot2017geodesic} on one dimensional distributions or the method of \citep{seguy2015principal} which extends to $\mathcal{P}_2(\mathbb{R}^d)$. Extrinsic methods rather exploit the linear structure of the tangent space on which the distribution are projected with the log map \citep{cazelles2018geodesic, pegoraro2022projected}. More recently, \citet{pont2022principal} adapted the framework to persistence diagrams endowed with the Wasserstein distance while \citet{beraha2023wasserstein} studied it for circular measures. We can also cite \citep{masarotto2022transportation} in which the focus is on the Bures-Wasserstein space, and \citep{niculae2023two} in which a method was proposed for distributions characterized by their first two moments.

\subsection{One Dimensional Gaussians}

Let $m_0,\sigma_0\in\mathbb{R}$ and $\mu_0 = \mathcal{N}(m_0, \sigma_0^2)$ be the origin distribution from which the geodesic rays will start. Typically, $\mu_0$ will be chosen as the barycenter of the data. And let $\nu_1=\mathcal{N}(m_1, \sigma_1^2), \dots, \nu_n=\mathcal{N}(m_n, \sigma_n^2)$ the dataset of Gaussian distributions. We note that the barycenter of $(\nu_i)_{k=1}^n$ is simply $\mathcal{N}(\Bar{m}, \Bar{\sigma}^2)$ where $\Bar{m}=\frac{1}{n}\sum_{k=1}^n m_k$ and $\Bar{\sigma} = \frac{1}{n}\sum_{k=1}^n \sigma_k$.

Let $\mu_1^{(1)} = \mathcal{N}(m_{(1)}, \sigma_{(1)}^2)$ be a suitable distribution for which $t\mapsto \mu_t^{(1)}$ is a unit-speed geodesic ray starting from $\mu_0$ and passing through $\mu_1^{(1)}$ at $t=1$, \emph{i.e.} which satisfies $\sigma_{(1)}\ge \sigma_0$ and $W_2^2(\mu_0,\mu_1) = (m_{(1)}-m_0)^2 + (\sigma_{(1)}-\sigma_0)^2 = 1$. We recall that the Busemann function evaluated at $\nu_k$ for any $k\in\{1,\dots,n\}$ can be obtained as
\begin{equation}
    B^\mu(\nu_k) = -(m_{(1)}-m_0)(m_k-m_0) - (\sigma_{(1)}-\sigma_0)(\sigma_k-\sigma_0) = - \left\langle \begin{pmatrix}
        m_{(1)} - m_0 \\ \sigma_{(1)} - \sigma_0
    \end{pmatrix}, \begin{pmatrix}
        m_k - m_0 \\
        \sigma_k - \sigma_0
    \end{pmatrix} \right\rangle.
\end{equation}
Moreover, if we denote $\mu^{(2)}$ a second geodesic ray starting from $\mu_0$ and passing through $\mu_1^{(2)}=\mathcal{N}(m_{(2)}, \sigma_{(2)}^2)$, and $T_1$ the OT map between $\mu_0$ and $\mu_1^{(1)}$ as well as $T_2$ the OT map between $\mu_0$ and $\mu_1^{(2)}$, then the orthogonality condition can be written as
\begin{equation}
    \begin{aligned}
        \langle T_1 - \id, T_2 - \id\rangle_{L^2(\mu_0)} &= \langle F_{(1)}^{-1} - F_0^{-1}, F_{(2)}^{-1}-F_0^{-1}\rangle_{L^2([0,1])} \\
        &= (\sigma_{(1)}-\sigma_0)(\sigma_{(2)}-\sigma_0) + (m_{(1)}-m_0)(m_{(2)}-m_0) \\
        &= 0,
    \end{aligned}
\end{equation}
using a change of variable and noting $F_{(1)}^{-1}$ and $F_{(2)}^{-1}$ the quantile functions of $\mu_1^{(1)}$ and  $\mu_1^{(2)}$ respectively.

Thus, to sum up, we need to solve the following optimization problem
\begin{equation} \label{eq:pca_1d_gaussians}
    \begin{aligned}
        &\forall i\ge 1,\ (m_{(i)}, \sigma_{(i)}) \in \argmax_{m,\sigma}\ \mathrm{Var}\left(\big((m-m_0)(m_k-m_0) + (\sigma-\sigma_0)(\sigma_k-\sigma_0)\big)_{k=1}^n\right) \\
        &\text{subject to} \quad \begin{cases}
        (m-m_0)^2 + (\sigma-\sigma_0)^2 = 1 \\
        \sigma \ge \sigma_0 \\
        \forall j \le i-1,\ (\sigma-\sigma_0)(\sigma_{(j)}-\sigma_0) + (m-m_0)(m_{(j)}-m_0) = 0.
        \end{cases}
    \end{aligned}
\end{equation}

In the next Proposition, we provide a closed-form formula for the first direction. As one dimensional Gaussians can be embedded into a 2D space $\mathbb{R}\times\mathbb{R}_+^*$ by representing each Gaussian $\mathcal{N}(m,\sigma^2)$ as $(m,\sigma)$ \citep{cho2023hyperbolic}, 
the set of constraint lies on the semi-circle $\{(m,\sigma)\in\mathbb{R}\times\mathbb{R}_+,\ \sigma\ge\sigma_0 \text{ and } (m-m_0)^2 + (\sigma-\sigma_0)^2 = 1\}$. Thus, the second direction is obtained as the only possible orthogonal projection.

\begin{proposition} \label{prop:1dgaussian_pca}
    Let $\mu_0 = \mathcal{N}(m_0, \sigma_0^2)$ and for all $k\in \{1,\dots,n\}$, $\nu_k = \mathcal{N}(m_k, \sigma_k^2)$. Denote for all $k\in\{1,\dots,n\}$, $x_k = \begin{pmatrix}
        m_k-m_0 \\ \sigma_k-\sigma_0
    \end{pmatrix}$ and $M = \frac{1}{n}\sum_{k=1}^n x_k x_k^T - \left(\frac{1}{n}\sum_{k=1}^n x_k \right)\left(\frac{1}{n} \sum_{k=1}^n x_k\right)^T$. Then, the first principal component obtained as the solution of \eqref{eq:pca_1d_gaussians} is given by $\mu^{(1)}_1 = \mathcal{N}(m_{(1)}, \sigma_{(1)}^2)$ where
    \begin{equation}
        \begin{cases}
            m_{(1)} = m_0 + \cos\left(\frac{\theta}{2}\right) \\
            \sigma_{(1)} = \sigma_0 + \sin\left(\frac{\theta}{2}\right),
        \end{cases}
    \end{equation}
    with $\theta = \arccos\left(\frac{M_{11}-M_{22}}{\sqrt{(M_{11}-M_{22})^2 + 4 M_{12}^2}}\right)$. By using the orthogonality condition between $\begin{pmatrix}m_{(1)}-m_0 \\ \sigma_{(1)}-\sigma_0\end{pmatrix}$ and $\begin{pmatrix}m_{(2)}-m_0 \\ \sigma_{(2)}-\sigma_0\end{pmatrix}$, the second component is obtained as $\mu_1^{(2)}=\mathcal{N}(m_{(2)}, \sigma_{(2)}^2)$ where
    \begin{equation}
        \begin{cases}
            m_{(2)} = m_0 + \cos\left(\frac{\theta - \mathrm{sign}(\theta-\pi)\pi}{2}\right) \\
            \sigma_{(2)} = \sigma_0 + \sin\left(\frac{\theta - \mathrm{sign}(\theta-\pi)\pi}{2}\right).
        \end{cases}
    \end{equation}
\end{proposition}

\begin{proof}
    See \Cref{proof:prop_1dgaussian_pca}.
\end{proof}

In the last Proposition, we reported the solutions in closed-form. We note that they could also be obtained as the eigenvectors of the matrix $M$ which is an empirical covariance matrix, as it is, similarly as the Euclidean PCA, an eigenvalue problem with the extra care of the constraint $\sigma-\sigma_0\ge 0$.

Before diving into some numerical applications, let us discuss some particular cases and analyze when the projections on the geodesic ray can be done. First, in the simple case where all distributions of the dataset have the same mean and only vary by their variance, \emph{i.e.} for all $k\ge 1$, $m_k=m_1$, then we notice that $M_{12} = M_{11} = 0$. Thus in this case, we obtain $\theta=\pi$ and $m_{(1)} = m_0$, $\sigma_{(1)}=\sigma_0 + 1$. Only the variance is captured by the first component. In the opposite case where all the distributions have the same variance, \emph{i.e.} for all $k\ge 1$, $\sigma_k=\sigma_1$, then we have $M_{12}=M_{22}=0$ and thus $\theta=0$, $m_{(1)} = m_0+1$, $\sigma_{(1)}=\sigma_0$. Only the mean is captured by the first component. This is the intuitive behavior that we would expect. For the projections, we show in the next Proposition that we can extend 1D Gaussian geodesic rays for $t<0$.

\begin{proposition} \label{lemma:extension_1dgaussian_ray}
    Let $\mu_0=\mathcal{N}(m_0,\sigma_0^2)$ and $\mu_1=\mathcal{N}(m_1, \sigma_1^2)$ two Gaussian defining a unit-speed geodesic ray starting from $\mu_0$ and passing through $\mu_1$ at $t=1$, \emph{i.e.} satisfying $\sigma_1 \ge \sigma_0$ and $(m_1-m_0)^2 + (\sigma_1-\sigma_0)^2 = 1$. Then, the underlying geodesic ray $t\mapsto \mu_t$ is well defined on $[-\frac{\sigma_0}{\sigma_1-\sigma_0}, +\infty[$.
\end{proposition}

\begin{proof}
    See \Cref{proof:lemma_extension_1dgaussian_ray}.
\end{proof}

This Proposition gives us a way to be sure that a Gaussian can be projected on the geodesic ray. For $\nu=\mathcal{N}(m,\sigma^2)$, if $B^\mu(\nu)>\frac{\sigma_0}{\sigma_1-\sigma_0}$, then $\nu$ will possibly not be projected on the geodesic. We note the two limiting cases: $\sigma_0=\sigma_1$ for which the geodesic ray is actually a line and can be extended to $\mathbb{R}$ which we recover here as $-\frac{\sigma_0}{\sigma_1-\sigma_0}\xrightarrow[\sigma_1 \to \sigma_0^+]{}-\infty$, and $\sigma_1 = 1+\sigma_0$ for which the ray can be extended to $[-\sigma_0, +\infty[$ and corresponds to a dilation. However, in this case, since $\sigma_1=1+\sigma_0$ and $m_1=m_0$, we note that any distribution can be projected on the geodesic since, for any $\nu=\mathcal{N}(m,\sigma^2)$, 
\begin{equation}
    \begin{aligned}
        B^\mu(\nu) &= -(m-m_0)(m_1-m_0) - (\sigma-\sigma_0)(\sigma_1-\sigma_0) = -(\sigma-\sigma_0),        
    \end{aligned}
\end{equation}
and thus the projection coordinate is $-B^\mu(\nu) = \sigma-\sigma_0 < -\sigma_0 \iff \sigma < 0$, which is not possible.

\paragraph{Numerical Examples.}

\begin{figure}[t]
    \centering
    \includegraphics[width=\linewidth]{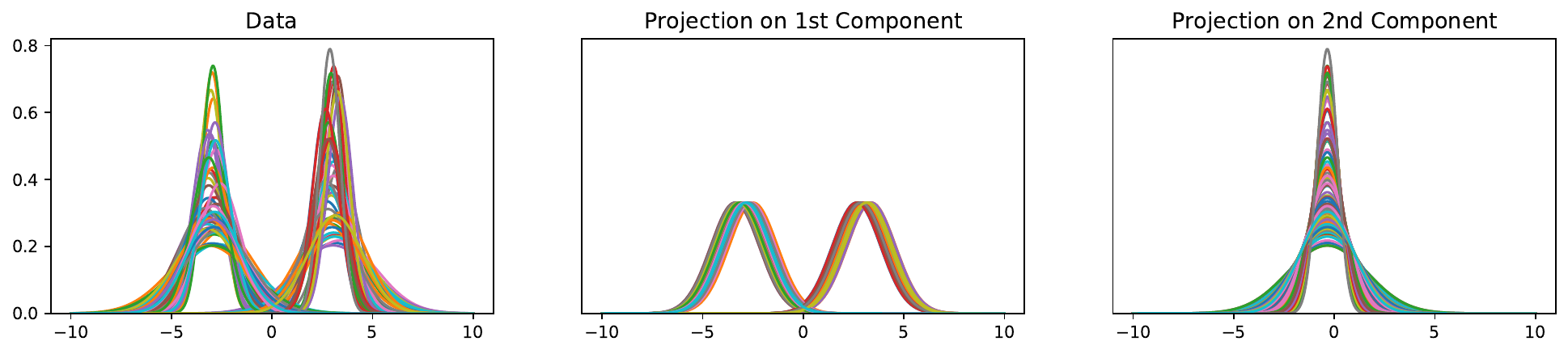}
    \caption{Projections for the datasets of clustered Gaussians.}
    \label{fig:proj_cluster_gaussians}
\end{figure}


\begin{figure}[t]
    \centering
    \includegraphics[width=\linewidth]{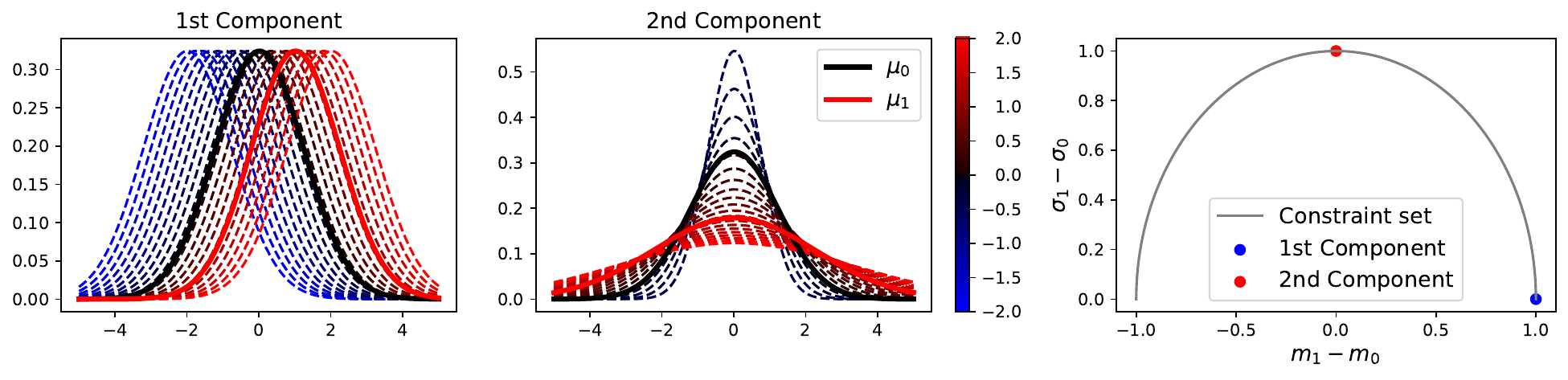}
    \caption{First and second principal components for the datasets of clustered Gaussians. (\textbf{Left}) 1st principal component for $t\in [-2,2]$. (\textbf{Center}) 2nd principal component for $t\in [-0.5,2]$ (for visibility). We plot in dashed lines the pdf of 20 evenly spaced measures $\mathcal{N}(m_t, \sigma_t^2)$ of the geodesic rays. The colors (from blue to red with black in the middle) encode the progression along the geodesic.}
    \label{fig:components_cluster_gaussians}
\end{figure}


As an illustration to assess the interpretability of the principal components found, we use the first simulation setting of \citep[Section 7.1]{pegoraro2022projected}. In this setting, we fix $n=100$ and generate the data randomly for all $k\ge 1$ as
\begin{equation}
    \begin{cases}
        m_k \sim \frac12 \mathcal{N}(0.3, 0.2^2) + \frac12 \mathcal{N}(-0.3, 0.2^2) \\
        \sigma_k \sim \mathrm{Unif}([0.5, 2]).
    \end{cases}
\end{equation}
\looseness=-1 We plot the densities of the data simulated on \Cref{fig:proj_cluster_gaussians}. As the major variability is on the mean of the data, we expect the first component to capture the change in the shift and the second component to capture the change in variance. We start from $\mu_0$ chosen as the barycenter and plot on \Cref{fig:components_cluster_gaussians} the principal components for $t\in [-2,2]$ for the first component and for $t\in [-0.5,2]$ for the second one for the sake of visibility as the variance quickly vanishes towards 0 when $t<-0.5$. 
We also plot the projections on the two components on \Cref{fig:proj_cluster_gaussians}. We did not observe any misspecified projection for the data which is justified by \Cref{lemma:extension_1dgaussian_ray}. Overall, the results are on par with what we expect and with previous PCA methods such as \citep{pegoraro2022projected}. 

\paragraph{With data through samples.} In the case where the data are in form of samples or histograms, and in which we want to find Gaussian geodesic rays as principal components, we cannot solve the problem in closed-form. Nonetheless, we can solve it numerically by parameterizing $\sigma$ as $\sigma=\sigma_0 + e^s$ in order to ensure $\sigma\ge \sigma_0$ and performing a projected gradient descent over $(m,s)$ in order to find the first component. Then, the second component can be found using the orthogonality condition. In this case, noting $F_k^{-1}$ the quantile functions of the data distributions $\nu_k$, the closed-form for the Busemann function can be computed as 
\begin{equation}
    \begin{aligned}
        B^\mu(\nu_k) &= - (m_1-m_0)\left(\int_0^1 F_k^{-1}(u)\ \mathrm{d}u - m_0\right) - (\sigma_1-\sigma_0)\left(\int_0^1 \phi^{-1}(u) F_k^{-1}(u)\ \mathrm{d}u - \sigma_0 \right) \\
        &= - (m_1-m_0)\left(m(\nu_k) - m_0\right) - (\sigma_1-\sigma_0)\left(\langle \phi^{-1}, F_k^{-1}\rangle_{L^2([0,1])} - \sigma_0 \right).
    \end{aligned}
\end{equation}
This formulation can be useful when we are only interested in the two first moments, but where each distribution is not necessarily Gaussian, and thus it would not be necessarily justified to approximate it as a Gaussian.

\subsection{One Dimensional Histograms}

In this section, we propose to deal with the general one dimensional case, without assuming any form for the geodesic rays or for the data. Thus, it would allow to handle any one dimensional histogram of real data. 

In this situation, denoting by $F_k^{-1}$ the quantile of $\nu_k\in\mathcal{P}_2(\mathbb{R})$, we want to solve
\begin{equation}
    \forall i \ge 1,\ F_{(i)}^{-1} \in \argmax_{F_{\mu_1}^{-1}}\ \mathrm{Var}\left(\big(B^\mu(\nu_k)\big)_{k=1}^n\right) \quad \text{subject to} \quad \begin{cases}
        W_2^2(\mu_0,\mu_1) = \|F_{\mu_1}^{-1} - F_{\mu_0}^{-1}\|_{L^2([0,1])}^2 = 1 \\
        F_{\mu_1}^{-1} - F_{\mu_0}^{-1} \text{ non-decreasing} \\
        \forall j < i,\ \langle F_{\mu_1}^{-1} - F_{\mu_0}^{-1}, F_{(j)}^{-1} - F_{\mu_0}^{-1}\rangle_{L^2([0,1])} = 0.
    \end{cases}
\end{equation}
To learn geodesic rays, a first solution could be to learn the quantiles by approximating them using \emph{e.g.} splines such as in \citep{pegoraro2022projected} at the expanse of solving a concave quadratic problem on the sphere, or monotone parametric transformations such as sum-of-squares polynomial flows \citep{jaini2019sum}, unconstrained monotonic neural networks \citep{wehenkel2019unconstrained} or monotone flows \citep{ahn2022invertible}. 

Instead, we propose to find $\mu_1$ by learning directly the Monge map and leveraging \Cref{prop:geodesic_rays} by modeling the map as the gradient of a 1-convex function and hence implicitely learning a geodesic ray. Modeling such functions with neural networks has recently received much attention and has been applied \emph{e.g.} to define Normalizing Flows \citep{huang2021convex} or to approximate the Monge map \citep{makkuva2020optimal, mokrov2021largescale, alvarez-melis2022optimizing, bunne2021jkonet}. Early works computed the gradient of Input Convex Neural Networks \citep{amos2017input}, but it has been observed that they have poor expressiveness (see \emph{e.g.} \citep{korotin2021wasserstein, korotin2021neural} or \Cref{chapter:swgf}). Hence, it has been recently advocated to rather model the gradient of a convex function directly with a neural network \citep{saremi2019approximating, richterpowell2021input, chaudhari2023learning}. Thus, we model directly the Monge map between $\mu_0$ and $\mu_1$ using a Cascaded Monotone Gradient Network (CMGN) introduced by \citet{chaudhari2023learning} and which is a neural network with positive semi-definite Jacobian and hence the gradient of a convex function. To ensure that it is the gradient of a 1-convex function, we add the identity to the output. In that case, the optimization problem we want to solve is
\begin{equation}
    \forall i,\ T_i \in \argmax_{T=\nabla u, \text{ u 1-convex}}\ \mathrm{Var}\left(\big(B^\mu(\nu_k)\big)_{k=1}^n\right) \quad \text{subject to} \quad \begin{cases}
        W_2^2(\mu_0, T_\#\mu_0) = \int \big(x-T(x)\big)^2\ \mathrm{d}\mu_0(x) = 1 \\
        \forall j<i,\ \int (T(x)-x)(T_j(x)-x)\ \mathrm{d}\mu_0(x) = 0. 
    \end{cases}
\end{equation}
With such modelization, the 1-convexity is enforced into the architecture of the networks, and hence the optimization is done over geodesic rays. Nonetheless, the unit-speed constraint and orthogonal constraints need to be relaxed through the Lagrangian to be incorporated into the loss, and might be tricky to optimize.
In practice, noting $T_\theta$ a 1-convex CMGN, and $\alpha$ and $(\lambda_j)_j$ Lagrange multipliers, we minimize the following loss for the i-th component:
\begin{equation}
    \begin{aligned}
        \mathcal{L}(\theta) &= -\mathrm{Var}\left(\big(B^\mu(\nu_k)\big)_{k=1}^n\right) + \alpha \left(1-\int \big(x-T_\theta(x)\big)^2\ \mathrm{d}\mu_0(x)\right)^2 \\ &+ \sum_{j=1}^{i-1} \lambda_j \left(\int (T_\theta(x)-x)(T_j(x)-x)\ \mathrm{d}\mu_0(x)\right)^2.
    \end{aligned}
\end{equation}

\begin{figure}[t]
    \centering
    \includegraphics[width=\linewidth]{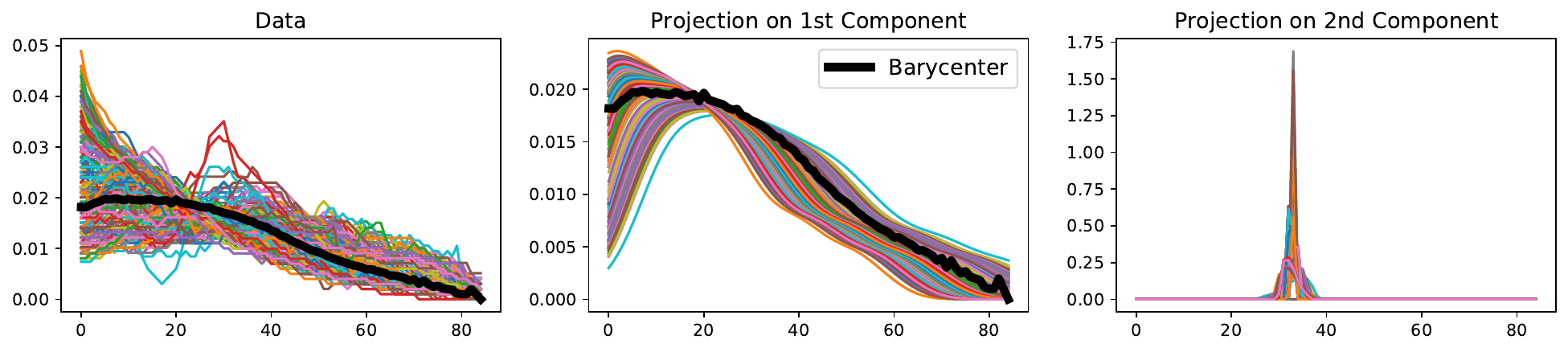}
    \caption{Projections for the dataset of population pyramid.}
    \label{fig:proj_pyramid_age}
\end{figure}

\begin{figure}[t]
    \centering
    \includegraphics[width=0.66\linewidth]{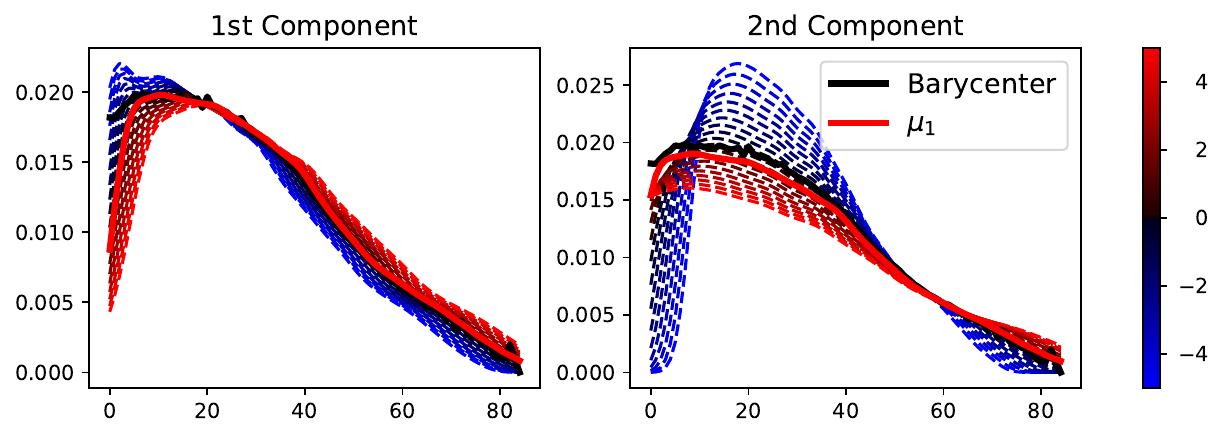}
    \caption{First and second principal components for the dataset of population pyramid interpolated for $t\in [-5,5]$.}
    \label{fig:components_pyramid_age}
\end{figure}


\paragraph{Population Pyramid.} 

We follow \citep{cazelles2018geodesic} and consider as real dataset the population pyramids of $n=217$ countries in the year 2000. Each histogram of the dataset represents the normalized frequency by age of people living in the country. Each bin represents one year and all peoples aged of more than 85 years belong to the last interval.

\begin{figure}[t]
    \centering
    \includegraphics[width=\linewidth]{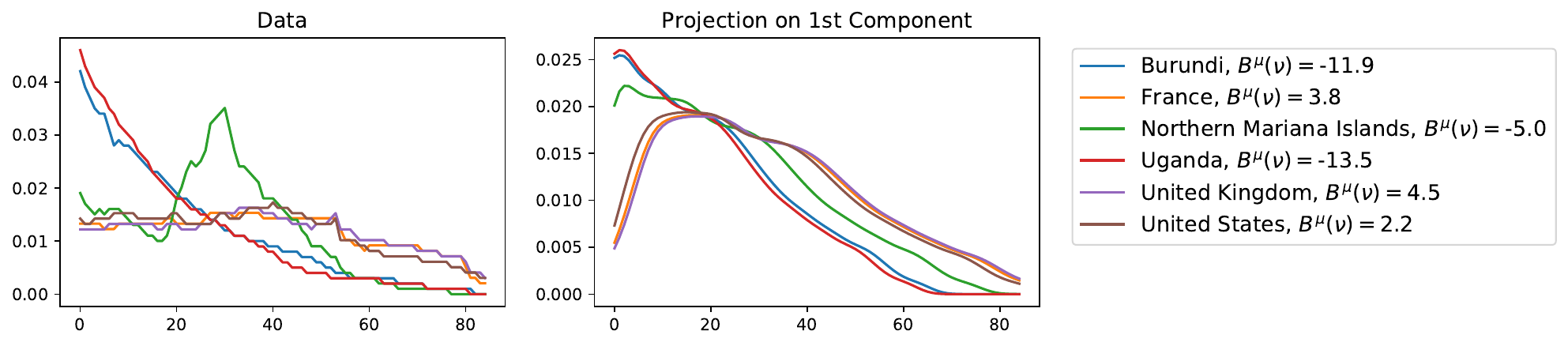}
    \caption{Projections on the first component for the dataset of population pyramid with the value of the Busemann function for selected countries. Countries for which the population is mostly young (such as Uganda or Burundi) have a low Busemann coordinate while more developed countries (such as France, UK, US) have a bigger one. Countries with a population in the middle such as the Northern Mariana Islands are projected around the origin.}
    \label{fig:proj_pyramid_age_selected}
\end{figure}

As origin measure, we choose the barycenter of the dataset which can be found as $F_{\Bar{\nu}}^{-1} = \frac{1}{n}\sum_{k=1}^n F_{\nu_k}^{-1}$. Then, we pass in the neural network the support of the barycenter histogram to obtain $\mu_1=(T_\theta)_\#\Bar{\nu}$. On \Cref{fig:proj_pyramid_age}, we plotted the histograms of the data of each country along their projection on the first and second components obtained. On \Cref{fig:components_pyramid_age}, we plotted the two first components interpolated for $t\in [-5,5]$. 
We observe that the projections on the first component capture the difference between less developed countries, whose population is mostly young and more developed countries. We report on \Cref{fig:proj_pyramid_age_selected} the results for some chosen countries which show clearly the differences between the projections obtained for developed and less developed countries. 

However, the second component does not seem to capture any additional useful information. We observe that the values of the projections are fairly low (around -19), and the projections do not seem to necessary lie on the geodesics ray, as we observe that the Busemann function evaluated on the projections can be different that the Busemann function of the original histograms. 
We also note that the optimization problem is relatively unstable. 
Thus, further work might be required to better optimize these problems or to better understand the behavior of the components learned.

\section{Conclusion and Discussion}

In this chapter, we studied the Busemann function on the space of probability measures endowed by the Wasserstein distance by first identifying geodesics for which it is well defined and then by computing its closed-form in the one dimensional case and in the Gaussian case. As an application, we used this function in order to perform a principal component analysis and applied it on synthetic and real one dimensional datasets.

\looseness=-1 Future works will try to leverage the closed-form on the Bures-Wasserstein space in order to perform PCA. One might also be interested in finding closed-forms for the Busemann function for more general probability distributions such as mixtures using the distance introduced in \citep{delon2020wasserstein, dusson2023wasserstein} or even to positive measures with for instance the Wasserstein On Positive measures (WOP) distance introduced in \citep{leblanc2023extending} or the unbalanced OT \citep{sejourne2022unbalanced}. 

However, deriving closed-forms for the Busemann function on the Wasserstein space is a relatively difficult problem and we have not yet found applications for which it would bring promising results. For PCA, several obstacles arise, hindering the use of the Busemann function: firstly, the projections might not always be projected on geodesic rays, which can be problematic in practice as it might skew the results. Furthermore, optimizing the objectives is a difficult task already in one dimension, and deriving an algorithm for Gaussian is not straightforward. Finding a task that could be solved with the Busemann function on the Wasserstein space is thus an important avenue of research.

\clearemptydoublepage
\cleartooddpage[\thispagestyle{empty}]
\chapter{Subspace Detours Meet Gromov-Wasserstein} \label{chapter:gw}

{
    \hypersetup{linkcolor=black} 
    \minitoc 
}

In the context of Optimal Transport (OT) methods, the subspace detour approach was recently proposed by \citet{muzellec2019subspace}.  
It consists of first finding an optimal plan between the measures projected on a wisely chosen subspace and then completing it in a nearly optimal transport plan on the whole space. The contribution of this chapter, based on \citep{bonet2021subspace}, is to extend this category of methods to the Gromov--Wasserstein problem, which is a particular type of OT distance involving the specific geometry of each distribution. After deriving the associated formalism and properties, we give an experimental illustration on a shape matching problem. We also discuss a specific cost for which we can show connections with the Knothe--Rosenblatt rearrangement.

\section{Introduction}

Classical Optimal Transport (OT) has received lots of attention recently, in~particular in Machine Learning for tasks such as generative networks \citep{arjovsky2017wasserstein} or domain adaptation \citep{courty2016optimal} to name a few. It generally relies on the Wasserstein distance, which builds an optimal coupling between distributions given a notion of distance between their samples. Yet, this metric cannot be used directly whenever the distributions lie in different metric spaces and lacks from potentially important properties, such as translation or rotation invariance of the supports of the distributions, which can be useful when comparing shapes or meshes \citep{memoli2011gromov, chowdhury2021quantized}. In~order to alleviate those problems, custom solutions have been proposed, such as \citep{alvarez2019towards}, in which invariances are enforced by optimizing over some class of transformations, or \citep{cai2020distances}, in which distributions lying in different spaces are compared by optimizing over the Stiefel manifold to project or embed one of the~measures.

Apart from these works, another meaningful OT distance to tackle these problems is the Gromov--Wasserstein (GW) distance, originally proposed in~\citep{memoli2007use, memoli2011gromov,sturm2012space}. It is a distance between metric spaces and has several appealing properties such as geodesics or invariances \citep{sturm2012space}. Yet, the~price to be paid lies in its computational complexity, which requires solving a nonconvex quadratic optimization problem with linear constraints. A~recent line of work tends to compute approximations or relaxations of the original problem in~order to spread its use in more data-intensive Machine Learning applications. For~example, \citet{peyre2016gromov} rely on entropic regularization and Sinkhorn iterations \citep{cuturi2013sinkhorn}, while recent methods impose coupling with low-rank constraints \citep{scetbon2021linear} or rely on a sliced approach \citep{vayer2019sliced} or on mini-batch estimators \citep{fatras2021minibatch} to approximate the  Gromov--Wasserstein distance. In~\citep{chowdhury2021quantized}, the authors propose to partition the space and to solve the Optimal Transport problem between a subset of points before~finding a coupling between all the points.

In this work, we study the \emph{subspace detour} approach for Gromov--Wasserstein. This class of method was first proposed for the Wasserstein setting by \citet{muzellec2019subspace} and consists of (1) projecting the measures onto a wisely chosen subspace and finding an optimal coupling between them (2) and then constructing a nearly optimal plan of the measures on the whole space using disintegration (see \Cref{section:subspace_detour}). 
Our main contribution is to generalize the subspace detour approach on different subspaces and to apply it for the GW distance. We derive some useful properties as well as closed-form solutions of this transport plan between Gaussians distributions. From~a practical side, we provide a novel closed-form expression of the one-dimensional GW problem that allows us to efficiently compute the subspace detours transport plan when the subspaces are one-dimensional. Illustrations of the method are given on a shape matching problem where we show good results with a cheaper computational cost compared to other GW-based methods.
Interestingly enough, we also propose a separable quadratic cost for the GW problem that can be related with a triangular coupling~\citep{bogachev2005triangular}, hence bridging the gap with Knothe--Rosenblatt (KR) rearrangements~\citep{rosenblatt1952remarks, knothe1957contributions}.

\section{Background}

In this section, we introduce all the necessary material to describe the subspace detours approach for classical Optimal Transport and relate it to the Knothe--Rosenblatt rearrangement. We show how to find couplings via the gluing lemma and measure disintegration. Then, we introduce the Gromov--Wasserstein problem for which we will derive the subspace detour in the next~sections.

\subsection{Classical Optimal~Transport}

We start by recalling some notions of classical transport problems introduced in \Cref{chapter:bg_ot}. Let $\mu,\nu\in\mathcal{P}(\mathbb{R}^d)$ be two probability measures. The~set of couplings between $\mu$ and $\nu$ is defined as: 
\begin{equation}
    \Pi(\mu,\nu) = \{\gamma\in\mathcal{P}(\mathbb{R}^d\times \mathbb{R}^d)|\ \pi^1_\#\gamma=\mu,\ \pi^2_\#\gamma=\nu\}
\end{equation}
where $\pi^1$ and $\pi^2$ are the projections on the first and second coordinate (i.e.,~$\pi^1(x,y)=x$), respectively.


\paragraph{Optimal coupling.}

There exists several types of coupling between probability measures for which a non-exhaustive list can be found in \citep[Chapter 1]{villani2009optimal}. Among~them, the~so called optimal coupling is the minimizer of the Kantorovich problem \eqref{eq:Kantorovich_problem} which we recall here: 
\begin{equation}
    \inf_{\gamma\in\Pi(\mu,\nu)}\ \int c(x,y)\ \mathrm{d}\gamma(x,y)
\end{equation}
with $c$ being some cost function. 

In one dimension, with~$\mu$ atomless, the~solution to \eqref{eq:Kantorovich_problem} when $c(x,y)=|x-y|^2$ is a deterministic coupling of the form $(\id, T)_\#\mu$ \citep[Theorem 2.5]{santambrogio2015optimal} with:
\begin{equation} \label{increasing_rearragnement}
    T=F_\nu^{-1}\circ F_\mu
\end{equation}
where $F_\mu$ is the cumulative distribution function of $\mu$, and $F_\nu^{-1}$ is the quantile function of $\nu$.
This map is also known as the increasing rearrangement map.

\paragraph{Knothe--Rosenblatt rearrangement.} \label{paragraph:knothe}

Another interesting coupling is the Knothe--Rosenblatt (KR) rearrangement, which takes advantage of the increasing rearrangement in one dimension by iterating over the dimension and using the disintegration of the measures. Concatenating all the increasing rearrangements between the conditional probabilities, the~KR rearrangement produces a nondecreasing triangular map (\emph{i.e.},~$T:\mathbb{R}^d\to\mathbb{R}^d$, for~all $x\in\mathbb{R}^d$, $T(x)=\big(T_1(x_1),\dots,T_j(x_1,\dots,x_j),\dots,T_d(x)\big)$, and for all $j$, $T_j$ is nondecreasing with respect to $x_j$), and~a deterministic coupling (\emph{i.e.},~$T_\#\mu=\nu$) \citep{villani2009optimal, santambrogio2015optimal, jaini2019sum}.

\citet{carlier2010knothe} made a connection between this coupling and Optimal Transport by showing that it can be obtained as the limit of OT plans for a degenerated cost :
\begin{equation} \label{eq:degenerated_cost}
    c_t(x,y)=\sum_{i=1}^d\lambda_i(t)(x_i-y_i)^2,
\end{equation}
where for all $i\in\{1,\dots,d\}$, $t>0$, $\lambda_i(t)>0$, and for all $i\ge 2$, $\frac{\lambda_i(t)}{\lambda_{i-1}(t)}\xrightarrow[t\to 0]{}0$. This cost can be recast as in \citep{bonnotte2013unidimensional} as $c_t(x,y)=(x-y)^T A_t (x-y)$, where $A_t = \mathrm{diag}\big(\lambda_1(t),\dots,\lambda_d(t)\big)$.
This formalizes into the following Theorem:
\begin{theorem}{\emph{\citep{carlier2010knothe, santambrogio2015optimal}.}} \label{theorem:KnotheToBrenier}
    Let $\mu$ and $\nu$ be two absolutely continuous measures on $\mathbb{R}^d$, with~compact supports. Let $\gamma_t$ be an Optimal Transport plan for the cost $c_t$, let $T_K$ be the Knothe--Rosenblatt map between $\mu$ and $\nu$, and~$\gamma_K = (\id, T_K)_\#\mu$ the associated transport plan. Then, we have $\gamma_t\xrightarrow[t\rightarrow 0]{\mathcal{D}}\gamma_K$. Moreover, if~$\gamma_t$ are induced by transport maps $T_t$, then $T_t$ converges in $L^2(\mu)$ when t tends to zero to the Knothe--Rosenblatt rearrangement $T_K$.
\end{theorem}

\subsection{Subspace Detours and~Disintegration} \label{section:subspace_detour}

\citet{muzellec2019subspace} proposed another OT problem by optimizing over the couplings which share a measure on a subspace. More precisely, they defined subspace-optimal plans for which the shared measure is the OT plan between projected measures.
\begin{definition}[Subspace-Optimal Plans, Definition 1 in \citep{muzellec2019subspace}]
    Let $\mu,\nu\in\mathcal{P}_2(\mathbb{R}^d)$ and let $E\subset \mathbb{R}^d$ be a $k$-dimensional subspace. Let $\gamma_{E}^*$ be an OT plan for the Wasserstein distance between $\mu_E=\pi^E_\#\mu$ and $\nu_E=\pi^E_\#\nu$ (with $\pi^E$ as the orthogonal projection on $E$). Then, the set of $E$-optimal plans between $\mu$ and $\nu$ is defined as $\Pi_E(\mu,\nu)=\{\gamma\in\Pi(\mu,\nu)|\ (\pi^E,\pi^E)_\#\gamma=\gamma_{E}^*\}$.
\end{definition}

In other words, the subspace OT plans are the transport plans of $\mu,\nu$ that agree on the subspace $E$ with the optimal transport plan $\gamma_E^{*}$ on this subspace. To~construct such coupling $\gamma\in\Pi(\mu,\nu)$, one can rely on the Gluing lemma \citep{villani2009optimal} or use the disintegration of the measure (see \Cref{def:disintegration}).

\paragraph{Coupling on the whole space.}

Let us note $\mu_{E^\bot|E}$ and $\nu_{E^\bot|E}$ as the disintegrated measures on the orthogonal spaces (\emph{i.e.}, such that $\mu=\mu_E \otimes \mu_{E^\bot|E}$ and $\nu=\nu_E\otimes \nu_{E^\bot|E}$ or if we have densities, $p(x_E,x_{E^\bot})=p_E(x_E)p_{E^\bot|E}(x_{E^\bot}|x_E)$). Then, to~obtain a transport plan between the two original measures on the whole space, we can look for another coupling between disintegrated measures $\mu_{E^\bot|E}$ and $\nu_{E^\bot|E}$. In~particular, two such couplings are proposed in~\citep{muzellec2019subspace}, the~Monge-Independent (MI) plan:
\begin{equation}
    \pi_{\mathrm{MI}} = \gamma_{E}^*\otimes (\mu_{E^\bot|E}\otimes \nu_{E^\bot |E})
\end{equation}
where we take the independent coupling between $\mu_{E^\bot|E}(x_E,\cdot)$ and $\nu_{E^\bot|E}(y_E,\cdot)$ for $\gamma_{E}^*$ almost every $(x_E,y_E)$, and~the Monge-Knothe (MK) plan:
\begin{equation}
    \pi_{\mathrm{MK}} = \gamma_{E}^*\otimes \gamma_{E^\bot|E}^*
\end{equation}
where $\gamma_{E^\bot |E}^*\big((x_E,y_E),\cdot\big)$ is an optimal plan between $\mu_{E^\bot|E}(x_E,\cdot)$ and $\nu_{E^\bot|E}(y_E,\cdot)$ for $\gamma_{E}^*$ almost every $(x_E,y_E)$.
\citet{muzellec2019subspace} observed that MI is more adapted to noisy environments since it only computes the OT plan of the subspace while MK is more suited for applications where we want to prioritize some subspace but where all the directions still contain relevant information.

This subspace detour approach can be of much interest following the popular assumption that two distributions on $\mathbb{R}^d$ differ only in a low-dimensional subspace as in the Spiked transport model \citep{niles2022estimation}. However, it is still required to find the adequate subspace. \citet{muzellec2019subspace} propose to either  rely on a priori knowledge to select the subspace (by using, \emph{e.g.}, a reference dataset and a principal component analysis) or to optimize over the Stiefel manifold.

\subsection{Gromov--Wasserstein}

 Formally, the~Gromov--Wasserstein distance allows us to compare metric measure spaces (mm-space), triplets $(X,d_X,\mu_X)$ and $(Y,d_Y,\mu_Y)$, where $(X,d_X)$ and $(Y,d_Y)$ are complete separable metric spaces and $\mu_X$ and $\mu_Y$ are Borel probability measures on $X$ and $Y$ \citep{sturm2012space}, respectively, by computing:
\begin{equation}
    GW(X,Y) = \inf_{\gamma\in\Pi(\mu_X,\mu_Y)}\iint L\big(d_X(x,x'),d_Y(y,y')\big) \ \mathrm{d}\gamma(x,y)\mathrm{d}\gamma(x',y')
\end{equation}
where $L$ is some loss on $\mathbb{R}$. It has actually been extended to other spaces by replacing the distances by cost functions $c_X$ and $c_Y$, as, \emph{e.g.}, in \citep{chowdhury2019gromov}. Furthermore, it has many appealing properties such as having invariances (which depend on the costs).

\citet{vayer2020contribution} notably studied this problem in the setting where $X$ and $Y$ are Euclidean spaces, with~$L(x,y)=(x-y)^2$ and $c(x,x')=\langle x,x'\rangle$ or $c(x,x')=\|x-x'\|_2^2$. In~particular, let $\mu\in\mathcal{P}(\mathbb{R}^p)$ and  $\nu\in\mathcal{P}(\mathbb{R}^q)$, and define the~inner-GW problem as:
\begin{equation} \label{eq:gw_ip}
    \mathrm{InnerGW}(\mu,\nu) = \inf_{\gamma\in\Pi(\mu,\nu)} \iint\big(\langle x,x'\rangle_p-\langle y,y'\rangle_q\big)^2\ \mathrm{d}\gamma(x,y)\mathrm{d}\gamma(x',y').
\end{equation}
For this problem, a~closed form in one dimension can be found when one of the distributions admits a density \emph{w.r.t.} the Lebesgue measure:
\begin{theorem}[Theorem 4.2.4 in \citep{vayer2020contribution}] \label{theorem:gw_ip_1d}
    Let $\mu,\nu\in\mathcal{P}(\mathbb{R})$, with~$\mu$ being absolutely continuous with respect to the Lebesgue measure. Let $F_\mu^\nearrow(x)\coloneqq F_\mu(x) =\mu(]-\infty,x])$ be the cumulative distribution function and $F_\mu^\searrow(x)=\mu(]-x,+\infty[)$ the anti-cumulative distribution function. Let $T_{asc}(x)=F_\nu^{-1}\big(F_\mu^\nearrow(x)\big)$ and $T_{desc}(x)=F_\nu^{-1}\big(F_\mu^\searrow(-x)\big)$. Then, an~optimal solution of \eqref{eq:gw_ip} is achieved either by $\gamma=(\id, T_{asc})_\#\mu$ or by $\gamma=(\id, T_{desc})_\#\mu$.
\end{theorem}

\section{Subspace Detours for $GW$} \label{section:subspace_detours_gw}

In this section, we propose to extend subspace detours from \citet{muzellec2019subspace} with Gromov--Wasserstein costs. We show that we can even take subspaces of different dimensions and~still obtain a coupling on the whole space using the Independent or the Monge--Knothe coupling. Then, we derive some properties analogously to \citet{muzellec2019subspace}, as~well as some closed-form solutions between Gaussians. We also provide a new closed-form expression of the inner-GW problem between one-dimensional discrete distributions and provide an illustration on a shape-matching~problem.

\subsection{Motivations}

\begin{figure}[t]
    \centering
    \includegraphics[width=\linewidth]{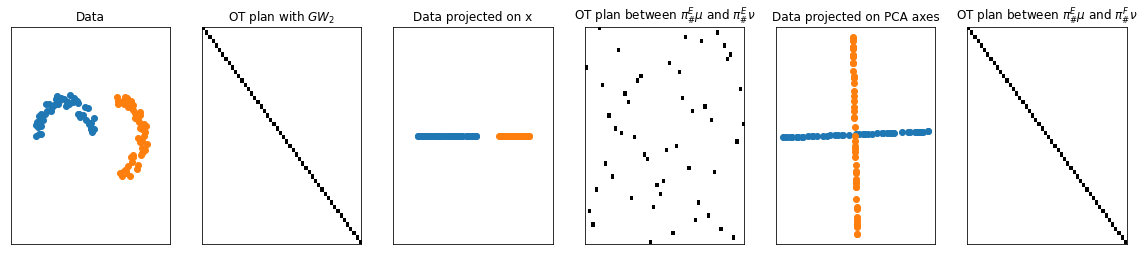}
    \caption{From left to right: Data (moons); OT plan obtained with GW for $c(x,x')=\|x-x'\|_2^2$; Data projected on the first axis; OT plan obtained between the projected measures; Data projected on their first PCA component; OT plan obtained between the the projected~measures.}
    \label{fig:GW}
\end{figure}

First, we adapt the definition of subspace optimal plans for \emph{different} subspaces. Indeed, since the GW distance is adapted to distributions that have their own geometry, we argue that if we project on the \emph{same} subspace, then it is likely that the resulting coupling would not be coherent with that of GW. 
To illustrate this point, we use as a source distribution $\mu$ one moon of the two moons dataset and~obtain a target $\nu$ by rotating $\mu$ by an angle of $\frac{\pi}{2}$ (see \Cref{fig:GW}). As~the GW with $c(x,x')=\|x-x'\|_2^2$ is invariant with respect to isometries, the~optimal coupling is diagonal, as recovered on the left side of the figure. However, when choosing one subspace to project both the source and target distributions, we completely lose the optimal coupling between them. Nonetheless, by~choosing one subspace for each measure more wisely (using here the first component of the principal component analysis (PCA) decomposition), we recover the diagonal coupling. This simple illustration underlines that the choice of both subspaces is important. A~way of choosing the subspaces could be to project on the subspace containing the more information for each dataset using, \emph{e.g.},~PCA independently on each distribution. \citet{muzellec2019subspace} proposed to optimize the optimal transport cost with respect to an orthonormal matrix with a projected gradient descent, which could be extended to an optimization over two orthonormal matrices in our~context.

By allowing for different subspaces, we obtain the following definition of subspace optimal~plans:

\begin{definition} \label{def_subspace_OT_plans}
    Let $\mu\in\mathcal{P}_2(\mathbb{R}^p)$, $\nu\in\mathcal{P}_2(\mathbb{R}^q)$, $E$ be a $k$-dimensional subspace of $\mathbb{R}^p$ and $F$ a $k'$-dimensional subspace of $\mathbb{R}^q$. Let $\gamma_{E\times F}^*$ be an optimal transport plan for $GW$ between $\mu_E=\pi^E_\#\mu$ and $\nu_F=\pi^F_\#\nu$ (with $\pi^E$ (resp. $\pi^F$) the orthogonal projection on $E$ (resp. $F$)). Then, the set of $(E,F)$-optimal plans between $\mu$ and $\nu$ is defined as $\Pi_{E,F}(\mu,\nu)=\{\gamma\in\Pi(\mu,\nu)|\ (\pi^E,\pi^F)_\#\gamma=\gamma_{E\times F}^*\}$.
\end{definition}

\looseness=-1 Analogously to \citet{muzellec2019subspace} (\Cref{section:subspace_detour}), we can obtain from $\gamma_{E\times F}^*$ a coupling on the whole space by either defining the Monge--Independent plan $\pi_{\mathrm{MI}} = \gamma_{E\times F}^*\otimes (\mu_{E^\bot|E}\otimes \nu_{F^\bot |F})$ or the Monge--Knothe plan $\pi_{\mathrm{MK}} = \gamma_{E\times F}^*\otimes \gamma_{E^\bot\times F^\bot|E\times F}^*$ where OT plans are taken with some OT cost, \emph{e.g.} $GW$.

\subsection{Properties}

Let $E\subset \mathbb{R}^p$ and $F\subset \mathbb{R}^q$ and denote:
\begin{equation} \label{GWEF}
    GW_{E,F}(\mu,\nu) = \inf_{\gamma\in\Pi_{E,F}(\mu,\nu)}\ \iint L(x,x',y,y')\ \mathrm{d}\gamma(x,y)\mathrm{d}\gamma(x',y')
\end{equation}
the Gromov--Wasserstein problem restricted to subspace optimal plans (\Cref{def_subspace_OT_plans}). In~the following, we show that Monge--Knothe couplings are optimal plans of this problem, which is a direct transposition of Proposition 1 in~\citep{muzellec2019subspace}.

\begin{proposition} \label{prop:prop1}
    Let $\mu\in\mathcal{P}(\mathbb{R}^p)$ and $\nu\in\mathcal{P}(\mathbb{R}^q)$, $E\subset \mathbb{R}^p$, $F\subset\mathbb{R}^q$, $\pi_{\mathrm{MK}}=\gamma_{E\times F}^*\otimes \gamma_{E^\bot\times F^\bot|E\times F}^*$, where $\gamma_{E\times F}^*$ is an optimal coupling between $\mu_E$ and $\nu_F$, and~for $\gamma_{E\times F}^*$, almost every $(x_E,y_F)$, $\gamma_{E^\bot\times F^\bot| E\times F}^*\big((x_E,y_F),\cdot\big)$ is an optimal coupling between $\mu_{E^\bot|E}(x_E,\cdot)$ and $\nu_{F^\bot|F}(y_F,\cdot)$. Then we have:
    \begin{equation}
        \pi_{\mathrm{MK}}\in\argmin_{\gamma\in\Pi_{E,F}(\mu,\nu)} \iint L(x,x',y,y')\  \mathrm{d}\gamma(x,y)\mathrm{d}\gamma(x',y').
    \end{equation}
\end{proposition}

\begin{proof}
    See \Cref{proof:prop1}.
\end{proof}

The key properties of $GW$ that we would like to keep are its invariances. We show in two particular cases that we conserve them on the orthogonal spaces (since the measure on $E\times F$ is fixed).

\begin{proposition} \label{prop:prop2}
    Let $\mu\in\mathcal{P}(\mathbb{R}^p)$, $\nu\in\mathcal{P}(\mathbb{R}^q)$, $E\subset \mathbb{R}^p$, $F\subset\mathbb{R}^q$. 
    For $L(x,x',y,y')=\big(\|x-x'\|_2^2 - \|y-y'\|_2^2\big)^2$ or $L(x,x',y,y') = \big(\langle x,x'\rangle_p - \langle y,y'\rangle_q\big)^2$, $GW_{E,F}$ \eqref{GWEF} is invariant with respect to isometries of the form $f=(\id_E,f_{E^\bot})$ (resp. $g=(\id_F,g_{F^\bot})$) with $f_{E^\bot}$ an isometry on $E^\bot$ (resp. $g_{F^\bot}$ an isometry on $F^\bot$) with respect to the corresponding cost ($c(x,x')=\|x-x'\|_2^2$ or $c(x,x')=\langle x,x'\rangle_p$).
\end{proposition}

\begin{proof}

    We propose a sketch of the proof. The~full proof can be found in \Cref{proof:prop2}. Let $L(x,x',y,y')=\big(\|x-x'\|_2^2-\|y-y'\|_2^2\big)^2$, let $f_{E^\bot}$ be an isometry \emph{w.r.t.} $c(x_{E^\bot},x'_{E^\bot})=\|x_{E^\bot}-x'_{E^\bot}\|_2^2$, and let $f:\mathbb{R}^p\to\mathbb{R}^p$ be defined as such for all $x\in\mathbb{R}^p$, $f(x)=(x_E,f_{E^\bot}(x_{E^\bot}))$.
    
    By using \Cref{lemma:paty}, we show that $\Pi_{E,F}(f_\#\mu,\nu)=\{(f,\id)_\#\gamma,\ \gamma\in\Pi_{E,F}(\mu,\nu)\}$.
    Hence, for~all $\gamma\in\Pi_{E,F}(f_\#\mu,\nu)$, there exists $\Tilde{\gamma}\in\Pi_{E,F}(\mu,\nu)$ such that $\gamma=(f,\id)_\#\Tilde{\gamma}$. By~disintegrating $\Tilde{\gamma}$ with respect to $\gamma_{E\times F}^*$ and using the properties of the pushforward, we can show that:
    \begin{equation}
        \begin{aligned}
            &\iint \big(\|x-x'\|_2^2-\|y-y'\|_2^2\big)^2\ \mathrm{d}(f,\id)_\#\Tilde{\gamma}(x,y) \mathrm{d}(f,\id)_\#\Tilde{\gamma}(x',y') \\ 
            &= \iint \big(\|x-x'\|_2^2-\|y-y'\|_2^2\big)^2\ \mathrm{d}\Tilde{\gamma}(x,y)\mathrm{d}\Tilde{\gamma}(x',y').
        \end{aligned}
    \end{equation}
    Finally, by~taking the infimum with respect to $\Tilde{\gamma}\in\Pi_{E,F}(\mu,\nu)$, we find:
    \begin{equation}
        GW_{E,F}(f_\#\mu,\nu) = GW_{E,F}(\mu,\nu).
    \end{equation}
\end{proof}

\subsection{Closed-Form between~Gaussians}

We can also derive explicit formulas between Gaussians in particular cases. Let $q\le p$, $\mu=\mathcal{N}(m_\mu,\Sigma)\in\mathcal{P}(\mathbb{R}^p)$, $\nu=\mathcal{N}(m_\nu,\Lambda)\in\mathcal{P}(\mathbb{R}^q)$ two Gaussian measures with $\Sigma=P_\mu D_\mu P_\mu^T$ and $\Lambda=P_\nu D_\nu P_\nu^T$. As~previously, let $E\subset\mathbb{R}^p$ and $F\subset\mathbb{R}^q$ be  $k$ and $k'$ dimensional subspaces, respectively.
Following \citet{muzellec2019subspace}, we represent $\Sigma$ in an orthonormal basis of $E\oplus E^\bot$ and~$\Lambda$ in an orthonormal basis of $F\oplus F^\bot$, \emph{i.e.} $\Sigma = \begin{pmatrix}\Sigma_E & \Sigma_{EE^\bot} \\ \Sigma_{E^\bot E} & \Sigma_{E^\bot}\end{pmatrix}$. Now, let us denote the following:
\begin{equation}
    \Sigma/\Sigma_E = \Sigma_{E^\bot}-\Sigma_{EE^\bot}^T\Sigma_E^{-1}\Sigma_{EE^\bot}    
\end{equation}
as the Schur complement of $\Sigma$ with respect to $\Sigma_E$. We know that the conditionals of Gaussians are Gaussians and~that their covariances are the Schur complements (see, \emph{e.g.} \citep{von2014mathematical, rasmussen2003gaussian}).

For $L(x,x',y,y')=\big(\|x-x'\|_2^2-\|y-y'\|_2^2\big)^2$, we have for now no certainty that the optimal transport plan is Gaussian. Let $\mathcal{N}_{p+q}$ denote the set of Gaussians in $\mathbb{R}^{p+q}$. By~restricting the minimization problem to Gaussian couplings, \emph{i.e.},~by solving:
\begin{equation} \label{GGW}
    \mathrm{GGW}(\mu,\nu) = \inf_{\gamma\in\Pi(\mu,\nu)\cap \mathcal{N}_{p+q}} \iint \big(\|x-x'\|_2^2-\|y-y'\|_2^2\big)^2\ \mathrm{d}\gamma(x,y)\mathrm{d}\gamma(x',y'),
\end{equation} 
\citet{salmona2021gromov} showed that there is a solution $\gamma^*=(\id,T)_\#\mu\in\Pi(\mu,\nu)$ with $\mu=\mathcal{N}(m_\mu,\Sigma)$, $\nu=\mathcal{N}(m_\nu,\Lambda)$ and
\begin{equation} \label{monge_map_gw}
    \forall x\in\mathbb{R}^d,\ T(x)=m_\nu+P_\nu A P_\mu^T (x-m_\mu)
\end{equation}
where $A=\begin{pmatrix} \Tilde{I}_q D_\nu^{\frac12}(D_\mu^{(q)})^{-\frac12} & 0_{q,p-q} \end{pmatrix}\in\mathbb{R}^{q\times p}$, and $\Tilde{I}_q$ is of the form $\mathrm{diag}\big((\pm 1)_{i\le q}\big)$.

By combining the results of \citet{muzellec2019subspace} and \citet{salmona2021gromov}, we obtain the following closed-form for Monge--Knothe~couplings:

\begin{proposition} \label{prop:closed_form_mk}
    Suppose $p\ge q$ and $k=k'$. For~the Gaussian-restricted GW problem \eqref{GGW}, a~Monge--Knothe transport map between $\mu=\mathcal{N}(m_\mu,\Sigma)\in\mathcal{P}(\mathbb{R}^p)$ and $\nu=\mathcal{N}(m_\nu,\Lambda)\in\mathcal{P}(\mathbb{R}^q)$ is, for~all $x\in\mathbb{R}^p$, $T_{\mathrm{MK}}(x) = m_\nu + B(x-m_\mu)$ where:
    \begin{equation}
        B = \begin{pmatrix}
        T_{E,F} & 0 \\
        C & T_{E^\bot,F^\bot|E,F}
    \end{pmatrix}
    \end{equation}
    with $T_{E,F}$ being an optimal transport map between $\mathcal{N}(0_E,\Sigma_E)$ and $\mathcal{N}(0_F,\Lambda_F)$ (of the form \eqref{monge_map_gw}), $T_{E^\bot,F^\bot|E,F}$ an optimal transport map between $\mathcal{N}(0_{E^\bot},\Sigma/\Sigma_E)$ and $\mathcal{N}(0_{F^\bot},\Lambda/\Lambda_F)$, and $C$ satisfies:
    \begin{equation}
        C = \big(\Lambda_{F^\bot F} (T_{E,F}^T)^{-1} - T_{E^\bot,F^\bot|E,F}\Sigma_{E^\bot E}\big)\Sigma_E^{-1}.
    \end{equation}
\end{proposition}

\begin{proof}
    See \Cref{proof:prop_closed_form_mk}.
\end{proof}

Suppose that $k\ge k'$, $m_\mu=0$, and $m_\nu=0$ and let $T_{E,F}$ be an optimal transport map between $\mu_E$ and $\nu_F$ (of the form \eqref{monge_map_gw}). We can derive a formula for the Monge--Independent coupling for the inner-GW problem and the Gaussian restricted GW problem.
\begin{proposition} \label{prop:closed_form_mi}
    $\pi_{\mathrm{MI}}=\mathcal{N}(0_{p+q},\Gamma)$ where $\Gamma = \begin{pmatrix} \Sigma & C \\ C^T & \Lambda \end{pmatrix}$ with
    \begin{equation}
        C = (V_E\Sigma_E+V_{E^\bot}\Sigma_{E^\bot E})T_{E,F}^T(V_F^T+\Lambda_F^{-1}\Lambda_{F^\bot F}^T V_{F^\bot}^T)
    \end{equation}
    where $T_{E,F}$ is an optimal transport map, either for the inner-GW problem or the Gaussian restricted problem.
\end{proposition}

\begin{proof}
    See \Cref{proof:prop_closed_form_mi}.
\end{proof}

\subsection{Computation of Inner-GW between One-Dimensional Empirical~Measures}

In practice, computing the Gromov--Wasserstein distance from samples of the distributions is costly. From~a computational point of view, the~subspace detour approach  provides an interesting method with better computational complexity when choosing 1D subspaces. Moreover, we have the intuition that the GW problem between measures lying on smaller dimensional subspaces has a better sample complexity than between the original measures, as~it is the case for the Wasserstein distance~\citep{weed2019sharp,lin2021projection}.

Below, we show that when both $E$ and $F$ are one-dimensional subspaces, then the resulting GW problem between the projected measures can be solved in linear time. This will rely on a new closed-form expression of the GW problem in 1D. 
\citet{vayer2020contribution} provided in Theorem 4.2.4 a closed-form for the inner-GW problem when one of the probability distributions is absolutely continuous with respect to the Lebesgue measure. However, we are interested here in computing inner-GW between discrete distributions.
We provide in the next proposition a closed-form expression for the inner-GW problem between any unidimensional discrete probability distributions:

\begin{proposition} \label{prop:closed_form_gw_1D}
    Consider $\Sigma_n=\{a \in \mathbb{R}^{n}_{+}, \sum_{i=1}^{n} a_i=1\}$ the $n$ probability simplex. For~a vector $a \in \mathbb{R}^{n}$, we denote $a^{-}$ as the vector with values reversed, \emph{i.e.} $a^- = (a_n,\dots,a_1)$.
    Let $\mu=\sum_{i=1}^{n} a_i\delta_{x_i},\nu=\sum_{j=1}^{m} b_j\delta_{y_j} \in \mathcal{P}(\mathbb{R})$ with $a \in \Sigma_n,\ b \in \Sigma_m$. Suppose that $x_{1} \leq \dots \leq x_{n}$ and $y_{1} \leq \dots \leq y_{m}$. Consider the problem:
    \begin{equation} \label{eq:gw1Dinner}
        \min_{\gamma \in \Pi(a,b)} \sum_{ijkl} (x_ix_k-y_jy_l)^{2} \gamma_{ij}\gamma_{kl}
    \end{equation}
    Then, there exists $\gamma \in \{ NW(a,b),NW(a^{-},b)\}$ such that $\gamma$ is an optimal solution of \eqref{eq:gw1Dinner} where $NW$ is the North-West corner rule defined in Algorithm~\ref{alg:nw}.
    As a corollary, an optimal solution of \eqref{eq:gw1Dinner} can be found in $O(n+m)$.
\end{proposition}

\begin{proof}
    See \Cref{proof:prop_closed_form_gw_1D}.
\end{proof}

\begin{algorithm}[t]
    \caption{\label{alg:nw} North-West corner rule $NW(a,b)$}
    \begin{algorithmic}
        \STATE $a \in \Sigma_n, b \in \Sigma_m$
        \WHILE{$i\le n$,\ $j\le m$}
        \STATE $\gamma_{ij}=\min\{a_i,b_j\}$ 
        \STATE $a_i=a_i-\gamma_{ij}$
        \STATE $b_j=b_j-\gamma_{ij}$ 
        \STATE{\text{If} $a_i=0$, $i=i+1$, \text{if} $b_j=0$, $j=j+1$}
        \ENDWHILE 
        \RETURN $\gamma \in \Pi(a,b)$
    \end{algorithmic}
\end{algorithm}


\Cref{theorem:gw_ip_1d} is not directly applicable to this setting since it requires having absolutely regular distributions, which is not the case here. Both results are, however, related, as the solution obtained by using the NW corner rule on the sorted samples is the same as that obtained by considering the coupling obtained from the quantile functions. Note that the previous result could also be used to define tractable alternatives to GW in the same manner as the Sliced Gromov--Wasserstein~\citep{vayer2019sliced}.

\subsection{Illustrations}

\begin{figure}[!htpb] 
    \includegraphics[width=0.98\linewidth]{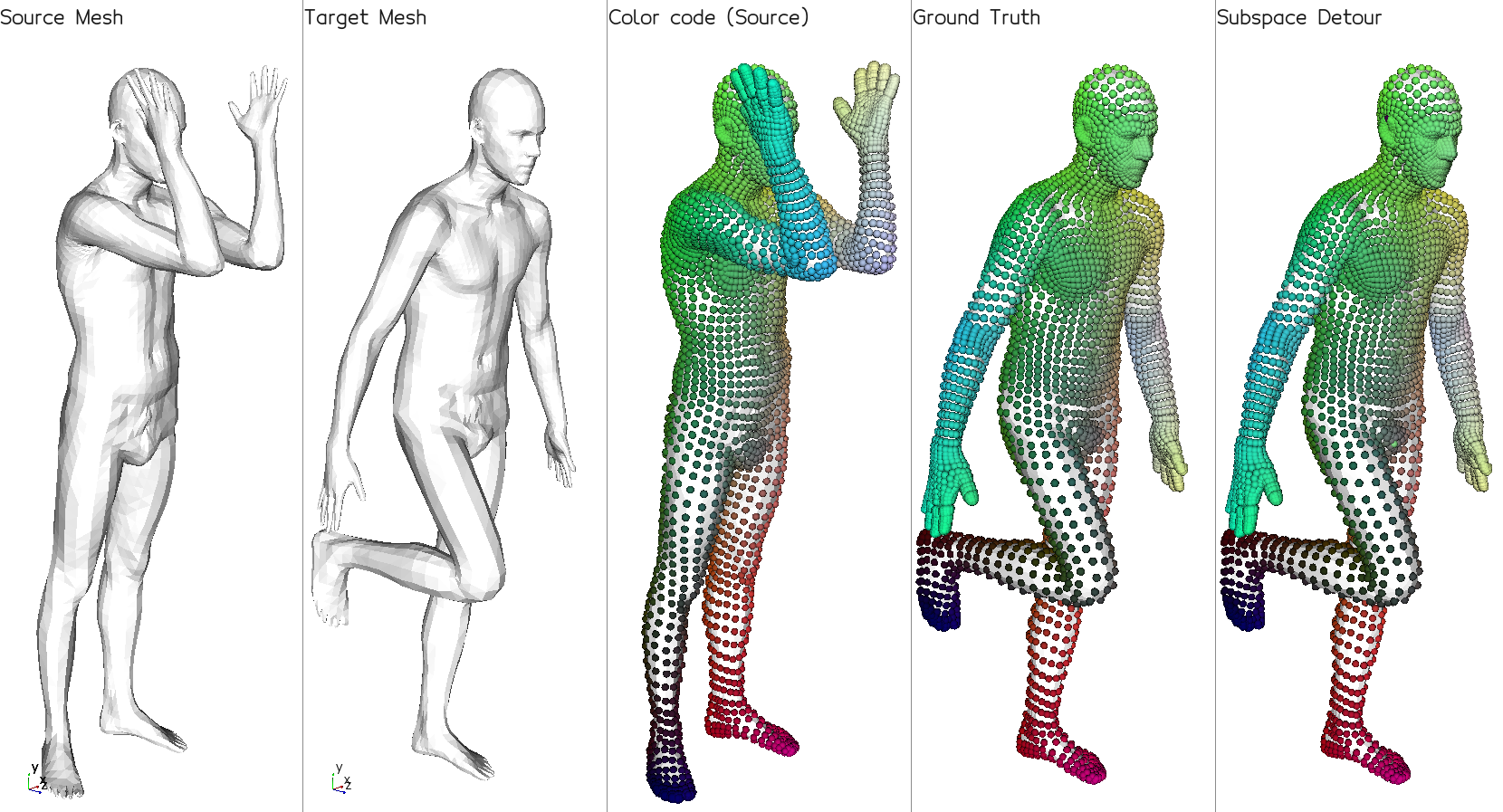}
    \includegraphics[width=0.98\linewidth]{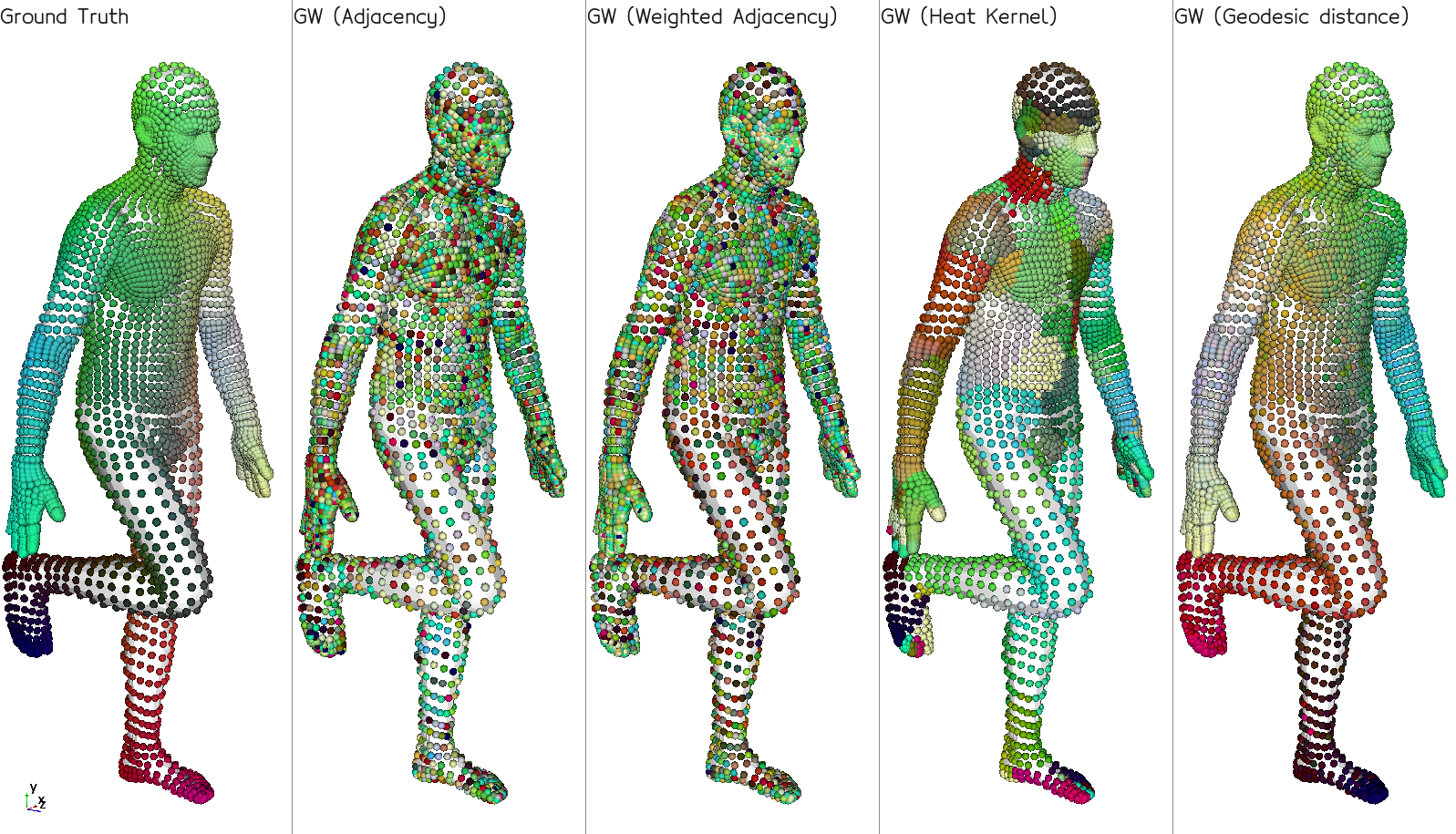}
    \caption{Three-dimensional mesh registration. (\textbf{First row}) source and target meshes, color code of the source, ground truth color code on the target, result of subspace detour using Fiedler vectors as subspace. (\textbf{Second row}) After recalling the expected ground truth for ease of comparison, we present results of different Gromov--Wasserstein mappings obtained with metrics based on adjacency, heat kernel, and geodesic~distances.}
    \label{fig:mesh}
\end{figure}

We use the Python Optimal Transport (POT) library \citep{flamary2021pot} to compute the different Optimal Transport problems involved in this illustration. We are interested here in solving a 3D mesh registration problem, which is a natural application of Gromov--Wasserstein~\citep{memoli2011gromov} since it enjoys invariances with respect to isometries such as permutations and~can also naturally exploit the topology of the meshes. For~this purpose, we selected two base meshes  from the {\sc Faust} dataset~\citep{bogo2014faust}, which provides ground
truth correspondences between shapes. The~information available from those meshes are geometrical ($6890$ vertices positions) and topological (mesh connectivity). These two meshes are represented, along with the visual results of the registration, in~Figure~\ref{fig:mesh}. In~order to visually depict the quality of the assignment induced by the transport map, we propagate through it a color code of the source vertices toward their associated counterpart vertices in the target mesh. Both the original color-coded source and the associated target ground truth are available on the first line of the illustration. To~compute our method, we simply use as a natural subspace for both meshes the algebraic connectivity of the mesh's topological information, also known as the Fiedler vector~\citep{fiedler1973algebraic} (eigenvector associated to the second smallest eigenvalue of the un-normalized Laplacian matrix). Fiedler vectors are computed in practice using \texttt{NetworkX}~\citep{hagberg2008exploring} but could also be obtained by using power methods~\citep{wu2014efficient}. Reduced to a 1D Optimal Transport problem (\ref{eq:gw1Dinner}), we used the Proposition \ref{prop:closed_form_gw_1D} to compute the optimal coupling in $O(n+m)$. 
Consequently, the~computation time is very low ($\sim5$ secs. on a standard laptop), and~the associated matching is very good, with more than $98\%$ of correct assignments. We qualitatively compare this result to Gromov--Wasserstein mappings induced by different cost functions, in~the second line of Figure~\ref{fig:mesh}: adjacency \citep{xu2019scalable}, weighted adjacency (weights are given by distances between vertices), heat kernel (derived from the un-normalized Laplacian) \citep{chowdhury2021generalized}, and, finally, geodesic distances over the meshes. On~average, computing the Gromov--Wasserstein mapping using POT took around $10$ minutes of time. Both methods based on adjacency fail to recover a meaningful mapping. Heat kernel allows us to map continuous areas of the source mesh but~fails in recovering a global structure. Finally, the~geodesic distance gives a much more coherent mapping but~has inverted left and right of the human figure. Notably, a~significant extra computation time was induced by the computation of the geodesic distances ($\sim1$h/mesh using the \texttt{NetworkX}~\citep{hagberg2008exploring} shortest path procedure). As~a conclusion, and~despite the simplification of the original problem, our method performs best with~a speed-up of two-orders of~magnitude. 

\section{Triangular Coupling as Limit of Optimal Transport Plans for Quadratic~Cost} \label{section:triangular_coupling}

Another interesting property derived in \citet{muzellec2019subspace} of the Monge--Knothe coupling is that it can be obtained as the limit of classic optimal transport plans, similar to \Cref{theorem:KnotheToBrenier}, using a separable cost of the form:
\begin{equation}
    c_t(x,y) = (x-y)^T P_t (x-y)
\end{equation}
with $P_t = V_EV_E^T + tV_{E^\bot}V_{E^\bot}^T$ and $(V_E,V_{E^\bot})$ as an orthonormal basis of $\mathbb{R}^p$. 

However, this property is not valid for the classical Gromov--Wasserstein cost (e.g., \linebreak $L(x,x',y,y')=\big(d_X(x,x')^2-d_Y(y,y')^2\big)^2$ or $L(x,x',y,y')=\big(\langle x,x'\rangle_p-\langle y,y'\rangle_q\big)^2$) as the cost is not separable. Motivated by this question, we ask ourselves in the following if we can derive a quadratic optimal transport cost for which we would have this~property.

Formally, we derive a new quadratic optimal transport problem using the Hadamard product. We show that this problem is well-defined and that it has interesting properties such as invariance with respect to axes. We also show that it can be related to a triangular coupling in a similar fashion to the classical Optimal Transport problem with the Knothe--Rosenblatt rearrangement.

\subsection{Construction of the Hadamard--Wasserstein~Problem}

In this part, we define the ``Hadamard--Wasserstein'' problem between $\mu\in\mathcal{P}(\mathbb{R}^d)$ and $\nu\in\mathcal{P}(\mathbb{R}^d)$ as:
\begin{equation} \label{HadamardWasserstein}
    \mathcal{HW}^2(\mu,\nu) = \inf_{\gamma\in\Pi(\mu,\nu)} \iint \|x\odot x'-y\odot y'\|_2^2\ \mathrm{d}\gamma(x,y)\mathrm{d}\gamma(x',y'),
\end{equation}
where $\odot$ is the Hadamard product (element-wise product). This problem is different than the Gromov--Wasserstein problem in the sense that we do not compare intradistance anymore bur rather the Hadamard products between vectors of the two spaces (in the same fashion as the classical Wasserstein distance). Hence, we need the two measures to belong in the same Euclidean space. Let us note $L$ as the cost defined as:
\begin{equation} \label{loss_HW}
    \forall x,x',y,y' \in\mathbb{R}^d,\ L(x,x',y,y') = \sum_{k=1}^d (x_kx_k'-y_ky_k')^2 = \|x\odot x' - y\odot y'\|_2^2.
\end{equation}
We observe that it coincides with the inner-GW \eqref{eq:gw_ip} loss in one dimension. Therefore, by \ref{theorem:gw_ip_1d}, we know that we have a closed-form solution in 1D.

\subsection{Properties} \label{section:properties_hgw}

First, we derive some useful properties of \eqref{HadamardWasserstein} which are usual for the regular Gromov--Wasserstein problem. Formally, we show that the problem is well defined and~that it is a pseudometric with invariances with respect to axes.

\begin{proposition} \label{prop_hadamard}
    Let $\mu,\nu\in\mathcal{P}(\mathbb{R}^d)$.
    
    \begin{enumerate}
        \item The problem \eqref{HadamardWasserstein} always admits a minimizer.
        \item $\mathcal{HW}$ is a pseudometric (i.e.,~it is symmetric, non-negative, $\mathcal{HW}(\mu,\mu)=0$, and it satisfies the triangle inequality).
        \item $\mathcal{HW}$ is invariant to reflection with respect to axes.
    \end{enumerate}
\end{proposition}

\begin{proof}
    See \Cref{proof:prop_hadamard}.
\end{proof}

$\mathcal{HW}$ loses some properties compared to $GW$. Indeed, it is only invariant with respect to axes, and it can only compare measures lying in the same Euclidean space in order for the distance to be well defined. Nonetheless, we show in the following that we can derive some links with triangular couplings in the same way as the Wasserstein distance with~KR.

Indeed, the~cost $L$ \eqref{loss_HW} is separable and reduces to the inner-GW loss in 1D, for~which we have a closed-form solution. We can therefore define a degenerated version of it:
\begin{equation} \label{degenerated_cost}
    \begin{aligned}
        \forall x,x',y,y'\in\mathbb{R}^d,\ L_t(x,x',y,y') &= \sum_{k=1}^d \Big( \prod_{i=1}^{k-1} \lambda_t^{(i)}\Big) (x_kx_k'-y_ky_k')^2 \\
        &= (x\odot x'-y\odot y')^T A_t(x\odot x'-y\odot y')
    \end{aligned}
\end{equation}
with $A_t=\mathrm{diag}(1,\lambda_t^{(1)},\lambda_t^{(1)}\lambda_t^{(2)}, \dots,\prod_{i=1}^{d-1}\lambda_t^{(i)})$, ~such as for all $t>0$, and~for all $i\in\{1,\dots,d-1\}$, $\lambda_t^{(i)}>0$, and $\lambda_t^{(i)}\xrightarrow[t\to 0]{}0$. We denote $\mathcal{HW}_t$ the problem \eqref{HadamardWasserstein} with the degenerated cost \eqref{degenerated_cost}. Therefore, we will be able to decompose the objective as: 
\begin{equation}
    \begin{aligned}
        \iint L_t(x,x',y,y')\ \mathrm{d}\gamma(x,y)\mathrm{d}\gamma(x',y') &= \iint (x_1x_1'-y_1y_1')^2\ \mathrm{d}\gamma(x,y)\mathrm{d}\gamma(x',y') \\ &+ \iint \sum_{k=2}^d \left(\prod_{i=1}^{k-1}\lambda_t^{(i)}\right)(x_kx_k'-y_ky_k')^2\ \mathrm{d}\gamma(x,y)\mathrm{d}\gamma(x',y')
    \end{aligned}
\end{equation}
and to use the same induction reasoning as~\citet{carlier2010knothe}.

Then, we can define a triangular coupling different from the Knothe--Rosenblatt rearrangement in the sense that each map will not be nondecreasing. Indeed, following \Cref{theorem:gw_ip_1d}, the~solution of each 1D problem: 
\begin{equation}
    \argmin_{\gamma\in\Pi(\mu,\nu)}\ \iint (xx'-yy')^2\ \mathrm{d}\gamma(x,y)\mathrm{d}\gamma(x',y')
\end{equation}
is either $(Id\times T_{\mathrm{asc}})_\#\mu$ or $(Id\times T_{\mathrm{desc}})_\#\mu$. Hence, at~each step $k\ge 1$, if~we disintegrate the joint law of the $k$ first variables as $\mu^{1:k}=\mu^{1:k-1}\otimes \mu^{k|1:k-1}$, the~optimal transport map $T(\cdot|x_1,\dots,x_{k-1})$ will be the solution of:
\begin{equation}
    \argmin_{T\in\{T_{\mathrm{asc}}, T_{\mathrm{desc}}\}}\ \iint \big(x_kx_k'-T(x_k)T(x_k')\big)^2\ \mu^{k\mid1:k-1}(\mathrm{d}x_k\mid x_{1:k-1})\mu^{k\mid1:k-1}(\mathrm{d}x_k'\mid x_{1:k-1}').
\end{equation}

We now state the main theorem, where we show that the limit of the OT plans obtained with the degenerated cost will be the triangular coupling we just~defined.

\begin{theorem} \label{ThGWKR}
    Let $\mu$ and $\nu$ be two absolutely continuous measures on $\mathbb{R}^d$ such that $\int\|x\|_2^4\ \mu(\mathrm{d}x) < +\infty$, $\int \|y\|_2^4\ \nu(\mathrm{d}y)<+\infty$ and with compact support. Let $\gamma_t$ be an optimal transport plan for $\mathcal{HW}_t$, let $T_K$ be the alternate Knothe--Rosenblatt map between $\mu$ and $\nu$ as defined in the last paragraph, and~let $\gamma_K=(Id\times T_K)_\#\mu$ be the associated transport plan. Then, we have $\gamma_t\xrightarrow[t\to 0]{\mathcal{D}}\gamma_K$. Moreover, if~$\gamma_t$ are induced by transport maps $T_t$, then $T_t\xrightarrow[t\to 0]{L^2(\mu)} T_K$.
\end{theorem}

\begin{proof}
    See \Cref{proof:thgwkr}.
\end{proof}

However, we cannot extend this Theorem to the subspace detour approach. Indeed, by~choosing $A_t=V_EV_E^T+tV_{E^\bot}V_{E^\bot}^T$ with $(V_E,V_{E^\bot})$ an orthonormal basis of $\mathbb{R}^d$, then we project $x\odot x'-y\odot y'$ on $E$ (respectively on $E^\bot$), which is generally different from $x_E\odot x_E'-y_E\odot y_E'$ (respectively $x_{E^\bot}\odot x_{E^\bot}'-y_{E^\bot}\odot y_{E^\bot}'$).

\subsection{Solving Hadamard--Wasserstein in the Discrete~Setting}

In this part, we derive formulas to solve numerically $\mathcal{HW}$ \eqref{HadamardWasserstein}. Let $x_1,\dots,x_n\in\mathbb{R}^d$, $y_1,\dots,y_m\in\mathbb{R}^d$, $\alpha\in\Sigma_n$, $\beta\in\Sigma_m$, $p=\sum_{i=1}^n \alpha_i\delta_{x_i}$ and $q=\sum_{j=1}^m\beta_j\delta_{y_j}$ two discrete measures in $\mathbb{R}^d$. The~Hadamard Wasserstein problem \eqref{HadamardWasserstein} becomes in the discrete setting:
\begin{equation}
    \begin{aligned}
        \mathcal{HW}^2(p,q) &= \inf_{\gamma\in\Pi(p,q)}\ \sum_{i,j}\sum_{k,\ell} \|x_i\odot x_k - y_j\odot y_\ell\|_2^2\ \gamma_{i,j}\gamma_{k,\ell} \\
        &= \inf_{\gamma\in\Pi(p,q)}\ \mathcal{E}(\gamma)
    \end{aligned}
\end{equation}
with $\mathcal{E}(\gamma)=\sum_{i,j}\sum_{k,\ell} \|x_i\odot x_k - y_j\odot y_\ell\|_2^2\ \gamma_{i,j}\gamma_{k,\ell}$.
As denoted in~\citep{peyre2016gromov}, if~we note: 
\begin{equation}
    \mathcal{L}_{i,j,k,\ell}=\|x_i\odot x_k-y_j\odot y_\ell\|_2^2,    
\end{equation}
then we have:
\begin{equation}
     \mathcal{E}(\gamma) = \langle \mathcal{L}\otimes \gamma,\gamma\rangle,
\end{equation}
where $\otimes$ is defined as:
\begin{equation}
    \mathcal{L}\otimes \gamma = \Big( \sum_{k,\ell} \mathcal{L}_{i,j,k,\ell}\gamma_{k,\ell}\Big)_{i,j}\in\mathbb{R}^{n\times m}.
\end{equation}

We show in the next proposition a decomposition of $\mathcal{L}\otimes \gamma$, which allows us to compute this tensor product more~efficiently.

\begin{proposition} \label{formulaHW}
    Let $\gamma\in\Pi(p,q)=\{M\in(\mathbb{R}_+)^{n\times m},\ M\mathbb{1}_m=p,\ M^T\mathbb{1}_n=q\}$, where $\mathbb{1}_n = (1,\dots,1)^T \in\mathbb{R}^n$. Let us note $X=(x_i\odot x_k)_{i,k}\in\mathbb{R}^{n\times n\times d}$, $Y=(y_j\odot y_\ell)_{j,\ell}\in\mathbb{R}^{m\times m\times d}$, $X^{(2)}=(\|X_{i,k}\|_2^2)_{i,k}\in \mathbb{R}^{n\times n}$, $Y^{(2)}=(\|Y_{j,l}\|_2^2)_{j,l}\in\mathbb{R}^{m\times m}$, and $\forall t\in\{1,\dots,d\},\ X_t=(X_{i,k,t})_{i,k}\in\mathbb{R}^{n\times n}$ and $Y_t=(Y_{j,\ell,t})_{j,\ell}\in\mathbb{R}^{m\times m}$. Then:
    \begin{equation}
        \mathcal{L}\otimes \gamma = X^{(2)}p\mathbb{1}_m^T+\mathbb{1}_n q^T (Y^{(2)})^T-2\sum_{t=1}^d X_t\gamma Y_t^T.
    \end{equation}
\end{proposition}

\begin{proof}
    See \Cref{proof:formulaHW}.
\end{proof}

From this decomposition, we can compute the tensor product $\mathcal{L}\otimes \gamma$ with a complexity of $O(d(n^2m+m^2n))$ using only multiplications of matrices (instead of $O(dn^2m^2)$ for a naive computation).

\begin{remark}
    For the degenerated cost function \eqref{degenerated_cost}, we just need to replace $X$ and $Y$ by $\Tilde{X_t}=A_t^{\frac12}X$ and $\Tilde{Y_t}=A_t^{\frac12}Y$ in the previous proposition.
\end{remark}

To solve this problem numerically, we can use the conditional gradient algorithm \citep[Algorithm 2]{vayer2019optimal}. This algorithm only requires to compute the gradient: 
\begin{equation}
    \nabla\mathcal{E}(\gamma) = 2(A+B+C) = 2(\mathcal{L}\otimes \gamma)
\end{equation}
at each step and a classical OT problem. This algorithm is more efficient than solving the quadratic problem directly. Moreover, while it is a non-convex problem, it actually converges to a local stationary point \citep{lacoste2016convergence}.

\begin{figure}[t]
    \includegraphics[width=\linewidth]{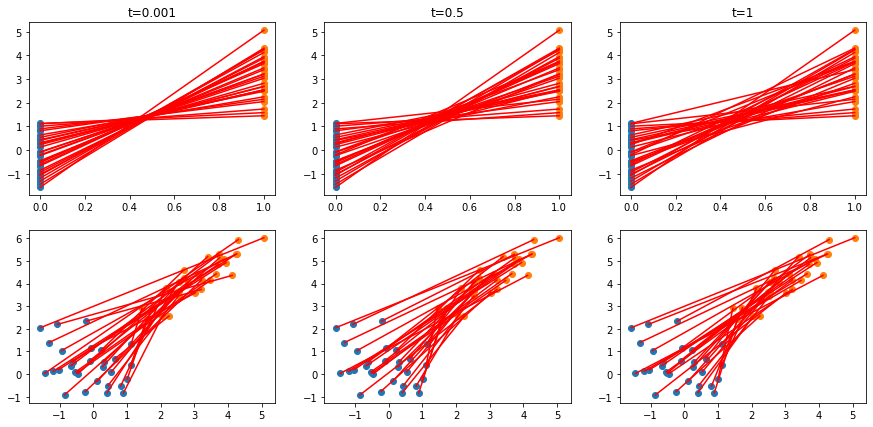}
    \caption{Degenerated coupling. On~the \textbf{first row}, the~points are projected on their first coordinate and we plot the optimal coupling. On~the \textbf{second row}, we plot the optimal coupling between the original points.}
    \label{fig:degnerated_coupling}
\end{figure}

On \Cref{fig:degnerated_coupling}, we generated 30 points of 2 Gaussian distributions, and~computed the optimal coupling of $\mathcal{HW}_t$ for several $t$. These points have the same uniform weight. We plot the couplings between the points on the second row, and~between the projected points on their first coordinate on the first row. 
Note that for discrete points, the~Knothe--Rosenblatt coupling amounts to sorting the points with respect to the first coordinate if there is no ambiguity (\emph{i.e.},~$x_1^{(1)}<\dots<x_n^{(1)}$) as it comes back to perform the Optimal Transport in one dimension \citep{peyre2019computational}~(Remark 2.28). For~our cost, the~optimal coupling in 1D can either be the increasing or the decreasing rearrangement. We observe on the first row of \Cref{fig:degnerated_coupling} that the optimal coupling when $t$ is close to 0 corresponds to the decreasing rearrangement, which corresponds well to the alternate Knothe--Rosenblatt map we defined in \Cref{section:properties_hgw}. It underlines the results provided in \Cref{ThGWKR}.

\section{Discussion}

We proposed in this work to extend the subspace detour approach to different subspaces, and~to other Optimal Transport costs such as Gromov--Wasserstein. Being able to project on different subspaces can be useful when the data are not aligned and do not share the same axes of interest, as~well as when we are working between different metric spaces as it is the case, for example, with graphs. However, a~question that arises is how to choose these subspaces. Since the method is mostly interesting when we choose one-dimensional subspaces, we proposed to use a PCA and to project on the first directions for data embedded in Euclidean spaces. For~more complicated data such as graphs, we projected onto the Fiedler vector and obtained good results in an efficient way on a 3D mesh registration problem. More generally, \citet{muzellec2019subspace} proposed to perform a gradient descent on the loss with respect to orthonormal matrices. This approach is non-convex and is only guaranteed to converge to a local minimum. Designing such an algorithm, which would minimize alternatively between two transformations in the Stiefel manifold, is left for future~works.

The subspace detour approach for transport problems is meaningful whenever one can identify subspaces that gather most of the information from the original distributions, while making the estimate more robust and with a better sample complexity as far as dimensions are lower. On~the computational complexity side, and~when we have only access to discrete data, the~subspace detour approach brings better computational complexity solely when the subspaces are chosen as one dimensional. Indeed, otherwise, we have the same complexity for solving the subspace detour and solving the OT problem directly (since the complexity only depends on the number of samples). 
In this case, the~1D projection often gives distinct values for all the samples (for continuous valued data) and hence the Monge--Knothe coupling is exactly the coupling in 1D. As~such, information is lost on the orthogonal spaces. It can be artificially recovered by quantizing the 1D values (as experimented in practice in \citep{muzellec2019subspace}), but~the added value is not clear and deserves broader studies. Absolutely continuous distributions \emph{w.r.t.} the Lebesgue measure being given, this limit however does not exist, but comes 
with the extra cost of being able to compute efficiently the projected measure onto the subspace, which might require discretization of the space and is therefore not practical in high~dimensions.

\looseness=-1 We also proposed a new quadratic cost $\mathcal{HW}$ that we call Hadamard--Wasserstein, which allows us to define a degenerated cost for which the optimal transport plan converges to a triangular coupling. However, this cost loses many properties compared to $W_2$ or $GW$, for~which we are inclined to use these problems. Indeed, while $\mathcal{HW}$ is a quadratic cost, it uses a Euclidean norm between the Hadamard product of vectors and requires the two spaces to be the same (in order to have the distance well defined). A~work around in the case $X=\mathbb{R}^p$ and $Y=\mathbb{R}^q$ with $p\le q$ would be to ``lift'' the vectors in $\mathbb{R}^p$ into vectors in $\mathbb{R}^q$ with padding as it is proposed in \citep{vayer2019sliced} or~to project the vectors in $\mathbb{R}^q$ on $\mathbb{R}^p$ as in \citep{cai2020distances}. Yet, for some applications where only the distance/similarity matrices are available, a~different strategy still needs to be found.
Another concern is the limited invariance properties (only with respect to axial symmetry symmetry in our case). 
Nevertheless, we expect that such a cost can be of interest in cases where invariance to symmetry is a desired property, such as in \citep{nagar2019detecting}.

\clearemptydoublepage
\cleartooddpage[\thispagestyle{empty}]
\bookmarksetup{startatroot} 

\chapter{Conclusion}

{
    \hypersetup{linkcolor=black} 
    \minitoc 
}

\chaptermark{Conclusion}

In this final chapter, we describe an overview of the contributions and discuss some perspectives which ensue from them.

\section{Contributions}

This thesis has focused on deriving efficient Optimal Transport methods based on projections on subspaces. In the first part, observing that many datasets have a representation on Riemannian manifolds, we defined new OT discrepancies on Riemannian manifolds by adapting the construction of the Euclidean Sliced-Wasserstein distance on such spaces. We first focused on Cartan-Hadamard manifolds on which we introduced two different Sliced-Wasserstein distances differing from their projection process onto the geodesics, and we provided some theoretical analysis of their properties. Then, we leveraged these two constructions and applied them to specific manifolds which have much interest in Machine Learning: Hyperbolic spaces and the space of Symmetric Positive Definite matrices (SPDs). First we demonstrated the computational efficiency compared to using more classical OT distances. Then, on Hyperbolic spaces, we compared the behavior of the two constructions on different tasks such as gradient flows or classification. On the space of SPDs, we used these new discrepancies on M/EEG data and performed brain-age prediction as well as domain adaptation for Brain Computer Interface data. We also studied the case of the hypersphere which is not a Cartan-Hadamard manifold and hence required a different strategy in order to define a SW distance on it. On this manifold, we applied SW to Wasserstein Autoencoders as well as density estimation tasks in order to show its benefit compared to just using the Euclidean SW between measures on the sphere embedded in the Euclidean space. 

As it can be beneficial for applications to use positive measures instead of probability measures, it motivated us to further define SW distances on positive measures. Thus, we introduced two new SW losses to compare efficiently positive measures and demonstrated their properties on document classification tasks as well as for computing barycenters of geoclimatic data. 

From another perspective, as SW is a real distance on the space of probability measures, it is possible to define gradient flows on this space with this distance \citep{ambrosio2008gradient}. We showed that it is indeed of interest from a computational point of view as the SW gradient flows of functionals can be more efficiently computed than when using the Wasserstein distance, while enjoying good empiric convergence properties. Moreover, we investigated empirically the underlying trajectory of the SW gradient flows and the popular conjecture that their trajectory is the same as Wasserstein gradient flows.

\looseness=-1 Besides studying the Sliced-Wasserstein distance, we have also been interested in the Busemann function which level sets provide generalizations of hyperplanes, and which has received much interest on certain Riemannian manifolds such as Hyperbolic spaces. Thus, it was fairly natural to study the Busemann function on the Wasserstein space. To do so, we first identified geodesics of the Wasserstein space for which the Busemann function is well defined when coupled with them. Then, we derived new closed-forms for the one dimensional case as well as the Gaussian case, making it possible to compute it in practice.
As a proof of concept, we proposed to use it in order to perform Principal Component Analysis on 1D measures.

\looseness=-1 Finally, we also studied the Gromov-Wasserstein distance which can be used to compare probability measures lying on incomparable spaces. While its sliced counterpart has been previously proposed by \citet{vayer2019sliced}, a major bottleneck of sliced methods is that they do not provide a coupling. Thus, we proposed to extend the subspace detour approach, first introduced by \citet{muzellec2019subspace} for the classical OT problem, to the Gromov-Wasserstein problem, and applied it on a shape matching problem.

\section{Perspectives}

The work done during this thesis can lead to different perspectives and open questions. We describe some of them in the following.

\paragraph{Sliced-Wasserstein on General Spaces.} Embedding data on Riemannian manifolds and then working directly on such space has become a prominent approach in Machine Learning. Thus, similarly as in the Euclidean space, we hope that the Sliced-Wasserstein distance on manifolds derived in this thesis will be used for ML tasks on such spaces, \emph{e.g.} as loss for Riemannian neural networks. This might require improving the expressive power of these distances, \emph{e.g.} by combining the original SW formulations with powerful ideas described in \Cref{section:variants_sw} to improve the Euclidean SW, for instance by changing the integration set, finding better estimators, projecting on higher-dimensional subspaces or on Hilbert curves adapted to Riemannian manifolds in a similar fashion as \citep{li2022hilbert}.

\looseness=-1 We focused in this work on specific manifolds to construct SW distances. But many different Riemannian manifolds have already been considered in ML, either to improve the quality of embeddings or to represent specific data structures. Some of them are Cartan-Hadamard manifolds, for which constructing SW distances could be done by following the framework proposed in \Cref{chapter:sw_hadamard}. For example, one might consider the space of SPDs with other metrics, such as more general pullback metrics \citep{chen2023adaptive}, for which, for $M\in S_d^{++}(\mathbb{R})$ and $A,B\in S_d(\mathbb{R})$, $g_M^\phi(A,B) = \langle \phi_{*,M}(A), \phi_{*,M}(B)\rangle_F$ where $\phi:S_d^{++}(\mathbb{R})\to S_d(\mathbb{R})$ is a diffeomorphism and $\phi_{*,M}$ the differential of $\phi$ at $M\in S_d^{++}(\mathbb{R})$. In this case, geodesic distances are of the form
\begin{equation}
    \forall X,Y\in S_d^{++}(\mathbb{R}),\ d_\phi(X,Y) = \|\phi(X)-\phi(Y)\|_F.
\end{equation}
Similarly as in the Log-Euclidean case (where $\phi=\log$), the space is of constant null curvature, and geodesic projections can be obtained as $P_\phi^A(M) = \langle A, \phi(M)\rangle_F$ for $A\in S_d(\mathbb{R})$ and $M\in S_d^{++}(\mathbb{R})$ (if we assume that $\phi(I_d)=0$ and that the differential at $I_d$ is the identity). Besides the Log-Euclidean distance, this framework includes the Log-Cholesky distance \citep{lin2019riemannian}, the $O(n)$-invariant Log-Euclidean metrics \citep{chen2023riemannian_multiclass} or the Adaptative Riemannian metric \citep{chen2023adaptive}. 
Another recent line of works consists of studying products of Riemannian manifolds which might be more flexible to embed data \citep{gu2019learning, Skopek2020Mixed-curvature, borde2023latent, lin2023hyperbolic}, as the resulting space is of non-constant curvature \citep{gu2019learning}. In particular, products of manifolds of non-positive curvature are still of non-positive curvatures \citep[Lemma 1]{gu2019learning}, and hence products of Cartan-Hadamard manifolds are still Cartan-Hadamard manifolds. For $\mathcal{M}=\mathcal{M}_1\times\mathcal{M}_2$,  \citet[Section II. 8.24]{bridson2013metric} provided the closed-form for the Busemann function associated to a geodesic ray $\gamma$ defined as $\gamma(t) = \big(\gamma_1(\cos(\theta) t), \gamma_2(\sin(\theta) t)\big)$ for $\gamma_1$ and $\gamma_2$ geodesic rays on $\mathcal{M}_1$ and $\mathcal{M}_2$ respectively, and $\theta\in ]0,\pi/2[$, as
\begin{equation}
    \forall x\in\mathcal{M},\ B^\gamma(x) = \cos(\theta) B^{\gamma_1}(x_1) + \sin(\theta) B^{\gamma_2}(x_2).
\end{equation}
This can be readily extended to $\mathcal{M}=\mathcal{M}_1\times\dots\times\mathcal{M}_n$, using $(\lambda_i)_{i=1}^n$ such that $\sum_{i=1}^n\lambda_i^2 = 1$ and a geodesic ray of the form $\gamma(t) = \big(\gamma_1(\lambda_1 t),\dots,\gamma_n(\lambda_n t)\big)$, as
\begin{equation}
    \forall x\in \mathcal{M},\ B^\gamma(x) = \sum_{i=1}^n \lambda_i B^{\gamma_i}(x_i).
\end{equation}
Another type of Riemannian manifolds with non-positive curvature are Siegel spaces \citep{nielsen2020siegel,cabanes2021classification,cabanes2022apprentissage}, which have recently received attention in ML \citep{lopez2021vector} for their capacity to leverage different curvatures.
It is also well known that one dimensional gaussians endowed with the Fisher information metric have a hyperbolic structure \citep{costa2015fisher}, and diagonal gaussians have the structure of a product of Hyperbolic spaces \citep{cho2023hyperbolic}. The space of parameters of Dirichlet distributions has also a Hadamard manifold structure \citep{le2021fisher}. Thus, developing sliced methods on parametric families of distribution might be possible through this framework.

Studying more complicated Riemannian manifolds such as tori, which have sections of positive, negative and null curvatures, or even more general closed manifolds as done in \citep{chen2023riemannian} is also an important avenue of research in order to be able to deal with \emph{e.g.} proteins \citep{huang2022riemannian, chen2023riemannian} or molecules \citep{jing2022torsional}.
Very recent works have also started to use more general spaces such as pseudo-Riemannian manifolds \citep{law2021ultrahyperbolic, xiong2022pseudoriemannian}, Finsler manifolds \citep{shen2001lectures, lopez2021symmetric} or more general metric spaces.
For example, CAT(0) spaces are metric spaces with non-positive curvature which have a structure very similar with Hadamard manifolds \citep{bridson2013metric} and which have recently received some attention in Optimal Transport \citep{berdellima2023existence}. \citet{lopez2021vector} proposed to endow the space of SPD matrices with vector-valued distance function, generalizing the Affine-Invariant distance, and allowing the use of Finsler metrics which are better suited to data structures such as graphs. In the same line of work, \citet{lopez2021symmetric} proposed to use Finsler metrics on the Siegel space and \citet{nielsen2022non} studied the Hilbert simplex which is a particular Finsler manifold \citep{troyanov2013funk}. Note that Finsler manifolds have also received attention in Optimal Transport \citep{ohta2010optimal, ohta2011displacement}.

However, extending SW to these different spaces might raise several challenges such as finding a meaningful set of curves on which to project the distributions or deriving efficient ways to project the distributions on the subspaces. Besides, it is important to study more closely the distance properties of the different SW distances introduced in this work in order to better justify theoretically their usefulness.

\paragraph{Gradient Flows.} 

A lot of open questions regarding Sliced-Wasserstein gradient flows still need to be handled such as the theoretical questions of convergences, and showing the links with Wasserstein gradient flows. Besides, this framework could be extended to other Sliced-Wasserstein distances such as Generalized SW versions \citep{kolouri2019generalized}, or Riemannian manifold versions derived in the first part of the thesis. Another direction would be to adapt the work of \citet{liutkus2019sliced} for Riemannian Sliced-Wasserstein distances in order to minimize these functionals using their Wasserstein gradient flows. A first step towards that direction has been made through \Cref{prop:chsw_1st_variation} in which the first variation of Cartan-Hadamard SW has been derived. This can be useful in order to derive the continuity equation of the underlying gradient flows, as well as practical algorithms for learning probability distributions on Riemannian manifolds through particle schemes, which would provide alternatives to MCMC algorithms such as the Riemannian Langevin algorithm \citep{girolami2011riemann, wang2020fast, gatmiry2022convergence}.

\paragraph{Unbalanced Sliced-Wasserstein.} 

In the thesis, we proposed two way of performing slicing with unbalanced OT. The first one consists of simply slicing the 1D UOT, and the second one consists of adding a regularization on the mass of the marginals. The second proposal has interesting properties as it allows to be more robust to outliers compared to the first one. However, \citet{leblanc2023extending} recently proposed a new OT distance between positive measures, which extends the Wasserstein distance in a proper way in the sense that its restriction to probability measures coincides with the Wasserstein distance, and geodesics between probability measures are well probability measures, which is not the case for UOT. This new OT loss between positive measures inherits many of the Wasserstein distance properties, but also its computational complexity and its statistical properties. Thus, it would be an interesting direction to derive its sliced version and to compare its properties with $\RSOT$ and $\SUOT$.

Another direction could be to study its gradient flows, either as a functional endowed with the Wasserstein-Fisher-Rao metric \citep{gallouet2017jko} or in a similar spirit of \Cref{chapter:swgf} by using the JKO scheme in the space of positive measures endowed by $\RSOT$ or $\SUOT$.

\paragraph{Busemann on Wasserstein Space.}

In our work, we only used the Busemann function to perform Principal Component Analysis in the one dimensional case where the geometry is flat and hence where the projections on the geodesics actually coincide with the geodesic projections. Thus, a natural next step is to study it in the Bures-Wasserstein space for Gaussians of higher dimension as we already have the closed-form for the Busemann function. 

An interesting direction would be to provide closed-forms in more general cases, or on the restriction on other classes of distributions, for example on Gaussian Mixture Models using the distance introduced by \citet{delon2020wasserstein} or by \citet{dusson2023wasserstein}. It would also be natural to study the case of positive measures using  either the Wasserstein distance on positive measures presented in \citep{leblanc2023extending} or Unbalanced OT distances \citep{sejourne2022unbalanced}, \emph{e.g.} relying on available closed-forms for Gaussians \citep{janati2020entropic}. Another direction to have closed-forms for arbitrary probability distributions would be to develop and study a sliced version, where \emph{e.g.} for $t\mapsto \mu_t$ a geodesic ray in $\mathcal{P}_2(\mathbb{R})$ and $\nu\in\mathcal{P}_2(\mathbb{R}^d)$, the Sliced-Busemann function would be defined as
\begin{equation}
    SB^\mu(\nu) = \int_{S^{d-1}} B^\mu(P^\theta_\#\nu)\ \mathrm{d}\lambda(\theta).
\end{equation}
Then studying the properties of this object, and how it differs from the regular Busemann function would be a natural avenue of research. 

Nonetheless, we note that despite the interesting theoretical properties, using the Busemann function to perform PCA on Wasserstein space does not seem very promising as the projections can be potentially out of the geodesic. Thus, finding an application for which it would be well suited, might be important to justify further studies.



\clearemptydoublepage
\cleartooddpage[\thispagestyle{empty}]
\chapter{Appendix} \label{chapter:appendix}

{
    \hypersetup{linkcolor=black} 
    \minitoc 
}

\section{Appendix of \Cref{chapter:sw_hadamard}}

\subsection{Lemmas}

We derive here some lemmas which will be useful for the proofs.

\begin{lemma}[Lemma 6 in \citep{paty2019subspace}] \label{lemma:paty}
    Let $\mathcal{M}$, $\mathcal{N}$ be two Riemannian manifolds. Let $f:\mathcal{M}\to \mathcal{N}$ be a measurable map and $\mu,\nu \in \mathcal{P}(\mathcal{M})$. Then,
    \begin{equation}
        \Pi(f_\#\mu,f_\#\nu) = \{(f\otimes f)_\#\gamma,\ \gamma\in\Pi(\mu,\nu)\}.
    \end{equation}
\end{lemma}

\begin{proof}
    This is a straightforward extension of \citep[Lemma 6]{paty2019subspace}.
\end{proof}

\begin{lemma} \label{lemma:lipschitz}
    Let $(\mathcal{M}, g)$ be a Hadamard manifold with origin $o$. Let $v\in T_o\mathcal{M}$, then
    \begin{enumerate}
        \item the geodesic projection $P^v$ is 1-Lipschitz.
        \item the Busemann function $B^v$ is 1-Lipschitz.
    \end{enumerate}
\end{lemma}

\begin{proof} \leavevmode
    \begin{enumerate}
        \item By \Cref{prop:isometry}, we know that
        \begin{equation}
            \forall x, y\in \mathcal{M},\ |P^v(x)-P^v(y)| = d\big(\Tilde{P}^v(x),\Tilde{P}^v(y)\big).
        \end{equation}
        Moreover, by \citep[Page 9]{ballmann2006manifolds}, $\Tilde{P}^v$ is 1-Lipschitz, so is $P^v$.
        \item The Busemann function is 1-Lipschitz, see \emph{e.g.} \citep[II. Proposition 8.22]{bridson2013metric}.
    \end{enumerate}
\end{proof}

\begin{lemma} \label{lemma:inequality_distance}
    Let $d$ be a metric on $\mathcal{M}$. Then, for any $p\ge1$,
    \begin{equation}
        \forall x,y \in\mathcal{M},\ d(x,y)^p \le 2^{p-1} \big( d(x,o)^p + d(o,y)^p\big).
    \end{equation}
\end{lemma}

\begin{lemma}[Lemma 1 in \citep{rakotomamonjy2021statistical} adapted from Theorem 2 in \citep{fournier2015rate}] \label{lemma:fournier}
    Let $p\ge 1$ and $\eta\in\mathcal{P}_p(\mathbb{R})$. Denote $\Tilde{M}_q(\eta)=\int |x|^q\ \mathrm{d}\eta(x)$ the moments of order $q$ and assume that $M_q(\eta)<\infty$ for some $q>p$. Then, there exists a constant $C_{p,q}$ depending only on $p,q$ such that for all $n\ge 1$,
    \begin{equation}
        \mathbb{E}[W_p^p(\hat{\eta}_n,\eta)] \le C_{p,q} \Tilde{M}_q(\eta)^{p/q}\left(n^{-1/2}\mathbb{1}_{\{q>2p\}} + n^{-1/2}\log(n) \mathbb{1}_{\{q=2p\}} + n^{-(q-p)/q} \mathbb{1}_{\{q\in(p,2p)\}}\right).
    \end{equation}
\end{lemma}


\begin{lemma} \label{lemma:derivative_geodesic_dist}
    Let $y\in \mathcal{M}$ and denote for all $x\in\mathcal{M}$, $f(x) = d(x,y)^2$. Then, $\mathrm{grad}_{\mathcal{M}} f(x) = -2\log_x(y)$.
\end{lemma}

For references about \Cref{lemma:derivative_geodesic_dist}, see \emph{e.g.} \citep[Appendix A]{chewi2020gradient} or \citep{goto2021approximated}.

\subsection{Proofs of \Cref{section:irsw}}

\subsubsection{Proof of \Cref{prop:isometry}} \label{proof:prop_isometry}

\begin{proof}[Proof of \Cref{prop:isometry}] 
    Let $x,y\in \mathcal{G}^v$. Then, there exists $s,t\in\mathbb{R}$ such that $x=\exp_o(sv)$ and $y=\exp_o(tv)$. By a simple calculation, we have on one hand that
    \begin{equation}
        \begin{aligned}
            \sign(\langle \log_o(x),v\rangle_o) &= \sign(\langle \log_o(\exp_o(sv)), v\rangle_o) \\
            &= \sign(s \|v\|_o^2) \\
            &= \sign(s),
        \end{aligned}
    \end{equation}
    using that $\log_o\circ \exp_o = \id$. And similarly, $\sign(\langle \log_o(y), v\rangle_o) = \sign(t)$.
    
    Then, by noting that $o=\exp_o(0)$, and recalling that $d(x,y) = d(\exp_o(tv), \exp_o(sv)) = |t-s|$,
    \begin{equation}
        \begin{aligned}
            |t^v(x)-t^v(y)| &= |\sign(\langle \log_o(x), v\rangle_o) d(x,o) - \sign(\langle \log_o(y), v\rangle_o d(y,o)| \\
            &= \big|\sign(s) d(\exp_o(sv),\exp_o(0)) - \sign(t) d(\exp_o(tv), \exp_o(0))\big| \\
            &= \big| \sign(s) |s| - \sign(t) |t|\big| \\
            &= |s-t| \\
            &= d(x,y).
        \end{aligned}    
    \end{equation}
\end{proof}

\subsubsection{Proof of \Cref{prop:charac_geod_proj}} \label{proof:prop_charac_geod_proj}

\begin{proof}[Proof of \Cref{prop:charac_geod_proj}]
    We want to solve:
    \begin{equation}
        P^v(x) = \argmin_{t\in\mathbb{R}}\ d\big(\gamma(t), x\big)^2,
    \end{equation}
    where $\gamma(t) = \exp_o(tv)$.
    For $t\in\mathbb{R}$, let $g(t) = d\big(\gamma(t), x\big)^2 = f\big(\gamma(t)\big)$ where $f(x) = d(x,y)^2$ for $x,y\in\mathcal{M}$. Then, by \Cref{lemma:derivative_geodesic_dist}, we have for any $t\in\mathbb{R}$,
    \begin{equation}
        \begin{aligned}
            g'(t) = 0 &\iff \langle \gamma'(t), \mathrm{grad}_\mathcal{M} f\big(\gamma(t)\big)\rangle_{\gamma(t)} = 0 \\
            &\iff \langle \gamma'(t), -2\log_{\gamma(t)}(x)\rangle_{\gamma(t)} = 0.
        \end{aligned}
    \end{equation}
\end{proof}

\subsubsection{Proof of \Cref{prop:eq_wasserstein}} \label{proof:prop_eq_wasserstein}

\begin{proof}[Proof of \Cref{prop:eq_wasserstein}]
    First, we note that $P^v = t^v\circ\Tilde{P}^v$. Then, by using \Cref{lemma:paty} which states that $\Pi(f_\#\mu, f_\#\nu)=\{(f\otimes f)_\#\gamma,\ \gamma\in\Pi(\mu,\nu)\}$ for any $f$ measurable, as well as that by \Cref{prop:isometry}, $|t^v(x)-t^v(y)| = d(x,y)$, we have:
    \begin{equation}
        \begin{aligned}
            W_p^p(P^v_\#\mu,P^v_\#\nu) &= \inf_{\gamma\in \Pi(P^v_\#\mu,P^v_\#\nu)}\ \int_{\mathbb{R}\times \mathbb{R}} |x-y|^p\ \mathrm{d}\gamma(x,y) \\
            &= \inf_{\gamma\in \Pi(\mu,\nu)}\ \int_{\mathbb{R}\times \mathbb{R}} |x-y|^p \ \mathrm{d}(P^v\otimes P^v)_\#\gamma(x,y) \\
            &= \inf_{\gamma\in \Pi(\mu,\nu)}\ \int_{\mathcal{M}\times \mathcal{M}}\ |P^v(x)-P^v(y)|^p\ \mathrm{d}\gamma(x,y) \\
            &= \inf_{\gamma\in \Pi(\mu,\nu)}\ \int_{\mathcal{M} \times \mathcal{M}} |t^v(\Tilde{P}^v(x))-t^v(\Tilde{P}^v(y))|^p\ \mathrm{d}\gamma(x,y) \\
            &= \inf_{\gamma\in \Pi(\mu,\nu)}\ \int_{\mathcal{M} \times \mathcal{M}} d\big(\Tilde{P}^v(x), \Tilde{P}^v(y)\big)^p\ \mathrm{d}\gamma(x,y) \\
            &= \inf_{\gamma\in \Pi(\mu,\nu)}\ \int_{\mathcal{M} \times \mathcal{M}} \ d(x, y)^p\ \mathrm{d}(\Tilde{P}^v\otimes \Tilde{P}^v)_\#\gamma(x,y) \\
            &= \inf_{\gamma\in\Pi(\Tilde{P}^v_\#\mu, \Tilde{P}^v_\#\nu)}\ \int_{\mathcal{G}^v\times \mathcal{G}^v} d(x,y)^p\ \mathrm{d}\gamma(x,y) \\
            &= W_p^p(\Tilde{P}^v_\#\mu, \Tilde{P}^v_\#\nu).
        \end{aligned}
    \end{equation}
\end{proof}

\subsubsection{Proof of \Cref{prop:eq_wasserstein_busemann}} \label{proof:prop_eq_wasserstein_busemann}

\begin{proof}[Proof of \Cref{prop:eq_wasserstein_busemann}]
    First, let us compute $t^v\circ \Tilde{B}^v$:
    \begin{equation}
        \begin{aligned}
            \forall x\in \mathcal{M},\ t^v(\Tilde{B}^v(x)) &= \sign(\langle \log_o(\Tilde{B}^v(x)), v \rangle_o) d(\Tilde{B}^v(x), o) \\
            &= \sign(-B^\gamma(x) \|v\|_o^2) d(\exp_o(-B^v(x)v), \exp_o(0)) \\
            &= \sign(-B^\gamma(x)) |-B^v(x)| \\
            &= - B^v(x).
        \end{aligned}
    \end{equation}
    Then, using the same computation as in the proof of \Cref{prop:eq_wasserstein}, we get
    \begin{equation}
        W_p^p(B^v_\#\mu, B^v_\#\nu) = W_p^p(\Tilde{B}^v_\#\mu, \Tilde{B}^v_\#\nu).
    \end{equation}
\end{proof}

\subsection{Proofs of \Cref{section:chsw_properties}}

\subsubsection{Proof of \Cref{prop:chsw_pseudo_distance}} \label{proof:prop_chsw_pseudo_distance}

\begin{proof}[Proof of \Cref{prop:chsw_pseudo_distance}]
    First, we will show that for any $\mu,\nu\in\mathcal{P}_p(\mathcal{M})$, $\chsw_p(\mu,\nu) < \infty$. Let $\mu,\nu \in \mathcal{P}_p(\mathcal{M})$, and let $\gamma\in \Pi(\mu,\nu)$ be an arbitrary coupling between them.
    Then by using first \Cref{lemma:paty} followed by the 1-Lipschitzness of the projections \Cref{lemma:lipschitz} and \Cref{lemma:inequality_distance}, we obtain
    \begin{equation}
        \begin{aligned}
            W_p^p(P^v_\#\mu, P^v_\#\nu) &= \inf_{\gamma\in\Pi(\mu,\nu)}\ \int |P^v(x)-P^v(y)|^p\ \mathrm{d}\gamma(x,y) \\
            &\le \int |P^v(x)-P^v(y)|^p\ \mathrm{d}\gamma(x,y) \\
            &\le \int d(x,y)^p\ \mathrm{d}\gamma(x,y) \\
            &\le 2^{p-1} \left(\int d(x,o)^p\ \mathrm{d}\mu(x) + \int d(o, y)^p\ \mathrm{d}\nu(y)\right) \\
            &< \infty.
        \end{aligned}
    \end{equation}
    Hence, we can conclude that $\chsw_p^p(\mu,\nu) < \infty$.

    Now, let us show that it is a pseudo-distance. First, it is straightforward to see that $\chsw_p(\mu,\nu)\ge 0$, that it is symmetric, \emph{i.e.} $\chsw_p(\mu,\nu)=\chsw_p(\nu,\mu)$, and that $\mu=\nu$ implies that $\chsw_p(\mu,\nu)=0$ using that $W_p$ is well a distance.

    For the triangular inequality, we can derive it using the triangular inequality for $W_p$ and the Minkowski inequality. Let $\mu,\nu,\alpha\in\mathcal{P}_p(\mathcal{M})$,
    \begin{equation}
        \begin{aligned}
            \chsw_p(\mu,\nu) &= \left(\int_{S_o} W_p^p(P^v_\#\mu, P^v_\#\nu)\ \mathrm{d}\lambda(v)\right)^{\frac{1}{p}} \\
            &\le \left(\int_{S_o} \big(W_p(P^v_\#\mu, P^v_\#\alpha) + W_p(P^v_\#\alpha, P^v_\#\nu)\big)^p\ \mathrm{d}\lambda(v)\right)^{\frac{1}{p}} \\ 
            &\le \left(\int_{S_o} W_p^p(P^v_\#\mu, P^v_\#\alpha)\ \mathrm{d}\lambda(v)\right)^{\frac{1}{p}} + \left(\int_{S_o} W_p^p(P^v_\#\alpha, P^v_\#\nu)\ \mathrm{d}\lambda(v)\right)^{\frac{1}{p}} \\
            &= \chsw_p(\mu,\alpha) + \chsw_p(\alpha,\nu).
        \end{aligned}
    \end{equation}
\end{proof}

\subsubsection{Proof of \Cref{prop:chsw_dual_rt}} \label{proof:prop_chsw_dual_rt}

\begin{proof}[Proof of \Cref{prop:chsw_dual_rt}]
    Let $f\in L^1(\mathcal{M})$, $g\in C_0(\mathbb{R}\times S_o)$, then by Fubini's theorem,
    \begin{equation}
        \begin{aligned}
            \langle \chr f, g\rangle_{\mathbb{R} \times S_o} &= \int_{S_o}\int_{\mathbb{R}} \chr f(t, v) g(t, v) \ \mathrm{d}t\mathrm{d}\lambda(v) \\
            &= \int_{S_o}\int_{\mathbb{R}} \int_{\mathcal{M}}f(x) \mathbb{1}_{\{t= P^v(x)\}} g(t,v) \ \mathrm{d}\vol(x)\mathrm{d}t\mathrm{d}\lambda(v) \\
            &= \int_{\mathcal{M}} f(x) \int_{S_o}\int_{\mathbb{R}} g(t,v)\mathbb{1}_{\{t= P^v(x)\}}\ \mathrm{d}t\mathrm{d}\lambda(v)\mathrm{d}\vol(x) \\
            &= \int_{\mathcal{M}} f(x) \int_{S_o} g\big(P^v(x), v\big)\ \mathrm{d}\lambda(v)\mathrm{d}\vol(x) \\
            &= \int_{\mathcal{M}} f(x) \chr^*g(x)\ \mathrm{d}\vol(x) \\
            &= \langle f, \chr^*g\rangle_{\mathcal{M}}.
        \end{aligned}
    \end{equation}
\end{proof}

\subsubsection{Proof of \Cref{prop:chr_vanish}} \label{proof:prop_chr_vanish}

\begin{proof}[Proof of \Cref{prop:chr_vanish}]
    We follow the proof of \citep[Lemma 1]{boman2009support}. On one hand, $g\in C_0(\mathbb{R}\times S_o)$, thus for all $\epsilon>0$, there exists $M>0$ such that $|t|\ge M$ implies $|g(t,v)|\le \epsilon$ for all $v\in S_o$.

    Let $\epsilon>0$ and $M>0$ which satisfies the previous property. Denote $E(x,M) = \{v\in S_o,\ |P^v(x)|<M\}$. Then, as $d(x,o)>0$, we have
    \begin{equation}
        E(x,M) = \{v\in S_o,\ |P^v(x)|<M\} = \left\{v\in S_p,\ \frac{P^v(x)}{d(x,o)} < \frac{M}{d(x,o)}\right\} \xrightarrow[d(x,o)\to\infty]{} \emptyset.
    \end{equation}
    Thus, $\lambda\big(E(x,M)\big) \xrightarrow[d(x,o)\to\infty]{} 0$. Choose $M'$ such that $d(x,o)>M'$ implies that $\lambda\big(E(x,M)\big) < \epsilon$.

    Then, for $x\in \mathcal{M}$ such that $|P^v(x)| \ge \max(M,M')$ (and thus $d(x,o)\ge M'$ since $|P^v(x)\le d(x,o)$ as $P^v$ is Lipschitz,
    \begin{equation}
        \begin{aligned}
            |\chr^*g(x)| &\le \left| \int_{E(x,M)} g(P^v(x),v)\ \mathrm{d}\lambda(v) \right| + \left| \int_{E(x,M)^c} g(P^v(x),v)\ \mathrm{d}\lambda(v) \right| \\
            &\le \|g\|_\infty\ \lambda\big(E(x,M)\big)  + \epsilon \lambda\big(E(x,M)^c\big) \\
            &\le \|g\|_\infty \epsilon + \epsilon.
        \end{aligned}
    \end{equation}
    Thus, we showed that $\chr^*g(x) \xrightarrow[d(x,o)\to\infty]{} 0$, and thus $\chr^*g\in C_0(\mathcal{M})$.
\end{proof}

\subsubsection{Proof of \Cref{prop:chsw_disintegration}} \label{proof:prop_chsw_disintegration}

\begin{proof}[Proof of \Cref{prop:chsw_disintegration}]
    Let $g\in C_0(\mathbb{R} \times S_o)$, as $\chr\mu=\lambda\otimes K$, we have by definition
    \begin{equation}
        \int_{S_o} \int_{\mathbb{R}} g(t, v) \ K(v,\cdot)_\#\mu(\mathrm{d}t)\ \mathrm{d}\lambda(v) = \int_{\mathbb{R} \times S_o} g(t,v)\ \mathrm{d}(\chr\mu)(t,v).
    \end{equation}
    Hence, using the property of the dual, we have for all $g\in C_o(\mathbb{R}\times S_o)$,
    \begin{equation}
        \begin{aligned}
            \int_{S_o} \int_{\mathbb{R}} g(t, v) \ K(v,\cdot)_\#\mu(\mathrm{d}t)\ \mathrm{d}\lambda(v) &= \int_{\mathbb{R} \times S_o} g(t,v)\ \mathrm{d}(\chr\mu)(t,v) \\
            &= \int_{\mathcal{M}} \chr^*g(x) \ \mathrm{d}\mu(x) \\
            &= \int_{\mathcal{M}} \int_{S_o} g(P^v(x), v)\ \mathrm{d}\lambda(v) \mathrm{d}\mu(x) \\
            &= \int_{S_o} \int_{\mathcal{M}} g(P^v(x), v) \ \mathrm{d}\mu(x) \mathrm{d}\lambda(v) \\
            &= \int_{S_o} \int_{\mathbb{R}} g(t, v) \ \mathrm{d}(P^v_\#\mu)(t)\mathrm{d}\lambda(v).
        \end{aligned}
    \end{equation}
    Hence, for $\lambda$-almost every $v\in S_o$, $K(v,\cdot)_\#\mu = P^v_\#\mu$.
\end{proof}

\subsubsection{Proof of \Cref{prop:chsw_upperbound}} \label{proof:prop_chsw_upperbound}

\begin{proof}[Proof of \Cref{prop:chsw_upperbound}]
    Using \Cref{lemma:paty} and that the projections are 1-Lipschitz (\Cref{lemma:lipschitz}), we can show that, for any $\mu,\nu\in\mathcal{P}_p(\mathcal{M})$,
    \begin{equation}
        \begin{aligned}
            \chsw_p^p(\mu,\nu) &= \inf_{\gamma\in \Pi(\mu,\nu)}\ \int |P^v(x)-P^v(y)|^p\ \mathrm{d}\gamma(x,y).
        \end{aligned}
    \end{equation}
    Let $\gamma^*\in \Pi(\mu,\nu)$ being an optimal coupling for the Wasserstein distance with ground cost $d$, then,
    \begin{equation}
        \begin{aligned}
            \chsw_p^p(\mu,\nu) &\le \int |P^v(x)-P^v(y)|^p\ \mathrm{d}\gamma^*(x,y) \\
            &\le \int d(x,y)^p\ \mathrm{d}\gamma^*(x,y) \\
            &= W_p^p(\mu,\nu).
        \end{aligned}
    \end{equation}
\end{proof}

\subsubsection{Proof of \Cref{prop:chsw_1st_variation}} \label{proof:prop_chsw_1st_variation}

\begin{proof}[Proof of \Cref{prop:chsw_1st_variation}]
    This proof follows the proof in the Euclidean case derived in \citep[Proposition 5.1.7]{bonnotte2013unidimensional} or in \citep[Proposition 1.33]{candau_tilh}.

    As $\mu$ is absolutely continuous, $P^v_\#\mu$ is also absolutely continuous and there is a Kantorovitch potential $\psi_v$ between $P^v_\#\mu$ and $P^v_\#\nu$. Moreover, as the support is restricted to a compact, it is Lipschitz and thus differentiable almost everywhere.
    
    First, using the duality formula, we obtain the following lower bound for all $\epsilon>0$,
    \begin{equation}
        \frac{\chsw_2^2\big((T_\epsilon)_\#\mu,\nu\big) - \chsw_2^2(\mu,\nu)}{2\epsilon} \ge \int_{S_o} \int_{\mathcal{M}} \frac{\psi_v(P^v(T_\epsilon(x))) - \psi_v(P^v(x))}{\epsilon}\ \mathrm{d}\mu(x)\mathrm{d}\lambda(v).
    \end{equation}
    Then, we know that the exponential map satisfies $\exp_x(0)=x$ and $\frac{\mathrm{d}}{\mathrm{d}t}\exp(tv)|_{t=0} = v$. Taking the limit $\epsilon\to 0$, the right term is equal to $\frac{\mathrm{d}}{\mathrm{d}t}g(t)|_{t=0}$ with $g(t) = \psi_v(P^v(T_t(x)))$ and is equal to
    \begin{equation}
        \frac{\mathrm{d}}{\mathrm{d}t}g(t)|_{t=0} = \psi_v'(P^v(T_0(x))) \langle \nabla P^v(T_0(x)), \frac{\mathrm{d}}{\mathrm{d}t}T_t(x)|_{t=0}\rangle_x = \psi_v'(P^v(x))\langle \mathrm{grad}_{\mathcal{M}} P^v(x),\xi(x)\rangle_x.
    \end{equation}
    Therefore, by the Lebesgue dominated convergence theorem (we have the convergence $\lambda$-almost surely and $|\psi_v(P^v(T_\epsilon(x)))-\psi_v(P^v(x))| \le \epsilon$ using that $\psi_v$ and $P^v$ are Lipschitz and that $d\big(\exp_x(\epsilon\xi(x)), \exp_x(0)\big) \le C\epsilon$),
    \begin{equation}
        \liminf_{\epsilon\to 0^+}\ \frac{\chsw_2^2\big((T_\epsilon)_\#\mu,\nu\big)-\chsw_2^2(\mu,\nu)}{2\epsilon} \ge \int_{S_o} \int_{\mathcal{M}} \psi_v'(P^v(x))\langle \mathrm{grad}_\mathcal{M} P^v(x), \xi(x)\rangle\ \mathrm{d}\mu(x)\mathrm{d}\lambda(v).
    \end{equation}

    For the upper bound, first, let $\pi^v\in\Pi(\mu,\nu)$ an optimal coupling. Then by \Cref{lemma:paty}, $\Tilde{\pi}^v = (P^v\otimes P^v)_\#\pi^v\in \Pi(P^v_\#\mu, P^v_\#\nu)$ is an optimal coupling for the regular quadratic cost and for $\Tilde{\pi}^v$-almost every $(x,y)$, $y=x-\psi_v'(x)$ and thus for $\pi^v$-almost every $(x,y)$, $P^v(y) = P^v(x) - \psi_v'\big(P^v(x)\big)$. Therefore,
    \begin{equation}
        \begin{aligned}
            \chsw_2^2(\mu,\nu) &= \int_{S_o} W_2^2(P^v_\#\mu, P^v_\#\nu)\ \mathrm{d}\lambda(v) \\
            &= \int_{S_o} \int_{\mathbb{R}\times\mathbb{R}} |x-y|^2 \ \mathrm{d}\Tilde{\pi}^v(x,y)\ \mathrm{d}\lambda(v) \\
            &= \int_{S_o} \int_{\mathcal{M}\times\mathcal{M}} |P^v(x) - P^v(y) |^2\ \mathrm{d}\pi(x,y)\ \mathrm{d}\lambda(v).
        \end{aligned}
    \end{equation}
    On the other hand, $((P^v\circ T_\epsilon)\otimes P^v)_\#\pi^v\in\Pi(P^v_\#(T_\epsilon)_\#\mu, P^v_\#\nu)$ and hence
    \begin{equation}
        \begin{aligned}
            \chsw_2^2\big((T_\epsilon)_\#\mu,\nu\big) &= \int_{S_o} W_2^2(P^v_\#(T_\epsilon)_\#\mu, P^v_\#\nu)\ \mathrm{d}\lambda(v) \\
            &\le \int_{S_o}\int_{\mathbb{R}\times\mathbb{R}} |P^v(T_\epsilon(x))-P^v(y)|^2\ \mathrm{d}\pi^v(x,y)\ \mathrm{d}\lambda(v).
        \end{aligned}
    \end{equation}
    Therefore,
    \begin{equation}
        \begin{aligned}
            \frac{\chsw_2^2\big((T_\epsilon)_\#\mu,\nu\big) - \chsw_2^2(\mu,\nu)}{2\epsilon} \le \int_{S_o} \int_{\mathbb{R}\times \mathbb{R}} \frac{|P^v(T_\epsilon(x))- P^v(y)|^2-|P^v(x)-P^v(y)|^2}{2\epsilon} \ \mathrm{d}\pi^v(x,y)\ \mathrm{d}\lambda(v).
        \end{aligned}
    \end{equation}
    Note $g(\epsilon) = \big(P^v(T_\epsilon(x)) - P^v(y)\big)^2$. Then, $\frac{\mathrm{d}}{\mathrm{d}\epsilon}g(\epsilon)|_{\epsilon=0} = 2\big(P^v(x)-P^v(y)\big) \langle \mathrm{grad}_\mathcal{M} P^v(x), \xi(x)\rangle_x$. But, as for $\pi^v$-almost every $(x,y)$, $P^v(y) = P^v(x) - \psi_v'(P^v(x))$, we have
    \begin{equation}
        \frac{\mathrm{d}}{\mathrm{d}\epsilon}g(\epsilon)|_{\epsilon=0} = 2 \psi_v'\big(P^v(x)\big) \langle\mathrm{grad}_{\mathcal{M}}P^v(x), \xi(x)\rangle_x.
    \end{equation}
    Finally, by the Lebesgue dominated convergence theorem, we obtain
    \begin{equation}
        \limsup_{\epsilon\to 0^+}\frac{\chsw_2^2\big((T_\epsilon)_\#\mu,\nu\big) - \chsw_2^2(\mu,\nu)}{2\epsilon} \le \int_{S_o} \int_{\mathcal{M}} \psi_v'(P^v(x))\langle \mathrm{grad}_\mathcal{M} P^v(x), \xi(x)\rangle_x\ \mathrm{d}\mu(x)\mathrm{d}\lambda(v).
    \end{equation}
\end{proof}

\subsubsection{Proof of \Cref{prop:chsw_hilbertian}} \label{proof:prop_chsw_hilbertian}

\begin{proof}[Proof of \Cref{prop:chsw_hilbertian}]
    Let $\mu,\nu\in\mathcal{P}_p(\mathcal{M})$, then
    \begin{equation}
        \begin{aligned}
            \chsw_p^p(\mu,\nu) &= \int_{S_o} W_p^p(P^v_\#\mu, P^v_\#\nu)\ \mathrm{d}\lambda(v) \\
            &= \int_{S_o} \|F_{P^v_\#\mu}^{-1} - F_{P^v_\#\nu}^{-1}\|^p_{L^p([0,1])}\ \mathrm{d}\lambda(v) \\
            &= \int_{S_o} \int_0^1 \big( F_{P^v_\#\mu}^{-1}(q) - F_{P^v_\#\nu}^{-1}(q)\big)^p\ \mathrm{d}q\ \mathrm{d}\lambda(v) \\
            &= \|\Phi(\mu)-\Phi(\nu)\|_{\mathcal{H}}^p.
        \end{aligned}
    \end{equation}
    Thus, $\chsw_p$ is Hilbertian.
\end{proof}

\subsubsection{Proof of \Cref{prop:chsw_sample_complexity}} \label{proof:prop_chsw_sample_complexity}

\begin{proof}[Proof of \Cref{prop:chsw_sample_complexity}]
    First, using the triangular inequality, the reverse triangular inequality and the Jensen inequality for $x\mapsto x^{1/p}$ (which is concave since $p\ge 1$),  we have the following inequality

    \begin{equation}
        \begin{aligned}
            \mathbb{E}[|\chsw_p(\hat{\mu}_n,\hat{\nu}_n) - \chsw_p(\mu,\nu)|] &= \mathbb{E}[|\chsw_p(\hat{\mu}_n,\hat{\nu}_n) - \chsw_p(\hat{\mu}_n,\nu) + \chsw_p(\hat{\mu}_n,\nu) - \chsw_p(\mu,\nu)|] \\
            &\le \mathbb{E}[|\chsw_p(\hat{\mu}_n,\hat{\nu}_n) - \chsw_p(\hat{\mu}_n,\nu)|] + \mathbb{E}[|\chsw_p(\hat{\mu}_n,\nu) - \chsw_p(\mu,\nu)|] \\
            &\le \mathbb{E}[\chsw_p(\nu,\hat{\nu}_n)] + \mathbb{E}[\chsw_p(\mu,\hat{\mu}_n)] \\
            &\le \mathbb{E}[\chsw_p^p(\nu,\hat{\nu}_n)]^{1/p} + \mathbb{E}[\chsw_p^p(\mu, \hat{\mu}_n)]^{1/p}.
        \end{aligned}
    \end{equation}

    Moreover, by Fubini-Tonelli,
    \begin{equation}
        \begin{aligned}
            \mathbb{E}[\chsw_p^p(\hat{\mu}_n,\mu)] &= \mathbb{E}\left[\int_{S_o} W_p^p(P^v_\#\hat{\mu}_n, \mu)\ \mathrm{d}\lambda(v)\right] \\
            &= \int_{S_o} \mathbb{E}[W_p^p(P^v_\#\hat{\mu}_n,P^v_\#\mu)]\ \mathrm{d}\lambda(v).
        \end{aligned}
    \end{equation}
    Then, by applying \Cref{lemma:fournier}, we get that for $q>p$, there exists a constant $C_{p,q}$ such that,
    \begin{equation}
        \mathbb{E}[W_p^p(P^v_\#\hat{\mu}_n, P^v_\#\nu)] \le C_{p,q} \Tilde{M}_q(P^v_\#\mu)^{p/q} \left(n^{-1/2}\mathbb{1}_{\{q>2p\}} + n^{-1/2}\log(n) \mathbb{1}_{\{q=2p\}} + n^{-(q-p)/q} \mathbb{1}_{\{q\in(p,2p)\}}\right).
    \end{equation}

    Then, noting that necessarily, $P^v(o)=0$ (for both the horospherical and geodesic projection, since the geodesic is of the form $\exp_o(tv)$), and using that $P^v$ is 1-Lipschitz \Cref{lemma:lipschitz}, we can bound the moments as
    \begin{equation}
        \begin{aligned}
            \Tilde{M}_q(P^v_\#\mu) &= \int_\mathbb{R} |x|^q\ \mathrm{d}(P^v_\#\mu)(x) \\
            &= \int_\mathcal{M} |P^v(x)|^q\ \mathrm{d}\mu(x) \\
            &= \int_\mathcal{M} |P^v(x)-P^v(o)|^q\ \mathrm{d}\mu(x) \\
            &\le \int_\mathcal{M} d(x,o)^q\ \mathrm{d}\mu(x) \\
            &= M_q(\mu).
        \end{aligned}
    \end{equation}

    Therefore, we have
    \begin{equation}
        \mathbb{E}[\chsw_p^p(\hat{\mu}_n, \mu)] \le C_{p,q} M_q(\mu)^{p/q} \left(n^{-1/2}\mathbb{1}_{\{q>2p\}} + n^{-1/2}\log(n) \mathbb{1}_{\{q=2p\}} + n^{-(q-p)/q} \mathbb{1}_{\{q\in(p,2p)\}}\right),
    \end{equation}
    and similarly,
    \begin{equation}
        \mathbb{E}[\chsw_p^p(\hat{\nu}_n, \nu)] \le C_{p,q} M_q(\nu)^{p/q} \left(n^{-1/2}\mathbb{1}_{\{q>2p\}} + n^{-1/2}\log(n) \mathbb{1}_{\{q=2p\}} + n^{-(q-p)/q} \mathbb{1}_{\{q\in(p,2p)\}}\right).
    \end{equation}

    Hence, we conclude that 
    \begin{equation}
        \mathbb{E}[|\chsw_p(\hat{\mu}_n,\hat{\nu}_n) - \chsw_p(\mu,\nu)|] \le 2 C_{p,q}^{1/p} M_q(\nu)^{1/q} \begin{cases}
            n^{-1/(2p)}\ \text{if } q>2p \\
            n^{-1/(2p)} \log(n)^{1/p}\ \text{if } q=2p \\
            n^{-(q-p)/(pq)}\ \text{if } q \in (p,2p).
        \end{cases}
    \end{equation}
\end{proof}

\subsubsection{Proof of \Cref{prop:chsw_proj_complexity}} \label{proof:prop_chsw_proj_complexity}

\begin{proof}[Proof of \Cref{prop:chsw_proj_complexity}]
    Let $(v_\ell)_{\ell=1}^L$ be iid samples of $\lambda$. Then, by first using Jensen inequality and then remembering that $\mathbb{E}_v[W_p^p(P^v_\#\mu,P^v_\#\nu)] = \chsw_p^p(\mu,\nu)$, we have
    \begin{equation}
        \begin{aligned}
            \mathbb{E}_v\left[|\widehat{\chsw}_{p,L}^p(\mu,\nu)-\chsw_p^p(\mu,\nu)|\right]^2 &\le \mathbb{E}_v\left[\left|\widehat{\chsw}_{p,L}^p(\mu,\nu)-\chsw_p^p(\mu,\nu)\right|^2\right]\\
            &= \mathbb{E}_v\left[\left|\frac{1}{L} \sum_{\ell=1}^L \big(W_p^p(P^{v_\ell}_\#\mu,P^{v_\ell}_\#\nu) - \chsw_p^p(\mu,\nu)\big)\right|^2\right] \\
            &= \frac{1}{L^2} \mathrm{Var}_v\left(\sum_{\ell=1}^L W_p^p(P^{v_\ell}_\#\mu,P^{v_\ell}_\#\nu)\right) \\
            &= \frac{1}{L} \mathrm{Var}_v\left(W_p^p(P^v_\#\mu,P^v_\#\nu)\right) \\
            &= \frac{1}{L} \int_{S_o} \left(W_p^p(P^v_\#\mu,P^v_\#\nu)-\chsw_p^p(\mu,\nu)\right)^2\ \mathrm{d}\lambda(v).
        \end{aligned}
    \end{equation}
\end{proof}

\newpage 

\section{Appendix of \Cref{chapter:hsw}}

\subsection{Proofs of \Cref{section:hsw}}

\subsubsection{Proof of \Cref{prop:hsw_geodesic_proj}} \label{proof:prop_hsw_geodesic_proj}

\begin{proof}[Proof of \Cref{prop:hsw_geodesic_proj}] \leavevmode
    \begin{enumerate}
        \item \textbf{Lorentz model.} Any point $y$ on the geodesic obtained by the intersection between $E=\mathrm{span}(x^0,v)$ and $\mathbb{L}^d$ can be written as
        \begin{equation}
            y = \cosh(t) x^0 + \sinh(t) v,
        \end{equation}
        where $t\in\mathbb{R}$. Moreover, as $\arccosh$ is an increasing function, we have
        \begin{equation}
            \begin{aligned}
                \Tilde{P}^v(x) &= \argmin_{y\in E\cap\mathbb{L}^d}\ d_\mathbb{L}(x,y) \\
                &= \argmin_{y\in E\cap \mathbb{L}^d}\ -\langle x,y\rangle_\mathbb{L}.
            \end{aligned}
        \end{equation}
        This problem is equivalent with solving
        \begin{equation}
            \argmin_{t\in\mathbb{R}}\ -\cosh(t)\langle x,x^0\rangle_\mathbb{L} - \sinh(t)\langle x,v\rangle_\mathbb{L}.
        \end{equation}
        Let $g(t)=-\cosh(t)\langle x,x^0\rangle_\mathbb{L} - \sinh(t)\langle x,v\rangle_\mathbb{L}$, then
        \begin{equation} \label{eq:hsw_coord_opt}
            g'(t) = 0 \iff \tanh(t) = - \frac{\langle x,v\rangle_\mathbb{L}}{\langle x,x^0\rangle_\mathbb{L}}.
        \end{equation}
        Finally, using that $1-\tanh^2(t)=\frac{1}{\cosh^2(t)}$ and $\cosh^2(t)-\sinh^2(t)=1$, and observing that necessarily, $\langle x,x^0\rangle_\mathbb{L} \le 0$, we obtain
        \begin{equation}
            \cosh(t) = \frac{1}{\sqrt{1-\left(-\frac{\langle x,v\rangle_\mathbb{L}}{\langle x,x^0\rangle_\mathbb{L}}\right)^2}} = \frac{-\langle x,x^0\rangle_\mathbb{L}}{\sqrt{\langle x,x^0\rangle_\mathbb{L}^2 - \langle x,v\rangle_\mathbb{L}^2}},
        \end{equation}
        and
        \begin{equation}
            \sinh(t) = \frac{-\frac{\langle x,v\rangle_\mathbb{L}}{\langle x,x^0\rangle_\mathbb{L}}}{\sqrt{1-\left(-\frac{\langle x,v\rangle_\mathbb{L}}{\langle x,x^0\rangle_\mathbb{L}}\right)^2}} = \frac{\langle x, v\rangle_\mathbb{L}}{\sqrt{\langle x,x^0\rangle_\mathbb{L}^2 - \langle x,v\rangle_\mathbb{L}^2}}.
        \end{equation}
        \item \textbf{Poincaré ball.} A geodesic passing through the origin on the Poincaré ball is of the form $\gamma(t) = tp$ for an ideal point $p\in S^{d-1}$ and $t\in ]-1,1[$. Using that $\arccosh$ is an increasing function, we find
        \begin{equation}
            \begin{aligned}
                \Tilde{P}^p(x) &= \argmin_{y \in \mathrm{span}(\gamma)}\ d_\mathbb{B}(x,y) \\
                &= \argmin_{tp}\ \arccosh\left(1 + 2 \frac{\|x-\gamma(t)\|_2^2}{(1-\|x\|_2^2)(1-\|\gamma(t)\|_2^2)}\right) \\
                &= \argmin_{tp}\ \log\big(\|x-\gamma(t)\|_2^2\big) - \log\big(1-\|x\|_2^2\big) - \log\big(1-\|\gamma(t)\|_2^2\big) \\
                &= \argmin_{tp}\ \log\big(\|x-tp\|_2^2\big) - \log\big(1-t^2\big).
            \end{aligned}
        \end{equation}
        Let $g(t) = \log\big(\|x-tp\|_2^2\big) - \log\big(1-t^2\big)$. Then,
        \begin{equation}
            g'(t) = 0 \iff  \left\{\begin{array}{ll}
          t^2 - \frac{1+\|x\|_2^2}{\langle x,p\rangle} t + 1 = 0 & \mbox{ if } \langle p,x\rangle \neq 0, \\
          t = 0 & \mbox{ if } \langle p,x\rangle = 0.
          \end{array}\right.
        \end{equation}
        Finally, if $\langle p,x\rangle \neq 0$, the solution is
        \begin{equation}
            t = \frac{1+\|x\|_2^2}{2\langle x,p\rangle} \pm \sqrt{\left(\frac{1+\|x\|_2^2}{2\langle x,p\rangle}\right)^2 - 1}.
        \end{equation}

        Now, let us suppose that $\langle x, p\rangle > 0$. Then, 
        \begin{equation}
            \begin{aligned}
                \frac{1+\|x\|_2^2}{2\langle x,p\rangle} + \sqrt{\left(\frac{1+\|x\|_2^2}{2\langle x,p\rangle}\right)^2 - 1} &\ge \frac{1+\|x\|_2^2}{2\langle x,p\rangle} \\
                & \ge 1,
            \end{aligned}
        \end{equation}
        because $\|x-p\|_2^2 \ge 0$ implies that $\frac{1+\|x\|_2^2}{2\langle x,p\rangle} \ge 1$, and therefore the solution is
        \begin{equation}
            t = \frac{1+\|x\|_2^2}{2\langle x,p\rangle} - \sqrt{\left(\frac{1+\|x\|_2^2}{2\langle x,p\rangle}\right)^2 - 1}.
        \end{equation}

        Similarly, if $\langle x, p\rangle < 0$, then
        \begin{equation}
            \begin{aligned}
                \frac{1+\|x\|_2^2}{2\langle x,p\rangle} - \sqrt{\left(\frac{1+\|x\|_2^2}{2\langle x,p\rangle}\right)^2 - 1} &\le  \frac{1+\|x\|_2^2}{2\langle x,p\rangle} \\
                &\le -1,
            \end{aligned}
        \end{equation}
        because $\|x+p\|_2^2 \ge 0$ implies $\frac{1+\|x\|_2^2}{2\langle x,p\rangle} \le -1$, and the solution is
        \begin{equation}
            \frac{1+\|x\|_2^2}{2\langle x,p\rangle} + \sqrt{\left(\frac{1+\|x\|_2^2}{2\langle x,p\rangle}\right)^2 - 1}.
        \end{equation}

        Thus,
        \begin{equation}
            \begin{aligned}
                s(x) &= \begin{cases}
                    \frac{1+\|x\|_2^2}{2\langle x, p\rangle} - \sqrt{\left(\frac{1+\|x\|_2^2}{2\langle x,p\rangle}\right)^2 -1} \quad \text{if } \langle x, p\rangle > 0 \\
                    \frac{1+\|x\|_2^2}{2\langle x, p\rangle} + \sqrt{\left(\frac{1+\|x\|_2^2}{2\langle x,p\rangle}\right)^2 -1} \quad \text{if } \langle x, p\rangle < 0.
                \end{cases} \\
                &= \frac{1+\|x\|_2^2}{2\langle x, p\rangle} - \mathrm{sign}(\langle x,p\rangle) \sqrt{\left(\frac{1+\|x\|_2^2}{2\langle x,p\rangle}\right)^2 -1} \\
                &= \frac{1+\|x\|_2^2}{2\langle x, p\rangle}  - \frac{\mathrm{sign}(\langle x,p\rangle)}{2\mathrm{sign}(\langle x,p\rangle)\langle x, p\rangle} \sqrt{(1+\|x\|_2^2)^2 - 4\langle x,p\rangle^2} \\
                &= \frac{1+\|x\|_2^2 - \sqrt{(1+\|x\|_2^2)^2 - 4 \langle x, p\rangle^2}}{2 \langle x, p\rangle}.
            \end{aligned}
        \end{equation}
    \end{enumerate}
\end{proof}

\subsubsection{Proof of \Cref{prop:hsw_coord_geod_proj}} \label{proof:prop_hsw_coord_geod_proj}

\begin{proof}[Proof of \Cref{prop:hsw_coord_geod_proj}] \leavevmode
    \begin{enumerate}
        \item \textbf{Lorentz model.} The coordinate on the geodesic can be obtained as 
        \begin{equation}
            P^v(x) = \argmin_{t\in\mathbb{R}}\ d_\mathbb{L}\big(\exp_{x^0}(tv), x\big).
        \end{equation}
        Hence, by using \eqref{eq:hsw_coord_opt}, we obtain that the optimal $t$ satisfies
        \begin{equation}
            \tanh(t) = - \frac{\langle x, v\rangle_\mathbb{L}}{\langle x,x^0\rangle_\mathbb{L}} \iff t = \arctanh\left(-\frac{\langle x, v\rangle_\mathbb{L}}{\langle x, x^0\rangle_\mathbb{L}}\right).
        \end{equation}
        \item \textbf{Poincaré ball.} As a geodesic is of the form $\gamma(t)=\tanh\left(\frac{t}{2}\right) p$ for all $t\in\mathbb{R}$, we deduce from \Cref{prop:hsw_geodesic_proj} that
        \begin{equation}
            s(x) = \tanh\left(\frac{t}{2}\right) \iff t = 2 \arctanh\big(s(x)\big).
        \end{equation}
    \end{enumerate}
\end{proof}

\subsubsection{Proof of \Cref{prop:busemann_closed_forms}} \label{proof:prop_busemann_closed_forms}

\begin{proof}[Proof of \Cref{prop:busemann_closed_forms}] \leavevmode
    \begin{enumerate}
        \item \textbf{Lorentz model.}
            The geodesic in direction $v$ can be characterized by
            \begin{equation}
                \forall t\in \mathbb{R},\ \gamma_v(t) = \cosh(t)x^0 + \sinh(t) v.
            \end{equation}
            Hence, we have
            \begin{equation}
                \begin{aligned}
                    \forall x\in \mathbb{L}^d,\ d_{\mathbb{L}}(\gamma_v(t), x) &= \arccosh(-\cosh(t)\langle x,x^0\rangle_\mathbb{L} - \sinh(t) \langle x, v\rangle_\mathbb{L} ) \\
                    &= \arccosh\left(-\frac{e^t + e^{-t}}{2} \langle x,x^0\rangle_\mathbb{L} - \frac{e^t - e^{-t}}{2}\langle x,v\rangle_\mathbb{L}\right) \\
                    &= \arccosh\left(\frac{e^t}{2}\big((-1-e^{-2t})\langle x, x^0\rangle_\mathbb{L} + (-1+e^{-2t})\langle x,v\rangle_\mathbb{L}\big)\right) \\
                    &= \arccosh\big(x(t)\big).
                \end{aligned}
            \end{equation}
            Then, on one hand, we have $x(t) \underset{t\to\infty}{\to} \pm \infty$, and using that $\arccosh(x)=\log\big(x+\sqrt{x^2-1}\big)$, we have
            \begin{equation}
                \begin{aligned}
                    d_{\mathbb{L}}(\gamma_v(t),x) - t &= \log\left(\big(x(t) + \sqrt{x(t)^2-1}\big) e^{-t}\right) \\
                    &= \log\left(e^{-t} x(t) + e^{-t}x(t) \sqrt{1-\frac{1}{x(t)^2}}\right) \\
                    &\underset{\infty}{=} \log\left(e^{-t}x(t) + e^{-t}x(t) \left(1-\frac{1}{2x(t)^2} + o\left(\frac{1}{x(t)^2}\right)\right)\right).
                \end{aligned}
            \end{equation}
            Moreover,
            \begin{equation}
                e^{-t}x(t) = \frac12 (-1-e^{-2t})\langle x,x^0\rangle_\mathbb{L} + \frac12(-1+e^{-2t})\langle x,v\rangle_\mathbb{L} \underset{t\to\infty}{\to} -\frac12 \langle x,  x^0 + v\rangle_\mathbb{L}.
            \end{equation}
            Hence,
            \begin{equation}
                B^v(x) = \log(-\langle x, x^0+v\rangle_\mathbb{L}).
            \end{equation}
            
        \item \textbf{Poincaré ball.}
        
        Note that this proof can be found \emph{e.g.} in the Appendix of \citep{ghadimi2021hyperbolic}. We report it for the sake of completeness.
        
            Let $p\in S^{d-1}$, then the geodesic from $0$ to $p$ is of the form $\gamma_p(t) = \exp_0(tp) = \tanh(\frac{t}{2})p$.
            Moreover, recall that $\arccosh(x) = \log(x+\sqrt{x^2 -1})$ and 
            \begin{equation}
                d_{\mathbb{B}}(\gamma_p(t),x) = \arccosh\left(1+2\frac{\|\tanh(\frac{t}{2})p-x\|_2^2}{(1-\tanh^2(\frac{t}{2}))(1-\|x\|_2^2)}\right) = \arccosh(1+x(t)),
            \end{equation}
            where
            \begin{equation}
                x(t) = 2\frac{\|\tanh(\frac{t}{2})p-x\|_2^2}{(1-\tanh^2(\frac{t}{2}))(1-\|x\|_2^2)}.
            \end{equation}
            Now, on one hand, we have
            \begin{equation}
                \begin{aligned}
                    B^p(x) &= \lim_{t\to\infty}\ (d_\mathbb{B}(\gamma_p(t),x)-t) \\
                    &= \lim_{t\to \infty}\ \log\big(1+x(t)+\sqrt{x(t)^2+2x(t)}\big)-t \\
                    &= \lim_{t\to \infty}\ \log\big(e^{-t}(1+x(t)+\sqrt{x(t)^2 + 2x(t)})\big).
                \end{aligned}
            \end{equation}
            On the other hand, using that $\tanh(\frac{t}{2}) = \frac{e^t-1}{e^t+1}$,
            \begin{equation}
                \begin{aligned}
                    e^{-t}x(t) &= 2e^{-t} \frac{\|\frac{e^t-1}{e^t+1}p-x\|_2^2}{(1-(\frac{e^t-1}{e^t+1})^2)(1-\|x\|_2^2)} \\
                    &= 2e^{-t} \frac{\|e^t p - p - e^t x - x\|_2^2}{4e^t (1-\|x\|_2^2)} \\
                    &= \frac12 \frac{\|p-e^{-t}p-x-e^{-t}x\|_2^2}{1-\|x\|_2^2} \\
                    &\underset{t\to\infty}{\to}\frac12 \frac{\|p-x\|_2^2}{1-\|x\|_2^2}.
                \end{aligned}
            \end{equation}
            Hence,
            \begin{equation}
                \begin{aligned}
                    B^p(x) &= \lim_{t\to\infty}\ \log\left( e^{-t} + e^{-t}x(t) + e^{-t}x(t)\sqrt{1+\frac{2}{x(t)}}\right) = \log\left(\frac{\|p-x\|_2^2}{1-\|x\|_2^2}\right),
                \end{aligned}
            \end{equation}
            using that $\sqrt{1+\frac{2}{x(t)}} = 1 + \frac{1}{x(t)} + o(\frac{1}{x(t)})$ and $\frac{1}{x(t)}\to_{t\to\infty} 0$.
    \end{enumerate}
\end{proof}

\subsubsection{Proof of \Cref{prop:horospherical_projection}} \label{proof:prop_horospherical_projections}

\begin{proof}[Proof of \Cref{prop:horospherical_projection}] \leavevmode
    \begin{enumerate}
        \item \textbf{Lorentz model.}
        
        First, a point on the geodesic $\gamma_v$ is of the form
        \begin{equation}
            y(t) = \cosh(t) x^0 + \sinh(t) v,
        \end{equation}
        with $t\in\mathbb{R}$.
        
        The projection along the horosphere amounts at following the level sets of the Busemann function $B^v$. And we have
        \begin{equation}
            \begin{aligned}
                B^v(x) = B^v(y(t)) &\iff \log(-\langle x,x^0+v\rangle_\mathbb{L}) = \log(-\langle \cosh(t)x^0 + \sinh(t) v, x^0+v\rangle_\mathbb{L}) \\
                &\iff \log (-\langle x,x^0+v\rangle_\mathbb{L}) = \log(-\cosh(t) \|x^0\|_\mathbb{L}^2 - \sinh(t)\|v\|_\mathbb{L}^2 ) \\
                &\iff \log(-\langle x,x^0+v\rangle_\mathbb{L} = \log (\cosh(t)-\sinh(t)) \\
                &\iff \langle x, x^0+v\rangle_\mathbb{L} = \sinh(t)-\cosh(t).
            \end{aligned}
        \end{equation}
        By noticing that $\cosh(t) = \frac{1+\tanh^2(\frac{t}{2})}{1-\tanh^2(\frac{t}{2})}$ and $\sinh(t) = \frac{2\tanh(\frac{t}{2})}{1-\tanh^2(\frac{t}{2})}$, let $u=\tanh(\frac{t}{2})$, then we have
        \begin{equation}
            \begin{aligned}
                B^v(x) = B^v(y(t)) &\iff \langle x,x^0+v\rangle_\mathbb{L} = \frac{2u}{1-u^2} - \frac{1+u^2}{1-u^2} = \frac{-(u-1)^2}{(1-u)(1+u)} = \frac{u-1}{u+1} \\
                &\iff u = \frac{1+\langle x,x^0+v\rangle_\mathbb{L}}{1-\langle x,x^0+v\rangle_\mathbb{L}}.
            \end{aligned}
        \end{equation}
        
        We can further continue the computation and obtain, by denoting $c=\langle x,x^0+v\rangle_\mathbb{L}$,
        \begin{equation} \label{eq:proj_horo_lorentz}
            \begin{aligned}
                \Tilde{B}^v(x) &= \frac{1+u^2}{1-u^2} x^0 + \frac{2u}{1-u^2} v \\
                &= \frac{1+\left(\frac{1+c}{1-c}\right)^2}{1-\left(\frac{1+c}{1-c}\right)^2} x^0 + 2 \frac{\left(\frac{1+c}{1-c}\right)}{1-\left(\frac{1+c}{1-c}\right)^2} v \\
                &= \frac{(1-c)^2 + (1+c)^2}{(1-c)^2 - (1+c)^2} x^0 + 2 \frac{(1+c)(1-c)}{(1-c)^2 - (1+c)^2} v \\
                &= - \frac{1+c^2}{2c} x^0 - \frac{1-c^2}{2c} v \\
                &= -\frac{1}{2\langle x, x^0 + v \rangle_\mathbb{L}} \big( (1+\langle x,x^0+v\rangle_\mathbb{L}^2)x^0 + (1-\langle x,x^0 +v\rangle_\mathbb{L}^2)v\big).
            \end{aligned}
        \end{equation}
        
        \item \textbf{Poincaré ball.}
        
        Let $p\in S^{d-1}$. First, we notice that points on the geodesic generated by $p$ and passing through 0 are of the form $x(\lambda)=\lambda p$ where $\lambda\in]-1,1[$.
        
        Moreover, there is a unique horosphere $S(p,x)$ passing through $x$ and starting from $p$. The points on this horosphere are of the form
        \begin{equation}
            \begin{aligned}
                y(\theta) &= \left(\frac{p+x(\lambda^*)}{2}\right) + \left\|\frac{p-x(\lambda^*)}{2}\right\|_2 \left(\cos(\theta) p + \sin(\theta) \frac{x-\langle x,p\rangle p}{\|x-\langle x,p\rangle p\|_2}\right) \\
                &= \frac{1+\lambda^*}{2}p + \frac{1-\lambda^2}{2} \left(\cos(\theta) p + \sin(\theta) \frac{x-\langle x,p\rangle p}{\|x-\langle x,p\rangle p\|_2}\right),
            \end{aligned}
        \end{equation}
        where $\lambda^*$ characterizes the intersection between the geodesic and the horosphere.
        
        Since the horosphere are the level sets of the Busemann function, we have $B^p(x)=B^p(\lambda^* p)$. Thus, we have
        \begin{equation}
            \begin{aligned}
                B^p(x) = B^p(\lambda^* p) &\iff \log\left(\frac{\|p-x\|_2^2}{1-\|x\|_2^2}\right) = \log\left(\frac{\|p-\lambda^* p\|_2^2}{1-\|\lambda^*p\|_2^2}\right) \\
                &\iff \frac{\|p-x\|_2^2}{1-\|x\|_2^2} = \frac{(1-\lambda^*)^2}{1-(\lambda^*)^2} \\
                &\iff \frac{\|p-x\|_2^2}{1-\|x\|_2^2} = \frac{1-\lambda^*}{1+\lambda^*} \\
                &\iff \lambda^* \left(\frac{\|p-x\|_2^2}{1-\|x\|_2^2} + 1\right) = 1 - \frac{\|p-x\|_2^2}{1-\|x\|_2^2} \\
                &\iff \lambda^* = \frac{1-\|x\|_2^2-\|p-x\|_2^2}{1-\|x\|_2^2+\|p-x\|_2^2}.
            \end{aligned}
        \end{equation}
    \end{enumerate}
\end{proof}

\subsubsection{Proof of \Cref{prop:equality_hhsw}} \label{proof:prop_equality_hhsw}

First, we show some Lemma.

\begin{lemma}[Commutation of projections.] \label{lemma:commute_projs}
    Let $v\in\mathrm{span}(x^0)^\bot \cap S^d$ of the form $v=(0,\Tilde{v})$ where $\Tilde{v}\in S^{d-1}$.
    Then, for all $x\in\mathbb{B}^d$, $y\in\mathbb{L}^d$,
    \begin{align}
        P_{\mathbb{B}\to\mathbb{L}}\big(\Tilde{B}^{\Tilde{v}}(x)\big) = \Tilde{B}^v\big(P_{\mathbb{B}\to\mathbb{L}}(x)\big), \label{eq:busemann_b_to_l} \\
        \Tilde{B}^{\Tilde{v}}(P_{\mathbb{L}\to \mathbb{B}}(y)) = P_{\mathbb{L}\to\mathbb{B}}(\Tilde{B}^{v}(y)) \label{eq:busemann_l_to_b} \\ 
        P_{\mathbb{B}\to\mathbb{L}}\big(\Tilde{P}^{\Tilde{v}}(x)\big) = \Tilde{P}^v\big(P_{\mathbb{B}\to\mathbb{L}}(x)\big), \label{eq:geodesic_b_to_l} \\
        \Tilde{P}^{\Tilde{v}}(P_{\mathbb{L}\to \mathbb{B}}(y)) = P_{\mathbb{L}\to\mathbb{B}}(\Tilde{P}^{v}(y)). \label{eq:geodesic_l_to_b}
    \end{align}
\end{lemma}

\begin{proof}
    \textbf{Proof of \eqref{eq:busemann_b_to_l}.} We first show \eqref{eq:busemann_b_to_l}. Let's recall the formula of the different projections.
    
    On one hand, 
    \begin{equation}
        \forall x\in\mathbb{B}^d,\ \Tilde{B}^{\Tilde{v}}(x) = \left(\frac{1-\|x\|_2^2-\|\Tilde{v}-x\|_2^2}{1-\|x\|_2^2+\|\Tilde{v}-x\|_2^2}\right)\Tilde{v},
    \end{equation}
    \begin{equation}
        \forall x\in \mathbb{L}^d,\ \Tilde{B}^v(x) = -\frac{1}{2\langle x, x^0 + v \rangle_\mathbb{L}} \big( (1+\langle x,x^0+v\rangle_\mathbb{L}^2)x^0 + (1-\langle x,x^0 +v\rangle_\mathbb{L}^2)v\big),
    \end{equation}
    and
    \begin{equation}
        \forall x\in\mathbb{B}^d,\ P_{\mathbb{B}\to\mathbb{L}}(x) = \frac{1}{1-\|x\|_2^2}(1+\|x\|_2^2, 2x_1,\dots, 2x_d).
    \end{equation}
    
    Let $x\in\mathbb{B}^d$.
    First, let's compute $P_{\mathbb{B}\to\mathbb{L}}\big(\Tilde{B}^{\Tilde{v}}(x)\big)$. We note that $\|\Tilde{v}\|_2^2=1$ and therefore
    \begin{equation}
        \|\Tilde{B}^{\Tilde{v}}(v)\|_2^2 = \left(\frac{1-\|x\|_2^2-\|\Tilde{v}-x\|_2^2}{1-\|x\|_2^2+\|\Tilde{v}-x\|_2^2}\right)^2.
    \end{equation}
    Then,
    \begin{equation}
        \begin{aligned}
            P_{\mathbb{B}\to\mathbb{L}}\big(\Tilde{B}^{\Tilde{v}}(x)\big) &= \frac{1}{1-\left(\frac{1-\|x\|_2^2-\|\Tilde{v}-x\|_2^2}{1-\|x\|_2^2+\|\Tilde{v}-x\|_2^2}\right)^2} \left(1+\left(\frac{1-\|x\|_2^2-\|\Tilde{v}-x\|_2^2}{1-\|x\|_2^2+\|\Tilde{v}-x\|_2^2}\right)^2, 2 \left(\frac{1-\|x\|_2^2-\|\Tilde{v}-x\|_2^2}{1-\|x\|_2^2+\|\Tilde{v}-x\|_2^2}\right) \Tilde{v}\right) \\
            &= \frac{1}{1-\left(\frac{1-\|x\|_2^2-\|\Tilde{v}-x\|_2^2}{1-\|x\|_2^2+\|\Tilde{v}-x\|_2^2}\right)^2} \left(\left(1+\left(\frac{1-\|x\|_2^2-\|\Tilde{v}-x\|_2^2}{1-\|x\|_2^2+\|\Tilde{v}-x\|_2^2}\right)^2\right)x^0 + 2 \left(\frac{1-\|x\|_2^2-\|\Tilde{v}-x\|_2^2}{1-\|x\|_2^2+\|\Tilde{v}-x\|_2^2}\right) v\right) \\
            &= \frac{\big(1-\|x\|_2^2+\|\Tilde{v}-x\|_2^2\big)^2}{4\|\Tilde{v}-x\|_2^2 (1-\|x\|_2^2)} \left(\frac{2(1-\|x\|_2^2)^2 + 2\|\Tilde{v}-x\|_2^4}{\big(1-\|x\|_2^2+\|\Tilde{v}-x\|_2^2\big)^2} x^0 + 2 \left(\frac{1-\|x\|_2^2-\|\Tilde{v}-x\|_2^2}{1-\|x\|_2^2+\|\Tilde{v}-x\|_2^2}\right) v\right) \\
            &= \frac{1}{2\|\Tilde{v}-x\|_2^2 (1-\|x\|_2^2)} \left( \big((1-\|x\|_2^2)^2 + \|\Tilde{v}-x\|_2^4\big)x^0 + (1-\|x\|_2^2-\|\Tilde{v}-x\|_2^2)(1-\|x\|_2^2+\|\Tilde{v}-x\|_2^2) v \right) \\
            &= \frac{1}{2\|\Tilde{v}-x\|_2^2 (1-\|x\|_2^2)} \left((1-\|x\|_2^2)^2 + \|\Tilde{v}-x\|_2^4\big)x^0 + \big((1-\|x\|_2^2)^2 - \|\Tilde{v}-x\|_2^4\big) v\right).
        \end{aligned}
    \end{equation}
    
    Now, let's compute $\Tilde{B}^v\big(P_{\mathbb{B}\to\mathbb{L}}(x)\big)$. First, let's remark that for all $y\in\mathbb{L}^d$, $\langle y,x^0 + v\rangle_\mathbb{L} = -y_0 + \langle y_{1:d}, \Tilde{v}\rangle$. Therefore, for all $x\in\mathbb{B}^d$,
    \begin{equation}
        \begin{aligned}
            \langle P_{\mathbb{B}\to\mathbb{L}}(x), x^0 + v\rangle_\mathbb{L} &= \langle \frac{1}{1-\|x\|_2^2}(1+\|x\|_2^2, 2x_1,\dots, 2x_d), x^0+v\rangle_\mathbb{L} \\
            &= \frac{1}{1-\|x\|_2^2} \left(-1-\|x\|_2^2 + 2 \langle x, \Tilde{v}\rangle\right) \\
            &= -\frac{1}{1-\|x\|_2^2} \|x-\Tilde{v}\|_2^2.
        \end{aligned}
    \end{equation}
    Moreover,
    \begin{equation}
        \begin{aligned}
            \langle P_{\mathbb{B}\to\mathbb{L}}(x), x^0 + v\rangle_\mathbb{L}^2 &= \frac{1}{(1-\|x\|_2^2)^2} \|\Tilde{v}-x\|_2^4.
        \end{aligned}
    \end{equation}
    
    Therefore, we have
    \begin{equation}
        \begin{aligned}
            \Tilde{B}^v\big(P_{\mathbb{B}\to\mathbb{L}}(x)\big) &= \Tilde{B}^v\left(\frac{1}{1-\|x\|_2^2}(1+\|x\|_2^2, 2x_1,\dots, 2x_d)\right) \\
            &= -\frac{1-\|x\|_2^2}{2 \left(-1-\|x\|_2^2 + 2 \langle x, \Tilde{v}\rangle\right)} \left( \big(1+\langle P_{\mathbb{B}\to\mathbb{L}}(x), x^0 + v\rangle_\mathbb{L}^2\big)x^0 + \big(1-\langle P_{\mathbb{B}\to\mathbb{L}}(x), x^0 + v\rangle_\mathbb{L}^2\big) v\right)\\
            &= \frac{1-\|x\|_2^2}{2 \|x-\Tilde{v}\|_2^2} \left( \frac{(1-\|x\|_2^2)^2 + \|\Tilde{v}-x\|^4}{(1-\|x\|_2^2)^2} x^0 + \frac{(1-\|x\|_2^2)^2 - \|\Tilde{v}-x\|^4}{(1-\|x\|_2^2)^2} v \right) \\
            &= \frac{1}{2\|x-\Tilde{v}\|_2^2 (1-\|x\|_2^2)} \left( (1-\|x\|_2^2)^2 + \|\Tilde{v}-x\|_2^4\big)x^0 + \big((1-\|x\|_2^2)^2 - \|\Tilde{v}-x\|_2^4\big) v\right) \\
            &= P_{\mathbb{B}\to\mathbb{L}}\big(\Tilde{B}^{\Tilde{v}}(x)\big).
        \end{aligned}
    \end{equation}

    \textbf{Proof of \eqref{eq:busemann_l_to_b}.} For \eqref{eq:busemann_l_to_b}, we use that $P_{\mathbb{B}\to\mathbb{L}}$ and $P_{\mathbb{L}\to\mathbb{B}}$ are inverse from each other. Hence, for all $x\in \mathbb{B}^d$, there exists $y\in\mathbb{L}^d$ such that $x=P_{\mathbb{L}\to\mathbb{B}}(y) \iff y = P_{\mathbb{B}\to\mathbb{L}}(x)$, and we obtain the second equality by plugging it into \eqref{eq:busemann_b_to_l}.

    \textbf{Proof of \eqref{eq:geodesic_b_to_l} and \eqref{eq:geodesic_l_to_b}.} Now, let's show \eqref{eq:geodesic_b_to_l}. The proof relies on the observation that $\{\exp_{x^0}(tv),\ t\in\mathbb{R}\}=P_{\mathbb{B}\to\mathbb{L}}\left(\{\exp_0(t\Tilde{v}),\ t\in\mathbb{R}\}\right)$ (\emph{i.e.} the images by $P_{\mathbb{B}\to\mathbb{L}}$ of geodesics in the Poincaré ball are geodesics in the Lorentz model). Thus,
    \begin{equation}
        \begin{aligned}
            \Tilde{P}^v(P_{\mathbb{B}\to\mathbb{L}}(x)) &= \argmin_{z\in \{\exp_{x^0}(tv),\ t\in\mathbb{R}\}}\ d_{\mathbb{L}}(P_{\mathbb{B}\to\mathbb{L}}(x), z) \\
            &= P_{\mathbb{B}\to\mathbb{L}}\big(\argmin_{z\in\{\exp_0(t\Tilde{v}),\ t\in\mathbb{R}\}}\ d_\mathbb{B}(P_{\mathbb{L}\to\mathbb{L}}(x), P_{\mathbb{B}\to\mathbb{L}}(z))\big) \\
            &= P_{\mathbb{B}\to\mathbb{L}}\big(\argmin_{z\in\{\exp_0(t\Tilde{v}),\ t\in\mathbb{R}\}}\ d_\mathbb{B}(x,z)\big) \\
            &= P_{\mathbb{B}\to\mathbb{L}}\big(\Tilde{P}^v(x)\big).
        \end{aligned}
    \end{equation}
    Similarly, we obtain \eqref{eq:geodesic_l_to_b}.
\end{proof}

\begin{proof}[Proof of \Cref{prop:equality_hhsw}] 


    Let $\mu,\nu\in\mathcal{P}(\mathbb{B}^d)$, $\Tilde{\mu} = (P_{\mathbb{B}\to\mathbb{L}})_\#\mu$, $\Tilde{\nu} = (P_{\mathbb{B}\to\mathbb{L}})_\#\nu$, $\Tilde{v}\in S^{d-1}$ an ideal point and $v=(0,\Tilde{v})\in\mathrm{span}(x^0)^\bot$. 
    
    Then, using \Cref{prop:eq_wasserstein_busemann}, \Cref{lemma:paty}, that $P_{\mathbb{B}\to\mathbb{L}}:\mathbb{B}^d\to\mathbb{L}^d$ is an isometry and Lemma \ref{lemma:commute_projs}, we have:
    \begin{equation}
        \begin{aligned}
            W_p^p(B^{\Tilde{v}}_\#\mu, B^{\Tilde{v}}_\#\nu) &= W_p^p(\Tilde{B}^{\Tilde{v}}_\#\mu, \Tilde{B}^{\Tilde{v}}_\#\nu) \quad \text{by \Cref{prop:eq_wasserstein_busemann}} \\
            &= \inf_{\gamma\in\Pi(\mu,\nu)}\ \int_{\mathbb{B}^d \times \mathbb{B}^d} d_\mathbb{B}\big(\Tilde{B}^{\Tilde{v}}(x), \Tilde{B}^{\Tilde{v}}(y)\big)^p\ \mathrm{d}\gamma(x,y) \quad \text{by \Cref{lemma:paty}} \\
            &= \inf_{\gamma\in\Pi(\mu,\nu)}\ \int_{\mathbb{B}^d\times\mathbb{B}^d} d_\mathbb{L}\big(P_{\mathbb{B}\to\mathbb{L}}(\Tilde{B}^{\Tilde{v}}(x)), P_{\mathbb{B}\to\mathbb{L}}(\Tilde{B}^{\Tilde{v}}(y))\big)^p\ \mathrm{d}\gamma(x,y) \quad \text{using that $P_{\mathbb{B}\to\mathbb{L}}$ is an isometry} \\
            &= \inf_{\gamma\in\Pi(\mu,\nu)}\ \int_{\mathbb{B}^d\times\mathbb{B}^d} d_\mathbb{L}\big(\Tilde{B}^v(P_{\mathbb{B}\to\mathbb{L}}(x)), \Tilde{B}^v(P_{\mathbb{B}\to\mathbb{L}}(y))\big)^p\ \mathrm{d}\gamma(x,y) \quad \text{using \Cref{lemma:commute_projs}} \\
            &= \inf_{\gamma\in\Pi(\Tilde{B}^v_\#\Tilde{\mu}, \Tilde{B}^v_\#\Tilde{\nu})}\ \int_{\mathbb{L}^d\times \mathbb{L}^d} d_\mathbb{L}(x,y)^p\ \mathrm{d}\gamma(x,y) \quad \text{by \Cref{lemma:paty}} \\
            &= W_p^p(\Tilde{B}^v_\#\Tilde{\mu}, \Tilde{B}^v_\#\Tilde{\nu}) \\
            &= W_p^p(B^v_\#\Tilde{\mu}, B^v_\#\Tilde{\nu}) \quad \text{by \Cref{prop:eq_wasserstein_busemann}}.
        \end{aligned}
    \end{equation}
    
    It is true for all $\Tilde{v}\in S^{d-1}$, and hence for $\lambda$-almost all $\Tilde{v}\in S^{d-1}$. Therefore, we have
    \begin{equation}
        \hhsw_p^p(\mu,\nu)=\hhsw_p^p(\Tilde{\mu},\Tilde{\nu}).
    \end{equation}

    Using the same proof with \Cref{prop:eq_wasserstein} instead of \Cref{prop:eq_wasserstein_busemann} and $P^v$ instead of $B^v$, we obtain
    \begin{equation}
        \ghsw_p^p(\mu,\nu)=\ghsw_p^p(\Tilde{\mu},\Tilde{\nu}).
    \end{equation}
\end{proof}

\subsubsection{Proof of \Cref{prop:hsw_integration_set}} \label{proof:prop_hsw_integration_set}

\begin{proof}[Proof of \Cref{prop:hsw_integration_set}]
    We will prove this proposition directly by working on the geodesics. As $t^v$ is a isometry (\Cref{prop:isometry}), for all $t\in\mathbb{R}$, there exists a unique $z$ on the geodesic $\mathrm{span}(x^0,v)\cap \mathbb{L}^d$ such that $t=t^v(z)$, and we can rewrite the set of integration as
    \begin{equation}
        \{x\in\mathbb{L}^d,\ P^v(x) = t\} = \{x\in \mathbb{L}^d,\ \Tilde{P}^v(x)=z\}.
    \end{equation}

    For the first inclusion, let $x\in\{x\in\mathbb{L}^d,\ \Tilde{P}^v(x)=z\}$. By Proposition \ref{prop:hsw_geodesic_proj} and hypothesis, we have that
    \begin{equation} \label{eq:proj}
        \Tilde{P}^v(x) = \frac{1}{\sqrt{\langle x,x^0\rangle_\mathbb{L}^2-\langle x,v\rangle_\mathbb{L}^2}} \big(-\langle x,x^0\rangle_\mathbb{L} x^0 + \langle x,v\rangle_\mathbb{L} v\big) = z.
    \end{equation}
    Let's denote $E=\mathrm{span}(v,x^0)$ the plan generating the geodesic. Then, by denoting $P^E$ the orthogonal projection on $E$, we have
    \begin{equation}
        \begin{aligned}
            P^E(x) &= \langle x, v\rangle v + \langle x,x^0\rangle x^0 \\
            &= \langle x,v\rangle_\mathbb{L} v - \langle x,x^0\rangle_\mathbb{L} x^0 \\
            &= \left(\sqrt{\langle x,x^0\rangle_\mathbb{L}^2-\langle x,v\rangle_\mathbb{L}^2}\right) z,
        \end{aligned}
    \end{equation}
    using that $v_0=0$ since $\langle x^0, v\rangle = v_0 = 0$, and hence $\langle x,v\rangle_\mathbb{L}=\langle x,v\rangle$, that $\langle x,x^0\rangle = x_0 = -\langle x,x^0\rangle_\mathbb{L}$ and \eqref{eq:proj}.
    Then, since $v_z\in\mathrm{span}(v, x^0)$ and $\langle z,v_z\rangle = 0$ (by construction of $R_z$), we have
    \begin{equation}
        \begin{aligned}
            \langle x,v_z\rangle &= \langle P^E(x), v_z\rangle  \\
            &= \langle \left(\sqrt{\langle x,x^0\rangle_\mathbb{L}^2-\langle x,v\rangle_\mathbb{L}^2}\right) z, v_z\rangle = 0.
        \end{aligned}
    \end{equation}
    Thus, $x\in \mathrm{span}(v_z)^\bot\cap \mathbb{L}^d$.
    
    For the second inclusion, let $x\in\mathrm{span}(v_z)^\bot \cap \mathbb{L}^d$. Since $z\in\mathrm{span}(v_z)^\bot$ (by construction of $R_z$), we can decompose $\mathrm{span}(v_z)^\bot$ as $\mathrm{span}(v_z)^\bot = \mathrm{span}(z)\oplus (\mathrm{span}(z)^\bot \setminus \mathrm{span}(v_z))$. Hence, there exists $\lambda\in\mathbb{R}$ such that $x=\lambda z + x^\bot$. Moreover, as $z\in\mathrm{span}(x^0,v)$, we have $\langle x,x^0\rangle_\mathbb{L} = \lambda \langle z, x^0\rangle_\mathbb{L}$ and $\langle x,v\rangle_\mathbb{L}=\langle x,v\rangle = \lambda \langle z,v\rangle = \lambda \langle z,v\rangle_\mathbb{L}$.
    Thus, the projection is
    \begin{equation}
        \begin{aligned}
            \Tilde{P}^v(x) &= \frac{1}{\sqrt{\langle x,x^0\rangle_\mathbb{L}^2-\langle x,v\rangle_\mathbb{L}^2}} \big(-\langle x,x^0\rangle_\mathbb{L} x^0 + \langle x,v\rangle_\mathbb{L} v\big) \\
            &= \frac{\lambda}{|\lambda|} \frac{1}{\sqrt{\langle z,x^0\rangle_\mathbb{L}^2-\langle z,v\rangle_\mathbb{L}^2}} \big(-\langle z,x^0\rangle_\mathbb{L} x^0 + \langle z,v\rangle_\mathbb{L} v\big) \\
            &= \frac{\lambda}{|\lambda|} z = \mathrm{sign}(\lambda) z.
        \end{aligned}
    \end{equation}
    But, $-z\notin \mathbb{L}^d$, hence necessarily, $\Tilde{P}^v(x) = z$.
    
    Finally, we can conclude that $\{x\in\mathbb{L}^d,\ \Tilde{P}^v(x)=z\} = \mathrm{span}(v_z)^\bot \cap \mathbb{L}^d$.
\end{proof}

\subsection{Details on Hyperbolic Spaces} \label{appendix:hyperbolic_space}

In this Section, we first recall different generalizations of the Gaussian distribution on Hyperbolic spaces, with a particular focus on Wrapped normal distributions. Then, we recall how to perform Riemannian gradient descent in the Lorentz model and in the Poincaré ball.

\subsubsection{Distributions on Hyperbolic Spaces}


\paragraph{Riemannian normal.} The first way of naturally generalizing Gaussian distributions to Riemannian manifolds is to use the geodesic distance in the density, which becomes
\begin{equation*}
    f(x) \propto \exp\left(-\frac{1}{2\sigma^2}d_M(x,\mu)^2\right).
\end{equation*}
It is actually the distribution maximizing the entropy \citep{pennec2006intrinsic, said2014new}. However, it is not straightforward to sample from such a distribution. For example, \citet{ovinnikov2019poincar} uses a rejection sampling algorithm.

\paragraph{Wrapped normal distribution.} A more convenient distribution, on which we can use the parameterization trick, is the Wrapped normal distribution \citep{nagano2019wrapped}. This distribution can be sampled from by first drawing $v\sim \mathcal{N}(0,\Sigma)$ and then transforming it into $v\in T_{x^0}\mathbb{L}^d$ by concatenating a 0 in the first coordinate. Then, we perform parallel transport to transport $v$ from the tangent space of $x^0$ to the tangent space of $\mu\in\mathbb{L}^d$. Finally, we can project the samples on the manifold using the exponential map. We recall the formula of parallel transport form $x$ to $y$:
\begin{equation}
    \forall v\in T_{x}\mathbb{L}^d,\ \mathrm{PT}_{x\to y}(v) = v + \frac{\langle y,v\rangle_\mathbb{L}}{1-\langle x,y\rangle_\mathbb{L}}(x+y).
\end{equation}

Since it only involves differentiable operations, we can perform the parameterization trick and \emph{e.g.} optimize directly over the mean and the variance. Moreover, by the change of variable formula, we can also derive the density \citep{nagano2019wrapped,bose2020latent}. Let $\Tilde{z}\sim\mathcal{N}(0,\Sigma)$, $z=(0,\Tilde{z})\in T_{x^0}\mathbb{L}^d$, $u=\mathrm{PT}_{x^0\to\mu}(z)$, then the density of $x=\exp_{\mu}(u)$ is:
\begin{equation}
    \log p(x) = \log p(\Tilde{z}) - (d-1)\log\left(\frac{\sinh(\|u\|_\mathbb{L})}{\|u\|_\mathbb{L}}\right).
\end{equation}
In the paper, we write $x\sim \mathcal{G}(\mu,\Sigma)$.

\subsubsection{Optimization on Hyperbolic Spaces} \label{appendix:optim}

For gradient descent on hyperbolic space, we refer to \citep[Section 7.6]{boumal2023introduction} and \citep{wilson2018gradient}. 

In general, for a functional $f:M\to\mathbb{R}$, Riemannian gradient descent is performed, analogously to the Euclidean space, by following the geodesics. Hence, the gradient descent reads as \citep{absil2009optimization, bonnabel2013stochastic}
\begin{equation}
    \forall k\ge 0,\ x_{k+1} = \exp_{x_k}\big(-\gamma \mathrm{grad} f(x_k)\big).
\end{equation}
Note that the exponential map can be replaced more generally by a retraction. We describe in the following paragraphs the different formulae in the Lorentz model and in the Poincaré ball.

\paragraph{Lorentz model.}

Let $f:\mathbb{L}^d \to \mathbb{R}$, then its Riemannian gradient is \citep[Proposition 7.7]{boumal2023introduction}
\begin{equation}
    \mathrm{grad}f(x) = \mathrm{Proj}_x(J\nabla f(x)),
\end{equation}
where $J=\mathrm{diag}(-1,1,\dots,1)$ and $\mathrm{Proj}_x(z) = z + \langle x,z\rangle_\mathbb{L} x$. Furthermore, the exponential map is
\begin{equation}
    \forall v\in T_x\mathbb{L}^d,\ \exp_x(v) = \cosh(\|v\|_\mathbb{L})x + \sinh(\|v\|_\mathbb{L}) \frac{v}{\|v\|_\mathbb{L}}.
\end{equation}

\paragraph{Poincaré ball.}

On $\mathbb{B}^d$, the Riemannian gradient of $f:\mathbb{B}^d\to\mathbb{R}$ can be obtained as \citep[Section 3]{nickel2017poincare}
\begin{equation}
    \mathrm{grad}f(x) = \frac{(1-\|\theta\|_2^2)^2}{4}\nabla f(x).
\end{equation}

\citet{nickel2017poincare} propose to use as retraction $R_x(v) = x+v$ instead of the exponential map, and add a projection, to constrain the value to remain within the Poincaré ball, of the form
\begin{equation}
    \mathrm{proj}(x) = \begin{cases}
        \frac{x}{\|x\|_2} - \epsilon \quad \text{if }\|x\|\ge 1 \\
        x \quad \text{otherwise},
    \end{cases}
\end{equation}
where $\epsilon=10^{-5}$ is a small constant ensuring numerical stability. Hence, the algorithm becomes
\begin{equation}
    x_{k+1} = \mathrm{proj}\left(x_k-\gamma_k \frac{(1-\|x_k\|_2^2)^2}{4} \nabla f(x_k)\right).
\end{equation}

A second solution is to compute directly the exponential map derived in \citep[Corollary 1.1]{ganea2018hyperbolic_cone}:
\begin{equation} \label{eq:exp_poincare}
    \exp_x(v) = \frac{\lambda_x \big(\cosh(\lambda_x\|v\|_2) + \langle x, \frac{v}{\|v\|_2}\rangle \sinh(\lambda_x \|v\|_2)\big) x + \frac{1}{\|v\|_2} \sinh(\lambda_x \|v\|_2)v}{1+(\lambda_x -1)\cosh(\lambda_x\|v\|_2) + \lambda_x \langle x, \frac{v}{\|v\|_2}\rangle \sinh(\lambda_x \|v\|_2)},
\end{equation}
where $\lambda_x = \frac{2}{1-\|x\|_2^2}$.

\subsection{Additional Details of Experiments} \label{appendix:hsw_xps}

\subsubsection{Gradient flows} \label{appendix:gradient_flows}

Denoting $\nu$ the target distribution from which we have access to samples $(y_i)_{i=1}^m$, we aim at learning $\nu$ by solving the following optimization problem:
\begin{equation}
    \mu = \argmin_{\mu}\ \hsw\left(\mu, \frac{1}{m}\sum_{i=1}^m \delta_{x_i}\right).
\end{equation}
As we cannot directly learn $\mu$, we model it as $\hat{\mu}=\frac{1}{n}\sum_{i=1}^n \delta_{x_i}$, and then learn the sample locations $(x_i)_{i=1}^n$ using a Riemannian gradient descent which we described in Appendix \ref{appendix:optim}. In practice, we take $n=500$ and use batches of $500$ target samples at each iteration. To compute the sliced discrepancies, we always use 1000 projections. On Figure \ref{fig:comparison_gradientflows}, we plot the log 2-Wasserstein with geodesic cost between the model measure $\hat{\mu}_k$ at each iteration $k$ and $\nu$. We average over 5 runs of each gradient descent. Now, we describe the specific setting for the different targets.

\paragraph{Wrapped normal distribution.} For the first experiment, we choose as target a wrapped normal distribution $\mathcal{G}(m,\Sigma)$. In the fist setting, we use $m=(1.5,1.25,0)\in\mathbb{L}^2$ and $\Sigma=0.1 I_2$. In the second, we use $m=(8,\sqrt{63},0)\in\mathbb{L}^2$ and $\Sigma = 0.1 I_2$. The learning rate is fixed as 5 for the different discrepancies, except for SWl on the second WND which lies far from origin, and for which we exhibit numerical instabilities with a learning rate too high. Hence, we reduced it to 0.1. We observed the same issue for HHSW on the Lorentz model. Fortunately, the Poincaré version, which is equal to the Lorentz version, did not suffer from these issues. It underlines the benefit of having both formulations. 

\paragraph{Mixture of wrapped normal distributions.} For the second experiment, the target is a mixture of 5 WNDs. The covariance are all taken equal as $0.01 I_2$. For the first setting, the outlying means are (on the Poincaré ball) $m_1=(0,-0.5)$, $m_2=(0,0.5)$, $m_3=(0.5,0)$, $m_4=(-0.5,0)$ and the center mean is $m_5 = (0,0.1)$. In the second setting, the outlying means are $m_1=(0,-0.9)$, $m_2=(0,0.9)$, $m_3=(0.9,0)$ and $m_4=(-0.9,0)$. We use the same $m_5$. The learning rate in this experiment is fixed at 1 for all discrepancies.

\subsubsection{Classification of Images with Busemann}

Denote $\{(x_i,y_i)_{i=1}^n\}$ the training set where $x_i\in\mathbb{R}^m$ and $y_i\in\{1,\dots,C\}$ is a label. The embedding is performed by using a neural network $f_\theta$ and the exponential map at the last layer, which projects the points on the Poincaré ball, \emph{i.e.} for $i\in\{1,\dots,n\}$, the embedding of $x_i$ is $z_i = \exp_0\big(f_\theta(z_i)\big)$, where $\exp_0$ is given by \eqref{eq:exp_poincare}, or more simply by
\begin{equation}
    \exp_0(x) = \tanh\left(\frac{\|x\|_2}{2}\right)\frac{x}{\|x\|_2}.
\end{equation}

The experimental setting of this experiment is the same as \citep{ghadimi2021hyperbolic}. That is, we use a Resnet-32 backbone and optimize it with Adam \citep{kingma2014adam}, a learning rate of 5e-4, weight decay of 5e-5, batch size of 128 and without pre-training. The network is trained for all experiments for 1110 epochs with learning rate decay of 10 after 1000 and 1100 epochs. Moreover, the $C$ prototypes are given by the algorithm of \citep{mettes2019hyperspherical} and are uniform on the sphere $S^{d-1}$.

For the additional hyperparameters in the loss \eqref{eq:loss_hsw}, we use by default $\lambda = 1$, and a mixture of $C$ wrapped normal distributions with means $\alpha p_c$, where $p_c\in S^{d-1}$ is a prototype, $c\in\{1,\dots,C\}$ and $\alpha=0.75$, and covariance matrix $\sigma I_d$ with $\sigma=0.1$. For CIFAR100, in dimension 50, we use $\lambda=1$, $\sigma=0.01$ and $\alpha=0.1$.
\newpage

\section{Appendix of \Cref{chapter:spdsw}}

\subsection{Proofs of \Cref{section:spdsw}}

\subsubsection{Proof of \Cref{prop:spdsw_geodesic_projection}} \label{proof:prop_spdsw_geodesic_projection}

\begin{proof}[Proof of \Cref{prop:spdsw_geodesic_projection}]
    Let $M \in S_d^{++}(\mathbb{R})$. We want to solve 
    \begin{equation}
        \geodproj^{\mathcal{G}_A}(M) = \argmin_{X \in \mathcal{G}_A}\ d_{LE}(X, M)^2.
    \end{equation}
    In the case of the Log-Euclidean metric, $\mathcal{G}_A = \{\exp(tA),\ t\in\mathbb{R}\}$.
    We have
    \begin{equation}
        \begin{aligned}
            d_{LE}(\exp(tA), M)^2 &= \|\log \exp(tA) - \log M\|_F^2\\
            &= \|tA - \log M \|_F^2\\
            &= t^2 \mathrm{Tr}(A^2) + \mathrm{Tr}(\log (M)^2) -2t\mathrm{Tr}(A\log M)\\
            &= g(t).
        \end{aligned}
    \end{equation}
    Hence
    \begin{equation}
        g'(t) = 0 \iff t = \frac{\mathrm{Tr}(A\log M)}{\mathrm{Tr}(A^2)}.
    \end{equation}
    Therefore
    \begin{equation}
        \geodproj^{\mathcal{G}_A}(M) = \exp\left(\frac{\mathrm{Tr}(A\log M)}{\mathrm{Tr}(A^2)} A\right) = \exp\left(\mathrm{Tr}(A\log M) A\right),
    \end{equation}
    since $\|A\|_F^2 = \mathrm{Tr}(A^2) = 1$.
\end{proof}

\subsubsection{Proof of \Cref{prop:spdsw_geodesic_coordinate}} \label{proof:prop_spdsw_geodesic_coordinate}

\begin{proof}[Proof of \Cref{prop:spdsw_geodesic_coordinate}]
    First, we give an orientation to the geodesic. This can be done by taking the sign of the inner product between $\log(\geodproj^{\mathcal{G}_A}(M))$ and $A$.
    \begin{equation}
        \begin{aligned}
            P^A(M) &= \mathrm{sign}(\langle A, \log(\geodproj^{\mathcal{G}_A}(M))\rangle_F) d\big(\geodproj^A(M), I\big) \\
            &= \mathrm{sign}(\langle A, \log(\geodproj^{\mathcal{G}_A}(M))\rangle_F) d\left(\exp\left( \mathrm{Tr}(A\log M) A\right), I\right) \\
            &= \mathrm{sign}(\langle A, \langle A, \log M\rangle_F A\rangle_F)  \|\langle A\log M\rangle_F A - \log I \|_F \\
            &= \mathrm{sign}(\langle A, \log M\rangle_F) |\langle A, \log M\rangle_F| \\
            &= \langle A, \log M\rangle_F \\
            &= \mathrm{Tr}(A\log M).
        \end{aligned}
    \end{equation}
\end{proof}

\subsubsection{Proof of \Cref{prop:spdsw_busemann_coords}} \label{proof:prop_spdsw_busemann_coords}

\begin{proof}[Proof of \Cref{prop:spdsw_busemann_coords}]
    First, following \citep{bridson2013metric}, we have for all $M\in S_d^{++}(\mathbb{R}),$
    \begin{equation}
        B^A(M) = \lim_{t\to\infty}\ \big(d_{LE}(\gamma_A(t),M) - t \big) = \lim_{t\to\infty} \frac{d_{LE}(\gamma_A(t), M)^2 - t^2}{2t},
    \end{equation}
    denoting $\gamma_A:t\mapsto \exp(tA)$ is the geodesic line associated to $\mathcal{G}_A$. Then, we get
    \begin{equation}
        \begin{aligned}
            \frac{d_{LE}(\gamma_A(t), M)^2 - t^2}{2t} &= \frac{1}{2t}\left( \|\log \gamma_A(t) - \log M\|_F^2 - t^2\right)\\ 
            &= \frac{1}{2t}\left(\|tA-\log M\|_F^2 - t^2\right) \\
            &= \frac{1}{2t} \left(t^2 \|A\|_F^2 + \|\log M\|_F^2 - 2t\langle A, \log M\rangle_F - t^2\right) \\
            &= -\langle A,\log M\rangle_F + \frac{1}{2t}\|\log M\|_F^2 ,
        \end{aligned}
    \end{equation}
    using that $\|A\|_F = 1$. Then, by passing to the limit $t\to \infty$, we find
    \begin{equation}
        B^A(t) = -\langle A,\log M\rangle_F = -\mathrm{Tr}(A\log M).
    \end{equation}
\end{proof}

\subsubsection{Proof of \Cref{prop:equivalence_swlog}} \label{proof:prop_equivalence_swlog}

\begin{proof}[Proof of \Cref{prop:equivalence_swlog}]
    Denoting $t^A(B)=\langle B,A\rangle_F$ for all $B\in S_d(\mathbb{R})$, we obtain using \Cref{lemma:paty}
    \begin{equation}
        \begin{aligned}
            W_p^p(t^A_\#\log_\#\mu, t^A_\#\log_\#\nu) &= \inf_{\gamma\in\Pi(\mu,\nu)}\ \int_{S_d^{++}(\mathbb{R})\times S_d^{++}(\mathbb{R})} |t^A(\log(X))-t^A(\log(Y))|^p\ \mathrm{d}\gamma(X,Y) \\
            &= \inf_{\gamma\in\Pi(\mu,\nu)}\ \int_{S_d^{++}(\mathbb{R})\times S_d^{++}(\mathbb{R})} |P^A(X)-P^A(Y)|^p\ \mathrm{d}\gamma(X,Y) \\
            &= W_p^p(P^A_\#\mu, P^A_\#\nu),
        \end{aligned}
    \end{equation}
    since $t^A(\log X) = \langle A, \log X\rangle_F = P^A(X)$. Hence,
    \begin{equation}
        \symsw_p^p(\log_\#\mu,\log_\#\nu) = \lespdsw_p^p(\mu,\nu)\enspace .
    \end{equation}
\end{proof}

\subsubsection{Proof of \Cref{lemma:uniform_distribution}} \label{proof:lemma_uniform_distribution}

\begin{proof}[Proof of \Cref{lemma:uniform_distribution}]
    A matrix in $S_d(\mathbb{R})$ has a unique decomposition $P\mathrm{diag}(\theta)P^T$ up to permutations of the columns of $P \in \mathcal{O}_d$ and coefficients of $\theta \in S^{d-1}$.
    Thus, there is a bijection between $\{A \in S_d(\mathbb{R}),\ \| A\|_F = 1\}$ and the set $S_{\mathcal(O),S^{d-1}}$ of $d!$-tuple $\{ (P_1, \theta_1), \dots, (P_{d!}, \theta_{d!})  \in (\mathcal{O}_d \times S^{d-1})^{d!}\} $ such that $(P_i, \theta_i)$ is a permutation of $(P_j, \theta_j)$.
    Therefore, the uniform distribution $\lambda_{S_{\mathcal(O),S^{d-1}}}$ on $S_{\mathcal(O),S^{d-1}}$, defined as $\mathrm{d}\lambda_{S_{\mathcal(O),S^{d-1}}}((P_1, \theta_1), \dots, (P_{d!}, \theta_{d!})) = \sum_{i=1}^{n!} \mathrm{d}(\lambda_O \otimes \lambda) (P_i, \theta_i) = d!\cdot \mathrm{d}(\lambda_O \otimes \lambda) (P_1, \theta_1)$, allows to define a uniform distribution $\lambda_S$ on $\{A \in S_d(\mathbb{R}),\ \| A\|_F = 1\}$. Let $A = P \mathrm{diag}\theta P^T$ with $(P, \theta) \in \mathcal{O}_d \times S^{d-1}$, then
    \begin{equation}
       \mathrm{d}\lambda_S(A) = d!\ \mathrm{d}(\lambda_O \otimes \lambda) (P, \theta). 
    \end{equation}
\end{proof}

\subsubsection{Proof of \Cref{prop:spdsw_distance}} \label{proof:prop_spdsw_distance}

\begin{proof}[Proof of \Cref{prop:spdsw_distance}]
    By \Cref{prop:chsw_pseudo_distance}, we know that $\lespdsw$ is a finite pseudo-distance on $\mathcal{P}_p(S_d^{++}(\mathbb{R}))$. We need here to show indiscernible property.

    Let $\mu,\nu\in\mathcal{P}_p(S_d^{++}(\mathbb{R}))$ such that $\lespdsw_p(\mu,\nu)=0$. Then, as for all $A\in S_d(\mathbb{R})$, $W_p^p(P^A_\#\mu,P^A_\#\nu)\ge 0$, it implies that for $\lambda_S$-almost every $A$, $W_p^p(P^A_\#\mu, P^A_\#\nu)=0$ which implies $P^A_\#\mu=P^A_\#\nu$ for $\lambda_S$-almost every $A$ since $W_p$ is a distance. By taking the Fourier transform, this implies that for all $s\in\mathbb{R}$, $\widehat{P^A_\#\mu}(s) = \widehat{P^A_\#\nu}(s)$. But, we have
    \begin{equation}
        \begin{aligned}
            \widehat{P^A_\#\mu}(s) &= \int_{\mathbb{R}} e^{-2i\pi ts}\ \mathrm{d}(P^A_\#\mu)(s) \\
            &= \int_{S_d^{++}(\mathbb{R})} e^{-2i\pi P^A(M) s}\ \mathrm{d}\mu(M) \\
            &= \int_{S_d^{++}(\mathbb{R})} e^{-2i\pi \langle sA, \log M\rangle_F}\ \mathrm{d}\mu(M) \\
            &= \int_{S_d(\mathbb{R})} e^{-2i\pi \langle sA, S\rangle_F}\ \mathrm{d}(\log_\#\mu)(S) \\
            &= \widehat{\log_\#\mu}(sA).
        \end{aligned}
    \end{equation}
    Hence, we get that $\lespdsw_p(\mu,\nu)=0$ implies that for $\lambda_S$-almost every $A$,
    \begin{equation}
        \forall s\in \mathbb{R},\ \widehat{\log_\#\mu}(sA) = \widehat{P^A_\#\mu}(s) = \widehat{P^A_\#\nu}(s) = \widehat{\log_\#\nu}(sA).
    \end{equation}
    By injectivity of the Fourier transform on $S_d(\mathbb{R})$, we get $\log_\#\mu=\log_\#\nu$. Then, as $\log$ is a bijection from $S_d^{++}(\mathbb{R})$ to $S_d(\mathbb{R})$, we have for all Borelian $C\subset S_d^{++}(\mathbb{R})$,
    \begin{equation}
        \begin{aligned}
            \mu(C) &= \int_{S_d^{++}(\mathbb{R})} \mathbb{1}_C(X)\ \mathrm{d}\mu(X) \\
            &= \int_{S_d(\mathbb{R})} \mathbb{1}_C(\exp(S))\ \mathrm{d}(\log_\#\mu)(S) \\
            &= \int_{S_d(\mathbb{R})} \mathbb{1}_C(\exp(S))\ \mathrm{d}(\log_\#\nu)(S) \\
            &= \int_{S_d^{++}(\mathbb{R})} \mathbb{1}_C(Y)\ \mathrm{d}\nu(Y) \\
            &= \nu(C).
        \end{aligned}
    \end{equation}
    Hence, we conclude that $\mu=\nu$ and that $\lespdsw_p$ is a distance.
\end{proof}

\subsubsection{Proof of \Cref{prop:spdsw_weakcv}} \label{proof:prop_spdsw_weakcv}

To prove \Cref{prop:spdsw_weakcv}, we will adapt the proof of \citet{nadjahi2020statistical} to our projection. First, we start to adapt \citet[Lemma S1]{nadjahi2020statistical}:
\begin{lemma}[Lemma S1 in \citet{nadjahi2020statistical}] \label{lemma:nadjahi}
    Let $(\mu_k)_k \in \mathcal{P}_p(S_d^{++}(\mathbb{R}))$ and $\mu\in \mathcal{P}_p(S_d^{++}(\mathbb{R}))$ such that $\lim_{k\to\infty}\ \mathrm{SPDSW}_1(\mu_k,\mu)=0$. Then, there exists $\varphi:\mathbb{N}\to\mathbb{N}$ non decreasing such that $\mu_{\varphi(k)} \xrightarrow[k\to\infty]{\mathcal{L}} \mu$.
\end{lemma}

\begin{proof}
    By \citet[Theorem 2.2.5]{bogachev2007measure}, 
    \begin{equation}
        \lim_{k\to\infty}\ \int_{S^d(\mathbb{R})} W_1(P^A_\#\mu_k, P^A_\#\mu)\ \mathrm{d}\lambda_S(A) = 0
    \end{equation}
    implies that there exists a subsequence $(\mu_{\varphi(k)})_k$ such that for $\lambda_S$-almost every $A\in S_d(\mathbb{R})$,
    \begin{equation}
        W_1(P^A_\#\mu_{\varphi(k)}, P^A_\#\mu) \xrightarrow[k\to \infty]{}0.
    \end{equation}
    As the Wasserstein distance metrizes the weak convergence, this is equivalent to $P^A_\#\mu_{\varphi(k)} \xrightarrow[k\to\infty]{\mathcal{L}} P^A_\#\mu$.
    
    Then, by Levy's characterization theorem, this is equivalent with the pointwise convergence of the characterization function, \emph{i.e.} for all $t\in \mathbb{R}$,\ $\phi_{P^A_\#\mu_{\varphi(k)}}(t) \xrightarrow[k\to \infty]{} \phi_{P^A_\#\mu}(t)$.
    Moreover, we have for all $s\in \mathbb{R}$,
    \begin{equation}
        \begin{aligned}
            \phi_{P^A_\#\mu_{\varphi(k)}}(s) &= \int_\mathbb{R} e^{-its} \mathrm{d}(P^A_\#\mu_{\varphi(k)})(t) \\
            &= \int_{S_d^{++}(\mathbb{R})} e^{-i P^A(M) s}\ \mathrm{d}\mu_{\varphi(k)}(M) \\
            &= \int_{S_d^{++}(\mathbb{R})} e^{-i \langle sA, \log M\rangle_F}\ \mathrm{d}\mu_{\varphi(k)}(M) \\
            &= \int_{S_d(\mathbb{R})} e^{-i\langle sA, S\rangle_F} \ \mathrm{d}(\log_\#\mu_{\varphi(k)})(S) \\
            &= \phi_{\log_\#\mu_{\varphi(k)}}(sA) \\
            &\xrightarrow[k\to \infty]{} \phi_{\log_\#\mu}(sA).
        \end{aligned}
    \end{equation}
    Then, working in $S_d(\mathbb{R})$ with the Frobenius norm, we can use the same proof of \citet{nadjahi2020statistical} by using a convolution with a gaussian kernel and show that it implies that $\log_\#\mu_{\varphi(k)}\xrightarrow[k\to\infty]{\mathcal{L}}\log_\#\mu$.
    
    Finally, let's show that it implies the weak convergence of $(\mu_{\varphi(k)})_k$ towards $\mu$. Let $f\in C_b(S_d^{++}(\mathbb{R}))$, then
    \begin{equation}
        \begin{aligned}
            \int_{S_d^{++}(\mathbb{R})} f \ \mathrm{d}\mu_{\varphi(k)} &= \int_{S_d(\mathbb{R})} f\circ \exp\ \mathrm{d}(\log_\#\mu_{\varphi(k)}) \\
            &\xrightarrow[k\to \infty]{} \int_{S_d(\mathbb{R})} f\circ \exp\ \mathrm{d}(\log_\#\mu) \\
            &= \int_{S_d^{++}(\mathbb{R})} f\ \mathrm{d}\mu.
        \end{aligned}
    \end{equation}
    Hence, we an conclude that $\mu_{\varphi(k)} \xrightarrow[k\to\infty]{\mathcal{L}} \mu$.
\end{proof}

\begin{proof}[Proof of \Cref{prop:spdsw_weakcv}]
    First, we suppose that $\mu_k \xrightarrow[k\to\infty]{\mathcal{L}} \mu$ in $\mathcal{P}_p(S_d(\mathbb{R}))$. Then, by continuity, we have that for $\lambda_S$ almost every $A\in S_d^{++}(\mathbb{R})$, $P^A_\#\mu_k \xrightarrow[k\to \infty]{} P^A_\#\mu$. Moreover, as the Wasserstein distance on $\mathbb{R}$ metrizes the weak convergence, $W_p(P^A_\#\mu_k, P^A_\#\mu) \xrightarrow[k\to\infty]{} 0$. Finally, as $W_p$ is bounded and it converges for $\lambda_S$-almost every $A$, we have by the Lebesgue convergence dominated theorem that $\lespdsw_p^p(\mu_k,\mu) \xrightarrow[k\to\infty]{} 0$.
        
    On the other hand, suppose that $\lespdsw_p(\mu_k,\mu)\xrightarrow[k\to\infty]{}0$. We first adapt Lemma S1 of \citep{nadjahi2020statistical} in Lemma \ref{lemma:nadjahi} and observe that by the Hölder inequality, 
    \begin{equation} \label{eq:holder}
        \lespdsw_1(\mu,\nu) \le \lespdsw_p(\mu,\nu),
    \end{equation}
    and hence $\lespdsw_1(\mu_k,\mu)\xrightarrow[k\to\infty]{} 0$.
    
    By the same contradiction argument as in \citet{nadjahi2020statistical}, let's suppose that $(\mu_k)_k$ does not converge to $\mu$. Then, by denoting $d_P$ the Lévy-Prokhorov metric, $\lim_{k\to\infty}d_P(\mu_k,\mu)\neq 0$. Hence, there exists $\epsilon>0$ and a subsequence $(\mu_{\varphi(k)})_k$ such that $d_P(\mu_{\varphi(k)},\mu)>\epsilon$.
    
    Then, we have first that $\lim_{k\to\infty}\lespdsw_1(\mu_{\varphi(k)},\mu) = 0$. Thus, by Lemma \ref{lemma:nadjahi}, there exists a subsequence $(\mu_{\psi(\varphi(k))})_k$ such that $\mu_{\psi(\varphi(k))}\xrightarrow[k\to\infty]{\mathcal{L}}\mu$ which is equivalent to $\lim_{k\to\infty} d_P(\mu_{\psi(\varphi(k))}, \mu) = 0$ which contradicts the hypothesis.
    
    We conclude that $(\mu_k)_k$ converges weakly to $\mu$.
\end{proof}

\subsubsection{Proof of \Cref{prop:spdsw_bound}} \label{proof:prop_spdsw_bound}

For the proof of \Cref{prop:spdsw_bound}, we will first recall the following Theorem:
\begin{theorem}[\citep{rivin2007surface}, Theorem 3] \label{th3}
    Let $f:\mathbb{R}^d\mapsto\mathbb{R}$ a homogeneous function of degree $p$ (\emph{i.e.} $\forall \alpha\in\mathbb{R},\ f(\alpha x)=\alpha^p f(x)$). Then,
    \begin{equation}
        \Gamma\Big(\frac{d+p}{2}\Big)\int_{S^{d-1}} f(x)\ \lambda(\mathrm{d}x) = \Gamma\Big(\frac{d}{2}\Big)\mathbb{E}[f(X)] \enspace,
    \end{equation}
    where $\forall i\in\{1,...,d\}$, $X_i\sim\mathcal{N}(0,\frac12)$ and $(X_i)_i$ are independent.
\end{theorem}
Then, making extensive use of this theorem, we show the following lemma:
\begin{lemma} \label{lemma_eq}
    \begin{equation}
        \forall S\in S_d(\mathbb{R}),\ \int_{S^{d-1}} |\langle \mathrm{diag}(\theta),S\rangle_F|^p\ \lambda(\mathrm{d}\theta) = \frac{1}{d} \left(\sum_i S_{ii}^2\right)^{\frac{p}{2}} \int_{S^{d-1}} \|\theta\|_p^p\ \lambda(\mathrm{d}\theta).
    \end{equation}
\end{lemma}

\begin{proof}
    Let $f:\theta\mapsto \|\theta\|_p^p=\sum_{i=1}^d \theta_i^p$, then we have $f(\alpha\theta)=\alpha^p f(\theta)$ and $f$ is $p$-homogeneous. By applying Theorem \ref{th3}, we have:
    \begin{equation}
        \begin{aligned}
            \int_{S^{d-1}} \|\theta\|_p^p\ \lambda(\mathrm{d}\theta) &= \frac{\Gamma\Big(\frac{d}{2}\Big)}{\Gamma\Big(\frac{d+p}{2}\Big)}\mathbb{E}[\|X\|_p^p] \text{ with $X_i \overset{\mathrm{iid}}{\sim}\mathcal{N}(0,\frac12)$}\\
            &= \frac{\Gamma\Big(\frac{d}{2}\Big)}{\Gamma\Big(\frac{d+p}{2}\Big)}d\ \mathbb{E}[|X_1|_p^p] \\
            &= \frac{\Gamma\Big(\frac{d}{2}\Big)}{\Gamma\Big(\frac{d+p}{2}\Big)}d\ \int |t|^p \frac{1}{\sqrt{\pi}} e^{-t^2} \mathrm{d}t.
        \end{aligned}
    \end{equation}
    
    On the other hand, let $\Tilde{f}:\theta\mapsto|\langle \mathrm{diag}(\theta),S\rangle_F|^p$, then $\Tilde{f}(\alpha\theta)=\alpha^p \Tilde{f}(\theta)$ and $\Tilde{f}$ is p-homogeneous. By applying Theorem \ref{th3}, we have:
    \begin{equation}
        \begin{aligned}
            \int_{S^{d-1}} |\langle \mathrm{diag}(\theta),S\rangle_F|^p\ \lambda(\mathrm{d}\theta) &= \frac{\Gamma\Big(\frac{d}{2}\Big)}{\Gamma\Big(\frac{d+p}{2}\Big)}\mathbb{E}[|\langle \mathrm{diag}(X),S\rangle_F|^p]\text{ with $X_i \overset{\mathrm{iid}}{\sim}\mathcal{N}(0,\frac12)$}\\
            &= \frac{\Gamma\Big(\frac{d}{2}\Big)}{\Gamma\Big(\frac{d+p}{2}\Big)} \int |t|^p \frac{1}{\sqrt{\sum_i S_{ii}^2 \pi}} e^{-\frac{t^2}{\sum_i z_{ii}^2}}\ \mathrm{d}t \text{ as $\langle \mathrm{diag}(X),S\rangle_F = \sum_i S_{ii} X_i \sim \mathcal{N}\Big(0,\frac{\sum_i S_{ii}^2}{2}\Big)$} \\
            &= \frac{\Gamma\Big(\frac{d}{2}\Big)}{\Gamma\Big(\frac{d+p}{2}\Big)} \left(\sum_i S_{ii}^2\right)^{\frac{p}{2}} \int |u|^p \frac{1}{\sqrt{\sum_i S_{ii}^2 \pi}} e^{-u^2} \sqrt{\sum_i S_{ii}^2} \mathrm{d}u \text{ by $u=\frac{t}{\sqrt{\sum_i S_{ii}^2}}$} \\
            &= \frac{\Gamma\Big(\frac{d}{2}\Big)}{\Gamma\Big(\frac{d+p}{2}\Big)} \left(\sum_i S_{ii}^2\right)^{\frac{p}{2}} \int |u|^p \frac{1}{\sqrt{\pi}} e^{-u^2} \mathrm{d}u.
        \end{aligned}
    \end{equation}
    
    Hence, we deduce that
    \begin{equation}
        \int_{S^{d-1}} |\langle \mathrm{diag}(\theta),S\rangle_F|^p\ \lambda(\mathrm{d}\theta) = \frac{1}{d} \left(\sum_i S_{ii}^2\right)^{\frac{p}{2}} \int_{S^{d-1}} \|\theta\|_p^p \ \mathrm{d}\lambda(\theta).
    \end{equation}
\end{proof}

\begin{proof}[Proof of \Cref{prop:spdsw_bound}]
    First, we show the upper bound of $\lespdsw_p$. Let $\mu,\nu\in\mathcal{P}_p(S_d^{++}(\mathbb{R})$ and $\gamma\in \Pi(\mu,\nu)$ an optimal coupling. Then, following the proof of \citet[Proposition 5.1.3]{bonnotte2013unidimensional}, and using \Cref{lemma:paty} combined with the fact that $(P^A\otimes P^A)_\#\gamma\in\Pi(P^A_\#\mu, P^A_\#\nu)$ for any $A\in S_d(\mathbb{R})$ such that $\|A\|_F=1$, we obtain
    \begin{equation} \label{eq:ineq_spdsw}
        \begin{aligned}
            \lespdsw_p^p(\mu,\nu) &= \int_{S_d(\mathbb{R})} W_p^p(P^A_\#\mu, P^A_\#\nu)\ \mathrm{d}\lambda_S(A) \\
            &\le \int_{S_d(\mathbb{R})} \int_{S_d^{++}(\mathbb{R})\times S_d^{++}(\mathbb{R})} |P^A(X)-P^A(Y)|^p\ \mathrm{d}\gamma(X,Y)\ \mathrm{d}\lambda_S(A) \\
            &= \int_{S_d(\mathbb{R})} \int_{S_d^{++}(\mathbb{R})\times S_d^{++}(\mathbb{R})} |\langle A, \log X - \log Y\rangle_F|^p\ \mathrm{d}\gamma(X,Y)\ \mathrm{d}\lambda_S(A) \\
            &= \int_{S^{d-1}}\int_{\mathcal{O}_d} \int_{S_d^{++}(\mathbb{R})\times S_d^{++}(\mathbb{R})} |\langle P\mathrm{diag}(\theta)P^T, \log X-\log Y\rangle_F|^p\ \mathrm{d}\gamma(X,Y)\ \mathrm{d}\lambda_O(P)\mathrm{d}\lambda(\theta) \\
            &= \int_{S^{d-1}}\int_{\mathcal{O}_d} \int_{S_d^{++}(\mathbb{R})\times S_d^{++}(\mathbb{R})} |\langle \mathrm{diag}(\theta), P^T(\log X-\log Y)P\rangle_F|^p\ \mathrm{d}\gamma(X,Y)\ \mathrm{d}\lambda_O(P)\mathrm{d}\lambda(\theta).
        \end{aligned}
    \end{equation}
    By Lemma \ref{lemma_eq}, noting $S=P^T(\log X-\log Y)P$, we have
    \begin{equation}
        \begin{aligned}
            \int_{S^{d-1}} |\langle \mathrm{diag}(\theta), S\rangle_F|^p\ \mathrm{d}\lambda(\theta) &= \frac{1}{d} \left(\sum_i S_{ii}^2\right)^{\frac{p}{2}} \int_{S^{d-1}} \|\theta\|_p^p\ \mathrm{d}\lambda(\theta) \\
            &\le \frac{1}{d} \|S\|_F^p \int_{S^{d-1}} \|\theta\|_p^p\ \mathrm{d}\lambda(\theta), 
        \end{aligned}
    \end{equation}
    since $\|S\|_F^2 = \sum_{i,j} S_{ij}^2 \ge \sum_i S_{ii}^2$. Moreover, $\|S\|_F = \|P^T(\log X-\log Y)P\|_F = \|\log X-\log Y\|_F$. Hence, coming back to \eqref{eq:ineq_spdsw}, we find
    \begin{equation} \label{eq:upperbound}
        \begin{aligned}
            \mathrm{SPDSW}_p^p(\mu,\nu) &\le \frac{1}{d} \int_{S^{d-1}} \|\theta\|_p^p \ \mathrm{d}\lambda(\theta) \int_{S_d^{++}(\mathbb{R})\times S_d^{++}(\mathbb{R})} \|\log X-\log Y\|_F^p\ \mathrm{d}\gamma(X,Y) \\
            &= \frac{1}{d}  \int_{S^{d-1}} \|\theta\|_p^p\ \mathrm{d}\lambda(\theta)\  W_p^p(\mu,\nu) \\
            &= c_{d,p}^p W_p^p(\mu,\nu).
        \end{aligned}
    \end{equation}
    since $\gamma$ is an optimal coupling between $\mu$ and $\nu$ for the Wasserstein distance with Log-Euclidean cost.

    For the lower bound, let us first observe that 
    \begin{equation}
        \begin{aligned}
            W_1(\mu,\nu) &= \inf_{\gamma\in\Pi(\mu,\nu)}\ \int_{S_d^{++}(\mathbb{R})\times S_d^{++}(\mathbb{R})} d_{LE}(X,Y)\ \mathrm{d}\gamma(X,Y) \\
            &= \inf_{\gamma\in\Pi(\mu,\nu)}\ \int_{S_d^{++}(\mathbb{R})\times S_d^{++}(\mathbb{R})} \|\log X - \log Y\|_F\ \mathrm{d}\gamma(X,Y) \\
            &= \inf_{\gamma\in\Pi(\mu,\nu)}\ \int_{S_d(\mathbb{R})\times S_d(\mathbb{R})} \|U-V\|_F\ \mathrm{d}(\log\otimes\log)_\#\gamma(U,V) \\
            &= \inf_{\gamma\in\Pi(\log_\#\mu,\log_\#\nu)}\ \int_{S_d(\mathbb{R})\times S_d(\mathbb{R})} \|U-V\|_F\ \mathrm{d}\gamma(U,V) \\
            &= W_1(\log_\#\mu,\log_\#\nu),
        \end{aligned}
    \end{equation}
    where we used \Cref{lemma:paty}. Here, note that $W_1$ must be understood with the groundcost metric which makes sense given the space, \emph{i.e.} $d_{LE}$ for $S_d^{++}(\mathbb{R})$ and $\|\cdot\|_F$ for $S_d(\mathbb{R})$.
    
    
    Using \Cref{prop:equivalence_swlog}, we have
    \begin{equation}
        \symsw_1(\log_\#\mu,\log_\#\nu) = \lespdsw_1(\mu,\nu).
    \end{equation}

    Therefore, as $S_d(\mathbb{R})$ is an Euclidean space of dimension $d(d+1)/2$, we can use \citep[Lemma 5.1.4]{bonnotte2013unidimensional} and we obtain that 
    \begin{equation}
        W_1(\log_\#\mu,\log_\#\nu) \le C_{d(d+1)/2} R^{d(d+1)/(d(d+1)+2)} \symsw_1(\log_\#\mu,\log_\#\nu)^{2/(d(d+1)+2)}.
    \end{equation}
    Then, using that $\symsw_1(\log_\#\mu,\log_\#\nu) = \lespdsw_1(\mu,\nu)$ and $W_1(\log_\#\mu,\log_\#\nu) = W_1(\mu,\nu)$, we obtain
    \begin{equation} \label{eq:ineq_w1}
        W_1(\mu,\nu) \le C_{d(d+1)/2} R^{d(d+1)/(d(d+1)+2)} \lespdsw_1(\mu,\nu)^{2/(d(d+1)+2)}.
    \end{equation}

    Now, following the proof of \citet[Theorem 5.1.5]{bonnotte2013unidimensional}, we use that on one hand, $W_p^p(\mu,\nu) \le (2R)^{p-1} W_1(\mu,\nu)$, and on the other hand, by Hölder, $\lespdsw_1(\mu,\nu)\le \lespdsw_p(\mu,\nu)$. Hence, using inequalities \eqref{eq:upperbound} and \eqref{eq:ineq_w1}, we get
    \begin{equation}
        \begin{aligned}
            \lespdsw_p^p(\mu,\nu) &\le c_{d,p}^p W_p^p(\mu,\nu) \\
            &\le (2R)^{p-1} W_1(\mu,\nu) \\
            &\le 2^{p-1} C_{d(d+1)/2} R^{p-1+d(d+1)/(d(d+1)+2)}\lespdsw_1(\mu,\nu)^{2/(d(d+1)/2)} \\
            &= C_{d,p}^d R^{p-2/(d(d+1))}\lespdsw_1(\mu,\nu)^{2/(d(d+1)+2)}.
        \end{aligned}
    \end{equation}
\end{proof}

\subsection{Complementary experiments}
\label{sec:comp_exp}

\subsubsection{Brain Age Prediction}

\begin{figure}[h!]
    \includegraphics[width=\linewidth]{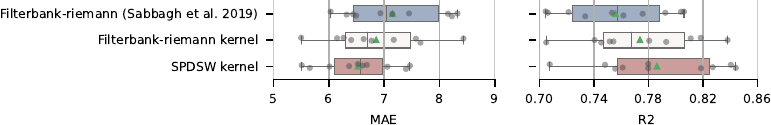}
    \caption{Results of 10-folds cross validation on the Cam-CAN data-set for one random seed.
    We display the Mean Absolute Error (MAE) and the $R^2$ coefficient.
    $\lespdsw$, with time-frames of 2s and 1000 projections, performs best.
    Note that Kernel Ridge regression based on the Log-Euclidean distance performs better than Ridge regression.}
    \label{fig:brain_age_boxplot}
\end{figure}

\paragraph{Performance of SPDSW-based brain age regression on 10-folds cross validation for one random seed.}
In \Cref{fig:brain_age_boxplot}, we display the Mean Absolute Error (MAE) and the $R^2$ coefficient on 10-folds cross validation with one random seed.
$\lespdsw$ is run with time-frames of 2s and 1000 projections.

\begin{figure}[h!]
    \centering
    \includegraphics[width=\linewidth]{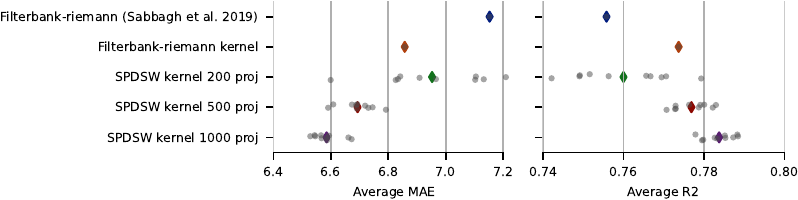}
    \caption{Average results for 10 random seeds with 200, 500 and 1000 projections for $\lespdsw$ compared to average MAE and $R^2$ obtained with Ridge and Kernel Ridge regression on features from covariance estimates \citep{sabbagh2019manifold}. With enough projections, $\lespdsw$ kernel does not suffer from variance and performs best.}
    \label{fig:variance_swspd}
\end{figure}

\paragraph{Performance of SPDSW-based brain age regression depending on number of projections.}
In \Cref{fig:variance_swspd}, we display the MAE and $R^2$ score on brain age regression with different numbers of projections for 10 random seeds.
In this example, the variance and scores are acceptable for 500 projections and more. 

\begin{figure}[h!]
    \centering
    \includegraphics[width=\linewidth]{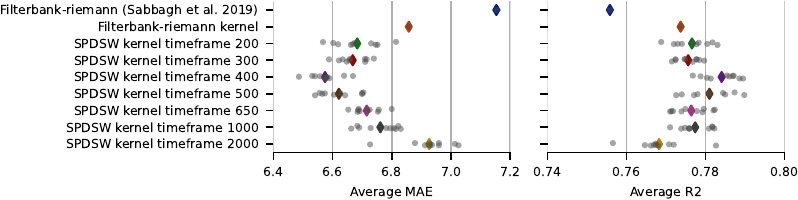}
    \caption{Average MAE and $R^2$ score on brain age regression with different time-frame lengths for 10 random seeds
    The performance depends on the time-frame length, and there is a trade-off to find between number of samples and noise in the samples.}
    \label{fig:meg_timeframes}
\end{figure}

\paragraph{Performance of SPDSW-based brain age regression depending on timeframe length.}
In \Cref{fig:meg_timeframes}, we display the MAE and $R^2$ score on brain age regression with different time-frame lengths for 10 random seeds.
The performance of $\lespdsw$-kernel Ridge regression depends on a trade-off between the number of samples in each distribution (smaller time-frames for more samples), and the level of noise in the covariance estimate (larger time-frame for less noise).
In this example, time-frames of $400$ samples seems to be a good choice.

\subsubsection{Domain Adaptation for BCI} \label{appendix:da}

\begin{figure}[h!]
    \centering
    \includegraphics[width=\columnwidth]{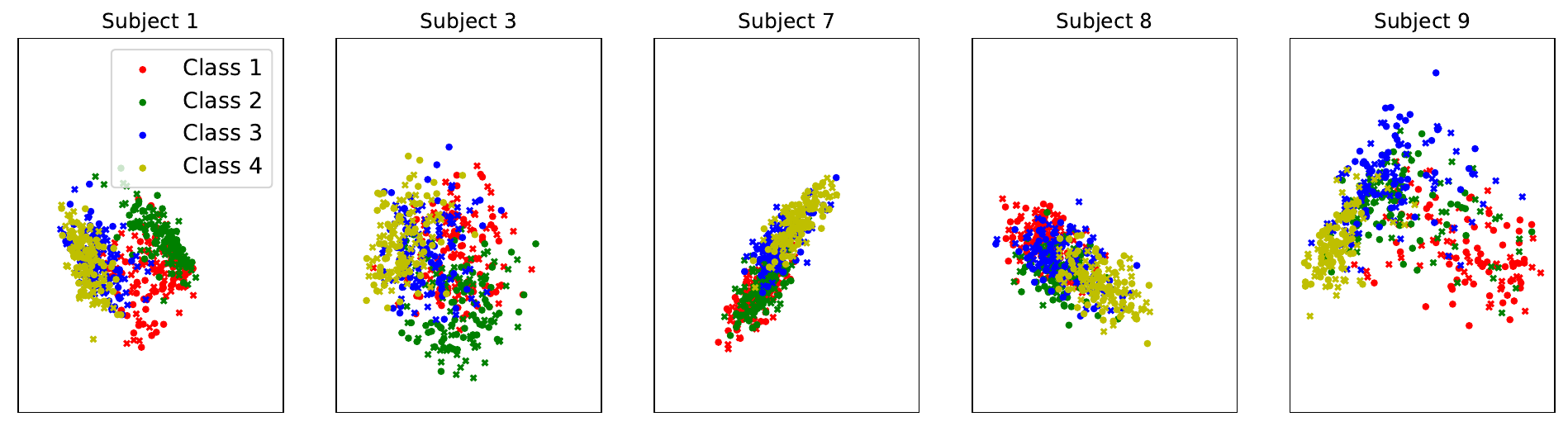}
    \caption{PCA representation on BCI data. Circles represent points from the target session and crosses points from the source after alignment.
    }
    \label{fig:classes_cross_session}
\end{figure}

\paragraph{Alignement.} We plot on \Cref{fig:classes_cross_session} the classes of the target session (circles) and of the source session after alignment (crosses) on each subject. We observe that the classes seem to be well aligned, which explains why simple transformations work on this data-set. Hence, minimizing a discrepancy allows to align the classes even without taking them into account in the loss. More complicated data-sets might require taking into account the classes for the alignment.

\paragraph{Cross Subject Task.}

In \Cref{tab:cross_subject}, we add the results obtained on the cross subject task. On the column ``subjects'', we denote the source subject, and we report in the table the mean of the accuracies obtained over all other subjects as targets.
The results for AISTODA are taken from \citet[Table 1.b, Alg.1 (u)]{yair2019domain}.
The preprocessing and hyperparameters might not be the same as in our setting.

\begin{table*}[t]
    \centering
    \caption{Accuracy and Runtime for Cross Subject.}
    \small
    \resizebox{\linewidth}{!}{
        \begin{tabular}{ccccccccccccc}
             Subjects & Source & AISOTDA & & SPDSW & LogSW & LEW & LES & & SPDSW & LogSW & LEW & LES \\
             & & \citep{yair2019domain} & & \multicolumn{4}{c}{Transformations in $S_d^{++}(\mathbb{R})$} & & \multicolumn{4}{c}{Descent over particles}\\ \toprule
            1 & 42.09 & 62.94 & & 61.91 & 60.50 & 62.89 & 63.64 & & 62.56 & 61.91 & 62.84 & 63.25 \\
            3 & 35.62 & 71.01 & & 66.40 & 66.53 & 66.34 & 66.30 & & 65.74 & 64.96 & 60.27 & 62.29 \\
            7 & 39.52 & 63.98 & & 60.42 & 57.29 & 60.89 & 60.43 & & 60.97 & 58.49 & 53.18 & 59.52 \\
            8 & 42.90 & 66.06 & & 61.09 & 60.19 & 61.29 & 62.14 & & 60.95 & 60.00 & 61.68 & 61.77 \\
            9 & 29.94 & 59.18 & & 53.31 & 50.63 & 54.79 & 54.89 & & 58.72 & 54.91 & 58.22 & 64.90\\
            \midrule 
            Avg. acc. & 38.01 & 64.43 & & 60.63 & 59.03 & 61.24 & 61.48 & & 61.79 & 60.05 & 59.24 & 62.55 \\
            Avg. time & - & - & & \textbf{4.34} & \textbf{4.31} & 11.76 & 11.21 & & \textbf{3.67} & \textbf{3.64} & 9.54 & 10.32 \\
            \bottomrule
        \end{tabular}
    }
    \label{tab:cross_subject}
\end{table*}

We add on Table \ref{tab:details_cross_subjects} 
the detailed accuracies between subjects (with on the rows the Table, and on the columns the targets) for SPDSW, LEW, 
and when applying the classifier on the source.

\begin{figure}[H]
    \centering
    \captionof{table}{Accuracy between subjects. The rows denote the source and the columns the targets.}
    \label{tab:details_cross_subjects}
    \begin{minipage}{0.32\linewidth}
        \centering
        \captionof{table}{Source.}
        \small
        \resizebox{\columnwidth}{!}{
            \begin{tabular}{cccccc}
                & 1 & 3 & 7 & 8 & 9    \\ 
                \toprule
                1 & - & 52.22 & 50.55 & 39.02 & 26.58 \\
                3 & 34.43 & - & 30.10 & 49.62 & 27.43 \\
                7 & 52.01 & 53.33 & - & 26.14 & 26.58 \\
                8 & 49.82 & 57.78 & 24.35 & 0 & 39.66 \\
                9 & 26.74 & 28.52 & 24.72 & 39.39 & - \\
                \bottomrule
            \end{tabular}
        }
        \label{tab:cross_subject_src}
    \end{minipage}
    \hfill
    \begin{minipage}{0.32\linewidth}
        \centering
        \captionof{table}{Particles + $\mathrm{SPDSW}$.}
        \small
        \resizebox{\columnwidth}{!}{
            \begin{tabular}{cccccc}
                & 1 & 3 & 7 & 8 & 9    \\ 
                \toprule
                1 & - & 69.04 & 60.89 & 68.18 & 52.15 \\
                3 & 66.23 & - & 70.18 & 70.83 & 55.70 \\
                7 & 58.02 & 71.04 & - & 61.82 & 53.00 \\
                8 & 57.73 & 70.44 & 58.16 & - & 57.47 \\
                9 & 55.24 & 61.85 & 52.10 & 65.68 & - \\
                \bottomrule
            \end{tabular}
        }
        \label{tab:cross_subject_particles_spdsw}
    \end{minipage}
    \hfill
    \begin{minipage}{0.32\linewidth}
        \centering
        \captionof{table}{Particles + LEW.}
        \small
        \resizebox{\columnwidth}{!}{
            \begin{tabular}{cccccc}
                & 1 & 3 & 7 & 8 & 9    \\ 
                \toprule
                1 & - & 72.59 & 55.42 & 69.32 & 54.01 \\
                3 & 63.37 & - & 61.99 & 62.12 & 53.59 \\
                7 & 50.18 & 62.96 & - & 48.11 & 51.48 \\
                8 & 61.54 & 74.07 & 53.87 & - & 57.22 \\
                9 & 48.35 & 63.33 & 57.20 & 64.02 & - \\
                \bottomrule
            \end{tabular}
        }
        \label{tab:cross_subject_particles_lew}
    \end{minipage}
    
    \vspace{15pt}
    
    \begin{minipage}{0.3\linewidth}
        \centering
        \small
        \resizebox{\columnwidth}{!}{
            \begin{tabular}{cccccc}
            \end{tabular}
        }
        \label{tab:cross_subject_aisotda}
    \end{minipage}
    \hfill
    \begin{minipage}{0.3\linewidth}
        \centering
        \captionof{table}{Transf. + $\mathrm{SPDSW}$.}
        \small
        \resizebox{\columnwidth}{!}{
            \begin{tabular}{cccccc}
                & 1 & 3 & 7 & 8 & 9    \\ 
                \toprule
                1 & - & 68.00 & 59.04 & 68.79 & 51.81 \\
                3 & 68.42 & - & 71.07 & 69.24 & 56.88 \\
                7 & 57.66 & 69.78 & - & 60.83 & 53.42 \\
                8 & 62.71 & 72.07 & 53.87 & - & 55.70 \\
                9 & 53.92 & 59.04 & 40.15 & 60.15 & - \\
                \bottomrule
            \end{tabular}
        }
        \label{tab:cross_subject_transfs_spdsw}
    \end{minipage}
    \hfill
    \begin{minipage}{0.3\linewidth}
        \centering
        \captionof{table}{Transf. + LEW.}
        \small
        \resizebox{\columnwidth}{!}{
            \begin{tabular}{cccccc}
                & 1 & 3 & 7 & 8 & 9    \\ 
                \toprule
                1 & - & 70.00 & 59.78 & 68.18 & 53.59 \\
                3 & 69.60 & - & 71.59 & 69.32 & 54.85 \\
                7 & 57.88 & 73.37 & - & 61.74 & 53.59 \\
                8 & 63.00 & 72.22 & 54.24 & - & 55.70 \\
                9 & 55.31 & 60.00 & 39.48 & 64.02 & - \\
                \bottomrule
            \end{tabular}
        }
        \label{tab:cross_subject_transfs_lew}
    \end{minipage}
\end{figure}

\paragraph{Evolution of the accuracy \emph{w.r.t} the number of projections.}

\begin{figure}[t]
    \centering
    \hspace*{\fill}
    \subfloat[Transformations on cross-subjects.]{\label{fig:acc_projs_transfs_subject}\includegraphics[width=0.3\columnwidth]{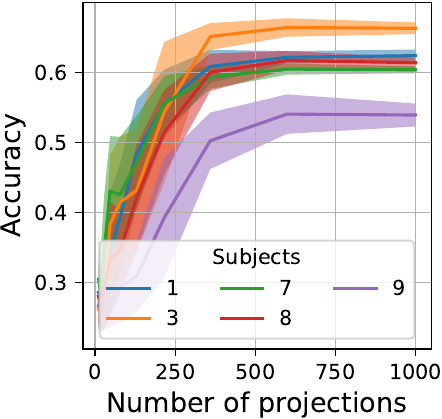}} \hfill
    \subfloat[Particles on cross-session.]{\label{fig:acc_projs_particles_session}\includegraphics[width=0.3\columnwidth]{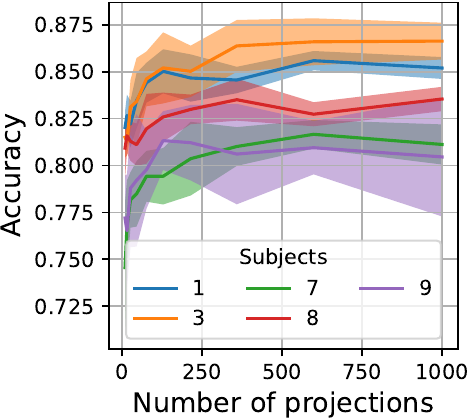}} \hfill
    \subfloat[Particles on cross-subject.]{\label{fig:acc_projs_particles_subject}\includegraphics[width=0.3\columnwidth]{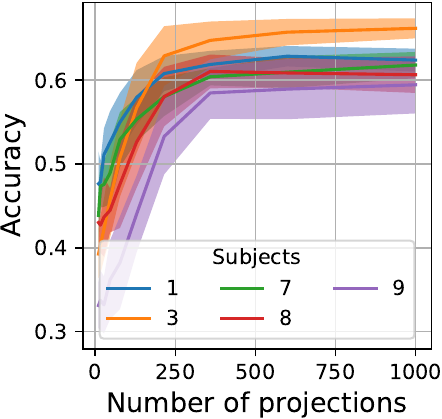}} \hfill
    \hspace*{\fill}
    \caption{Accuracy \emph{w.r.t} the number of projections when optimizing over particles or transformations, and for the cross-session task and cross subject task.
    In all cases, the accuracy converges for 500 projections.}
    \label{fig:acc_projs}
\end{figure}

On \Cref{fig:pca_acc_projs}, we plot the evolution of the accuracy obtained by learning transformations on $S_d^{++}(\mathbb{R})$ on the cross session task. We report on Figure \ref{fig:acc_projs} the plot for the other cases. We compared the results for $L\in \{10, 16, 27, 46, 77, 129, 215, 359, 599, 1000\}$ projections, which are evenly spaced in log scale. Other parameters are the same as in \Cref{tab:cross_session} and are detailed in \Cref{sec:exp_details_bci}. The results were averaged over 10 runs, and we report the standard deviation.

\subsubsection{Illustrations}

\begin{figure}[t]
    \centering
    \hspace*{\fill}
    \subfloat[Sample complexity of $D=\lespdsw$ and $D=\mathrm{LEW}$ for $d=2$ and $d=50$.]{\label{fig:sample_complexity}\includegraphics[width=0.48\columnwidth]{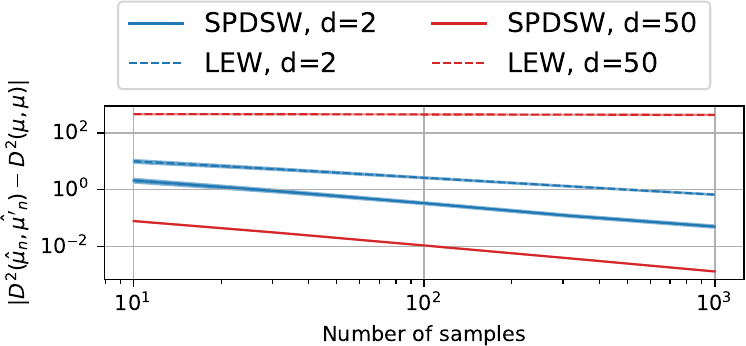}} \hfill
    \subfloat[Projection complexity of $\lespdsw$ and the $\logsw$ for $d=2$ and $d=20$.]{\label{fig:proj_complexity}\includegraphics[width=0.48\columnwidth]{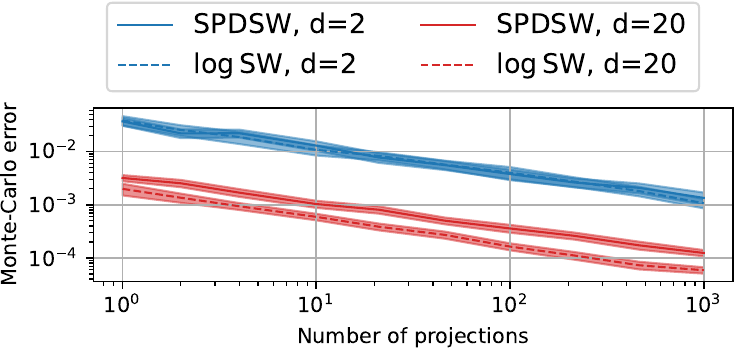}} \hfill
    \hspace*{\fill}
    \caption{Sample and projection complexity. Experiments are replicated 100 times and we report the 95\% confidence intervals. We note $\hat{\mu}_n$ and $\hat{\mu}'_n$ two different empirical distributions of $\mu$. The sample complexity of $\lespdsw$ does not depend on the dimension contrary to Wasserstein. The projections complexity has a slope which decreases in $O(\frac{1}{\sqrt{L}})$.}
    \label{fig:sample_proj_complexities}
\end{figure}

\paragraph{Sample Complexity.} We illustrate \Cref{prop:chsw_sample_complexity} in the particular case of $\lespdsw$ in \Cref{fig:sample_complexity} by plotting $\lespdsw$ and the Wasserstein distance with Log-Euclidean ground cost (LEW) between samples drawn from the same Wishart distribution, for $d=2$ and $d=50$. $\lespdsw$ is computed with $L=1000$ projections. We observe that $\lespdsw$ converges with the same speed in both dimensions while LEW converges slower in dimension 50.


\paragraph{Projection Complexity.} We illustrate \Cref{prop:chsw_proj_complexity} in the particular case of $\lespdsw$ in \Cref{fig:proj_complexity} by plotting the absolute error between $\widehat{\lespdsw}_{2,L}^2$ and $\widehat{\lespdsw}_{2,L^*}^2$. We fix $L^*$ at 10000 which gives a good idea of the true value of $\lespdsw$ and we vary $L$ between $1$ and $10^3$ evenly in log scale. We average the results over 100 runs and plot 95\% confidence intervals. We observe that the Monte-Carlo error converges to 0 with a convergence rate of $O(\frac{1}{\sqrt{L}})$.


\subsection{Experimental details}
\label{sec:exp_details}

\subsubsection{Runtime}

In \Cref{fig:spdsw_runtime}, we plot the runtime \emph{w.r.t} the number of samples for different OT discrepancies. Namely, we compare $\lespdsw$, $\aispdsw$, $\logsw$, the Wasserstein distance with Affine-Invariant ground cost, the Wasserstein distance with Log-Euclidean ground cost, and the Sinkhorn algorithm used to compute the entropic regularized OT problem with Log-Euclidean ground cost. The distance ground costs are computed with \texttt{geoopt} \citep{kochurov2020geoopt} while Wasserstein and Sinkhorn are computed with \texttt{POT} \citep{flamary2021pot}. All computations are done on a A6000 GPU. We average the results over 20 runs and for $n\in\{100,215,464,1000,2154,4641,10000,21544,46415,100000\}$ samples, which are evenly spaced in log scale, from a Wishart distribution in dimension $d=20$. For the sliced methods, we fix $L=200$ projections. For the Sinkhorn algorithm, we use a stopping threshold of $10^{-10}$ with maximum $10^5$ iterations and a regularization parameter of $\epsilon = 1$.

\subsubsection{Brain Age Prediction}
\label{subsec:brain_age_prediction_details}

We reuse the code for preprocessing steps and benchmarking procedure described in \citet{engemann2022reusable} for the CamCAN data-set, and available at \url{https://github.com/meeg-ml-benchmarks/brain-age-benchmark-paper}, which we recall here.

The data consist of measurements from 102 magnetometers and 204 gradiometers.
First, we apply a band-pass filtering between 0.1Hz and 49Hz.
Then, the signal is subsampled with a decimation factor of 5, leading to a sample frequency of 200Hz.
Then, we apply the temporal signal-space-separation (tSSS).
Default settings were applied for the harmonic decomposition (8 components of the internal sources, 3 for the external sources) on a 10-s sliding window.
To discard segments for which inner and outer signal components were poorly distinguishable, we applied a correlation threshold of 98\%.

For analysis, the band frequencies used are the following: (0.1Hz, 1Hz), (1Hz, 4Hz), (4Hz, 8Hz), (8Hz, 15Hz), (15Hz, 26Hz), (26Hz, 35Hz), (35Hz, 49Hz).
The rank of the covariance matrices obtained after OAS is reduced to 53 with a PCA, which leads to the best score on this problem as mentioned in \citet{sabbagh2020predictive}.

The code for the MEG experiments is essentially based on the work by \citet{engemann2022reusable}, the class \texttt{SPDSW} available in the supplementary material, and the Kernel Ridge Regression of \texttt{scikit-learn}.
The full version will be added later in order to respect anonymity.

\subsubsection{Domain Adaptation for BCI} \label{sec:exp_details_bci}

For both the optimization over particles and over transformations, we use \texttt{geoopt} \citep{kochurov2020geoopt} with the Riemannian gradient descent. We now detail the hyperparameters and the procedure.

First, the data from the BCI Competition IV 2a are preprocessed using the code from \citet{hersche2018fast} available at \url{https://github.com/MultiScale-BCI/IV-2a}. We applied a band-pass filter between 8 and 30 Hz. With these hyper-parameters, we get one regularized covariance matrix per subject.

For all experiments, we report the results averaged over 5 runs. For the sliced discrepancies, we always use $L=500$ projections which we draw only once a the beginning. When optimizing over particles, we used a learning rate of $1000$ for the sliced methods and of $10$ for Wasserstein and Sinkhorn. The number of epochs was fixed at 500 for the cross-session task and for the cross-subject tasks. For the basic transformations, we always use 500 epochs and we choose a learning rate of $1e^{-1}$ on cross session and $5e^{-1}$ on cross subject for sliced methods, and of $1e^{-2}$ for Wasserstein and Sinkhorn. For the Sinkhorn algorithm, we use $\epsilon=10$ with the default hyperparameters from the \texttt{POT} implementation. Moreover, we only use one translation and rotation for the transformation.

Furthermore, the results reported for AISOTDA in \Cref{tab:cross_session} and \Cref{tab:cross_subject} are taken from \citet{yair2019domain} (Table 1.a, column Alg.1 (u)). We note however that they may not have used the same preprocessing and hyperparameters to load the covariance matrices.

\newpage

\section{Appendix of \Cref{chapter:ssw}}

\subsection{Proofs of \Cref{section:ssw}}

\subsubsection{Proof of \Cref{prop:w2_unif_circle}} \label{proof:prop_w2_unif_circle}

\begin{proof}[Proof of \Cref{prop:w2_unif_circle}] 
    \textbf{Optimal $\alpha$.} Let $\mu\in \mathcal{P}_2(S^1)$, $\nu=\mathrm{Unif}(S^1)$. Since $\nu$ is the uniform distribution on $S^1$, its cdf is the identity on $[0,1]$ (where we identified $S^1$ and $[0,1]$). We can extend the cdf $F$ on the real line as in \citep{rabin2011transportation} with the convention $F(y+1)=F(y)+1$. Therefore, $F_\nu = \mathrm{Id}$ on $\mathbb{R}$. Moreover, we know that for all $x\in S^1$, $(F_\nu-\alpha)^{-1}(x)=F_\nu^{-1}(x+\alpha)=x+\alpha$ and 
    \begin{equation}
        W_2^2(\mu,\nu) = \inf_{\alpha\in\mathbb{R}}\ \int_0^1 |F_\mu^{-1}(t)-(F_\nu-\alpha)^{-1}(t)|^2 \ \mathrm{d}t.
    \end{equation}
    
    For all $\alpha\in\mathbb{R}$, let $f(\alpha) = \int_0^1 \big(F_\mu^{-1}(t) - (F_\nu-\alpha)^{-1}(t)\big)^2 \ \mathrm{d}t$. Then, we have:
    \begin{equation}
        \begin{aligned}
            \forall \alpha\in \mathbb{R},\ f(\alpha) &= \int_0^1 \big(F_\mu^{-1}(t)-t-\alpha\big)^2\ \mathrm{d}t \\
            &= \int_0^1 \big(F_\mu^{-1}(t)-t\big)^2\ \mathrm{d}t + \alpha^2 - 2\alpha \int_0^1 (F_\mu^{-1}(t)-t)\ \mathrm{d}t \\
            &= \int_0^1 \big(F_\mu^{-1}(t)-t\big)^2\ \mathrm{d}t + \alpha^2 -2\alpha \left( \int_0^1 x\  \mathrm{d}\mu(x) - \frac{1}{2}\right),
        \end{aligned}
    \end{equation}
    where we used that $(F_\mu^{-1})_\#\mathrm{Unif}([0,1]) = \mu$.
    
    Hence, $f'(\alpha)=0 \iff \alpha = \int_0^1 x\ \mathrm{d}\mu(x) - \frac12$.
    
    \noindent \textbf{Closed-form for empirical distributions.} Let $(x_i)_{i=1}^n \in [0,1[^n$ such that $x_1<\dots<x_n$ and let $\mu_n=\frac{1}{n}\sum_{i=1}^n \delta_{x_i}$ a discrete distribution.
    
    To compute the closed-form of $W_2$ between $\mu_n$ and $\nu=\mathrm{Unif}(S^1)$, we first have that the optimal $\alpha$ is $\alpha_n = \frac{1}{n}\sum_{i=1}^n x_i - \frac12$. Moreover, we also have:
    \begin{equation} \label{eq:w2_proof}
        \begin{aligned}
            W_2^2(\mu_n,\nu) &= \int_0^1 \big(F_{\mu_n}^{-1}(t) - (t+\hat{\alpha}_n)\big)^2 \ \mathrm{d}t \\
            &= \int_0^1 F_{\mu_n}^{-1}(t)^2\ \mathrm{d}t - 2\int_0^1 tF_{\mu_n}^{-1}(t)\mathrm{d}t - 2\hat{\alpha}_n \int_0^1 F_{\mu_n}^{-1}(t)\mathrm{d}t + \frac13 + \hat{\alpha}_n + \hat{\alpha}_n^2.
        \end{aligned}
    \end{equation}
    Then, by noticing that $F_{\mu_n}^{-1}(t) = x_i$ for all $t\in [F(x_i),F(x_{i+1})[$, we have
    \begin{equation}
        \begin{aligned}
            \int_0^1 t F_{\mu_n}^{-1}(t)\mathrm{d}t = \sum_{i=1}^n \int_{\frac{i-1}{n}}^{\frac{i}{n}} t x_i \mathrm{d}t = \frac{1}{2n^2} \sum_{i=1}^n x_i(2i-1),
        \end{aligned}
    \end{equation}
    \begin{equation}
        \int_0^1 F_{\mu}^{-1}(t)^2 \mathrm{d}t = \frac{1}{n}\sum_{i=1}^n x_i^2, \quad \int_0^1 F_\mu^{-1}(t)\mathrm{d}t = \frac{1}{n}\sum_{i=1}^n x_i,
    \end{equation}
    and we also have:
    \begin{equation}
        \hat{\alpha}_n + \hat{\alpha}_n^2 = \frac{1}{n} \sum_{i=1}^n x_i - \frac12 + \Big(\frac{1}{n}\sum_{i=1}^n x_i\Big)^2 + \frac14 - \frac{1}{n}\sum_{i=1}^n x_i =  \Big(\frac{1}{n}\sum_{i=1}^n x_i\Big)^2 -\frac14.
    \end{equation}
    Then, by plugging these results into \eqref{eq:w2_proof}, we obtain
    \begin{equation}
        \begin{aligned}
            W_2^2(\mu_n,\nu) &= \frac{1}{n}\sum_{i=1}^n x_i^2 - \frac{1}{n^2}\sum_{i=1}^n (2i-1)x_i - 2 \Big(\frac{1}{n}\sum_{i=1}^n x_i\Big)^2 + \frac{1}{n}\sum_{i=1}^n x_i + \frac13 + \Big(\frac{1}{n}\sum_{i=1}^n x_i\Big)^2 - \frac14 \\
            &= \frac{1}{n}\sum_{i=1}^n x_i^2 - \Big(\frac{1}{n}\sum_{i=1}^n x_i\Big)^2 + \frac{1}{n^2}\sum_{i=1}^n (n+1-2i)x_i + \frac{1}{12}. 
        \end{aligned}
    \end{equation}
\end{proof}

\subsubsection{Proof of the closed-form \eqref{eq:proj_S1}}

Here, we show the equality derived in \eqref{eq:proj_S1}, which we recall:
\begin{equation}
    \forall U\in\mathbb{V}_{d,2}, \forall x\in S^{d-1},\ P^U(x) = U^T\argmin_{y \in \mathrm{span}(UU^T)\cap S^{d-1}}\ d_{S^{d-1}}(x,y) = \argmin_{z\in S^1}\ d_{S^{d-1}}(x,Uz).
\end{equation}

\begin{proof}
    Let $U\in\mathbb{V}_{d,2}$. Then the great circle generated by $U\in\mathbb{V}_{d,2}$ is defined as the intersection between $\mathrm{span}(UU^T)$ and $S^{d-1}$. And we have the following characterization:
    \begin{equation*}
        \begin{aligned}
            x \in \mathrm{span}(UU^T)\cap S^{d-1} &\iff \exists y\in \mathbb{R}^d,\ x = UU^T y\ \text{and}\ \|x\|_2^2 = 1 \\
            &\iff \exists y \in \mathbb{R}^d,\ x=UU^T y\ \text{and}\ \|UU^T y\|_2^2 = y^T UU^T y = \|U^Ty\|_2^2 = 1 \\
            &\iff \exists z\in S^{1},\ x=Uz.
        \end{aligned}
    \end{equation*}
    
    And we deduce that 
    \begin{equation}
        \forall U\in \mathbb{V}_{d,2}, x\in S^{d-1}, \ P^U(x) = \argmin_{z\in S^1}\ d_{S^{d-1}} (x, Uz).
    \end{equation}
\end{proof}

\subsubsection{Proof of \Cref{lemma:proj_closeform}} \label{proof:lemma_proj_closedform_ssw}

\begin{proof}[Proof of \Cref{lemma:proj_closeform}]
    Let $U\in\mathbb{V}_{d,2}$ and $x\in S^{d-1}$ such that $U^Tx\neq 0$. Denote $U=(u_1\ u_2)$, \emph{i.e.} the $2$-plane $E$ is $E=\mathrm{span}(UU^T) = \mathrm{span}(u_1, u_2)$ and $(u_1,u_2)$ is an orthonormal basis of $E$. Then, for all $x\in S^{d-1}$, the projection on $E$ is $p^E(x)=\langle u_1,x\rangle u_1 + \langle u_2,x\rangle u_2 = UU^T x$.
    
    Now, let us compute the geodesic distance between $x\in S^{d-1}$ and $\frac{p^E(x)}{\|p^E(x)\|_2}\in E\cap S^{d-1}$:
    \begin{equation}
        d_{S^{d-1}}\left(x, \frac{p^E(x)}{\|p^E(x)\|_2}\right) = \arccos\left(\langle x, \frac{p^E(x)}{\|p^E(x)\|_2}\rangle\right) = \arccos(\|p^E(x)\|_2),
    \end{equation}
    using that $x=p^E(x)+p^{E^\bot}(x)$.
    
    Let $y\in E\cap S^{d-1}$ another point on the great circle. By the Cauchy-Schwarz inequality, we have
    \begin{equation}
        \langle x,y\rangle = \langle p^E(x), y\rangle \le \|p^E(x)\|_2\|y\|_2 = \|p^E(x)\|_2.
    \end{equation}
    Therefore, using that $\arccos$ is decreasing on $(-1,1)$,
    \begin{equation}
        d_{S^{d-1}}(x,y) = \arccos(\langle x,y\rangle) \ge \arccos(\|p^E(x)\|_2) = d_{S^{d-1}}\left(x,\frac{p^E(x)}{\|p^E(x)\|_2}\right).
    \end{equation}
    Moreover, we have equality if and only if $y=\lambda p^E(x)$. And since $y\in S^{d-1}$, $|\lambda| = \frac{1}{\|p^E(x)\|_2}$. Using again that $\arccos$ is decreasing, we deduce that the minimum is well attained in $y=\frac{p^E(x)}{\|p^E(x)\|_2} = \frac{UU^Tx}{\|UU^Tx\|_2}$.
    
    Finally, using that $\|UU^Tx\|_2 = x^T UU^T UU^Tx = x^T UU^T x= \|U^Tx\|_2$, we deduce that
    \begin{equation}
        P^U(x) = \frac{U^Tx}{\|U^Tx\|_2}.
    \end{equation}
    
    Finally, by noticing that the projection is unique if and only if $U^Tx=0$, and using \citep[Proposition 4.2]{bardelli2017probability} which states that there is a unique projection for a.e. $x$, we deduce that $\{x\in S^{d-1},\ U^Tx=0\}$ is of measure null and hence, for a.e. $x\in S^{d-1}$, we have the result.
\end{proof}

\subsection{Proofs of \Cref{section:spherical_radon}}

\subsubsection{Proof of \Cref{prop:ssw_distance}} \label{proof:prop_ssw_distance}

\begin{proof}[Proof of \Cref{prop:ssw_distance}]
    Let $p\ge 1$.
    First, it is straightforward to see that for all $\mu,\nu\in\mathcal{P}_p(S^{d-1})$, $\ssw_p(\mu,\nu) \ge 0$, $\ssw_p(\mu,\nu)=\ssw_p(\nu,\mu)$, $\mu=\nu \implies \ssw_p(\mu,\nu)=0$ and that we have the triangular inequality since
    \begin{equation}
        \begin{aligned}
            \forall \mu,\nu,\alpha\in\mathcal{P}_p(S^{d-1}),\ \ssw_p(\mu,\nu) &= \Big(\int_{\mathbb{V}_{d,2}} W_p^p(P^U_\#\mu,P^U_\#\nu)\ \mathrm{d}\sigma(U)\Big)^{\frac{1}{p}} \\
            &\le \Big(\int_{\mathbb{V}_{d,2}} \big(W_p(P^U_\#\mu,P^U_\#\alpha) + W_p(P^U_\#\alpha,P^U_\#\nu)\big)^p\ \mathrm{d}\sigma(U)\Big)^{\frac{1}{p}} \\
            &\le \Big(\int_{\mathbb{V}_{d,2}} W_p^p(P^U_\#\mu,P^U_\#\alpha)\ \mathrm{d}\sigma(U)\Big)^{\frac{1}{p}} \\ &+ \Big(\int_{\mathbb{V}_{d,2}} W_p^p(P^U_\#\alpha,P^U_\#\nu)\ \mathrm{d}\sigma(U)\Big)^{\frac{1}{p}} \\
            &= \ssw_p(\mu,\alpha)+\ssw_p(\alpha,\nu),
        \end{aligned}
    \end{equation}
    using the triangular inequality for $W_p$ and the Minkowski inequality.
    Therefore, it is at least a pseudo-distance.
    
\end{proof}

\subsubsection{Proof of \Cref{prop:integration_set_ssw}} \label{proof:prop_integration_set_ssw}

\begin{proof}[Proof of \Cref{prop:integration_set_ssw}]
    Let $U\in\mathbb{V}_{d,2}$, $z\in S^1$. Denote $E=\mathrm{span}(UU^T)$ the 2-plane generating the great circle, and $E^\bot$ its orthogonal complementary. Hence, $E\oplus E^\bot = \mathbb{R}^d$ and $\mathrm{dim}(E^\bot) = d-2$. Now, let $F=E^\bot \oplus \mathrm{span}(Uz)$. Since $Uz = UU^TUz\in E$, we have that $\mathrm{dim}(F)=d-1$. Hence, $F$ is a hyperplane and $F\cap S^{d-1}$ is a $(d-2)$-dimensional subsphere of $S^{d-1}$.
    
    Now, for the first inclusion, let $x\in\{x\in S^{d-1},\ P^U(x)=z\}$. First, we show that $x\in F\cap S^{d-1}$. By Lemma \ref{lemma:proj_closeform} and hypothesis, we know that $P^U(x)=\frac{U^Tx}{\|U^Tx\|_2} = z$. By denoting by $p^E$ the projection on $E$, we have:
    \begin{equation}
        p^E(x) = UU^Tx=U(\|U^Tx\|_2 z) = \|U^Tx\|_2 Uz \in \mathrm{span}(Uz).
    \end{equation}
    Hence, $x=p^E(x)+x_{E^\bot} = \|U^Tx\|_2 Uz + x_{E^\bot} \in F$. Moreover, as 
    \begin{equation}
        \langle x,Uz\rangle = \|U^Tx\|_2 \langle Uz,Uz\rangle = \|U^Tx\|_2 >0,
    \end{equation}
    we deduce that $x\in \{F\cap S^{d-1},\ \langle x,Uz\rangle>0\}$.
    
    For the other inclusion, let $x\in \{F\cap S^{d-1},\ \langle x,Uz\rangle>0\}$. Since $x\in F$, we have $x=x_{E^\bot} + \lambda Uz$, $\lambda\in\mathbb{R}$. Hence, using Lemma \ref{lemma:proj_closeform},
    \begin{equation}
        P^U(x) = \frac{U^Tx}{\|U^Tx\|_2} = \frac{\lambda}{|\lambda|} \frac{z}{\|z\|_2} = \mathrm{sign}(\lambda) z.
    \end{equation}
    But, we also have $\langle x,Uz\rangle=\lambda \|Uz\|_2^2 = \lambda >0$. Therefore, $\mathrm{sign}(\lambda)=1$ and $P^U(x)=z$.
    
    Finally, we conclude that $\{x\in S^{d-1},\ P^U(x)=z\rangle\} = \{x\in F\cap S^{d-1},\ \langle x,Uz\rangle >0\}$.
\end{proof}

\subsubsection{Proof of \Cref{prop:radon_dual_ssw}} \label{proof:prop_radon_dual_ssw}

\begin{proof}[Proof of \Cref{prop:radon_dual_ssw}]

    Let $f\in L^1(S^{d-1})$, $g\in C_b(S^1\times \mathbb{V}_{d,2})$, and denote $\Tilde{f}$ such that for all $y\in S^{d-1}$, $\Tilde{f}(y) = f(O_U^T y)$ with $O_U\in SO(d)$ the rotation defined such that for any $z\in S^1$, $O_U Uz\in\mathrm{span}(e_{d-1},e_{d})$. Also, note that $P^{U_0}(O_U x) = \frac{U_0^T O_U x}{\|U_0^T O_U x\|_2} = \frac{U^T x}{\|U^Tx\|_2} = P^U(x)$. Then,
    \begin{equation}
        \begin{aligned}
            &\langle \Tilde{R}f, g\rangle_{S^1\times \mathbb{V}_{d,2}} \\ &= \int_{S^1}\int_{\mathbb{V}_{d,2}} \Tilde{R}f(z,U) g(z,U)\ \mathrm{d}\sigma(U)\mathrm{d}\sigma_1(z) \\
            &= \int_{S^1} \int_{\mathbb{V}_{d,2}} \Tilde{R}\Tilde{f}_U(z,U_0) g(z,U)\ \mathrm{d}\sigma(U) \mathrm{d}\sigma_1(z) \\
            &= \int_0^{2\pi} \int_{\mathbb{V}_{d,2}} \Tilde{R}\Tilde{f}_U((\cos\theta_{d-1}, \sin\theta_{d-1}), U_0) g((\cos\theta_{d-1},\sin\theta_{d-1}), U)\ \mathrm{d}\sigma(U)\mathrm{d}\theta_{d-1} \\
            &= \int_0^{2\pi} \int_{\mathbb{V}_{d,2}} \int_{[0,\pi]^{d-2}} \Tilde{f}_U(\varphi(\theta_1,\dots,\theta_{d-1})) g((\cos\theta_{d-1},\sin\theta_{d-1}), U) \prod_{i=1}^{d-2} \sin(\theta_i)^{d-i-1}\ \mathrm{d}\theta_1\dots\mathrm{d}\theta_{d-2} \mathrm{d}\sigma(U) \mathrm{d}\theta_{d-1} \\
            &= \int_{\mathbb{V}_{d,2}} \int_{S^{d-1}} \Tilde{f}_U(y) g(P^{U_0}(y), U) \ \mathrm{d}\sigma_d(y)\mathrm{d}\sigma(U) \quad \text{ using $y=\varphi(\theta_1,\dots,\theta_{d-1})$} \\
            &= \int_{\mathbb{V}_{d,2}} \int_{S^{d-1}} f(O_U^T y) g(P^{U_0}(y), U) \ \mathrm{d}\sigma_d(y)\mathrm{d}\sigma(U) \\
            &= \int_{\mathbb{V}_{d,2}} \int_{S^{d-1}} f(x) g(P^{U_0}(O_Ux), U)\ \mathrm{d}\sigma_d(x)\mathrm{d}\sigma(U) \quad \text{using $x=O_U^T y$ and rotational invariance of $\sigma_d$} \\
            &= \int_{\mathbb{V}_{d,2}} \int_{S^{d-1}} f(x) g(P^U(x), U)\ \mathrm{d}\sigma_d(x)\mathrm{d}\sigma(U) \quad \text{using that $U=O_U^T U_0$} \\
            &= \int_{S^{d-1}} f(x) \Tilde{R}^*g(x)\ \mathrm{d}\sigma_d(x) \\
            &= \langle f, \Tilde{R}^*g\rangle_{S^{d-1}}.
        \end{aligned}
    \end{equation}

\end{proof}

\subsubsection{Proof of \Cref{prop:radon_disintegrated}} \label{proof:prop_radon_disintegrated}

\begin{proof}[Proof of \Cref{prop:radon_disintegrated}]
    Let $g\in C_b(S^1 \times \mathbb{V}_{d,2})$,b y applying the Fubini theorem,
    \begin{equation}
        \begin{aligned}
            \ \int_{\mathbb{V}_{d,2}} \int_{S^1} g(z,U) \ (\Tilde{R}\mu)^U(\mathrm{d}z)\ \mathrm{d}\sigma(U) &= \int_{S^1 \times \mathbb{V}_{d,2}} g(z,U)\ \mathrm{d}(\Tilde{R}\mu)(z,U) \\
            &= \int_{S^{d-1}} \Tilde{R}^*g(x) \ \mathrm{d}\mu(x) \\
            &= \int_{S^{d-1}} \int_{\mathbb{V}_{d,2}} g(P^U(x), U)\ \mathrm{d}\sigma(U) \mathrm{d}\mu(x) \\
            &= \int_{\mathbb{V}_{d,2}} \int_{S^{d-1}} g(P^U(x), U) \ \mathrm{d}\mu(x) \mathrm{d}\sigma(U) \\
            &= \int_{\mathbb{V}_{d,2}} \int_{S^1} g(z, U) \ \mathrm{d}(P^U_\#\mu)(z)\mathrm{d}\sigma(U).
        \end{aligned}
    \end{equation}
    Hence, for $\sigma$-almost every $U\in\mathbb{V}_{d,2}$, $(\Tilde{R}\mu)^U = P^U_\#\mu$.
\end{proof}

\subsubsection{Proof of \Cref{prop:link_hemispherical}} \label{proof:prop_link_hemispherical}

\begin{proof}[Proof of \Cref{prop:link_hemispherical}]
    Let $f\in L^1(S^{d-1})$, $z\in S^1$, $U\in\mathbb{V}_{d,2}$, then by \Cref{prop:integration_set_ssw},
    \begin{equation}
        \begin{aligned}
            \Tilde{R}f(z,U) &= \int_{S^{d-1}\cap F} f(x) \mathbb{1}_{\{\langle x,Uz\rangle>0\}}\mathrm{d}\vol (x).
        \end{aligned}
    \end{equation}
    $F$ is a hyperplane. Let $O\in\mathbb{R}^{d\times d}$ be the rotation such that for all $x\in F$, $Ox\in\mathrm{span}(e_1,\dots,e_{d-1})=\Tilde{F}$ where $(e_1,\dots,e_d)$ is the canonical basis. By applying the change of variable $Ox=y$, and since the surface measure is rotationally invariant, we obtain
    \begin{equation}
        \Tilde{R}f(z,U) = \int_{O(F\cap S^{d-1})} f(O^Ty) \mathbb{1}_{\{\langle O^Ty, Uz\rangle>0\}} \mathrm{d}\vol (y) = \int_{\Tilde{F}\cap S^{d-1}} f(O^Ty)\mathbb{1}_{\{\langle y, OUz\rangle>0\}}\mathrm{d}\vol(y).
    \end{equation}
    Now, we have that $OU\in\mathbb{V}_{d,2}$ since $(OU)^T(OU) = I_2$, and since $Uz\in F$, $OUz\in \Tilde{F}$. For all $y\in \Tilde{F}$, we have $\langle y,e_d\rangle = y_d = 0$. Let $J=\begin{pmatrix}I_{d-1} \\ 0_{1, d-1} \end{pmatrix}\in\mathbb{R}^{d\times (d-1)}$, then for all $y\in \Tilde{F}\cap S^{d-1}$, $y=J\Tilde{y}$ where $\Tilde{y}\in S^{d-2}$ is composed of the $d-1$ first coordinates of $y$.
    
    Let's define, for all $\Tilde{y}\in S^{d-2}$, $\Tilde{f}(\Tilde{y}) = f(O^T J\Tilde{y})$, $\Tilde{U}=J^TOU$.
    
    Then, since $\Tilde{F}\cap S^{d-1} \cong S^{d-2}$, we can write:
    \begin{equation}
        \Tilde{R}f(z,U) = \int_{S^{d-2}} \Tilde{f}(\Tilde{y}) \mathbb{1}_{\{\langle \Tilde{y}, \Tilde{U}z\rangle > 0\}}\mathrm{d}\vol(\Tilde{y}) = \mathcal{H}_{d-2}\Tilde{f}(\Tilde{U}z).
    \end{equation}
\end{proof}

\subsubsection{Proof of \Cref{prop:kernel_radon_ssw}} \label{proof:prop_kernel_radon_ssw}

First, we recall Lemma 2.3 of \citep{rubin1999inversion} on $S^{d-2}$. In the following, we omit the indices for $\mathcal{H}$ which is always on $S^{d-2}$. Note that for $\mu\in\mathcal{P}(S^{d-2})$, $\mathcal{H}\mu(x) = \mu\big(\{x\in S^{d-2},\ \langle x, y\rangle\ge 0\}\big)$ and for $f\in L^1(S^{d-2})$, $\langle \mathcal{H}\mu, f\rangle = \langle \mu, \mathcal{H}f\rangle$.

\begin{lemma}[Lemma 2.3 \citep{rubin1999inversion}] \label{lemma:rubin}
    $\mathrm{ker}(\mathcal{H}) = \{\mu\in\mathcal{M}_{\mathrm{even}}(S^{d-2}),\ \mu(S^{d-2})=0\}$ where $\mathcal{M}_{\mathrm{even}}$ is the set of even measures, \emph{i.e.} measures such that for all $f\in C(S^{d-2})$, $\langle \mu, f\rangle = \langle \mu, f^-\rangle$ where $f^-(x)=f(-x)$ for all $x\in S^{d-2}$.
\end{lemma}

\begin{proof}[Proof of \Cref{prop:kernel_radon_ssw}]
    Let $\mu\in\mathcal{M}_{ac}(S^{d-1})$. First, we notice that the density of $\Tilde{R}\mu$ \emph{w.r.t.} $\lambda\otimes \sigma$ is, for all $z\in S^1$, $U\in\mathbb{V}_{d,2}$, 
    \begin{equation}
        (\Tilde{R}\mu)(z,U)= \frac{1}{2\pi}\int_{S^{d-1}} \mathbb{1}_{\{P^U(x)=z\}}\ \mathrm{d}\mu(x) = \frac{1}{2\pi} \int_{F\cap S^{d-1}} \mathbb{1}_{\{\langle x,Uz\rangle>0\}}\ \mathrm{d}\mu(x).    
    \end{equation}
    Indeed, using \Cref{prop:radon_dual_ssw}, and \Cref{prop:integration_set_ssw}, we have for all $g\in C_b(S^1\times \mathbb{V}_{d,2})$,
    \begin{equation}
        \begin{aligned}
            \langle \Tilde{R}\mu, g\rangle_{S^1\times \mathbb{V}_{d,2}} = \langle \mu, \Tilde{R}^*g\rangle_{S^{d-1}} &= \int_{S^{d-1}} \Tilde{R}^*g(x)\ \mathrm{d}\mu(x) \\
            &= \int_{S^{d-1}} \int_{\mathbb{V}_{d,2}} g(P^U(x), U)\ \mathrm{d}\sigma(U)\mathrm{d}\mu(x) \\
            &= \frac{1}{2\pi} \int_{S^{d-1}} \int_{S^1} \int_{\mathbb{V}_{d,2}} g(z,U) \mathbb{1}_{\{ z=P^U(x)\}}\ \mathrm{d}\sigma(U) \mathrm{d}\vol(z) \mathrm{d}\mu(x) \\
            &= \frac{1}{2\pi}\int_{\mathbb{V}_{d,2}\times S^1} g(z, U) \int_{S^{d-1}} \mathbb{1}_{\{z=P^U(x)\}} \mathrm{d} \mu(x)\ \mathrm{d}\vol(z)\mathrm{d}\sigma(U) \\
            &= \frac{1}{2\pi}\int_{\mathbb{V}_{d,2}\times S^1} g(z, U) \int_{F\cap S^{d-1}} \mathbb{1}_{\{\langle x,Uz\rangle>0\}}\mathrm{d}\mu(x) \ \mathrm{d}\vol(z)\mathrm{d}\sigma(U).
        \end{aligned}
    \end{equation}
    
    Hence, using \Cref{prop:link_hemispherical}, we can write $(\Tilde{R}\mu)(z,U) = \frac{1}{2\pi}(\mathcal{H}\Tilde{\mu})(\Tilde{U}z)$ where $\Tilde{\mu} = J^T_\#O_\#\mu$.
    
    Now, let $\mu\in\mathrm{ker}(\Tilde{R})$, then for all $z\in S^1$, $U\in\mathbb{V}_{d,2}$, $\Tilde{R}\mu(z,U) = \mathcal{H}\Tilde{\mu}(\Tilde{U}z)=0$ and hence $\Tilde{\mu}\in\mathrm{ker}(\mathcal{H})=\{\Tilde{\mu}\in\mathcal{M}_{\mathrm{even}}(S^{d-2}),\ \Tilde{\mu}(S^{d-2})=0\}$.
    
    First, let's show that $\mu\in\mathcal{M}_{\mathrm{even}}(S^{d-1})$. Let $f\in C(S^{d-1})$ and $U\in\mathbb{V}_{d,2}$, then, by using the same notation as in Propositions \ref{prop:integration_set_ssw} and \ref{prop:link_hemispherical}, we have
    \begin{equation}
        \begin{aligned}
            \langle \mu, f\rangle_{S^{d-1}} = \int_{S^{d-1}} f(x)\ \mathrm{d}\mu(x) 
            &= \frac{1}{2\pi} \int_{S^{1}} \int_{S^{d-1}} f(x) \mathbb{1}_{\{z=P^U(x)\}}\mathrm{d}\mu(x)\mathrm{d}\vol(z) \\
            &= \frac{1}{2\pi}\int_{S^1} \int_{F\cap S^{d-1}} f(x) \mathbb{1}_{\{\langle x,Uz\rangle>0\}}\mathrm{d}\mu(x)\mathrm{d}\vol(z) \quad \text{ by Prop. \ref{prop:integration_set_ssw}} \\
            &= \frac{1}{2\pi}\int_{S^1} \int_{S^{d-2}} \Tilde{f}(y) \mathbb{1}_{\{\langle y,\Tilde{U}z\rangle > 0\}} \mathrm{d}\Tilde{\mu}(y)\mathrm{d}\vol(z) \\
            &= \frac{1}{2\pi}\int_{S^1} \langle \mathcal{H}\Tilde{\mu}(\Tilde{U}z), \Tilde{f}\rangle_{S^{d-2}}\ \mathrm{d}\vol(z) \\
            &= \frac{1}{2\pi}\int_{S^1} \langle \Tilde{\mu}, \mathcal{H}\Tilde{f}(\Tilde{U}z)\rangle_{S^{d-2}}\ \mathrm{d}\vol(z) \\
            &= \frac{1}{2\pi}\int_{S^1} \langle \Tilde{\mu}, (\mathcal{H}\Tilde{f})^-(\Tilde{U}z) \rangle_{S^{d-2}}\ \mathrm{d}\vol(z) \quad \text{ since $\Tilde{\mu}\in\mathcal{M}_{\mathrm{even}}$} \\
            &= \int_{S^{d-1}} f^-(x)\ \mathrm{d}\mu(x) = \langle \mu, f^-\rangle_{S^{d-1}},
        \end{aligned}
    \end{equation}
    using for the last line all the opposite transformations. Therefore, $\mu\in\mathcal{M}_{\mathrm{even}}(S^{d-1})$.
    
    Now, we need to find on which set the measure is null. We have
    \begin{equation}
        \begin{aligned}
            &\forall z\in S^1, U\in \mathbb{V}_{d,2},\ \Tilde{\mu}(S^{d-2})=0 \\&\iff \forall z\in S^1, U\in\mathbb{V}_{d,2},\ \mu(O^{-1}((J^T)^{-1}(S^{d-2}))) = \mu(F\cap S^{d-1}) = 0.
        \end{aligned}
    \end{equation}
    
    Hence, we deduce that
    \begin{equation}
        \begin{aligned}
            \mathrm{ker}(\Tilde{R}) = \{&\mu\in\mathcal{M}_{\mathrm{even}}(S^{d-1}),\ \forall U \in\mathbb{V}_{d,2}, \forall z\in S^1, F=\mathrm{span}(UU^T)^\bot\cap\mathrm{span}(Uz),\\ & \mu(F\cap S^{d-1})=0\}.
        \end{aligned}
    \end{equation}
    Moreover, we have that $\cup_{U,z} F_{U,z}\cap S^{d-1} = \{H\cap S^{d-1} \subset\mathbb{R}^d,\ \mathrm{dim}(H)=d-1\}$.
    
    Indeed, on the one hand, let H an hyperplane, $x\in H\cap S^{d-1}$, $U\in\mathbb{V}_{d,2}$, and note $z=P^U(x)$. Then, $x\in F\cap S^{d-1}$ by \Cref{prop:integration_set_ssw} and $H\cap S^{d-1}\subset \cup_{U,z}F_{U,z}$.
    
    On the other hand, let $U\in\mathbb{V}_{d,2}$, $z\in S^1$, $F$ is a hyperplane since $\mathrm{dim}(F)=d-1$ and therefore $F\cap S^{d-1}\subset \{H,\ \mathrm{dim}(H)=d-1\}$. 
    
    Finally, we deduce that
    \begin{equation}
        \mathrm{ker}(\Tilde{R}) = \big\{\mu\in \mathcal{M}_{\mathrm{even}}(S^{d-1}),\ \forall H\in\mathcal{G}_{d,d-1},\ \mu(H\cap S^{d-1}) = 0\big\}.
    \end{equation}
\end{proof}

\subsection{Proofs of \Cref{section:properties_ssw}}

\subsubsection{Proof of \Cref{prop:ssw_cv}} \label{proof:prop_ssw_cv}

\begin{proof}[Proof of \Cref{prop:ssw_cv}]
    Since the Wasserstein distance metrizes the weak convergence (Corollary 6.11 \citep{villani2009optimal}), we have $P^U_\#\mu_k\xrightarrow[k\to\infty]{}P^U_\#\mu$ (by continuity) $\iff W_p^p(P^U_\#\mu_k,P^U_\#\mu)\xrightarrow[k\to\infty]{}0$ and hence by the dominated convergence theorem, $\ssw_p^p(\mu_k,\mu)\xrightarrow[k\to\infty]{}0$.
\end{proof}

\subsubsection{Proof of \Cref{prop:samplecomplexity_ssw}} \label{proof:prop_samplecomplexity_ssw}

\begin{proof}[Proof of \Cref{prop:samplecomplexity_ssw}]
    By using the triangle inequality, Fubini-Tonelli, and the hypothesis on the sample complexity of $W_p^p$ on $S^1$, we obtain:
    \begin{equation}
        \begin{aligned}
            \mathbb{E}[|\ssw_p^p(\hat{\mu}_n,\hat{\nu}_n) - \ssw_p^p(\mu,\nu)|] &= \mathbb{E}\left[\left|\int_{\mathbb{V}_{d,2}} \big(W_p^p(P^U_\#\hat{\mu}_n, P^U_\#\hat{\nu}_n) - W_p^p(P^U_\#\mu,P^U_\#\nu)\big)\ \mathrm{d}\sigma(U)\right|\right] \\
            &\le \mathbb{E}\left[\int_{\mathbb{V}_{d,2}} \big|W_p^p(P^U_\#\hat{\mu}_n, P^U_\#\hat{\nu}_n) - W_p^p(P^U_\#\mu,P^U_\#\nu)\big|\ \mathrm{d}\sigma(U)\right] \\
            &= \int_{\mathbb{V}_{d,2}} \mathbb{E}\left[\big|W_p^p(P^U_\#\hat{\mu}_n, P^U_\#\hat{\nu}_n) - W_p^p(P^U_\#\mu,P^U_\#\nu)\big|\right]\ \mathrm{d}\sigma(U) \\
            &\le \int_{\mathbb{V}_{d,2}} \beta(p,n) \ \mathrm{d}\sigma(U) \\
            &= \beta(p,n).
        \end{aligned}
    \end{equation}
\end{proof}

\subsubsection{Proof of \Cref{prop:proj_complexity_ssw}} \label{proof:prop_proj_complexity_ssw}

\begin{proof}[Proof of \Cref{prop:proj_complexity_ssw}]
    Let $(U_i)_{i=1}^L$ be iid samples of $\sigma$. Then, by first using Jensen inequality and then remembering that $\mathbb{E}_U[W_p^p(P^U_\#\mu,P^U_\#\nu)] = \ssw_p^p(\mu,\nu)$, we have
    \begin{equation}
        \begin{aligned}
            \mathbb{E}_U\left[|\widehat{\ssw}_{p,L}^p(\mu,\nu)-\ssw_p^p(\mu,\nu)|\right]^2 &\le \mathbb{E}_U\left[\left|\widehat{\ssw}_{p,L}^p(\mu,\nu)-\ssw_p^p(\mu,\nu)\right|^2\right]\\
            &= \mathbb{E}_U\left[\left|\frac{1}{L} \sum_{i=1}^L \big(W_p^p(P^{U_i}_\#\mu,P^{U_i}_\#\nu) - \ssw_p^p(\mu,\nu)\big)\right|^2\right] \\
            &= \frac{1}{L^2} \mathrm{Var}_U\left(\sum_{i=1}^L W_p^p(P^{U_i}_\#\mu,P^{U_i}_\#\nu)\right) \\
            &= \frac{1}{L} \mathrm{Var}_U\left(W_p^p(P^U_\#\mu,P^U_\#\nu)\right) \\
            &= \frac{1}{L} \int_{\mathbb{V}_{d,2}} \left(W_p^p(P^U_\#\mu,P^U_\#\nu)-\ssw_p^p(\mu,\nu)\right)^2\ \mathrm{d}\sigma(U).
        \end{aligned}
    \end{equation}
\end{proof}

\subsection{Background on the Sphere}

\subsubsection{Uniqueness of the Projection} \label{appendix:uniqueness_projection}

Here, we discuss the uniqueness of the projection $P^U$ for almost every $x$. For that, we recall some results of \citep{bardelli2017probability}.

Let $M$ be a closed subset of a complete finite-dimensional Riemannian manifold $N$. Let $d$ be the Riemannian distance on $N$. Then, the distance from the set $M$ is defined as
\begin{equation}
    d_M(x) = \inf_{y\in M}\ d(x,y).
\end{equation}
The infimum is a minimum since $M$ is closed and $N$ locally compact, but the minimum might not be unique. When it is unique, let's denote the point which attains the minimum as $\pi(x)$, \emph{i.e.} $d(x,\pi(x))=d_M(x)$.
\begin{proposition}[Proposition 4.2 in \citep{bardelli2017probability}] \label{prop:uniqueness_proj}
    Let $M$ be a closed set in a complete $m$-dimensional Riemannian manifold $N$. Then, for almost every $x$, there exists a unique point $\pi(x)\in M$ that realizes the minimum of the distance from $x$.
\end{proposition}
From this Proposition, they further deduce that the measure $\pi_\#\gamma$ is well defined on $M$ with $\gamma$ a locally absolutely continuous measure \emph{w.r.t.} the Lebesgue measure.

In our setting, for all $U\in\mathbb{V}_{d,2}$, we want to project a measure $\mu\in\mathcal{P}(S^{d-1})$ on the great circle $\mathrm{span}(UU^T)\cap S^{-1}$. Hence, we have $N=S^{d-1}$ which is a complete finite-dimensional Riemannian manifold and $M=\mathrm{span}(UU^T)\cap S^{d-1}$ a closed set in $N$. Therefore, we can apply Proposition \ref{prop:uniqueness_proj} and the push-forward measures are well defined for absolutely continuous measures.

\subsubsection{Optimization on the Sphere} \label{appendix:optim_sphere}

Let $F:S^{d-1}\to\mathbb{R}$ be some functional on the sphere. Then, we can perform a gradient descent on a Riemannian manifold by following the geodesics, which are the counterpart of straight lines in $\mathbb{R}^d$. Hence, the gradient descent algorithm \citep{absil2009optimization, bonnabel2013stochastic} reads as
\begin{equation}
    \forall k\ge 0,\ x_{k+1} = \exp_{x_k}\big(-\gamma \mathrm{grad}f(x)\big),
\end{equation}
where for all $x\in S^{d-1}$, $\exp_x:T_xS^{d-1}\to S^{d-1}$ is a map from the tangent space $T_xS^{d-1}=\{v\in \mathbb{R}^d,\ \langle x,v\rangle=0\}$ to $S^{d-1}$ such that for all $v\in T_xS^{d-1}$, $\exp_x(v)=\gamma_v(1)$ with $\gamma_v$ the unique geodesic starting from $x$ with speed $v$, $\emph{i.e.}$ $\gamma(0)=x$ and $\gamma'(0)=v$.

For $S^{d-1}$, the exponential map is known and is
\begin{equation}
    \forall x\in S^{d-1}, \forall v\in T_xS^{d-1},\ \exp_x(v)= \cos(\|v\|_2)x + \sin(\|v\|_2)\frac{v}{\|v\|_2}.
\end{equation}

Moreover, the Riemannian gradient on $S^{d-1}$ is known as \citep[Eq. 3.37]{absil2009optimization}
\begin{equation}
    \mathrm{grad}f(x) = \mathrm{Proj}_x(\nabla f(x)) = \nabla f(x) - \langle \nabla f(x),x\rangle x,
\end{equation}
$\mathrm{Proj}_x$ denoting the orthogonal projection on $T_xS^{d-1}$.

For more details, we refer to \citep{absil2009optimization, boumal2023introduction}.

\subsubsection{Von Mises-Fisher Distribution} \label{appendix:vmf}

The von Mises-Fisher (vMF) distribution is a distribution on $S^{d-1}$ characterized by a concentration parameter $\kappa>0$ and a location parameter $\mu\in S^{d-1}$ through the density
\begin{equation}
    \forall \theta\in S^{d-1},\ f_{\mathrm{vMF}}(\theta;\mu,\kappa) = \frac{\kappa^{d/2-1}}{(2\pi)^{d/2} I_{d/2-1}(\kappa)} \exp(\kappa \mu^T\theta),
\end{equation}
where $I_\nu(\kappa)=\frac{1}{2\pi}\int_0^\pi \exp(\kappa\cos(\theta))\cos(\nu\theta)\mathrm{d}\theta$ is the modified Bessel function of the first kind.

Several algorithms allow to sample from it, see \emph{e.g.} \citep{wood1994simulation, ulrich1984computer} for algorithms using rejection sampling or \citep{kurz2015stochastic} without rejection sampling.

For $d=1$, the vMF coincides with the von Mises (vM) distribution, which has for density
\begin{equation}
    \forall \theta\in [-\pi,\pi[,\ f_{\mathrm{vM}}(\theta;\mu,\kappa) = \frac{1}{I_0(\kappa)} \exp(\kappa \cos(\theta-\mu)),
\end{equation}
with $\mu\in[0,2\pi[$ the mean direction and $\kappa>0$ its concentration parameter. We refer to \citep[Section 3.5 and Chapter 9]{mardia2000directional} for more details on these distributions.

In particular, for $\kappa=0$, the vMF (resp. vM) distribution coincides with the uniform distribution on the sphere (resp. the circle).

\citet[]{jung2021geodesic} studied the law of the projection of a vMF on a great circle. In particular, they showed that, while the vMF plays the role of the normal distributions for directional data, the projection actually does not follow a von Mises distribution. More precisely, they showed the following theorem:

\begin{theorem}[Theorem 3.1 in \citep{jung2021geodesic}]
    Let $d\ge 3$, $X\sim \mathrm{vMF}(\mu,\kappa)\in S^{d-1}$, $U\in\mathbb{V}_{d,2}$ and $T=P^U(X)$ the projection on the great circle generated by $U$. Then, the density function of $T$ is
    \begin{equation}
        \forall t\in [-\pi,\pi[,\ f(t) = \int_0^1 f_R(r) f_{\mathrm{vM}}(t;0,\kappa\cos(\delta)r)\ \mathrm{d}r,
    \end{equation}
    where $\delta$ is the deviation of the great circle (geodesic) from $\mu$ and the mixing density is
    \begin{equation}
        \forall r\in ]0,1[,\ f_R(r) = \frac{2}{I_\nu^*(\kappa)} I_0(\kappa\cos(\delta)r) r (1-r^2)^{\nu-1} I_{\nu-1}^*(\kappa\sin(\delta)\sqrt{1-r^2}),
    \end{equation}
    with $\nu=(d-2)/2$ and $I_\nu^*(z)=(\frac{z}{2})^{-\nu}I_\nu(z)$ for $z>0$, $I_\nu^*(0)=1/\Gamma(\nu+1)$.
\end{theorem}

Hence, as noticed by \citet[]{jung2021geodesic}, in the particular case $\kappa=0$, \emph{i.e.} $X\sim\mathrm{Unif}(S^{d-1})$, then 
\begin{equation}
    f(t) = \int_0^1 f_R(r) f_\mathrm{vM}(t;0,0)\ \mathrm{d}r = f_{\mathrm{vM}(t;0,0)} \int_0^1 f_R(r)\mathrm{d}r = f_{\mathrm{vM}}(t;0,0),
\end{equation}
and hence $T\sim\mathrm{Unif}(S^1)$.

\subsubsection{Normalizing Flows on the Sphere} \label{appendix:nf_sphere}

Normalizing flows \citep{papamakarios2021normalizing} are invertible transformations. There has been a recent interest in defining such transformations on manifolds, and in particular on the sphere \citep{rezende2020normalizing, cohen2021riemannian, rezende2021implicit}.

\paragraph{Exponential map normalizing flows.}

Here, we implemented the Exponential map normalizing flows introduced in \citep{rezende2020normalizing}. The transformation $T$ is 
\begin{equation}
    \forall x\in S^{d-1},\ z=T(x)=\exp_x\big(\mathrm{Proj}_x(\nabla\phi(x))\big),
\end{equation}
where $\phi(x)=\sum_{i=1}^K \frac{\alpha_i}{\beta_i}e^{\beta_i(x^T\mu_i-1)}$, $\alpha_i\ge 0$, $\sum_i \alpha_i\le 1$, $\mu_i\in S^{d-1}$ and $\beta_i>0$ for all $i$. $(\alpha_i)_i$, $(\beta_i)_i$ and $(\mu_i)_i$ are the learnable parameters.

The density of $z$ can be obtained as
\begin{equation}
    p_Z(z) = p_X(x)\det\big(E(x)^TJ_T(x)^TJ_T(x)E(x)\big)^{-\frac12},
\end{equation}
where $J_f$ is the Jacobian in the embedded space and $E(x)$ it the matrix whose columns form an orthonormal basis of $T_xS^{d-1}$.

The common way of training normalizing flows is to use either the reverse or forward KL divergence. Here, we use them with a different loss, namely SSW.

\paragraph{Stereographic projection.}

The stereographic projection $\rho:S^{d-1}\to\mathbb{R}^{d-1}$ maps the sphere $S^{d-1}$ to the Euclidean space. A strategy first introduced in \citep{gemici2016normalizing} is to use it before applying a normalizing flows in the Euclidean space in order to map some prior, and which allows to perform density estimation.

More precisely, the stereographic projection is defined as 
\begin{equation}
    \forall x\in S^{d-1},\ \rho(x) = \frac{x_{2:d}}{1+x_1},
\end{equation}
and its inverse is
\begin{equation}
    \forall u \in \mathbb{R}^{d-1},\ \rho^{-1}(u) = \begin{pmatrix} 2 \frac{u}{\|u\|_2^2 + 1} \\ 1-\frac{2}{\|u\|_2^2 + 1} \end{pmatrix}.
\end{equation}
\citet{gemici2016normalizing} derived the change of variable formula for this transformation, which comes from the theory of probability between manifolds. If we have a transformation $T=f\circ \rho$, where $f$ is a normalizing flows on $\mathbb{R}^{d-1}$, \emph{e.g.} a RealNVP \citep{dinh2017density}, then the log density of the target distribution can be obtained as 
\begin{equation}
    \begin{aligned}
        \log p(x) &= \log p_Z(z) + \log |\det J_f(z)| - \frac12 \log |\det J_{\rho^{-1}}^T J_{\rho^{-1}}(\rho(x))| \\
        &= \log p_Z(z) + \log |\det J_f(z)| - d \log\left(\frac{2}{\|\rho(x)\|_2^2 + 1}\right),
    \end{aligned}
\end{equation}
where we used the formula of \citep{gemici2016normalizing} for the change of variable formula of $\rho$, and where $p_Z$ is the density of some prior on $\mathbb{R}^{d-1}$, typically of a standard Gaussian.
We refer to \citep{gemici2016normalizing, mathieu2020riemannian} for more details about these transformations.

\subsection{Details of the Experiments} \label{appendix:experiments_ssw}

\subsubsection{Gradient Flows on Mixture of von Mises-Fisher Distributions}

For the experiment in \Cref{xp:gradient_flows_ssw}, we use as target distribution a mixture of 6 vMF distributions from which we have access to samples. We refer to \Cref{appendix:vmf} for background on vMF distributions.

The 6 vMF distributions have weights $1/6$, concentration parameter $\kappa=10$ and location parameters $\mu_1=(1,0,0)$, $\mu_2=(0,1,0)$, $\mu_3=(0,0,1)$, $\mu_4=(-1,0,0)$, $\mu_5=(0,-1,0)$ and $\mu_6=(0,0,-1)$.

We approximate the distribution using the empirical distribution, \emph{i.e.} $\hat{\mu} = \frac{1}{n}\sum_{i=1}^n \delta_{x_i}$ and we optimize over the particles $(x_i)_{i=1}^n$. To optimize over particles, we can either use a projected gradient descent:
\begin{equation}
    \begin{cases}
        x^{(k+1)} = x^{(k)}-\gamma \nabla_{x^{(k)}} SSW_2^2(\hat{\mu}_k, \nu) \\
        x^{(k+1)} = \frac{x^{(k+1)}}{\|x^{(k+1)}\|_2},
    \end{cases}
\end{equation}
or a Riemannian gradient descent on the sphere \citep{absil2009optimization} (see \Cref{appendix:optim_sphere} for more details). Note that the projected gradient descent is a Riemannian gradient descent with retraction \citep{boumal2023introduction}.

\subsubsection{Earth data estimation}

Let $T$ be a normalizing flow (NF). For a density estimation task, we have access to a distribution $\mu$ through samples $(x_i)_{i=1}^n$, \emph{i.e.} through the empirical measure $\hat{\mu}_n = \frac{1}{n}\sum_{i=1}^n \delta_{x_i}$ . And the goal is to find an invertible transformation $T$ such that $T_\#\mu = p_Z$, where $p_Z$ is a prior distribution for which we know the density. In that case, indeed, the density of $\mu$, denoted as $f_\mu$ can be obtained as
\begin{equation}
    \forall x,\ f_\mu(x) = p_Z(T(x)) |\det J_T(x)|.
\end{equation}
For the invertible transform, we propose to use normalizing flows on the sphere (see Appendix \ref{appendix:nf_sphere}). We use two different normalizing flows, exponential map normalizing flows \citep{rezende2020normalizing} and Real NVP \citep{dinh2017density} + stereographic projection \citep{gemici2016normalizing} which we call ``Stereo'' in Table \ref{tab:nll}.

To fit $T_\#\mu = p_Z$, we use either SSW, SW on the sphere, or SW on $\mathbb{R}^{d-1}$ for the stereographic projection based NF. For the exponential map normalizing flow, we compose 48 blocks, each one with 100 components. These transformations have 24000 parameters. For Real NVP, we compose 10 blocks of Real NVPs, with shifting and scaling as multilayer perceptrons, composed of 10 layers, 25 hidden units and with Leaky ReLU of parameters 0.2 for the activation function. The number of parameters of these networks is 27520.

For the training process, we perform 20000 epochs with full batch size. We use Adam as an optimizer with a learning rate of $10^{-1}$. For the stereographic NF, we use a learning rate of $10^{-3 }$.

We report in Table \ref{tab:earh_datasets} details of the datasets.

\begin{table}[H]
    \centering
    \caption{Details of Earth datasets.}
    \small
    \begin{tabular}{cccc}
        & Earthquake & Flood & Fire \\ \toprule
        Train set size & 4284 & 3412 & 8966 \\
        Test set size & 1836 & 1463 & 3843 \\
        Data size & 6120 & 4875 & 12809 \\
        \bottomrule
    \end{tabular}
    \label{tab:earh_datasets}
\end{table}

\subsubsection{Sliced-Wasserstein Autoencoder} \label{appendix:swae}

We recall that in the WAE framework, we want to minimize
\begin{equation}
    \mathcal{L}(f,g) = \int c\big(x,g(f(x))\big)\mathrm{d}\mu(x) + \lambda D(f_\#\mu,p_Z),
\end{equation}
where $f$ is an encoder, $g$ a decoder, $p_Z$ a prior distribution, $c$ some cost function and $D$ is a divergence in the latent space. Several $D$ were proposed. For example, \citet[]{tolstikhin2018wasserstein} proposed to use the MMD, \citet[]{kolouri2018sliced} used the SW distance, \citet[]{patrini2020sinkhorn} used the Sinkhorn divergence, \citet[]{kolouri2019generalized} used the generalized SW distance. Here, we use $D=SSW_2^2$.

\paragraph{Architecture and procedure.} We first detail the hyperparameters and architectures of neural networks for MNIST and Fashion MNIST. For the encoder $f$ and the decoder $g$, we use the same architecture as \citet{kolouri2018sliced}.

For both the encoder and the decoder architecture, we use fully convolutional architectures with 3x3 convolutional filters. More precisely, the architecture of the encoder is
\begin{equation*}
    \begin{aligned}
        x\in \mathbb{R}^{28\times 28} &\to \mathrm{Conv2d}_{16} \to \mathrm{LeakyReLU}_{0.2} \\
        &\to \mathrm{Conv2d}_{16} \to \mathrm{LeakyReLU}_{0.2} \to \mathrm{AvgPool}_2 \\
        &\to \mathrm{Conv2d}_{32} \to \mathrm{LeakyReLU}_{0.2} \\
        &\to \mathrm{Conv2d}_{32} \to \mathrm{LeakyReLU}_{0.2} \to \mathrm{AvgPool}_2 \\
        &\to \mathrm{Conv2d}_{64} \to \mathrm{LeakyReLU}_{0.2} \\
        &\to \mathrm{Conv2d}_{64} \to \mathrm{LeakyReLU}_{0.2} \to \mathrm{AvgPool}_2 \\
        &\to \mathrm{Flatten} \to \mathrm{FC}_{128} \to \mathrm{ReLU} \\
        &\to \mathrm{FC}_{d_Z} \to \ell^2 \ \text{normalization}
    \end{aligned}
\end{equation*}
where $d_Z$ is the dimension of the latent space (either $11$ for $S^{10}$ or $3$ for $S^2$). 

The architecture of the decoder is
\begin{equation*}
    \begin{aligned}
        z\in\mathbb{R}^{d_Z} &\to \mathrm{FC}_{128} \to \mathrm{FC}_{1024} \to \mathrm{ReLU} \\
        &\to \mathrm{Reshape(64 x 4 x 4)} \to \mathrm{Upsample}_2 \to \mathrm{Conv}_{64} \to \mathrm{LeakyReLU}_{0.2} \\
        &\to \mathrm{Conv}_{64} \to \mathrm{LeakyReLU}_{0.2} \\
        &\to \mathrm{Upsample}_2 \to \mathrm{Conv}_{64} \to \mathrm{LeakyReLU}_{0.2} \\
        &\to \mathrm{Conv}_{32} \to \mathrm{LeakyReLU}_{0.2} \\
        &\to \mathrm{Upsample}_2 \to \mathrm{Conv}_{32} \to \mathrm{LeakyReLU}_{0.2} \\
        &\to \mathrm{Conv}_1 \to \mathrm{Sigmoid}
    \end{aligned}
\end{equation*}

To compare the different autoencoders, we used as the reconstruction loss the binary cross entropy, $\lambda=10$, Adam \citep[]{kingma2014adam} as optimizer with a learning rate of $10^{-3}$ and Pytorch's default momentum parameters for 800 epochs with batch of size $n=500$. Moreover, when using SW type of distance, we approximated it with $L=1000$ projections. 

For the experiment on CIFAR10, we use the same architecture as \citet{tolstikhin2018wasserstein}. More precisely, the architecture of the encoder is
\begin{equation*}
    \begin{aligned}
        x\in\mathbb{R}^{3\times 32 \times 32} & \to \mathrm{Conv2d}_{128} \to \mathrm{BatchNorm} \to \mathrm{ReLU} \\
        &\to \mathrm{Conv2d}_{256} \to \mathrm{BatchNorm} \to \mathrm{ReLU} \\
        &\to \mathrm{Conv2d}_{512} \to \mathrm{BatchNorm} \to \mathrm{ReLU} \\
        &\to \mathrm{Conv2d}_{1024} \to \mathrm{BatchNorm} \to \mathrm{ReLU} \\
        &\to \mathrm{FC}_{d_z} \to \ell^2\ \text{normalization}
    \end{aligned}
\end{equation*}
where $d_z = 65$.

The architecture of the decoder is 
\begin{equation*}
    \begin{aligned}
        z\in\mathbb{R}^{d_z} &\to \mathrm{FC}_{4096} \to \mathrm{Reshape}(1024\times 2 \times 2) \\
        &\to  \mathrm{Conv2dT}_{512} \to \mathrm{BatchNorm} \to \mathrm{ReLU} \\
        &\to \mathrm{Conv2dT}_{256} \to \mathrm{BatchNorm} \to \mathrm{ReLU} \\
        &\to \mathrm{Conv2dT}_{128} \to \mathrm{BatchNorm} \to \mathrm{ReLU} \\
        &\to \mathrm{Conv2dT}_{3} \to \mathrm{Sigmoid}
    \end{aligned}
\end{equation*}

We use here a batch size of $n=128$, $\lambda=0.1$, the binary cross entropy as reconstruction loss and Adam as optimizer with a learning rate of $10^{-3}$.

We report in Table \ref{tab:fid} the FID obtained using 10000 samples and we report the mean over 5 trainings.

For SSW, we used the formulation using the uniform distribution \eqref{eq:w2_unif_circle}. To compute SW, we used the POT library \citep[]{flamary2021pot}. To compute the Sinkhorn divergence, we used the GeomLoss package \citep{feydy2019interpolating}.
\newpage

\section{Appendix of \Cref{chapter:swgf}}

\subsection{Proofs} \label{appendix:proofs_swgf}

First, we recall some propositions about the continuity and convexity of the Sliced-Wasserstein distance as well as on the existence of the minimizer at each step of the SW-JKO scheme. These results were derived in \citep{candau_tilh}. In the following, we restrain ourselves to measures supported on a compact domain $K$. 

\begin{proposition}[Proposition 3.4 in \citep{candau_tilh}] \label{prop:candautilh_34}
    Let $\nu\in\mathcal{P}_2(K)$. Then, $\mu\mapsto \sw_2^2(\mu,\nu)$ is continuous \emph{w.r.t.} the weak convergence.
\end{proposition}

\begin{proposition}[Proposition 3.5 in \citep{candau_tilh}] \label{prop:candautilh_35}
    Let $\nu\in\mathcal{P}_2(K)$, then $\mu\mapsto \sw_2^2(\mu,\nu)$ is convex and strictly convex whenever $\nu$ is absolutely continuous \emph{w.r.t.} the Lebesgue measure.
\end{proposition}

\begin{proposition}[Proposition 3.7 in \citep{candau_tilh}] \label{prop:candautilh_37}
    Let $\tau>0$ and $\mu_k^\tau\in\mathcal{P}_2(K)$. Then, there exists a unique solution $\mu_{k+1}^\tau\in\mathcal{P}_2(K)$ to the minimization problem
    \begin{equation}
        \min_{\mu\in\mathcal{P}_2(K)}\ \frac{\sw_2^2(\mu,\mu_k^\tau)}{2\tau} + \int V\mathrm{d}\mu + \mathcal{H}(\mu).
    \end{equation}
    The solution is even absolutely continuous.
\end{proposition}

\subsubsection{Proof of \Cref{prop:minimizer}} \label{proof:prop_minimizer}

\begin{proof}[Proof of \Cref{prop:minimizer}]
    
    Let $\tau>0$,\ $k\in\mathbb{N}$,\ $\mu_k^\tau\in\mathcal{P}_2(K)$. Let's note $J(\mu)=\frac{\sw_2^2(\mu,\mu_k^\tau)}{2\tau}+\mathcal{F}(\mu)$. 
    
    According to \Cref{prop:candautilh_34}, $\mu\mapsto \sw_2^2(\mu,\mu_k^\tau)$ is continuous with respect to the weak convergence. Indeed, let $\mu\in\mathcal{P}_2(K)$ and let $(\mu_n)_n$ converging weakly to $\mu$, \emph{i.e.} $\mu_n\xrightarrow[n\to\infty]{\mathcal{L}} \mu$. Then, by the reverse triangular inequality, we have
    \begin{equation}
        |\sw_2(\mu_n,\mu_k^\tau)-\sw_2(\mu,\mu_k^\tau)|\le \sw_2(\mu_n,\mu) \le W_2(\mu_n,\mu).
    \end{equation}
    Since the Wasserstein distance metrizes the weak convergence \citep{villani2009optimal}, we have that $W_2(\mu_n,\mu)\to 0$. And therefore, $\mu\mapsto \sw_2(\mu,\mu_k^\tau)$ is continuous \emph{w.r.t.} the weak convergence.
    
    By hypothesis, $\mathcal{F}$ is lower semi continuous, hence $\mu\mapsto J(\mu)$ is lower semi continuous. Moreover, $\mathcal{P}_2(K)$ is compact for the weak convergence, thus we can apply the Weierstrass theorem (Box 1.1 in \citep{santambrogio2015optimal}) and there exists a minimizer $\mu_{k+1}^\tau$ of $J$.
    
    By \Cref{prop:candautilh_35}, $\mu\mapsto \sw_2^2(\mu,\nu)$ is convex and strictly convex whenever $\nu$ is absolutely continuous \emph{w.r.t.} the Lebesgue measure. Hence, for the uniqueness, if $\mathcal{F}$ is strictly convex then $\mu\mapsto J(\mu)$ is also strictly convex and the minimizer is unique. And if $\rho_k^\tau$ is absolutely continuous, then according to \Cref{prop:candautilh_35}, $\mu\mapsto \sw_2^2(\mu,\mu_k^\tau)$ is strictly convex, and hence $\mu\mapsto J(\mu)$ is also strictly convex since $\mathcal{F}$ was taken convex by hypothesis.
\end{proof}

\subsubsection{Proof of Proposition \ref{prop:nonincreasing}} \label{proof:prop_nonincreasing}

\begin{proof}[Proof of \Cref{prop:nonincreasing}]
    Let $k\in\mathbb{N}$, then since $\mu_{k+1}^\tau$ is the minimizer of \eqref{eq:swjko},
    \begin{equation}
        \mathcal{F}(\mu_{k+1}^\tau)+\frac{\sw_2^2(\mu_{k+1}^\tau,\mu_k^\tau)}{2\tau} \le \mathcal{F}(\mu_{k}^\tau)+\frac{\sw_2^2(\mu_{k}^\tau,\mu_k^\tau)}{2\tau} = \mathcal{F}(\mu_k^\tau).
    \end{equation}
    Hence, as $\sw_2^2(\mu_{k+1}^\tau,\mu_k^\tau)\ge 0$, 
    \begin{equation}
        \mathcal{F}(\mu_{k+1}^\tau)\le\mathcal{F}(\mu_k^\tau).
    \end{equation}
\end{proof}

\subsection{Relations between Sliced-Wasserstein and Wasserstein} \label{appendix:sw}

\paragraph{Link for 1D supported measures.}

Let $\mu,\nu\in\mathcal{P}(\mathbb{R}^d)$ supported on a line. For simplicity, we suppose that the measures are supported on an axis, \emph{i.e.} $\mu(x)=\mu_1(x_1)\prod_{i=2}^d \delta_0(x_i)$ and $\nu(x) = \nu_1(x_1)\prod_{i=2}^d \delta_0(x_i)$.

In this case, we have that
\begin{equation}
    W_2^2(\mu,\nu) = W_2^2(P^{e_1}_\#\mu,P^{e_1}_\#\nu) = \int_0^1 |F_{P^{e_1}_\#\mu}^{-1}(x) - F_{P^{e_1}_\#\nu}^{-1}(x)|^2\ \mathrm{d}x.
\end{equation}

On the other hand, let $\theta\in S^{d-1}$, then we have
\begin{equation}
    \begin{aligned}
        \forall y\in\mathbb{R},\ F_{P^\theta_\#\mu}(y) &= \int_{\mathbb{R}} \mathbb{1}_{]-\infty,y]}(x)\ P^\theta_\#\mu(\mathrm{d}x) \\
        &= \int_{\mathbb{R}^d} \mathbb{1}_{]-\infty, y]}(\langle \theta,x\rangle)\ \mu(\mathrm{d}x) \\
        &= \int_{\mathbb{R}} \mathbb{1}_{]-\infty, y]}(x_1\theta_1)\ \mu_1(\mathrm{d}x_1) \\
        &= \int_{\mathbb{R}} \mathbb{1}_{]-\infty, \frac{y}{\theta_1}]}(x_1)\ \mu_1(\mathrm{d}x_1) \\
        &= F_{P^{e_1}_\#\mu}\left(\frac{y}{\theta_1}\right).
    \end{aligned}
\end{equation}
Therefore, $F_{P^\theta_\#\mu}^{-1}(z) = \theta_1 F_{P^{e_1}_\#\mu}^{-1}(z)$ and 
\begin{equation}
    \begin{aligned}
        W_2^2(P^\theta_\#\mu,P^\theta_\#\nu) &= \int_0^1 |\theta_1 F_{P^{e_1}_\#\mu}^{-1}(z) - \theta_1 F_{P^{e_1}_\#\nu}^{-1}(z)|^2 \ \mathrm{d}z \\
        &= \theta_1^2 \int_0^1 |F_{P^{e_1}_\#\mu}^{-1}(z) - F_{P^{e_1}_\#\nu}^{-1}(z)|^2 \ \mathrm{d}z \\
        &= \theta_1^2 W_2^2(\mu,\nu).
    \end{aligned}
\end{equation}
Finally, using that $\int_{S^{d-1}}\theta\theta^T\ \mathrm{d}\lambda(\theta) = \frac{1}{d} I_d$, we can conclude that
\begin{equation}
    SW_2^2(\mu,\nu) = \int_{S^{d-1}} \theta_1^2 W_2^2(\mu,\nu)\ \mathrm{d}\lambda(\theta) = \frac{W_2^2(\mu,\nu)}{d}.
\end{equation}

\paragraph{Closed-form between Gaussians.} \label{sw_gaussians}

It is well known that there is a closed-form for the Wasserstein distance between Gaussians \citep{givens1984class}. If we take $\alpha=\mathcal{N}(\mu,\Sigma)$ and $\beta=\mathcal{N}(m,\Lambda)$ with $m,\mu\in\mathbb{R}^d$ and $\Sigma,\Lambda\in\mathbb{R}^{d\times d}$ two symmetric positive definite matrices, then 
\begin{equation}
    W_2^2(\alpha,\beta) = \|m-\mu\|_2^2 + \mathrm{Tr}\big(\Sigma+\Lambda - 2(\Sigma^{\frac12}\Lambda\Sigma^{\frac12})^{\frac12}\big).
\end{equation}

Let $\alpha=\mathcal{N}(\mu,\sigma^2 I_d)$ and $\beta=\mathcal{N}(m,s^2 I_d)$ two isotropic Gaussians. Here, we have
\begin{equation}
    \begin{aligned}
        W_2^2(\alpha,\beta) &= \|\mu-m\|^2_2 + \mathrm{Tr}(\sigma^2 I_d + s^2 I_d - 2(\sigma s^2\sigma I_d)^\frac12) \\
        &= \|\mu-m\|^2_2 + (\sigma-s)^2\ \mathrm{Tr}(I_d) \\
        &= \|\mu-m\|^2_2 + d(\sigma-s)^2.
    \end{aligned}
\end{equation}
On the other hand, \citet{nadjahi2021fast} showed (Equation 73) that
\begin{equation}
    SW_2^2(\alpha,\beta) = \frac{\|\mu-m\|_2^2}{d} + (\sigma-s)^2 = \frac{W_2^2(\alpha,\beta)}{d}.
\end{equation}

In that case, the dilation of factor $d$ between WGF and SWGF clearly appears.

\subsection{Algorithms to solve the SW-JKO scheme} \label{algorithms_swjko}

We provide here the algorithms used to solve the SW-JKO scheme \eqref{eq:swjko} for the discrete grid and for the particles (\Cref{section:swjko_practice}).

\paragraph{Discrete grid.}

\begin{algorithm}[t]
   \caption{SW-JKO with Discrete Grid}
   \label{alg:swjko_discrete}
    \begin{algorithmic}
       \STATE {\bfseries Input:} $\mu_0$ the initial distribution with density $\rho_0$, $K$ the number of SW-JKO steps, $\tau$ the step size, $\mathcal{F}$ the functional, $N_e$ the number of epochs to solve each SW-JKO step, $(x_j)_{j=1}^N$ the grid
       \STATE Let $\rho^{(0)}=\left(\frac{\rho_0(x_j)}{\sum_{\ell=1}^N\rho_0(x_\ell)}\right)_{j=1}^N$
       \FOR{$k=1$ {\bfseries to} $K$}
       \STATE Initialize the weights $\rho^{(k+1)}$ (with for example a copy of $\rho^{(k)}$)
       \STATE // Denote $\mu_{k+1}^\tau = \sum_{j=1}^N \rho_j^{(k+1)}\delta_{x_j}$ and $\mu_k^\tau = \sum_{j=1}^N \rho_j^{(k)} \delta_{x_j}$
       \FOR{$i=1$ {\bfseries to} $N_e$}
       \STATE Compute $J(\mu_{k+1}^\tau) = \frac{1}{2\tau} SW_2^2(\mu_k^\tau, \mu_{k+1}^\tau)+\mathcal{F}(\mu_{k+1}^\tau)$
       \STATE Backpropagate through $J$ with respect to $\rho^{(k+1)}$
       \STATE Perform a gradient step
       \STATE Project on the simplex $\rho^{(k+1)}$ using the algorithm of \citet{condat2016fast}
       \ENDFOR
       \ENDFOR
    \end{algorithmic}
\end{algorithm}

We recall that in that case, we model the distributions as $\mu_k^\tau = \sum_{i=1}^N \rho_i^{(k)}\delta_{x_i}$ where we use $N$ samples located at $(x_i)_{i=1}^N$ and $(\rho_i^{(k)})_{i=1}^N$ belongs to the simplex $\Sigma_n$. Hence, the SW-JKO scheme at step $k+1$ rewrites 
\begin{equation}
    \min_{(\rho_i)_i\in\Sigma_N}\ \frac{SW_2^2(\sum_{i=1}^N\rho_i\delta_{x_i},\mu_k^\tau)}{2\tau}+\mathcal{F}(\sum_{i=1}^N\rho_i\delta_{x_i}).
\end{equation}

We report in Algorithm \ref{alg:swjko_discrete} the whole procedure.

\paragraph{Particle scheme.}

In this case, we model the distributions as empirical distributions and we try to optimize the positions of the particles. Hence, we have $\mu_k^\tau=\frac{1}{N}\sum_{i=1}^N \delta_{x_i^{(k)}}$ and the problem \eqref{eq:swjko} becomes
\begin{equation}
    \min_{(x_i)_i}\ \frac{SW_2^2(\frac{1}{N}\sum_{i=1}^N \delta_{x_i}, \mu_k^\tau)}{2\tau}+\mathcal{F}(\frac{1}{N}\sum_{i=1}^N \delta_{x_i}).
\end{equation}

In this case, we provide the procedure in Algorithm \ref{alg:swjko_particles}.

\begin{algorithm}[t]
   \caption{SW-JKO with Particles}
   \label{alg:swjko_particles}
    \begin{algorithmic}
       \STATE {\bfseries Input:} $\mu_0$ the initial distribution, $K$ the number of SW-JKO steps, $\tau$ the step size, $\mathcal{F}$ the functional, $N_e$ the number of epochs to solve each SW-JKO step, $N$ the batch size
       \STATE Sample $(x^{(0)}_j)_{j=1}^N\sim \mu_0$ i.i.d
       \FOR{$k=1$ {\bfseries to} $K$}
       \STATE Initialize $N$ particles $(x^{(k+1)}_j)_{j=1}^N$ (with for example a copy of $(x^{(k)}_j)_{j=1}^N$)
       \STATE // Denote $\mu_{k+1}^\tau = \frac{1}{N}\sum_{j=1}^N \delta_{x_j^{(k+1)}}$ and $\mu_k^\tau = \frac{1}{N}\sum_{j=1}^N \delta_{x_j^{(k)}}$
       \FOR{$i=1$ {\bfseries to} $N_e$}
       \STATE Compute $J(\mu_{k+1}^\tau) = \frac{1}{2\tau} SW_2^2(\mu_k^\tau, \mu_{k+1}^\tau)+\mathcal{F}(\mu_{k+1}^\tau)$
       \STATE Backpropagate through $J$ with respect to $(x_j^{(k+1)})_{j=1}^N$
       \STATE Perform a gradient step
       \ENDFOR
       \ENDFOR
    \end{algorithmic}
\end{algorithm}

\subsection{Experimental Details}

\subsubsection{Convergence to stationary distribution} \label{appendix_cv_stationary}

Here, we want to demonstrate that, through the SW-JKO scheme, we are able to find good minima of functionals using simple generative models.

\begin{figure}[t]
    \centering
    \includegraphics[width=\linewidth]{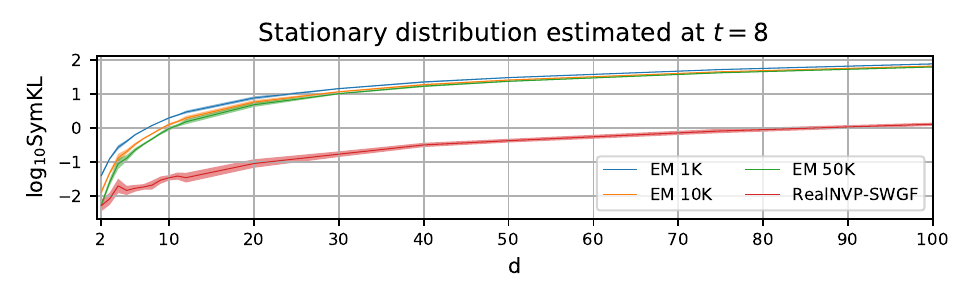}
    \caption{Symmetric KL divergence between the learned distribution at time $t=8$ and the true stationary solution on Gaussians}
    \label{fig:gaussian_stationary1}
\end{figure}

In this experiment, we generate 15 Gaussians for $d$ between 2 and 12, and we quantify how well the SW-JKO scheme, using a Real NVP to approximate distributions, is able to learn the target Gaussians. We start from $\mu_0=\mathcal{N}(0,I)$ and use a step size of $\tau=0.1$ for 80 iterations in order to match the stationary distribution. In this case, the functional is
\begin{equation}
    \mathcal{F}(\mu)=\int V(x)\mathrm{d}\mu(x)+\mathcal{H}(\mu)
\end{equation}
with $V(x)=-\frac12 (x-b)^T A (x-b)$, and the stationary distribution is $\rho^*(x) \propto e^{-V(x)}$, hence $\rho^* = \mathcal{N}(b,A^{-1})$. This functional is approximated using \eqref{eq:approx_fokker_planck_nf}. In \Cref{fig:unstabilities_jkoicnn}, we showed the results for $d\in\{2,\dots,12\}$ and the unstability of JKO-ICNN. We add the results for $d\in\{20,30,40,50,75,100\}$ in \Cref{fig:gaussian_stationary1}.

\paragraph{Symmetric Kullback-Leibler divergence.} To quantify the closeness between the learned distributions and the targets, we compute the symmetric Kullback-Leibler divergence between the ground truth of WGF $\mu^*$ and the distribution $\hat{\mu}$ approximated by the different schemes. The symmetric Kullback-Leibler divergence is obtained as
\begin{equation}
    \mathrm{SymKL}(\mu^*,\hat{\mu})=\mathrm{KL}(\mu^*||\hat{\mu}) + \mathrm{KL}(\hat{\mu}||\mu^*).
\end{equation}
To approximate it, we generate $10^4$ samples of each distribution and evaluate the density at those samples.

\paragraph{Normalizing flows.} If we note $g_\theta$ a normalizing flows, $p_Z$ the distribution in the latent space and $\rho=(g_\theta)_\# p_Z$, then we can evaluate the log density of $\rho$ by using the change of variable formula. Let $x=g_\theta(z)$, then
\begin{equation}
    \log(\rho(x))=\log(p_Z(z))-\log|\det J_{g_\theta}(z)|.
\end{equation}
We choose RealNVPs \citep{dinh2017density} for the simplicity of the transformations and the fact that we can compute efficiently the determinant of the Jacobian (since we have a closed-form). A RealNVP flow is a composition of transformations $T$ of the form
\begin{equation}
    \forall z\in\mathbb{R}^d,\ x = T(z) = \big(z^1, \exp(s(z^1))\odot z^2 + t(z^1)\big)
\end{equation}
where we write $z=(z^1,z^2)$ and with $s$ and $t$ some neural networks. To modify all the components, we use also swap transformations (\emph{i.e.} $(z^1,z^2)\mapsto (z^2,z^1)$). This transformation is invertible with $\log \det J_T(z) = \sum_i s(z^1_i)$.

In our experiments, we use RealNVPs with 5 affine coupling layers, using fully connected neural networks for the scaling and shifting networks with 100 hidden units and 5 layers.

\paragraph{Optimization hyperparameters.} We use 200 epochs of each inner optimization and an Adam optimizer \citep{kingma2014adam} with a learning rate of $5\cdot 10^{-3}$ for the first iteration and $10^{-3}$ for the rest. We also use a batch size of 1000 samples. To approximate $\sw$, we always use $L=1000$ projections.

\paragraph{Euler-Maruyama.} For Euler-Maruyama, as in \citep{mokrov2021largescale}, we use kernel density estimation in order to approximate the density. We use the \texttt{Scipy} implementation \citep{2020SciPy-NMeth} ``gaussian\_kde'' with the Scott's rule to choose the bandwidth. We run the different chains with a step size of $10^{-3}$.

\subsubsection{Bayesian logistic regression} \label{Appendix_BLR}

For the Bayesian logistic regression, we have access to covariates $s_1,\dots,s_n\in\mathbb{R}^d$ with their associated labels $y_1,\dots,y_n\in\{-1,1\}$. Following \citep{liu2016stein, mokrov2021largescale}, we put as prior on the regression weights $w$, $p_0(w|\alpha)=\mathcal{N}(w;0,\frac{1}{\alpha})$ with $p_0(\alpha)=\Gamma(\alpha;1,0.01)$. Therefore, we aim at learning the posterior $p(w,\alpha|y)$:
\begin{equation}
    p(w,\alpha|y) \propto p(y|w,\alpha) p_0(w|\alpha) p_0(\alpha) = p_0(\alpha) p_0(w|\alpha) \prod_{i=1}^n p(y_i|w,\alpha)
\end{equation}
where $p(y_i|w,\alpha) = \sigma(w^T s_i)^{\frac{1+y_i}{2}}(1-\sigma(w^T s_i))^{\frac{1-y}{2}}$ with $\sigma$ the sigmoid. To evaluate $\mathcal{V}(\mu)=\int V(x)\ \mathrm{d}\mu(x)$, we resample data uniformly.

In our context, let $V(x) = -\log\big(p_0(\alpha)p_0(w|\alpha)p(y|w,\alpha)\big)$, then using $\mathcal{F}(\mu) = \int V \mathrm{d}\mu + \mathcal{H}(\mu)$ as functional, we know that the limit of the stationary solution of Fokker-Planck is proportional to $e^{-V} = p(w,\alpha|y)$.

Following \cite{liu2016stein, mokrov2021largescale}, we use the 8 datasets of \cite{mika1999fisher} and the covertype dataset (\url{https://www.csie.ntu.edu.tw/~cjlin/libsvmtools/datasets/binary.html}).

We report in Table \ref{tab:datasets} the characteristics of the different datasets. The datasets are loaded using the code of \cite{mokrov2021largescale} (\url{https://github.com/PetrMokrov/Large-Scale-Wasserstein-Gradient-Flows}). We split the dataset between train set and test set with a 4:1 ratio.

\begin{table}[t]
    \centering
    \caption{Number of features, of samples and batch size of each dataset.}
    \begin{tabular}{cccccccccc}
         & covtype & german & diabetis & twonorm & ringnorm & banana & splice & waveform & image  \\ \toprule
        features & 54 & 20 & 8 & 20 & 20 & 2 & 60 & 21 & 18 \\
        samples & 581012 & 1000 & 768 & 7400 & 7400 & 5300 & 2991 & 5000 & 2086 \\
        batch size & 512 & 800 & 614 & 1024 & 1024 & 1024 & 512 & 512 & 1024 \\
        \bottomrule
    \end{tabular}
    \label{tab:datasets}
\end{table}

We report in Table \ref{tab:hyperparams_mine} the hyperparameters used for the results reported in Table \ref{tab:blr}. We also tuned the time step $\tau$ since for too big $\tau$, we observed bad results, as the SW-JKO scheme should be a good approximation of the SWGF only for small enough $\tau$. 

Moreover, we reported in Table \ref{tab:blr} the mean over 5 training. For the results obtained with JKO-ICNN, we used the same hyperparameters as \citet{mokrov2021largescale}.

\begin{table}[t]
    \centering
    \caption{Hyperparameters for SWGFs with RealNVPs. nl: number of coupling layers in RealNVP, nh: number of hidden units of conditioner neural networks, lr: learning rate using Adam, JKO steps: number of SW-JKO steps, Iters by step: number of epochs for each SW-JKO step, $\tau$: the time step, batch size: number of samples taken to approximate the functional.}
    \begin{tabular}{cccccccccc}
         & covtype & german & diabetis & twonorm & ringnorm & banana & splice & waveform & image  \\ \toprule
        nl & 2 & 2 & 2 & 2 & 2 & 2 & 5 & 5 & 2 \\
        nh & 512 & 512 & 512 & 512 & 512 & 512 & 128 & 128 & 512 \\
        lr & $2e^{-5}$ & $1e^{-4}$ & $5e^{-4}$ & $1e^{-4}$ & $5e^{-5}$ & $1e^{-4}$ & $5e^{-4}$ & $1e^{-4}$ & $5e^{-5}$ \\
        JKO steps & 5 & 5 & 10 & 20 & 5 & 5 & 5 & 5 & 5 \\
        Iters by step & 1000 & 500 & 500 & 500 & 1000 & 500 & 500 & 500 & 500 \\
        $\tau$  & 0.1 & $10^{-6}$ & $5\cdot 10^{-6}$ & $10^{-8}$ & $10^{-6}$ & 0.1 & $10^{-6}$ & $10^{-8}$ & 0.1 \\
        batch size & 1024 & 1024 & 1024 & 1024 & 1024 & 1024 & 1024 & 512 & 1024 \\
        \bottomrule
    \end{tabular}
    \label{tab:hyperparams_mine}
\end{table}

\subsubsection{Influence of the number of projections} \label{appendix:impact_projs}

It is well known that the approximation of Sliced-Wasserstein is subject to the curse of dimensionality through the Monte-Carlo approximation \citep{nadjahi2020statistical}. We provide here some experiments to quantify this influence. However, first note that the goal is not to minimize the Sliced-Wasserstein distance, but rather the functional, SW playing mostly a regularizer role. Experiments on the influence of the number of experiments to approximate the SW have already been conducted (see \emph{e.g.} Figure 2 in \citep{nadjahi2020statistical} or Figure 1 in \citep{deshpande2019max}).

Here, we take the same setting of Section \ref{section:xp_stationary_FKP}, \emph{i.e.} we generate 15 random Gaussians, and then vary the number of projections and report the Symmetric Kullback-Leibler divergence on Figure \ref{fig:impact_projs}. We observe that the results seem to improve with the number of projections until they reach a certain plateau. The plateau seems to be attained for a bigger number of dimensions in high dimensions.
\newpage

\section{Appendix of \Cref{chapter:usw}}

\subsection{Document Classification} \label{appendix:doc_classif}



\subsubsection{Datasets}

\begin{table}[H]
    \centering
    \caption{Dataset characteristics.}
    \begin{tabular}{ccccc}
         & \textbf{BBCSport} & \textbf{Movies} & \textbf{Goodreads genre} & \textbf{Goodreads like} \\ \toprule
        Doc & 737 & 2000 & 1003 & 1003 \\
        Train & 517 & 1500 & 752 & 752 \\
        Test & 220 & 500 & 251 & 251 \\
        Classes & 5 & 2 & 8 & 2 \\
        Mean words by doc & $116\pm 54$ & $182 \pm 65$ & $1491 \pm 538$ & $1491 \pm 538$\\
        Median words by doc & 104 & 175 & 1518 & 1518 \\
        Max words by doc & 469 & 577 & 3499 & 3499 \\
        \bottomrule
    \end{tabular}
    \label{tab:summary_docs}
\end{table}

We sum up the statistics of the different datasets in \Cref{tab:summary_docs}.

\paragraph{BBCSport.} The BBCSport dataset contains articles between 2004 and 2005, and is composed of 5 classes. We average over the 5 same train/test split of \cite{kusner2015word}. The dataset can be found in \url{https://github.com/mkusner/wmd/tree/master}.

\paragraph{Movie Reviews.} The movie reviews dataset is composed of 1000 positive and 1000 negative reviews. We take five different random 75/25 train/test split. The data can be found in \url{http://www.cs.cornell.edu/people/pabo/movie-review-data/}.

\paragraph{Goodreads.} This dataset, proposed in \citep{maharjan2017multi}, and which can be found at \url{https://ritual.uh.edu/multi_task_book_success_2017/}, is composed of 1003 books from 8 genres. A first possible classification task is to predict the genre. A second task is to predict the likability, which is a binary task where a book is said to have success if it has an average rating $\geq 3.5$ on the website Goodreads (\url{https://www.goodreads.com}). The five train/test split are randomly drawn with 75/25 proportions.

\subsubsection{Technical Details} \label{xp_docs:technicals}

All documents are embedded with the Word2Vec model \citep{mikolov2013distributed} in dimension $d=300$. The embedding can be found in \url{https://drive.google.com/file/d/0B7XkCwpI5KDYNlNUTTlSS21pQmM/view?resourcekey=0-wjGZdNAUop6WykTtMip30g}.

In this experiment, we report the results averaged over 5 random train/test split. For discrepancies which are approximated using random projections, we additionally average the results over 3 different computations, and we report this standard deviation in \Cref{tab:table_acc}. Furthermore, we always use 500 projections to approximate the sliced discrepancies. For Frank-Wolfe based methods, we use 10 iterations, which we found to be enough to have a good accuracy. We added an ablation of these two hyperparameters in Figure \ref{fig:ablations_docs}. We report the results obtained with the best $\rho$ for $\RSOT$ and $\SUOT$ computed among a grid $\rho\in\{10^{-4}, 5\cdot 10^{-4}, 10^{-3}, 5\cdot 10^{-3}, 10^{-2}, 10^{-1}, 1\}$. For $\RSOT$, the best $\rho$ is consistently $5\cdot 10^{-3}$ for the Movies and Goodreads datasets, and $5\cdot 10^{-4}$ for the BBCSport dataset. For $\SUOT$, the best $\rho$ obtained was $0.01$ for the BBCSport dataset, $1.0$ for the movies dataset and $0.5$ for the goodreads dataset. For UOT, we used $\rho=1.0$ on the BBCSport dataset. For the movies dataset, the best $\rho$ obtained on a subset was 50, but it took an unreasonable amount of time to run on the full dataset as the runtime increases with $\rho$ (see \cite[Figure 3]{chapel2021unbalanced}). 
On the goodreads dataset, it took too much memory on the GPU. For Sinkhorn UOT, we used $\epsilon=0.001$ and $\rho=0.1$ on the BBCSport and Goodreads datasets, and $\epsilon=0.01$ on the Movies dataset. For each method, the number of neighbors used for the k-NN method is obtained via cross-validation.
\newpage 

\section{Appendix of \Cref{chapter:busemann}}

\subsection{Proofs of \Cref{section:geodesic_rays}}

\subsubsection{Proof of \Cref{prop:geodesic_rays}} \label{proof:prop_geodesic_rays}

\begin{proof}[Proof of \Cref{prop:geodesic_rays}]
    In the setting $c(x,y)=\frac12 \|x-y\|_2^2$ and $\mu_0$ absolutely continuous with respect to the Lebesgue measure, we can apply Brenier's theorem (\Cref{th:brenier}) and hence there is a unique OT map $T$ between $\mu_0$ and $\mu_1$, and $T$ is the gradient of a convex function, \emph{i.e.} $T=\nabla u$ with $u$ convex.

    First, let us suppose that the OT map $T$ between $\mu_0$ and $\mu_1$ is the gradient of a 1-convex function $u$. Let $\mu:t\mapsto \big((1-t)\id + tT)_\#\mu_0 = \big((1-t)\id + t\nabla u\big)_\#\mu_0$. Then, on one hand, we have
    \begin{equation}
        W_2^2(\mu_s,\mu_t) \le (t-s)^2W_2^2(\mu_0,\mu_1).
    \end{equation}
    Indeed, let $\gamma^*\in\Pi(\mu_0,\mu_1)$ be an optimal coupling. Then, necessarily, denoting $\pi^s(x,y) = (1-s) x + s y$, we have for any $s,t\in\mathbb{R}$, $(\pi^s,\pi^t)_\#\gamma^* \in \Pi(\mu_s,\mu_t)$. Therefore,
    \begin{equation}
        \begin{aligned}
            W_2^2(\mu_s,\mu_t) &\le \int \|x-y\|_2^2\ \mathrm{d}(\pi^s,\pi^t)_\#\gamma^*(x,y) \\
            &= \int \|(1-s)x+sy - (1-t)x-ty\|_2^2\ \mathrm{d}\gamma^*(x,y) \\
            &= (s-t)^2 W_2^2(\mu_0,\mu_1).
        \end{aligned}
    \end{equation}
    Then, let $\alpha\ge 1$ and $0\le s<t\le \alpha$. By the triangular inequality and the previous inequality, we have
    \begin{equation}
        \begin{aligned}
            W_2(\mu_0, \mu_\alpha) &\le W_2(\mu_0, \mu_s) + W_2(\mu_s,\mu_t) + W_2(\mu_t, \mu_\alpha) \\
            &= (s+\alpha-t)W_2(\mu_0,\mu_1) + W_2(\mu_s,\mu_t).
        \end{aligned}
    \end{equation}
    If $x\mapsto (1-\alpha)\frac{\|x\|_2^2}{2} + \alpha u(x)$ is convex (\emph{i.e.} $u$ is $\frac{\alpha-1}{\alpha}$-convex), then its gradient which is equal to $x\mapsto (1-\alpha) x + \alpha\nabla u(x)$ is the Monge map between $\mu_0$ and $\mu_\alpha$ as $\mu_\alpha = \big((1-\alpha)\id + \alpha \nabla u\big)_\#\mu_0$, and thus $W_2^2(\mu_0, \mu_\alpha) = \alpha^2 W_2(\mu_0, \mu_1)$. Hence, we obtain
    \begin{equation}
        W_2(\mu_0, \mu_\alpha) = \alpha W_2(\mu_0,\mu_1) \le (s+\alpha-t)W_2(\mu_0,\mu_1) + W_2(\mu_s,\mu_t) \iff (t-s)W_2(\mu_0,\mu_1) \le W_2(\mu_s,\mu_t).
    \end{equation}
    It allows to conclude that $W_2(\mu_s,\mu_t) = |t-s|W_2(\mu_0,\mu_1)$ for all $s,t\in [0,\alpha ]$.
    To extend it on $\mathbb{R}_+$, we need it to be true for all $\alpha \ge 1$, which is true as $u$ is 1-convex. Thus, we can conclude that $t\mapsto \mu_t$ is a geodesic ray.

    For the opposite direction, suppose that $\mu_t = \big((1-t)\id +t T\big)_\#\mu_0$ is a geodesic ray. Then, for all $s\ge 0$,
    \begin{equation}
        \begin{aligned}
            W_2^2(\mu_s, \mu_0) &= s^2 W_2^2(\mu_0, \mu_1) \\
            &= \int \| s (x - \nabla u(x))\|_2^2 \ \mathrm{d}\mu_0(x) \\
            &= \int \| x - (1-s)x - s \nabla u(x) \|_2^2 \ \mathrm{d}\mu_0(x) \\
            &= \int \| x - T_s(x) \|_2^2 \ \mathrm{d}\mu_0(x),
        \end{aligned}
    \end{equation}
    where $(T_s)_\#\mu_0 = \mu_s$ with $T_s:x\mapsto (1-s)x+\nabla u(x)$. By Brenier's theorem, since the OT map is unique and necessarily the gradient of a convex functions, $T_s = \nabla u_s$ with $u_s: x \mapsto (1-s)\frac{\|x\|_2^2}{2} + s u(x) = \frac{\|x\|_2^2}{2} + s \left(u(x)-\frac{\|x\|_2^2}{2} \right)$ convex. 
    Thus, for all $s\ge 0$, 
    \begin{equation}
        I + s (\nabla^2 u - I) \succeq 0 \iff \nabla^2 u - I \succeq - \frac{1}{s} I.
    \end{equation}
    It is true for all $s\ge 0$, hence taking the limit $s\to\infty$, we obtain well $\nabla^2 u - I \succeq 0$, \emph{i.e.} $u$ is 1-convex.
\end{proof}

\subsubsection{Proof of \Cref{prop:1d_geodesic_rays}} \label{proof:prop_1d_geodesic_rays}

\begin{proof}[Proof of \Cref{prop:1d_geodesic_rays}]
    By \citep[Equation 7.2.8]{ambrosio2008gradient}, the quantile of $\mu_t$ is $F_t^{-1} = (1-t)F_0^{-1} + t F_1^{-1}$. Then, we know that $F_t^{-1}$ is a quantile function if and only if it is non-decreasing and continuous. As a linear combination of continuous function, it is always continuous. We only need to find conditions for which it is non-decreasing for all $t\ge 0$.
    Let $0<m<m'<1$, then
    \begin{equation}
        \begin{aligned}
            F_t^{-1}(m) -F_t^{-1}(m') = F_0^{-1}(m)-F_0^{-1}(m') + t\big(F_1^{-1}(m)-F_0^{-1}(m) - F_1^{-1}(m') + F_0^{-1}(m')\big),
        \end{aligned}
    \end{equation}
    and hence,
    \begin{equation}
        \begin{aligned}
            \forall t\ge 0, m'>m,\ F_t^{-1}(m)-F_t^{-1}(m') \le 0 &\iff \forall m'>m,\  F_1^{-1}(m)-F_0^{-1}(m) \le F_1^{-1}(m')-F_0^{-1}(m') \\ &\iff F_1^{-1}-F_0^{-1} \text{ non-decreasing}. \\
        \end{aligned}
    \end{equation}
\end{proof}

\subsection{Proof of \Cref{section:busemann}}

\subsubsection{Proof of \Cref{prop:busemann_closed_1d}} \label{proof:prop_busemann_closed_1d}

\begin{proof}[Proof of \Cref{prop:busemann_closed_1d}]
    $(\mu_t)_{t\ge 0}$ is a unit-speed ray.  Therefore, $W_2(\mu_0,\mu_1)=1$.
    
    Then, we have, for any $\nu\in\mathcal{P}_2(\mathbb{R})$, $t\ge 0$,
    \begin{equation}
        \begin{aligned}
            &W_2(\nu,\mu_t) - t \\ &= \left(\int_0^1 \big(F_\nu^{-1}(u)-F_{\mu_t}^{-1}(u)\big)^2\ \mathrm{d}u\right)^{\frac12} - t \\
            &= \left(\int_0^1 \big(F_\nu^{-1}(u) - (1-t) F_{\mu_0}^{-1}(u) - t F_{\mu_1}^{-1}(u)\big)^2\ \mathrm{d}u\right)^{\frac12} - t \\
            &= \left(\int_0^1 \big(F_\nu^{-1}(u)-F_{\mu_0}^{-1}(u) + t(F_{\mu_0}^{-1}(u)-F_{\mu_1}^{-1}(u))\big)^2\ \mathrm{d}u \right)^{\frac12} - t \\
            &= \Big(\int_0^1 \big(F_\nu^{-1}(u)-F_{\mu_0}^{-1}(u)\big)^2\ \mathrm{d}u + 2t \int_0^1 \big(F_\nu^{-1}(u)-F_{\mu_0}^{-1}(u)\big)\big(F_{\mu_0}^{-1}(u)-F_{\mu_1}^{-1}(u)\big)\ \mathrm{d}u \\ &\quad + t^2 \int_0^1 \big(F_{\mu_0}^{-1}(u)-F_{\mu_1}^{-1}(u)\big)^2\ \mathrm{d}u \Big)^{\frac12} - t \\
            &= t \left(\frac{1}{t^2} \int_0^1 \big(F_\nu^{-1}(u)-F_{\mu_0}^{-1}(u)\big)^2\ \mathrm{d}u + \frac{2}{t} \int_0^1 \big(F_\nu^{-1}(u)-F_{\mu_0}^{-1}(u)\big)\big(F_{\mu_0}^{-1}(u)-F_{\mu_1}^{-1}(u)\big)\ \mathrm{d}u + W_2^2(\mu_0,\mu_1) \right)^{\frac12} - t\\
            &\underset{t\to\infty}{=} t \left(1 + \frac{1}{t} \int_0^1 \big(F_\nu^{-1}(u)-F_{\mu_0}^{-1}(u)\big)\big(F_{\mu_0}^{-1}(u)-F_{\mu_1}^{-1}(u)\big)\ \mathrm{d}u + o\left(\frac{1}{t}\right) \right) - t \\
            &= \int_0^1 \big(F_\nu^{-1}(u)-F_{\mu_0}^{-1}(u)\big)\big(F_{\mu_0}^{-1}(u)-F_{\mu_1}^{-1}(u)\big)\ \mathrm{d}u.
        \end{aligned}
    \end{equation}
\end{proof}

\subsubsection{Proofs of \Cref{prop:closed_form_general_gaussian}} \label{proof:prop_closed_form_general_gaussian}

\begin{proof}[Proof of \Cref{prop:closed_form_general_gaussian}]
    We will use here that for any geodesic ray $\gamma$, $\lim_{t\to\infty}\ \frac{d(x,\gamma(t))+t}{2t} = 1$ (cf \citep[II. 8.24]{bridson2013metric}. Then we know that 
    \begin{equation}
        \lim_{t\to \infty}\ \frac{d(x,\gamma(t))^2-t^2}{2t} = \lim_{t\to\infty}\ \big(d(x,\gamma(t))-t\big),
    \end{equation}
    since
    \begin{equation}
        \frac{d(x,\gamma(t))^2-t^2}{2t} = \frac{(d(x,\gamma(t))-t)(d(x,\gamma(t))+t)}{2t} = \big(d(x,\gamma(t))-t\big) \frac{d(x,\gamma(t))+t}{2t} \xrightarrow[t\to \infty]{} B_\gamma(x).
    \end{equation}

    In our case, we have for any $t\ge 0$, $\mu_t =\mathcal{N}(m_t, \Sigma_t)$ where
    \begin{equation}
        \begin{cases}
            m_t = (1-t)\mo+t \m1 \\
            \Sigma_t = \big((1-t)I_d+t A\big)\So \big((1-t)I_d + t A\big),
        \end{cases}
    \end{equation}
    with $A=\So^{-\frac12}(\So^\frac12\S1\So^\frac12)^\frac12\So^{-\frac12}$. Then, we have, using in particular that $A\So A = \S1$, for any $t\ge 0$,
    \begin{align}
        &\frac{\|m_t-m\|_2^2}{2t} = \frac{t}{2} \|\m1-\mo\|_2^2 + \langle \m1-\mo, \mo-m\rangle + O\left(\frac{1}{t}\right), \\
        &\frac{\mathrm{Tr}(\Sigma_t)}{2t} = \frac{t}{2} \tr\big(\So-2\So A+\S1\big) + \tr\big(\So A-\So\big) + O\left(\frac{1}{t}\right) \\
        &\frac{\tr\big((\Sigma^\frac12 \Sigma_t \Sigma^\frac12)^\frac12\big)}{2t} = \frac12 \tr\left(\left(\Sigma^\frac12 (\So-\So A-A\So+\S1)\Sigma^\frac12 + O\left(\frac{1}{t}\right) \right)^\frac12\right).
    \end{align}
    Then, by remembering that
    \begin{equation}
        W_2^2(\mu_0,\mu_1) = \|\m1-\mo\|_2^2 + \tr(\So+\S1-2(\So^\frac12 \S1 \So^{\frac12})^\frac12) = 1,
    \end{equation}
    and using that
    \begin{equation}
        \So A = \So^{\frac12}(\So^\frac12 \S1 \So^\frac12)^\frac12 \So^{-\frac12}
    \end{equation}
    and hence
    \begin{equation}
        \tr(\So A) = \tr\big((\So^\frac12 \S1 \So^{\frac12})^\frac12\big),
    \end{equation}
    we obtain:
    \begin{equation}
        \begin{aligned}
            &\frac{W_2^2(\nu,\mu_t)-t^2}{2t} \\ &= \frac{\|m_t-m\|_2^2 + \tr\left(\Sigma_t+\Sigma-2(\Sigma^\frac12 \Sigma_t \Sigma^\frac12)^\frac12\right)-t^2}{2t} \\ 
            &= \frac{t}{2}\left(\|\m1-\mo\|_2^2 + \tr(\So + \S1 - 2 \So A)\right) + \langle \m1-\mo, \mo-m\rangle + \tr\big(\So A-\So\big) \\ &\quad -  \tr\left(\left(\Sigma^\frac12 (\So-\So A-A\So+\S1)\Sigma^\frac12 + O\left(\frac{1}{t}\right) \right)^\frac12\right) - \frac{t}{2} + O\left(\frac{1}{t}\right) \\
            &= \frac{t}{2} W_2^2(\mu_0,\mu_1) + \langle \m1-\mo, \mo-m\rangle + \tr\big(\So A-\So\big) \\ &\quad -  \tr\left(\left(\Sigma^\frac12 (\So-\So A-A\So+\S1)\Sigma^\frac12 + O\left(\frac{1}{t}\right) \right)^\frac12\right) - \frac{t}{2} + O\left(\frac{1}{t}\right) \\
            &= \langle \m1-\mo, \mo-m\rangle + \tr\big(\So A-\So\big) \\ &\quad -  \tr\left(\left(\Sigma^\frac12 (\So-\So A-A\So+\S1)\Sigma^\frac12 + O\left(\frac{1}{t}\right) \right)^\frac12\right) + O\left(\frac{1}{t}\right) \\
            &\xrightarrow[t\to\infty]{}  \langle \m1-\mo, \mo-m\rangle + \tr\big(\So(A-I_d) \big) - \tr\big((\Sigma^\frac12(\So-\So A - A\So + \S1)\Sigma^\frac12)^\frac12\big).
        \end{aligned}
    \end{equation}
\end{proof}

\subsection{Proofs of \Cref{section:pca}}

\subsubsection{Proof of \Cref{prop:1dgaussian_pca}} \label{proof:prop_1dgaussian_pca}

\begin{proof}[Proof of \Cref{prop:1dgaussian_pca}]
    First, let us find the first component. We want to solve:
    \begin{equation}
        \begin{aligned}
            & \max_{(m,\sigma)}\ \frac{1}{n}\sum_{i=1}^n\big((m-m_0)(m_i-m_0) + (\sigma-\sigma_0)(\sigma_i-\sigma_0)\big)^2 - \left(\frac{1}{n}\sum_{i=1}^n \big((m-m_0)(m_i-m_0) + (\sigma-\sigma_0)(\sigma_i-\sigma_0)\big)\right)^2 \\
            &\text{subject to }
            \begin{cases}
                (m-m_0)^2 + (\sigma-\sigma_0)^2 = 1 \\
                \sigma-\sigma_0 \ge 0.
            \end{cases}
        \end{aligned}
    \end{equation}
    Let's note for all $i$, $x_i = \begin{pmatrix} m_i-m_0 \\ \sigma_i-\sigma_0\end{pmatrix}$ and $x=\begin{pmatrix} m-m_0 \\ \sigma-\sigma_0\end{pmatrix}$. Then, the objective can be rewritten as 
    \begin{equation}
        \max_x\ \frac{1}{n} x^T\left(\sum_{i=1}^n x_i x_i^T\right) x - x^T \left(\frac{1}{n}\sum_{i=1}^n x_i\right)\left(\frac{1}{n}\sum_{i=1}^n x_i\right)^T x \quad  \text{subject to} \quad  \begin{cases}
            \|x\|_2^2 = 1 \\
            [x]_2 \ge 0.
        \end{cases}
    \end{equation}

    This is a convex objective on a compact space. Let's encode the constraints by using a parametrization on the circle. Indeed, as $\|x\|_2^2 = 1$ and $[x]_2 \ge 0$, there exists $\theta\in[0,\pi]$ such that $x = x_\theta = \begin{pmatrix} \cos\theta \\ \sin\theta\end{pmatrix}$.
    Now, let $M = \frac{1}{n}\sum_{i=1}^n x_i x_i^T - \left(\frac{1}{n}\sum_{i=1}^n x_i\right) \left(\frac{1}{n}\sum_{i=1}^n x_i\right)^T$ and rewrite the objective as
    \begin{equation}
        \begin{aligned}
            \frac{1}{n} x_\theta^T\left(\sum_{i=1}^n x_i x_i^T\right) x_\theta - x_\theta^T \left(\frac{1}{n}\sum_{i=1}^n x_i\right)\left(\frac{1}{n}\sum_{i=1}^n x_i\right)^T x_\theta &= x_\theta^T M x_\theta \\
            &= \cos^2(\theta) M_{11} + \sin^2(\theta) M_{22} + 2 \cos(\theta) \sin(\theta) M_{12} \\
            &=\frac{1+\cos(2\theta)}{2} M_{11} + \frac{1-\cos(2\theta)}{2} M_{22} + \sin(2\theta) M_{12} \\
            &= \frac{M_{11}-M_{22}}{2} \cos(2\theta) + \frac12 (M_{11}+M_{22}) + \sin(2\theta) M_{12}.        
        \end{aligned}
    \end{equation}

    Thus the problem is equivalent with
    \begin{equation}
        \max_{\Tilde{\theta}\in [0,2\pi[}\ \frac{M_{11}-M_{22}}{2} \cos(\Tilde{\theta}) + M_{12} \sin(\Tilde{\theta}) = f(\Tilde{\theta}).
    \end{equation}
    Noting $a=\frac{M_{11}-M_{22}}{2}$, $b=M_{12}$, and posing $\phi$ such that $\cos\phi = \frac{a}{\sqrt{a^2 + b^2}}$ and $\sin\phi = \frac{b}{\sqrt{a^2+b^2}}$, we can rewrite $f$ as 
    \begin{equation}
        f(\Tilde{\theta}) = \sqrt{a^2 + b^2} \big(\cos\phi \cos\Tilde{\theta} + \sin\phi \sin \Tilde{\theta}\big) = \sqrt{a^2 + b^2} \cos(\Tilde{\theta}-\phi).
    \end{equation}
    Hence, $f$ is maximal for $\Tilde{\theta}=\phi=\arccos\left(\frac{a}{\sqrt{a^2 + b^2}}\right) = \arccos\left(\frac{M_{11}-M_{22}}{\sqrt{(M_{11}-M_{22})^2 + 4M_{12}^2}}\right)$.
    Finally, we get the optimum as
    \begin{equation}
        \begin{cases}
            m^{(1)}_1 = m_0 + \cos\left(\frac{\Tilde{\theta}}{2}\right) \\
            \sigma^{(1)}_1 = \sigma_0 + \sin\left(\frac{\Tilde{\theta}}{2}\right).
        \end{cases}
    \end{equation}

    The second component is fully characterized by the orthogonal condition. Noting $\psi\in [0,\pi[$ the angle such that $x_\psi = \begin{pmatrix}\cos\phi \\ \sin\psi \end{pmatrix} = \begin{pmatrix} m-m_0 \\ \sigma-\sigma_0\end{pmatrix}$ is the solution of 
    \begin{equation}
        \max_{\psi\in [0,\pi]}\ x_\psi^T M x_\psi \quad \text{subject to} \quad \langle x_\theta, x_\psi\rangle = 0.
    \end{equation}
    Then, $\langle x_\theta, x_\psi\rangle = 0 \iff \cos(\theta-\psi)=0 \iff \psi = \theta \pm \frac{\pi}{2}$. Since $\psi\in [0,\pi[$, if $\theta\ge\frac{\pi}{2}$ then $\psi = \theta - \frac{\pi}{2}$. If $\theta<\frac{\pi}{2}$, then $\psi = \theta + \frac{\pi}{2}$. To conclude, the second component is obtained with
    \begin{equation}
        \begin{cases}
            m_1^{(2)} = m_0 + \cos\left(\frac{\Tilde{\theta} - \mathrm{sign}(\Tilde{\theta} - \pi)\pi}{2}\right) \\
            \sigma_1^{(2)} = \sigma_0 + \sin\left(\frac{\Tilde{\theta} - \mathrm{sign}(\Tilde{\theta} - \pi)\pi}{2}\right).
        \end{cases}
    \end{equation}
\end{proof}

\subsubsection{Proof of \Cref{lemma:extension_1dgaussian_ray}} \label{proof:lemma_extension_1dgaussian_ray}

\begin{proof}[Proof of \Cref{lemma:extension_1dgaussian_ray}]
    Let $\mu_0=\mathcal{N}(m_0,\sigma_0^2)$ and $\mu_1=\mathcal{N}(m_1, \sigma_1^2)$ such that $(m_1-m_0)^2+(\sigma_1-\sigma_0)^2=1$ and $\sigma_1\ge \sigma_0$. Extending the geodesic between $\mu_0$ and $\mu_1$ on $[1-\alpha, 0]$ for $\alpha \ge 1$ is equivalent to extending the geodesic between $\mu_1$ and $\mu_0$ on $[0, \alpha]$. Thus, we first find a condition to extend the geodesic between $\mu_1$ and $\mu_0$.

    The Monge map $\Tilde{T}$ between $\mu_1$ and $\mu_0$ is defined for all $x\in\mathbb{R}$ as $\Tilde{T}(x) = \frac{\sigma_0}{\sigma_1}(x-m_1) + m_0 = h'(x)$ with $h:x\mapsto \frac{\sigma_0}{\sigma_1}(x-m_1)^2 + m_0 x$.  Then, by \citep[Section 4]{natale2022geodesic}, we know that we can extend the geodesic linking $\mu_1$ to $\mu_0$ on $[0,\alpha]$ for $\alpha\ge 1$ if and only if $h$ is $\frac{\alpha-1}{\alpha}$-convex, \emph{i.e.} if and only if
    \begin{equation} \label{eq:inequality_alpha}
        h''(x) - \frac{\alpha -1}{\alpha} \ge 0 \iff \frac{\sigma_0}{\sigma_1} \ge \frac{\alpha-1}{\alpha} \iff \frac{\sigma_1}{\sigma_0} \le \frac{\alpha}{\alpha-1}.
    \end{equation}

    Therefore, we deduce that we can extend the geodesic ray starting from $\mu_0$ and passing through $\mu_1$ at $t=1$ on $[-(\alpha-1), +\infty[$ if and only if $\frac{\alpha}{\alpha - 1} \ge \frac{\sigma_1}{\sigma_0} \ge 1$ (the last inequality comes from the condition to have a geodesic ray $\sigma_1\ge \sigma_0$).

    Since $(m_1-m_0)^2 + (\sigma_1-\sigma_0)^2 = 1$, it implies that necessarily, $\sigma_1 - \sigma_0 \le 1 \iff \frac{\sigma_1}{\sigma_0} \le \frac{1}{\sigma_0} + 1$. Thus, we find that the biggest possible $\alpha\ge 1$ satisfying the inequality \eqref{eq:inequality_alpha} is $\frac{\sigma_1}{\sigma_1-\sigma_0}$ as
    \begin{equation}
        \begin{aligned}
            \frac{\alpha}{\alpha -1} \ge \frac{\sigma_1}{\sigma_0} \iff \alpha \frac{\sigma_0-\sigma_1}{\sigma_0} \ge - \frac{\sigma_1}{\sigma_0} \iff \alpha \le \frac{\sigma_1}{\sigma_1-\sigma_0},
        \end{aligned}
    \end{equation}
    and for $\alpha=\frac{\sigma_1}{\sigma_1-\sigma_0}$, $\frac{\alpha}{\alpha-1} = \frac{\sigma_1}{\sigma_0}$. In this case, $1-\alpha = -\frac{\sigma_0}{\sigma_1-\sigma_0}$ and hence the geodesic ray can be extended at least to $[-\frac{\sigma_0}{\sigma_1-\sigma_0}, +\infty[$.
\end{proof}
\newpage

\section{Appendix of \Cref{chapter:gw}}

\subsection{Proofs of \Cref{section:subspace_detours_gw}} \label{appendix:proofs_gw}

\subsubsection{Proof of \Cref{prop:prop1}} \label{proof:prop1}

\begin{proof}[Proof of \Cref{prop:prop1}]
    Let $\gamma\in\Pi_{E,F}(\mu,\nu)$, then:
    \begin{equation}
        \begin{aligned}
             &\iint L(x,x',y,y') \mathrm{d}\gamma(x,y)\mathrm{d}\gamma(x',y') \\ 
             &= \iint \Big(\iint L(x,x',y,y') \gamma_{E^\bot\times F^\bot|E\times F}\big((x_E,y_F),(\mathrm{d}x_{E^\bot},\mathrm{d}y_{F^\bot})\big) \\ &\gamma_{E^\bot\times F^\bot|E\times F}\big((x_E',y_F'),(\mathrm{d}x_{E^\bot}',\mathrm{d}y_{F^\bot}')\big)\Big)\mathrm{d}\gamma_{E\times F}^*(x_E,y_F) \mathrm{d}\gamma_{E\times F}^*(x_E',y_F').
        \end{aligned}
    \end{equation}
    However, for~$\gamma_{E\times F}^*$ a.e. $(x_E,y_F),(x_E',y_F')$,
\begingroup\makeatletter\def\f@size{9.8}\check@mathfonts
\def\maketag@@@#1{\hbox{\m@th\normalsize\normalfont#1}}%
    \begin{equation}
        \begin{aligned}
            &\iint L(x,x',y,y') \gamma_{E^\bot\times F^\bot|E\times F}\big((x_E,y_F),(\mathrm{d}x_{E^\bot},\mathrm{d}y_{F^\bot})\big) \gamma_{E^\bot\times F^\bot|E\times F}\big((x_E',y_F'),(\mathrm{d}x_{E^\bot}',\mathrm{d}y_{F^\bot}')\big) \\
            &\geq \iint L(x,x',y,y') \gamma_{E^\bot\times F^\bot|E\times F}^*\big((x_E,y_F),(\mathrm{d}x_{E^\bot},\mathrm{d}y_{F^\bot})\big) \gamma_{E^\bot\times F^\bot|E\times F}^*\big((x_E',y_F'),(\mathrm{d}x_{E^\bot}',\mathrm{d}y_{F^\bot}')\big)
        \end{aligned}
    \end{equation}\endgroup
    by definition of the Monge--Knothe coupling. 
    By integrating with respect to $\gamma_{E\times F}^*$, we obtain:
    \begin{equation}
        \iint L(x,x',y,y') \mathrm{d}\gamma(x,y)\mathrm{d}\gamma(x',y') \ge \iint L(x,x',y,y')\mathrm{d}\pi_{\mathrm{MK}}(x,y)\mathrm{d}\pi_{\mathrm{MK}}(x',y').
    \end{equation}
    Therefore, $\pi_{\mathrm{MK}}$ is optimal for subspace optimal plans.
\end{proof}

\subsubsection{Proof of \Cref{prop:prop2}} \label{proof:prop2}

\begin{proof}[Proof of \Cref{prop:prop2}]
    We first deal with $L(x,x',y,y')=\big(\|x-x'\|_2^2 - \|y-y'\|_2^2\big)^2$. Let $f_{E^\bot}$ be an isometry \emph{w.r.t} $c(x_{E^\bot},x'_{E^\bot})=\|x_{E^\bot}-x'_{E^\bot}\|_2^2$, and let $f:\mathbb{R}^p\to\mathbb{R}^p$ be defined such as for all $x\in\mathbb{R}^p$, $f(x)=(x_E,f_{E^\bot}(x_{E^\bot}))$.
    
    From \Cref{lemma:paty}, we know that $\Pi(f_\#\mu,\nu)=\{(f,Id)_\#\gamma|\ \gamma\in\Pi(\mu,\nu)\}$. We can rewrite:
    \begin{equation}
        \begin{aligned}
            \Pi_{E,F}(f_\#\mu,\nu)&=\{\gamma\in\Pi(f_\#\mu,\nu)|(\pi^E,\pi^F)_\#\gamma=\gamma_{E\times F}^*\} \\
            &= \{(f,Id)_\#\gamma | \gamma\in\Pi(\mu,\nu), (\pi^E,\pi^F)_\#(f,Id)_\#\gamma=\gamma_{E\times F}^*\} \\
            &=  \{(f,Id)_\#\gamma | \gamma\in\Pi(\mu,\nu), (\pi^E,\pi^F)_\#\gamma=\gamma_{E\times F}^*\} \\
            &= \{(f,Id)_\#\gamma|\gamma\in\Pi_{E,F}(\mu,\nu)\}
        \end{aligned}
    \end{equation}
    using $f=(Id_E, f_{E^\bot})$, $\pi^E\circ f = Id_E$ and $(\pi^E,\pi^F)_\#(f,Id)_\#\gamma=(\pi^E,\pi^F)_\#\gamma$.
    
    Now, for~all $\gamma\in\Pi_{E,F}(f_\#\mu,\nu)$, there exists $\Tilde{\gamma}\in\Pi_{E,F}(\mu,\nu)$ such that $\gamma=(f,Id)_\#\Tilde{\gamma}$, and we can disintegrate $\Tilde{\gamma}$ with respect to $\gamma_{E\times F}^*$:
    \begin{equation}
        \Tilde{\gamma} = \gamma_{E\times F}^* \otimes K
    \end{equation}
    with $K$ a probability kernel on $(E\times F, \mathcal{B}(E^\bot)\otimes \mathcal{B}(F^\bot))$.
    
    For $\gamma_{E\times F}^*$ almost every $(x_E,y_F),\ (x_E',y_F')$, we have:
    \begin{equation}
        \begin{aligned}
            &\iint \big(\|x_E-x_E'\|_2^2+\|x_{E^\bot}-x_{E^\bot}'\|_2^2-\|y_F-y_F'\|_2^2-\|y_{F^\bot}-y_{F^\bot}'\|_2^2\big)^2 \\
            &\ (f_{E^\bot},Id)_\# K((x_E,y_F),(\mathrm{d}x_{E^\bot},\mathrm{d}y_{F^{\bot}})) (f_{E^\bot},Id)_\# K((x_E',y_F'),(\mathrm{d}x_{E^\bot}',\mathrm{d}y_{F^{\bot}}')) \\
            &= \iint \big(\|x_E-x_E'\|_2^2+\|f_{E^\bot}(x_{E^\bot})-f_{E^\bot}(x_{E^\bot}')\|_2^2-\|y_F-y_F'\|_2^2-\|y_{F^\bot}-y_{F^\bot}'\|_2^2\big)^2 \\
            &\ K((x_E,y_F),(\mathrm{d}x_{E^\bot},\mathrm{d}y_{F^{\bot}})) K((x_E',y_F'),(\mathrm{d}x_{E^\bot}',\mathrm{d}y_{F^{\bot}}')) \\
            &= \iint \big(\|x_E-x_E'\|_2^2+\|x_{E^\bot}-x_{E^\bot}'\|_2^2-\|y_F-y_F'\|_2^2-\|y_{F^\bot}-y_{F^\bot}'\|_2^2\big)^2 \\
            &\ K((x_E,y_F),(\mathrm{d}x_{E^\bot},\mathrm{d}y_{F^{\bot}})) K((x_E',y_F'),(\mathrm{d}x_{E^\bot}',\mathrm{d}y_{F^{\bot}}'))
        \end{aligned}
    \end{equation}
    using in the last line that $\|f_{E^\bot}(x_{E^\bot})-f_{E^\bot}(x_{E^\bot}')\|_2 = \|x_{E^\bot}-x_{E^\bot}'\|_2$ since $f_{E^\bot}$ is an isometry.
    
    By integrating with respect to $\gamma_{E\times F}^*$, we obtain:
\begin{equation} \label{eq:equality_K}
        \begin{aligned}
        &\iint\Big(\iint \big(\|x-x'\|_2^2 - \|y-y'\|_2^2\big)^2 \\
        &\ (f_{E^\bot},Id)_\# K((x_E,y_F),(\mathrm{d}x_{E^\bot},\mathrm{d}y_{F^{\bot}})) (f_{E^\bot},Id)_\# K((x_E',y_F'),(\mathrm{d}x_{E^\bot}',\mathrm{d}y_{F^{\bot}}')) \Big) \\ &\mathrm{d}\gamma_{E\times F}^*(x_E,y_F) \mathrm{d}\gamma_{E\times F}^*(x_E',y_F') \\
        &= \iint \big(\|x-x'\|_2^2 - \|y-y'\|_2^2\big)^2\ \mathrm{d}\Tilde{\gamma}(x,y)\mathrm{d}\Tilde{\gamma}(x',y').
        \end{aligned}
    \end{equation}
    
    Now, we show that $\gamma=(f,Id)_\#\Tilde{\gamma} = \gamma_{E\times F}^*\otimes (f_{E^\bot},Id)_\# K$. Let $\phi$ be some bounded measurable function on $\mathbb{R}^p\times\mathbb{R}^q$:
    \begin{equation}
        \begin{aligned}
            \int \phi(x,y)\mathrm{d}\gamma(x,y) &= \int \phi(x,y) \mathrm{d}((f,Id)_\#\Tilde{\gamma}(x,y)) \\
            &= \int \phi(f(x),y) \mathrm{d}\Tilde{\gamma}(x,y) \\
            &= \iint \phi(f(x),y) K\big((x_E,y_F),(\mathrm{d}x_{E^\bot}, \mathrm{d}y_{F^\bot})\big)\ \mathrm{d}\gamma_{E\times F}^*(x_E,y_F) \\
            &= \iint \phi((x_E,f_{E^\bot}(x_{E^\bot})),y) K\big((x_E,y_F),(\mathrm{d}x_{E^\bot}, \mathrm{d}y_{F^\bot})\big)\ \mathrm{d}\gamma_{E\times F}^*(x_E,y_F) \\
            &= \iint \phi(x,y) (f_{E^\bot},Id)_\# K\big((x_E,y_F),(\mathrm{d}x_{E^\bot}, \mathrm{d}y_{F^\bot})\big)\ \mathrm{d}\gamma_{E\times F}^*(x_E,y_F).
        \end{aligned}
    \end{equation}
    Hence, we can rewrite \eqref{eq:equality_K} as:
    \begin{equation}
        \begin{aligned}
            &\iint \big(\|x-x'\|_2^2-\|y-y'\|_2^2\big)^2\ \mathrm{d}(f,Id)_\#\Tilde{\gamma}(x,y) \mathrm{d}(f,Id)_\#\Tilde{\gamma}(x',y') \\ 
            &= \iint \big(\|x-x'\|_2^2-\|y-y'\|_2^2\big)^2\     \mathrm{d}\Tilde{\gamma}(x,y)\mathrm{d}\Tilde{\gamma}(x',y').
        \end{aligned}
    \end{equation}
    
    Now, by~taking the infimum with respect to $\Tilde{\gamma}\in\Pi_{E,F}(\mu,\nu)$, we find:
    \begin{equation}
        GW_{E,F}(f_\#\mu,\nu) = GW_{E,F}(\mu,\nu).
    \end{equation}
    
    For the inner product case, we can do the same proof for linear isometries on $E^\bot$.
\end{proof}

\subsubsection{Proof of \Cref{prop:closed_form_mk}} \label{proof:prop_closed_form_mk}

For $GW$ with $c(x,x')=\|x-x'\|_2^2$, we have for now no guarantee that there exists an optimal coupling which is a transport map. \citet{salmona2021gromov} proposed to restrict the problem to the set of Gaussian couplings $\pi(\mu,\nu)\cap \mathcal{N}_{p+q}$ where $\mathcal{N}_{p+q}$ denotes the set of Gaussians in $\mathbb{R}^{p+q}$. In~that case, the~problem becomes:
\begin{equation} \label{ggw}
    GGW(\mu,\nu)=\inf_{\gamma\in\Pi(\mu,\nu)\cap \mathcal{N}_{p+q}}\ \iint \big(\|x-x'\|_2^2 - \|y-y'\|_2^2\big)^2 \mathrm{d}\gamma(x,y)\mathrm{d}\gamma(x',y').
\end{equation}

In that case, they showed that an optimal solution is of the form $T(x)=m_\nu+P_\nu A P_\mu^T(x-m_\mu)$ with $A=\begin{pmatrix} \Tilde{I}_q D_\nu^{\frac12}(D_\mu^{(q)})^{-\frac12} & 0_{q,p-q} \end{pmatrix}$ and $\Tilde{I}_q$ of the form $\mathrm{diag}\big((\pm 1)_{i\le q}\big)$.

\begin{proof}[Proof of \Cref{prop:closed_form_mk}]
    Since the problem is translation invariant, we can always solve the problem between the centered~measures.
    
    In the following, we suppose that $k=k'$. Let us denote $T_{E,F}$ as the optimal transport map for \eqref{ggw} between $\mathcal{N}(0,\Sigma_E)$ and $\mathcal{N}(0,\Lambda_F)$. According to \citet[Theorem 4.1]{salmona2021gromov}, such a solution exists and is of the form \eqref{monge_map_gw}. We also denote $T_{E^\bot,F^\bot}$ as the optimal transport map between $\mathcal{N}(0,\Sigma/\Sigma_E)$ and $\mathcal{N}(0,\Lambda/\Lambda_F)$ (which is well defined since we assumed $p\ge q$ and hence $p-k\ge q-k'$ since $k=k'$).
    
    We know that the Monge--Knothe transport map will be a linear map $T_{\mathrm{MK}}(x)=Bx$ with $B$ a block triangular matrix of the form:
    \begin{equation}
        B = \begin{pmatrix}
            T_{E,F} & 0_{k',p-k} \\
            C & T_{E^\bot,F^\bot}
        \end{pmatrix} \in \mathbb{R}^{q\times p},
    \end{equation}
    with $C\in\mathbb{R}^{(q-k')\times k}$ and~such that $B\Sigma B^T = \Lambda$ (to have well a transport map between $\mu$ and~$\nu$).
    
    Actually, 
    \begin{equation}
        B\Sigma B^T = \begin{pmatrix}
            T_{E,F}\Sigma_E T_{E,F}^T & T_{E,F}\Sigma_E C^T + T_{E,F}\Sigma_{EE^\bot} T_{E^\bot,F^\bot}^T \\
            (C\Sigma_E + T_{E^\bot,F^\bot}\Sigma_{E^\bot E})T_{E,F}^T & (C\Sigma_E+T_{E^\bot,F^\bot}\Sigma_{E^\bot E})C^T + (C\Sigma_{EE^\bot}+T_{E^\bot,F^\bot}\Sigma_{E^\bot})T_{E^\bot,F^\bot}^T
        \end{pmatrix}.
    \end{equation}

    First, we have well $T_{E,F}\Sigma_E T_{E,F}^T = \Lambda_F$, as $T_{E,F}$ is a transport map between $\mu_E$ and $\nu_F$. Then:
    \begin{equation}
        B\Sigma B^T = \Lambda \iff \begin{cases}
            T_{E,F}\Sigma_E T_{E,F}^T = \Lambda_F \\
            T_{E,F}\Sigma_E C^T + T_{E,F}\Sigma_{EE^\bot} T_{E^\bot,F^\bot}^T = \Lambda_{FF^\bot} \\
            (C\Sigma_E + T_{E^\bot,F^\bot}\Sigma_{E^\bot E})T_{E,F}^T = \Lambda_{F^\bot F} \\
            (C\Sigma_E+T_{E^\bot,F^\bot}\Sigma_{E^\bot E})C^T + (C\Sigma_{EE^\bot}+T_{E^\bot,F^\bot}\Sigma_{E^\bot})T_{E^\bot,F^\bot}^T = \Lambda_{F^\bot}.
        \end{cases}
    \end{equation}
    
    We have: 
    \begin{equation}
        (C\Sigma_E + T_{E^\bot,F^\bot}\Sigma_{E^\bot E})T_{E,F}^T = \Lambda_{F^\bot F} \iff C\Sigma_E T_{E,F}^T = \Lambda_{F^\bot F} - T_{E^\bot, F^\bot}\Sigma_{E^\bot E} T_{E,F}^T.
    \end{equation}
    
    As $k=k'$, $\Sigma_E T_{E,F}^T\in\mathbb{R}^{k\times k}$ and is invertible (as $\Sigma_E$ and $\Lambda_F$ are positive definite and $T_{E,F}=P_{\mu_E}A_{E,F}P_{\nu_F}$ with $A_{E,F}=\begin{pmatrix} \Tilde{I}_k D_{\nu_F}^{\frac11} D_{\mu_E}^{-\frac12} \end{pmatrix}$ with positive values on the diagonals. Hence, we have: 
    \begin{equation}
        C = (\Lambda_{F^\bot F} (T_{E,F}^T)^{-1} - T_{E^\bot,F^\bot}\Sigma_{E^\bot E})\Sigma_E^{-1}.
    \end{equation}
    
    Now, we still have to check the last two equations. First:
    \begin{equation}
        \begin{aligned}
            T_{E,F}\Sigma_E C^T + T_{E,F}\Sigma_{EE^\bot} T_{E^\bot,F^\bot}^T &= T_{E,F}\Sigma_E \Sigma_E^{-1} T_{E,F}^{-1}\Lambda_{F^\bot F}^T  - T_{E,F}\Sigma_E \Sigma_E^{-1} \Sigma_{E^\bot E}^T T_{E^\bot, F^\bot}^T+ T_{E,F}\Sigma_{EE^\bot} T_{E^\bot,F^\bot}^T \\
            &= \Lambda_{FF^\bot}.
        \end{aligned}
    \end{equation}\vspace{3pt}

    For the last equation:\vspace{-9pt}
    
    \begingroup\makeatletter\def\f@size{9.8}\check@mathfonts
    \def\maketag@@@#1{\hbox{\m@th\normalsize\normalfont#1}}%
    \begin{equation}
        \begin{aligned}
            &(C\Sigma_E+T_{E^\bot,F^\bot}\Sigma_{E^\bot E})C^T + (C\Sigma_{EE^\bot}+T_{E^\bot,F^\bot}\Sigma_{E^\bot})T_{E^\bot,F^\bot}^T \\
            &= (\Lambda_{F^\bot F} (T_{E,F}^T)^{-1} - T_{E^\bot,F^\bot}\Sigma_{E^\bot E}+T_{E^\bot,F^\bot}\Sigma_{E^\bot E}) \Sigma_E^{-1} (T_{E,F}^{-1}\Lambda_{F^\bot F}^T-\Sigma_{E^\bot E}^T T_{E^\bot,F^\bot}^T) \\
            &\ +\Lambda_{F^\bot F} (T_{E,F}^T)^{-1} \Sigma_E^{-1}\Sigma_{EE^\bot} T_{E^\bot, F^\bot}^T - T_{E^\bot,F^\bot}\Sigma_{E^\bot E}\Sigma_E^{-1} \Sigma_{EE^\bot} T_{E^\bot, F^\bot}^T + T_{E^\bot, F^\bot}\Sigma_{E^\bot} T_{E^\bot, F^\bot}^T \\
            &= \Lambda_{F^\bot F} (T_{E,F}^T )^{-1} \Sigma_E^{-1} T_{E,F}^{-1} \Lambda_{F^\bot F}^T - \Lambda_{F^\bot F} (T_{E,F}^T)^{-1} \Sigma_E^{-1}\Sigma_{E^\bot E}^T T_{E^\bot, F^\bot}^T - T_{E^\bot,F^\bot}\Sigma_{E^\bot E}\Sigma_E^{-1}T_{E,F}^{-1}\Lambda_{F^\bot F}^T \\
            &\ + T_{E^\bot,F^\bot}\Sigma_{E^\bot E} \Sigma_E^{-1} \Sigma_{E^\bot E}^T T_{E^\bot,F^\bot}^T + T_{E^\bot,F^\bot}\Sigma_{E^\bot E} \Sigma_E^{-1} T_{E,F}^{-1} \Lambda_{F^\bot F}^T - T_{E^\bot,F^\bot} \Sigma_{E^\bot E} \Sigma_E^{-1} \Sigma_{E^\bot E}^T T_{E^\bot F^\bot}^T \\
            &\ + \Lambda_{F^\bot F} (T_{E,F}^T)^{-1}\Sigma_E^{-1}\Sigma_{EE^\bot} T_{E^\bot,F^\bot}^T - T_{E^\bot,F^\bot}\Sigma_{E^\bot E}\Sigma_E^{-1}\Sigma_{E^\bot E}^T T_{E^\bot,F^\bot}^T + T_{E^\bot,F^\bot}\Sigma_{ E^\bot}T_{E^\bot,F^\bot}^T \\
            &= \Lambda_{F^\bot F} (T_{E,F}^T)^{-1}\Sigma_E^{-1}T_{E,F}^{-1}\Lambda_{F^\bot F}^T - T_{E^\bot,F^\bot} \Sigma_{E^\bot E}\Sigma_E^{-1} \Sigma_{E^\bot E}^T T_{E^\bot,F^\bot}^T + T_{E^\bot,F^\bot}\Sigma_{E^\bot} T_{E^\bot,F^\bot}^T
        \end{aligned}
    \end{equation}\endgroup
    Now, using that $(T_{E,F}^T)^{-1}\Sigma_E^{-1}T_{E,F}^{-1} = (T_{E,F}\Sigma_E T_{E,F}^T)^{-1} = \Lambda_F^{-1}$ and $\Sigma_{E^\bot}-\Sigma_{E^\bot E}\Sigma_E^{-1}\Sigma_{E^\bot E}^T = \Sigma/\Sigma_E$, we have:
    \begin{equation}
        \begin{aligned}
            &(C\Sigma_E+T_{E^\bot,F^\bot}\Sigma_{E^\bot E})C^T + (C\Sigma_{EE^\bot}+T_{E^\bot,F^\bot}\Sigma_{E^\bot})T_{E^\bot,F^\bot}^T \\
            &= \Lambda_{F^\bot F}\Lambda_F^{-1}\Lambda_{F^\bot F}^T + T_{E^\bot,F^\bot} (\Sigma_{E^\bot}-\Sigma_{E^\bot E}\Sigma_E^{-1}\Sigma_{E^\bot E}^T) T_{E^\bot, F^\bot}^T \\
            &= \Lambda_{F^\bot F}\Lambda_F^{-1}\Lambda_{F^\bot F}^T + \Lambda/\Lambda_F \\
            &= \Lambda_{F^\bot}
        \end{aligned}
    \end{equation}

    Then, $\pi_{\mathrm{MK}}$ is of the form $(Id,T_{\mathrm{MK}})_\#\mu$ with:
    \begin{equation}
        T_{\mathrm{MK}}(x) = m_\nu+B(x-m_\mu).
    \end{equation}
\end{proof}

\subsubsection{Proof of \Cref{prop:closed_form_mi}} \label{proof:prop_closed_form_mi}

\begin{proof}[Proof of \Cref{prop:closed_form_mi}]
    Suppose $k\ge k'$ in order to be able to define the OT map between $\mu_E$ and $\nu_F$.
    
    For the Monge--Independent plan, $\pi_{\mathrm{MI}}=\gamma_{E\times F}^*\otimes(\mu_{E^\bot|E}\otimes \nu_{F^\bot|F})$, let $(X,Y)\sim \pi_{\mathrm{MI}}$. We know that $\pi_{\mathrm{MI}}$ is a degenerate Gaussian with a covariance of the form:
    \begin{equation}
        \mathrm{Cov}(X,Y) = \begin{pmatrix}
            \mathrm{Cov}(X) & C \\
            C^T & \mathrm{Cov}(Y)
        \end{pmatrix}
    \end{equation}
    where $\mathrm{Cov}(X)=\Sigma$ and $\mathrm{Cov}(Y)=\Lambda$. Moreover, we know that $C$ is of the form:
    \begin{equation}
        \begin{pmatrix}
            \mathrm{Cov}(X_E,Y_F) & \mathrm{Cov}(X_E,Y_{F^\bot}) \\
            \mathrm{Cov}(X_{E^\bot},Y_F) & \mathrm{Cov}(X_{E^\bot},Y_{F^\bot})
        \end{pmatrix}.
    \end{equation}
    
    Let us assume that $m_\mu=m_\nu=0$, then: 
    \begin{equation}
        \begin{aligned}
            \mathrm{Cov}(X_E,Y_F) &= \mathrm{Cov}(X_E,T_{E,F}X_E) = \mathbb{E}[X_EX_E^T]T_{E,F}^T = \Sigma_E T_{E,F}^T,
        \end{aligned}
    \end{equation}
    \begin{equation}
        \begin{aligned}
            \mathrm{Cov}(X_E,Y_{F^\bot}) &= \mathbb{E}[X_E Y_{F^\bot}^T] \\
            &= \mathbb{E}[\mathbb{E}[X_E Y_{F^\bot}^T|X_E,Y_F]] \\
            &= \mathbb{E}[X_E\mathbb{E}[Y_{F^\bot}^T|Y_F]]
        \end{aligned}
    \end{equation}
    since $Y_F=T_{E,F}X_E$, $X_E$ is $\sigma(Y_F)$-measurable. Now, using the equation (A.6) from \citet{rasmussen2003gaussian}, we have:
    \begin{equation}
        \begin{aligned}
            \mathbb{E}[Y_{F^\bot}|Y_F] &= \Lambda_{F^\bot F}\Lambda_F^{-1}Y_F \\
            &= \Lambda_{F^\bot F}\Lambda_F^{-1}T_{E,F} X_E
        \end{aligned}
    \end{equation}
    and 
    \begin{equation}
        \mathbb{E}[X_{E^\bot}|X_E] = \Sigma_{E^\bot E} \Sigma_E^{-1} X_E.
    \end{equation}
    
    Hence:
    \begin{equation}
        \begin{aligned}
            \mathrm{Cov}(X_E,Y_{F^\bot}) &=\mathbb{E}[X_E\mathbb{E}[Y_{F^\bot}^T|Y_F]] \\
            &= \mathbb{E}[X_E X_E^T] T_{E,F}^T \Lambda_F^{-1} \Lambda_{F^\bot F}^T \\
            &= \Sigma_E T_{E,F}^T \Lambda_F^{-1} \Lambda_{F^\bot F}^T.
        \end{aligned}
    \end{equation}
    
    We also have:
    \begin{equation}
        \mathrm{Cov}(X_{E^\bot},Y_F) = \mathbb{E}[X_{E^\bot} X_E^T T_{E,F}^T] = \Sigma_{E^\bot E}T_{E,F}^T,
    \end{equation}
    and
    \begin{equation}
        \begin{aligned}
            \mathrm{Cov}(X_{E^\bot}, Y_{F^\bot}) &= \mathbb{E}[X_{E^\bot} Y_{F^\bot}^T] \\
            &= \mathbb{E}[\mathbb{E}[X_{E^\bot}Y_{F^\bot}^T|X_E,Y_F]] \\
            &= \mathbb{E}[\mathbb{E}[X_{E^\bot}|X_E]\mathbb{E}[Y_{F^\bot}^T|Y_F]]\ \text{ by independence} \\
            &= \mathbb{E}[\Sigma_{E^\bot E}\Sigma_E^{-1} X_E X_E^T T_{E,F}^T \Lambda_F^{-1}\Lambda_{F^\bot F}^T] \\
            &= \Sigma_{E^\bot E}T_{E,F}^T \Lambda_F^{-1}\Lambda_{F^\bot F}^T.
        \end{aligned}
    \end{equation}
    
    Finally, we find:
    \begin{equation}
        C = \begin{pmatrix}
            \Sigma_E T_{E,F}^T & \Sigma_E T_{E,F}^T \Lambda_F^{-1} \Lambda_{F^\bot F}^T \\
            \Sigma_{E^\bot E}T_{E,F}^T & \Sigma_{E^\bot E}T_{E,F}^T \Lambda_F^{-1}\Lambda_{F^\bot F}^T
        \end{pmatrix}.
    \end{equation}
    
    By taking orthogonal bases $(V_E,V_{E^\bot})$ and $(V_F,V_{F^\bot})$, we can put it in a more compact way, such as in Proposition 4 in \citet{muzellec2019subspace}:
    \begin{equation}
        C = (V_E\Sigma_E+V_{E^\bot}\Sigma_{E^\bot E})T_{E,F}^T(V_F^T+\Lambda_F^{-1}\Lambda_{F^\bot F}^T V_{F^\bot}^T).
    \end{equation}
    
    To check it, just expand the terms and see that $C_{E,F} = V_E C V_F^T$.
\end{proof}

\subsubsection{Proof of \Cref{prop:closed_form_gw_1D}} \label{proof:prop_closed_form_gw_1D}

\begin{proof}[Proof of \Cref{prop:closed_form_gw_1D}]
    Let $\gamma \in \Pi(a,b)$. Then: 
    \begin{equation}
        \begin{split}
            \sum_{ijkl} (x_ix_k-y_jy_l)^{2} \gamma_{ij}\gamma_{kl}=\sum_{ijkl} (x_ix_k)^{2} \gamma_{ij}\gamma_{kl}+\sum_{ijkl} (y_jy_l)^{2} \gamma_{ij}\gamma_{kl}-2\sum_{ijkl} x_ix_ky_jy_l \gamma_{ij}\gamma_{kl}
        \end{split}
    \end{equation}
    However, $\sum_{ijkl} (x_ix_k)^{2} \gamma_{ij}\gamma_{kl}=\sum_{ik} (x_ix_k)^{2} a_ia_k$, and $\sum_{ijkl} (y_jy_l)^{2} \gamma_{ij}\gamma_{kl}=\sum_{jl} (y_jy_l)^{2} b_jb_l$, so this does not depend on $\gamma$. Moreover $2\sum_{ijkl} x_ix_ky_jy_l \gamma_{ij}\gamma_{kl}=2(\sum_{ij} x_iy_j \gamma_{ij})^{2}$. Hence, the problem \eqref{eq:gw1Dinner} is equivalent to $\max_{\gamma \in \Pi(a,b)} (\sum_{ij} x_iy_j \gamma_{ij})^{2}$ (in terms of the OT plan), which is also equivalent to solving $\max_{\gamma \in \Pi(a,b)} |\sum_{ij} x_iy_j \gamma_{ij}|$ or equivalently:
    \begin{equation} \label{eq:thelinearprob}
        \max_{\gamma \in \Pi(a,b)}  \pm 1 \sum_{ij} x_iy_j \gamma_{ij}
    \end{equation}
    We have two cases to consider: If $\pm 1=1$, we have to solve $-\min_{\gamma \in \Pi(a,b)} \sum_{ij} (-x_i)y_j \gamma_{ij}$. Since the points are sorted, the matrix $c_{ij}=-x_iy_j$ satisfies the Monge property~\citep{burkard95perspectives}:
    \begin{equation}
        \forall (i,j) \in \{1,\dots,n-1\} \times \{1,\dots,m-1\}, \ c_{i,j}+c_{i+1,j+1} \leq c_{i+1,j}+c_{i,j+1}
    \end{equation}
    To see this, check that:
    \begin{equation}
        \begin{split}
            &(-x_i)y_j+(-x_{i+1})y_{j+1}-(-x_{i+1})y_j-(-x_{i})y_{j+1}\\
            &=(-x_{i})(y_{j}-y_{j+1})+(-x_{i+1})(y_{j+1}-y_j)=(y_{j}-y_{j+1})(x_{i+1}-x_{i})\leq 0
        \end{split}
    \end{equation}
    In this case, the North-West corner rule $NW(a,b)$ defined in \Cref{alg:nw} is known to produce an optimal solution to the linear problem \eqref{eq:thelinearprob}  \citep{burkard95perspectives}. If~$\pm=-1$, then changing $x_i$ to $-x_i$ concludes.
\end{proof}

\subsection{Proofs of \Cref{section:triangular_coupling}}

\subsubsection{Proof of \Cref{prop_hadamard}} \label{proof:prop_hadamard}

\begin{proof}[Proof of \Cref{prop_hadamard}]
    Let $\mu,\nu\in\mathcal{P}(\mathbb{R}^d)$,
    \begin{enumerate}
        \item $(x,x')\mapsto x\odot x'$ is a continuous map, therefore, $L$ is less semi-continuous. Hence, by~applying Lemma 2.2.1 of \citep{vayer2020contribution}, we observe that $\gamma\mapsto \iint L(x,x',y,y')\mathrm{d}\gamma(x,y)\mathrm{d}\gamma(x',y')$ is less semi-continuous for the weak convergence of~measures.
        
        Now, as~$\Pi(\mu,\nu)$ is a compact set (see the proof of Theorem 1.7 in~\citep{santambrogio2015optimal} for the Polish space case and~of Theorem 1.4 for the compact metric space) and~$\gamma\mapsto \iint L\mathrm{d}\gamma\mathrm{d}\gamma$ is less semi-continuous for the weak convergence, we can apply the Weierstrass theorem \citep[Memo 2.2.1]{vayer2020contribution}, which states that \eqref{HadamardWasserstein} always admits a minimizer.
        \item See \citep[Theorem 16]{chowdhury2019gromov}.
        \item For invariances, we first look at the properties that must be satisfied by $T$ in order to have: $\forall x,x',\ f(x,x')=f(T(x),T(x'))$ where $f:(x,x')\mapsto x\odot x'$.
        
        We find that $\forall x\in\mathbb{R}^d,\ \forall 1\le i\le d,\ |[T(x)]_i|=|x_i|$ because, denoting $(e_i)_{i=1}^d$ as the canonical basis, we have:
        \begin{equation}
            x\odot e_i = T(x)\odot T(e_i),
        \end{equation}
        which implies that:
        \begin{equation}
            x_i = [T(x)]_i [T(e_i)]_i.
        \end{equation}
        However, $f(e_i,e_i)=f(T(e_i),T(e_i))$ implies $[T(e_i)]_i^2 = 1$, and~therefore:
        \begin{equation}
            |[T(x)]_i| = |x_i|.
        \end{equation}
        If we take for $T$ the reflection with respect to axes, then it satisfies $f(x,x')=f(T(x),T(x'))$ well.
        Moreover, it is a good equivalence relation, and~therefore, we have a distance on the quotient space.
    \end{enumerate}
\end{proof}

\subsubsection{Proof of \Cref{ThGWKR}} \label{proof:thgwkr}

We first recall a useful~theorem.

\begin{theorem}[Theorem 2.8 in \citet{billingsley2013convergence}] \label{billingsley}
    Let $\Omega=X\times Y$ be a separable space, and~let $P,P_n\in\mathcal{P}(\Omega)$ with marginals $P_X$ (respectively $P_{n,X}$) and $P_Y$ (respectively $P_{n,Y}$). Then, $P_{n,X}\otimes P_{n,Y} \xrightarrow[]{\mathcal{L}}P$ if and only if $P_{n,X}\xrightarrow[]{\mathcal{L}}P_X$, $P_{n,Y}\xrightarrow[]{\mathcal{L}}P_Y$ and $P=P_X\otimes P_Y$. 
\end{theorem}

\begin{proof}[Proof of \Cref{ThGWKR}]
    The following proof is mainly inspired by the proof of \Cref{theorem:KnotheToBrenier} in \citep[Theorem 2.1]{carlier2010knothe}, \citep[Theorem 3.1.6]{bonnotte2013unidimensional} and \citep[Theorem 2.23]{santambrogio2015optimal}.
    
    Let $\mu,\nu\in\mathcal{P}(\mathbb{R}^d)$, absolutely continuous, with~finite fourth moments and compact supports. We recall the problem $\mathcal{HW}_t$:
    \begin{equation}
        \mathcal{HW}_t^2(\mu,\nu) = \inf_{\gamma\in\Pi(\mu,\nu)}\ \iint \sum_{k=1}^d \Big( \prod_{i=1}^{k-1} \lambda_t^{(i)}\Big)\ (x_kx_k'-y_ky_k')^2\ \mathrm{d}\gamma_t(x,y)\mathrm{d}\gamma_t(x',y'),
    \end{equation}
    with $\forall t>0,\ \forall i\in\{1,\dots,d-1\}$, $\lambda_t^{(i)}>0$ and $\lambda_t^{(i)}\xrightarrow[t\to 0]{}0$.
    
    First, let us denote $\gamma_t$ the optimal coupling for $\mathcal{HW}_t$ for all $t>0$. We want to show that $\gamma_t\xrightarrow[t\rightarrow 0]{\mathcal{L}}\gamma_K$ with $\gamma_K=(Id\times T_K)_\#\mu$ and $T_K$ our alternate Knothe-Rosenblatt rearrangement. Let $\gamma\in\Pi(\mu,\nu)$ such that $\gamma_t\xrightarrow[t\rightarrow 0]{\mathcal{L}}\gamma$ (true up to subsequence as $\{\mu\}$ and $\{\nu\}$ are tight in $\mathcal{P}(X)$ and $\mathcal{P}(Y)$ if $X$ and $Y$ are polish space, therefore, by~\citep[Lemma 4.4]{villani2009optimal}, $\Pi(\mu,\nu)$ is a tight set, and~we can apply the Prokhorov theorem \citep[Box 1.4]{santambrogio2015optimal} on $(\gamma_t)_t$ and extract a subsequence)).
    
    \proofpart{1}{}
    
    First, let us notice that:
    \begin{equation}
        \begin{aligned}
            \mathcal{HW}_t^2(\mu,\nu) &= \iint \sum_{k=1}^d\Big( \prod_{i=1}^{k-1} \lambda_t^{(i)}\Big)\ (x_kx_k'-y_ky_k')^2\ \mathrm{d}\gamma_t(x,y)\mathrm{d}\gamma_t(x',y') \\
            &= \iint (x_1x_1'-y_1y_1')^2\ \mathrm{d}\gamma_t(x,y)\mathrm{d}\gamma_t(x',y') \\
            &+ \iint \sum_{k=2}^d\Big( \prod_{i=1}^{k-1} \lambda_t^{(i)}\Big)\ (x_kx_k'-y_ky_k')^2\ \mathrm{d}\gamma_t(x,y)\mathrm{d}\gamma_t(x',y'). \\
        \end{aligned}
    \end{equation}
    Moreover, as~$\gamma_t$ is the optimal coupling between $\mu$ and $\nu$, and~$\gamma_K\in\Pi(\mu,\nu)$,
    \begin{equation}
        \begin{aligned}
            \mathcal{HW}_t^2(\mu,\nu) &\le \iint \sum_{k=1}^d\Big( \prod_{i=1}^{k-1} \lambda_t^{(i)}\Big)\ (x_kx_k'-y_ky_k')^2\ \mathrm{d}\gamma_K(x,y)\mathrm{d}\gamma_K(x',y') \\
            &= \iint (x_1x_1'-y_1y_1')^2\ \mathrm{d}\gamma_K(x,y)\mathrm{d}\gamma_K(x',y') \\
            &+ \iint \sum_{k=2}^d\Big( \prod_{i=1}^{k-1} \lambda_t^{(i)}\Big)\ (x_kx_k'-y_ky_k')^2\ \mathrm{d}\gamma_K(x,y)\mathrm{d}\gamma_K(x',y'). \\
        \end{aligned}
    \end{equation}
    
    In our case, we have $\gamma_t\xrightarrow[t\rightarrow 0]{\mathcal{L}}\gamma$, thus, by~Theorem \ref{billingsley}, we have $\gamma_t\otimes\gamma_t\xrightarrow[t\rightarrow0]{\mathcal{L}}\gamma\otimes\gamma$. Using the fact that $\forall i,\ \lambda_t^{(i)}\xrightarrow[t\rightarrow0]{}0$ (and Lemma 1.8 of \citet{santambrogio2015optimal}, since we are on compact support, we can bound the cost (which is continuous) by its max), we obtain the following inequality
    \begin{equation}
        \iint (x_1x_1'-y_1y_1')^2\ \mathrm{d}\gamma(x,y)\mathrm{d}\gamma(x',y') \le \iint (x_1x_1'-y_1y_1')^2\ \mathrm{d}\gamma_K(x,y)\mathrm{d}\gamma_K(x',y').
    \end{equation}
    
    By denoting $\gamma^1$ and $\gamma_K^1$ the marginals on the first variables, we can use the projection $\pi^1(x,y)=(x_1,y_1)$, such as $\gamma^1=\pi^1_\#\gamma$ and $\gamma_K^1 = \pi^1_\#\gamma_K$. Hence, we get
    \begin{equation}
        \iint (x_1x_1'-y_1y_1')^2\ \mathrm{d}\gamma^1(x_1,y_1)\mathrm{d}\gamma^1(x_1',y_1') \le \iint (x_1x_1'-y_1y_1')^2\ \mathrm{d}\gamma^1_K(x_1,y_1)\mathrm{d}\gamma^1_K(x_1',y_1').
    \end{equation}
    However, $\gamma_K^1$ was constructed in order to be the unique optimal map for this cost (either $T_{asc}$ or $T_{desc}$ according to theorem \citep[Theorem 4.2.4]{vayer2020contribution}). Thus, we can deduce that $\gamma^1=(Id\times T_K^1)_\#\mu^1 = \gamma_K^1$.
    
    \proofpart{2}{}
    
    We know that for any $t>0$, $\gamma_t$ and $\gamma_K$ share the same marginals. Thus, as~previously, $\pi^1_\#\gamma_t$ should have a cost worse than $\pi^1_\#\gamma_K$, which translates to
    \begin{equation}
        \begin{aligned}
            \iint (x_1x_1'-y_1y_1')^2\ \mathrm{d}\gamma^1_K(x_1,y_1)\mathrm{d}\gamma^1_K(x_1',y_1') &= \iint (x_1x_1'-y_1y_1')^2\ \mathrm{d}\gamma^1(x_1,y_1)\mathrm{d}\gamma^1(x_1',y_1')\\
            &\le \iint (x_1x_1'-y_1y_1')^2\ \mathrm{d}\gamma^1_t(x_1,y_1)\mathrm{d}\gamma^1_t(x_1',y_1').
        \end{aligned}
    \end{equation}
    Therefore, we have the following inequality,
\begingroup\makeatletter\def\f@size{9.5}\check@mathfonts
\def\maketag@@@#1{\hbox{\m@th\normalsize\normalfont#1}}%
    \begin{equation}
        \begin{aligned}
            &\iint (x_1x_1'-y_1y_1')^2\ \mathrm{d}\gamma^1(x,y)\mathrm{d}\gamma^1(x',y') + \iint \sum_{k=2}^d\Big( \prod_{i=1}^{k-1} \lambda_t^{(i)}\Big)\ (x_kx_k'-y_ky_k')^2\ \mathrm{d}\gamma_t(x,y)\mathrm{d}\gamma_t(x',y') \\
            &\le \mathcal{HW}_t^2(\mu,\nu) \\
            &\le \iint (x_1x_1'-y_1y_1')^2\ \mathrm{d}\gamma^1(x,y)\mathrm{d}\gamma^1(x',y') \\ &+ \iint \sum_{k=2}^d\Big( \prod_{i=1}^{k-1} \lambda_t^{(i)}\Big)\ (x_kx_k'-y_ky_k')^2\ \mathrm{d}\gamma_K(x,y)\mathrm{d}\gamma_K(x',y').
        \end{aligned}
    \end{equation}\endgroup
    We can substract the first term and factorize by $\lambda_t^{(1)}>0$,
    \begin{equation}
        \begin{aligned}
            &\iint \sum_{k=2}^d\Big( \prod_{i=1}^{k-1} \lambda_t^{(i)}\Big)\ (x_kx_k'-y_ky_k')^2\ \mathrm{d}\gamma_t(x,y)\mathrm{d}\gamma_t(x',y') \\
            &= \lambda_t^{(1)}\Big( \iint (x_2x_2'-y_2y_2')^2\ \mathrm{d}\gamma_t(x,y)\mathrm{d}\gamma_t(x',y') \\ &+\iint \sum_{k=3}^d\Big( \prod_{i=2}^{k-1} \lambda_t^{(i)}\Big)\ (x_kx_k'-y_ky_k')^2\ \mathrm{d}\gamma_t(x,y)\mathrm{d}\gamma_t(x',y') \Big) \\
            &\le \lambda_t^{(1)}\Big( \iint (x_2x_2'-y_2y_2')^2\ \mathrm{d}\gamma_K(x,y)\mathrm{d}\gamma_K(x',y') \\ &+\iint \sum_{k=3}^d\Big( \prod_{i=2}^{k-1} \lambda_t^{(i)}\Big)\ (x_kx_k'-y_ky_k')^2\ \mathrm{d}\gamma_K(x,y)\mathrm{d}\gamma_K(x',y') \Big).
        \end{aligned}
    \end{equation}
    By dividing by $\lambda_t^{(1)}$ and by taking the limit $t\rightarrow 0$ as in the first part, we get
\begin{equation} \label{inequalityBeforeDisintegration}
        \iint (x_2x_2'-y_2y_2')^2\ \mathrm{d}\gamma(x,y)\mathrm{d}\gamma(x',y') \le \iint (x_2x_2'-y_2y_2')^2\ \mathrm{d}\gamma_K(x,y)\mathrm{d}\gamma_K(x',y').
    \end{equation}
    
    Now, the~2 terms depend only on $(x_2,y_2)$ and $(x_2',y_2')$. We will project on the two first coordinates, i.e.,~let $\pi^{1,2}(x,y)=((x_1,x_2),(y_1,y_2))$ and $\gamma^{1,2}=\pi^{1,2}_\#\gamma$, $\gamma^{1,2}_K=\pi^{1,2}_\#\gamma_K$. Using the disintegration of measures, we know that there exist kernels $\gamma^{2|1}$ and $\gamma^{2|1}_K$ such that $\gamma^{1,2}=\gamma^1 \otimes \gamma^{2|1}$ and $\gamma^{1,2}_K = \gamma_K^1 \otimes \gamma^{2|1}_K$, where
    \begin{equation}
        \forall A\in\mathcal{B}(X\times Y),\ \mu\otimes K(A) = \iint \mathbb{1}_A(x,y) K(x,\mathrm{d}y)\mu(\mathrm{d}x).
    \end{equation}
    We can rewrite the previous Equation~(\ref{inequalityBeforeDisintegration}) as
\begin{equation} \label{inequality_1}
        \begin{aligned}
            &\iint (x_2x_2'-y_2y_2')^2\ \mathrm{d}\gamma(x,y)\mathrm{d}\gamma(x',y')  \\
            &= \iiiint (x_2x_2'-y_2y_2')^2\ \gamma^{2|1}((x_1,y_1),(\mathrm{d}x_2,\mathrm{d}y_2))\gamma^{2|1}((x'_1,y'_1),(\mathrm{d}x'_2,\mathrm{d}y'_2)) \\ &\mathrm{d}\gamma^1(x_1,y_1)\mathrm{d}\gamma^1(x_1',y_1') \\
            &\le \iiiint (x_2x_2'-y_2y_2')^2\ \gamma_K^{2|1}((x_1,y_1),(\mathrm{d}x_2,\mathrm{d}y_2))\gamma_K^{2|1}((x'_1,y'_1),(\mathrm{d}x'_2,\mathrm{d}y'_2)) \\ &\mathrm{d}\gamma_K^1(x_1,y_1)\mathrm{d}\gamma_K^1(x_1',y_1').
        \end{aligned}
    \end{equation}
    
    Now, we will assume at first that the marginals of $\gamma^{2|1}((x_1,y_1),\cdot)$ are well $\mu^{2|1}(x_1,\cdot)$ and $\nu^{2|1}(y_1,\cdot)$. Then, by~definition of $\gamma_K^{2|1}$, as~it is optimal for the $GW$ cost with inner products, we have for all $(x_1,y_1), (x_1',y_1')$,
\begin{equation} \label{inequality_KR}
        \begin{aligned}
            &\iint (x_2x_2'-y_2y_2')^2\ \gamma_K^{2|1}((x_1,y_1),(\mathrm{d}x_2,\mathrm{d}y_2))\gamma_K^{2|1}((x'_1,y'_1),(\mathrm{d}x'_2,\mathrm{d}y'_2)) \\ 
            &\le \iint (x_2x_2'-y_2y_2')^2\ \gamma^{2|1}((x_1,y_1),(\mathrm{d}x_2,\mathrm{d}y_2))\gamma^{2|1}((x'_1,y'_1),(\mathrm{d}x'_2,\mathrm{d}y'_2)).
        \end{aligned}
    \end{equation}
    Moreover, we know from the first part that $\gamma^1=\gamma_K^1$, then by integrating with respect to $(x_1,y_1)$ and $(x_1',y_1')$, we have
\begin{equation} \label{inequality_2}
        \begin{aligned}
            &\iiiint (x_2x_2'-y_2y_2')^2\ \gamma_K^{2|1}((x_1,y_1),(\mathrm{d}x_2,\mathrm{d}y_2))\gamma_K^{2|1}((x'_1,y'_1),(\mathrm{d}x'_2,\mathrm{d}y'_2))\\ &\mathrm{d}\gamma^1(x_1,y_1)\mathrm{d}\gamma^1(x_1',y_1')  \\ 
            &\le \iiiint (x_2x_2'-y_2y_2')^2\ \gamma^{2|1}((x_1,y_1),(\mathrm{d}x_2,\mathrm{d}y_2))\gamma^{2|1}((x'_1,y'_1),(\mathrm{d}x'_2,\mathrm{d}y'_2))\\ &\mathrm{d}\gamma^1(x_1,y_1)\mathrm{d}\gamma^1(x_1',y_1'). 
        \end{aligned}
    \end{equation}
    By \eqref{inequality_1} and \eqref{inequality_2}, we deduce that we have an equality and we get
\begin{equation} \label{equality}
        \begin{aligned}
            &\iint\Big(\iint (x_2x_2'-y_2y_2')^2\ \gamma^{2|1}((x_1,y_1),(\mathrm{d}x_2,\mathrm{d}y_2))\gamma^{2|1}((x'_1,y'_1),(\mathrm{d}x'_2,\mathrm{d}y'_2)) \\
            &- \iint(x_2x_2'-y_2y_2')^2\ \gamma^{2|1}_K((x_1,y_1),(\mathrm{d}x_2,\mathrm{d}y_2))\gamma^{2|1}_K((x'_1,y'_1),(\mathrm{d}x'_2,\mathrm{d}y'_2)) \Big)\\ &\mathrm{d}\gamma^1(x_1,y_1)\mathrm{d}\gamma^1(x_1',y_1') = 0.
        \end{aligned}
    \end{equation}
    However, we know by \eqref{inequality_KR} that the middle part of \eqref{equality} is nonnegative, thus we have for $\gamma^1$-a.e. $(x_1,y_1),(x_1',y_1')$, 
    \begin{equation}
        \begin{aligned}
            &\iint(x_2x_2'-y_2y_2')^2\ \gamma^{2|1}_K((x_1,y_1),(\mathrm{d}x_2,\mathrm{d}y_2))\gamma^{2|1}_K((x'_1,y'_1),(\mathrm{d}x'_2,\mathrm{d}y'_2)) \\ 
            &= \iint(x_2x_2'-y_2y_2')^2\ \gamma^{2|1}((x_1,y_1),(\mathrm{d}x_2,\mathrm{d}y_2))\gamma^{2|1}((x'_1,y'_1),(\mathrm{d}x'_2,\mathrm{d}y'_2)).
        \end{aligned}
    \end{equation}
    From that, we can conclude as in the first part that $\gamma^{2|1}=\gamma_K^{2|1}$ (by unicity of the optimal map). And~thus $\gamma^{1,2}=\gamma_K^{1,2}$.
    
    Now, we still have to show that the marginals of $\gamma^{2|1}((x_1,y_1),\cdot)$ and $\gamma_K^{2,1}((x_1,y_1),\cdot)$ are well the same, i.e.,~$\mu^{2|1}(x_1,\cdot)$ and $\nu^{2|1}(y_1,\cdot)$. Let $\phi$ and $\psi$ be continuous functions, then we have to show that for $\gamma^1$-a.e. $(x_1,y_1)$, we have
    \begin{equation}
        \begin{cases}
            \int \phi(x_2)\gamma^{2|1}((x_1,y_1),(\mathrm{d}x_2,\mathrm{d}y_2)) = \int \phi(x_2)\mu^{2|1}(x_1,\mathrm{d}x_2) \\
            \int \psi(y_2)\gamma^{2|1}((x_1,y_1),(\mathrm{d}x_2,\mathrm{d}y_2)) = \int \psi(y_2)\nu^{2|1}(y_1,\mathrm{d}y_2).
        \end{cases}
    \end{equation}
    As we want to prove it for $\gamma^1$-a.e. $(x_1,y_1)$, it is sufficient to prove that for all continuous function $\xi$,
    \begin{equation}
        \begin{cases}
            \iint \xi(x_1,y_1)\phi(x_2)\gamma^{2|1}((x_1,y_1),(\mathrm{d}x_2,\mathrm{d}y_2)) \mathrm{d}\gamma^1(x_1,y_1) \\ = \iint \xi(x_1,y_1) \phi(x_2)\mu^{2|1}(x_1,\mathrm{d}x_2)\mathrm{d}\gamma^1(x_1,y_1)  \\
            \iint \xi(x_1,y_1) \psi(y_2)\gamma^{2|1}((x_1,y_1),(\mathrm{d}x_2,\mathrm{d}y_2)) \mathrm{d}\gamma^1(x_1,y_1) \\ = \iint \xi(x_1,y_1) \psi(y_2)\nu^{2|1}(y_1,\mathrm{d}y_2) \mathrm{d}\gamma^1(x_1,y_1).
        \end{cases}
    \end{equation}
    
    First, we can use the projections $\pi_x(x,y)=x$ and $\pi_y(x,y)=y$. Moreover, we know that $\gamma^1 = (Id\times T_K^1)_\#\mu^1$. The~alternate Knothe--Rosenblatt rearrangement is, as~the usual one, bijective (because $\mu$ and $\nu$ are absolutely continuous), and~thus, as~we suppose that $\nu$ satisfies the same hypothesis than $\mu$, we also have $\gamma^1 = ((T_K^1)^{-1},Id)_\#\nu^1$. Let us note $\Tilde{T}_K^1=(T_K^1)^{-1}$. Then, the~equalities that we want to show are:
    \begin{equation}
        \begin{cases}
            \iint \xi(x_1,T_K^1(x_1))\phi(x_2)\gamma^{2|1}_x((x_1,T_K^1(x_1)),\mathrm{d}x_2) \mathrm{d}\mu^1(x_1) \\ = \iint \xi(x_1,T_K^1(x_1)) \phi(x_2)\mu^{2|1}(x_1,\mathrm{d}x_2)\mathrm{d}\mu^1(x_1)  \\
            \iint \xi(\Tilde{T}_K^1(y_1),y_1) \psi(y_2)\gamma_y^{2|1}((\Tilde{T}_K^1(y_1),y_1),\mathrm{d}y_2) \mathrm{d}\nu^1(y_1) \\ = \iint \xi(\Tilde{T}_K^1(y_1),y_1) \psi(y_2)\nu^{2|1}(y_1,\mathrm{d}y_2) \mathrm{d}\nu^1(y_1).
        \end{cases}
    \end{equation}
    In addition, we have indeed
    \begin{equation}
        \begin{aligned}
             &\iint \xi(x_1,T_K^1(x_1))\phi(x_2)\gamma^{2|1}_x((x_1,T_K^1(x_1)),\mathrm{d}x_2) \mathrm{d}\mu^1(x_1) \\ &= \iint \xi(x_1,T_K^1(x_1))\phi(x_2) \mathrm{d}\gamma^{1,2}((x_1,x_2),(y_1,y_2)) \\
             &= \iint \xi(x_1,T_K^1(x_1))\phi(x_2) \mathrm{d}\gamma_x^{1,2}(x_1,x_2) \\
             &= \iint \xi(x_1,T_K^1(x_1))\phi(x_2) \mu^{2|1}(x_1,\mathrm{d}x_2)\mathrm{d}\mu^1(x_1).
        \end{aligned}
    \end{equation}
    We can do the same for the $\nu$ part by~symmetry.
    
    \proofpart{3}{}
    
    Now, we can proceed the same way by induction. Let $\ell\in\{2,\dots,d\}$ and suppose that the result is true in dimension $\ell-1$ (i.e.,~$\gamma^{1:\ell-1}=\pi^{1:\ell-1}_\#\gamma = \gamma_K^{1:\ell-1}$).
    
    For this part of the proof, we rely on \citep[Theorem 2.23]{santambrogio2015optimal}. We can build a measure $\gamma_K^t\in\mathcal{P}(\mathbb{R}^d\times\mathbb{R}^d)$ such that:
\begin{equation} \label{conditions}
        \begin{cases}
            \pi^x_\#\gamma_K^t = \mu \\
            \pi^y_\#\gamma_K^t = \nu \\
            \pi^{1:\ell-1}_\#\gamma_K^t = \eta_{t,\ell}
        \end{cases}
    \end{equation}
    where $\eta_{t,\ell}$ is the optimal transport plan between $\mu^\ell=\pi^{1:\ell-1}_\#\mu$ and $\nu^\ell=\pi^{1:\ell-1}_\#\nu$ for the objective:
    \begin{equation}
        \iint \sum_{k=1}^{\ell-1}\Big( \prod_{i=1}^{k-1} \lambda_t^{(i)}\Big) (x_kx_k'-y_ky_k')^2\ \mathrm{d}\gamma(x,y)\mathrm{d}\gamma(x',y').
    \end{equation}
    By induction hypothesis, we have $\eta_{t,\ell}\xrightarrow[t\to 0]{\mathcal{L}} \pi^{1:\ell-1}_\#\gamma_K$. To~build such a measure, we can first disintegrate $\mu$ and $\nu$:
    \begin{equation}
        \begin{cases}
            \mu = \mu^{1:\ell-1}\otimes \mu^{\ell:d|1:\ell-1} \\
            \nu = \nu^{1:\ell-1}\otimes \nu^{\ell:d|1:\ell-1},
        \end{cases}
    \end{equation}
    then we pick the Knothe transport $\gamma_K^{\ell:d|1:\ell-1}$ between $\mu^{\ell:d|1:\ell-1}$ and $\nu^{\ell:d|1:\ell-1}$. Thus, by~taking $\gamma_K^T = \eta_{t,\ell}\otimes \gamma_K^{\ell:d|1:\ell-1}$, $\gamma_K^T$ satisfies the conditions well (\ref{conditions}).
    
    Hence, we have:
    \begin{equation}
        \begin{aligned}
            &\iint \sum_{k=1}^{\ell-1}\Big( \prod_{i=1}^{k-1} \lambda_t^{(i)}\Big) (x_kx_k'-y_ky_k')^2\ \mathrm{d}\gamma_K^t(x,y)\mathrm{d}\gamma_K^t(x',y') \\ &= \iint \sum_{k=1}^{\ell-1}\Big( \prod_{i=1}^{k-1} \lambda_t^{(i)}\Big) (x_kx_k'-y_ky_k')^2\ \mathrm{d}\eta_{t,\ell}(x_{1:\ell-1},y_{1:\ell-1})\mathrm{d}\eta_{t,\ell}(x'_{1:\ell-1},y'_{1:\ell-1}) \\
            &\leq \iint \sum_{k=1}^{\ell-1}\Big( \prod_{i=1}^{k-1} \lambda_t^{(i)}\Big) (x_kx_k'-y_ky_k')^2\ \mathrm{d}\gamma_t(x,y)\mathrm{d}\gamma_t(x',y'),
        \end{aligned}
    \end{equation}
    and therefore:
    \begin{equation}
        \begin{aligned}
            &\iint\sum_{k=1}^{\ell-1}\Big( \prod_{i=1}^{k-1} \lambda_t^{(i)}\Big) (x_kx_k'-y_ky_k')^2\ \mathrm{d}\gamma_K^t(x,y)\mathrm{d}\gamma_K^t(x',y') \\ &+ \iint \sum_{k=\ell}^d\Big( \prod_{i=1}^{k-1} \lambda_t^{(i)}\Big)\ (x_kx_k'-y_ky_k')^2\ \mathrm{d}\gamma_t(x,y)\mathrm{d}\gamma_t(x',y') \\
            &\le \mathcal{HW}_t^2(\mu,\nu) \\
            &\le \iint \sum_{k=1}^{\ell-1}\Big( \prod_{i=1}^{k-1} \lambda_t^{(i)}\Big) (x_kx_k'-y_ky_k')^2\ \mathrm{d}\gamma_K^t(x,y)\mathrm{d}\gamma_K^t(x',y') \\ &+ \iint \sum_{k=\ell}^d\Big( \prod_{i=1}^{k-1} \lambda_t^{(i)}\Big)\ (x_kx_k'-y_ky_k')^2\ \mathrm{d}\gamma_K^t(x,y)\mathrm{d}\gamma_K^t(x',y'). \\
        \end{aligned}
    \end{equation}
    
    As before, by~subtracting the first term, dividing by $\prod_{i=1}^{\ell-1}\lambda_t^{(i)}$ and taking the limit, we obtain:
    \begin{equation}
        \iint (x_\ell x_\ell'-y_\ell y_\ell')^2\mathrm{d}\gamma_t(x,y)\mathrm{d}\gamma_t(x',y') \leq \iint (x_\ell x_\ell'-y_\ell y_\ell')^2\mathrm{d}\gamma_K^t(x,y)\mathrm{d}\gamma_K^t(x',y').
    \end{equation}
    For the right hand side, using that $\gamma_K^t = \eta_{t,\ell}\otimes \gamma_K^{\ell:d|1:\ell-1}$, we have:
    \begin{equation}
        \begin{aligned}
            &\iint (x_\ell x_\ell'-y_\ell y_\ell')^2\mathrm{d}\gamma_K^t(x,y)\mathrm{d}\gamma_K^t(x',y') \\ &= \iiiint (x_\ell x_\ell'-y_\ell y_\ell')^2\ \gamma_K^{\ell:d|1:\ell-1}((x_{1:\ell-1},y_{1:\ell-1}),(\mathrm{d}x_{\ell:d},\mathrm{d}y_{\ell:d}))\\ &\gamma_K^{\ell:d|1:\ell-1}((x_{1:\ell-1}',y_{1:\ell-1}'),(\mathrm{d}x_{\ell:d}',\mathrm{d}y_{\ell:d}'))\mathrm{d}\eta_{t,\ell}(x_{1:\ell-1},y_{1:\ell-1})\mathrm{d}\eta_{t,\ell}(x_{1:\ell-1}',y_{1:\ell-1}') \\
            &= \iiiint (x_\ell x_\ell'-y_\ell y_\ell')^2\ \gamma_K^{\ell|1:\ell-1}((x_{1:\ell-1},y_{1:\ell-1}),(\mathrm{d}x_{\ell},\mathrm{d}y_{\ell})) \\ &\gamma_K^{\ell|1:\ell-1}((x_{1:\ell-1}',y_{1:\ell-1}'),(\mathrm{d}x_{\ell}',\mathrm{d}y_{\ell}'))\mathrm{d}\eta_{t,\ell}(x_{1:\ell-1},y_{1:\ell-1})\mathrm{d}\eta_{t,\ell}(x_{1:\ell-1}',y_{1:\ell-1}').
        \end{aligned}
    \end{equation}
    Let us note for $\eta_{t,\ell}$ almost every $(x_{1:\ell-1},y_{1:\ell-1}),(x_{1:\ell-1}',y_{1:\ell-1}')$
\begingroup\makeatletter\def\f@size{9.8}\check@mathfonts
\def\maketag@@@#1{\hbox{\m@th\normalsize\normalfont#1}}%
    \begin{equation}
        \begin{aligned}
            &GW(\mu^{\ell|1:\ell-1}, \nu^{\ell|1:\ell-1}) \\ &=\iint (x_\ell x_\ell'-y_\ell y_\ell')^2\gamma_K^{\ell|1:\ell-1}((x_{1:\ell-1},y_{1:\ell-1}),(\mathrm{d}x_{\ell},\mathrm{d}y_{\ell}))\gamma_K^{\ell|1:\ell-1}((x_{1:\ell-1}',y_{1:\ell-1}'),(\mathrm{d}x_{\ell}',\mathrm{d}y_{\ell}')),
        \end{aligned}
    \end{equation}\endgroup
    then
    \begin{equation}
        \begin{aligned}
            &\iint (x_\ell x_\ell'-y_\ell y_\ell')^2\mathrm{d}\gamma_K^t(x,y)\mathrm{d}\gamma_K^t(x',y') \\&= \iint GW(\mu^{\ell|1:\ell-1}, \nu^{\ell|1:\ell-1}) \mathrm{d}\eta_{t,\ell}(x_{1:\ell-1},y_{1:\ell-1})\mathrm{d}\eta_{t,\ell}(x_{1:\ell-1}',y_{1:\ell-1}').
        \end{aligned}
    \end{equation}
    By Theorem \ref{billingsley}, we have $\eta_{t,\ell}\otimes \eta_{t,\ell}\xrightarrow[t\to 0]{\mathcal{L}} \pi^{1:\ell-1}_\#\gamma_K \otimes \pi^{1:\ell-1}_\#\gamma_K$. So, if~\begin{equation}
        \eta\mapsto \iint GW(\mu^{\ell|1:\ell-1}, \nu^{\ell|1:\ell-1})\mathrm{d}\eta\mathrm{d}\eta
    \end{equation}
    is continuous over the transport plans between $\mu^{1:\ell-1}$ and $\nu^{1:\ell-1}$, we have
    \begin{equation}
        \begin{aligned}
            &\iint (x_\ell x_\ell'-y_\ell y_\ell')^2\mathrm{d}\gamma_K^t(x,y)\mathrm{d}\gamma_K^t(x',y') \\
            \ &\xrightarrow[t\to 0]{} \iint GW(\mu^{\ell|1:\ell-1}, \nu^{\ell|1:\ell-1}) \pi^{1:\ell-1}_\#\gamma_K(\mathrm{d}x_{1:\ell-1},\mathrm{d}y_{1:\ell-1}) \pi^{1:\ell-1}_\#\gamma_K(\mathrm{d}x_{1:\ell-1}',\mathrm{d}y_{1:\ell-1}')
        \end{aligned}
    \end{equation}
    and
    \begin{equation}
        \begin{aligned}
            &\iint GW(\mu^{\ell|1:\ell-1}, \nu^{\ell|1:\ell-1}) \pi^{1:\ell-1}_\#\gamma_K(\mathrm{d}x_{1:\ell-1},\mathrm{d}y_{1:\ell-1}) \pi^{1:\ell-1}_\#\gamma_K(\mathrm{d}x_{1:\ell-1}',\mathrm{d}y_{1:\ell-1}') \\
            &= \iint (x_\ell x_\ell'-y_\ell y_\ell')^2 \mathrm{d}\gamma_K(x,y)\mathrm{d}\gamma_K(x',y')
        \end{aligned}
    \end{equation}
    by replacing the true expression of $GW$ and using the disintegration $\gamma_K = (\pi_K^{1:\ell-1})_\#\gamma_K\otimes \gamma_K^{\ell|1:\ell-1}$.
    
    For the continuity, we can apply \citep[Lemma 1.8]{santambrogio2015optimal} (as in the \citep[Corollary 2.24]{santambrogio2015optimal} with $X=Y=\mathbb{R}^{\ell-1}\times\mathbb{R}^{\ell-1}$, $\Tilde{X}=\Tilde{Y}=\mathcal{P}(\Omega)$ with $\Omega\subset \mathbb{R}^{d-\ell+1}\times \mathbb{R}^{d-\ell+1}$ and $c(a,b)=GW(a,b)$, which can be bounded on compact supports by $\max |c|$. Moreover, we use Theorem \ref{billingsley} and the fact that $\eta_t\otimes\eta_t \xrightarrow[t\to0]{\mathcal{L}} \gamma_K^{1:\ell-1}\otimes \gamma_K^{1:\ell-1}$.
    
    By taking the limit $t\to 0$, we now obtain:
    \begin{equation}
        \iint (x_\ell x_\ell'-y_\ell y_\ell')^2\mathrm{d}\gamma(x,y)\mathrm{d}\gamma(x',y') \leq \iint (x_\ell x_\ell'-y_\ell y_\ell')^2\mathrm{d}\gamma_K(x,y)\mathrm{d}\gamma_K(x',y').
    \end{equation}
    
    We can now disintegrate with respect to $\gamma^{1:\ell-1}$ as before. We just need to prove that the marginals coincide, which is performed by taking for test functions:
    \begin{equation}
        \begin{cases}
            \xi(x_1,\dots,x_{\ell-1},y_1,\dots,y_{\ell-1})\phi(x_\ell) \\
            \xi(x_1,\dots,x_{\ell-1},y_1,\dots,y_{\ell-1})\psi(y_\ell)
        \end{cases}
    \end{equation}
    and using the fact that the measures are concentrated on $y_k=T_K(x_k)$.
    
    \proofpart{4}{} 
    
    Therefore, we have well $\gamma_t\xrightarrow[t\to 0]{\mathcal{L}}\gamma_K$.
    Finally, for~the $L^2$ convergence, we have:
    \begin{equation}
        \int \|T_t(x)-T_K(x)\|_2^2\ \mu(\mathrm{d}x) = \int \|y-T_K(x)\|_2^2\ \mathrm{d}\gamma_t(x,y) \to \int \|y-T_K(x)\|_2^2\ \mathrm{d}\gamma_K(x,y) = 0
    \end{equation}
    as $\gamma_t=(Id\times T_t)_\#\mu$ and $\gamma_K=(Id\times T_K)_\#\mu$. Hence, $T_t\xrightarrow[t\to 0]{L^2}T_K$.
\end{proof}

\subsubsection{Proof of \Cref{formulaHW}} \label{proof:formulaHW}

\begin{proof}[Proof of \Cref{formulaHW}]
    First, we can start by writing:
    \begin{equation}
        \begin{aligned}
            \mathcal{L}_{i,j,k,\ell} &= \|x_i\odot x_k-y_j\odot y_\ell\|_2^2 \\
            &= \|X_{i,k}-Y_{j,\ell}\|_2^2 \\
            &= \|X_{i,k}\|_2^2+\|Y_{j,\ell}\|_2^2-2\langle X_{i,k},Y_{j,\ell}\rangle \\
            &= [X^{(2)}]_{i,k}+[Y^{(2)}]_{j,\ell}-2\langle X_{i,k},Y_{j,\ell}\rangle.
        \end{aligned}
    \end{equation}
    We cannot directly apply proposition 1 from \citep{peyre2016gromov} (as the third term is a scalar product), but~by performing the same type of computation, we obtain:
    \begin{equation}
        \mathcal{L}\otimes \gamma = A+B+C 
    \end{equation}
    with
    \begin{equation}
        A_{i,j} = \sum_{k,\ell} [X^{(2)}]_{i,k}\gamma_{k,\ell} = \sum_k [X^{(2)}]_{i,k}\sum_\ell\gamma_{k,\ell} = \sum_k [X^{(2)}]_{i,k}[\gamma\mathbb{1}_m]_{k,1} = [X^{(2)}\gamma\mathbb{1}_m]_{i,1} = [X^{(2)}p]_{i,1}
    \end{equation}
    \begin{equation}
        B_{i,j} = \sum_{k,\ell} [Y^{(2)}]_{j,\ell}\gamma_{k,\ell} = \sum_\ell [Y^{(2)}]_{j,\ell}\sum_k \gamma_{k,\ell} = \sum_\ell [Y^{(2)}]_{j,\ell} [\gamma^T\mathbb{1}_n]_{\ell,1} = [Y^{(2)}\gamma^T\mathbb{1}_n]_{j,1} = [Y^{(2)}q]_{j,1}
    \end{equation}
    and
    \begin{equation}
        \begin{aligned}
            C_{i,j} = -2\sum_{k,\ell}\langle X_{i,k},Y_{j,\ell}\rangle \gamma_{k,\ell} &= -2\sum_{k,\ell}\sum_{t=1}^d X_{i,k,t}Y_{j,\ell,t}\gamma_{k,\ell} \\ 
            &= -2\sum_{t=1}^d \sum_k [X_t]_{i,k}\sum_\ell [Y_t]_{j,\ell}\gamma_{\ell,k}^T \\
            &= -2 \sum_{t=1}^d \sum_k [X_t]_{i,k}[Y_t\gamma^T]_{j,k} \\
            &= -2\sum_{t=1}^d [X_t(Y_t\gamma^T)^T]_{i,j}.
        \end{aligned}
    \end{equation}
    Finally, we have:
    \begin{equation}
        \mathcal{L}\otimes \gamma = X^{(2)}p\mathbb{1}_m^T+\mathbb{1}_n q^T (Y^{(2)})^T-2\sum_{t=1}^d X_t\gamma Y_t^T.
    \end{equation}
\end{proof}

\clearemptydoublepage
\cleartooddpage[\thispagestyle{empty}]
\chapter{Introduction (Français)}
{
    \hypersetup{linkcolor=black} 
    \minitoc 
}

\chaptermark{Introduction}

\looseness=-1 En Machine Learning (ML), l'objectif est d'apprendre le meilleur modèle possible pour une tâche donnée à partir d'un ensemble de données d'entraînement. Les données peuvent avoir différentes structures, de nuages de points en passant par des images ou des graphes, et peuvent reposer dans différents espaces. Une manière pratique de modéliser les données est d'assumer qu'elles suivent une probabilité de distribution sous-jacente inconnue. Ainsi, il est important de développer des outils pour gérer des probabilités de distributions, comme des métriques pour les comparer ou des algorithmes pour les apprendre. De plus, étant donné le nombre de données disponible et leur potentielle grande dimensionnalité, ces méthodes ont besoin d'être capable de passer à l'échelle avec le nombre d'échantillons de données ainsi qu'avec la dimension.

Par exemple, la modélisation générative est une tâche populaire en ML, qui a récemment reçu beaucoup d'attention via les ``Large Language Models'' (LLM) qui ont pour objectif de générer du texte \citep{brown2020language, touvron2023llama, openai2023gpt4}, ou via les modèles de diffusion qui visent à générer des images \citep{rombach2022high, ramesh2022hierarchical, saharia2022photorealistic}. Généralement, l'objectif de ces tâches est d'apprendre la distribution inconnue des données afin de pouvoir générer de nouveaux exemples. Cela revient à minimiser une divergence bien choisie entre des distributions de probabilité. Pour modéliser les distributions de probabilité inconnues, il est commun d'exploiter le Deep Learning en utilisant des réseaux de neurones. Des frameworks populaires incluent les réseaux antagonistes génératifs (GANs) \citep{goodfellow2014generative}, les autoencodeurs variationnels (VAEs) \citep{kingma2013auto}, les flots normalisants (NFs) \citep{papamakarios2021normalizing} ou plus récemment les modèles génératifs basés sur le score \citep{sohl2015deep, song2019generative}.

Des fonctions de perte typiques pour apprendre des distributions de probabilités sont les $f$-divergences \citep{nowozin2016f} (incluant la divergence de Kullback-Leibler par exemple) ou la ``Maximum Mean Discrepancy'' (MMD) \citep{li2017mmd, binkowski2018demystifying, mroueh2021convergence}. Cependant, ces différents objectifs requièrent généralement que les distributions aient des densités, qu'elles partagent le même support et/ou elles ne respectent pas nécessairement bien la géométrie des données \citep{arjovsky2017wasserstein}. Une alternative populaire pour manipuler des distributions de probabilités tout en respectant la géométrie des données à travers un coût spécifié, et pour comparer des distributions qui ne partagent pas forcément le même support est le Transport Optimal (OT) \citep{villani2009optimal}, qui permet de comparer des distributions en trouvant la façon la moins coûteuse de bouger la masse d'une distribution à l'autre. Ainsi, les fonctions de pertes d'OT ont été utilisées comme une autre alternative pour les modèles génératifs à travers par exemple les Wasserstein GANs \citep{arjovsky2017wasserstein} ou les Wasserstein Autoencodeurs \citep{tolstikhin2018wasserstein}.

Cependant, dans sa formulation originale, le Transport Optimal souffre d'un gros coût computationnel ainsi que de la malédiction de la dimension, ce qui peut réduire son utilité pour des applications de ML. Ainsi, cette thèse se concentre sur le développement et l'analyse de méthodes efficaces de Transport Optimal avec pour objectif de les appliquer sur des problèmes de ML.

\section{Transport Optimal pour le Machine Learning}

Le Transport Optimal \citep{villani2009optimal} fournit une façon de comparer des distributions de probabilités tout en prenant en compte leur géométrie sous-jacente. Ce problème, d'abord introduit par \citet{monge1781memoire}, consiste originellement à trouver la meilleure façon de bouger une distribution de probabilité sur une autre par rapport à une certaine fonction de coût. Cela fournit deux quantités d'intérêt. La première est la fonction de transport optimal (et plus généralement le plan de transport optimal) qui permet de pousser une distribution sur une distribution cible. La seconde est la valeur optimale du problème sous-jacent, qui quantifie à quel point deux distributions de probabilités sont proches et définit une distance entre elles appelée généralement la distance de Wasserstein.

Le problème de Transport Optimal a reçu beaucoup d'attention récemment. La fonction de transport optimal, aussi appelée la fonction de Monge, peut être utilisée dans plusieurs problèmes comme l'adaptation de domaine \citep{courty2016optimal}, où l'objectif est de classifier des données d'une distribution de probabilité cible pour laquelle nous n'avons pas accès à des exemples d'entraînement grâce à un autre jeu de données que l'on utilise comme ensemble d'entraînement. Ainsi, la fonction de transport optimal permet d'aligner le jeu de données source avec le jeu de données cible, ce qui permet ensuite d'utiliser un classifieur appris sur le jeu de données source. La fonction de transport a aussi été utile pour la traduction, où l'on veut aligner deux embeddings de différents langages \citep{grave2019unsupervised}, pour la biologie computationnelle \citep{schiebinger2019optimal}, en vision par ordinateur \citep{makkuva2020optimal} ou pour des applications physiques comme la cosmologie \citep{nikakhtar2022optimal, panda2023semi}.

Dans cette thèse, nous allons surtout être intéressés par les propriétés de distance du problème de Transport Optimal. Comme il fournit un bon moyen de comparer des distributions de probabilité, il a été utilisé, par exemple, pour classifier des documents qui peuvent être vus comme des distributions de probabilités sur des mots \citep{kusner2015word, huang2016supervised}, pour faire de la réduction de dimension de jeux de données d'histogrammes ou plus généralement de distributions de probabilités en utilisant une analyse en composante principale (ACP) \citep{seguy2015principal, bigot2017geodesic, cazelles2018geodesic} ou de l'apprentissage de dictionnaires \citep{rolet2016fast, schmitz2018wasserstein, mueller2022geometric}, ou pour faire du clustering \citep{cuturi2014fast} avec l'algorithme de Wasserstein K-Means \citep{domazakis2019clustering, zhuang2022wasserstein}.
Il fournit aussi des fonctions de perte efficaces pour des problèmes d'apprentissage supervisés \citep{frogner2015learning} ou pour des tâches de modèles génératifs avec les Wasserstein GANs \citep{arjovsky2017wasserstein, gulrajani2017improved, genevay2017gan} ou les Wasserstein Autoencoders \citep{tolstikhin2018wasserstein}. Le coût de transport optimal a aussi été utilisé pour obtenir des trajectoires de flots plus droites, permettant de réaliser une meilleure inférence de manière plus rapide \citep{finlay2020train, onken2021ot, tong2023conditional}. De plus, l'espace des probabilités de distribution muni de la distance de Wasserstein a une structure géodésique \citep{otto2001geometry}, qui permet de dériver une théorie complète de flots gradients \citep{ambrosio2008gradient}. Cela a mené à la dérivation de plusieurs algorithmes qui apportent des moyens efficaces de minimiser des fonctionnelles sur l'espace des mesures de probabilités \citep{arbel2019maximum, salim2020wasserstein, glaser2021kale, altekruger2023neural} et qui sont reliés avec des algorithmes d'échantillonnage dérivés par exemple dans la communauté des Markov chain Monte-Carlo (MCMC) \citep{jordan1998variational, wibisono2018sampling}.

\section{Motivations}

En Machine Learning, nous sommes souvent amenés à manipuler des problèmes avec de grandes quantités de données. Dans ce cas, un des inconvénients du problème de Transport Optimal est sa complexité computationnelle par rapport au nombre d'échantillons pour calculer la distance de transport optimal. Pour réduire ce coût computationnel, différentes solutions ont été proposées ces dernières années qui ont rendu le Transport Optimal très populaire en ML.

\paragraph{Alternatives au problème original de Transport Optimal.}

\citet{cuturi2013sinkhorn} a proposé d'ajouter une régularisation entropique au problème de transport optimal, ce qui permet de dériver un algorithme avec une meilleure complexité computationnelle et qui peut être utilisé sur des GPUs \citep{feydy2020geometric}, ce qui a permis de populariser le transport optimal dans la communauté de ML \citep{torres2021survey}. Cet objectif a notamment été utilisé pour la modélisation générative en utilisant l'auto-différentiation \citep{genevay2018learning}. Pour des problèmes d'apprentissage où l'objectif est d'apprendre implicitement la distribution des données, une autre alternative beaucoup utilisée en Deep Learning est l'approche par minibatch \citep{genevay2016stochastic, fatras2020learning, fatras2021minibatch} qui n'utilise à chaque étape qu'une petite portion des données. Une autre famille d'approches utilise des alternatives à la formulation classique du problème de transport optimal en considérant des projections sur des sous-espaces. Ces approches peuvent être motivées d'une part par le fait que des distributions dans des espaces de grande dimension sont souvent supposées être supportées sur un sous-espace de faible dimension, ou que deux distributions ne diffèrent que sur sous-espace de faible dimension \citep{niles2022estimation}. D'autre part, ces approches peuvent être calculées plus efficacement que le problème de transport optimal classique tout en conservant certaines de ses propriétés et en offrant souvent de meilleures propriétés statistiques en grande dimension. Dans cette thèse, nous allons surtout nous intéresser à des méthodes qui reposent sur des projections sur des sous-espaces.

\paragraph{Sliced-Wasserstein.} L'exemple principal de ce genre de méthodes est la distance de Sliced-Wasserstein (SW) \citep{rabin2012wasserstein, bonnotte2013unidimensional, bonneel2015sliced}, qui est définie comme la moyenne de la distance de Wasserstein entre les projections unidimensionnelles des mesures sur toutes les directions. Cette distance a beaucoup de bonnes propriétés, notamment celle d'avoir un faible coût computationnel. Elle s'est avérée être une alternative appropriée à la distance de Wasserstein ou au problème de transport avec régularisation entropique. Comme il s'agit d'une fonction de perte différentiable, elle a été utilisée dans de nombreux problèmes d'apprentissage tels que la modélisation générative pour apprendre l'espace latent des autoencodeurs avec les autoencodeurs Sliced-Wasserstein \citep{kolouri2018sliced}, pour apprendre des générateurs avec les générateurs Sliced-Wasserstein \citep{deshpande2018generative, wu2019sliced, lezama2021run}, pour entraîner des flots normalisants \citep{coeurdoux2022sliced, coeurdoux2023learning}, pour de l'inférence variationnelle \citep{yi2023sliced}, ou comme un objectif pour des algorithmes non paramétriques \citep{liutkus2019sliced, dai2021sliced, du2023nonparametric}. Elle a aussi été utilisée dans de nombreuse applications telle que la texture de synthèse \citep{tartavel2016wasserstein, heitz2021sliced}, l'adaptation de domaine \citep{lee2019sliced, rakotomamonjy2021statistical, xu2022unsupervised}, la reconstruction de nuage de points \citep{nguyen2023self}, des tests \citep{wang2021two, wang2021two_kernel, xu2022central} ou pour évaluer la performance de GANs \citep{karras2018progressive}. En outre, c'est une distance hilbertienne qui peut ainsi être utilisée pour définir des noyaux entre distributions de probabilités, qui peuvent être utilisés dans des méthodes à noyaux \citep{hofmann2008kernel}, par exemple pour le kernel K-Means, PCA, SVM \citep{kolouri2016sliced} ou pour faire de la régression \citep{meunier2022distribution}.

Comme SW est très populaire, beaucoup de variantes ont été proposées afin de pouvoir être utilisées avec des structures de données spécifiques \citep{nguyen2022revisiting} ou pour améliorer son pouvoir discriminatif en échantillonnant plus attentivement les directions des projections \citep{deshpande2019max, rowland2019orthogonal, nguyen2020distributional, nguyen2020improving, dai2021sliced, nguyen2023markovian, nguyen2023energy, ohana2022shedding}, en changeant la façon de projeter \citep{kolouri2019generalized, chen2020augmented, nguyen2022hierarchical} ou les sous-espaces sur lesquels projeter \citep{paty2019subspace, lin2021projection, li2022hilbert}. D'autres travaux proposent des estimateurs de SW, soit pour réduire la variance \citep{nguyen2023control}, soit pour réduire la malédiction de la dimension par rapport aux projections \citep{nadjahi2021fast}.

Le processus de slicing a aussi reçu beaucoup d'attentions pour d'autres divergences. \citet{nadjahi2020statistical} a étudié les propriétés de différentes divergences entre probabilités slicées, comprenant par exemple la distance de Sliced-Wasserstein, mais aussi la divergence de Sinkhorn slicée ou la Sliced-MMD. Cela a été aussi utilisé par exemple pour fournir une variante slicée de la distance ``Tree-Wasserstein'' \citep{le2019tree}, pour généraliser des divergences qui ne sont originellement bien définies qu'entre distributions de probabilité unidimensionnelles à des distributions de plus grande dimension comme la distance de Cramér \citep{Kolouri2020Sliced} ou pour alléger la malédiction de la dimension de la ``Kernelized Stein discrepancy'' \citep{gong2021sliced}, de l'information mutuelle \citep{goldfeld2021sliced, goldfeld2022k} ou de la variation totale et de la distance de Kolmogorov-Smirnov pour comparer des chaînes MCMC \citep{grenioux2023sampling}. Cela a aussi été utilisé pour la tâche de score matching \citep{song2020sliced} qui a récemment été mise en avant à travers les modèles de diffusion. Il a été aussi appliqué à différents problèmes de transport optimal comme le problème multi-marginal \citep{cohen2021sliced} ou le problème de transport optimal partiel \citep{figalli2010optimal} dans \citep{bonneel2019spot, bai2022sliced}, qui peut être utilisé entre des mesures avec différentes masses, et qui est un cas particulier du problème de transport optimal non balancé \citep{benamou2003numerical, sejourne2022unbalanced}.

\looseness=-1 Ces précédents travaux se sont focalisés principalement sur des espaces euclidiens. Cependant, beaucoup de données ont une structure connue qui n'est pas euclidienne. En effet, par la ``manifold hypothesis'', il est communément accepté que les données reposent sur une variété de plus faible dimension \citep{dong2012nonlinear, bengio2013representation, fefferman2016testing, pope2021the}. Dans certains cas, il est possible de connaître exactement la structure Riemannienne des données. Par exemple, les données terrestres sont sur une sphère, ou des données hiérarchiques peuvent être représentées efficacement sur des espaces hyperboliques \citep{nickel2017poincare}. Le problème de transport optimal est bien défini sur ces espaces \citep{villani2009optimal}. Ainsi, en ML, le transport optimal a récemment été utilisé pour des données reposant dans des variétés riemanniennes \citep{alvarez2020unsupervised, hoyos2020aligning}. Mais l'accent a été mis sur l'utilisation de la distance de Wasserstein ou du transport optimal avec régularisation entropique, au lieu de méthodes reposant sur des projections sur des sous-espaces. Afin de combler cette lacune, l'un des principaux objectifs de la thèse sera de développer des distances de Sliced-Wasserstein sur des variétés riemanniennes.
 
\looseness=-1 Une des limitations de SW est le manque de plan de transport optimal, qui peut être très utile pour des applications telles que l'adaptation de domaine \citep{courty2016optimal}, d'alignements d'embeddings de mots avec le problème de Wasserstein Procrutes \citep{grave2019unsupervised, ramirez2020novel} ou l'alignement de cellules \citep{demetci2022scot}. Pour pallier à cela, on pourrait utiliser la projection barycentrique, mais qui ne donnerait pas forcément un bon plan de transport car beaucoup de projections ne sont pas vraiment significatives. Trouver un plan de transport optimal requiert donc de résoudre le problème de transport optimal, qui peut être insoluble en pratique pour des problèmes à grande échelle. \citet{muzellec2019subspace} ont proposé de projeter les distributions sur un sous-espace, puis de se reposer sur la désintégration de mesures pour retrouver le plan de transport optimal. Plus récemment, \citet{li2022hilbert} ont plutôt utilisé des plans possiblement sous-optimaux obtenus entre des projections sur des courbes de Hilbert.

\paragraph{Transport Optimal entre des Données Incomparables.}

Quand on a des données incomparables, par exemple des données qui ne peuvent pas être représentées dans le même espace ou qui ne peuvent pas être bien comparées entre elles avec des distances à cause d'invariances entre données qui ne sont pas prises en compte par la distance, le problème de transport optimal classique n'est plus applicable, ou en tout cas pas très performant. Alors qu'il a été proposé d'apprendre de manière simultanée des transformations latentes pour calculer la distance de transport optimal \citep{alvarez2019towards} ou de représenter les deux distributions dans un espace euclidien commun \citep{alaya2021heterogeneous, alaya2022theoretical}, une méthode populaire qui prend directement en compte les invariances tout en permettant de comparer des distributions sur des espaces différents est la distance de Gromov-Wasserstein \citep{memoli2011gromov}. Cette distance a récemment reçu beaucoup d'intérêts en ML, par exemple pour comparer des données génomiques \citep{demetci2022scot} ou des graphes \citep{xu2019scalable, chowdhury2021generalized}. Cependant, cette distance souffre d'un coût computationnel encore plus grand que le problème de transport original \citep{peyre2016gromov}, et ne peut ainsi qu'être difficilement calculable dans un contexte de grande échelle. Alors que ce problème n'a pas toujours une forme close en une dimension \citep{dumont2022existence, beinert2022assignment}, une forme close est disponible dans certains cas particuliers \citep{vayer2020contribution} et une version sliced a été précédemment proposée \citep{vayer2019sliced}.

\paragraph{Objectifs.}

Ici, nous résumons les objectifs de la thèse avant de décrire plus en détail dans la prochaine section les contributions.

\begin{itemize}
    \item[$\bullet$] Tout d'abord, comme beaucoup de données ont une structure riemannienne, nous aurons pour objectif de définir de nouvelles distances de Sliced-Wasserstein sur des variétés riemanniennes.
    \item[$\bullet$] Comme SW fournit une distance efficace entre des distributions de probabilités qui partage beaucoup de propriétés avec la distance de Wasserstein, une question naturelle est d'étudier les propriétés des flots gradients sous-jacents comparés aux flots gradients Wasserstein.
    \item[$\bullet$] Motivé par les propriétés de robustesse du transport optimal non balancé, et des récentes méthodes de Sliced Partial OT, nous explorerons comment étendre le processus de slicing au transport optimal non balancé dans le but de comparer des mesures positives.
    \item[$\bullet$] Un autre objectif de cette thèse sera de fournir de nouveaux outils pour projeter sur des sous-espaces de l'espace des mesures de probabilités, dans l'objectif de l'appliquer à des jeux de données composés de distributions de probabilités.
    \item[$\bullet$] Comme une limitation de SW est de ne pas fournir de plan de transport, nous explorerons comment calculer efficacement des plans de transport entre des espaces incomparables en utilisant le problème de Gromov-Wasserstein.    
\end{itemize}

\section{Aperçu de la Thèse et Contributions}

Cette thèse se concentre sur les distances de transport optimal basées sur des projections sur des sous-espaces. Le chapitre \ref{chapter:bg_ot} fournit le contexte général sur le Transport Optimal requis pour comprendre le reste de la thèse, ainsi qu'un aperçu de la littérature.

Ensuite, la partie \ref{part:sw_riemannian} introduit des distances de Sliced-Wasserstein sur des variétés riemanniennes et les applique à différents problèmes de Machine Learning ainsi qu'à différentes variétés. La partie \ref{part:ot} couvre soit des applications du Transport Optimal basées sur la distance de Wasserstein, ou des variantes de Transport Optimal basées sur des projections sur des sous-espaces. Nous détaillons maintenant plus en profondeur le contenu et les contributions de chaque chapitre. Nous mentionnons aussi les collaborateurs en dehors du laboratoire d'accueil de l'auteur de la thèse.

\subsection{Partie \ref{part:sw_riemannian}: Sliced-Wasserstein sur Variétés Riemanniennes}

Dans la partie \ref{part:sw_riemannian}, nous étudions l'extension de la distance de Sliced-Wasserstein, originellement définie sur des espaces euclidiens, à des variétés riemanniennes. Plus précisément, nous introduisons d'abord dans le chapitre \ref{chapter:sw_hadamard} une façon de construire des distances de Sliced-Wasserstein sur des variétés de (Cartan)-Hadamard et nous introduisons certaines de leurs propriétés. Ensuite, nous prenons avantage de ces constructions dans les chapitres \ref{chapter:hsw} et \ref{chapter:spdsw} afin de construire des distances de Sliced-Wasserstein sur des variétés de Hadamard spécifiques : les espaces hyperboliques et les espaces de matrice symétrique définies positives. Finalement, dans le chapitre \ref{chapter:ssw}, nous étudions le cas de la sphère, qui ne rentre pas dans le cadre précédent car ce n'est pas une variété de Hadamard.

\subsubsection{Chapitre \ref{chapter:sw_hadamard}: Sliced-Wasserstein sur variétés de Cartan-Hadamard} 

Dans ce chapitre, en considérant $\mathbb{R}^d$ comme un cas particulier d'une variété riemannienne, nous dérivons les outils pour étendre la distance de Sliced-Wasserstein sur des variétés Riemanniennes géodésiquement complètes. Plus précisément, nous identifions les lignes comme des géodésiques, et proposons de projeter les mesures sur les géodésiques de variétés.

Nous nous concentrons ici sur des variétés Riemanniennes géodésiquement complètes de courbure négative, qui ont pour particularité d'avoir leurs géodésiques isométriques à $\mathbb{R}$. Cela permet de projeter les mesures sur la ligne réelle où la distance de Wasserstein peut être facilement calculée. De plus, nous proposons deux façons de projeter sur la ligne réelle. Ces deux manières de projeter sont des extensions naturelles de la projection dans le cas Euclidien. Tout d'abord, nous considérons la projection géodésique, qui consiste à projeter le long des chemins les plus courts, et qui permet de définir la distance Geodesic Cartan-Hadamard Sliced-Wasserstein ($\gchsw$). La seconde projection est la projection horosphérique, qui projette le long des horosphères en utilisant les lignes de niveau de la fonction de Busemann, et qui permet de définir la distance Horospherical Cartan-Hadamard Sliced-Wasserstein ($\hchsw$).

Ensuite, nous analysons théoriquement ces deux constructions et montrons que plusieurs propriétés importantes de la distance de Sliced-Wasserstein euclidienne sont encore valables sur des variétés de Hadamard. Plus précisément, nous discutons de leurs propriétés de distance, dérivons leurs premières variations et montrons qu'elles peuvent être représentées dans des espaces de Hilbert. Puis, nous dérivons leur complexité de projection ainsi que leur complexité par rapport aux échantillons, qui de manière similaire au cas euclidien, est indépendant de la dimension.

\subsubsection{Chapitre \ref{chapter:hsw}: Hyperbolic Sliced-Wasserstein}

Dans ce chapitre, nous prenons avantage de la construction dérivée dans le chapitre \ref{chapter:sw_hadamard} et l'appliquons aux espaces hyperboliques, qui sont des cas particuliers de variété de Hadamard, caractérisés par une courbure (constante) strictement négative.

Puisqu'il y a plusieurs paramétrisations équivalentes des espaces hyperboliques, nous étudions le cas du modèle de Lorentz et de la boule de Poincaré, et dérivons des formes closes pour définir et calculer efficacement la distance Geodesic Hyperbolic Sliced-Wasserstein ($\ghsw$) et Horospherical Hyperbolic Sliced-Wasserstein ($\hhsw$). Nous montrons aussi que ces deux formulations peuvent être utilisées indifféremment sur la boule de Poincaré et le modèle de Lorentz.

Nous comparons ensuite les comportements de $\ghsw$, $\hhsw$ et les distances de Sliced-Wasserstein euclidiennes sur la boule de Poincaré et le modèle de Lorentz sur différentes tâches comme la descente de gradient ou des problèmes de classification avec des réseaux de neurones.

Ce chapitre est basé sur \citep{bonet2022hyperbolic} et a été présenté au workshop ``Topology, Algebra and Geometry in Machine Learning'' (TAG-ML) à la conférence ICML 2023. Le code est disponible à \url{https://github.com/clbonet/Hyperbolic_Sliced-Wasserstein_via_Geodesic_and_Horospherical_Projections}.

\subsubsection{Chapitre \ref{chapter:spdsw}: Sliced-Wasserstein sur les Matrices Symétriques Définies Positives}

Dans ce chapitre, nous introduisons des distances de Sliced-Wasserstein sur l'espace des matrices symétriques définies positives (SPD). Muni de métriques spécifiques, l'espace des SPDs est de courbure négative et donc une variété de Hadamard. Ainsi, nous pouvons aussi utiliser la théorie introduite dans le chapitre \ref{chapter:sw_hadamard} afin de définir des distances de Sliced-Wasserstein.

Nous étudions l'espace des SPDs muni de deux métriques spécifiques : la métrique Affine-Invariante et la métrique Log-Euclidienne. Avec la métrique Affine-Invariante, l'espace des SPDs est de courbure variable négative. Comme dériver une forme close pour la projection géodésique est difficile, nous nous concentrons sur la projection de Busemann et introduisons la distance de Sliced-Wasserstein horosphérique $\aispdsw$. Cependant, $\aispdsw$ est coûteux computationnellement. Ainsi, cela nous motive à utiliser la métrique Log-Euclidienne, qui peut être vue comme une approximation du premier ordre de la métrique Affine-Invariante \citep{arsigny2005fast, pennec2020manifold} et qui est plus facile à calculer en pratique. Muni de cette métrique, l'espace des SPDs est de courbure nulle et nous pouvons dériver la distance de Sliced-Wasserstein correspondante $\lespdsw$.

Nous dérivons quelques propriétés complémentaires pour $\lespdsw$. Puis, nous appliquons cette distances à des problèmes de Magnetoencéphalographie et de Electroencéphalographie (M/EEG) comme la prédiction de l'âge du cerveau ou l'adaptation de domaine appliqué à des problèmes d'interfaces neuronales directes.

Ce chapitre est basé sur \citep{bonet2023sliced} et a été accepté à ICML 2023. Le code est disponible à \url{https://github.com/clbonet/SPDSW}. Ce travail a été fait en collaboration avec Benoît Malézieux (Inria).

\subsubsection{Chapitre \ref{chapter:ssw}: Spherical Sliced-Wasserstein}

Nous étudions dans ce chapitre une manière de définir une distance de Sliced-Wasserstein sur la sphère. Contrairement aux chapitres précédents, la sphère est de courbure strictement positive, et n'est donc pas une variété de Hadamard. Ainsi, nous ne pouvons pas utiliser les constructions introduites dans le chapitre \ref{chapter:sw_hadamard}.

Prenant en compte les particularités de la sphère, nous introduisons une distance de Sliced-Wasserstein sphérique ($\ssw$) en projetant les mesures sur des grands cercles, qui sont les géodésiques de la sphère. Pour l'implémentation pratique, nous dérivons une forme close de la projection géodésique, et utilisons l'algorithme de \citet{delon2010fast} pour calculer la distance de Wasserstein sur le cercle. De plus, nous introduisons une forme close pour calculer la distance de Wasserstein sur le cercle entre une mesure de probabilité arbitraire et la distribution uniform sur $S^1$. Concernant la partie théorique, nous étudions quelques connections avec une transformée de Radon sphérique permettant d'investiguer les propriétés de distance.

Ensuite, nous illustrons l'utilité de cette distance sur des tâches de Machine Learning comme l'échan\-tillonnage, l'estimation de densité ou pour apprendre des modèles génératifs.

Ce chapitre est basé sur \citep{bonet2023spherical} et a été accepté à ICLR 2023. Le code est disponible à \url{https://github.com/clbonet/Spherical_Sliced-Wasserstein}. De plus, l'implémentation de $\ssw$ a été ajoutée à la librairie Python Optimal Transport (POT) \citep{flamary2021pot}.

\subsection{Partie \ref{part:ot}: Transport Optimal et Variantes via des Projections}

Dans la partie \ref{part:ot}, nous étudions différents problèmes qui impliquent des projections sur des sous-espaces et du transport optimal. Dans le chapitre \ref{chapter:swgf}, nous investiguons les flots gradients dans l'espace des mesures de probabilités muni de la distance de Sliced-Wasserstein comparé avec l'espace des mesures de probabilité muni de la distance de Wasserstein. Ensuite, dans le chapitre \ref{chapter:usw}, nous développons un framework pour comparer des mesures positives avec des méthodes de sliced. Dans le chapitre \ref{chapter:busemann}, nous investiguons la fonction de Busemann dans l'espace des mesures de probabilité muni de la distance de Wasserstein. Finalement, nous développons dans le chapitre \ref{chapter:gw} une approche basée sur des détours par sous-espace pour le problème de Gromov-Wasserstein.

\subsubsection{Chaptire \ref{chapter:swgf}: Flots Gradients dans l'espace de Sliced-Wasserstein}

Une façon de minimiser des fonctionnelles sur l'espace de mesures de probabilités est d'utiliser les flots gradients Wasserstein, qui peuvent être approximés par le schéma d'Euler implicite, aussi appelé le schéma de Jordan-Kinderlehrer-Otto (JKO). Cependant, cela peut être coûteux computationnellement en pratique. Ainsi, dans ce chapitre, nous proposons de remplacer la distance de Wasserstein dans le schéma JKO par la distance de Sliced-Wasserstein afin de réduire le coût computationnel. Cela revient à calculer des flots gradients dans l'espace des mesures de probabilités muni de la distance de Sliced-Wasserstein. En modélisant les distributions de probablité avec des réseaux de neurones, nous proposons d'approximer la trajectoire des flots gradients Sliced-Wasserstein de plusieurs fonctionnelles, et de comparer leurs trajectoires avec celles des flots gradients Wasserstein.

Nous étudions différents types de fonctionnelles. Tout d'abord, nous étudions la divergence de Kullback-Leibler qui nécessite d'utiliser des réseaux de neurones inversibles, appelés flots normalisants, afin de l'approximer en pratique. Avec une cible gaussienne, nous connaissons exactement son flot gradient Wasserstein, et pouvons donc comparer sa trajectoire avec le flot gradient Sliced-Wasserstein approximé. Ensuite, nous étudions la capacité de notre méthode d'approximer des lois cibles avec des données réelles dans un contexte de régression logistique bayésienne. De plus, nous étudions la minimisation de la distance de Sliced-Wasserstein afin d'apprendre des mesure cibles en grande dimension, comme des distributions d'images.

Ce chapitre est basé sur \citep{bonet2022efficient} et a été publié dans le journal Transactions on Machine Learning Research (TMLR). Le code est disponible en ligne à \url{https://github.com/clbonet/Sliced-Wasserstein_Gradient_Flows}.

\subsubsection{Chapitre \ref{chapter:usw}: Le Transport non-balancé rencontre Sliced-Wasserstein}

Dans certains cas, il peut être bénéfique de comparer des mesures positives au lieu de distributions de probabilités. Cela a mené au développement du problème de transport optimal non balancé qui relâche les contraintes du coût de transport afin de comparer des mesures positives.

Nous étudions dans ce chapitre comment slicer efficacement ces méthodes de deux façons. Tout d'abord, nous proposons naïvement de moyenner le problème de transport non balancé entre les mesures projetées (SUOT), de la même manière que \citep{bonneel2019spot, bai2022sliced} pour le transport optimal partiel. Comme l'une des caractéristiques principales du transport non-balancé est d'enlever les exemples aberrants des marginales originales, nous introduisons aussi la distance de Sliced-Wasserstein non balancée (USW) qui régularise les marginales originales. L'implémentation pratique est basée sur l'algorithme de Frank-Wolfe en s'appuyant sur \citep{sejourne2022faster}.

Ce chapitre est basé sur un papier soumis \citep{sejourne2023unbalanced} et est un effort collaboratif avec Thibault Séjourné (EPFL), Kimia Nadjahi (MIT), Kilian Fatras (Mila) et Nicolas Courty. La contribution principale de l'auteur de la thèse est sur la partie expérimentale, où nous montrons sur une tâche de classification de document les bénéfices d'utiliser USW à la place de SUOT. L'algorithme est aussi assez flexible pour être utilisé avec n'importe quelle variante de Sliced-Wasserstein, et nous illustrons cela en calculant la distance Unbalanced Hyperbolic Sliced-Wasserstein en s'appuyant sur le chapitre \ref{chapter:hsw}.

\subsubsection{Chapitre \ref{chapter:busemann}: Busemann Function dans l'espace Wasserstein}

La fonction de Busemann, associée à des géodésiques bien choisies, fournit une généralisation naturelle du produit scalaire sur des variétés. Ainsi, ses lignes de niveaux peuvent être vues comme des contreparties naturelles aux hyperplans. Cela a récemment été beaucoup utilisé sur des variétés de Hadamard comme les espaces hyperboliques afin de faire de l'analyse en composantes principales (ACP) ou pour des tâches de classification \citep{chami2021horopca, ghadimi2021hyperbolic}.

Afin de pouvoir analyser des jeux de données composés de mesures de probabilités, ce chapitre étudie la fonction de Busemann sur l'espace des mesures de probabilités munie de la distance de Wasserstein (espace de Wasserstein). Dans l'espace de Wasserstein, cette fonction n'est pas définie pour toutes les géodésiques. Ainsi, nous identifions d'abord pour quelles géodésiques cette fonction est bien définie. Ensuite, nous dérivons des formes closes dans les cas particuliers des mesures de probabilités sur la ligne réelle et des gaussiennes. Nous illustrons ensuite l'utilisation de cette fonction pour effectuer une analyse en composante principale de jeux de données de distributions unidimensionnelles.

Ce travail est effectué en collaboration avec Elsa Cazelles (IRIT).

\subsubsection{Chapitre \ref{chapter:gw}: Les détours par sous-espaces rencontrent Gromov-Wasserstein}

Dans ce chapitre, nous sommes intéressés à la réduction du coût computationnel du problème de Gromov-Wasserstein en étant encore capable de calculer un plan de transport optimal entre les mesures originales. Ainsi, nous proposons d'étendre l'approche de détours par sous-espaces, originellement introduite par \citet{muzellec2019subspace} pour le problème de transport optimal, au problème de Gromov-Wasserstein. Comme le problème de Gromov-Wasserstein requiert seulement de calculer les distances dans chaque espace, nous proposons de projeter sur un sous-espace différent la source et la cible, ce qui peut permettre de mieux conserver le vrai plan de transport optimal. Nous dérivons quelques propriétés théoriques du problème, et notamment une forme close pour les couplings basés sur des détours par sous-espaces quand les deux mesures sont gaussiennes et le problème est restreint à des couplings gaussiens. Ensuite, nous illustrons cette approche à un problème de matching de forme.

Dans une seconde partie, nous introduisons un nouveau coût de transport optimal, qui partage la propriété du problème de transport optimal original d'être connecté formellement au coupling de Knothe-Rosenblatt via un coût dégénéré.

Ce chapitre est basé sur \citep{bonet2021subspace} et a été présenté au workshop de Neurips OTML2021 et publié dans le journal Algorithms. Il a été fait en collaboration avec Titouan Vayer (Inria).

\clearemptydoublepage
\cleartooddpage[\thispagestyle{empty}]
\phantomsection 
\addcontentsline{toc}{chapter}{Bibliography}

\bibliography{biblio/biblio}

\clearemptydoublepage
\cleartoevenpage[\thispagestyle{empty}]
\markboth{}{}
\newgeometry{inner=30mm,outer=20mm,top=40mm,bottom=20mm}

\backcoverheader

\selectfontbackcover{ 

\titleFR{Tirer parti du transport optimal via des projections sur des sous-espaces pour des applications d'apprentissage automatique}

\keywordsFR{Transport Optimal, Sliced-Wasserstein, Variétés Riemanniennes, Flots Gradients}

\abstractFR{Le problème de transport optimal a reçu beaucoup d'attention en Machine Learning car il permet de comparer des distributions de probabilités en exploitant la géométrie de l'espace sous-jacent. Cependant, dans sa formulation originale, résoudre ce problème souffre d'un coût computationnel important. Ainsi, tout un champ de travail consiste à proposer des alternatives pour réduire ce coût tout en continuant de bénéficier de ses propriétés. Dans cette thèse, nous nous concentrons sur des alternatives qui utilisent des projections sur des sous-espaces. L'alternative principale est la distance de Sliced-Wasserstein, que nous proposons d'étendre à des variétés Riemanniennes afin de l'utiliser dans des applications de Machine Learning pour lesquelles ce genre d'espace a été prouvé bénéfique. Nous proposons aussi de nouvelles variantes de distance sliced entre des mesures positives dans le problème de transport non balancé. Pour revenir à la distance originale de Sliced-Wasserstein entre mesures de probabilités, nous étudions la dynamique de flots gradients quand cet espace est munis de cette distance à la place de la distance de Wasserstein. Ensuite, nous investiguons la fonction de Busemann, une généralisation du produit scalaire dans des espaces métriques, dans l'espace des mesures de probabilité. Finalement, nous étendons une approche basée sur des détours sur des sous-espaces à des espaces incomparables en utilisant la distance de Gromov-Wasserstein.}

\titleEN{Leveraging Optimal Transport via Projections on Subspaces for Machine Learning Applications}

\keywordsEN{Optimal Transport, Sliced-Wasserstein, Riemannian Manifolds, Gradient Flows}

\abstractEN{Optimal Transport has received much attention in Machine Learning as it allows to compare probability distributions by exploiting the geometry of the underlying space. However, in its original formulation, solving this problem suffers from a significant computational burden. Thus, a meaningful line of work consists at proposing alternatives to reduce this burden while still enjoying its properties. In this thesis, we focus on alternatives which use projections on subspaces. The main such alternative is the Sliced-Wasserstein distance, which we first propose to extend to Riemannian manifolds in order to use it in Machine Learning applications for which using such spaces has been shown to be beneficial in the recent years. We also study sliced distances between positive measures in the so-called unbalanced OT problem. Back to the original Euclidean Sliced-Wasserstein distance between probability measures, we study the dynamic of gradient flows when endowing the space with this distance in place of the usual Wasserstein distance. Then, we investigate the use of the Busemann function, a generalization of the inner product in metric spaces, in the space of probability measures. Finally, we extend the subspace detour approach to incomparable spaces using the Gromov-Wasserstein distance.}

}


\restoregeometry

\let\clearpage\relax 

\end{document}